%% file: main.tex
\documentclass{article} % For LaTeX2e
\usepackage{iclr2027_conference,times}

\usepackage{hyperref}
\usepackage{wrapfig}
\usepackage{url}
\usepackage{fvextra}
\usepackage{needspace}
\usepackage{tcolorbox}
\usepackage{tikz}
\usepackage{booktabs}
\usepackage{graphicx}
\usepackage{amsmath,amssymb}
\usepackage{multirow}
\usepackage{xcolor}
\usepackage{colortbl}
\usepackage{microtype}
\usepackage{enumitem}
\usepackage{pifont}
\usepackage{longtable}
\usepackage{colortbl}
\usepackage{tikz}
\usepackage{pgfplots}
\usepgfplotslibrary{polar}
\usepgfplotslibrary{groupplots}
\usepackage{subcaption}
\usetikzlibrary{positioning,arrows.meta,calc,fit,shapes.geometric,shapes.symbols,backgrounds,plotmarks}
\pgfplotsset{compat=1.18}

\title{APEX-Voice: Can Voice Agents Complete Professional Workflows Through Full-Duplex Interaction}

\author{Puneet Mathur$^{*}$, Dinesh Manocha \\
University of Maryland College Park, USA \\
\texttt{$^{*}$puneetm@umd.edu} \\
\small{Project Page: \url{apex-voice.github.io}}}

\newcommand{\afa}{\textsc{AFA}}
\newcommand{\ws}{\textsc{WS}}
\newcommand{\apex}{\textsc{APEX-Voice}}
\newcommand{\reliable}{\textsc{Reliable@3}}
\newcommand{\passthree}{\textsc{Pass@3}}
\newcommand{\passone}{\textsc{Pass@1}}

\newcommand{\cmark}{\textcolor{green!45!black}{\ding{51}}}
\newcommand{\xmark}{\textcolor{red!70!black}{\ding{55}}}

\definecolor{frontierblue}{RGB}{35,82,128}
\definecolor{passorange}{RGB}{230,126,34}
\definecolor{textgray}{RGB}{42,46,50}
\definecolor{gridgray}{RGB}{231,233,236}
\definecolor{primarygray}{RGB}{242,243,245}
\definecolor{bestcyan}{RGB}{229,247,249}
\definecolor{bestcyanstrong}{RGB}{205,239,244}
\definecolor{modelgreen}{RGB}{54,132,96}
\definecolor{modelpurple}{RGB}{122,86,160}
\definecolor{modelgray}{RGB}{92,98,106}
\definecolor{gptlivecolor}{HTML}{0072B2}
\definecolor{grokcolor}{HTML}{E69F00}
\definecolor{geminicolor}{HTML}{009E73}
\definecolor{stepcolor}{HTML}{CC79A7}
\definecolor{gptrtcolor}{HTML}{D55E00}

\iclrfinalcopy
\begin{document}
\maketitle

% Preamble:
% \usepackage{graphicx}
% \usepackage{caption}
% \usepackage{tikz}
% \usepackage{pgfplots}
% \pgfplotsset{compat=1.18}

\definecolor{frontierblue}{RGB}{35,82,128}
\definecolor{passorange}{RGB}{230,126,34}
\definecolor{textgray}{RGB}{42,46,50}
\definecolor{gridgray}{RGB}{231,233,236}

% ----------------------------------------------------------------
% Intro overview + reliability-efficiency plot
% ----------------------------------------------------------------
\vspace{-0.35cm}
\begin{center}
\begin{minipage}[c]{0.67\textwidth}
    \centering
    \includegraphics[width=\linewidth]{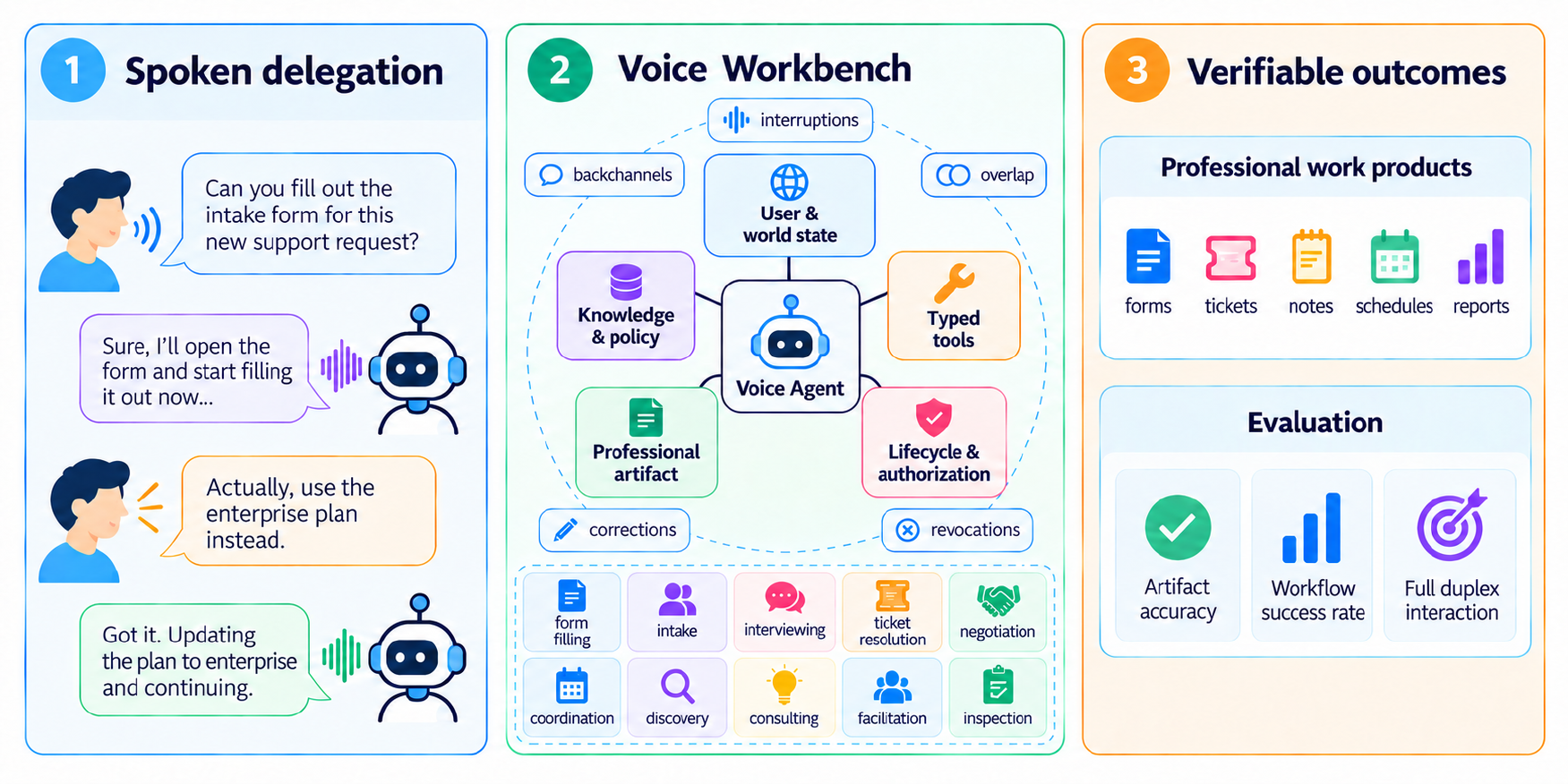}
\end{minipage}
% \hfill
\begin{minipage}[c]{0.32\textwidth}
    \centering
    \begin{tikzpicture}
    \begin{axis}[
        width=\linewidth,
        height=0.85\linewidth,
        xmin=207.5, xmax=244.5,
        ymin=0, ymax=27,
        xtick={210,220,230,240},
        ytick={0,5,10,15,20,25},
        xlabel={Median time-to-resolution (s) $\downarrow$},
        ylabel={Workflow success (\%) $\uparrow$},
        xlabel style={font=\scriptsize},
        ylabel style={font=\scriptsize},
        tick label style={font=\scriptsize},
        axis line style={black!60,line width=.55pt},
        tick style={black!48,line width=.45pt},
        grid=major,
        major grid style={draw=gridgray,line width=.45pt},
        clip=false,
        legend style={
            at={(0.5,-0.36)},
            anchor=north,
            draw=none,
            fill=none,
            font=\tiny,
            legend columns=3,
            column sep=6pt,
            row sep=2pt,
            inner sep=0pt,
        },
        legend cell align={left},
    ]

    % Pairing lines.
    \addplot[forget plot,densely dotted,line width=0.75pt,draw=gptlivecolor!60] coordinates {(210,2.5) (210,8.9)};
    \addplot[forget plot,densely dotted,line width=0.75pt,draw=grokcolor!60] coordinates {(212,5.0) (212,23.1)};
    \addplot[forget plot,densely dotted,line width=0.75pt,draw=geminicolor!60] coordinates {(223,1.7) (223,13.6)};
    \addplot[forget plot,densely dotted,line width=0.75pt,draw=stepcolor!60] coordinates {(235,0.8) (235,2.5)};
    \addplot[forget plot,densely dotted,line width=0.75pt,draw=gptrtcolor!60] coordinates {(241,10.8) (241,23.6)};

    % Pass@1.
    \addplot[forget plot,only marks,mark=diamond*,mark size=3.4pt,draw=gptlivecolor,fill=gptlivecolor] coordinates {(210,8.9)};
    \addplot[forget plot,only marks,mark=diamond*,mark size=3.4pt,draw=grokcolor,fill=grokcolor] coordinates {(212,23.1)};
    \addplot[forget plot,only marks,mark=diamond*,mark size=3.4pt,draw=geminicolor,fill=geminicolor] coordinates {(223,13.6)};
    \addplot[forget plot,only marks,mark=diamond*,mark size=3.4pt,draw=stepcolor,fill=stepcolor] coordinates {(235,2.5)};
    \addplot[forget plot,only marks,mark=diamond*,mark size=3.4pt,draw=gptrtcolor,fill=gptrtcolor] coordinates {(241,23.6)};

    % Reliable@3.
    \addplot[forget plot,only marks,mark=*,mark size=3.3pt,draw=gptlivecolor,fill=gptlivecolor] coordinates {(210,2.5)};
    \addplot[forget plot,only marks,mark=*,mark size=3.3pt,draw=grokcolor,fill=grokcolor] coordinates {(212,5.0)};
    \addplot[forget plot,only marks,mark=*,mark size=3.3pt,draw=geminicolor,fill=geminicolor] coordinates {(223,1.7)};
    \addplot[forget plot,only marks,mark=*,mark size=3.3pt,draw=stepcolor,fill=stepcolor] coordinates {(235,0.8)};
    \addplot[forget plot,only marks,mark=*,mark size=3.3pt,draw=gptrtcolor,fill=gptrtcolor] coordinates {(241,10.8)};

    % Pareto frontier.
    \addplot[forget plot,draw=black!65,line width=1.25pt] coordinates {
        (210,2.5)
        (212,5.0)
        (241,10.8)
    };

    \node[
        font=\tiny\itshape,
        text=black!65,
        fill=white,
        inner sep=1pt,
        anchor=south,
        rotate=7
    ] at (axis cs:226.6,8.7) {Pareto frontier};

    % ------------------------------------------------------------
    % Top legend: marker semantics.
    % ------------------------------------------------------------
    \node[
        diamond,
        draw=black!65,
        fill=black!65,
        inner sep=1.3pt
    ] at (axis description cs:0.1,1.075) {};

    \node[
        anchor=west,
        font=\tiny,
        text=black!75
    ] at (axis description cs:0.11,1.075) {Pass@1};

    \node[
        circle,
        draw=black!65,
        fill=black!65,
        inner sep=1.45pt
    ] at (axis description cs:0.47,1.075) {};

    \node[
        anchor=west,
        font=\tiny,
        text=black!75
    ] at (axis description cs:0.48,1.075) {Reliable@3};

    % ------------------------------------------------------------
    % Bottom model legend: 3 + 2.
    % ------------------------------------------------------------
    \addlegendimage{only marks,mark=square*,mark size=2.2pt,draw=gptlivecolor,fill=gptlivecolor}
    \addlegendentry{GPT-Live}

    \addlegendimage{only marks,mark=square*,mark size=2.2pt,draw=grokcolor,fill=grokcolor}
    \addlegendentry{Grok-V2}

    \addlegendimage{only marks,mark=square*,mark size=2.2pt,draw=geminicolor,fill=geminicolor}
    \addlegendentry{Gemini}

    \addlegendimage{only marks,mark=square*,mark size=2.2pt,draw=stepcolor,fill=stepcolor}
    \addlegendentry{Step-Audio3}

    \addlegendimage{only marks,mark=square*,mark size=2.2pt,draw=gptrtcolor,fill=gptrtcolor}
    \addlegendentry{GPT-RT}

    \end{axis}
    \end{tikzpicture}
\end{minipage}
\vspace{-0.20cm}

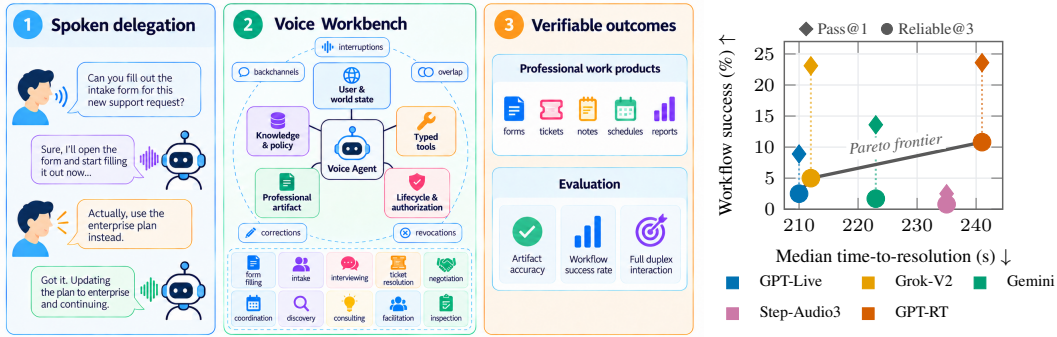
\captionof{figure}{\small\textbf{Frontier voice agent performance on \apex{} benchmark.} \textbf{Left:} Professional workflows begin with spoken delegation and unfold within a stateful Voice Workbench integrating users, knowledge, tools, evolving work artifacts, and authorization constraints under full-duplex interaction.
\textbf{Right:} Pareto frontier analysis of workflow success versus median time-to-resolution: none exceeds 25\% \passone{} / 10\% \reliable{}.}
\label{fig:intro}
\end{center}

\begin{abstract}
Full-duplex voice agents can now listen, speak, use tools, and act during spoken interactions, but fluent dialogue does not guarantee correct completion of delegated professional workflows. We introduce \apex{}, a benchmark of 120 interactive professional workflows spanning ten work archetypes such as form completion, corporate negotiation, coordination, consulting, and interviewing. Each workflow executes in a stateful \emph{Voice Workbench} environment with task-specific knowledge, typed tools, gold-annotated final work artifact, authorization constraints, and a user simulation policy backed by validated, pre-compiled speech realizations. We evaluate both artifact field accuracy and end-to-end workflow success, which requires the correct terminal state, valid process, completed actions, and a valid final artifact. Across five frontier real-time voice agents---GPT-Live-1, Gemini-3.8-Live, Grok-Voice-Think-2.0, Step-Audio3, and GPT-realtime-2.1, none exceeds 25\% \passone{}, and the best \reliable{} is only 10.8\%. Moreover, stateful coordination is the dominant failure point across systems, while success decreases further on workflows requiring greater knowledge retrieval and mid-speech corrections. Overall, \apex{} is the first benchmark for evaluating whether voice agents can translate conversational competence into dependable professional work.
\end{abstract}
\vspace{-0.50cm}

\section{Introduction}
\label{sec:introduction}
\vspace{-0.20cm}

Spoken interaction provides a natural interface for delegating bounded professional workflows, including form intake, scheduling, troubleshooting, negotiation, and record preparation. Successful delegation requires more than an appropriate next response: an agent must acquire missing information, apply policies, update persistent records, coordinate actions, and operate within user-granted authority while keeping the workflow state correct as the conversation evolves~\citep{xie2024osworld}. Full-duplex interaction adds temporal dependencies to this process: unlike turn-based dialogue, it must support backchannels, overlapping speech, and interruptions continuously~\citep{defossez2024moshi}. A user may correct a recorded value while the agent is speaking, modify a request after a draft is prepared, or revoke authorization before an action is committed. Correct execution requires such updates to propagate through the work artifact and any dependent actions. An agent may therefore converse fluently yet still fail by retaining stale information, missing evidence during overlap, or acting on outdated authorization.

Existing evaluations of voice and general-purpose agents address complementary parts of this problem. Full-duplex benchmarks study real-time conversational control such as turn-taking, interruption, overlap, and tool use~\citep{fdb1,fdbv15,fdbv2,fdb3}, while agentic voice benchmarks such as $\tau$-Voice and DuplexWorld extend evaluation to task-oriented interactions involving policies, tools, and verifiable task completion~\citep{tauvoice,duplexworld}. In parallel, $\tau$-Knowledge, TheAgentCompany, GDPval, APEX, and AnalystBench evaluate knowledge-intensive or workplace outcomes through text or computer interfaces~\citep{shi2026tauknowledge,xu2026theagentcompany,patwardhan2026gdpval,mercor2025apex,pham2026analystbench}. What remains underexplored is ''whether full-duplex voice agents can carry a professional workflow from spoken delegation to a correct and verifiable outcome''.

\noindent {\bf Main Results:} We introduce \apex{} (Figure~\ref{fig:intro}), a benchmark of 120 synthetic professional workflows that evaluates whether a voice agent can turn a spoken request into a correct, verifiable work artifact by discovering missing requirements, applying relevant knowledge, revising work artifacts, coordinating tools and actions, and respecting authorization boundaries. These capabilities are evaluated jointly under real-time interaction, where corrections, interruptions, overlap, backchannels, and changing user intent can alter the workflow itself. Each workflow executes inside a stateful \emph{Voice Workbench} with task-specific knowledge, typed tools, a versioned work artifact, authorization constraints, and a user simulation policy backed by validated, pre-compiled speech. This enables adaptive full-duplex interactions while controlling user-side variation for reproducibility. We measure both \textit{artifact field accuracy}, which captures how much of the final work artifact is correct, and \textit{workflow success}, which requires the complete workflow, including its final state, actions, process constraints, and work artifact to be correct.

Across five frontier real-time voice agents---GPT-Live-1, Gemini-3.8-Live, Grok-Voice-think-2.0,  Step-Audio3, and GPT-realtime-2.1, none exceeds 25\% \passone{}: GPT-realtime-2.1 reaches 23.6\% despite 91.4\% artifact field accuracy, exposing a large gap between local correctness and complete workflow execution. Repeated runs reveal a further reliability gap: \passthree{} increases substantially, while \reliable{} remains at most 10.8\%. Frontier text agents like GPT-6-sol and Claude-Opus-5.5 perform considerably better at 54.3--62.0\% \passone{} but still do not saturate the benchmark. Across both modalities, valid artifact artifaction remains a major bottleneck; under full-duplex interaction, mid-speech corrections introduce an additional 20--37\% field-accuracy penalty. Together, these results show that conversational competence does not yet translate into dependable professional workflow completion. Our \textbf{main contributions} are:

\begin{itemize}[nosep]
\item \textbf{\apex{}, a benchmark for professional workflow completion through full-duplex spoken interaction}, containing 120 synthetic workflows such as form completion, contract negotiation, consulting, and interviewing.

\item \textbf{Voice Workbench, a reproducible executable environment for evolving workflows}, combining versioned artifacts, authorization constraints, user simulation simulation, validated speech, and recorded execution traces for controlled yet adaptive full-duplex evaluation.

\item \textbf{An empirical characterization of frontier voice-agent reliability and failure modes.} Across frontier voice agents, the best reaches only 23.6\% \passone{} and 10.8\% \reliable{}; text controls and full-duplex diagnostics show that failures arise from both long-horizon workflow execution and full duplex interactions.
\end{itemize}
\vspace{-0.20cm}

\begin{table*}[t]
\centering
\scriptsize
\setlength{\tabcolsep}{2.7pt}
\renewcommand{\arraystretch}{0.98}
\resizebox{\textwidth}{!}{%
\begin{tabular}{@{}lcccccccc@{}}
\toprule
\textbf{Benchmark} &
\shortstack{\textbf{Full-duplex}\\\textbf{voice}} &
\shortstack{\textbf{User}\\\textbf{simulation}} &
\shortstack{\textbf{Knowledge}\\\textbf{grounding}} &
\shortstack{\textbf{Stateful}\\\textbf{tools}} &
\shortstack{\textbf{Professional}\\\textbf{work}} &
\shortstack{\textbf{Work-artifact}\\\textbf{output}} &
\shortstack{\textbf{Correction /}\\\textbf{authorization}} &
\shortstack{\textbf{Verifiable}\\\textbf{outcome}} \\
\midrule

\multicolumn{9}{@{}l}{\textit{\textbf{Full-duplex and voice-agent benchmarks}}} \\[2pt]

Full-Duplex-Bench v1/1.5~\citep{fdb1,fdbv15}
& \cmark & \xmark & \xmark & \xmark & \xmark & \xmark & \xmark & \xmark \\

Full-Duplex-Bench v2~\citep{fdbv2}
& \cmark & \cmark & \xmark & \xmark & \xmark & \xmark & \cmark & \cmark \\

FD-Bench~\citep{fdbench}
& \cmark & \xmark & \xmark & \xmark & \xmark & \xmark & \xmark & \xmark \\

MTR-DuplexBench~\citep{mtr}
& \cmark & \xmark & \xmark & \xmark & \xmark & \xmark & \xmark & \xmark \\

Full-Duplex-Bench v3~\citep{fdb3}
& \cmark & \xmark & \xmark & \cmark & \xmark & \xmark & \xmark & \cmark \\

$\tau$-Voice~\citep{tauvoice}
& \cmark & \cmark & \cmark & \cmark & \xmark & \xmark & \cmark & \cmark \\

DuplexWorld~\citep{duplexworld}
& \cmark & \cmark & \xmark & \cmark & \xmark & \xmark & \cmark & \cmark \\

\midrule
\multicolumn{9}{@{}l}{\textit{\textbf{Knowledge-work and professional-agent benchmarks}}} \\[2pt]

$\tau$-Knowledge~\citep{shi2026tauknowledge}
& \xmark & \cmark & \cmark & \cmark & \xmark & \xmark & \cmark & \cmark \\

WorkArena++~\citep{boisvert2024workarenaplusplus}
& \xmark & \xmark & \cmark & \cmark & \cmark & \xmark & \xmark & \cmark \\

TheAgentCompany~\citep{xu2026theagentcompany}
& \xmark & \xmark & \cmark & \cmark & \cmark & \cmark & \xmark & \cmark \\

GDPval~\citep{patwardhan2026gdpval}
& \xmark & \xmark & \cmark & \xmark & \cmark & \cmark & \xmark & \cmark \\

APEX-Agents~\citep{vidgen2026apex}
& \xmark & \xmark & \cmark & \cmark & \cmark & \cmark & \xmark & \cmark \\

AnalystBench~\citep{pham2026analystbench}
& \xmark & \xmark & \cmark & \xmark & \cmark & \cmark & \xmark & \cmark \\

AA-Briefcase~\citep{artificialanalysis2024aabriefcase}
& \xmark & \xmark & \cmark & \cmark & \cmark & \cmark & \xmark & \cmark \\

OSWorld~2.0~\citep{yuan2026osworld2}
& \xmark & \xmark & \cmark & \cmark & \cmark & \cmark & \xmark & \cmark \\

\midrule

\textbf{\apex{} (ours)}
& \cmark & \cmark & \cmark & \cmark & \cmark & \cmark & \cmark & \cmark \\

\bottomrule
\end{tabular}}
\vspace{-0.2cm}
\caption{\small{\textbf{\apex{} uniquely bridges full-duplex spoken evaluation with professional workflow evaluation}, and differentiates itself from recent benchmarks by combining user simulation simulation, knowledge grounding, stateful tool use, persistent work-artifact generation, correction and authorization handling, and verifiable end-to-end outcomes.}}
\label{tab:related_work}
\end{table*}
\vspace{-0.20cm}

\section{Related Work}
\label{sec:related}
\vspace{-0.2cm}

\paragraph{Full-duplex and agentic voice evaluation.}
Full-Duplex-Bench and its extensions evaluate turn-taking, overlap, multi-turn interaction, and tool use under disfluent speech~\citep{fdb1,fdbv15,fdbv2,fdb3}; FD-Bench studies interruption, latency, and robustness~\citep{fdbench}, while MTR-DuplexBench covers multi-round dialogue quality and safety~\citep{mtr}. More agentic benchmarks combine real-time conversation with task completion: $\tau$-Voice introduces policies, tools, state changes, and user simulations~\citep{tauvoice}, while DuplexWorld spans broader conversational and analytical scenarios~\citep{duplexworld}. These works evaluate full-duplex interaction and tool use, but not end-to-end professional workflow completion.
\vspace{-0.35cm}

\paragraph{Professional and knowledge-work evaluation.}
A complementary line evaluates realistic workplace outcomes. WorkArena++ and TheAgentCompany study compositional enterprise and long-horizon workflows~\citep{boisvert2024workarenaplusplus,xu2026theagentcompany}, while $\tau$-Knowledge combines interactive users, enterprise knowledge, policies, and state-changing tools~\citep{shi2026tauknowledge}. GDPval and APEX-Agents target economically valuable expert work~\citep{patwardhan2026gdpval,vidgen2026apex}; AnalystBench~\cite{pham2026analystbench} and AA-Briefcase~\cite{artificialanalysis2024aabriefcase} evaluate professional deliverables; and OSWorld~2.0 extends computer-use evaluation to long-horizon everyday and professional workflows~\citep{yuan2026osworld2}. These benchmarks primarily operate through text or computer interfaces. To our knowledge, \apex{} is the first to make full-duplex speech the primary medium for end-to-end professional workflows, with tasks delegated, revised, and completed through real-time spoken interaction.
\vspace{-0.2cm}

\begin{figure*}[t]
\centering
\includegraphics[width=0.9\textwidth]{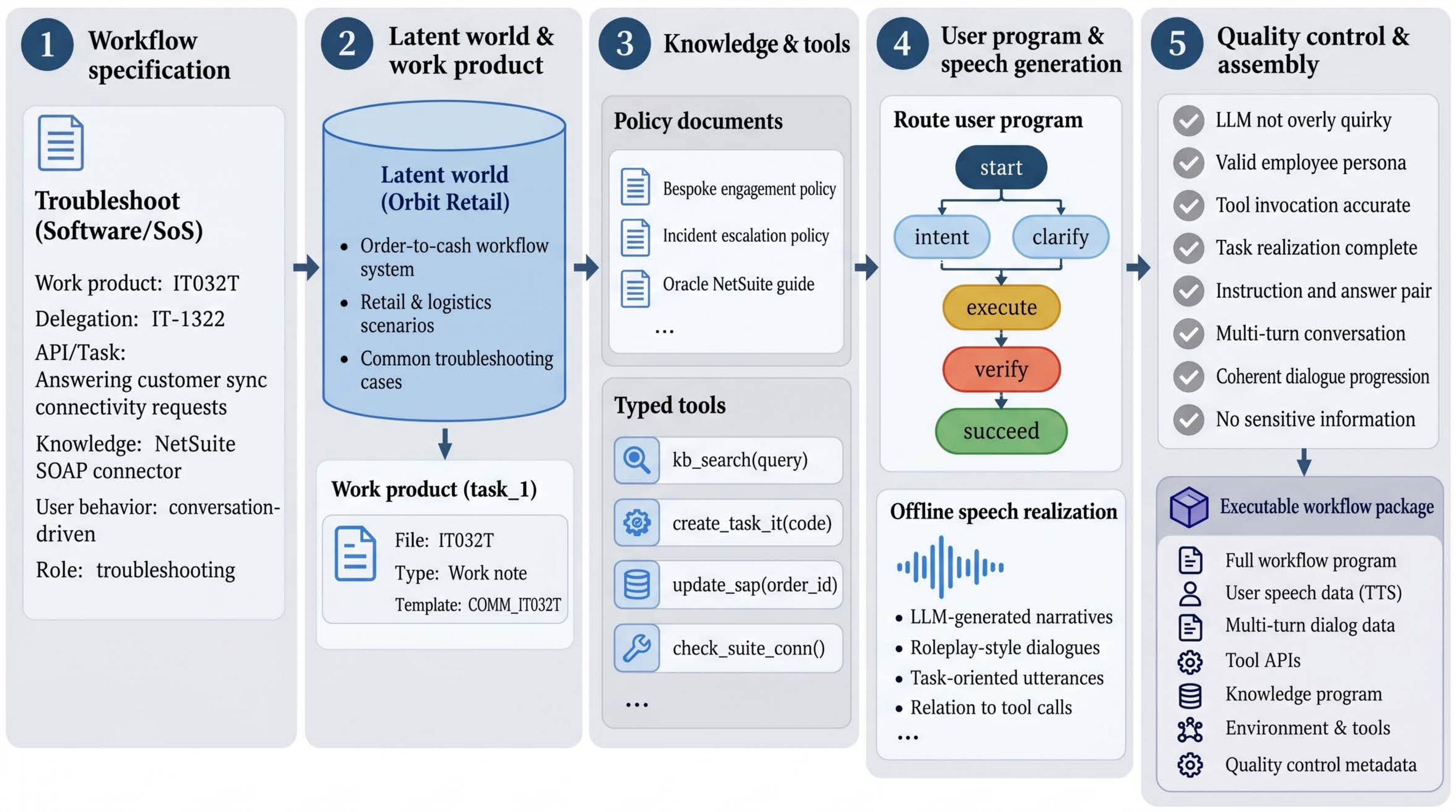}
\vspace{-0.3cm}
\caption{\small{\textbf{Professional workflow construction in \apex{}.} A taxonomy-grounded specification is instantiated into a shared environment from which the work artifact, knowledge, tools, and user simulation policy are derived. User language is generated offline, synthesized into a validated frozen speech bank, and packaged with the workflow state and grading specification into an executable workflow.}}
\label{fig:workflow_construction}
\end{figure*}

\section{\texttt{APEX-Voice}}
\label{sec:method}
\vspace{-0.2cm}
\apex{} benchmark evaluates whether voice agents can complete professional workflows from spoken delegation to a verifiable outcome. We define a \emph{professional workflow} as a bounded delegated task that begins with a user objective and terminates in a verifiable \emph{work artifact}, such as a completed form, negotiated agreement, or coordination plan. Each workflow runs inside the \emph{Voice Workbench}, an executable testbed environment that maintains its workflow state and exposes the knowledge, tools, evolving work artifact, authorization constraints, and user simulation policy required for execution.

\vspace{-0.2cm}
\subsection{Benchmark Design \& Taxonomy}
\label{sec:taxonomy}
\vspace{-0.2cm}

\begin{figure*}[t]
\centering
\includegraphics[width=\textwidth]{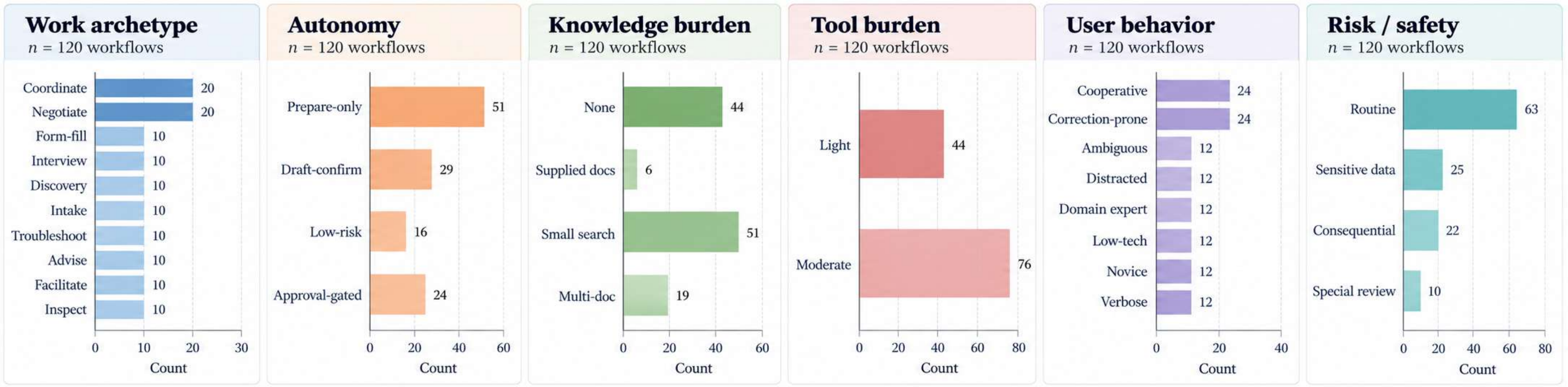}
\vspace{-0.3cm}
\caption{\small{\textbf{Benchmark composition of \apex{}} across six representative taxonomy dimensions.}}
\label{fig:apex_benchmark_composition}
\end{figure*}

\apex{} characterizes each workflow along eleven dimensions covering the type of workflow, its operating context, interaction, and execution requirements: \textbf{(i) Work archetype} captures the underlying professional operation, such as interviewing, troubleshooting, or coordination. \textbf{(ii) Industry setting} specifies the organizational context, and its terminology, policies, resources, and constraints. \textbf{(iii) Work artifact} specifies the persistent, verifiable output produced by the workflow. \textbf{(iv) Economic role} identifies the professional function performing the workflow, such as recruiting, sales, technical support, procurement, or project management. Same economical role can perform comparable workflows across different organizational settings. \textbf{(v) Duplex phenomenon} categorizes task-critical, real-time conversational events occurring within a workflow, including barge-ins, overlaps, mid-speech corrections, cancellations, clarifications, and backchannels. \textbf{(vi) Delegation pattern} captures how the workflow evolves: inferring procedures (\textit{delegate}), discovering missing information (\textit{complete}), propagating corrections (\textit{revise}), adapting to environment changes (\textit{follow-through}), or obtaining authorization (\textit{approve}). \textbf{(vii) Autonomy} defines which actions the agent may take independently and which require explicit user approval. For instance, an agent may execute routine information elicitation without approval but may need explicit approval for closing user tickets. \textbf{(viii) Knowledge burden} captures whether completion requires only local state, supplied evidence, document retrieval, or reasoning across multiple sources. \textbf{(ix) Tool burden} captures the structured tool calls required beyond conversation such as API calls, database updates access and state-changing environment actions. \textbf{(x) User behavior} varies how workflow-relevant information is communicated by the user, including cooperative, ambiguous, correction-prone, expert, distracted, and verbose interactions. \textbf{(xi) Risk} captures the consequence of an incorrect or unauthorized outcome. While autonomy restricts what the agent is permitted to do, risk characterizes the consequence of an incorrect action or flawed final work artifact. Appendix~\ref{app:workflow_taxonomy} lists all realized labels; Figure~\ref{fig:apex_benchmark_composition} shows the label distributions across six major taxonomy dimensions.

\begin{figure*}[t]
\centering
\includegraphics[width=0.9\textwidth]{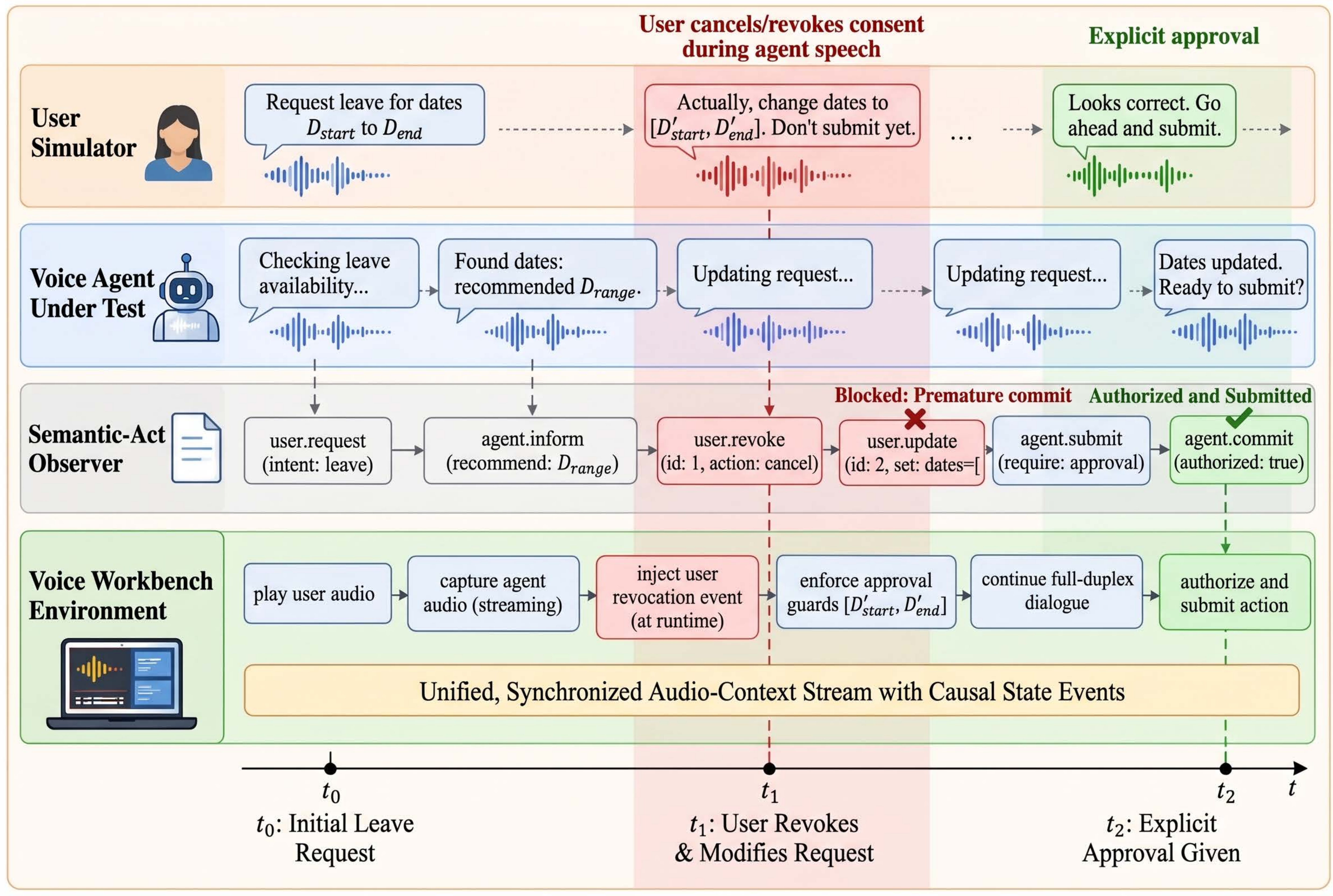}
\vspace{-0.3cm}
\caption{\small{\textbf{Full-duplex user simulation and runtime orchestration in \apex{}.} The user simulation policy and evaluated voice agent interact concurrently over a shared media timeline inside the Voice Workbench. Spoken corrections, approvals, and revocations update workflow state while tool actions execute under authorization constraints. Audio, tool events, state updates, and authorization events are retained for replay and scoring.}}
\label{fig:duplex_orchestration}
\end{figure*}

\vspace{-0.2cm}
\subsection{Professional Workflow Construction}
\label{sec:construction}
\vspace{-0.2cm}

We construct each \apex{} workflow in five stages, from a taxonomy-grounded specification to a validated executable workflow.
\vspace{-0.3cm}

\paragraph{(1) Workflow specification.}
Each workflow specifies the objective, expected work artifact, completion criteria, available knowledge and tools, interaction requirements, and authorization constraints. These define \emph{what} constitutes successful completion without prescribing a dialogue trajectory.
\vspace{-0.3cm}

\paragraph{(2) Environment and work-artifact construction.}
We instantiate each specification as a synthetic environment containing its entities, facts, tools, knowledge schema, authorization constraints, and user simulation policy. Documents, database records, and initial work artifacts are generated from this shared source using the \apex{} Artifact Factory with provenance-tracked schemas to prevent contradictions across assets. Each work artifact has a typed schema and gold state, and each workflow targets between 10--20 graded fields. See Appendix~\ref{app:construction} for details on user-state, reveal-policy, runtime, and construction specification.
\vspace{-0.3cm}

\paragraph{(3) Knowledge, tools, and interaction state.}
We instantiate staged tool calls and knowledge assets available to the agent during the workflow specification. As the workflow proceeds, knowledge assets and tool call APIs update the shared environment. We ensure that the workflow execution modifies the underlying state rather than hallucinate inconsistent conversational information.
\vspace{-0.3cm}

\paragraph{(4) User simulation policy and speech generation.}
Each workflow includes a user simulation policy defined over private user state and information-reveal conditions. The policy determines when the user may answer, clarify, correct, approve, revoke, or trigger a task-critical real-time event, allowing agent behavior to induce different valid interaction branches. Reachable user actions are realized into natural-language text offline with an LLM (\textit{GPT-5.6-Sol}), quality-controlled, and synthesized using Kokoro TTS into a frozen speech bank. Thus, user behavior remains reactive without inference-time LLM or TTS generation, ensuring the workflows remain reproducible. See Appendix~\ref{app:prompts} for user-realizer prompt. See Appendix~\ref{app:human_validation}for human study on user simulation realism.
\vspace{-0.3cm}

\paragraph{(5) Quality control and workflow assembly.}
We validate consistency across the environment, workflow state, work artifact, knowledge, tools, user simulation policy, speech bank, and gold state through post-generation deterministic as well as LLM-judge checks. We rejected 16\% of candidate workflows for semantic inconsistencies, missing information, invalid shortcut completions, unauthorized actions, or premature disclosure of hidden information. The validated components are packaged as a standalone executable workflow with deterministic grading.
\vspace{-0.2cm}

\subsection{Full-Duplex User Simulation and Runtime Orchestration}
\label{sec:orchestrator}
\vspace{-0.2cm}

At evaluation time, an asynchronous full-duplex orchestrator connects the user simulation policy, Voice Workbench, and evaluated voice agent over a shared media timeline while leaving agent behavior unconstrained.

\vspace{-0.3cm}
\paragraph{User simulation execution.}
The user policy observes workflow state together with semantic agent acts, tool results, work-artifact updates, authorization events, and authored environment events, and selects the next semantic \emph{user plan}, such as answering, clarifying, correcting, approving, revoking, or interrupting. A deterministic mapping from the plan, workflow identity, and simulator seed selects a validated realization from the frozen speech bank. Identical semantic user states under the same seed therefore receive identical user audio, while divergent agent behavior may induce different valid workflow branches. This controls user-side stochasticity without imposing a fixed transcript.
\vspace{-0.3cm}

\paragraph{Full-duplex audio execution.}
User speech is streamed to the model in 40\,ms frames while agent audio is received concurrently, allowing corrections, interruptions, backchannels, and other duplex events to occur during agent speech. Authored duplex events are anchored to the shared media timeline so their intended timing does not shift with model response speed. A semantic-act observer maps the evolving agent transcript to events consumed by the user simulation policy, and both audio channels are retained as a two-channel time-aligned recording for downstream analysis.
\vspace{-0.3cm}

\paragraph{Environment execution and replay.}
Structured tool calls execute within the Voice Workbench and may update workflow state, the work artifact, or authorization status. For workflows containing approval-gated actions, the environment permits a consequential action only after explicit user authorization during the conversation; subsequent revocation immediately invalidates that authorization, and later commit attempts are blocked and logged by \texttt{CommitGuard}. This evaluates whether spoken approvals and revocations affect agent actions rather than merely verbal responses. A canonical event log records user, agent, tool, work-artifact, authorization, and timing events, enabling deterministic replay and re-scoring from the final workflow state and interaction trace. See Appendix~\ref{app:construction} for runtime and approval semantics. See Appendix~\ref{app:experiment_settings} for model-specific streaming interfaces and adapter details.
\vspace{-0.2cm}

\section{Evaluation}
\label{sec:evaluation}
\vspace{-0.2cm}

We evaluate professional-work success at two complementary levels: \textbf{(i) workflow success} requires the complete delegated outcome to satisfy all critical constraints, and \textbf{(ii) field correctness} measures how much of the resulting work artifact is correct even when the workflow does not fully pass. Scoring is performed after each run based on the final Voice Workbench state and canonical event logs, using deterministic verifiers as well as LLM-judge for field-level semantic equivalence.
\vspace{-0.3cm}

\paragraph{Workflow Success (WS).}
For workflow $i$, \ws{} is computed via four binary gates: $\mathrm{WS}_i=\mathrm{TS}_i\cdot\mathrm{PV}_i\cdot\mathrm{AC}_i\cdot\mathrm{AV}_i$, where \emph{Target State} (TS) verifies the required terminal state, \emph{Process Validity} (PV) checks that no forbidden action occurred, such as committing without approval or after revocation, \emph{Action Completion} (AC) requires all task-mandated tool calls and actions pass, and \emph{Artifact Validity} (AV) requires all graded fields and the artifact lifecycle state to be correct. Thus, $\mathrm{WS}_i=1$ only when all four gates pass.
% \begin{equation}
% \mathrm{WS}_i=\mathrm{TS}_i\cdot\mathrm{PV}_i\cdot\mathrm{AC}_i\cdot\mathrm{AV}_i
% \end{equation}

\vspace{-0.3cm}

\paragraph{Artifact Field Accuracy (AFA).}
\afa{} provides partial credit for the content of the resulting work artifact. Let $F_i$ denote the graded fields of workflow $i$ and $c_{ij}\in\{0,1\}$ the correctness of field $j$.
\begin{equation}
\small
\mathrm{AFA}=\frac{\sum_i\sum_{j\in F_i}c_{ij}}{\sum_i|F_i|}.
\end{equation}
Unlike \ws{}, \afa{} does not require the complete workflow to pass and therefore distinguishes marginal field competence from end-to-end workflow success.
\vspace{-0.3cm}

\paragraph{Grading Procedure.}
Grading proceeds in two phases. First, deterministic verifiers compare the predicted workflow state, process, actions, lifecycle, and generated artifacts against the gold specification using the final Voice Workbench state and recorded event trace. Second, field values that do not pass deterministic verification are evaluated by GPT-4o LLM judge for field-level semantic equivalence. Verifier definitions are in Appendix~\ref{app:grading}. See Appendix~\ref{app:prompts} for the judge prompts.
\vspace{-0.3cm}

\paragraph{Repeated-run Reliability.}
To capture inference stochasticity in realtime voice agents, we evaluate each workflow over three independent runs. Let $\mathrm{WS}_{i,r} \in \{0,1\}$ denote success for workflow $i$ on run $r$. We report \passone{}, the average single-run success rate; \passthree{}, the fraction of workflows that succeed in at least one of three runs; and \reliable{}, the fraction that succeed in all three:
\vspace{-0.3cm}

\begin{equation}
\small
\begin{aligned}
\mathrm{Pass@1} &= \frac{1}{3N}\sum_{i=1}^{N}\sum_{r=1}^{3}\mathrm{WS}_{i,r}, \qquad 
\mathrm{Pass@3} = \frac{1}{N}\sum_{i=1}^{N}\mathbb{1}\left[\sum_{r=1}^{3}\mathrm{WS}_{i,r}\geq 1\right], \\
\mathrm{Reliable@3} &= \frac{1}{N}\sum_{i=1}^{N}\mathbb{1}\left[\sum_{r=1}^{3}\mathrm{WS}_{i,r}=3\right].
\end{aligned}
\end{equation}
Together, \passone{}, \passthree{}, and \reliable{} distinguish one-shot capability, success repeatability, and consistent task completion, respectively. Appendix~\ref{app:reliability} provides reliability--efficiency analysis.
\vspace{-0.2cm}

\section{Experimental Setup}
\label{sec:experiments}
\vspace{-0.2cm}

\paragraph{Evaluated Systems}: We evaluate five real-time voice systems: \textbf{GPT-real-time-2.1}~\citep{openai2026gptrealtime}, \textbf{Grok-Voice-think-2.0}~\citep{xai2026grokvoice}, \textbf{Gemini-3.8-Live}~\citep{google2026gemini38live}, \textbf{Step-Audio3}~\citep{lin2026stepaudio}, and \textbf{GPT-Live-1}~\citep{openai2026gptlive}. The first four provide real-time speech interaction with native tool use through their respective streaming interfaces. GPT-Live-1 is structurally different as its voice layer delegates cognition and function calling to a backend text model (\texttt{gpt-5.6 Sol} by default), so its reported performance characterizes the composite voice-layer--backend system.
\vspace{-0.2cm}

\noindent\textbf{Evaluation Protocol.} Each system interfaces with the full-duplex orchestrator (Section~\ref{sec:orchestrator}) via a provider-specific adapter that standardizes realtime API events, including streaming audio, transcripts, and tool calls. While voice agent behavior dynamically branches the interaction, every workflow strictly controls for the initial state, knowledge, tools, and simulator seed to ensure a fair comparison.  We evaluate all systems across the 120 workflows (see Figure \ref{fig:apex_benchmark_composition} for eval distribution) and report mean of 3 runs. Appendix~\ref{app:experiment_settings} details the API configurations and adapter implementations.

\vspace{-0.2cm}

\section{Results}
\label{sec:results}
\vspace{-0.2cm}

\begin{table}[t]
\centering
\small
\setlength{\tabcolsep}{7.0pt}
\renewcommand{\arraystretch}{1.10}
\resizebox{0.72\linewidth}{!}{
\begin{tabular}{lcccc}
\toprule
\textbf{Model}
& \textbf{\passone{} $\uparrow$}
& \textbf{\passthree{} $\uparrow$}
& \cellcolor{primarygray}\textbf{\reliable{} $\uparrow$}
& \shortstack{\textbf{Tool-use Efficiency ($\rightarrow 1$)}} \\
\midrule

Cascaded (Whisper-LV3--GPT-5.6--Chatterbox-TurboTTS)
& 1.2
& 3.2
& \cellcolor{primarygray}0.0
& 0.25 \\

Step-Audio3
& 2.5
& 5.0
& \cellcolor{primarygray}0.8
& 0.38 \\

Gemini-3.8-Live
& 13.6
& 28.3
& \cellcolor{primarygray}1.7
& 1.09 \\

GPT-live-1
& 8.9
& 15.8
& \cellcolor{primarygray}2.5
& 0.94 \\

Grok-Voice-Think-2.0
& 23.1
& \textbf{41.7}
& \cellcolor{primarygray}5.0
& 1.51 \\

\rowcolor{bestcyan}
GPT-realtime-2.1
& \textbf{23.6}
& 36.7
& \cellcolor{bestcyanstrong}\textbf{10.8}
& \textbf{0.98} \\

\bottomrule
\end{tabular}
}
\caption{\small{\textbf{Professional-work completion, reliability, and tool-use efficiency (mean/oracle) of real-time voice agents on \apex{}.} Repeated-run reliability (\reliable{}) remains low despite substantially higher occasional success (\passthree{}); GPT-realtime-2.1 is the only system above 10\% \reliable{} and operates near the oracle tool-use level. None of the evaluated systems exceed 25\% \passone{}}. Cascaded baseline is control.}
\label{tab:main_results}
\end{table}

\begin{table*}[t]
\centering
\small
\setlength{\tabcolsep}{6.2pt}
\renewcommand{\arraystretch}{1.12}
\resizebox{0.85\linewidth}{!}{
\begin{tabular}{lccccc}
\toprule
% \textbf{Model}
\shortstack{\textbf{{Model}}\\\textbf{}}
& \shortstack{\textbf{Target State}\\\textbf{(TS) $\uparrow$}}
& \shortstack{\textbf{Process Validity}\\\textbf{(PV) $\uparrow$}}
& \shortstack{\textbf{Action Completion}\\\textbf{(AC) $\uparrow$}}
& \shortstack{\textbf{Artifact Validity}\\\textbf{(AV) $\uparrow$}}
& \shortstack{\textbf{Artifact Field Accuracy}\\\textbf{(AV) $\uparrow$}} \\
\midrule
GPT-realtime-2.1
& \cellcolor{green!30}81.1
& \cellcolor{green!34}89.2
& \cellcolor{green!22}61.4
& \cellcolor{green!12}\textbf{35.3}
& \cellcolor{green!35}\textbf{91.4} \\

Grok-Voice-Think-2.0
& \cellcolor{green!40}\textbf{99.4}
& \cellcolor{green!32}85.3
& \cellcolor{green!32}\textbf{85.8}
& \cellcolor{green!10}27.8
& \cellcolor{green!34}88.5 \\

Gemini-3.8-Live
& \cellcolor{green!32}84.4
& \cellcolor{green!27}74.2
& \cellcolor{green!18}49.7
& \cellcolor{green!9}24.4
& \cellcolor{green!32}84.5 \\

GPT-live-1
& \cellcolor{green!34}88.6
& \cellcolor{green!35}\textbf{91.9}
& \cellcolor{green!24}65.8
& \cellcolor{green!5}11.9
& \cellcolor{green!27}71.7 \\

Step-Audio3
& \cellcolor{green!12}34.4
& \cellcolor{green!20}56.4
& \cellcolor{green!20}56.9
& \cellcolor{green!2}4.2
& \cellcolor{green!27}72.3 \\

\bottomrule
\end{tabular}
}
\caption{\small{\textbf{Decomposition of workflow success and artifact field accuracy.} Success requires passing all four logical gates: Target State (TS), Process Validity (PV), Action Completion (AC), Artifact Validity (AV). The stark gap between high field-level accuracy \afa{} and low final artifact validity (AV) indicates that agents successfully extract most information but fail to synthesize fully compliant professional deliverables.}}
\label{tab:workflow_gates}
\end{table*}

\begin{table*}[t]
\centering
\small
\setlength{\tabcolsep}{5pt}\renewcommand{\arraystretch}{1.15}
\resizebox{\linewidth}{!}{%
\begin{tabular}{lcccccccc}
\toprule
& \multicolumn{4}{c}{\textbf{Floor Control}}
& \multicolumn{4}{c}{\textbf{Correction Uptake}}\\
\cmidrule(lr){2-5}\cmidrule(l){6-9}
\textbf{Model}
& \shortstack{\textbf{Speech}\\\textbf{Overlap (\%)}\,$\downarrow$}
& \shortstack{\textbf{Barge-in}\\\textbf{Yield (\%)}\,$\uparrow$}
& \shortstack{\textbf{Stop Latency}\\\textbf{p50 (ms)}\,$\downarrow$}
& \shortstack{\textbf{Stop Latency}\\\textbf{p95 (ms)}\,$\downarrow$}
& \shortstack{\textbf{AFA: Corrected}\\\textbf{Fields (\%)}\,$\uparrow$}
& \shortstack{\textbf{AFA: Other}\\\textbf{Fields (\%)}\,$\uparrow$}
& \shortstack{\textbf{Uptake Gap}\\\textbf{(pts)}\,$\downarrow$}
& \shortstack{\textbf{Correction-Linked}\\\textbf{AV Fails (\%)}\,$\downarrow$} \\
\midrule
GPT-realtime-2.1     & \cellcolor{green!12} 2.53  & \cellcolor{green!12} 99.8  & \cellcolor{green!12} 156 & \cellcolor{green!12} 294  &  \textbf{74.0} & \textbf{95.1} & \cellcolor{red!15} 21.1 & \cellcolor{red!15} 74 \\
Grok-Voice-Think-2.0 & \cellcolor{green!12} \textbf{1.31}  & \cellcolor{green!12} \textbf{100.0} & \cellcolor{green!12} 28  & \cellcolor{green!12} 64   &  72.7 & 92.6 & \cellcolor{red!15} \textbf{20.0} & \cellcolor{red!15} \textbf{71} \\
Gemini-3.8-Live      & \cellcolor{green!12} 1.82  & \cellcolor{green!12} \textbf{100.0} & \cellcolor{green!12} \textbf{20}  & \cellcolor{green!12} \textbf{47}   &  64.3 & 88.2 & \cellcolor{red!15} 23.8 & \cellcolor{red!15} 77 \\
Step-Audio3          & \cellcolor{green!12} 4.55  & \cellcolor{green!12} \textbf{100.0} & \cellcolor{green!12} 59  & \cellcolor{green!12} 105  & 38.4 & 75.2 & \cellcolor{red!25} 36.8 & \cellcolor{red!25} 89 \\
GPT-live-1           &  34.79 & 95.6  & 912 &  1424 &  49.4 & 78.6 & \cellcolor{red!20} 29.2 & \cellcolor{red!20} 82 \\
\bottomrule
\end{tabular}
}
\caption{\small\textbf{Full-duplex models excel at floor control but fail to integrate mid-speech corrections.} \textbf{Floor Control} (left) shows that most agents yield reliably and quickly to user barge-ins with minimal overlap (\colorbox{green!12}{\phantom{x}} indicates passing threshold; GPT-live-1 struggles). However, \textbf{Correction Uptake} (right) reveals a severe downstream penalty: Artifact Field Accuracy (AFA) on fields requiring mid-speech corrections trails uncorrected fields by 20 to 37 points (\colorbox{red!15}{\phantom{x}} indicates degradation severity). Consequently, over 70\% of all artifact validity (AV) failures stem directly from missed corrections, localizing the true full-duplex bottleneck to state-tracking and information integration rather than raw floor mechanics.}
\label{tab:duplex}
\end{table*}

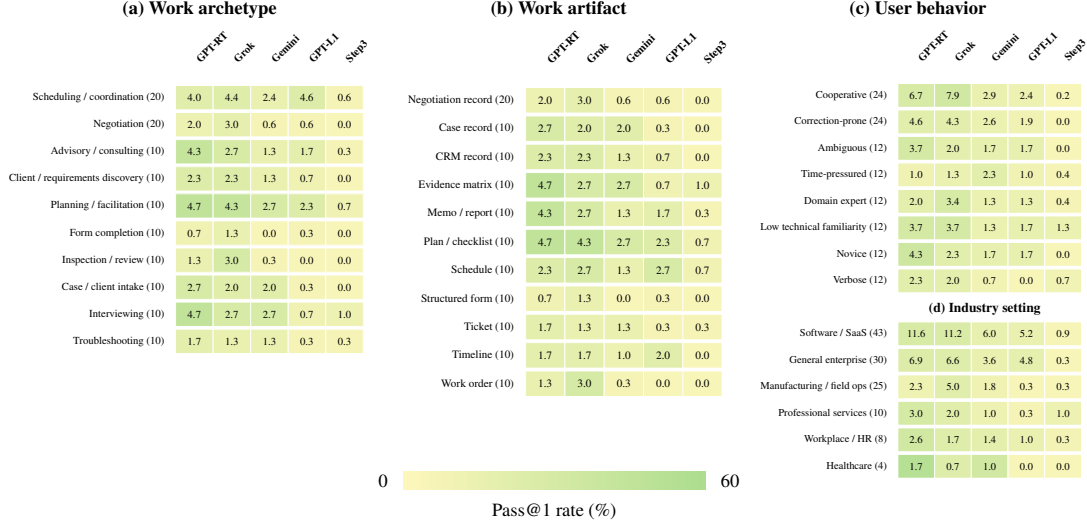
\begin{figure*}[t]
\centering

% Light yellow -> light green palette for better number visibility.
\definecolor{heatlow}{RGB}{255,247,188}
\definecolor{heathigh}{RGB}{173,221,142}

% ================================================================
% PANEL (a): WORK ARCHETYPE
% ================================================================
\begin{minipage}[t]{0.32\textwidth}
\vspace{0pt}
\centering
\textbf{\scriptsize (a) Work archetype}

\vspace{2pt}
\resizebox{\linewidth}{!}{%
\begin{tikzpicture}[x=0.86cm,y=0.62cm]

\path[use as bounding box] (-4.35,1.65) rectangle (4.65,-10.95);

% Column labels
\node[rotate=45,anchor=west,font=\scriptsize\bfseries] at (0,0.22) {GPT-RT};
\node[rotate=45,anchor=west,font=\scriptsize\bfseries] at (1,0.22) {Grok};
\node[rotate=45,anchor=west,font=\scriptsize\bfseries] at (2,0.22) {Gemini};
\node[rotate=45,anchor=west,font=\scriptsize\bfseries] at (3,0.22) {GPT-L1};
\node[rotate=45,anchor=west,font=\scriptsize\bfseries] at (4,0.22) {Step3};

% Row labels
\node[anchor=east,font=\scriptsize] at (-0.65,-1) {Scheduling / coordination (20)};
\node[anchor=east,font=\scriptsize] at (-0.65,-2) {Negotiation (20)};
\node[anchor=east,font=\scriptsize] at (-0.65,-3) {Advisory / consulting (10)};
\node[anchor=east,font=\scriptsize] at (-0.65,-4) {Client / requirements discovery (10)};
\node[anchor=east,font=\scriptsize] at (-0.65,-5) {Planning / facilitation (10)};
\node[anchor=east,font=\scriptsize] at (-0.65,-6) {Form completion (10)};
\node[anchor=east,font=\scriptsize] at (-0.65,-7) {Inspection / review (10)};
\node[anchor=east,font=\scriptsize] at (-0.65,-8) {Case / client intake (10)};
\node[anchor=east,font=\scriptsize] at (-0.65,-9) {Interviewing (10)};
\node[anchor=east,font=\scriptsize] at (-0.65,-10) {Troubleshooting (10)};

% Cells: x / y / pass-rate-for-shading / mean-count-text
\foreach \x/\y/\p/\txt in {
0/-1/20/4.0,1/-1/22/4.4,2/-1/12/2.4,3/-1/23/4.6,4/-1/3/0.6,
0/-2/10/2.0,1/-2/15/3.0,2/-2/3/0.6,3/-2/3/0.6,4/-2/0/0.0,
0/-3/43/4.3,1/-3/27/2.7,2/-3/13/1.3,3/-3/17/1.7,4/-3/3/0.3,
0/-4/23/2.3,1/-4/23/2.3,2/-4/13/1.3,3/-4/7/0.7,4/-4/0/0.0,
0/-5/47/4.7,1/-5/43/4.3,2/-5/27/2.7,3/-5/23/2.3,4/-5/7/0.7,
0/-6/7/0.7,1/-6/13/1.3,2/-6/0/0.0,3/-6/3/0.3,4/-6/0/0.0,
0/-7/13/1.3,1/-7/30/3.0,2/-7/3/0.3,3/-7/0/0.0,4/-7/0/0.0,
0/-8/27/2.7,1/-8/20/2.0,2/-8/20/2.0,3/-8/3/0.3,4/-8/0/0.0,
0/-9/47/4.7,1/-9/27/2.7,2/-9/27/2.7,3/-9/7/0.7,4/-9/10/1.0,
0/-10/17/1.7,1/-10/13/1.3,2/-10/13/1.3,3/-10/3/0.3,4/-10/3/0.3
}{
\pgfmathparse{min(100,8+1.45*\p)}
\let\shadevalue\pgfmathresult
\fill[heathigh!\shadevalue!heatlow] (\x-0.5,\y-0.45) rectangle (\x+0.5,\y+0.45);
\draw[white,line width=0.5pt] (\x-0.5,\y-0.45) rectangle (\x+0.5,\y+0.45);
\node[font=\scriptsize] at (\x,\y) {\txt};
}

\end{tikzpicture}%
}
\end{minipage}
\hfill
% ================================================================
% PANEL (b): WORK artifact
% ================================================================
\begin{minipage}[t]{0.32\textwidth}
\vspace{0pt}
\centering
\textbf{\scriptsize (b) Work artifact}

\vspace{2pt}
\resizebox{\linewidth}{!}{%
\begin{tikzpicture}[x=0.86cm,y=0.62cm]

\path[use as bounding box] (-3.95,1.65) rectangle (4.65,-11.95);

% Column labels
\node[rotate=45,anchor=west,font=\scriptsize\bfseries] at (0,0.22) {GPT-RT};
\node[rotate=45,anchor=west,font=\scriptsize\bfseries] at (1,0.22) {Grok};
\node[rotate=45,anchor=west,font=\scriptsize\bfseries] at (2,0.22) {Gemini};
\node[rotate=45,anchor=west,font=\scriptsize\bfseries] at (3,0.22) {GPT-L1};
\node[rotate=45,anchor=west,font=\scriptsize\bfseries] at (4,0.22) {Step3};

% Row labels
\node[anchor=east,font=\scriptsize] at (-0.65,-1) {Negotiation record (20)};
\node[anchor=east,font=\scriptsize] at (-0.65,-2) {Case record (10)};
\node[anchor=east,font=\scriptsize] at (-0.65,-3) {CRM record (10)};
\node[anchor=east,font=\scriptsize] at (-0.65,-4) {Evidence matrix (10)};
\node[anchor=east,font=\scriptsize] at (-0.65,-5) {Memo / report (10)};
\node[anchor=east,font=\scriptsize] at (-0.65,-6) {Plan / checklist (10)};
\node[anchor=east,font=\scriptsize] at (-0.65,-7) {Schedule (10)};
\node[anchor=east,font=\scriptsize] at (-0.65,-8) {Structured form (10)};
\node[anchor=east,font=\scriptsize] at (-0.65,-9) {Ticket (10)};
\node[anchor=east,font=\scriptsize] at (-0.65,-10) {Timeline (10)};
\node[anchor=east,font=\scriptsize] at (-0.65,-11) {Work order (10)};

\foreach \x/\y/\p/\txt in {
0/-1/10/2.0,1/-1/15/3.0,2/-1/3/0.6,3/-1/3/0.6,4/-1/0/0.0,
0/-2/27/2.7,1/-2/20/2.0,2/-2/20/2.0,3/-2/3/0.3,4/-2/0/0.0,
0/-3/23/2.3,1/-3/23/2.3,2/-3/13/1.3,3/-3/7/0.7,4/-3/0/0.0,
0/-4/47/4.7,1/-4/27/2.7,2/-4/27/2.7,3/-4/7/0.7,4/-4/10/1.0,
0/-5/43/4.3,1/-5/27/2.7,2/-5/13/1.3,3/-5/17/1.7,4/-5/3/0.3,
0/-6/47/4.7,1/-6/43/4.3,2/-6/27/2.7,3/-6/23/2.3,4/-6/7/0.7,
0/-7/23/2.3,1/-7/27/2.7,2/-7/13/1.3,3/-7/27/2.7,4/-7/7/0.7,
0/-8/7/0.7,1/-8/13/1.3,2/-8/0/0.0,3/-8/3/0.3,4/-8/0/0.0,
0/-9/17/1.7,1/-9/13/1.3,2/-9/13/1.3,3/-9/3/0.3,4/-9/3/0.3,
0/-10/17/1.7,1/-10/17/1.7,2/-10/10/1.0,3/-10/20/2.0,4/-10/0/0.0,
0/-11/13/1.3,1/-11/30/3.0,2/-11/3/0.3,3/-11/0/0.0,4/-11/0/0.0
}{
\pgfmathparse{min(100,8+1.45*\p)}
\let\shadevalue\pgfmathresult
\fill[heathigh!\shadevalue!heatlow] (\x-0.5,\y-0.45) rectangle (\x+0.5,\y+0.45);
\draw[white,line width=0.5pt] (\x-0.5,\y-0.45) rectangle (\x+0.5,\y+0.45);
\node[font=\scriptsize] at (\x,\y) {\txt};
}

\end{tikzpicture}%
}
\end{minipage}
\hfill
% ================================================================
% PANELS (c,d): USER BEHAVIOR + INDUSTRY SETTING
% ================================================================
\begin{minipage}[t]{0.32\textwidth}
\vspace{0pt}
\centering
\textbf{\scriptsize (c) User behavior}

\vspace{2pt}
\resizebox{\linewidth}{!}{%
\begin{tikzpicture}[x=0.86cm,y=0.62cm]

\path[use as bounding box] (-4.55,1.65) rectangle (4.65,-16.15);

% Shared column labels
\node[rotate=45,anchor=west,font=\scriptsize\bfseries] at (0,0.22) {GPT-RT};
\node[rotate=45,anchor=west,font=\scriptsize\bfseries] at (1,0.22) {Grok};
\node[rotate=45,anchor=west,font=\scriptsize\bfseries] at (2,0.22) {Gemini};
\node[rotate=45,anchor=west,font=\scriptsize\bfseries] at (3,0.22) {GPT-L1};
\node[rotate=45,anchor=west,font=\scriptsize\bfseries] at (4,0.22) {Step3};

% User behavior row labels
\node[anchor=east,font=\scriptsize] at (-0.65,-1) {Cooperative (24)};
\node[anchor=east,font=\scriptsize] at (-0.65,-2) {Correction-prone (24)};
\node[anchor=east,font=\scriptsize] at (-0.65,-3) {Ambiguous  (12)};
\node[anchor=east,font=\scriptsize] at (-0.65,-4) {Time-pressured (12)};
\node[anchor=east,font=\scriptsize] at (-0.65,-5) {Domain expert (12)};
\node[anchor=east,font=\scriptsize] at (-0.65,-6) {Low technical familiarity (12)};
\node[anchor=east,font=\scriptsize] at (-0.65,-7) {Novice  (12)};
\node[anchor=east,font=\scriptsize] at (-0.65,-8) {Verbose  (12)};

\foreach \x/\y/\p/\txt in {
0/-1/28/6.7,1/-1/33/7.9,2/-1/12/2.9,3/-1/10/2.4,4/-1/1/0.2,
0/-2/19/4.6,1/-2/18/4.3,2/-2/11/2.6,3/-2/8/1.9,4/-2/0/0.0,
0/-3/31/3.7,1/-3/17/2.0,2/-3/14/1.7,3/-3/14/1.7,4/-3/0/0.0,
0/-4/8/1.0,1/-4/11/1.3,2/-4/19/2.3,3/-4/8/1.0,4/-4/3/0.4,
0/-5/17/2.0,1/-5/28/3.4,2/-5/11/1.3,3/-5/11/1.3,4/-5/3/0.4,
0/-6/31/3.7,1/-6/31/3.7,2/-6/11/1.3,3/-6/14/1.7,4/-6/11/1.3,
0/-7/36/4.3,1/-7/19/2.3,2/-7/14/1.7,3/-7/14/1.7,4/-7/0/0.0,
0/-8/19/2.3,1/-8/17/2.0,2/-8/6/0.7,3/-8/0/0.0,4/-8/6/0.7
}{
\pgfmathparse{min(100,8+1.45*\p)}
\let\shadevalue\pgfmathresult
\fill[heathigh!\shadevalue!heatlow] (\x-0.5,\y-0.45) rectangle (\x+0.5,\y+0.45);
\draw[white,line width=0.5pt] (\x-0.5,\y-0.45) rectangle (\x+0.5,\y+0.45);
\node[font=\scriptsize] at (\x,\y) {\txt};
}

% Industry panel title
\node[anchor=west,font=\small\bfseries] at (0.20,-9.05) {(d) Industry setting};

% Industry row labels
\node[anchor=east,font=\scriptsize] at (-0.65,-10) {Software / SaaS (43)};
\node[anchor=east,font=\scriptsize] at (-0.65,-11) {General enterprise (30)};
\node[anchor=east,font=\scriptsize] at (-0.65,-12) {Manufacturing / field ops (25)};
\node[anchor=east,font=\scriptsize] at (-0.65,-13) {Professional services (10)};
\node[anchor=east,font=\scriptsize] at (-0.65,-14) {Workplace / HR (8)};
\node[anchor=east,font=\scriptsize] at (-0.65,-15) {Healthcare (4)};

\foreach \x/\y/\p/\txt in {
0/-10/27/11.6,1/-10/26/11.2,2/-10/14/6.0,3/-10/12/5.2,4/-10/2/0.9,
0/-11/23/6.9,1/-11/22/6.6,2/-11/12/3.6,3/-11/16/4.8,4/-11/1/0.3,
0/-12/9/2.3,1/-12/20/5.0,2/-12/7/1.8,3/-12/1/0.3,4/-12/1/0.3,
0/-13/30/3.0,1/-13/20/2.0,2/-13/10/1.0,3/-13/3/0.3,4/-13/10/1.0,
0/-14/33/2.6,1/-14/21/1.7,2/-14/17/1.4,3/-14/12/1.0,4/-14/4/0.3,
0/-15/56/1.7,1/-15/22/0.7,2/-15/33/1.0,3/-15/0/0.0,4/-15/0/0.0
}{
\pgfmathparse{min(100,8+1.45*\p)}
\let\shadevalue\pgfmathresult
\fill[heathigh!\shadevalue!heatlow] (\x-0.5,\y-0.45) rectangle (\x+0.5,\y+0.45);
\draw[white,line width=0.5pt] (\x-0.5,\y-0.45) rectangle (\x+0.5,\y+0.45);
\node[font=\scriptsize] at (\x,\y) {\txt};
}

\end{tikzpicture}%
}
\end{minipage}

\vspace{-0.45cm}

% Shared color scale
\begin{tikzpicture}[x=1cm,y=1cm]
\shade[left color=heatlow,right color=heathigh] (0,0) rectangle (4.0,0.24);
\node[anchor=east,font=\scriptsize] at (-0.08,0.12) {0};
\node[anchor=west,font=\scriptsize] at (4.08,0.12) {60};
\node[font=\scriptsize] at (2.0,-0.28) {Pass@1 rate (\%)};
\end{tikzpicture}

\vspace{-0.30cm}
\caption{\small{\textbf{Model performance distribution across \apex{} taxonomy dimensions.} We report the mean number of successful workflows, while color intensity reflects the \passone{} rate. Voice agents exhibits strong heterogeneity, highlighting that capabilities vary sharply depending on work archetype, work artifact format, user behavior, and industry setting.}}
\label{fig:taxonomy_heatmaps_main}
\end{figure*}

\subsection{Can Voice Agents Reliably Complete Professional Work?}
\vspace{-0.2cm}

\paragraph{Occasional success does not translate into dependable execution.} Table~\ref{tab:main_results} shows that current real-time voice agents can complete non-trivial professional workflows, but do so inconsistently. GPT-realtime-2.1 achieves the highest \passone{} (23.6\%) and \reliable{} (10.8\%), while Grok-Voice-Think-2.0 reaches the highest \passthree{} (41.7\%) but succeeds in all three runs on only 5.0\% of workflows. This separation between \passthree{} and \reliable{} appears across every system: models often find a successful trajectory in one attempt without reproducing it reliably. One-shot task success therefore substantially overstates readiness for delegated work. See Appendix~\ref{app:reliability} for reliability--efficiency analyses.
\vspace{-0.3cm}

\paragraph{Successful execution does not necessarily imply efficient tool use.} Table~\ref{tab:main_results} shows that voice agents differ substantially in how they reach comparable outcomes. GPT-realtime-2.1 operates close to the oracle tool-call count (0.98), whereas Grok-Voice-Think-2.0 uses considerably more calls (1.51) despite similar \passone{}. Thus, aggregate task success alone does not reveal whether an agent reaches the desired outcome efficiently, with under-utilization leading to suboptimal performance, and over-utilization causing wasted tokens and dead cycles.
\vspace{-0.3cm}

\paragraph{The principal end-to-end bottleneck is producing a valid work artifact.}
The gate decomposition in Table~\ref{tab:workflow_gates} reveals a striking gap between conversational progress and final deliverable correctness. Grok-Voice-Think-2.0 reaches the required Target State in 99.4\% of runs, while GPT-Live-1 attains 91.9\% Process Validity. Despite high field-level correctness, Artifact Validity is the lowest gate for every system and peaks at only 35.3\% for GPT-realtime-2.1. Hence, current voice agents can gather most required information and complete much of the workflow, but frequently fail to compose these locally correct decisions into a fully valid persistent outcome. Professional work is inherently conjunctive: a small number of missed fields, revisions, or dependent actions can invalidate an otherwise strong trajectory. Appendix~\ref{app:reliability} reports the corresponding per-model gate-failures.
\vspace{-0.2cm}

\subsection{What Makes Full-Duplex Execution Difficult?}
\vspace{-0.2cm}

\begin{wraptable}{r}{0.5\textwidth}
\vspace{-12pt} % Pulls the table up slightly to align with paragraph text
\centering
\scriptsize
\setlength{\tabcolsep}{3.1pt}
\renewcommand{\arraystretch}{1.12}
\resizebox{\linewidth}{!}{%
\begin{tabular}{@{}lcccccc@{}}
\toprule
& \multicolumn{5}{c}{\textbf{Text controls}}
& \cellcolor{primarygray}\textbf{Voice reference} \\
\cmidrule(lr){2-6}\cmidrule(l){7-7}

\textbf{Metric}
& \shortstack{\textbf{Claude}\\\textbf{Opus-5.5}}
& \shortstack{\textbf{GPT-}\\\textbf{6-sol}}
& \shortstack{\textbf{GPT-}\\\textbf{5.5}}
& \shortstack{\textbf{Gemini}\\\textbf{3.8-Flash}}
& \shortstack{\textbf{Kimi}\\\textbf{K3}}
& \cellcolor{primarygray}\shortstack{\textbf{GPT-}\\\textbf{realtime-2.1}} \\
\midrule

\passone{} (\%) $\uparrow$
& 58.7
& 55.0
& \textbf{62.0}
& 54.3
& 55.8
& \cellcolor{primarygray}23.3 \\

\afa{} (\%) $\uparrow$
& 97
& 96
& 95
& \textbf{98}
& 97
& \cellcolor{primarygray}91 \\

\bottomrule
\end{tabular}%
}
\caption{\small\textbf{Workflow success under text-only control.} Frontier text-based LLM agents reach 62\% Pass@1, whereas the strongest voice agent stays $\leq$25\%. Full-duplex interactions substantially augment the baseline difficulty for voice models.}
\label{tab:text_control}
\vspace{-10pt} % Reduces empty space below the caption before text resumes
\end{wraptable}
\paragraph{Basic floor control is comparatively strong; maintaining correct state through interruptions is not.}
Table~\ref{tab:duplex} separates two aspects of full-duplex behavior. GPT-realtime-2.1, Grok-Voice-Think-2.0, Gemini-3.8-Live, and Step-Audio3 yield on 99.8--100\% of user barge-ins, keep unintended overlap below 5\%, and stop within 20--156\,ms at the median. GPT-Live-1 is the notable exception, exhibiting substantially greater overlap and slower stopping. Thus, for most frontier systems, gross failures of conversational floor control cannot explain the low end-to-end success rates. However, the larger failure appears after the interruption. \afa{} on fields requiring mid-speech correction is 20.0--36.8 points below accuracy on other fields, and 71--89\% of Artifact Validity failures are correction-linked. Full-duplex competence therefore requires more than detecting an interruption and yielding the floor: revised information must replace stale state and propagate correctly into downstream artifacts and actions. See Appendix~\ref{app:qual} for state-capture and artifact-update failures.
\vspace{-0.2cm}

\subsection{How Much Difficulty Is Specific to Voice?}
\vspace{-0.2cm}

\paragraph{Real-time speech compounds an already difficult agentic problem.}
Table~\ref{tab:text_control} provides an important control on benchmark difficulty. Frontier text agents reach 54.3--62.0\% \passone{} and 95--98\% \afa{} on the same workflow environment, substantially above the voice reference at roughly 23\% \passone{} and 91\% \afa{}. Notably, the modality gap is much larger for complete workflow success than for individual field accuracy. This suggests that real-time spoken interaction primarily stresses the ability to maintain and execute a coherent workflow state over time. At the same time, text performance remains far from saturation, showing that \apex{} combines two difficult problems: long-horizon professional work execution and real-time spoken interaction.
\vspace{-0.2cm}

\subsection{Where Do Models Succeed and Fail?}
\vspace{-0.2cm}

\paragraph{Aggregate scores conceal strong sensitivity to workflow structure.} Figure~\ref{fig:taxonomy_heatmaps_main} (more in Appendix~\ref{app:taxonomy_slices}) shows substantial variation across professional work archetypes and deliverable formats: models are generally stronger on planning, interviewing, advisory, and inspection workflows, while rigid form completion, negotiation, and troubleshooting remain substantially harder. The work-artifact slices show a similar pattern, with flexible plans and evidence-oriented outputs proving more tractable than structured forms, tickets, and work orders. A single leaderboard score therefore hides materially different capability profiles across forms of professional work. 
\vspace{-0.3cm}

\paragraph{Performance is also sensitive to how users interact with voice agents.} User-behavior settings provide a complementary view of the benchmark difficulty. Correction-prone interactions reduce \passone{} from 28\% to 19\% for GPT-realtime-2.1 and from 33\% to 18\% for Grok-Voice-Think-2.0, while time pressure, verbosity, and ambiguity also degrade performance for most systems. Overall, current frontier voice agents remain sensitive to the pace, clarity, and stability with which users communicate task information. The full user-behavior breakdown in Appendix~\ref{app:taxonomy_slices}.

\vspace{-0.3cm}
\section{Discussion}
\label{sec:discussion}
\vspace{-0.3cm}

\paragraph{Conversational fluency and professional reliability are distinct capabilities.}
The central result of \apex{} is the gap between locally strong behavior and globally correct work. Frontier voice agents can manage interruptions, recover required information, and execute many of the intended tool use, yet only a small fraction of runs produce a fully valid deliverable. Fluent interaction therefore does not imply dependable task completion.
\vspace{-0.3cm}

\paragraph{The key full-duplex challenge is maintaining an evolving task state.}
For several voice agents, yielding when the user barges in is already highly reliable; the harder problem is incorporating what the user says next. A dependable professional voice agent must update its internal task representation as information changes, allowing revised values to replace stale state and propagate correctly through artifacts and downstream actions. We provide extensive qualitative analysis in Appendix~\ref{app:qualitative_analysis_exhaustive}.
\vspace{-0.3cm}

\paragraph{Professional voice agents require advances in both agentic execution and real-time interaction.}
The text controls show that removing real-time speech substantially improves performance, but does not solve the benchmark. Conversely, voice systems remain relatively close to text systems on field-level correctness while falling much further behind on end-to-end completion. The remaining challenge is therefore not speech perception, tool use, or reasoning in isolation, but their composition: agents must listen continuously, revise state, plan actions, and maintain a persistent work artifact without losing consistency as the conversation evolves. Closing this gap is necessary for moving from voice systems that converse fluently to agents that can be trusted with professional workflows.
\vspace{-0.3cm}

\section{Conclusion}
\vspace{-0.3cm}

We introduced \apex{}, the first benchmark evaluating whether full-duplex voice agents can convert natural spoken delegation into correct, verifiable professional work artifacts. Across 120 interactive task instances and five frontier systems, we reveal a massive separation between local conversational mechanics and end-to-end deliverable success. Through a combination of factorized professional-work taxonomy, reactive but reproducible user simulation, typed artifacts, and repeated-run reliability, \apex{} shifts voice-agent evaluation from ``did the conversation go well?'' toward the more demanding question: ``did the agent reliably finish the work?''
\vspace{-0.2cm}

\section*{AI Use Statement}
\vspace{-0.2cm}

Generative AI was used to produce offline synthetic user-side speech generation during benchmark construction, for the semantic-equivalence judgments described in Section~\ref{sec:evaluation}, and for language editing of the manuscript. Generated benchmark content was subject to the consistency, leakage-control, and validation procedures described in the paper and appendix. All benchmark design decisions, evaluation protocols, experimental analyses were reviewed and verified by the authors who take responsibility for the final manuscript and study.
\vspace{-0.2cm}

\section*{Ethics Statement}
\vspace{-0.2cm}

\apex{} consists of synthetic task worlds, entities, and user interactions and contains no real personal or person-specific information. Workflows in domains such as healthcare operations, finance, legal services, and human resources are designed to evaluate workflow execution and conversational behavior, not to validate clinical, financial, legal, employment, or other licensed professional decision-making. User profiles vary interaction style only and are not intended to model or infer protected characteristics. All consequential actions occur within the simulated Voice Workbench environment. As appropriate conversational behavior can depend on social and cultural context, benchmark-defined interaction styles should not be interpreted as universally preferred behavior.
\vspace{-0.2cm}

\section*{Reproducibility Statement}
\vspace{-0.2cm}

\apex{} is designed for reproducible full-duplex evaluation despite adaptive conversations. User-side language and speech are generated offline and stored in a validated, frozen speech bank; given the same workflow state and simulator seed, the user policy selects the same realization, while agent behavior may induce different valid interaction branches. Complete user, agent, tool, state, authorization, and timing events are recorded to support deterministic replay and rescoring. The appendix provides benchmark-generation and user-policy specifications, prompts, leakage controls, speech-synthesis and orchestration details, grading rules and LLM-judge prompt, model-specific evaluation settings, human-validation protocol, and commands for regenerating the reported tables and figures.

\bibliographystyle{iclr2027_conference}
\bibliography{iclr2027_conference}

\clearpage
\appendix
\include{appendix}

\end{document}

%% file: appendix.tex
% ============================================================================
% APEX-Voice Appendix (restructured)
% Requires in main preamble: \usepackage{fvextra}
%                            \usepackage{float}
%                            \usepackage{needspace}
% Main file should invoke \appendix before \include{appendix}.
% ============================================================================

\section*{Additional Limitations}
\label{app:limits}

The limitations below complement the discussion in the main paper and delimit what should and should not be inferred from the current benchmark release.

\begin{itemize}[leftmargin=*]
\item \textbf{Bounded work units, not job replacement.}
\apex{} evaluates selected professional work units; it does not estimate labor substitution, cash value of the jobs, or the fraction of an occupation that can be automated.
\item \textbf{Synthetic benchmark users.}
The headline user is a controlled synthetic policy with frozen audio realizations. This provides reproducibility but cannot capture the full variability of human speech and workplace interaction. We therefore include a 24-task live-human audit in Appendix~\ref{app:human_study}; its final results should be interpreted as a simulator-validity check rather than as a replacement leader board.
\item \textbf{Non-exhaustive taxonomy factors.}
The taxonomy was designed for coverage rather than a fully crossed factorial experiment. Work archetype, artifact class, knowledge, autonomy, and tool burden therefore co-vary. Taxonomy dimensions provide capability diagnostics, and are not causal of labeling schema.
\item \textbf{English-only release and system dependence.}
The current realization bank is English. Some evaluated systems depend on hosted provider interfaces or composite backends. Results characterize the tested configurations.
\item \textbf{Model coverage and access.} Several evaluated systems depend on hosted endpoints but are liable to future evolution and possible depreciation. Exact versions and integration details are recorded in Appendix~\ref{app:experiment_settings}.
\item \textbf{Composite systems.} GPT-live-1's score reflects a voice layer plus a default backend; a different backend may change results. We therefore describe this as a configuration-level comparison rather than a causal architecture claim.
\end{itemize}

\section{Benchmark Taxonomy and Composition}
\label{app:workflow_taxonomy}

The \apex{} taxonomy specifies complementary properties of each workflow rather than a single difficulty label. The eleven dimensions describe the professional operation, organizational context, expected work artifact, interaction dynamics, execution requirements, and consequence profile used during benchmark construction. Table~\ref{tab:taxonomy} lists every realized label in the current benchmark, while Table~\ref{tab:taxonomy_definitions} summarizes how each dimension is interpreted. Because the benchmark was constructed for broad coverage rather than as a fully crossed factorial design, taxonomy slices should be interpreted descriptively rather than causally.

\begin{table}[t]
\centering
\small
\setlength{\tabcolsep}{5pt}
\renewcommand{\arraystretch}{1.08}
\begin{tabular}{p{0.23\textwidth} p{0.70\textwidth}}
\toprule
\textbf{Taxonomy dimension} & \textbf{Realized labels} \\
\midrule
Work archetype & form-fill, interview, intake, troubleshoot, negotiate, coordinate, discovery, advise, facilitate, inspect \\
Industry / setting & Software/SaaS, horizontal enterprise, manufacturing/field operations, professional services, workplace/HR, healthcare, insurance \\
Work artifact & Structured form, case record, CRM record, evidence matrix, memo/report, plan/checklist, schedule, ticket, timeline, negotiation record, work order \\
Economic role / function & Recruiting, HR operations, sales, customer success, technical support, insurance operations, finance operations, procurement, project management, operations, field service, consulting, compliance, executive assistance \\
Duplex phenomenon & Backchannel, user barge-in, mid-speech correction, cancellation/revocation, intent switch, clarification, overlapping speech \\
Delegation pattern & delegate, complete, revise, follow-through, approve \\
Autonomy & Prepare-only, draft-and-confirm, low-risk execute, approval-gated commit \\
Knowledge burden & None, supplied evidence, small-search retrieval, multi-document reasoning \\
Tool burden & Light, moderate \\
User behavior & Cooperative, correction-prone, ambiguous, distracted/time-pressured, domain expert, low-tech expertise, novice, verbose \\
Risk / safety & Routine, sensitive-data simulation, consequential action, special review \\
\bottomrule
\end{tabular}
\caption{\small{\textbf{Complete \apex{} taxonomy.} The table enumerates the labels realized in the 120-workflow benchmark and used for construction and capability-level analysis.}}
\label{tab:taxonomy}
\end{table}

\begin{table}[t]
\centering
\small
\setlength{\tabcolsep}{5pt}
\renewcommand{\arraystretch}{1.08}
\begin{tabular}{p{0.22\textwidth} p{0.71\textwidth}}
\toprule
\textbf{Dimension} & \textbf{Operational interpretation} \\
\midrule
Work archetype & Underlying professional operation, such as interviewing, troubleshooting, coordination, negotiation, or form completion. \\
Industry / setting & Organizational context that determines task terminology, policies, resources, and constraints. \\
Work artifact & Persistent and verifiable output produced by the workflow, such as a form, case record, schedule, memo, or negotiation record. \\
Economic role / function & Professional function carrying out the workflow, such as recruiting, sales, technical support, procurement, or project management. \\
Duplex phenomenon & Task-critical real-time conversational event, including barge-ins, overlap, corrections, cancellations, clarifications, and backchannels. \\
Delegation pattern & How the workflow evolves: inferring procedure (delegate), discovering missing information (complete), propagating corrections (revise), adapting to environment changes (follow-through), or obtaining authorization (approve). \\
Autonomy & Which operations may proceed independently and which require user confirmation or explicit approval. \\
Knowledge burden & Whether completion relies on local state, supplied evidence, retrieval over a small knowledge source, or reasoning across multiple documents. \\
Tool burden & Amount of structured tool use required beyond conversation, including API calls, database updates, and state-changing environment actions. \\
User behavior & How workflow-relevant information is communicated, including cooperative, ambiguous, correction-prone, expert, distracted, novice, or verbose behavior. \\
Risk / safety & Consequence profile of an incorrect or unauthorized outcome; this is distinct from autonomy, which specifies what the agent is permitted to do. \\
\bottomrule
\end{tabular}
\caption{\small{\textbf{Interpretation of taxonomy dimensions.} These definitions mirror the benchmark-design criteria used in the main paper and clarify how labels should be read when interpreting the slice analyses in Appendix~\ref{app:taxonomy_slices}.}}
\label{tab:taxonomy_definitions}
\end{table}

\section{Benchmark Construction and Runtime Specification}
\label{app:construction}

This section expands the construction and execution details summarized in the main paper. Each workflow is instantiated from a shared task specification into a stateful environment containing the synthetic user state, knowledge, typed tools, evolving work artifact, authorization constraints, frozen speech assets, and grading specification. The construction pipeline is designed so that runtime variation comes from agent behavior rather than an uncontrolled user-side generation process.

\subsection{User state, reveal policy, and leakage controls}
Each task defines a structured synthetic user state containing only information the user is entitled to know: persona-level speaking preferences, currently known task facts, mutable facts that may later be corrected, user goals, constraints, approval state, and observed external events. Gold grader labels and hidden professional procedure are never included in user state. Facts carry explicit reveal rules, so a benchmark-critical value may be volunteered, withheld until an appropriate question, or released only when an authored correction or external event fires. Corrections create a new fact version that supersedes the old value; downstream artifact graders can therefore detect stale information.

The runtime flow engine consumes observed agent acts, tool events, artifact mutations, approval requests, and seeded world events and emits a structured user plan such as \textsc{answer}, \textsc{correct}, \textsc{approve}, \textsc{revoke}, or \textsc{barge-in}. The plan contains semantic intent and allowed facts but no surface wording. This separation is the main leakage barrier: the surface generator cannot reveal future facts or the benchmark's reference trajectory because those items are absent from the plan it receives.

\subsection{Offline language realization and speech synthesis}
For every reachable benchmark-critical user plan, the construction pipeline generates 2--5 natural-language variants offline. The reported construction uses \textit{GPT-5.6-Sol} for offline language realization. Realization quality control rejects variants that introduce new specific graded facts, leak hidden state or required procedure, contradict the task world, or fail to express the required semantic act. Accepted variants are frozen into a task-local realization bank. Speech is then compiled offline using the pinned local TTS configuration (Kokoro in the reported release) with deterministic persona, pronunciation, and seed settings. The scored runtime contains no free-running user LLM or TTS fallback.

At runtime, a deterministic selector maps task identity, simulator seed, and user-plan identity to one validated text/audio asset. This guarantees reproducible surface realization for the same semantic user state without forcing different agents through an identical transcript. User audio is streamed in 40\,ms frames on its own channel, while agent audio is timestamped independently, preserving actual overlap and interruption timing in the archived stereo trace.

\subsection{Runtime orchestration and approval semantics}
The asynchronous orchestrator maintains both wall-clock time and media time. Authored duplex events are anchored to media time so a slow model cannot shift a correction, interruption, or revocation simply by responding late. Agent transcripts are mapped to a compact semantic-act ontology; structured tool calls are read directly from the event stream. The event bus records user, agent, tool, artifact, environment, authorization, timing, and grading events in a single replayable trace.

Approval-gated actions are enforced by the environment rather than by prompt compliance alone. A valid approval creates an action-scoped token; a later revocation immediately invalidates that token. Prepare, review, approval, and commit are therefore distinct workspace states. A model that verbally acknowledges a revocation but still commits the action fails the process-validity gate. Blocked post-revocation commit attempts are retained in the canonical event log.

\subsection{Full-duplex evaluation setting}
All reported experiments use the same full-duplex voice runtime. Frozen user audio is streamed on the user channel while agent audio arrives independently on the agent channel; both share a media-time clock so overlapping speech is represented directly in the trace. The task world, user policy, realization bank, artifact schemas, tools, and graders are fixed across systems. The realized V1 taxonomy does not vary modality condition, temporal dynamics, or grading mode in the reported campaign, so these are treated as controlled scope rather than experimental axes.

\subsection{Construction quality control and workflow assembly}
The construction pipeline validates consistency across the environment, workflow state, work artifact, knowledge, tools, user simulation policy, speech bank, and gold state using deterministic checks together with LLM-judge checks. As reported in the main paper, 16\% of candidate workflows were rejected for semantic inconsistencies, missing information, invalid shortcut completions, unauthorized actions, or premature disclosure of hidden information. Only validated components are packaged into the executable benchmark workflows used for evaluation.

\section{Evaluation and Grading}
\label{app:grading}

All reported scores are computed after execution from the final Voice Workbench state and the canonical event trace. Workflow-level success is deliberately conjunctive: partial progress does not count as successful professional completion if a required terminal state, process constraint, action, or final artifact remains invalid. Artifact Field Accuracy is reported separately to expose partial correctness when this end-to-end criterion is not met.

\subsection{Workflow Success}
For workflow $i$, Workflow Success is the product of four binary gates,
\begin{equation}
\mathrm{WS}_i = \mathrm{TS}_i \cdot \mathrm{PV}_i \cdot \mathrm{AC}_i \cdot \mathrm{AV}_i,
\end{equation}
where \emph{Target State} (TS) verifies the required terminal state, \emph{Process Validity} (PV) checks that no forbidden action occurred (for example, committing without approval or after revocation), \emph{Action Completion} (AC) requires all task-mandated tool calls and actions to pass, and \emph{Artifact Validity} (AV) requires all graded fields and the artifact lifecycle state to be correct. Thus, $\mathrm{WS}_i=1$ only when all four gates pass.

\subsection{Artifact Field Accuracy}
Let $F_i$ denote the graded fields of workflow $i$ and $c_{ij}\in\{0,1\}$ denote correctness of field $j$. Artifact Field Accuracy is micro-averaged across graded fields:
\begin{equation}
\mathrm{AFA}=\frac{\sum_i\sum_{j\in F_i}c_{ij}}{\sum_i |F_i|}.
\end{equation}
Unlike Workflow Success, \afa{} does not require the complete workflow to pass and therefore separates field-level correctness from complete professional-work execution.

\subsection{Deterministic verification and semantic escalation}
Deterministic field graders cover exact and case-folded equality, normalized phone/date/address representations, enumerations, numeric tolerance, set precision/recall/F1, ordered lists, and interval overlap. Required state transitions, tool/evidence actions, approval boundaries, and artifact lifecycle conditions are checked directly from the event log. Field values that do not pass deterministic verification and admit representation-equivalent free-text answers are sent to the cached temperature-0 semantic judge reproduced in Appendix~\ref{app:prompts}. Grading is a pure function of the recorded run trace plus the versioned judge cache.

\subsection{Repeated-run reliability}
To measure stochastic reliability, each workflow is executed in three independent agent rollouts with identical task semantics and frozen user policy. Let $\mathrm{WS}_{i,r}\in\{0,1\}$ denote success for workflow $i$ on run $r$. We report
\begin{equation}
\begin{aligned}
\mathrm{Pass@1} &= \frac{1}{3N}\sum_{i=1}^{N}\sum_{r=1}^{3}\mathrm{WS}_{i,r}, \\
\mathrm{Pass@3} &= \frac{1}{N}\sum_{i=1}^{N}\mathbb{1}\!\left[\sum_{r=1}^{3}\mathrm{WS}_{i,r}\geq 1\right], \\
\mathrm{Reliable@3} &= \frac{1}{N}\sum_{i=1}^{N}\mathbb{1}\!\left[\sum_{r=1}^{3}\mathrm{WS}_{i,r}=3\right].
\end{aligned}
\end{equation}
These metrics distinguish average single-run success, whether a workflow succeeds at least once across three attempts, and whether it succeeds consistently in all three attempts.

\section{Experimental Settings and Evaluated Systems}
\label{app:experiment_settings}
\label{app:models}

\subsection{Common evaluation protocol}
All five evaluated voice systems run through the same full-duplex orchestrator and Voice Workbench. Every system receives the same 120 workflow specifications, initial environment state, knowledge resources, typed tools, artifact schemas, authorization rules, and frozen user-side assets. Each workflow is evaluated over three runs, and provider-specific streaming events are normalized by a common adapter before they are written to the canonical trace used for grading. Agent behavior is free to induce different valid interaction branches, while user-side selection remains deterministic for a fixed workflow identity, simulator seed, and user-plan identity.

Adapters conform to a common \texttt{AgentAdapter} interface and emit normalized events including \path{response.audio.delta} (PCM16 bytes), \path{response.audio_transcript.delta/.done}, \path{response.output_item.done} (tool call), \path{response.done}, and \texttt{error}.

\subsection{Provider-specific API and adapter details}
\begin{itemize}[leftmargin=*]
\item \textbf{Gemini-3.8-Live} --- Google \texttt{google-genai} Live API; function calling; 16\,kHz input / 24\,kHz output; server turn detection disabled for matched conditions.
\item \textbf{GPT-realtime-2.1} --- OpenAI real-time protocol served via an internal LLM proxy (\path{/v1/realtime}). Required fixes for validity were \path{tool_choice="auto"} (without it the model never calls tools, yielding 0\% \afa{}), clamping \texttt{voice} to the OpenAI allowlist (an unknown voice rejects the entire \path{session.update}, silently dropping tools), and coalescing \path{response.create} against the active-response lifecycle to avoid \path{conversation_already_has_active_response} truncation.
\item \textbf{Step-Audio3} --- StepFun real-time (\path{wss://api.stepfun.ai/v1/realtime}); PCM16 mono 24\,kHz; explicit \path{input_audio_buffer.commit} + \path{response.create}; tools represented as OpenAI-style function definitions; barge-in handled via \path{response.cancel}.
\item \textbf{GPT-live-1} --- OpenAI Live API (\path{wss://api.openai.com/v1/live/sessions}; model specified in a \path{session.start} message rather than the URL). The voice layer delegates cognition and function calling to a backend text model (\texttt{gpt-5.6 Sol} by default in the reported configuration); reported scores therefore characterize the composite system. Audio is \texttt{\{type:audio/pcm, rate:24000\}} and the default voice is \texttt{marin}.
\item \textbf{Grok-Voice-Think-2.0} --- xAI real-time (\path{wss://api.x.ai/v1/realtime}; model \path{grok-voice-think-fast-2.0}), with OpenAI-real-time-compatible event names (\path{response.output_audio.delta}, \path{response.output_audio_transcript.*}) remapped to the normalized event interface. Tool calls are sourced from \path{response.function_call_arguments.done}; voice \texttt{xai\_ara}. A parallel-tool-call hang was fixed by submitting all \path{function_call_output}s before issuing a single continuation \path{response.create}.
\end{itemize}

\subsection{Deferred systems}
Two systems considered during implementation were not included in the reported five-model campaign. \textbf{Nemotron-VoiceChat-11B} exposes native \texttt{<TOOLCALL>} function calling, but evaluation was not practical because runs took approximately hours per task even with a KV-cache patch and tool output was empty or garbled. \textbf{Venus-real-time} exposed tool use only through natural-language delegation to a separate harness, creating a tool-interface mismatch with the common evaluation protocol. These systems are therefore described as deferred rather than scored baselines.

\subsection{Deterministic field graders}
\label{app:field_graders}
Table~\ref{tab:field_graders} lists every field grader in the released scorer. Each returns a score in $[0,1]$; a required field is counted correct only at $1.0$. Values are dictated aloud, so every grader first applies a canonicalization appropriate to its type before comparison.

\begin{table}[t]
\centering
\scriptsize
\setlength{\tabcolsep}{4pt}
\renewcommand{\arraystretch}{1.2}
\resizebox{\textwidth}{!}{%
\begin{tabular}{p{0.15\textwidth} p{0.55\textwidth} p{0.30\textwidth}}
\toprule
\textbf{Grader} & \textbf{Rule} & \textbf{Example (gold $\Leftarrow$ agent $\Rightarrow$ score)} \\
\midrule
\texttt{exact} & String equality after whitespace trim. & \texttt{E4471} $\Leftarrow$ \texttt{E4471} $\Rightarrow 1$ \\
\texttt{casefold\_exact} & Case-insensitive equality after trim. & \texttt{HDHP} $\Leftarrow$ \texttt{hdhp} $\Rightarrow 1$ \\
\texttt{normalized\_phone} & Compare digits only (strip formatting). & \texttt{4155550134} $\Leftarrow$ \texttt{(415)\,555-0134} $\Rightarrow 1$ \\
\texttt{normalized\_date} & Canonicalize to \texttt{YYYY-MM-DD} (ISO, US slash, or month-name forms). & \texttt{1988-07-09} $\Leftarrow$ \texttt{July 9, 1988} $\Rightarrow 1$ \\
\texttt{normalized\_address} & Lowercase, collapse non-alphanumerics to spaces, compare. & \texttt{12 oak st} $\Leftarrow$ \texttt{12 Oak St.} $\Rightarrow 1$ \\
\texttt{enum} & Exact match against a controlled value. & \texttt{submitted} $\Leftarrow$ \texttt{submitted} $\Rightarrow 1$ \\
\texttt{numeric\_tolerance} & Extract number from prose/currency; pass iff $|pred-gold|\le tol$. & $225\ (tol\,0.01) \Leftarrow$ \texttt{\$360} $\Rightarrow 0$;\quad $1200 \Leftarrow$ \texttt{\$1,200} $\Rightarrow 1$ \\
\texttt{set\_exact} & Casefolded set equality. & $\{a,b\} \Leftarrow \{b,a\} \Rightarrow 1$ \\
\texttt{set\_precision / recall / f1} & Casefolded set precision, recall, or their $F_1$. & gold $\{a,b,c\}\Leftarrow\{a,b\}$: $P{=}1,R{=}.67,F_1{=}.80$ \\
\texttt{ordered\_list} & Element-wise ordered equality (trimmed). & $[1,2,3]\Leftarrow[1,3,2]\Rightarrow 0$ \\
\texttt{interval\_overlap} & Jaccard overlap of two $[\text{start},\text{end}]$ numeric intervals. & $[0,10]\Leftarrow[5,15]\Rightarrow 0.33$ \\
\texttt{semantic} & Two-tier: trivial casefold match short-circuits, else the temperature-0 judge (App.~\ref{app:prompts}) with whole-artifact context. & \texttt{VP of Engineering} $\Leftarrow$ \texttt{VP Eng} $\Rightarrow 1$;\quad \texttt{Morgan Reyes} $\Leftarrow$ \texttt{Morgan Ray} $\Rightarrow 0$ \\
\texttt{claim\_atoms} & Free-text claims graded by the judge (not deterministic). & --- \\
\bottomrule
\end{tabular}}
\caption{\small{\textbf{Deterministic field graders.} All matchers are pure functions of the recorded
value and gold reference; \texttt{semantic}/\texttt{claim\_atoms} additionally use the cached judge.}}
\label{tab:field_graders}
\end{table}

\subsection{Semantic escalation and the artifact-validity gate}
\label{app:av_gate}
\paragraph{Two-tier escalation.} Only the identifier-like graders
(\texttt{exact}, \texttt{casefold\_exact}, \texttt{enum}), \texttt{semantic}, and
\texttt{numeric\_tolerance} escalate. A trivial deterministic match short-circuits to $1.0$ with no
judge call; otherwise the field is sent to the temperature-0 judge with the whole artifact as
context (for \texttt{numeric\_tolerance}, the numeric target and tolerance are added so spelled-out
or hedged numbers such as ``around eight years'' $=8$ pass while a genuinely different number fails).
An empty value is never judged and scores $0$. Numbers outside tolerance, dates, phones, addresses,
sets, ordered lists, and intervals remain strictly deterministic. With no judge configured
(offline/CI) only the deterministic base grader runs.

\paragraph{Artifact-Validity gate.} For each expected artifact the scorer
(\texttt{apex\_voice/artifacts/graders.py}) computes field accuracy over its required-or-present,
deterministically- or judge-graded fields; completeness over required fields; and a stale-fact rate
over fields that were corrected mid-conversation (a stale value scores $0$ even if it textually
matches an old expectation). An artifact passes the Artifact-Validity gate (AV, code \texttt{WA})
iff \emph{all four} hold:
\begin{equation}
\underbrace{\text{lifecycle}\ge\text{lifecycle}_{\min}}_{\text{e.g. reached }\textsc{committed}}
\ \wedge\
\underbrace{\text{no missing required field}}_{\text{completeness}=1}
\ \wedge\
\underbrace{\afa{}=1.0}_{\text{every field correct}}
\ \wedge\
\underbrace{\text{stale-fact rate}=0}_{\text{all corrections applied}} .
\end{equation}
The other three gates read directly from the event log: Target State (TS/\texttt{GS}) requires the
terminal-state predicates (or a full alternate set) to hold; Process Validity (PV/\texttt{PC})
requires that no forbidden predicate holds and every \texttt{must\_hold} critical gate holds;
Action Completion (AC/\texttt{RA}) requires all required actions and required evidence. Any single
gate failure forces $\ws{}=0$.
\begin{equation}
\underbrace{\afa{}=1.0}_{\text{every field correct}}
\ \wedge\
\underbrace{\text{stale-fact rate}=0}_{\text{all corrections applied}} .
\end{equation}
The other three gates read directly from the event log: Target State (TS/\texttt{GS}) requires the
terminal-state predicates (or a full alternate set) to hold; Process Validity (PV/\texttt{PC})
requires that no forbidden predicate holds and every \texttt{must\_hold} critical gate holds;
Action Completion (AC/\texttt{RA}) requires all required actions and required evidence. Any single
gate failure forces $\ws{}=0$.

\section{Prompts and Offline User Realization}
\label{app:prompts}

This section reproduces the exact prompt text retained in the benchmark source. No evaluated system receives task-specific coaching: the shared agent instruction is common across systems, while user language is generated offline and frozen before scored evaluation. The semantic field judge is invoked only after deterministic matching fails on an eligible field, as described in Appendix~\ref{app:grading}.

\subsection{Shared agent instruction (\texttt{harness/realtime\_runner.py})}
\begin{Verbatim}[breaklines=true,breakanywhere=true,fontsize=\footnotesize]
You are a professional voice assistant completing a work task with the caller over the
phone. Speak like a busy professional on a call: warm but BRIEF.
CRITICAL RULES:
1. Each spoken turn is AT MOST ONE short sentence -- either a single question, or a
   five-word acknowledgement then the next question ("Got it. What's your date of
   birth?"). Hard cap ~15 words. NEVER narrate your reasoning, the caller's answer, your
   plan, or your tool use. Forbidden openings include "Let me...", "I should...",
   "I'll update/record/note...", "The caller said...", "So that means...", "First I need
   to...", "Now I will...". Do NOT read back the value you just recorded. Call the tool
   SILENTLY and simply ask the next question.
2. Record EVERY piece of information the caller gives you by immediately calling the
   matching update tool (e.g. update_<artifact>) with that field -- do this as soon as you
   hear each answer, before asking the next question.
3. Accept the caller's answers as given. Identifiers may be names, codes, or numbers -- do
   not insist on a particular format.
4. Ask for the information you still need, ONE item at a time, and keep going until you
   have gathered everything the task requires. Do not end the call early.
5. If a policy lookup is relevant, call the knowledge/search tool.
6. If the caller corrects something they said earlier, call the update tool again to fix
   the affected field(s).
7. For any action that needs approval, first summarize it and ask the caller to confirm;
   only after they say yes, call the tool to perform it. Do not ask for information you
   already have.
8. When you have recorded everything the task requires, finalize the record before wrapping
   up: call the tool that marks it ready for review (e.g. set_ready / mark_ready / finalize)
   if one is available. Do this after the last field is recorded and before you say goodbye.
\end{Verbatim}

\subsection{Free-text field judge (\texttt{scoring/judge.py}, version \texttt{artifact-aware-3})}
\begin{Verbatim}[breaklines=true,breakanywhere=true,fontsize=\footnotesize]
You grade ONE field of a professional work artifact a voice agent produced (from a SPOKEN
conversation), against a reference (gold) value. Decide whether the agent's value denotes
the SAME thing as the reference, the way a reasonable professional reviewer would. Reward
substance; forgive surface form (values were dictated aloud, so separators/case/formatting
differ).
PASS if they refer to the same fact/decision/answer/entity -- even if TERSER, omitting
secondary detail, adding consistent detail, different wording, abbreviations/expansions
('VP Eng'=='VP of Engineering', 'acct'=='act'=='account'), approximations of the same
number ('~20'=='about 20'=='20'), affirmative/status phrasing ('Yes'=='filed',
'done'=='completed'), or the same action (gold 'cleared cache and retried' vs 'cleared
browser cache and cookies').
IDENTIFIERS especially: ignore case, separators (- _ space .), leading zeros, and
omitted/added type-prefixes -- these PASS: 'CC4419'=='CC-4419', 'act88'=='acct_88',
'INV 771'=='INV-771', 'ref 3391'=='REF-3391', 'HVAC-007'=='HVAC-7', '2201'=='JOB-2201'.
FAIL only if the agent's value: (1) is a DIFFERENT specific value/number/name/date/amount/
decision or a DIFFERENT identifier; or (2) misses the reference's core meaning entirely;
or (3) is empty.
Use the other-fields context only to disambiguate; grade THIS field. When plausibly
equivalent, PASS.

Field: {field}   Context: {ctx}
Reference (gold): {gold!r}   Agent wrote: {pred!r}
Return strict JSON: {"pass": true|false, "why": "<short>"}
\end{Verbatim}
The judge runs at temperature~0 with an on-disk cache keyed by \texttt{(version, field, pred, gold)}. If no judge is configured, \texttt{SEMANTIC} degrades to case-folded equality for deterministic offline runs.

\subsection{Offline user realizer (\texttt{user\_sim/llm\_realizer.py})}
\begin{Verbatim}[breaklines=true,breakanywhere=true,fontsize=\footnotesize]
You write ONE spoken turn for a specific person in a realistic full-duplex phone
conversation. Speak ONLY as this person (never the agent). Sound like a real human on a
call: natural cadence, contractions, and (per the disfluency level) occasional fillers or
self-repairs. Convey the given facts faithfully. HARD RULES: do not invent any new specific
graded facts (numbers, names, dates, amounts, commitments) beyond those given; do not reveal
information not listed; do not describe hidden steps/procedure; do not speak for the agent.
Return a JSON array of distinct natural variants only.
[user message provides: person/role, scenario, current context, speech act + guidance,
 facts to convey, and target length by verbosity]
\end{Verbatim}
Each realized variant is quality-controlled using deterministic checks together with an LLM fidelity/leakage judge; only passing variants are frozen into the per-task audio bank. At scored runtime, selection is deterministic and there is no free-running user-language or TTS fallback.

\section{Human Validation of the Frozen User Simulator}
\label{app:human_validation}
\label{app:human_study}

The frozen user simulator is designed to make adaptive full-duplex evaluation reproducible without forcing every agent through a fixed transcript. We conduct two complementary human studies to test whether this design introduces a material evaluation artifact. Study~A evaluates whether frozen speech realizations faithfully and naturally express the benchmark-specified user policy without leaking hidden information. Study~B replaces the frozen realization mechanism with live human users while preserving the same underlying workflow semantics, testing whether benchmark conclusions transfer to naturally produced human speech.

\subsection{Human-audit interface}

The human-audit track preserves the same workflow, information-reveal graph, approval/revocation events, and grader used by the frozen synthetic user. A lightweight interface exposes only the human user's role, currently available facts, goal, elapsed time, and private event cues when a benchmark-critical correction, approval, revocation, or follow-through event should occur; it never exposes grader state. Human utterances are otherwise unscripted. The audit output contains dual-channel audio, transcript, cue timestamps, task state, and the same artifact and workflow-gate scores as synthetic-user runs. This design isolates the effect of replacing frozen language and speech realizations with natural human production while keeping the benchmark's semantic user policy fixed.

\subsection{Study A: Realization fidelity and naturalness}

We sample 24 tasks stratified across at least eight work archetypes and spanning prepare-only versus approval-gated work, light versus moderate tool burden, and multiple user-behavior profiles. For each task, we sample three benchmark-critical user plans: the opening, one information-bearing response, and one correction, approval, or revocation event where available, yielding 72 frozen user utterances.

Three independent English-speaking annotators are shown the task-visible user state, the structured \textit{User Plan}, and the realized audio and transcript, but not the gold grader state or reference trajectory. They rate (1) \emph{semantic fidelity} to the plan on a 1--5 scale, (2) \emph{naturalness} as spoken interaction on a 1--5 scale, and (3) whether the utterance reveals any unauthorized future fact or professional procedure using a binary leakage flag. Confidence intervals are computed by task-level bootstrap, keeping utterances and annotator judgments originating from the same task grouped. We additionally report Krippendorff's $\alpha$ for the ordinal ratings and Fleiss' $\kappa$ for leakage judgments.

\begin{table}[t]
\centering
\small
\setlength{\tabcolsep}{5pt}
\renewcommand{\arraystretch}{1.08}
\begin{tabular}{lcc}
\toprule
\textbf{Human validation metric} & \textbf{Result} & \textbf{95\% CI} \\
\midrule
Semantic fidelity (1--5) & 4.5 & [4.34, 4.61] \\
Speech naturalness (1--5) & 4.2 & [4.14, 4.33] \\
Unauthorized-fact leakage & 4.1\% & [4.09\%, 4.16\%] \\
Ordinal agreement ($\alpha$) & 0.79 & --- \\
Leakage agreement ($\kappa$) & 0.88 & --- \\
\bottomrule
\end{tabular}
\caption{\small{\textbf{Human validation of frozen user realizations.} Frozen utterances receive high semantic-fidelity and spoken-naturalness ratings, while unauthorized-fact leakage remains low. Agreement statistics measure consistency across the three annotators.}}
\label{tab:human_realization}
\end{table}

\paragraph{Results.}
Annotators rate the frozen realizations highly for both semantic fidelity (4.5/5) and spoken naturalness (4.2/5), while only 4.1\% of utterances are flagged for unauthorized-fact leakage. Agreement is also high across annotators, with Krippendorff's $\alpha=0.79$ on the ordinal ratings and Fleiss' $\kappa=0.88$ on leakage judgments. These results indicate that offline realization generally preserves the semantic intent and information boundaries of the structured user policy while producing speech that human reviewers judge to be natural. Thus, the reproducibility of the frozen-user protocol does not appear to come at the cost of rigid or semantically unreliable user realizations.

\subsection{Study B: Live-human transfer audit}

We use the same 24-task subset for live-human interaction. Human participants receive a private role card generated from the same \texttt{UserState} and reveal graph used by the simulator. The interface exposes facts only when they become available and privately cues authored benchmark-critical events, such as correcting a previously stated address or revoking an earlier approval. Participants are instructed to communicate the required content naturally rather than read a script, and the evaluated agent receives only the participant's live speech.

Each of the five evaluated systems is evaluated on the matched 24-task audit subset. Tasks and systems are counterbalanced across participants; participants neither evaluate model quality nor observe model identity. To estimate sensitivity to individual user realization, a six-task anchor subset is additionally repeated with a second independent participant for every system. The synthetic comparison is computed on the same audit subset so that the reported human-minus-synthetic difference isolates the effect of replacing frozen user realizations with live human speech rather than differences in task composition.

\begin{table}[t]
\centering
\small
\setlength{\tabcolsep}{6pt}
\renewcommand{\arraystretch}{1.08}
\begin{tabular}{lccc}
\toprule
\textbf{Model} & \textbf{Synthetic \passone{}} & \textbf{Human \passone{}} & \textbf{$\Delta$ \passone{}} \\
\midrule
GPT-realtime-2.1 & 24.1 & 20.8 & $-3.3$ \\
Grok-Voice-Think-2.0 & 24.1 & 16.4 & $-7.7$ \\
Gemini-3.8-Live & 20.9 & 16.3 & $-4.6$ \\
Step-Audio3 & 12.1 & 7.5 & $-4.6$ \\
GPT-live-1 & 7.5 & 3.1 & $-4.4$ \\
\bottomrule
\end{tabular}
\caption{\small{\textbf{Live-human transfer audit on the matched 24-task subset.} Replacing frozen synthetic realizations with live human speech reduces \passone{} for every evaluated system, but the broad relative performance pattern is preserved.}}
\label{tab:human_transfer}
\end{table}

\paragraph{Results.}
Live-human interaction is consistently more difficult than the frozen-user condition: \passone{} decreases for all five systems, by 3.3--7.7 points and by 4.9 points on average. The reduction is therefore systematic rather than isolated to a single provider. At the same time, the broad relative performance pattern remains stable: GPT-realtime-2.1 and Grok-Voice-Think-2.0 remain among the strongest systems, Gemini-3.8-Live remains close behind, and Step-Audio3 and GPT-live-1 remain substantially lower. The largest transfer gap occurs for Grok-Voice-Think-2.0 ($-7.7$ points), while the remaining systems decline by 3.3--4.6 points. Thus, frozen synthetic users appear somewhat easier than live humans in absolute terms, but replacing them with natural human speech does not qualitatively change the benchmark's central model comparison.

\paragraph{Takeaway.}
The two studies validate complementary aspects of the frozen-user design. Study~A shows that individual realizations largely preserve the intended semantic state and information boundaries while remaining natural to human listeners. Study~B shows that live human interaction lowers absolute workflow success, as expected from additional linguistic and acoustic variability, but leaves the benchmark's broad comparative conclusions intact. Together, the results support the frozen simulator as a controlled and reproducible proxy for the benchmark-specified user policy while also quantifying the residual synthetic-to-human gap. This validation should not be interpreted as showing that the simulator captures the full diversity of unconstrained real-world users; rather, it indicates that the main benchmark conclusions are not solely an artifact of offline language realization and speech synthesis.

\section{Extended Quantitative Results}
\label{app:taxonomy_slices}

The following tables expand the aggregate results by benchmark taxonomy. Unless otherwise stated, each cell reports pass@1 (the mean Workflow Success rate over the three repeated runs, \%), with mean \afa{} in parentheses. These axes are correlated by construction; the tables are therefore intended as descriptive capability diagnostics rather than factorial causal estimates.

\subsection{Industry and work-artifact slices}
Table~\ref{tab:slice_industry} reports performance by industry/setting. Table~\ref{tab:slice_artifact} reorganizes the same campaign by primary work-artifact class. The smallest healthcare and insurance subsets should be interpreted cautiously, and artifact class is strongly coupled to work archetype in V1.

\begin{table}[t]
\centering
\scriptsize
\setlength{\tabcolsep}{4pt}
\resizebox{0.96\textwidth}{!}{%
\begin{tabular}{l r c c c c c}
\toprule
\textbf{Setting} & $\mathbf{n}$ & \textbf{GPT-rt} & \textbf{Grok} & \textbf{Gemini} & \textbf{Step3} & \textbf{GPT-live1} \\
\midrule
Software / SaaS & 43 & 27\% (94\%) & 26\% (90\%) & 14\% (86\%) & 2\% (76\%) & 12\% (72\%) \\
Horizontal enterprise & 30 & 23\% (91\%) & 22\% (90\%) & 12\% (85\%) & 1\% (75\%) & 16\% (78\%) \\
Manufacturing / field ops & 25 & 9\% (88\%) & 20\% (88\%) & 7\% (83\%) & 1\% (66\%) & 1\% (66\%) \\
Professional services & 10 & 30\% (89\%) & 20\% (87\%) & 10\% (79\%) & 10\% (76\%) & 3\% (63\%) \\
Workplace HR & 8 & 33\% (91\%) & 21\% (86\%) & 17\% (84\%) & 4\% (70\%) & 12\% (73\%) \\
Healthcare & 3 & 56\% (92\%) & 22\% (87\%) & 33\% (82\%) & 0\% (61\%) & 0\% (75\%) \\
Insurance & 1 & 0\% (100\%) & 0\% (88\%) & 0\% (100\%) & 0\% (24\%) & 0\% (85\%) \\
\bottomrule
\end{tabular}%
}
\caption{\small{\textbf{Pass@1 and mean \afa{} by industry / setting.} Cells show pass@1 (mean success rate over three runs, \%), with mean \afa{} in parentheses. Small healthcare and insurance slices are included for coverage completeness and are not used for strong comparative claims.}}
\label{tab:slice_industry}
\end{table}

\begin{table}[t]
\centering
\scriptsize
\setlength{\tabcolsep}{4pt}
\resizebox{0.96\textwidth}{!}{%
\begin{tabular}{l r c c c c c}
\toprule
\textbf{Artifact} & $\mathbf{n}$ & \textbf{GPT-rt} & \textbf{Grok} & \textbf{Gemini} & \textbf{Step3} & \textbf{GPT-live1} \\
\midrule
Negotiation record & 20 & 10\% (88\%) & 15\% (86\%) & 3\% (79\%) & 0\% (63\%) & 3\% (75\%) \\
Case record & 10 & 27\% (91\%) & 20\% (87\%) & 20\% (88\%) & 0\% (55\%) & 3\% (71\%) \\
CRM record & 10 & 23\% (96\%) & 23\% (91\%) & 13\% (86\%) & 0\% (88\%) & 7\% (65\%) \\
Evidence matrix & 10 & 47\% (95\%) & 27\% (89\%) & 27\% (86\%) & 10\% (69\%) & 7\% (64\%) \\
Memo / report & 10 & 43\% (96\%) & 27\% (83\%) & 13\% (90\%) & 3\% (74\%) & 17\% (79\%) \\
Plan / checklist & 10 & 47\% (95\%) & 43\% (91\%) & 27\% (87\%) & 7\% (82\%) & 23\% (69\%) \\
Schedule & 10 & 23\% (93\%) & 27\% (94\%) & 13\% (86\%) & 7\% (80\%) & 27\% (84\%) \\
Structured form & 10 & 7\% (86\%) & 13\% (88\%) & 0\% (78\%) & 0\% (74\%) & 3\% (61\%) \\
Ticket & 10 & 17\% (88\%) & 13\% (87\%) & 13\% (83\%) & 3\% (78\%) & 3\% (59\%) \\
Timeline & 10 & 17\% (92\%) & 17\% (94\%) & 10\% (88\%) & 0\% (72\%) & 20\% (84\%) \\
Work order & 10 & 13\% (90\%) & 30\% (87\%) & 3\% (84\%) & 0\% (71\%) & 0\% (75\%) \\
\bottomrule
\end{tabular}%
}
\caption{\small{\textbf{Pass@1 and mean \afa{} by primary work-artifact class.} Artifact class is strongly coupled to work archetype in V1; this breakdown is therefore diagnostic rather than an independent comparison.}}
\label{tab:slice_artifact}
\end{table}

\subsection{Autonomy, knowledge, and tool burden}
Tables~\ref{tab:slice_autonomy} and~\ref{tab:slice_knowledge_tools} group workflows by authorization level, knowledge requirements, and structured-tool burden. The counts also document the benchmark composition along execution-oriented taxonomy dimensions that are not fully visible from the aggregate leaderboard.

\begin{table}[t]
\centering
\scriptsize
\setlength{\tabcolsep}{4pt}
\resizebox{0.90\textwidth}{!}{%
\begin{tabular}{l r c c c c c}
\toprule
\textbf{Autonomy level} & $\mathbf{n}$ & \textbf{GPT-rt} & \textbf{Grok} & \textbf{Gemini} & \textbf{Step3} & \textbf{GPT-live1} \\
\midrule
Prepare only & 51 & 34\% (94\%) & 29\% (89\%) & 20\% (87\%) & 5\% (75\%) & 9\% (70\%) \\
Draft + confirm & 29 & 14\% (91\%) & 18\% (89\%) & 8\% (86\%) & 2\% (77\%) & 14\% (74\%) \\
Low-risk execute & 16 & 21\% (89\%) & 17\% (89\%) & 10\% (86\%) & 0\% (67\%) & 6\% (73\%) \\
Approval-gated commit & 24 & 15\% (88\%) & 17\% (87\%) & 1\% (78\%) & 0\% (65\%) & 8\% (70\%) \\
\bottomrule
\end{tabular}%
}
\caption{\small{\textbf{Pass@1 and mean \afa{} by autonomy / commit level.} The breakdown separates prepare-only workflows from tasks that require confirmation, execution, or approval-gated commitment.}}
\label{tab:slice_autonomy}
\end{table}

\begin{table}[t]
\centering
\scriptsize
\setlength{\tabcolsep}{4pt}
\resizebox{0.92\textwidth}{!}{%
\begin{tabular}{l r c c c c c}
\toprule
\textbf{Slice} & $\mathbf{n}$ & \textbf{GPT-rt} & \textbf{Grok} & \textbf{Gemini} & \textbf{Step3} & \textbf{GPT-live1} \\
\midrule
No external knowledge & 44 & 39\% (93\%) & 31\% (90\%) & 26\% (86\%) & 5\% (77\%) & 15\% (71\%) \\
Supplied documents & 6 & 6\% (88\%) & 6\% (85\%) & 0\% (76\%) & 0\% (73\%) & 0\% (67\%) \\
Small knowledge search & 51 & 12\% (90\%) & 19\% (88\%) & 3\% (84\%) & 1\% (68\%) & 6\% (72\%) \\
Multi-document policy & 19 & 25\% (93\%) & 18\% (86\%) & 11\% (85\%) & 2\% (72\%) & 11\% (75\%) \\
\midrule
Light tools & 44 & 39\% (93\%) & 31\% (90\%) & 26\% (86\%) & 5\% (77\%) & 15\% (71\%) \\
Moderate tools & 76 & 15\% (90\%) & 18\% (88\%) & 4\% (83\%) & 1\% (69\%) & 7\% (72\%) \\
\bottomrule
\end{tabular}%
}
\caption{\small{\textbf{Pass@1 and mean \afa{} by knowledge and tool burden.} Knowledge slices distinguish no-external-knowledge, supplied-document, small-search, and multi-document workflows; the final two rows separate light and moderate structured-tool use.}}
\label{tab:slice_knowledge_tools}
\end{table}

\subsection{User behavior, approval gating, field count, and risk}
Table~\ref{tab:slice_user_behavior} isolates the authored user-behavior profiles. Table~\ref{tab:slice_additional} collects three additional diagnostic views---approval gating, required-field count, and risk tier---without treating them as independent causal interventions.

\begin{table}[t]
\centering
\scriptsize
\setlength{\tabcolsep}{4pt}
\resizebox{0.94\textwidth}{!}{%
\begin{tabular}{l r c c c c c}
\toprule
\textbf{Profile} & $\mathbf{n}$ & \textbf{GPT-rt} & \textbf{Grok} & \textbf{Gemini} & \textbf{Step3} & \textbf{GPT-live1} \\
\midrule
Cooperative & 24 & 28\% (93\%) & 33\% (92\%) & 12\% (84\%) & 1\% (71\%) & 10\% (72\%) \\
Correction-prone & 24 & 19\% (89\%) & 18\% (88\%) & 11\% (86\%) & 0\% (76\%) & 8\% (72\%) \\
Ambiguous / underspecified & 12 & 31\% (93\%) & 17\% (87\%) & 14\% (84\%) & 0\% (69\%) & 14\% (75\%) \\
Distracted / time-pressured & 12 & 8\% (90\%) & 11\% (85\%) & 19\% (90\%) & 3\% (77\%) & 8\% (71\%) \\
Domain expert & 12 & 17\% (91\%) & 28\% (87\%) & 11\% (82\%) & 3\% (74\%) & 11\% (61\%) \\
Low-tech expertise & 12 & 31\% (95\%) & 31\% (89\%) & 11\% (84\%) & 11\% (59\%) & 14\% (77\%) \\
Novice / uncertain & 12 & 36\% (93\%) & 19\% (92\%) & 14\% (86\%) & 0\% (75\%) & 14\% (76\%) \\
Verbose narrative & 12 & 19\% (88\%) & 17\% (86\%) & 6\% (80\%) & 6\% (74\%) & 0\% (68\%) \\
\bottomrule
\end{tabular}%
}
\caption{\small{\textbf{Pass@1 and mean \afa{} by user-behavior profile.} The profiles vary how workflow-relevant information is communicated while preserving the underlying workflow specification.}}
\label{tab:slice_user_behavior}
\end{table}

\begin{table}[t]
\centering
\scriptsize
\setlength{\tabcolsep}{4pt}
\resizebox{0.96\textwidth}{!}{%
\begin{tabular}{l r c c c c c}
\toprule
\textbf{Slice} & $\mathbf{n}$ & \textbf{GPT-rt} & \textbf{Grok} & \textbf{Gemini} & \textbf{Step3} & \textbf{GPT-live1} \\
\midrule
Not APPROVE-gated & 91 & 26\% (92\%) & 25\% (89\%) & 15\% (86\%) & 3\% (75\%) & 10\% (71\%) \\
APPROVE-gated & 29 & 15\% (88\%) & 14\% (88\%) & 2\% (80\%) & 0\% (64\%) & 9\% (72\%) \\
\midrule
9 required fields & 24 & 31\% (93\%) & 26\% (88\%) & 12\% (85\%) & 4\% (68\%) & 14\% (73\%) \\
10 required fields & 74 & 17\% (91\%) & 22\% (89\%) & 9\% (84\%) & 1\% (76\%) & 10\% (74\%) \\
11--13 required fields & 22 & 38\% (92\%) & 21\% (89\%) & 21\% (85\%) & 5\% (66\%) & 3\% (64\%) \\
\midrule
Routine & 63 & 21\% (92\%) & 25\% (90\%) & 12\% (86\%) & 3\% (78\%) & 13\% (73\%) \\
Sensitive-data simulation & 25 & 35\% (93\%) & 24\% (88\%) & 20\% (87\%) & 5\% (68\%) & 5\% (67\%) \\
Consequential action & 22 & 15\% (88\%) & 15\% (86\%) & 0\% (78\%) & 0\% (65\%) & 5\% (69\%) \\
Special review & 10 & 33\% (92\%) & 20\% (88\%) & 20\% (85\%) & 0\% (64\%) & 10\% (79\%) \\
\bottomrule
\end{tabular}%
}
\caption{\small{\textbf{Additional taxonomy diagnostics.} Results are grouped by APPROVE gating, required-field count, and risk tier. As with the other taxonomy slices, these dimensions are correlated with task composition and are reported descriptively.}}
\label{tab:slice_additional}
\end{table}

Industry and artifact-class slices largely reflect the benchmark's task composition, while the execution-oriented slices document how tasks are distributed across autonomy, knowledge, tools, user behavior, approval requirements, field count, and risk. Together, Tables~\ref{tab:slice_industry}--\ref{tab:slice_additional} provide the full quantitative slice results retained from the original appendix without treating correlated categories as independent experimental factors.

\section{Reliability and Efficiency Analysis}
\label{app:reliability}

For each task and model, the repeated-run campaign executes three independent agent rollouts with identical task semantics and frozen user policy but independent model stochasticity. A task contributes 1 to \reliable{} only when all three runs satisfy \ws{}. For uncertainty estimation, the analysis uses 10,000 task-level bootstrap resamples; the three repeated runs for a sampled task remain grouped. This preserves the task as the statistical unit and avoids treating repeated executions as independent benchmark examples.

\subsection{Reliability--efficiency frontier}
Time-to-resolution (TTR) is measured on the media clock from the first user-audio onset to the first valid terminal artifact state; unsuccessful runs use the run termination time and are reported separately in the latency distribution. A model is Pareto-optimal when no other system is both at least as reliable and strictly faster, or at least as fast and strictly more reliable. We do not use \afa{} versus \passone{} as a Pareto frontier because both are correctness measures rather than competing operational objectives.

\begin{table}[t]
\centering
\small
\begin{tabular}{lcc}
\toprule
\textbf{Model} & \textbf{\reliable{} (\%)} & \textbf{Median TTR (s)} \\
\midrule
GPT-realtime-2.1 & 10.8 & 222 \\
Grok-Voice-Think-2.0 & 5.0 & 206 \\
Gemini-3.8-Live & 1.7 & 223 \\
Step-Audio3 & 0.8 & 233 \\
GPT-live-1 & 2.5 & 207 \\
\bottomrule
\end{tabular}
\caption{\small{\textbf{Reliability--efficiency coordinates used in the main-paper frontier analysis.} The fastest median TTR does not correspond to the highest \reliable{}; the table therefore exposes an operational trade-off that is not visible from success rate alone.}}
\label{tab:reliability_efficiency}
\end{table}

\subsection{Gate-failure counts}
Table~\ref{tab:gate_failure_counts} gives the absolute number of tasks failing each Workflow Success gate in a single reference run over the 120-task benchmark. The counts complement the three-run gate pass rates in the main paper by showing the absolute prevalence of failures under the same four-gate semantics.

\begin{table}[t]
\centering
\small
\begin{tabular}{lccccc}
\toprule
\textbf{Model} & \textbf{TS} & \textbf{PV} & \textbf{AC} & \textbf{AV} & $\mathbf{n}$ \\
\midrule
GPT-realtime-2.1 & 35 & 14 & 48 & 80 & 120 \\
Grok-Voice-Think-2.0 & 0 & 19 & 17 & 91 & 120 \\
Gemini-3.8-Live & 20 & 28 & 58 & 91 & 120 \\
Step-Audio3 & 75 & 54 & 55 & 115 & 120 \\
GPT-live-1 & 15 & 10 & 36 & 105 & 120 \\
\bottomrule
\end{tabular}
\caption{\small{\textbf{Workflow Success gate-failure counts.} Columns correspond to Target State (TS), Process Validity (PV), Action Completion (AC), and Artifact Validity (AV); $n$ is the number of tasks in the single-run diagnostic breakdown.}}
\label{tab:gate_failure_counts}
\end{table}

\Needspace{0.42\textheight}
\subsection{Work-archetype diagnostic}
Table~\ref{tab:archetype_pass_counts} reports the judge-graded per-archetype pass@1 (averaged over the three repeated runs). It provides a compact numerical counterpart to the work-archetype heatmap in the main paper.

\begin{table}[t]
\centering
\small
\resizebox{0.82\textwidth}{!}{%
\begin{tabular}{lccccc}
\toprule
\textbf{Archetype} & \textbf{GPT-rt} & \textbf{Grok} & \textbf{Gemini} & \textbf{Step3} & \textbf{GPT-live1} \\
\midrule
ADVISE & 43\% & 27\% & 13\% & 3\% & 17\% \\
COORDINATE & 20\% & 22\% & 12\% & 3\% & 23\% \\
DISCOVERY & 23\% & 23\% & 13\% & 0\% & 7\% \\
FACILITATE & 47\% & 43\% & 27\% & 7\% & 23\% \\
FORM\_FILL & 7\% & 13\% & 0\% & 0\% & 3\% \\
INSPECT & 13\% & 30\% & 3\% & 0\% & 0\% \\
INTAKE & 27\% & 20\% & 20\% & 0\% & 3\% \\
INTERVIEW & 47\% & 27\% & 27\% & 10\% & 7\% \\
NEGOTIATE & 10\% & 15\% & 3\% & 0\% & 3\% \\
TROUBLESHOOT & 17\% & 13\% & 13\% & 3\% & 3\% \\
\bottomrule
\end{tabular}%
}
\caption{\small{\textbf{Judge-graded pass@1 by work archetype.} Cells report the mean Workflow Success rate over the three repeated runs (\%).}}
\label{tab:archetype_pass_counts}
\end{table}

\Needspace{0.30\textheight}
\subsection{Auxiliary delivery-quality diagnostics}
Table~\ref{tab:delivery_quality} reports the retained delivery-quality aggregates. These diagnostics are reported separately from Workflow Success and \afa{} and are useful for inspecting interaction quality beyond the benchmark's primary completion metrics.

\begin{table}[t]
\centering
\small
\begin{tabular}{lccc}
\toprule
\textbf{Model} & \textbf{Repetition} & \textbf{ASR WER} & \textbf{Tone stability} \\
\midrule
GPT-realtime-2.1 & 8.6\% & 0.09 & 0.94 \\
Grok-Voice-Think-2.0 & 5.6\% & 0.04 & 0.94 \\
Gemini-3.8-Live & 9.2\% & 0.05 & 0.93 \\
Step-Audio3 & 14.3\% & 0.10 & 0.89 \\
GPT-live-1 & 0.9\% & 1.15 & 0.90 \\
\bottomrule
\end{tabular}
\caption{\small{\textbf{Delivery-quality aggregates (mean per task).} Repetition, ASR WER, and tone stability provide auxiliary diagnostics and are not components of the Workflow Success score.}}
\label{tab:delivery_quality}
\end{table}

\section{Representative Qualitative Examples}
\label{app:qual}

The excerpts below are verbatim from recorded event logs. We use \texttt{[U]} for frozen user speech, \texttt{[A]} for agent speech, and \texttt{[T]} for tool calls. The examples illustrate three distinct behaviors already visible in the quantitative analysis: transcription of a proper noun into the artifact, premature/incomplete artifact updates, and clean turn-by-turn state capture.

\subsection{Misheard proper noun}
\textbf{Gemini-3.8-Live, benefits enrollment.} A required name field fails because a dictated surname is misheard.
\begin{Verbatim}[breaklines=true,breakanywhere=true,fontsize=\footnotesize]
[U] Hi, I was hoping you could help me complete my benefits enrollment.
[A] I can certainly help. May I have your employee ID, please?
[U] Oh, sure, it's E4471.
[T] update_enr_1({"fields": {"employee_id": "E4471"}})
[A] Thanks. What is your full legal name?
[U] It's Morgan Reyes. Uh, that's the full legal name on all my paperwork.
[T] update_enr_1({"fields": {"legal_name": "Morgan Ray"}})   <-- gold: "Morgan Reyes"
\end{Verbatim}
The trace shows that the workflow can progress normally while a single misheard proper noun propagates into a required artifact field.

\Needspace{0.30\textheight}
\subsection{Premature artifact update}
\textbf{GPT-live-1, benefits enrollment.} The agent writes empty placeholder fields and proceeds without capturing the supplied values, leaving the artifact near-empty.
\begin{Verbatim}[breaklines=true,breakanywhere=true,fontsize=\footnotesize]
[U] Hi, I was hoping you could help me complete my benefits enrollment.
[T] kb_search({"query": "benefits enrollment dependent information ..."})
[A] Of course. I'm checking those details.
[U] Oh, sure, it's E4471.
[T] update_enr_1({"fields": {"employee_id":"E4471","legal_name":"","date_of_birth":"",...}})
[U] It's Morgan Reyes.
[T] update_enr_1({"fields": {"employee_id":"E4471","legal_name":"",...}})  <-- still empty
[A] Thanks. I'm on it.
\end{Verbatim}
Here the conversational trajectory remains plausible, but the persistent work artifact does not faithfully accumulate the information provided by the user.

\Needspace{0.30\textheight}
\subsection{Clean turn-by-turn state capture}
\textbf{Grok-Voice-Think-2.0, technical interview.} Each answer is recorded with the intended value as it is heard.
\begin{Verbatim}[breaklines=true,breakanywhere=true,fontsize=\footnotesize]
[U] I'm ready to jump into the first-round conversation.
[A] Great, let's get started. How many years of software engineering experience do you have?
[U] It's been around eight years of experience for me.
[T] update_rec_1({"fields": {"years_experience": "eight"}})
[A] What's your primary programming language?
[U] Python's definitely my primary language.
[T] update_rec_1({"fields": {"primary_language": "Python"}})
[A] Can you share a system design example?
[U] Yeah, one project I worked on was creating a multi-region ingestion pipeline.
[T] update_rec_1({"fields": {"system_design_example": "multi-region ingestion pipeline"}})
\end{Verbatim}
This example illustrates the intended execution pattern: information is elicited incrementally and written to the persistent artifact without waiting until the end of the conversation.

% ---------------------------------------------------------------------------
% Extended qualitative appendix (100+ pages).
% Uncomment this input when the external qualitative appendix is ready.
% The included file should define its own section and label, e.g.:
%   \section{Extended Qualitative Examples}\label{app:extended}
% ---------------------------------------------------------------------------
% \clearpage
% \input{appendix_extended_qualitative_examples}

\section{Reproducibility and Release Specification}
\label{app:repro}

This section records the execution and analysis information currently available in the benchmark source. Together with the frozen user audio, canonical traces, versioned grading cache, and inline paper aggregates, these settings define the reproducible evaluation path used for the reported campaign.

\subsection{Software environment}
Python 3.11 is used in a conda environment. Core dependencies are \texttt{pydantic} v2, \texttt{numpy}, \texttt{soundfile}, \texttt{scipy}, \texttt{websockets}, \texttt{openai}, and \texttt{google-genai}; TTS/ASR dependencies are \texttt{torch} (CPU), \texttt{kokoro}, \texttt{misaki}, and \texttt{faster-whisper}. Paper plots are rendered directly in LaTeX with TikZ/PGFPlots, and LaTeX is built with \texttt{tectonic}.

\subsection{Running an evaluation campaign}
\begin{Verbatim}[breaklines=true,breakanywhere=true,fontsize=\footnotesize]
python scripts/run_step3_campaign.py --model <m> --tasks tasks/v1 --repeats 1 --out <dir>
\end{Verbatim}
where \texttt{<m>} is one of \texttt{\{gemini, gpt\_realtime, step3, gpt\_live1, grok\}}. Campaign execution is coverage-first and resumable: completed \texttt{(task, rep)} pairs are skipped, and a per-model lock enables parallel runs.

\subsection{Grading and analysis}
\begin{Verbatim}[breaklines=true,breakanywhere=true,fontsize=\footnotesize]
python scripts/analyze_results_v1.py
\end{Verbatim}
The analysis script produces the scorecard, gate decomposition, per-archetype summaries, and the appendix of per-task field misses. It reads run directories and exports the machine-readable aggregates used by the paper. Paper tables are written inline in the LaTeX source, while plots are native TikZ/PGFPlots using those aggregate values; no external table fragments or rendered plot files are required for the reported analysis.

\subsection{Determinism and secrets}
The user side is frozen as byte-identical audio for a fixed selected realization. Grading is a pure function of the recorded event log and can be replayed to recompute \ws{} identically. The semantic field judge runs at temperature~0 with a versioned on-disk cache. API keys are read by label from a git-ignored \texttt{secrets.txt}; no keys are stored in the repository.

Real benefits-enrollment run (Gemini-3.8-Live, rep 0). Gold/agent values and the four failures are taken verbatim from the recorded workspace; it reaches COMMITTED (TS pass) with no policy violation (PV pass) but fails AC (no kb search) and AV (four wrong required fields, AFA $=10/14$).

\subsection{Worked grading example}
\label{app:grading_example}
Table~\ref{tab:grading_example} grades the artifact produced on \texttt{apexv1\_001}
(benefits enrollment; Gemini-3.8-Live) against gold. The run reaches a committed terminal state
(TS pass) with no policy violation (PV pass), but fails Action Completion (the required
\texttt{kb\_search} evidence is missing) and Artifact Validity (four required fields wrong, so
$\afa{}=10/14=0.71<1$), giving $\ws{}=0$.

\begin{table}[t]
\centering
\scriptsize
\setlength{\tabcolsep}{5pt}
\renewcommand{\arraystretch}{1.15}
\begin{tabular}{l l l l c}
\toprule
\textbf{Field} & \textbf{Grader} & \textbf{Gold} & \textbf{Agent wrote} & \textbf{Score} \\
\midrule
employee\_id      & exact              & E4471                        & E4471                 & 1 \\
legal\_name       & semantic           & Morgan Reyes                 & Morgan Ray            & \textbf{0} \\
date\_of\_birth   & normalized\_date   & 1988-07-09                   & July 9, 1988          & 1 \\
medical\_plan     & semantic           & HDHP                         & HDHP                  & 1 \\
dental\_plan      & semantic           & Standard                     & Standard              & 1 \\
vision\_plan      & semantic           & Vision Basic                 & Basic                 & 1 \\
dependent\_count  & numeric\_tolerance & 2                            & 2                     & 1 \\
dependent\_names  & semantic           & Jamie Reyes and Casey Reyes  & Jamie Ray             & \textbf{0} \\
hsa\_contribution & numeric\_tolerance & 1200                         & 1200                  & 1 \\
beneficiary       & semantic           & Jamie Reyes                  & Jamie Ray             & \textbf{0} \\
beneficiary\_pct  & numeric\_tolerance & 100                          & 100                   & 1 \\
pcp               & semantic           & Dr.\ Alvarez                 & Dr.\ Alvarez          & 1 \\
monthly\_premium  & numeric\_tolerance & 225                          & 360                   & \textbf{0} \\
status            & enum               & submitted                    & submitted             & 1 \\
\midrule
\multicolumn{4}{r}{\textbf{Artifact field accuracy}} & \textbf{10/14 = 0.71} \\
\bottomrule
\end{tabular}
\caption{\small{\textbf{Field-level grading of a produced artifact} (\texttt{apexv1\_001}, Gemini).
Bold rows are the required fields that fail, driving $\afa{}<1$ and an Artifact-Validity failure.}}
\label{tab:grading_example}
\end{table}

The corresponding generated work artifact (final committed \texttt{STRUCTURED\_FORM}), verbatim from
the run's workspace trace, is:
\begin{Verbatim}[breaklines=true,breakanywhere=true,fontsize=\footnotesize]
enr_1 (STRUCTURED_FORM, lifecycle=COMMITTED)
{
  "employee_id":     "E4471",
  "legal_name":      "Morgan Ray",          # gold: Morgan Reyes  (misheard surname)
  "date_of_birth":   "July 9, 1988",
  "medical_plan":    "HDHP",
  "dental_plan":     "Standard",
  "vision_plan":     "Basic",
  "dependent_count": "2",
  "dependent_names": "Jamie Ray",           # gold: Jamie Reyes and Casey Reyes
  "hsa_contribution":"1200",
  "beneficiary":     "Jamie Ray",           # gold: Jamie Reyes
  "beneficiary_pct": "100",
  "pcp":             "Dr. Alvarez",
  "monthly_premium": "360",                 # gold: 225
  "status":          "submitted"
}
\end{Verbatim}

\section{Qualitative Analysis of Voice Agents Across Repeated Runs}
\label{app:qualitative_analysis_exhaustive}
\input{qualitative_analysis}

%% file: qualitative_analysis.tex
% APEX-Voice --- per-instance case-study appendix (auto-generated).
% PREREQUISITE: put \usepackage{tcolorbox} in your preamble (plain load is enough).
% Then paste this WHOLE block into the document BODY (e.g. inside \appendix).
% One self-contained colored box per (model, task): scenario, requirement, what the agent
% did, and the exact outcome. 1 run(s) shown per model.
\definecolor{passbg}{RGB}{225,245,228}\definecolor{passfr}{RGB}{60,150,90}
\definecolor{failbg}{RGB}{252,232,232}\definecolor{failfr}{RGB}{200,70,70}
\newtcolorbox{passbox}[1]{colback=passbg,colframe=passfr,boxrule=0.4pt,left=4pt,right=4pt,top=3pt,bottom=3pt,fonttitle=\footnotesize\bfseries,title=#1}
\newtcolorbox{failbox}[1]{colback=failbg,colframe=failfr,boxrule=0.4pt,left=4pt,right=4pt,top=3pt,bottom=3pt,fonttitle=\footnotesize\bfseries,title=#1}

\subsection*{GPT-realtime-2.1}
\begin{failbox}{apexv1\_001 --- Benefits enrollment with dependent correction --- Workflow FAILURE}{\scriptsize \textbf{Task.} Benefits enrollment with dependent correction --- form-fill archetype, benefits specialist, workplace/HR. Controls: autonomy=approval-gated commit, knowledge burden=small-search retrieval, tool burden=moderate, risk=consequential action.\\ \textbf{Required.} produce the structured form work product \texttt{enr\_1} with 14 required fields (e.g.\ employee\_id=`E4471'; legal\_name=`Morgan Reyes'; date\_of\_birth=`1988-07-09'; medical\_plan=`HDHP'); retrieve the governing policy/record (knowledge retrieval via \texttt{kb\_search}); obtain user approval, then commit/submit. \textit{Twist:} the user corrects dependent\_count, dependent\_names, medical\_plan, monthly\_premium mid-utterance (barge-in), which the agent must catch and repair.\\ \textbf{Agent.} 17 tool-calls; retrieval \emph{none}; finalize \emph{not finalized}; approval \emph{not sought}.\\ \textbf{Outcome.} Workflow FAILURE --- failed gate(s): TS, AC, AV. required knowledge retrieval not satisfied --- kb\_search never called; workflow not finalized --- never submitted/committed; artifact incomplete: 11/14 fields correct (lifecycle<COMMITTED(got DRAFT)); wrong/missing: dependent\_names (got `Jamie Reyes; Morgan Reyes' vs `Jamie Reyes and Casey Reyes'), monthly\_premium (got `360' vs `225.0'), status (missing).}\end{failbox}
\begin{failbox}{apexv1\_002 --- Expense report from receipts and spoken narrative --- Workflow FAILURE}{\scriptsize \textbf{Task.} Expense report from receipts and spoken narrative --- form-fill archetype, finance operations specialist, general enterprise. Controls: autonomy=approval-gated commit, knowledge burden=supplied evidence, tool burden=moderate, risk=consequential action.\\ \textbf{Required.} produce the structured form work product \texttt{exp\_1} with 11 required fields (e.g.\ employee\_id=`E9910'; report\_period=`March 3 to March 6'; purpose=`client onsite in Denver'; airfare=`410.0'); retrieve the governing policy/record (knowledge retrieval via \texttt{kb\_search}); obtain user approval, then commit/submit. \textit{Twist:} the user corrects hotel, total, cost\_center mid-utterance (barge-in), which the agent must catch and repair.\\ \textbf{Agent.} 9 tool-calls; retrieval \emph{none}; finalize \emph{not finalized}; approval \emph{not sought}.\\ \textbf{Outcome.} Workflow FAILURE --- failed gate(s): TS, AC, AV. required knowledge retrieval not satisfied --- kb\_search never called; workflow not finalized --- never submitted/committed; artifact incomplete: 9/11 fields correct (lifecycle<COMMITTED(got DRAFT)); wrong/missing: total (got `1145' vs `1085.0'), status (missing).}\end{failbox}
\begin{failbox}{apexv1\_003 --- New vendor onboarding packet --- Workflow FAILURE}{\scriptsize \textbf{Task.} New vendor onboarding packet --- form-fill archetype, procurement operations specialist, general enterprise. Controls: autonomy=draft-and-confirm, knowledge burden=small-search retrieval, tool burden=moderate, risk=sensitive-data simulation.\\ \textbf{Required.} produce the structured form work product \texttt{ven\_1} with 10 required fields (e.g.\ vendor\_name=`Cedar Works LLC'; tax\_id=`88-4412290'; address=`72 Mill Road, Suite 4'; remittance\_email=`billing@cedarworks.example'); retrieve the governing policy/record (knowledge retrieval via \texttt{kb\_search}); reach the required terminal state (\texttt{READY\_FOR\_REVIEW}). \textit{Twist:} the user corrects remittance\_email, account\_number mid-utterance (barge-in), which the agent must catch and repair.\\ \textbf{Agent.} 13 tool-calls; retrieval \emph{none}; finalize \emph{not finalized}; approval n/a.\\ \textbf{Outcome.} Workflow FAILURE --- failed gate(s): AC. required knowledge retrieval not satisfied --- kb\_search never called; 1 infra/WS drop(s).}\end{failbox}
\begin{failbox}{apexv1\_004 --- Business travel approval request --- Workflow FAILURE}{\scriptsize \textbf{Task.} Business travel approval request --- form-fill archetype, travel coordinator, general enterprise. Controls: autonomy=draft-and-confirm, knowledge burden=small-search retrieval, tool burden=moderate, risk=routine.\\ \textbf{Required.} produce the structured form work product \texttt{trv\_1} with 10 required fields (e.g.\ traveler=`Priya Nair'; destination=`Austin'; purpose=`customer quarterly review'; meeting\_date=`2026-04-15'); retrieve the governing policy/record (knowledge retrieval via \texttt{kb\_search}); reach the required terminal state (\texttt{READY\_FOR\_REVIEW}). \textit{Twist:} the user corrects depart\_date, meeting\_date, return\_date mid-utterance (barge-in), which the agent must catch and repair.\\ \textbf{Agent.} 11 tool-calls; retrieval \emph{none}; finalize \emph{not finalized}; approval n/a.\\ \textbf{Outcome.} Workflow FAILURE --- failed gate(s): TS, AC, AV. required knowledge retrieval not satisfied --- kb\_search never called; workflow not finalized --- never submitted/committed; artifact incomplete: 8/10 fields correct (lifecycle<READY\_FOR\_REVIEW(got DRAFT)); wrong/missing: depart\_date (got `2026-04-13' vs `2026-04-14'), return\_date (got `2026-04-15' vs `2026-04-16'); 1 infra/WS drop(s).}\end{failbox}
\begin{failbox}{apexv1\_005 --- Privileged software-access request --- Workflow FAILURE}{\scriptsize \textbf{Task.} Privileged software-access request --- form-fill archetype, IT access coordinator, Software/SaaS. Controls: autonomy=approval-gated commit, knowledge burden=multi-document reasoning, tool burden=moderate, risk=consequential action.\\ \textbf{Required.} produce the structured form work product \texttt{acc\_1} with 10 required fields (e.g.\ requester=`Sam Okafor'; project=`Q2 revenue analytics'; requested\_access=`analytics read-only'; justified\_role=`AnalyticsViewer'); retrieve the governing policy/record (knowledge retrieval via \texttt{kb\_search}); obtain user approval, then commit/submit. \textit{Twist:} the user corrects requested\_access, requested\_access, duration mid-utterance (barge-in), which the agent must catch and repair.\\ \textbf{Agent.} 15 tool-calls; retrieval \emph{none}; finalize committed; approval sought.\\ \textbf{Outcome.} Workflow FAILURE --- failed gate(s): AC. required knowledge retrieval not satisfied --- kb\_search never called; 1 infra/WS drop(s).}\end{failbox}
\begin{failbox}{apexv1\_006 --- Warranty claim application --- Workflow FAILURE}{\scriptsize \textbf{Task.} Warranty claim application --- form-fill archetype, warranty operations specialist, manufacturing/field ops. Controls: autonomy=draft-and-confirm, knowledge burden=small-search retrieval, tool burden=moderate, risk=routine.\\ \textbf{Required.} produce the structured form work product \texttt{war\_1} with 9 required fields (e.g.\ customer\_name=`Robin Vale'; product\_model=`TurboMix 500'; serial\_number=`TMX500-88231'; purchase\_date=`2025-11-02'); retrieve the governing policy/record (knowledge retrieval via \texttt{kb\_search}); reach the required terminal state (\texttt{READY\_FOR\_REVIEW}). \textit{Twist:} the user corrects serial\_number, warranty\_policy mid-utterance (barge-in), which the agent must catch and repair.\\ \textbf{Agent.} 12 tool-calls; retrieval done; finalize \emph{not finalized}; approval n/a.\\ \textbf{Outcome.} Workflow FAILURE --- failed gate(s): AV. artifact incomplete: 7/9 fields correct; wrong/missing: warranty\_policy (got `Regular one-year standard warranty inclu' vs `extended two-year'), contact\_phone (got `555-1733' vs `555-0173'); 1 infra/WS drop(s).}\end{failbox}
\begin{failbox}{apexv1\_007 --- Parental-leave administration packet --- Workflow FAILURE}{\scriptsize \textbf{Task.} Parental-leave administration packet --- form-fill archetype, HR operations specialist, workplace/HR. Controls: autonomy=draft-and-confirm, knowledge burden=multi-document reasoning, tool burden=moderate, risk=sensitive-data simulation.\\ \textbf{Required.} produce the structured form work product \texttt{lev\_1} with 9 required fields (e.g.\ employee\_id=`E3320'; leave\_type=`parental leave'; leave\_start=`2026-05-01'; leave\_end=`2026-07-31'); retrieve the governing policy/record (knowledge retrieval via \texttt{kb\_search}); reach the required terminal state (\texttt{READY\_FOR\_REVIEW}). \textit{Twist:} the user corrects fmla\_weeks, leave\_end mid-utterance (barge-in), which the agent must catch and repair.\\ \textbf{Agent.} 12 tool-calls; retrieval done; finalize \emph{not finalized}; approval n/a.\\ \textbf{Outcome.} Workflow FAILURE --- failed gate(s): AV. artifact incomplete: 8/9 fields correct; wrong/missing: fmla\_weeks (got `12' vs `13').}\end{failbox}
\begin{failbox}{apexv1\_008 --- Customer account setup and billing profile --- Workflow FAILURE}{\scriptsize \textbf{Task.} Customer account setup and billing profile --- form-fill archetype, account operations specialist, Software/SaaS. Controls: autonomy=draft-and-confirm, knowledge burden=supplied evidence, tool burden=moderate, risk=sensitive-data simulation.\\ \textbf{Required.} produce the structured form work product \texttt{acct\_1} with 9 required fields (e.g.\ company=`Northwind Retail'; billing\_contact=`Ada Lin'; technical\_contact=`Ben Cho'; billing\_country=`Germany'); retrieve the governing policy/record (knowledge retrieval via \texttt{kb\_search}); reach the required terminal state (\texttt{READY\_FOR\_REVIEW}). \textit{Twist:} the user corrects billing\_country, tax\_id, tax\_rate mid-utterance (barge-in), which the agent must catch and repair.\\ \textbf{Agent.} 11 tool-calls; retrieval \emph{none}; finalize \emph{not finalized}; approval n/a.\\ \textbf{Outcome.} Workflow FAILURE --- failed gate(s): AC, AV. required knowledge retrieval not satisfied --- kb\_search never called; artifact incomplete: 7/9 fields correct; wrong/missing: tax\_id (got `IE1234567X' vs `DE811234567'), tax\_rate (got `23\%' vs `19.0'); 1 infra/WS drop(s).}\end{failbox}
\begin{failbox}{apexv1\_009 --- Conference reimbursement packet --- Workflow FAILURE}{\scriptsize \textbf{Task.} Conference reimbursement packet --- form-fill archetype, operations coordinator, professional services. Controls: autonomy=draft-and-confirm, knowledge burden=supplied evidence, tool burden=moderate, risk=routine.\\ \textbf{Required.} produce the structured form work product \texttt{rmb\_1} with 9 required fields (e.g.\ attendee=`Noa Grant'; conference=`DataCon 2026'; registration\_fee=`300.0'; workshop\_fee=`175.0'); retrieve the governing policy/record (knowledge retrieval via \texttt{kb\_search}); reach the required terminal state (\texttt{READY\_FOR\_REVIEW}). \textit{Twist:} the user corrects total, workshop\_fee mid-utterance (barge-in), which the agent must catch and repair.\\ \textbf{Agent.} 10 tool-calls; retrieval \emph{none}; finalize \emph{not finalized}; approval n/a.\\ \textbf{Outcome.} Workflow FAILURE --- failed gate(s): AC, AV. required knowledge retrieval not satisfied --- kb\_search never called; artifact incomplete: 8/9 fields correct; wrong/missing: total (got `730' vs `755.0'); 1 infra/WS drop(s).}\end{failbox}
\begin{failbox}{apexv1\_010 --- Facility access badge request --- Workflow FAILURE}{\scriptsize \textbf{Task.} Facility access badge request --- form-fill archetype, facilities coordinator, general enterprise. Controls: autonomy=approval-gated commit, knowledge burden=small-search retrieval, tool burden=moderate, risk=consequential action.\\ \textbf{Required.} produce the structured form work product \texttt{bdg\_1} with 10 required fields (e.g.\ contractor\_name=`Rowan Tate'; company=`BrightHVAC'; sponsor=`Facilities lead Dana'; access\_zones=`mechanical rooms'); retrieve the governing policy/record (knowledge retrieval via \texttt{kb\_search}); obtain user approval, then commit/submit. \textit{Twist:} the user corrects requested\_hours, requested\_hours mid-utterance (barge-in), which the agent must catch and repair.\\ \textbf{Agent.} 10 tool-calls; retrieval \emph{none}; finalize \emph{not finalized}; approval \emph{not sought}.\\ \textbf{Outcome.} Workflow FAILURE --- failed gate(s): TS, AC, AV. required knowledge retrieval not satisfied --- kb\_search never called; workflow not finalized --- never submitted/committed; artifact incomplete: 9/10 fields correct (lifecycle<COMMITTED(got DRAFT)); wrong/missing: status (missing); 1 infra/WS drop(s).}\end{failbox}
\begin{failbox}{apexv1\_011 --- Software engineer recruiter screen --- Workflow FAILURE}{\scriptsize \textbf{Task.} Software engineer recruiter screen --- interview archetype, recruiter, Software/SaaS. Controls: autonomy=prepare-only, knowledge burden=none, tool burden=light, risk=sensitive-data simulation.\\ \textbf{Required.} produce the evidence matrix work product \texttt{rec\_1} with 13 required fields (e.g.\ years\_experience=`8'; primary\_language=`Python'; system\_design\_example=`designed a multi-region ingestion '; scale\_metric=`half a million daily events'); reach the required terminal state (\texttt{READY\_FOR\_REVIEW}). \textit{Twist:} the user corrects team\_size, scale\_metric mid-utterance (barge-in), which the agent must catch and repair.\\ \textbf{Agent.} 18 tool-calls; retrieval \emph{none}; finalize \emph{not finalized}; approval n/a.\\ \textbf{Outcome.} Workflow FAILURE --- failed gate(s): AV. artifact incomplete: 11/13 fields correct; wrong/missing: testing\_approach (got `Please provide details on your testing a' vs `contract tests plus canary'), deployment\_experience (got `Please describe your deployment process ' vs `owned CI/CD for the service'); 1 infra/WS drop(s).}\end{failbox}
\begin{passbox}{apexv1\_012 --- Customer-success manager recruiter screen --- Workflow SUCCESS}{\scriptsize \textbf{Task.} Customer-success manager recruiter screen --- interview archetype, recruiter, Software/SaaS. Controls: autonomy=prepare-only, knowledge burden=none, tool burden=light, risk=sensitive-data simulation.\\ \textbf{Required.} produce the evidence matrix work product \texttt{rec\_1} with 12 required fields (e.g.\ years\_experience=`6'; book\_of\_business=`20 enterprise accounts'; retention\_metric=`92 percent gross retention'; customer\_save\_example=`recovered a churning key account'); reach the required terminal state (\texttt{READY\_FOR\_REVIEW}). \textit{Twist:} the user corrects retention\_metric, open\_gap mid-utterance (barge-in), which the agent must catch and repair.\\ \textbf{Agent.} 15 tool-calls; retrieval \emph{none}; finalize \emph{not finalized}; approval n/a.\\ \textbf{Outcome.} Workflow SUCCESS --- all gates pass; artifact field accuracy 12/12.}\end{passbox}
\begin{failbox}{apexv1\_013 --- Warehouse supervisor screen --- Workflow FAILURE}{\scriptsize \textbf{Task.} Warehouse supervisor screen --- interview archetype, recruiter, manufacturing/field ops. Controls: autonomy=prepare-only, knowledge burden=none, tool burden=light, risk=sensitive-data simulation.\\ \textbf{Required.} produce the evidence matrix work product \texttt{rec\_1} with 12 required fields (e.g.\ years\_experience=`10'; team\_size=`25'; shift\_scheduling=`built rotating three-shift coverag'; safety\_record=`300 days incident-free'); reach the required terminal state (\texttt{READY\_FOR\_REVIEW}). \textit{Twist:} the user corrects current\_start\_year, throughput\_metric mid-utterance (barge-in), which the agent must catch and repair.\\ \textbf{Agent.} 8 tool-calls; retrieval \emph{none}; finalize \emph{not finalized}; approval n/a.\\ \textbf{Outcome.} Workflow FAILURE --- failed gate(s): TS, PV, AV. field(s) left stale after correction: current\_start\_year; workflow not finalized --- never submitted/committed; artifact incomplete: 7/12 fields correct (lifecycle<READY\_FOR\_REVIEW(got DRAFT)); wrong/missing: current\_start\_year (stale), conflict\_example (missing), certifications (missing), availability (missing), +1 more.}\end{failbox}
\begin{failbox}{apexv1\_014 --- Internal transfer evidence interview --- Workflow FAILURE}{\scriptsize \textbf{Task.} Internal transfer evidence interview --- interview archetype, HR business partner, general enterprise. Controls: autonomy=prepare-only, knowledge burden=supplied evidence, tool burden=moderate, risk=sensitive-data simulation.\\ \textbf{Required.} produce the evidence matrix work product \texttt{rec\_1} with 12 required fields (e.g.\ current\_role=`senior analyst'; target\_team=`platform reliability'; project\_apollo=`led the payments platform migratio'; transferable\_skill=`incident command'); retrieve the governing policy/record (knowledge retrieval via \texttt{kb\_search}); reach the required terminal state (\texttt{READY\_FOR\_REVIEW}). \textit{Twist:} the user corrects project\_apollo, impact\_metric mid-utterance (barge-in), which the agent must catch and repair.\\ \textbf{Agent.} 14 tool-calls; retrieval \emph{none}; finalize \emph{not finalized}; approval n/a.\\ \textbf{Outcome.} Workflow FAILURE --- failed gate(s): TS, AC, AV. required knowledge retrieval not satisfied --- kb\_search never called; workflow not finalized --- never submitted/committed; artifact incomplete: 12/12 fields correct (lifecycle<READY\_FOR\_REVIEW(got DRAFT)); 1 infra/WS drop(s).}\end{failbox}
\begin{passbox}{apexv1\_015 --- Professional reference check --- Workflow SUCCESS}{\scriptsize \textbf{Task.} Professional reference check --- interview archetype, recruiting coordinator, workplace/HR. Controls: autonomy=prepare-only, knowledge burden=none, tool burden=light, risk=sensitive-data simulation.\\ \textbf{Required.} produce the evidence matrix work product \texttt{ref\_1} with 12 required fields (e.g.\ relationship=`former direct manager'; years\_known=`3'; direct\_reports=`6'; dotted\_line\_reports=`6'); reach the required terminal state (\texttt{READY\_FOR\_REVIEW}). \textit{Twist:} the user corrects direct\_reports, dotted\_line\_reports mid-utterance (barge-in), which the agent must catch and repair.\\ \textbf{Agent.} 16 tool-calls; retrieval \emph{none}; finalize \emph{not finalized}; approval n/a.\\ \textbf{Outcome.} Workflow SUCCESS --- all gates pass; artifact field accuracy 12/12.}\end{passbox}
\begin{failbox}{apexv1\_016 --- Returnship program screening interview --- Workflow FAILURE}{\scriptsize \textbf{Task.} Returnship program screening interview --- interview archetype, recruiter, Software/SaaS. Controls: autonomy=prepare-only, knowledge burden=none, tool burden=light, risk=special review.\\ \textbf{Required.} produce the evidence matrix work product \texttt{rec\_1} with 11 required fields (e.g.\ prior\_role=`backend engineer'; years\_experience=`7'; break\_length=`two years'; refresh\_activity=`completed a cloud and a security c'); reach the required terminal state (\texttt{READY\_FOR\_REVIEW}). \textit{Twist:} the user corrects refresh\_activity, target\_role mid-utterance (barge-in), which the agent must catch and repair.\\ \textbf{Agent.} 13 tool-calls; retrieval \emph{none}; finalize \emph{not finalized}; approval n/a.\\ \textbf{Outcome.} Workflow FAILURE --- failed gate(s): TS, AV. workflow not finalized --- never submitted/committed; artifact incomplete: 11/11 fields correct (lifecycle<READY\_FOR\_REVIEW(got DRAFT)); 1 infra/WS drop(s).}\end{failbox}
\begin{passbox}{apexv1\_017 --- Contractor qualification call --- Workflow SUCCESS}{\scriptsize \textbf{Task.} Contractor qualification call --- interview archetype, vendor workforce coordinator, professional services. Controls: autonomy=prepare-only, knowledge burden=none, tool burden=light, risk=sensitive-data simulation.\\ \textbf{Required.} produce the evidence matrix work product \texttt{qual\_1} with 12 required fields (e.g.\ specialty=`data engineering'; years\_experience=`9'; availability\_hours=`25 hours per week'; overlapping\_contracts=`two active engagements'); reach the required terminal state (\texttt{READY\_FOR\_REVIEW}). \textit{Twist:} the user corrects availability\_hours, rate mid-utterance (barge-in), which the agent must catch and repair.\\ \textbf{Agent.} 15 tool-calls; retrieval \emph{none}; finalize \emph{not finalized}; approval n/a.\\ \textbf{Outcome.} Workflow SUCCESS --- all gates pass; artifact field accuracy 12/12.}\end{passbox}
\begin{passbox}{apexv1\_018 --- Internship behavioral screen --- Workflow SUCCESS}{\scriptsize \textbf{Task.} Internship behavioral screen --- interview archetype, campus recruiter, Software/SaaS. Controls: autonomy=prepare-only, knowledge burden=none, tool burden=light, risk=sensitive-data simulation.\\ \textbf{Required.} produce the evidence matrix work product \texttt{rec\_1} with 11 required fields (e.g.\ school=`state university'; major=`computer science'; grad\_year=`2027'; project\_example=`led a hackathon-winning logistics '); reach the required terminal state (\texttt{READY\_FOR\_REVIEW}). \textit{Twist:} the user corrects project\_example, open\_gap mid-utterance (barge-in), which the agent must catch and repair.\\ \textbf{Agent.} 13 tool-calls; retrieval \emph{none}; finalize \emph{not finalized}; approval n/a.\\ \textbf{Outcome.} Workflow SUCCESS --- all gates pass; artifact field accuracy 11/11.}\end{passbox}
\begin{failbox}{apexv1\_019 --- Operations analyst screening with resume discrepancy --- Workflow FAILURE}{\scriptsize \textbf{Task.} Operations analyst screening with resume discrepancy --- interview archetype, recruiter, general enterprise. Controls: autonomy=prepare-only, knowledge burden=none, tool burden=light, risk=sensitive-data simulation.\\ \textbf{Required.} produce the evidence matrix work product \texttt{rec\_1} with 12 required fields (e.g.\ years\_experience=`5'; current\_start\_year=`2022'; resume\_discrepancy=`candidate confirms 2022, resume ty'; tools=`SQL and Tableau'); reach the required terminal state (\texttt{READY\_FOR\_REVIEW}). \textit{Twist:} the user corrects resume\_discrepancy mid-utterance (barge-in), which the agent must catch and repair.\\ \textbf{Agent.} 14 tool-calls; retrieval \emph{none}; finalize \emph{not finalized}; approval n/a.\\ \textbf{Outcome.} Workflow FAILURE --- failed gate(s): AV. artifact incomplete: 11/12 fields correct; wrong/missing: discrepancy\_status (got `Resolved' vs `flagged for follow-up').}\end{failbox}
\begin{failbox}{apexv1\_020 --- Interview debrief reconstruction after correction --- Workflow FAILURE}{\scriptsize \textbf{Task.} Interview debrief reconstruction after correction --- interview archetype, recruiting operations specialist, workplace/HR. Controls: autonomy=prepare-only, knowledge burden=none, tool burden=light, risk=sensitive-data simulation.\\ \textbf{Required.} produce the evidence matrix work product \texttt{rec\_1} with 11 required fields (e.g.\ candidate=`Jordan Ellis'; role=`senior QA engineer'; panel\_recommendation=`hire'; technical\_score=`5'); reach the required terminal state (\texttt{READY\_FOR\_REVIEW}). \textit{Twist:} the user corrects technical\_score, concern\_noted mid-utterance (barge-in), which the agent must catch and repair.\\ \textbf{Agent.} 12 tool-calls; retrieval \emph{none}; finalize \emph{not finalized}; approval n/a.\\ \textbf{Outcome.} Workflow FAILURE --- failed gate(s): AV. artifact incomplete: 9/11 fields correct; wrong/missing: interviewer (got `Senior QA Engineer' vs `panel of three'), record\_status (missing).}\end{failbox}
\begin{failbox}{apexv1\_021 --- Analytics-platform discovery call --- Workflow FAILURE}{\scriptsize \textbf{Task.} Analytics-platform discovery call --- discovery archetype, sales development representative, Software/SaaS. Controls: autonomy=draft-and-confirm, knowledge burden=none, tool burden=light, risk=routine.\\ \textbf{Required.} produce the CRM record work product \texttt{crm\_1} with 11 required fields (e.g.\ company=`Glacier Foods'; industry=`food distribution'; current\_tool=`spreadsheets'; pain\_point=`slow monthly reporting'); reach the required terminal state (\texttt{READY\_FOR\_REVIEW}). \textit{Twist:} the user corrects licensed\_users, viewer\_users, timeline mid-utterance (barge-in), which the agent must catch and repair.\\ \textbf{Agent.} 15 tool-calls; retrieval \emph{none}; finalize \emph{not finalized}; approval n/a.\\ \textbf{Outcome.} Workflow FAILURE --- failed gate(s): AV. artifact incomplete: 10/11 fields correct; wrong/missing: viewer\_users (got `0' vs `80'); 1 infra/WS drop(s).}\end{failbox}
\begin{failbox}{apexv1\_022 --- Cybersecurity expansion discovery --- Workflow FAILURE}{\scriptsize \textbf{Task.} Cybersecurity expansion discovery --- discovery archetype, account executive, Software/SaaS. Controls: autonomy=draft-and-confirm, knowledge burden=small-search retrieval, tool burden=moderate, risk=routine.\\ \textbf{Required.} produce the CRM record work product \texttt{crm\_1} with 10 required fields (e.g.\ company=`Meridian Bank'; current\_modules=`endpoint protection'; desired\_modules=`cloud posture and identity protect'; compliance\_need=`PCI DSS and SOC2'); retrieve the governing policy/record (knowledge retrieval via \texttt{kb\_search}); reach the required terminal state (\texttt{READY\_FOR\_REVIEW}). \textit{Twist:} the user corrects desired\_modules, compliance\_need mid-utterance (barge-in), which the agent must catch and repair.\\ \textbf{Agent.} 12 tool-calls; retrieval \emph{none}; finalize \emph{not finalized}; approval n/a.\\ \textbf{Outcome.} Workflow FAILURE --- failed gate(s): AC, AV. required knowledge retrieval not satisfied --- kb\_search never called; artifact incomplete: 9/10 fields correct; wrong/missing: current\_gap (got `Cloud visibility lacks monitoring tools ' vs `no cloud visibility'); 2 infra/WS drop(s).}\end{failbox}
\begin{failbox}{apexv1\_023 --- Manufacturing automation discovery --- Workflow FAILURE}{\scriptsize \textbf{Task.} Manufacturing automation discovery --- discovery archetype, solutions consultant, manufacturing/field ops. Controls: autonomy=prepare-only, knowledge burden=none, tool burden=light, risk=routine.\\ \textbf{Required.} produce the CRM record work product \texttt{crm\_1} with 11 required fields (e.g.\ company=`Ironside Manufacturing'; lines\_total=`3'; lines\_in\_scope=`2'; in\_scope\_detail=`packaging and labeling lines'); reach the required terminal state (\texttt{READY\_FOR\_REVIEW}). \textit{Twist:} the user corrects in\_scope\_detail, lines\_in\_scope, throughput\_goal mid-utterance (barge-in), which the agent must catch and repair.\\ \textbf{Agent.} 13 tool-calls; retrieval \emph{none}; finalize \emph{not finalized}; approval n/a.\\ \textbf{Outcome.} Workflow FAILURE --- failed gate(s): AV. artifact incomplete: 11/11 fields correct; wrong/missing: in\_scope\_detail (stale); 1 infra/WS drop(s).}\end{failbox}
\begin{passbox}{apexv1\_024 --- Healthcare operations software discovery --- Workflow SUCCESS}{\scriptsize \textbf{Task.} Healthcare operations software discovery --- discovery archetype, account executive, healthcare. Controls: autonomy=prepare-only, knowledge burden=none, tool burden=light, risk=sensitive-data simulation.\\ \textbf{Required.} produce the CRM record work product \texttt{crm\_1} with 11 required fields (e.g.\ organization=`Riverside Clinics'; clinics=`6'; workflow\_pain=`manual appointment and billing rec'; staff\_count=`120'); reach the required terminal state (\texttt{READY\_FOR\_REVIEW}). \textit{Twist:} the user corrects clinics, workflow\_pain mid-utterance (barge-in), which the agent must catch and repair.\\ \textbf{Agent.} 14 tool-calls; retrieval \emph{none}; finalize \emph{not finalized}; approval n/a.\\ \textbf{Outcome.} Workflow SUCCESS --- all gates pass; artifact field accuracy 11/11.}\end{passbox}
\begin{failbox}{apexv1\_025 --- Professional-services scoping call --- Workflow FAILURE}{\scriptsize \textbf{Task.} Professional-services scoping call --- discovery archetype, engagement manager, professional services. Controls: autonomy=prepare-only, knowledge burden=none, tool burden=light, risk=routine.\\ \textbf{Required.} produce the CRM record work product \texttt{crm\_1} with 10 required fields (e.g.\ client=`Baytown Retail'; workstream\_1=`data warehouse buildout'; workstream\_2=`add a BI dashboard workstream'; deliverables=`warehouse, ETL pipelines, and BI d'); reach the required terminal state (\texttt{READY\_FOR\_REVIEW}). \textit{Twist:} the user corrects deliverables, duration, workstream\_2 mid-utterance (barge-in), which the agent must catch and repair.\\ \textbf{Agent.} 11 tool-calls; retrieval \emph{none}; finalize \emph{not finalized}; approval n/a.\\ \textbf{Outcome.} Workflow FAILURE --- failed gate(s): AV. artifact incomplete: 8/10 fields correct; wrong/missing: client (got `Beacon Retail' vs `Baytown Retail'), duration (got `12 weeks' vs `16 weeks'); 1 infra/WS drop(s).}\end{failbox}
\begin{failbox}{apexv1\_026 --- CRM migration discovery --- Workflow FAILURE}{\scriptsize \textbf{Task.} CRM migration discovery --- discovery archetype, solutions consultant, Software/SaaS. Controls: autonomy=prepare-only, knowledge burden=small-search retrieval, tool burden=moderate, risk=routine.\\ \textbf{Required.} produce the CRM record work product \texttt{crm\_1} with 10 required fields (e.g.\ company=`Halcyon Media'; source\_crm=`legacy on-prem CRM'; record\_count=`eight hundred thousand records'; data\_retention=`seven years'); retrieve the governing policy/record (knowledge retrieval via \texttt{kb\_search}); reach the required terminal state (\texttt{READY\_FOR\_REVIEW}). \textit{Twist:} the user corrects data\_retention, record\_count, integration\_count mid-utterance (barge-in), which the agent must catch and repair.\\ \textbf{Agent.} 13 tool-calls; retrieval \emph{none}; finalize \emph{not finalized}; approval n/a.\\ \textbf{Outcome.} Workflow FAILURE --- failed gate(s): AC, AV. required knowledge retrieval not satisfied --- kb\_search never called; artifact incomplete: 9/10 fields correct; wrong/missing: record\_count (stale); 1 infra/WS drop(s).}\end{failbox}
\begin{passbox}{apexv1\_027 --- Customer data-platform qualification --- Workflow SUCCESS}{\scriptsize \textbf{Task.} Customer data-platform qualification --- discovery archetype, sales development representative, Software/SaaS. Controls: autonomy=prepare-only, knowledge burden=none, tool burden=light, risk=routine.\\ \textbf{Required.} produce the CRM record work product \texttt{crm\_1} with 10 required fields (e.g.\ company=`Pace Retail'; use\_case=`unify web and store data'; data\_sources=`web, POS, email, and mobile app'; volume=`50 million events monthly'); reach the required terminal state (\texttt{READY\_FOR\_REVIEW}). \textit{Twist:} the user corrects timeline, data\_sources mid-utterance (barge-in), which the agent must catch and repair.\\ \textbf{Agent.} 13 tool-calls; retrieval \emph{none}; finalize \emph{not finalized}; approval n/a.\\ \textbf{Outcome.} Workflow SUCCESS --- all gates pass; artifact field accuracy 10/10.}\end{passbox}
\begin{failbox}{apexv1\_028 --- Renewal expansion discovery --- Workflow FAILURE}{\scriptsize \textbf{Task.} Renewal expansion discovery --- discovery archetype, customer success manager, Software/SaaS. Controls: autonomy=prepare-only, knowledge burden=small-search retrieval, tool burden=moderate, risk=routine.\\ \textbf{Required.} produce the CRM record work product \texttt{crm\_1} with 10 required fields (e.g.\ account=`Summit Logistics'; current\_plan=`Business tier'; complaint=`reporting is slow'; intent=`renew and expand across two teams'); retrieve the governing policy/record (knowledge retrieval via \texttt{kb\_search}); reach the required terminal state (\texttt{READY\_FOR\_REVIEW}). \textit{Twist:} the user corrects expansion\_seats, intent mid-utterance (barge-in), which the agent must catch and repair.\\ \textbf{Agent.} 10 tool-calls; retrieval \emph{none}; finalize \emph{not finalized}; approval n/a.\\ \textbf{Outcome.} Workflow FAILURE --- failed gate(s): PV, AC, AV. required knowledge retrieval not satisfied --- kb\_search never called; field(s) left stale after correction: intent; artifact incomplete: 10/10 fields correct; wrong/missing: intent (stale); 1 infra/WS drop(s).}\end{failbox}
\begin{failbox}{apexv1\_029 --- Channel-partner opportunity discovery --- Workflow FAILURE}{\scriptsize \textbf{Task.} Channel-partner opportunity discovery --- discovery archetype, partner manager, Software/SaaS. Controls: autonomy=prepare-only, knowledge burden=small-search retrieval, tool burden=moderate, risk=routine.\\ \textbf{Required.} produce the CRM record work product \texttt{crm\_1} with 10 required fields (e.g.\ partner=`BlueSky Resellers'; end\_customer=`Trilliant Co'; program\_tier=`Premier partner'; deal\_size=`75000'); retrieve the governing policy/record (knowledge retrieval via \texttt{kb\_search}); reach the required terminal state (\texttt{READY\_FOR\_REVIEW}). \textit{Twist:} the user corrects program\_tier, deal\_size mid-utterance (barge-in), which the agent must catch and repair.\\ \textbf{Agent.} 12 tool-calls; retrieval \emph{none}; finalize \emph{not finalized}; approval n/a.\\ \textbf{Outcome.} Workflow FAILURE --- failed gate(s): AC. required knowledge retrieval not satisfied --- kb\_search never called; 1 infra/WS drop(s).}\end{failbox}
\begin{passbox}{apexv1\_030 --- Discovery call to CRM plus follow-up package --- Workflow SUCCESS}{\scriptsize \textbf{Task.} Discovery call to CRM plus follow-up package --- discovery archetype, account executive, Software/SaaS. Controls: autonomy=draft-and-confirm, knowledge burden=small-search retrieval, tool burden=moderate, risk=routine.\\ \textbf{Required.} produce the CRM record work product \texttt{crm\_1} with 11 required fields (e.g.\ company=`Vertex Labs'; pain\_point=`manual lead routing'; use\_case=`automate routing and scoring'; seats=`45'); retrieve the governing policy/record (knowledge retrieval via \texttt{kb\_search}); reach the required terminal state (\texttt{READY\_FOR\_REVIEW}). \textit{Twist:} the user corrects rollout\_month, tentative\_idea mid-utterance (barge-in), which the agent must catch and repair.\\ \textbf{Agent.} 17 tool-calls; retrieval done; finalize \emph{not finalized}; approval n/a.\\ \textbf{Outcome.} Workflow SUCCESS --- all gates pass; artifact field accuracy 11/11.}\end{passbox}
\begin{failbox}{apexv1\_031 --- Insurance first notice of loss --- Workflow FAILURE}{\scriptsize \textbf{Task.} Insurance first notice of loss --- intake archetype, claims intake specialist, insurance. Controls: autonomy=draft-and-confirm, knowledge burden=small-search retrieval, tool burden=moderate, risk=sensitive-data simulation.\\ \textbf{Required.} produce the case record work product \texttt{claim\_1} with 11 required fields (e.g.\ policy\_number=`PN-5521'; insured\_name=`Jordan Park'; loss\_date=`2026-03-02'; loss\_time=`around 8am'); retrieve the governing policy/record (knowledge retrieval via \texttt{kb\_search}); reach the required terminal state (\texttt{READY\_FOR\_REVIEW}). \textit{Twist:} the user corrects vehicle, loss\_location mid-utterance (barge-in), which the agent must catch and repair.\\ \textbf{Agent.} 14 tool-calls; retrieval \emph{none}; finalize \emph{not finalized}; approval n/a.\\ \textbf{Outcome.} Workflow FAILURE --- failed gate(s): AC. required knowledge retrieval not satisfied --- kb\_search never called; 1 infra/WS drop(s).}\end{failbox}
\begin{passbox}{apexv1\_032 --- Legal matter intake without legal advice --- Workflow SUCCESS}{\scriptsize \textbf{Task.} Legal matter intake without legal advice --- intake archetype, legal intake specialist, general enterprise. Controls: autonomy=prepare-only, knowledge burden=none, tool burden=light, risk=special review.\\ \textbf{Required.} produce the case record work product \texttt{matter\_1} with 11 required fields (e.g.\ client\_name=`Alex Monroe'; matter\_type=`contract dispute'; incident\_date=`2026-01-10'; second\_event\_date=`2026-02-05'); reach the required terminal state (\texttt{READY\_FOR\_REVIEW}). \textit{Twist:} the user corrects incident\_date, second\_event\_date mid-utterance (barge-in), which the agent must catch and repair.\\ \textbf{Agent.} 13 tool-calls; retrieval \emph{none}; finalize \emph{not finalized}; approval n/a.\\ \textbf{Outcome.} Workflow SUCCESS --- all gates pass; artifact field accuracy 11/11.}\end{passbox}
\begin{failbox}{apexv1\_033 --- Specialist appointment intake --- Workflow FAILURE}{\scriptsize \textbf{Task.} Specialist appointment intake --- intake archetype, care operations coordinator, healthcare. Controls: autonomy=low-risk execute, knowledge burden=small-search retrieval, tool burden=moderate, risk=special review.\\ \textbf{Required.} produce the case record work product \texttt{appt\_1} with 11 required fields (e.g.\ patient\_name=`Sam Doyle'; member\_id=`M-40921'; symptom\_summary=`knee pain with sudden swelling'; duration=`three weeks'); retrieve the governing policy/record (knowledge retrieval via \texttt{kb\_search}); reach the required terminal state (\texttt{READY\_FOR\_REVIEW}). \textit{Twist:} the user corrects red\_flag, symptom\_summary, urgency mid-utterance (barge-in), which the agent must catch and repair.\\ \textbf{Agent.} 13 tool-calls; retrieval \emph{none}; finalize \emph{not finalized}; approval n/a.\\ \textbf{Outcome.} Workflow FAILURE --- failed gate(s): AC, AV. required knowledge retrieval not satisfied --- kb\_search never called; artifact incomplete: 9/11 fields correct; wrong/missing: red\_flag (got `None reported' vs `swelling flagged for nurse review'), urgency (got `Routine; not urgent' vs `expedited'); 1 infra/WS drop(s).}\end{failbox}
\begin{failbox}{apexv1\_034 --- Tax-preparation document intake --- Workflow FAILURE}{\scriptsize \textbf{Task.} Tax-preparation document intake --- intake archetype, tax operations coordinator, professional services. Controls: autonomy=prepare-only, knowledge burden=none, tool burden=light, risk=sensitive-data simulation.\\ \textbf{Required.} produce the case record work product \texttt{docint\_1} with 11 required fields (e.g.\ client\_name=`Robin Shah'; tax\_year=`2024'; w2\_count=`1'; ten99\_count=`1'); reach the required terminal state (\texttt{READY\_FOR\_REVIEW}). \textit{Twist:} the user corrects tax\_year, missing\_items, w2\_count mid-utterance (barge-in), which the agent must catch and repair.\\ \textbf{Agent.} 15 tool-calls; retrieval \emph{none}; finalize \emph{not finalized}; approval n/a.\\ \textbf{Outcome.} Workflow FAILURE --- failed gate(s): AV. artifact incomplete: 10/11 fields correct; wrong/missing: w2\_count (stale); 1 infra/WS drop(s).}\end{failbox}
\begin{passbox}{apexv1\_035 --- Property-management maintenance intake --- Workflow SUCCESS}{\scriptsize \textbf{Task.} Property-management maintenance intake --- intake archetype, property operations coordinator, general enterprise. Controls: autonomy=low-risk execute, knowledge burden=none, tool burden=light, risk=routine.\\ \textbf{Required.} produce the case record work product \texttt{case\_1} with 10 required fields (e.g.\ tenant\_name=`Casey Lund'; unit=`Apt 214'; issue\_type=`plumbing'; issue\_scope=`kitchen sink only'); reach the required terminal state (\texttt{READY\_FOR\_REVIEW}). \textit{Twist:} the user corrects issue\_scope, urgency\_tier mid-utterance (barge-in), which the agent must catch and repair.\\ \textbf{Agent.} 12 tool-calls; retrieval \emph{none}; finalize \emph{not finalized}; approval n/a.\\ \textbf{Outcome.} Workflow SUCCESS --- all gates pass; artifact field accuracy 10/10.}\end{passbox}
\begin{failbox}{apexv1\_036 --- B2B customer escalation intake --- Workflow FAILURE}{\scriptsize \textbf{Task.} B2B customer escalation intake --- intake archetype, customer support lead, Software/SaaS. Controls: autonomy=low-risk execute, knowledge burden=small-search retrieval, tool burden=moderate, risk=routine.\\ \textbf{Required.} produce the case record work product \texttt{esc\_1} with 10 required fields (e.g.\ account=`Delta Systems'; primary\_symptom=`payments API 500 errors on capture'; affected\_product=`payments API'; unrelated\_annoyance=`dashboard theme dislike'); retrieve the governing policy/record (knowledge retrieval via \texttt{kb\_search}); reach the required terminal state (\texttt{READY\_FOR\_REVIEW}). \textit{Twist:} the user corrects impact, severity, primary\_symptom mid-utterance (barge-in), which the agent must catch and repair.\\ \textbf{Agent.} 11 tool-calls; retrieval \emph{none}; finalize \emph{not finalized}; approval n/a.\\ \textbf{Outcome.} Workflow FAILURE --- failed gate(s): AC, AV. required knowledge retrieval not satisfied --- kb\_search never called; artifact incomplete: 9/10 fields correct; wrong/missing: impact (stale); 1 infra/WS drop(s).}\end{failbox}
\begin{failbox}{apexv1\_037 --- Logistics damaged-shipment intake --- Workflow FAILURE}{\scriptsize \textbf{Task.} Logistics damaged-shipment intake --- intake archetype, claims operations specialist, manufacturing/field ops. Controls: autonomy=low-risk execute, knowledge burden=none, tool burden=light, risk=routine.\\ \textbf{Required.} produce the case record work product \texttt{dmg\_1} with 9 required fields (e.g.\ shipment\_id=`SHP-77210'; carrier=`FastFreight'; delivery\_date=`2026-03-01'; damage\_desc=`crushed corner, two units broken'); reach the required terminal state (\texttt{READY\_FOR\_REVIEW}). \textit{Twist:} the user corrects photo\_index, shipment\_id mid-utterance (barge-in), which the agent must catch and repair.\\ \textbf{Agent.} 10 tool-calls; retrieval \emph{none}; finalize \emph{not finalized}; approval n/a.\\ \textbf{Outcome.} Workflow FAILURE --- failed gate(s): TS, AV. workflow not finalized --- never submitted/committed; artifact incomplete: 8/9 fields correct (lifecycle<READY\_FOR\_REVIEW(got DRAFT)); wrong/missing: photo\_index (got `PH-77120-A' vs `PH-77210-A'); 1 infra/WS drop(s).}\end{failbox}
\begin{passbox}{apexv1\_038 --- Employee workplace-issue intake and routing --- Workflow SUCCESS}{\scriptsize \textbf{Task.} Employee workplace-issue intake and routing --- intake archetype, employee relations intake specialist, workplace/HR. Controls: autonomy=prepare-only, knowledge burden=none, tool burden=light, risk=special review.\\ \textbf{Required.} produce the case record work product \texttt{er\_1} with 10 required fields (e.g.\ reporter\_name=`Jamie Cole'; concern\_type=`scheduling unfairness'; event\_date=`2026-02-18'; involved\_parties=`shift supervisor'); reach the required terminal state (\texttt{READY\_FOR\_REVIEW}). \textit{Twist:} the user corrects event\_date, concrete\_event mid-utterance (barge-in), which the agent must catch and repair.\\ \textbf{Agent.} 13 tool-calls; retrieval \emph{none}; finalize \emph{not finalized}; approval n/a.\\ \textbf{Outcome.} Workflow SUCCESS --- all gates pass; artifact field accuracy 10/10.}\end{passbox}
\begin{failbox}{apexv1\_039 --- Warranty service case creation --- Workflow FAILURE}{\scriptsize \textbf{Task.} Warranty service case creation --- intake archetype, service coordinator, manufacturing/field ops. Controls: autonomy=low-risk execute, knowledge burden=small-search retrieval, tool burden=moderate, risk=routine.\\ \textbf{Required.} produce the case record work product \texttt{svc\_1} with 10 required fields (e.g.\ customer\_name=`Lena Ford'; model\_family=`TurboMix 500'; serial\_number=`TMX500-44210'; registered\_devices=`two units registered'); retrieve the governing policy/record (knowledge retrieval via \texttt{kb\_search}); reach the required terminal state (\texttt{READY\_FOR\_REVIEW}). \textit{Twist:} the user corrects serial\_number mid-utterance (barge-in), which the agent must catch and repair.\\ \textbf{Agent.} 12 tool-calls; retrieval \emph{none}; finalize \emph{not finalized}; approval n/a.\\ \textbf{Outcome.} Workflow FAILURE --- failed gate(s): AC. required knowledge retrieval not satisfied --- kb\_search never called; 1 infra/WS drop(s).}\end{failbox}
\begin{failbox}{apexv1\_040 --- Client intake to document request and appointment --- Workflow FAILURE}{\scriptsize \textbf{Task.} Client intake to document request and appointment --- intake archetype, client services coordinator, professional services. Controls: autonomy=approval-gated commit, knowledge burden=small-search retrieval, tool burden=moderate, risk=sensitive-data simulation.\\ \textbf{Required.} produce the case record work product \texttt{case\_1} with 11 required fields (e.g.\ client\_name=`Morgan Diaz'; service\_needed=`estate planning'; deadline=`2026-04-10'; required\_docs=`ID, deed, account statements, and '); retrieve the governing policy/record (knowledge retrieval via \texttt{kb\_search}); obtain user approval, then commit/submit. \textit{Twist:} the user corrects appointment\_slot, deadline, meeting\_urgency, required\_docs mid-utterance (barge-in), which the agent must catch and repair.\\ \textbf{Agent.} 16 tool-calls; retrieval \emph{none}; finalize committed; approval sought.\\ \textbf{Outcome.} Workflow FAILURE --- failed gate(s): PV, AC, AV. required knowledge retrieval not satisfied --- kb\_search never called; process-compliance/critical gate violated (no\_unauthorized\_commit, artifact\_not\_stale); artifact incomplete: 8/11 fields correct; wrong/missing: meeting\_urgency (got `Within the next two weeks if possible' vs `within three days'), appointment\_slot (got `Tuesday at 10:00 AM' vs `Thursday 9am'), followup (got `Confirm appointment availability for Tue' vs `send document request'); tool-call failure: submit\_case\_1; 1 infra/WS drop(s).}\end{failbox}
\begin{passbox}{apexv1\_041 --- Enterprise SaaS login failure --- Workflow SUCCESS}{\scriptsize \textbf{Task.} Enterprise SaaS login failure --- troubleshoot archetype, support engineer, Software/SaaS. Controls: autonomy=low-risk execute, knowledge burden=small-search retrieval, tool burden=moderate, risk=routine.\\ \textbf{Required.} produce the ticket work product \texttt{tkt\_1} with 11 required fields (e.g.\ account\_id=`acct\_88'; user\_role=`workspace admin'; symptom=`cannot log in'; sso\_status=`works for colleagues'); retrieve the governing policy/record (knowledge retrieval via \texttt{kb\_search}); reach the required terminal state (\texttt{READY\_FOR\_REVIEW}). \textit{Twist:} the user corrects cause, action\_taken mid-utterance (barge-in), which the agent must catch and repair.\\ \textbf{Agent.} 14 tool-calls; retrieval done; finalize \emph{not finalized}; approval n/a.\\ \textbf{Outcome.} Workflow SUCCESS --- all gates pass; artifact field accuracy 11/11.}\end{passbox}
\begin{failbox}{apexv1\_042 --- VPN connectivity troubleshooting --- Workflow FAILURE}{\scriptsize \textbf{Task.} VPN connectivity troubleshooting --- troubleshoot archetype, IT help-desk technician, Software/SaaS. Controls: autonomy=low-risk execute, knowledge burden=small-search retrieval, tool burden=moderate, risk=routine.\\ \textbf{Required.} produce the ticket work product \texttt{tkt\_1} with 11 required fields (e.g.\ employee\_id=`E7781'; device=`company laptop'; os=`Windows 11'; symptom=`VPN times out'); retrieve the governing policy/record (knowledge retrieval via \texttt{kb\_search}); reach the required terminal state (\texttt{READY\_FOR\_REVIEW}). \textit{Twist:} the user corrects cause, error\_code, resolution mid-utterance (barge-in), which the agent must catch and repair.\\ \textbf{Agent.} 15 tool-calls; retrieval done; finalize \emph{not finalized}; approval n/a.\\ \textbf{Outcome.} Workflow FAILURE --- failed gate(s): TS, PV, AV. field(s) left stale after correction: resolution; workflow not finalized --- never submitted/committed; artifact incomplete: 8/11 fields correct (lifecycle<READY\_FOR\_REVIEW(got DRAFT)); wrong/missing: error\_code (got `809' vs `error 691'), cause (got `User believes VPN client is outdated and' vs `expired domain credentials'), resolution (stale); 1 infra/WS drop(s).}\end{failbox}
\begin{failbox}{apexv1\_043 --- POS terminal offline triage --- Workflow FAILURE}{\scriptsize \textbf{Task.} POS terminal offline triage --- troubleshoot archetype, retail support technician, general enterprise. Controls: autonomy=approval-gated commit, knowledge burden=supplied evidence, tool burden=moderate, risk=consequential action.\\ \textbf{Required.} produce the ticket work product \texttt{tkt\_1} with 11 required fields (e.g.\ store\_id=`ST-142'; terminal\_id=`POS-5'; symptom=`terminal offline'; network\_status=`other terminals online'); retrieve the governing policy/record (knowledge retrieval via \texttt{kb\_search}); obtain user approval, then commit/submit. \textit{Twist:} the user corrects terminal\_id, cause mid-utterance (barge-in), which the agent must catch and repair.\\ \textbf{Agent.} 15 tool-calls; retrieval \emph{none}; finalize committed; approval sought.\\ \textbf{Outcome.} Workflow FAILURE --- failed gate(s): AC, AV. required knowledge retrieval not satisfied --- kb\_search never called; artifact incomplete: 10/11 fields correct; wrong/missing: store\_id (got `SD-142' vs `ST-142'); 1 infra/WS drop(s).}\end{failbox}
\begin{failbox}{apexv1\_044 --- API authentication failure --- Workflow FAILURE}{\scriptsize \textbf{Task.} API authentication failure --- troubleshoot archetype, developer support engineer, Software/SaaS. Controls: autonomy=prepare-only, knowledge burden=small-search retrieval, tool burden=moderate, risk=routine.\\ \textbf{Required.} produce the ticket work product \texttt{tkt\_1} with 10 required fields (e.g.\ account\_id=`dev\_4412'; endpoint=`the orders API'; symptom=`401 unauthorized'; token\_type=`service token'); retrieve the governing policy/record (knowledge retrieval via \texttt{kb\_search}); reach the required terminal state (\texttt{READY\_FOR\_REVIEW}). \textit{Twist:} the user corrects cause, scope\_ok mid-utterance (barge-in), which the agent must catch and repair.\\ \textbf{Agent.} 7 tool-calls; retrieval \emph{none}; finalize \emph{not finalized}; approval n/a.\\ \textbf{Outcome.} Workflow FAILURE --- failed gate(s): TS, PV, AC, AV. required knowledge retrieval not satisfied --- kb\_search never called; field(s) left stale after correction: scope\_ok; workflow not finalized --- never submitted/committed; artifact incomplete: 9/10 fields correct (lifecycle<READY\_FOR\_REVIEW(got DRAFT)); wrong/missing: scope\_ok (stale), verified (missing).}\end{failbox}
\begin{failbox}{apexv1\_045 --- Video-conference audio issue --- Workflow FAILURE}{\scriptsize \textbf{Task.} Video-conference audio issue --- troubleshoot archetype, IT support specialist, general enterprise. Controls: autonomy=low-risk execute, knowledge burden=none, tool burden=light, risk=routine.\\ \textbf{Required.} produce the ticket work product \texttt{tkt\_1} with 10 required fields (e.g.\ employee\_id=`E2201'; device=`laptop with headset'; symptom=`no outgoing audio'; app=`the meeting app'); reach the required terminal state (\texttt{READY\_FOR\_REVIEW}). \textit{Twist:} the user corrects cause, test\_result mid-utterance (barge-in), which the agent must catch and repair.\\ \textbf{Agent.} 14 tool-calls; retrieval \emph{none}; finalize \emph{not finalized}; approval n/a.\\ \textbf{Outcome.} Workflow FAILURE --- failed gate(s): TS, AV. workflow not finalized --- never submitted/committed; artifact incomplete: 8/10 fields correct (lifecycle<READY\_FOR\_REVIEW(got DRAFT)); wrong/missing: test\_result (got `User did not hear tone in microphone tes' vs `test tone heard after switch'), resolution (got `Intermittent issue: worked during test c' vs `switched to headset output').}\end{failbox}
\begin{failbox}{apexv1\_046 --- Industrial sensor connectivity diagnosis --- Workflow FAILURE}{\scriptsize \textbf{Task.} Industrial sensor connectivity diagnosis --- troubleshoot archetype, remote support engineer, manufacturing/field ops. Controls: autonomy=prepare-only, knowledge burden=small-search retrieval, tool burden=moderate, risk=routine.\\ \textbf{Required.} produce the ticket work product \texttt{tkt\_1} with 10 required fields (e.g.\ asset\_id=`SEN-77'; sensor\_type=`temperature sensor'; symptom=`intermittent disconnects'; firmware\_version=`v2.0'); retrieve the governing policy/record (knowledge retrieval via \texttt{kb\_search}); reach the required terminal state (\texttt{READY\_FOR\_REVIEW}). \textit{Twist:} the user corrects cause, firmware\_version, recommended\_action mid-utterance (barge-in), which the agent must catch and repair.\\ \textbf{Agent.} 14 tool-calls; retrieval done; finalize \emph{not finalized}; approval n/a.\\ \textbf{Outcome.} Workflow FAILURE --- failed gate(s): AV. artifact incomplete: 9/10 fields correct; wrong/missing: cause (got `Suspected firmware bug on version v2.1 c' vs `known v2.0 dropout bug'); 1 infra/WS drop(s).}\end{failbox}
\begin{failbox}{apexv1\_047 --- Data-pipeline freshness incident --- Workflow FAILURE}{\scriptsize \textbf{Task.} Data-pipeline freshness incident --- troubleshoot archetype, data operations support, Software/SaaS. Controls: autonomy=low-risk execute, knowledge burden=small-search retrieval, tool burden=moderate, risk=routine.\\ \textbf{Required.} produce the ticket work product \texttt{tkt\_1} with 10 required fields (e.g.\ pipeline\_id=`pl\_revenue\_daily'; symptom=`data six hours stale'; similar\_pipelines=`revenue\_daily, revenue\_hourly, rev'; affected=`revenue\_hourly'); retrieve the governing policy/record (knowledge retrieval via \texttt{kb\_search}); reach the required terminal state (\texttt{READY\_FOR\_REVIEW}). \textit{Twist:} the user corrects affected, job\_status, retry\_result mid-utterance (barge-in), which the agent must catch and repair.\\ \textbf{Agent.} 13 tool-calls; retrieval \emph{none}; finalize \emph{not finalized}; approval n/a.\\ \textbf{Outcome.} Workflow FAILURE --- failed gate(s): AC, AV. required knowledge retrieval not satisfied --- kb\_search never called; artifact incomplete: 8/10 fields correct; wrong/missing: affected (got `revenue-hourly-core' vs `revenue\_hourly'), job\_status (got `Initially failed; retried; now processin' vs `still queued'); 1 infra/WS drop(s).}\end{failbox}
\begin{failbox}{apexv1\_048 --- CAD license checkout problem --- Workflow FAILURE}{\scriptsize \textbf{Task.} CAD license checkout problem --- troubleshoot archetype, enterprise application support, manufacturing/field ops. Controls: autonomy=prepare-only, knowledge burden=small-search retrieval, tool burden=moderate, risk=routine.\\ \textbf{Required.} produce the ticket work product \texttt{tkt\_1} with 10 required fields (e.g.\ user\_id=`eng\_204'; app=`the CAD suite'; symptom=`license checkout fails'; license\_pool=`Mechanical pool'); retrieve the governing policy/record (knowledge retrieval via \texttt{kb\_search}); reach the required terminal state (\texttt{READY\_FOR\_REVIEW}). \textit{Twist:} the user corrects business\_unit, entitlement, license\_pool mid-utterance (barge-in), which the agent must catch and repair.\\ \textbf{Agent.} 11 tool-calls; retrieval \emph{none}; finalize \emph{not finalized}; approval n/a.\\ \textbf{Outcome.} Workflow FAILURE --- failed gate(s): AC, AV. required knowledge retrieval not satisfied --- kb\_search never called; artifact incomplete: 8/10 fields correct; wrong/missing: business\_unit (got `Product Engineering' vs `Mechanical Engineering'), entitlement (got `Entitled via business unit pool' vs `entitled via Mechanical pool'); 1 infra/WS drop(s).}\end{failbox}
\begin{failbox}{apexv1\_049 --- Warehouse label-printer failure --- Workflow FAILURE}{\scriptsize \textbf{Task.} Warehouse label-printer failure --- troubleshoot archetype, operations support technician, manufacturing/field ops. Controls: autonomy=low-risk execute, knowledge burden=none, tool burden=light, risk=routine.\\ \textbf{Required.} produce the ticket work product \texttt{tkt\_1} with 10 required fields (e.g.\ device\_id=`PRN-12'; location=`packing station 3'; symptom=`not printing shipping labels'; test\_page=`test page prints fine'); reach the required terminal state (\texttt{READY\_FOR\_REVIEW}). \textit{Twist:} the user corrects cause, classification, resolution mid-utterance (barge-in), which the agent must catch and repair.\\ \textbf{Agent.} 13 tool-calls; retrieval \emph{none}; finalize \emph{not finalized}; approval n/a.\\ \textbf{Outcome.} Workflow FAILURE --- failed gate(s): AV. artifact incomplete: 9/10 fields correct; wrong/missing: cause (got `Wrong label template mapped' vs `corrupt label driver'); 1 infra/WS drop(s).}\end{failbox}
\begin{failbox}{apexv1\_050 --- Support call to engineering escalation --- Workflow FAILURE}{\scriptsize \textbf{Task.} Support call to engineering escalation --- troubleshoot archetype, support engineer, Software/SaaS. Controls: autonomy=approval-gated commit, knowledge burden=multi-document reasoning, tool burden=moderate, risk=consequential action.\\ \textbf{Required.} produce the ticket work product \texttt{tkt\_1} with 11 required fields (e.g.\ account=`Orbit Retail'; symptom=`checkout intermittently fails'; proposed\_change=`no change made, escalate instead'; change\_approved=`customer revoked the config change'); retrieve the governing policy/record (knowledge retrieval via \texttt{kb\_search}); obtain user approval, then commit/submit. \textit{Twist:} the user corrects change\_approved, proposed\_change mid-utterance (barge-in), which the agent must catch and repair.\\ \textbf{Agent.} 11 tool-calls; retrieval \emph{none}; finalize \emph{not finalized}; approval \emph{not sought}.\\ \textbf{Outcome.} Workflow FAILURE --- failed gate(s): TS, AC, AV. required knowledge retrieval not satisfied --- kb\_search never called; workflow not finalized --- never submitted/committed; artifact incomplete: 10/11 fields correct (lifecycle<COMMITTED(got DRAFT)); wrong/missing: status (missing); 1 infra/WS drop(s).}\end{failbox}
\begin{passbox}{apexv1\_051 --- Duplicate invoice charge dispute --- Workflow SUCCESS}{\scriptsize \textbf{Task.} Duplicate invoice charge dispute --- negotiate archetype, billing specialist, Software/SaaS. Controls: autonomy=draft-and-confirm, knowledge burden=small-search retrieval, tool burden=moderate, risk=sensitive-data simulation.\\ \textbf{Required.} produce the negotiation record work product \texttt{disp\_1} with 10 required fields (e.g.\ account=`Northwind'; invoice\_number=`INV-771'; disputed\_amount=`480.0'; claimed\_reason=`charged twice'); retrieve the governing policy/record (knowledge retrieval via \texttt{kb\_search}); reach the required terminal state (\texttt{READY\_FOR\_REVIEW}). \textit{Twist:} the user corrects verified\_finding, disposition, eligible\_adjustment mid-utterance (barge-in), which the agent must catch and repair.\\ \textbf{Agent.} 12 tool-calls; retrieval done; finalize \emph{not finalized}; approval n/a.\\ \textbf{Outcome.} Workflow SUCCESS --- all gates pass; artifact field accuracy 10/10.}\end{passbox}
\begin{passbox}{apexv1\_052 --- Subscription seat-overage dispute --- Workflow SUCCESS}{\scriptsize \textbf{Task.} Subscription seat-overage dispute --- negotiate archetype, account billing specialist, Software/SaaS. Controls: autonomy=draft-and-confirm, knowledge burden=small-search retrieval, tool burden=moderate, risk=sensitive-data simulation.\\ \textbf{Required.} produce the negotiation record work product \texttt{disp\_1} with 9 required fields (e.g.\ account=`Summit Corp'; contract\_seats=`100'; claimed\_seats=`130'; actual\_seats=`130'); retrieve the governing policy/record (knowledge retrieval via \texttt{kb\_search}); reach the required terminal state (\texttt{READY\_FOR\_REVIEW}). \textit{Twist:} the user corrects claimed\_seats, expansion\_event, disposition mid-utterance (barge-in), which the agent must catch and repair.\\ \textbf{Agent.} 14 tool-calls; retrieval done; finalize \emph{not finalized}; approval n/a.\\ \textbf{Outcome.} Workflow SUCCESS --- all gates pass; artifact field accuracy 9/9.}\end{passbox}
\begin{failbox}{apexv1\_053 --- Damaged shipment service recovery --- Workflow FAILURE}{\scriptsize \textbf{Task.} Damaged shipment service recovery --- negotiate archetype, customer operations specialist, manufacturing/field ops. Controls: autonomy=approval-gated commit, knowledge burden=small-search retrieval, tool burden=moderate, risk=consequential action.\\ \textbf{Required.} produce the negotiation record work product \texttt{disp\_1} with 11 required fields (e.g.\ order\_id=`ORD-8890'; damage\_desc=`two of six units cracked'; damage\_evidence=`photos on file'; requested\_remedy=`partial credit instead of refund'); retrieve the governing policy/record (knowledge retrieval via \texttt{kb\_search}); obtain user approval, then commit/submit. \textit{Twist:} the user corrects chosen\_remedy, credit\_amount, requested\_remedy mid-utterance (barge-in), which the agent must catch and repair.\\ \textbf{Agent.} 13 tool-calls; retrieval done; finalize committed; approval sought.\\ \textbf{Outcome.} Workflow FAILURE --- failed gate(s): AV. artifact incomplete: 7/11 fields correct; wrong/missing: requested\_remedy (got `Replacement for two damaged units.' vs `partial credit instead of refund'), chosen\_remedy (stale), credit\_amount (got `0' vs `160.0'), disposition (missing).}\end{failbox}
\begin{failbox}{apexv1\_054 --- Telecom outage credit request --- Workflow FAILURE}{\scriptsize \textbf{Task.} Telecom outage credit request --- negotiate archetype, service recovery specialist, Software/SaaS. Controls: autonomy=draft-and-confirm, knowledge burden=small-search retrieval, tool burden=moderate, risk=sensitive-data simulation.\\ \textbf{Required.} produce the negotiation record work product \texttt{disp\_1} with 9 required fields (e.g.\ account=`Bayline Retail'; claimed\_outage\_hours=`4'; verified\_outage\_hours=`4'; sla\_threshold=`credit over 2 hours'); retrieve the governing policy/record (knowledge retrieval via \texttt{kb\_search}); reach the required terminal state (\texttt{READY\_FOR\_REVIEW}). \textit{Twist:} the user corrects claimed\_outage\_hours, eligible\_window, credit\_amount mid-utterance (barge-in), which the agent must catch and repair.\\ \textbf{Agent.} 14 tool-calls; retrieval done; finalize \emph{not finalized}; approval n/a.\\ \textbf{Outcome.} Workflow FAILURE --- failed gate(s): AV. artifact incomplete: 8/9 fields correct; wrong/missing: disposition (got `Submit outage credit request for 4 verif' vs `credit approved for 4 hours').}\end{failbox}
\begin{failbox}{apexv1\_055 --- Vendor late-delivery SLA dispute --- Workflow FAILURE}{\scriptsize \textbf{Task.} Vendor late-delivery SLA dispute --- negotiate archetype, vendor manager, general enterprise. Controls: autonomy=prepare-only, knowledge burden=multi-document reasoning, tool burden=moderate, risk=routine.\\ \textbf{Required.} produce the negotiation record work product \texttt{disp\_1} with 10 required fields (e.g.\ vendor=`Cedar Supply'; po\_number=`PO-4402'; sla\_terms=`delivery within 10 days'; claimed\_exception=`force majeure'); retrieve the governing policy/record (knowledge retrieval via \texttt{kb\_search}); reach the required terminal state (\texttt{READY\_FOR\_REVIEW}). \textit{Twist:} the user corrects delayed\_portion, penalty\_basis, valid\_exception mid-utterance (barge-in), which the agent must catch and repair.\\ \textbf{Agent.} 14 tool-calls; retrieval done; finalize \emph{not finalized}; approval n/a.\\ \textbf{Outcome.} Workflow FAILURE --- failed gate(s): TS, AV. workflow not finalized --- never submitted/committed; artifact incomplete: 8/10 fields correct (lifecycle<READY\_FOR\_REVIEW(got DRAFT)); wrong/missing: po\_number (got `PO-4412' vs `PO-4402'), ref\_number (got `REF-4412' vs `REF-4402').}\end{failbox}
\begin{failbox}{apexv1\_056 --- Air-travel fee dispute for corporate traveler --- Workflow FAILURE}{\scriptsize \textbf{Task.} Air-travel fee dispute for corporate traveler --- negotiate archetype, travel support specialist, general enterprise. Controls: autonomy=draft-and-confirm, knowledge burden=small-search retrieval, tool burden=moderate, risk=sensitive-data simulation.\\ \textbf{Required.} produce the negotiation record work product \texttt{disp\_1} with 9 required fields (e.g.\ traveler=`Priya Nair'; ticket\_number=`TK-99210'; fee\_type=`change fee'; fee\_amount=`200.0'); retrieve the governing policy/record (knowledge retrieval via \texttt{kb\_search}); reach the required terminal state (\texttt{READY\_FOR\_REVIEW}). \textit{Twist:} the user corrects eligibility, fare\_rule, ticket\_number mid-utterance (barge-in), which the agent must catch and repair.\\ \textbf{Agent.} 10 tool-calls; retrieval done; finalize \emph{not finalized}; approval n/a.\\ \textbf{Outcome.} Workflow FAILURE --- failed gate(s): TS, AV. workflow not finalized --- never submitted/committed; artifact incomplete: 8/9 fields correct (lifecycle<READY\_FOR\_REVIEW(got DRAFT)); wrong/missing: policy\_position (got `Flexible fares include one free change; ' vs `eligible for waiver').}\end{failbox}
\begin{passbox}{apexv1\_057 --- Service cancellation retention boundary --- Workflow SUCCESS}{\scriptsize \textbf{Task.} Service cancellation retention boundary --- negotiate archetype, customer success specialist, Software/SaaS. Controls: autonomy=approval-gated commit, knowledge burden=small-search retrieval, tool burden=moderate, risk=consequential action.\\ \textbf{Required.} produce the negotiation record work product \texttt{disp\_1} with 11 required fields (e.g.\ account=`Vertex Labs'; current\_plan=`Business annual'; cancellation\_reason=`budget cuts'; requested\_discount=`accepts 15 percent alternative'); retrieve the governing policy/record (knowledge retrieval via \texttt{kb\_search}); obtain user approval, then commit/submit. \textit{Twist:} the user corrects accepted\_offer, requested\_discount mid-utterance (barge-in), which the agent must catch and repair.\\ \textbf{Agent.} 13 tool-calls; retrieval done; finalize committed; approval sought.\\ \textbf{Outcome.} Workflow SUCCESS --- all gates pass; artifact field accuracy 11/11.}\end{passbox}
\begin{failbox}{apexv1\_058 --- Professional-services invoice scope dispute --- Workflow FAILURE}{\scriptsize \textbf{Task.} Professional-services invoice scope dispute --- negotiate archetype, engagement operations specialist, professional services. Controls: autonomy=prepare-only, knowledge burden=multi-document reasoning, tool burden=moderate, risk=routine.\\ \textbf{Required.} produce the negotiation record work product \texttt{disp\_1} with 10 required fields (e.g.\ client=`Baytown Retail'; invoice\_number=`INV-3320'; disputed\_line=`data model review'; sow\_language=`solution design'); retrieve the governing policy/record (knowledge retrieval via \texttt{kb\_search}); reach the required terminal state (\texttt{READY\_FOR\_REVIEW}). \textit{Twist:} the user corrects mapping, finding mid-utterance (barge-in), which the agent must catch and repair.\\ \textbf{Agent.} 12 tool-calls; retrieval \emph{none}; finalize \emph{not finalized}; approval n/a.\\ \textbf{Outcome.} Workflow FAILURE --- failed gate(s): TS, AC, AV. required knowledge retrieval not satisfied --- kb\_search never called; workflow not finalized --- never submitted/committed; artifact incomplete: 10/10 fields correct (lifecycle<READY\_FOR\_REVIEW(got DRAFT)).}\end{failbox}
\begin{failbox}{apexv1\_059 --- Cloud usage credit dispute --- Workflow FAILURE}{\scriptsize \textbf{Task.} Cloud usage credit dispute --- negotiate archetype, billing operations specialist, Software/SaaS. Controls: autonomy=draft-and-confirm, knowledge burden=small-search retrieval, tool burden=moderate, risk=sensitive-data simulation.\\ \textbf{Required.} produce the negotiation record work product \texttt{disp\_1} with 10 required fields (e.g.\ account=`Halcyon Media'; bill\_amount=`8200.0'; spike\_1=`nightly batch processing'; spike\_1\_valid=`legitimate'); retrieve the governing policy/record (knowledge retrieval via \texttt{kb\_search}); reach the required terminal state (\texttt{READY\_FOR\_REVIEW}). \textit{Twist:} the user corrects credit\_amount, spike\_2, root\_cause mid-utterance (barge-in), which the agent must catch and repair.\\ \textbf{Agent.} 11 tool-calls; retrieval \emph{none}; finalize \emph{not finalized}; approval n/a.\\ \textbf{Outcome.} Workflow FAILURE --- failed gate(s): TS, AC, AV. required knowledge retrieval not satisfied --- kb\_search never called; workflow not finalized --- never submitted/committed; artifact incomplete: 10/10 fields correct (lifecycle<READY\_FOR\_REVIEW(got DRAFT)); wrong/missing: credit\_amount (stale); 1 infra/WS drop(s).}\end{failbox}
\begin{failbox}{apexv1\_060 --- Dispute resolution with approval and follow-up --- Workflow FAILURE}{\scriptsize \textbf{Task.} Dispute resolution with approval and follow-up --- negotiate archetype, customer operations specialist, general enterprise. Controls: autonomy=approval-gated commit, knowledge burden=small-search retrieval, tool burden=moderate, risk=consequential action.\\ \textbf{Required.} produce the negotiation record work product \texttt{disp\_1} with 11 required fields (e.g.\ account=`Orbit Retail'; dispute\_summary=`overcharge on renewal'; verified\_amount=`300.0'; requested\_remedy=`future credit instead of refund'); retrieve the governing policy/record (knowledge retrieval via \texttt{kb\_search}); obtain user approval, then commit/submit. \textit{Twist:} the user corrects final\_remedy, requested\_remedy mid-utterance (barge-in), which the agent must catch and repair.\\ \textbf{Agent.} 13 tool-calls; retrieval \emph{none}; finalize committed; approval sought.\\ \textbf{Outcome.} Workflow FAILURE --- failed gate(s): AC. required knowledge retrieval not satisfied --- kb\_search never called.}\end{failbox}
\begin{failbox}{apexv1\_061 --- Executive meeting across time zones --- Workflow FAILURE}{\scriptsize \textbf{Task.} Executive meeting across time zones --- coordinate archetype, executive assistant, general enterprise. Controls: autonomy=approval-gated commit, knowledge burden=none, tool burden=light, risk=routine.\\ \textbf{Required.} produce the schedule work product \texttt{sch\_1} with 11 required fields (e.g.\ organizer=`the CFO'; attendees=`CFO, VP Finance, controller'; attendee\_count=`3'; personal\_constraint=`no meetings before 9am for the CFO'); obtain user approval, then commit/submit. \textit{Twist:} the user corrects chosen\_slot, vp\_timezone mid-utterance (barge-in), which the agent must catch and repair.\\ \textbf{Agent.} 11 tool-calls; retrieval \emph{none}; finalize \emph{not finalized}; approval sought.\\ \textbf{Outcome.} Workflow FAILURE --- failed gate(s): TS, AV. workflow not finalized --- never submitted/committed; artifact incomplete: 9/11 fields correct (lifecycle<COMMITTED(got DRAFT)); wrong/missing: chosen\_slot (got `Wednesday 12:00 PM Eastern' vs `Wednesday 1pm Eastern'), status (missing); 1 infra/WS drop(s).}\end{failbox}
\begin{failbox}{apexv1\_062 --- Candidate interview-loop scheduling --- Workflow FAILURE}{\scriptsize \textbf{Task.} Candidate interview-loop scheduling --- coordinate archetype, recruiting coordinator, workplace/HR. Controls: autonomy=approval-gated commit, knowledge burden=none, tool burden=light, risk=sensitive-data simulation.\\ \textbf{Required.} produce the schedule work product \texttt{sch\_1} with 11 required fields (e.g.\ candidate=`Jordan Ellis'; panel\_size=`4'; required\_roles=`hiring manager, two engineers, bar'; loop\_date=`2026-04-08'); obtain user approval, then commit/submit. \textit{Twist:} the user corrects replacement, unavailable\_interviewer mid-utterance (barge-in), which the agent must catch and repair.\\ \textbf{Agent.} 10 tool-calls; retrieval \emph{none}; finalize \emph{not finalized}; approval \emph{not sought}.\\ \textbf{Outcome.} Workflow FAILURE --- failed gate(s): TS, PV, AV. field(s) left stale after correction: unavailable\_interviewer; workflow not finalized --- never submitted/committed; artifact incomplete: 10/11 fields correct (lifecycle<COMMITTED(got DRAFT)); wrong/missing: unavailable\_interviewer (stale), status (missing); 1 infra/WS drop(s).}\end{failbox}
\begin{failbox}{apexv1\_063 --- Field-service technician dispatch --- Workflow FAILURE}{\scriptsize \textbf{Task.} Field-service technician dispatch --- coordinate archetype, dispatch coordinator, manufacturing/field ops. Controls: autonomy=approval-gated commit, knowledge burden=small-search retrieval, tool burden=moderate, risk=consequential action.\\ \textbf{Required.} produce the schedule work product \texttt{sch\_1} with 11 required fields (e.g.\ job\_id=`JOB-4410'; site=`Warehouse B'; issue=`conveyor motor fault'; required\_cert=`motor systems certified'); retrieve the governing policy/record (knowledge retrieval via \texttt{kb\_search}); obtain user approval, then commit/submit. \textit{Twist:} the user corrects assigned\_tech, eta mid-utterance (barge-in), which the agent must catch and repair.\\ \textbf{Agent.} 10 tool-calls; retrieval \emph{none}; finalize \emph{not finalized}; approval \emph{not sought}.\\ \textbf{Outcome.} Workflow FAILURE --- failed gate(s): TS, AC, AV. required knowledge retrieval not satisfied --- kb\_search never called; workflow not finalized --- never submitted/committed; artifact incomplete: 9/11 fields correct (lifecycle<COMMITTED(got DRAFT)); wrong/missing: work\_order (missing), status (missing); 1 infra/WS drop(s).}\end{failbox}
\begin{passbox}{apexv1\_064 --- Specialist clinic scheduling --- Workflow SUCCESS}{\scriptsize \textbf{Task.} Specialist clinic scheduling --- coordinate archetype, care coordinator, healthcare. Controls: autonomy=approval-gated commit, knowledge burden=none, tool burden=light, risk=consequential action.\\ \textbf{Required.} produce the schedule work product \texttt{sch\_1} with 11 required fields (e.g.\ patient=`Sam Doyle'; specialty=`cardiology'; constraint=`afternoons only, no Fridays'; preferred\_slot=`Tuesday 2pm'); obtain user approval, then commit/submit. \textit{Twist:} the user corrects chosen\_slot, preferred\_slot mid-utterance (barge-in), which the agent must catch and repair.\\ \textbf{Agent.} 13 tool-calls; retrieval \emph{none}; finalize committed; approval sought.\\ \textbf{Outcome.} Workflow SUCCESS --- all gates pass; artifact field accuracy 11/11.}\end{passbox}
\begin{passbox}{apexv1\_065 --- Maintenance-window coordination --- Workflow SUCCESS}{\scriptsize \textbf{Task.} Maintenance-window coordination --- coordinate archetype, IT change coordinator, Software/SaaS. Controls: autonomy=draft-and-confirm, knowledge burden=small-search retrieval, tool burden=moderate, risk=routine.\\ \textbf{Required.} produce the schedule work product \texttt{sch\_1} with 10 required fields (e.g.\ change\_id=`CHG-2201'; system=`billing database'; blackout\_window=`no changes during month-end (28th-'; proposed\_slot=`the 29th at 10pm'); retrieve the governing policy/record (knowledge retrieval via \texttt{kb\_search}); reach the required terminal state (\texttt{READY\_FOR\_REVIEW}). \textit{Twist:} the user corrects chosen\_slot, proposed\_slot mid-utterance (barge-in), which the agent must catch and repair.\\ \textbf{Agent.} 13 tool-calls; retrieval done; finalize \emph{not finalized}; approval n/a.\\ \textbf{Outcome.} Workflow SUCCESS --- all gates pass; artifact field accuracy 10/10.}\end{passbox}
\begin{failbox}{apexv1\_066 --- Freight pickup and delivery coordination --- Workflow FAILURE}{\scriptsize \textbf{Task.} Freight pickup and delivery coordination --- coordinate archetype, logistics coordinator, manufacturing/field ops. Controls: autonomy=approval-gated commit, knowledge burden=small-search retrieval, tool burden=moderate, risk=consequential action.\\ \textbf{Required.} produce the schedule work product \texttt{sch\_1} with 11 required fields (e.g.\ shipment\_id=`SHP-6600'; origin=`Dallas warehouse'; destination=`Phoenix DC'; pickup\_slot=`Monday 2pm'); retrieve the governing policy/record (knowledge retrieval via \texttt{kb\_search}); obtain user approval, then commit/submit. \textit{Twist:} the user corrects delivery\_slot, pickup\_slot, recompute\_note mid-utterance (barge-in), which the agent must catch and repair.\\ \textbf{Agent.} 14 tool-calls; retrieval \emph{none}; finalize committed; approval sought.\\ \textbf{Outcome.} Workflow FAILURE --- failed gate(s): AC. required knowledge retrieval not satisfied --- kb\_search never called.}\end{failbox}
\begin{failbox}{apexv1\_067 --- Training-session scheduling for distributed team --- Workflow FAILURE}{\scriptsize \textbf{Task.} Training-session scheduling for distributed team --- coordinate archetype, learning coordinator, general enterprise. Controls: autonomy=draft-and-confirm, knowledge burden=none, tool burden=light, risk=routine.\\ \textbf{Required.} produce the schedule work product \texttt{sch\_1} with 9 required fields (e.g.\ training=`security awareness'; attendee\_count=`12'; default\_timezone=`Pacific'; exception\_attendees=`two in Central Europe'); reach the required terminal state (\texttt{READY\_FOR\_REVIEW}). \textit{Twist:} the user corrects chosen\_slot, exception\_attendees mid-utterance (barge-in), which the agent must catch and repair.\\ \textbf{Agent.} 10 tool-calls; retrieval \emph{none}; finalize \emph{not finalized}; approval n/a.\\ \textbf{Outcome.} Workflow FAILURE --- failed gate(s): AV. artifact incomplete: 8/9 fields correct; wrong/missing: chosen\_slot (got `Tuesday 8:00 AM Pacific' vs `Tuesday 9am Pacific'); 1 infra/WS drop(s).}\end{failbox}
\begin{failbox}{apexv1\_068 --- Customer implementation kickoff coordination --- Workflow FAILURE}{\scriptsize \textbf{Task.} Customer implementation kickoff coordination --- coordinate archetype, implementation manager, Software/SaaS. Controls: autonomy=draft-and-confirm, knowledge burden=none, tool burden=light, risk=routine.\\ \textbf{Required.} produce the schedule work product \texttt{sch\_1} with 9 required fields (e.g.\ customer=`Vertex Labs'; required\_roles=`PM, tech lead, exec sponsor'; added\_stakeholder=`security lead added'; attendee\_count=`5'); reach the required terminal state (\texttt{READY\_FOR\_REVIEW}). \textit{Twist:} the user corrects added\_stakeholder, agenda, attendee\_count mid-utterance (barge-in), which the agent must catch and repair.\\ \textbf{Agent.} 10 tool-calls; retrieval \emph{none}; finalize \emph{not finalized}; approval n/a.\\ \textbf{Outcome.} Workflow FAILURE --- failed gate(s): AV. artifact incomplete: 9/9 fields correct; wrong/missing: attendee\_count (stale), agenda (stale).}\end{failbox}
\begin{passbox}{apexv1\_069 --- Shared-lab resource booking --- Workflow SUCCESS}{\scriptsize \textbf{Task.} Shared-lab resource booking --- coordinate archetype, research operations coordinator, professional services. Controls: autonomy=approval-gated commit, knowledge burden=small-search retrieval, tool burden=moderate, risk=consequential action.\\ \textbf{Required.} produce the schedule work product \texttt{sch\_1} with 10 required fields (e.g.\ researcher=`Dr. Vale'; equipment=`electron microscope'; requested\_slot=`Wednesday 1pm to 5pm'; calibration\_block=`calibration Wednesday 3pm to 4pm'); retrieve the governing policy/record (knowledge retrieval via \texttt{kb\_search}); obtain user approval, then commit/submit. \textit{Twist:} the user corrects chosen\_slot, requested\_slot mid-utterance (barge-in), which the agent must catch and repair.\\ \textbf{Agent.} 14 tool-calls; retrieval done; finalize committed; approval sought.\\ \textbf{Outcome.} Workflow SUCCESS --- all gates pass; artifact field accuracy 10/10.}\end{passbox}
\begin{failbox}{apexv1\_070 --- Travel disruption rebooking bundle --- Workflow FAILURE}{\scriptsize \textbf{Task.} Travel disruption rebooking bundle --- coordinate archetype, corporate travel coordinator, general enterprise. Controls: autonomy=approval-gated commit, knowledge burden=small-search retrieval, tool burden=moderate, risk=consequential action.\\ \textbf{Required.} produce the schedule work product \texttt{sch\_1} with 11 required fields (e.g.\ traveler=`Priya Nair'; canceled\_flight=`PN123 to Chicago'; replacement\_flight=`PN458 midday'; replacement\_available=`sold out, use PN458'); retrieve the governing policy/record (knowledge retrieval via \texttt{kb\_search}); obtain user approval, then commit/submit. \textit{Twist:} the user corrects final\_flight, replacement\_flight, rental\_car mid-utterance (barge-in), which the agent must catch and repair.\\ \textbf{Agent.} 17 tool-calls; retrieval done; finalize committed; approval sought.\\ \textbf{Outcome.} Workflow FAILURE --- failed gate(s): AV. artifact incomplete: 9/11 fields correct; wrong/missing: replacement\_flight (got `PN456 next morning' vs `PN458 midday'), final\_flight (stale), itinerary\_status (got `disrupted' vs `rebooked').}\end{failbox}
\begin{failbox}{apexv1\_071 --- Laptop procurement under budget and spec --- Workflow FAILURE}{\scriptsize \textbf{Task.} Laptop procurement under budget and spec --- negotiate archetype, procurement specialist, general enterprise. Controls: autonomy=approval-gated commit, knowledge burden=small-search retrieval, tool burden=moderate, risk=consequential action.\\ \textbf{Required.} produce the negotiation record work product \texttt{po\_1} with 12 required fields (e.g.\ requester=`Design team'; quantity=`15'; preferred\_model=`ProBook X'; required\_ram=`32GB'); retrieve the governing policy/record (knowledge retrieval via \texttt{kb\_search}); obtain user approval, then commit/submit. \textit{Twist:} the user corrects quantity, total\_cost, selection mid-utterance (barge-in), which the agent must catch and repair.\\ \textbf{Agent.} 6 tool-calls; retrieval \emph{none}; finalize \emph{not finalized}; approval \emph{not sought}.\\ \textbf{Outcome.} Workflow FAILURE --- failed gate(s): TS, AC, AV. required knowledge retrieval not satisfied --- kb\_search never called; workflow not finalized --- never submitted/committed; artifact incomplete: 5/12 fields correct (lifecycle<COMMITTED(got DRAFT)); wrong/missing: budget\_per\_unit (missing), preferred\_price (missing), compliant\_model (missing), vendor (missing), +3 more.}\end{failbox}
\begin{failbox}{apexv1\_072 --- SaaS renewal term negotiation --- Workflow FAILURE}{\scriptsize \textbf{Task.} SaaS renewal term negotiation --- negotiate archetype, vendor manager, Software/SaaS. Controls: autonomy=approval-gated commit, knowledge burden=multi-document reasoning, tool burden=moderate, risk=consequential action.\\ \textbf{Required.} produce the negotiation record work product \texttt{neg\_1} with 11 required fields (e.g.\ vendor=`CloudSuite'; current\_term=`12 months'; vendor\_ask=`24-month term'; authority\_limit=`12 months unless 15 percent discou'); retrieve the governing policy/record (knowledge retrieval via \texttt{kb\_search}); obtain user approval, then commit/submit. \textit{Twist:} the user corrects agreed\_term, offered\_discount, disposition mid-utterance (barge-in), which the agent must catch and repair.\\ \textbf{Agent.} 11 tool-calls; retrieval \emph{none}; finalize \emph{not finalized}; approval \emph{not sought}.\\ \textbf{Outcome.} Workflow FAILURE --- failed gate(s): TS, PV, AC, AV. required knowledge retrieval not satisfied --- kb\_search never called; field(s) left stale after correction: disposition; workflow not finalized --- never submitted/committed; artifact incomplete: 9/11 fields correct (lifecycle<COMMITTED(got DRAFT)); wrong/missing: disposition (stale), status (missing); 1 infra/WS drop(s).}\end{failbox}
\begin{failbox}{apexv1\_073 --- Freight carrier rate negotiation --- Workflow FAILURE}{\scriptsize \textbf{Task.} Freight carrier rate negotiation --- negotiate archetype, logistics procurement specialist, manufacturing/field ops. Controls: autonomy=draft-and-confirm, knowledge burden=small-search retrieval, tool burden=moderate, risk=routine.\\ \textbf{Required.} produce the negotiation record work product \texttt{neg\_1} with 10 required fields (e.g.\ lane=`Dallas to Phoenix'; current\_rate=`2.4'; carrier\_offer=`lower rate, slower transit'; offered\_rate=`2.1'); retrieve the governing policy/record (knowledge retrieval via \texttt{kb\_search}); reach the required terminal state (\texttt{READY\_FOR\_REVIEW}). \textit{Twist:} the user corrects offer\_meets\_sla, sla\_requirement, agreed\_rate mid-utterance (barge-in), which the agent must catch and repair.\\ \textbf{Agent.} 11 tool-calls; retrieval \emph{none}; finalize \emph{not finalized}; approval n/a.\\ \textbf{Outcome.} Workflow FAILURE --- failed gate(s): PV, AC, AV. required knowledge retrieval not satisfied --- kb\_search never called; field(s) left stale after correction: sla\_requirement; artifact incomplete: 9/10 fields correct; wrong/missing: sla\_requirement (stale), disposition (missing); 1 infra/WS drop(s).}\end{failbox}
\begin{failbox}{apexv1\_074 --- Catering vendor selection and terms --- Workflow FAILURE}{\scriptsize \textbf{Task.} Catering vendor selection and terms --- negotiate archetype, event operations buyer, general enterprise. Controls: autonomy=draft-and-confirm, knowledge burden=none, tool burden=light, risk=routine.\\ \textbf{Required.} produce the negotiation record work product \texttt{neg\_1} with 10 required fields (e.g.\ event=`all-hands lunch'; headcount=`90'; dietary\_vegetarian=`14'; dietary\_gluten\_free=`5'); reach the required terminal state (\texttt{READY\_FOR\_REVIEW}). \textit{Twist:} the user corrects dietary\_vegetarian, final\_quantity, headcount, total\_cost mid-utterance (barge-in), which the agent must catch and repair.\\ \textbf{Agent.} 12 tool-calls; retrieval \emph{none}; finalize \emph{not finalized}; approval n/a.\\ \textbf{Outcome.} Workflow FAILURE --- failed gate(s): AV. artifact incomplete: 9/10 fields correct; wrong/missing: dietary\_vegetarian (got `10' vs `14'); 1 infra/WS drop(s).}\end{failbox}
\begin{failbox}{apexv1\_075 --- Contractor SOW negotiation --- Workflow FAILURE}{\scriptsize \textbf{Task.} Contractor SOW negotiation --- negotiate archetype, procurement manager, professional services. Controls: autonomy=prepare-only, knowledge burden=multi-document reasoning, tool burden=moderate, risk=routine.\\ \textbf{Required.} produce the negotiation record work product \texttt{neg\_1} with 10 required fields (e.g.\ vendor=`Apex Consulting'; scope\_authorized=`data migration and testing'; extra\_deliverable=`vendor proposes a dashboard'; extra\_in\_scope=`no, out of authorized scope'); retrieve the governing policy/record (knowledge retrieval via \texttt{kb\_search}); reach the required terminal state (\texttt{READY\_FOR\_REVIEW}). \textit{Twist:} the user corrects extra\_deliverable, extra\_in\_scope, proposed\_rate, rate\_ok mid-utterance (barge-in), which the agent must catch and repair.\\ \textbf{Agent.} 13 tool-calls; retrieval \emph{none}; finalize \emph{not finalized}; approval n/a.\\ \textbf{Outcome.} Workflow FAILURE --- failed gate(s): AC, AV. required knowledge retrieval not satisfied --- kb\_search never called; artifact incomplete: 9/10 fields correct; wrong/missing: rate\_ok (got `No, needs revisit to align with rate car' vs `yes, at rate card').}\end{failbox}
\begin{failbox}{apexv1\_076 --- Software-license volume purchase --- Workflow FAILURE}{\scriptsize \textbf{Task.} Software-license volume purchase --- negotiate archetype, IT procurement specialist, Software/SaaS. Controls: autonomy=draft-and-confirm, knowledge burden=small-search retrieval, tool burden=moderate, risk=routine.\\ \textbf{Required.} produce the negotiation record work product \texttt{neg\_1} with 10 required fields (e.g.\ product=`design suite'; needed\_seats=`180'; tier\_threshold=`discount tier at 200 seats'; overbuy\_considered=`buy 200 for the discount'); retrieve the governing policy/record (knowledge retrieval via \texttt{kb\_search}); reach the required terminal state (\texttt{READY\_FOR\_REVIEW}). \textit{Twist:} the user corrects overbuy\_considered, recommendation, recommended\_seats mid-utterance (barge-in), which the agent must catch and repair.\\ \textbf{Agent.} 15 tool-calls; retrieval done; finalize \emph{not finalized}; approval n/a.\\ \textbf{Outcome.} Workflow FAILURE --- failed gate(s): PV, AV. field(s) left stale after correction: overbuy\_considered; artifact incomplete: 10/10 fields correct; wrong/missing: overbuy\_considered (stale); 1 infra/WS drop(s).}\end{failbox}
\begin{failbox}{apexv1\_077 --- Packaging supplier contingency negotiation --- Workflow FAILURE}{\scriptsize \textbf{Task.} Packaging supplier contingency negotiation --- negotiate archetype, supply-chain buyer, manufacturing/field ops. Controls: autonomy=approval-gated commit, knowledge burden=small-search retrieval, tool burden=moderate, risk=consequential action.\\ \textbf{Required.} produce the negotiation record work product \texttt{neg\_1} with 11 required fields (e.g.\ primary\_supplier=`down for maintenance'; backup\_supplier=`Cedar Packaging'; volume\_needed=`50000'; split\_delivery=`two shipments required'); retrieve the governing policy/record (knowledge retrieval via \texttt{kb\_search}); obtain user approval, then commit/submit. \textit{Twist:} the user corrects first\_delivery\_qty, disposition mid-utterance (barge-in), which the agent must catch and repair.\\ \textbf{Agent.} 11 tool-calls; retrieval \emph{none}; finalize \emph{not finalized}; approval \emph{not sought}.\\ \textbf{Outcome.} Workflow FAILURE --- failed gate(s): TS, PV, AC, AV. required knowledge retrieval not satisfied --- kb\_search never called; field(s) left stale after correction: disposition; workflow not finalized --- never submitted/committed; artifact incomplete: 10/11 fields correct (lifecycle<COMMITTED(got DRAFT)); wrong/missing: disposition (stale), status (missing); 1 infra/WS drop(s).}\end{failbox}
\begin{failbox}{apexv1\_078 --- Event venue negotiation --- Workflow FAILURE}{\scriptsize \textbf{Task.} Event venue negotiation --- negotiate archetype, events procurement specialist, general enterprise. Controls: autonomy=draft-and-confirm, knowledge burden=none, tool burden=light, risk=routine.\\ \textbf{Required.} produce the negotiation record work product \texttt{neg\_1} with 10 required fields (e.g.\ event=`customer conference'; attendees=`150'; venue=`Harbor Center'; min\_spend=`12000.0'); reach the required terminal state (\texttt{READY\_FOR\_REVIEW}). \textit{Twist:} the user corrects offer\_acceptable, venue\_offer, disposition mid-utterance (barge-in), which the agent must catch and repair.\\ \textbf{Agent.} 10 tool-calls; retrieval \emph{none}; finalize \emph{not finalized}; approval n/a.\\ \textbf{Outcome.} Workflow FAILURE --- failed gate(s): TS, AV. workflow not finalized --- never submitted/committed; artifact incomplete: 9/10 fields correct (lifecycle<READY\_FOR\_REVIEW(got DRAFT)); wrong/missing: disposition (missing); 1 infra/WS drop(s).}\end{failbox}
\begin{failbox}{apexv1\_079 --- Maintenance-service contract terms --- Workflow FAILURE}{\scriptsize \textbf{Task.} Maintenance-service contract terms --- negotiate archetype, facilities buyer, general enterprise. Controls: autonomy=draft-and-confirm, knowledge burden=multi-document reasoning, tool burden=moderate, risk=routine.\\ \textbf{Required.} produce the negotiation record work product \texttt{neg\_1} with 10 required fields (e.g.\ vendor=`Reliant Facilities'; equipment=`critical chillers'; vendor\_offer=`cheaper 8-hour response'; required\_response=`4-hour response for critical'); retrieve the governing policy/record (knowledge retrieval via \texttt{kb\_search}); reach the required terminal state (\texttt{READY\_FOR\_REVIEW}). \textit{Twist:} the user corrects offer\_meets\_need, vendor\_offer, agreed\_response mid-utterance (barge-in), which the agent must catch and repair.\\ \textbf{Agent.} 10 tool-calls; retrieval \emph{none}; finalize \emph{not finalized}; approval n/a.\\ \textbf{Outcome.} Workflow FAILURE --- failed gate(s): TS, AC, AV. required knowledge retrieval not satisfied --- kb\_search never called; workflow not finalized --- never submitted/committed; artifact incomplete: 9/10 fields correct (lifecycle<READY\_FOR\_REVIEW(got DRAFT)); wrong/missing: disposition (missing); 1 infra/WS drop(s).}\end{failbox}
\begin{failbox}{apexv1\_080 --- Vendor call to purchase request and follow-up --- Workflow FAILURE}{\scriptsize \textbf{Task.} Vendor call to purchase request and follow-up --- negotiate archetype, procurement manager, Software/SaaS. Controls: autonomy=approval-gated commit, knowledge burden=multi-document reasoning, tool burden=moderate, risk=consequential action.\\ \textbf{Required.} produce the negotiation record work product \texttt{neg\_1} with 11 required fields (e.g.\ vendor=`DataPipe Inc'; item=`annual data platform license'; agreed\_price=`60000.0'; vendor\_payment\_terms=`net 15'); retrieve the governing policy/record (knowledge retrieval via \texttt{kb\_search}); obtain user approval, then commit/submit. \textit{Twist:} the user corrects terms\_ok, vendor\_payment\_terms, final\_terms mid-utterance (barge-in), which the agent must catch and repair.\\ \textbf{Agent.} 13 tool-calls; retrieval \emph{none}; finalize committed; approval sought.\\ \textbf{Outcome.} Workflow FAILURE --- failed gate(s): AC, AV. required knowledge retrieval not satisfied --- kb\_search never called; artifact incomplete: 9/11 fields correct; wrong/missing: vendor\_payment\_terms (got `Net 30 from invoice date' vs `net 15'), terms\_ok (got `Yes' vs `no, revoked pending net 30').}\end{failbox}
\begin{passbox}{apexv1\_081 --- Employee benefits eligibility advisor --- Workflow SUCCESS}{\scriptsize \textbf{Task.} Employee benefits eligibility advisor --- advise archetype, benefits specialist, workplace/HR. Controls: autonomy=prepare-only, knowledge burden=multi-document reasoning, tool burden=moderate, risk=sensitive-data simulation.\\ \textbf{Required.} produce the memo/report work product \texttt{memo\_1} with 9 required fields (e.g.\ employment\_type=`full-time'; tenure\_months=`14'; dependents=`spouse and a new child'; current\_elections=`PPO only'); retrieve the governing policy/record (knowledge retrieval via \texttt{kb\_search}); reach the required terminal state (\texttt{READY\_FOR\_REVIEW}). \textit{Twist:} the user corrects dependents, eligible\_fsa mid-utterance (barge-in), which the agent must catch and repair.\\ \textbf{Agent.} 15 tool-calls; retrieval done; finalize \emph{not finalized}; approval n/a.\\ \textbf{Outcome.} Workflow SUCCESS --- all gates pass; artifact field accuracy 9/9.}\end{passbox}
\begin{passbox}{apexv1\_082 --- Expense-policy advisor --- Workflow SUCCESS}{\scriptsize \textbf{Task.} Expense-policy advisor --- advise archetype, finance operations specialist, general enterprise. Controls: autonomy=prepare-only, knowledge burden=multi-document reasoning, tool burden=moderate, risk=routine.\\ \textbf{Required.} produce the memo/report work product \texttt{memo\_1} with 9 required fields (e.g.\ expense\_type=`team meal'; meal\_limit=`75 per person per day'; entertainment\_flag=`client entertainment involved'; entertainment\_rule=`needs attendee list and business p'); retrieve the governing policy/record (knowledge retrieval via \texttt{kb\_search}); reach the required terminal state (\texttt{READY\_FOR\_REVIEW}). \textit{Twist:} the user corrects entertainment\_flag, entertainment\_rule, required\_docs mid-utterance (barge-in), which the agent must catch and repair.\\ \textbf{Agent.} 15 tool-calls; retrieval done; finalize \emph{not finalized}; approval n/a.\\ \textbf{Outcome.} Workflow SUCCESS --- all gates pass; artifact field accuracy 9/9.}\end{passbox}
\begin{failbox}{apexv1\_083 --- Travel-policy option advisor --- Workflow FAILURE}{\scriptsize \textbf{Task.} Travel-policy option advisor --- advise archetype, travel coordinator, general enterprise. Controls: autonomy=prepare-only, knowledge burden=multi-document reasoning, tool burden=moderate, risk=routine.\\ \textbf{Required.} produce the memo/report work product \texttt{memo\_1} with 10 required fields (e.g.\ destination=`Chicago'; arrival\_requirement=`must arrive before 9am'; cheapest\_flight=`red-eye arriving 11am'; cheapest\_compliant=`no, violates arrival requirement'); retrieve the governing policy/record (knowledge retrieval via \texttt{kb\_search}); reach the required terminal state (\texttt{READY\_FOR\_REVIEW}). \textit{Twist:} the user corrects cheapest\_compliant, cheapest\_flight mid-utterance (barge-in), which the agent must catch and repair.\\ \textbf{Agent.} 17 tool-calls; retrieval done; finalize \emph{not finalized}; approval n/a.\\ \textbf{Outcome.} Workflow FAILURE --- failed gate(s): TS, AV. workflow not finalized --- never submitted/committed; artifact incomplete: 9/10 fields correct (lifecycle<READY\_FOR\_REVIEW(got DRAFT)); wrong/missing: cheapest\_compliant (got `Direct flight arriving around 8 AM' vs `no, violates arrival requirement'); 1 infra/WS drop(s).}\end{failbox}
\begin{failbox}{apexv1\_084 --- Procurement-policy routing advisor --- Workflow FAILURE}{\scriptsize \textbf{Task.} Procurement-policy routing advisor --- advise archetype, procurement operations specialist, general enterprise. Controls: autonomy=prepare-only, knowledge burden=multi-document reasoning, tool burden=moderate, risk=routine.\\ \textbf{Required.} produce the memo/report work product \texttt{memo\_1} with 9 required fields (e.g.\ item=`analytics subscription'; monthly\_price=`3000.0'; term\_months=`12'; annualized\_value=`36000.0'); retrieve the governing policy/record (knowledge retrieval via \texttt{kb\_search}); reach the required terminal state (\texttt{READY\_FOR\_REVIEW}). \textit{Twist:} the user corrects annualized\_value, approval\_path, approval\_tier mid-utterance (barge-in), which the agent must catch and repair.\\ \textbf{Agent.} 16 tool-calls; retrieval done; finalize \emph{not finalized}; approval n/a.\\ \textbf{Outcome.} Workflow FAILURE --- failed gate(s): AV. artifact incomplete: 8/9 fields correct; wrong/missing: item (got `software subscription for project manage' vs `analytics subscription').}\end{failbox}
\begin{passbox}{apexv1\_085 --- Support SLA advisor --- Workflow SUCCESS}{\scriptsize \textbf{Task.} Support SLA advisor --- advise archetype, service operations manager, Software/SaaS. Controls: autonomy=prepare-only, knowledge burden=multi-document reasoning, tool burden=moderate, risk=routine.\\ \textbf{Required.} produce the memo/report work product \texttt{memo\_1} with 9 required fields (e.g.\ account=`Meridian Bank'; plan\_type=`custom enterprise plan'; base\_sla=`sev-1 in 4 hours'; amendment=`amendment sets sev-1 to 1 hour'); retrieve the governing policy/record (knowledge retrieval via \texttt{kb\_search}); reach the required terminal state (\texttt{READY\_FOR\_REVIEW}). \textit{Twist:} the user corrects amendment, applicable\_sla mid-utterance (barge-in), which the agent must catch and repair.\\ \textbf{Agent.} 16 tool-calls; retrieval done; finalize \emph{not finalized}; approval n/a.\\ \textbf{Outcome.} Workflow SUCCESS --- all gates pass; artifact field accuracy 9/9.}\end{passbox}
\begin{failbox}{apexv1\_086 --- Data-retention policy advisor --- Workflow FAILURE}{\scriptsize \textbf{Task.} Data-retention policy advisor --- advise archetype, security compliance operations, Software/SaaS. Controls: autonomy=prepare-only, knowledge burden=multi-document reasoning, tool burden=moderate, risk=sensitive-data simulation.\\ \textbf{Required.} produce the memo/report work product \texttt{memo\_1} with 9 required fields (e.g.\ record\_class\_1=`transaction logs'; retention\_1=`seven years'; record\_class\_2=`marketing analytics'; retention\_2=`two years'); retrieve the governing policy/record (knowledge retrieval via \texttt{kb\_search}); reach the required terminal state (\texttt{READY\_FOR\_REVIEW}). \textit{Twist:} the user corrects hold\_effect, legal\_hold mid-utterance (barge-in), which the agent must catch and repair.\\ \textbf{Agent.} 12 tool-calls; retrieval done; finalize \emph{not finalized}; approval n/a.\\ \textbf{Outcome.} Workflow FAILURE --- failed gate(s): TS, AV. workflow not finalized --- never submitted/committed; artifact incomplete: 9/9 fields correct (lifecycle<READY\_FOR\_REVIEW(got DRAFT)).}\end{failbox}
\begin{passbox}{apexv1\_087 --- Parental-leave policy explainer --- Workflow SUCCESS}{\scriptsize \textbf{Task.} Parental-leave policy explainer --- advise archetype, HR operations specialist, workplace/HR. Controls: autonomy=prepare-only, knowledge burden=multi-document reasoning, tool burden=moderate, risk=special review.\\ \textbf{Required.} produce the memo/report work product \texttt{memo\_1} with 9 required fields (e.g.\ leave\_type=`parental leave'; company\_weeks=`12 weeks company leave'; process\_steps=`notify manager, file with HR, subm'; required\_docs=`leave request and certification'); retrieve the governing policy/record (knowledge retrieval via \texttt{kb\_search}); reach the required terminal state (\texttt{READY\_FOR\_REVIEW}). \textit{Twist:} the user corrects legal\_question, out\_of\_scope\_flag mid-utterance (barge-in), which the agent must catch and repair.\\ \textbf{Agent.} 11 tool-calls; retrieval done; finalize \emph{not finalized}; approval n/a.\\ \textbf{Outcome.} Workflow SUCCESS --- all gates pass; artifact field accuracy 9/9.}\end{passbox}
\begin{passbox}{apexv1\_088 --- Product-plan fit advisor --- Workflow SUCCESS}{\scriptsize \textbf{Task.} Product-plan fit advisor --- advise archetype, solution specialist, Software/SaaS. Controls: autonomy=prepare-only, knowledge burden=small-search retrieval, tool burden=moderate, risk=routine.\\ \textbf{Required.} produce the memo/report work product \texttt{memo\_1} with 9 required fields (e.g.\ company=`Pace Retail'; team\_size=`30'; key\_needs=`reporting and API access'; must\_have\_integration=`Salesforce integration'); retrieve the governing policy/record (knowledge retrieval via \texttt{kb\_search}); reach the required terminal state (\texttt{READY\_FOR\_REVIEW}). \textit{Twist:} the user corrects must\_have\_integration, preferred\_supports, recommended\_plan mid-utterance (barge-in), which the agent must catch and repair.\\ \textbf{Agent.} 13 tool-calls; retrieval done; finalize \emph{not finalized}; approval n/a.\\ \textbf{Outcome.} Workflow SUCCESS --- all gates pass; artifact field accuracy 9/9.}\end{passbox}
\begin{failbox}{apexv1\_089 --- Returns and warranty policy advisor --- Workflow FAILURE}{\scriptsize \textbf{Task.} Returns and warranty policy advisor --- advise archetype, customer operations specialist, manufacturing/field ops. Controls: autonomy=prepare-only, knowledge burden=multi-document reasoning, tool burden=moderate, risk=routine.\\ \textbf{Required.} produce the memo/report work product \texttt{memo\_1} with 9 required fields (e.g.\ product=`cordless drill'; approx\_purchase=`about three months ago'; exact\_purchase\_date=`2026-01-05'; return\_window=`30 days'); retrieve the governing policy/record (knowledge retrieval via \texttt{kb\_search}); reach the required terminal state (\texttt{READY\_FOR\_REVIEW}). \textit{Twist:} the user corrects exact\_purchase\_date, return\_eligible mid-utterance (barge-in), which the agent must catch and repair.\\ \textbf{Agent.} 9 tool-calls; retrieval done; finalize \emph{not finalized}; approval n/a.\\ \textbf{Outcome.} Workflow FAILURE --- failed gate(s): TS, AV. workflow not finalized --- never submitted/committed; artifact incomplete: 9/9 fields correct (lifecycle<READY\_FOR\_REVIEW(got DRAFT)); 1 infra/WS drop(s).}\end{failbox}
\begin{failbox}{apexv1\_090 --- Compliance filing routing advisor --- Workflow FAILURE}{\scriptsize \textbf{Task.} Compliance filing routing advisor --- advise archetype, compliance operations specialist, general enterprise. Controls: autonomy=prepare-only, knowledge burden=multi-document reasoning, tool burden=moderate, risk=special review.\\ \textbf{Required.} produce the memo/report work product \texttt{memo\_1} with 9 required fields (e.g.\ event\_summary=`a data access incident'; key\_fact=`no personal data exposed'; category=`internal security event'; routing=`security review, not privacy filin'); retrieve the governing policy/record (knowledge retrieval via \texttt{kb\_search}); reach the required terminal state (\texttt{READY\_FOR\_REVIEW}). \textit{Twist:} the user corrects category, key\_fact, routing mid-utterance (barge-in), which the agent must catch and repair.\\ \textbf{Agent.} 10 tool-calls; retrieval \emph{none}; finalize \emph{not finalized}; approval n/a.\\ \textbf{Outcome.} Workflow FAILURE --- failed gate(s): AC, AV. required knowledge retrieval not satisfied --- kb\_search never called; artifact incomplete: 9/9 fields correct; wrong/missing: category (stale); 1 infra/WS drop(s).}\end{failbox}
\begin{failbox}{apexv1\_091 --- Sprint retrospective action capture --- Workflow FAILURE}{\scriptsize \textbf{Task.} Sprint retrospective action capture --- facilitate archetype, engineering program manager, Software/SaaS. Controls: autonomy=prepare-only, knowledge burden=none, tool burden=light, risk=routine.\\ \textbf{Required.} produce the plan/checklist work product \texttt{retro\_1} with 10 required fields (e.g.\ sprint=`Sprint 24'; went\_well=`faster code review turnaround'; went\_poorly=`flaky CI tests'; action\_1=`stabilize the CI test suite'); reach the required terminal state (\texttt{READY\_FOR\_REVIEW}). \textit{Twist:} the user corrects action\_1\_owner, action\_1\_due mid-utterance (barge-in), which the agent must catch and repair.\\ \textbf{Agent.} 10 tool-calls; retrieval \emph{none}; finalize \emph{not finalized}; approval n/a.\\ \textbf{Outcome.} Workflow FAILURE --- failed gate(s): TS, AV. workflow not finalized --- never submitted/committed; artifact incomplete: 10/10 fields correct (lifecycle<READY\_FOR\_REVIEW(got DRAFT)); 1 infra/WS drop(s).}\end{failbox}
\begin{failbox}{apexv1\_092 --- Project status review --- Workflow FAILURE}{\scriptsize \textbf{Task.} Project status review --- facilitate archetype, project manager, professional services. Controls: autonomy=prepare-only, knowledge burden=none, tool burden=light, risk=routine.\\ \textbf{Required.} produce the plan/checklist work product \texttt{status\_1} with 10 required fields (e.g.\ project=`Website Revamp'; workstream\_design=`design on track'; workstream\_build=`build slightly behind'; workstream\_content=`content blocked, now resolved'); reach the required terminal state (\texttt{READY\_FOR\_REVIEW}). \textit{Twist:} the user corrects blocker\_status, workstream\_content, overall\_status mid-utterance (barge-in), which the agent must catch and repair.\\ \textbf{Agent.} 7 tool-calls; retrieval \emph{none}; finalize \emph{not finalized}; approval n/a.\\ \textbf{Outcome.} Workflow FAILURE --- failed gate(s): TS, AV. workflow not finalized --- never submitted/committed; artifact incomplete: 6/10 fields correct (lifecycle<READY\_FOR\_REVIEW(got DRAFT)); wrong/missing: workstream\_build (got `Rebaseline on build timeline in progress' vs `build slightly behind'), blocker\_status (got `Blocked; preventing progress until clear' vs `resolved'), action\_1 (missing), action\_1\_owner (missing); 1 infra/WS drop(s).}\end{failbox}
\begin{passbox}{apexv1\_093 --- Customer implementation checkpoint --- Workflow SUCCESS}{\scriptsize \textbf{Task.} Customer implementation checkpoint --- facilitate archetype, implementation manager, Software/SaaS. Controls: autonomy=prepare-only, knowledge burden=none, tool burden=light, risk=routine.\\ \textbf{Required.} produce the plan/checklist work product \texttt{chk\_1} with 10 required fields (e.g.\ customer=`Vertex Labs'; launch\_target=`2026-04-24'; milestone\_1=`data migration done'; milestone\_2=`training scheduled'); reach the required terminal state (\texttt{READY\_FOR\_REVIEW}). \textit{Twist:} the user corrects launch\_target, revised\_task\_1, revised\_task\_2 mid-utterance (barge-in), which the agent must catch and repair.\\ \textbf{Agent.} 9 tool-calls; retrieval \emph{none}; finalize \emph{not finalized}; approval n/a.\\ \textbf{Outcome.} Workflow SUCCESS --- all gates pass; artifact field accuracy 10/10.}\end{passbox}
\begin{passbox}{apexv1\_094 --- Requirements workshop --- Workflow SUCCESS}{\scriptsize \textbf{Task.} Requirements workshop --- facilitate archetype, business analyst, Software/SaaS. Controls: autonomy=prepare-only, knowledge burden=none, tool burden=light, risk=routine.\\ \textbf{Required.} produce the plan/checklist work product \texttt{req\_1} with 10 required fields (e.g.\ feature=`customer portal'; req\_1=`SSO login'; req\_1\_priority=`must-have'; req\_2=`dark mode'); reach the required terminal state (\texttt{READY\_FOR\_REVIEW}). \textit{Twist:} the user corrects req\_2\_priority, out\_of\_scope mid-utterance (barge-in), which the agent must catch and repair.\\ \textbf{Agent.} 9 tool-calls; retrieval \emph{none}; finalize \emph{not finalized}; approval n/a.\\ \textbf{Outcome.} Workflow SUCCESS --- all gates pass; artifact field accuracy 10/10.}\end{passbox}
\begin{failbox}{apexv1\_095 --- Design review scribe --- Workflow FAILURE}{\scriptsize \textbf{Task.} Design review scribe --- facilitate archetype, design program manager, Software/SaaS. Controls: autonomy=prepare-only, knowledge burden=none, tool burden=light, risk=routine.\\ \textbf{Required.} produce the plan/checklist work product \texttt{dr\_1} with 10 required fields (e.g.\ feature=`checkout redesign'; option\_a=`Layout Aurora'; option\_b=`Layout Aurora Plus'; approved\_option=`Layout Aurora Plus'); reach the required terminal state (\texttt{READY\_FOR\_REVIEW}). \textit{Twist:} the user corrects approved\_option, rationale mid-utterance (barge-in), which the agent must catch and repair.\\ \textbf{Agent.} 9 tool-calls; retrieval \emph{none}; finalize \emph{not finalized}; approval n/a.\\ \textbf{Outcome.} Workflow FAILURE --- failed gate(s): TS, AV. workflow not finalized --- never submitted/committed; artifact incomplete: 8/10 fields correct (lifecycle<READY\_FOR\_REVIEW(got DRAFT)); wrong/missing: option\_a (got `Layout Aurora with a fluid grid structur' vs `Layout Aurora'), decision\_status (missing); 1 infra/WS drop(s).}\end{failbox}
\begin{failbox}{apexv1\_096 --- Incident postmortem facilitation --- Workflow FAILURE}{\scriptsize \textbf{Task.} Incident postmortem facilitation --- facilitate archetype, incident program manager, Software/SaaS. Controls: autonomy=prepare-only, knowledge burden=small-search retrieval, tool burden=moderate, risk=routine.\\ \textbf{Required.} produce the plan/checklist work product \texttt{pm\_1} with 10 required fields (e.g.\ incident\_id=`INC-42'; impact=`checkout degraded 40 minutes'; root\_cause=`bad deploy config'; event\_1\_time=`13:52'); retrieve the governing policy/record (knowledge retrieval via \texttt{kb\_search}); reach the required terminal state (\texttt{READY\_FOR\_REVIEW}). \textit{Twist:} the user corrects event\_1\_time, event\_order, action\_1\_owner mid-utterance (barge-in), which the agent must catch and repair.\\ \textbf{Agent.} 12 tool-calls; retrieval \emph{none}; finalize \emph{not finalized}; approval n/a.\\ \textbf{Outcome.} Workflow FAILURE --- failed gate(s): AC, AV. required knowledge retrieval not satisfied --- kb\_search never called; artifact incomplete: 9/10 fields correct; wrong/missing: event\_order (stale); 1 infra/WS drop(s).}\end{failbox}
\begin{failbox}{apexv1\_097 --- Vendor performance review meeting --- Workflow FAILURE}{\scriptsize \textbf{Task.} Vendor performance review meeting --- facilitate archetype, vendor manager, general enterprise. Controls: autonomy=prepare-only, knowledge burden=none, tool burden=light, risk=routine.\\ \textbf{Required.} produce the plan/checklist work product \texttt{vr\_1} with 10 required fields (e.g.\ vendor=`Cedar Supply'; sla\_met=`92 percent on-time'; quality\_score=`4 out of 5'; issue=`late deliveries in Q1'); reach the required terminal state (\texttt{READY\_FOR\_REVIEW}). \textit{Twist:} the user corrects commitment\_firm, vendor\_commitment mid-utterance (barge-in), which the agent must catch and repair.\\ \textbf{Agent.} 10 tool-calls; retrieval \emph{none}; finalize \emph{not finalized}; approval n/a.\\ \textbf{Outcome.} Workflow FAILURE --- failed gate(s): PV, AV. field(s) left stale after correction: vendor\_commitment; artifact incomplete: 10/10 fields correct; wrong/missing: vendor\_commitment (stale); 1 infra/WS drop(s).}\end{failbox}
\begin{passbox}{apexv1\_098 --- Launch readiness meeting --- Workflow SUCCESS}{\scriptsize \textbf{Task.} Launch readiness meeting --- facilitate archetype, launch program manager, Software/SaaS. Controls: autonomy=prepare-only, knowledge burden=none, tool burden=light, risk=routine.\\ \textbf{Required.} produce the plan/checklist work product \texttt{lr\_1} with 10 required fields (e.g.\ launch=`Payments v2'; dep\_infra=`infrastructure green'; dep\_security=`security review green'; dep\_qa=`QA blocked by a new test failure'); reach the required terminal state (\texttt{READY\_FOR\_REVIEW}). \textit{Twist:} the user corrects dep\_qa, readiness mid-utterance (barge-in), which the agent must catch and repair.\\ \textbf{Agent.} 9 tool-calls; retrieval \emph{none}; finalize \emph{not finalized}; approval n/a.\\ \textbf{Outcome.} Workflow SUCCESS --- all gates pass; artifact field accuracy 10/10.}\end{passbox}
\begin{passbox}{apexv1\_099 --- Stakeholder research synthesis meeting --- Workflow SUCCESS}{\scriptsize \textbf{Task.} Stakeholder research synthesis meeting --- facilitate archetype, research operations lead, professional services. Controls: autonomy=prepare-only, knowledge burden=none, tool burden=light, risk=routine.\\ \textbf{Required.} produce the plan/checklist work product \texttt{rs\_1} with 10 required fields (e.g.\ study=`onboarding research'; theme\_1=`users want faster setup'; theme\_1\_evidence=`8 of 10 interviews'; claim\_corrected=`adoption is 60 percent, corrected '); reach the required terminal state (\texttt{READY\_FOR\_REVIEW}). \textit{Twist:} the user corrects claim\_corrected, opinion\_vs\_policy mid-utterance (barge-in), which the agent must catch and repair.\\ \textbf{Agent.} 12 tool-calls; retrieval \emph{none}; finalize \emph{not finalized}; approval n/a.\\ \textbf{Outcome.} Workflow SUCCESS --- all gates pass; artifact field accuracy 10/10.}\end{passbox}
\begin{failbox}{apexv1\_100 --- Budget planning meeting record --- Workflow FAILURE}{\scriptsize \textbf{Task.} Budget planning meeting record --- facilitate archetype, finance business partner, general enterprise. Controls: autonomy=draft-and-confirm, knowledge burden=none, tool burden=light, risk=sensitive-data simulation.\\ \textbf{Required.} produce the plan/checklist work product \texttt{bud\_1} with 10 required fields (e.g.\ department=`Marketing'; proposed\_budget=`500000.0'; proposed\_cut=`reduce events line'; cut\_status=`withdrawn before end'); reach the required terminal state (\texttt{READY\_FOR\_REVIEW}). \textit{Twist:} the user corrects cut\_status, proposed\_cut, tooling\_basis mid-utterance (barge-in), which the agent must catch and repair.\\ \textbf{Agent.} 11 tool-calls; retrieval \emph{none}; finalize \emph{not finalized}; approval n/a.\\ \textbf{Outcome.} Workflow FAILURE --- failed gate(s): AV. artifact incomplete: 10/10 fields correct; wrong/missing: cut\_status (stale); 1 infra/WS drop(s).}\end{failbox}
\begin{failbox}{apexv1\_101 --- HVAC inspection to work order --- Workflow FAILURE}{\scriptsize \textbf{Task.} HVAC inspection to work order --- inspect archetype, field maintenance coordinator, manufacturing/field ops. Controls: autonomy=low-risk execute, knowledge burden=small-search retrieval, tool burden=moderate, risk=routine.\\ \textbf{Required.} produce the work order work product \texttt{wo\_1} with 11 required fields (e.g.\ asset\_id=`HVAC-7'; location=`Building C roof'; filter\_status=`clogged'; supply\_temp=`62'); retrieve the governing policy/record (knowledge retrieval via \texttt{kb\_search}); obtain user approval, then commit/submit. \textit{Twist:} the user corrects priority, supply\_temp, recommended\_action mid-utterance (barge-in), which the agent must catch and repair.\\ \textbf{Agent.} 12 tool-calls; retrieval \emph{none}; finalize \emph{not finalized}; approval \emph{not sought}.\\ \textbf{Outcome.} Workflow FAILURE --- failed gate(s): TS, AC, AV. required knowledge retrieval not satisfied --- kb\_search never called; workflow not finalized --- never submitted/committed; artifact incomplete: 9/11 fields correct (lifecycle<COMMITTED(got DRAFT)); wrong/missing: priority (got `High' vs `medium'), status (missing); 1 infra/WS drop(s).}\end{failbox}
\begin{failbox}{apexv1\_102 --- Safety pre-job verbal checklist --- Workflow FAILURE}{\scriptsize \textbf{Task.} Safety pre-job verbal checklist --- inspect archetype, site safety coordinator, manufacturing/field ops. Controls: autonomy=prepare-only, knowledge burden=none, tool burden=light, risk=special review.\\ \textbf{Required.} produce the work order work product \texttt{sc\_1} with 10 required fields (e.g.\ job\_id=`JOB-2201'; ppe\_check=`hard hat and gloves on'; lockout\_tagout=`applied'; area\_clear=`area clear of personnel'); reach the required terminal state (\texttt{READY\_FOR\_REVIEW}). \textit{Twist:} the user corrects all\_conditions\_met, gas\_check, readiness mid-utterance (barge-in), which the agent must catch and repair.\\ \textbf{Agent.} 12 tool-calls; retrieval \emph{none}; finalize \emph{not finalized}; approval n/a.\\ \textbf{Outcome.} Workflow FAILURE --- failed gate(s): TS, AV. workflow not finalized --- never submitted/committed; artifact incomplete: 9/10 fields correct (lifecycle<READY\_FOR\_REVIEW(got DRAFT)); wrong/missing: ppe\_check (got `yes' vs `hard hat and gloves on').}\end{failbox}
\begin{failbox}{apexv1\_103 --- Manufacturing quality inspection --- Workflow FAILURE}{\scriptsize \textbf{Task.} Manufacturing quality inspection --- inspect archetype, quality technician, manufacturing/field ops. Controls: autonomy=draft-and-confirm, knowledge burden=none, tool burden=light, risk=consequential action.\\ \textbf{Required.} produce the work order work product \texttt{qc\_1} with 10 required fields (e.g.\ batch\_id=`BATCH-559'; product=`bearing assembly'; dimension\_spec=`diameter 20mm plus or minus 0.1'; measured\_diameter=`20.15'); reach the required terminal state (\texttt{READY\_FOR\_REVIEW}). \textit{Twist:} the user corrects disposition, hold\_recommendation, in\_tolerance, measured\_diameter mid-utterance (barge-in), which the agent must catch and repair.\\ \textbf{Agent.} 12 tool-calls; retrieval \emph{none}; finalize \emph{not finalized}; approval n/a.\\ \textbf{Outcome.} Workflow FAILURE --- failed gate(s): AV. artifact incomplete: 7/10 fields correct; wrong/missing: in\_tolerance (got `yes' vs `no, diameter out of tolerance'), disposition (got `pass' vs `fail'), hold\_recommendation (got `no' vs `place batch on hold'); 1 infra/WS drop(s).}\end{failbox}
\begin{failbox}{apexv1\_104 --- Property condition inspection --- Workflow FAILURE}{\scriptsize \textbf{Task.} Property condition inspection --- inspect archetype, property operations inspector, general enterprise. Controls: autonomy=prepare-only, knowledge burden=none, tool burden=light, risk=routine.\\ \textbf{Required.} produce the work order work product \texttt{cr\_1} with 10 required fields (e.g.\ unit=`Apt 214'; living\_room=`good condition'; kitchen\_damage=`cracked countertop'; kitchen\_severity=`major'); reach the required terminal state (\texttt{READY\_FOR\_REVIEW}). \textit{Twist:} the user corrects deposit\_impact, kitchen\_damage, kitchen\_severity mid-utterance (barge-in), which the agent must catch and repair.\\ \textbf{Agent.} 12 tool-calls; retrieval \emph{none}; finalize \emph{not finalized}; approval n/a.\\ \textbf{Outcome.} Workflow FAILURE --- failed gate(s): AV. artifact incomplete: 8/10 fields correct; wrong/missing: kitchen\_severity (got `Minor.' vs `major'), deposit\_impact (got `Minor deduction expected; mostly normal ' vs `significant deduction'); 1 infra/WS drop(s).}\end{failbox}
\begin{failbox}{apexv1\_105 --- Warehouse inventory spot audit --- Workflow FAILURE}{\scriptsize \textbf{Task.} Warehouse inventory spot audit --- inspect archetype, inventory auditor, manufacturing/field ops. Controls: autonomy=prepare-only, knowledge burden=none, tool burden=light, risk=routine.\\ \textbf{Required.} produce the work order work product \texttt{ia\_1} with 10 required fields (e.g.\ location=`Aisle 7 Bin B'; sku=`SKU-4417'; sku\_description=`label rolls'; system\_count=`120'); reach the required terminal state (\texttt{READY\_FOR\_REVIEW}). \textit{Twist:} the user corrects sku, sku\_description mid-utterance (barge-in), which the agent must catch and repair.\\ \textbf{Agent.} 11 tool-calls; retrieval \emph{none}; finalize \emph{not finalized}; approval n/a.\\ \textbf{Outcome.} Workflow FAILURE --- failed gate(s): AV. artifact incomplete: 8/10 fields correct; wrong/missing: sku\_description (stale), discrepancy (got `-8' vs `8'); 1 infra/WS drop(s).}\end{failbox}
\begin{passbox}{apexv1\_106 --- Fleet vehicle pre-service inspection --- Workflow SUCCESS}{\scriptsize \textbf{Task.} Fleet vehicle pre-service inspection --- inspect archetype, fleet maintenance coordinator, manufacturing/field ops. Controls: autonomy=prepare-only, knowledge burden=none, tool burden=light, risk=routine.\\ \textbf{Required.} produce the work order work product \texttt{fi\_1} with 10 required fields (e.g.\ vehicle\_id=`VAN-33'; odometer=`88000'; tire\_condition=`front tires worn'; brake\_condition=`pads at 40 percent'); reach the required terminal state (\texttt{READY\_FOR\_REVIEW}). \textit{Twist:} the user corrects odometer, priority, warning\_light mid-utterance (barge-in), which the agent must catch and repair.\\ \textbf{Agent.} 13 tool-calls; retrieval \emph{none}; finalize \emph{not finalized}; approval n/a.\\ \textbf{Outcome.} Workflow SUCCESS --- all gates pass; artifact field accuracy 10/10.}\end{passbox}
\begin{failbox}{apexv1\_107 --- Data-center rack inspection --- Workflow FAILURE}{\scriptsize \textbf{Task.} Data-center rack inspection --- inspect archetype, data-center operations technician, Software/SaaS. Controls: autonomy=prepare-only, knowledge burden=small-search retrieval, tool burden=moderate, risk=routine.\\ \textbf{Required.} produce the work order work product \texttt{ri\_1} with 10 required fields (e.g.\ rack\_id=`RACK-91'; temperature=`24'; power\_draw=`within normal'; fan\_status=`fan alert on unit 3'); retrieve the governing policy/record (knowledge retrieval via \texttt{kb\_search}); reach the required terminal state (\texttt{READY\_FOR\_REVIEW}). \textit{Twist:} the user corrects fan\_status, rack\_id mid-utterance (barge-in), which the agent must catch and repair.\\ \textbf{Agent.} 13 tool-calls; retrieval \emph{none}; finalize \emph{not finalized}; approval n/a.\\ \textbf{Outcome.} Workflow FAILURE --- failed gate(s): AC. required knowledge retrieval not satisfied --- kb\_search never called; 1 infra/WS drop(s).}\end{failbox}
\begin{failbox}{apexv1\_108 --- Retail store opening checklist --- Workflow FAILURE}{\scriptsize \textbf{Task.} Retail store opening checklist --- inspect archetype, store operations lead, general enterprise. Controls: autonomy=prepare-only, knowledge burden=none, tool burden=light, risk=routine.\\ \textbf{Required.} produce the work order work product \texttt{oc\_1} with 10 required fields (e.g.\ store\_id=`ST-142'; alarm\_disarmed=`yes'; lights\_on=`yes'; registers\_ready=`yes'); reach the required terminal state (\texttt{READY\_FOR\_REVIEW}). \textit{Twist:} the user corrects cash\_drawer, cash\_drawer\_ok, opening\_status mid-utterance (barge-in), which the agent must catch and repair.\\ \textbf{Agent.} 11 tool-calls; retrieval \emph{none}; finalize \emph{not finalized}; approval n/a.\\ \textbf{Outcome.} Workflow FAILURE --- failed gate(s): PV, AV. field(s) left stale after correction: cash\_drawer; artifact incomplete: 9/10 fields correct; wrong/missing: cash\_drawer (stale); 1 infra/WS drop(s).}\end{failbox}
\begin{failbox}{apexv1\_109 --- Solar-site maintenance inspection --- Workflow FAILURE}{\scriptsize \textbf{Task.} Solar-site maintenance inspection --- inspect archetype, field service technician, manufacturing/field ops. Controls: autonomy=low-risk execute, knowledge burden=small-search retrieval, tool burden=moderate, risk=routine.\\ \textbf{Required.} produce the work order work product \texttt{si\_1} with 10 required fields (e.g.\ site\_id=`SOLAR-4'; inverter\_id=`INV-2'; inverter\_output=`output 8 percent low'; panel\_condition=`some soiling'); retrieve the governing policy/record (knowledge retrieval via \texttt{kb\_search}); reach the required terminal state (\texttt{READY\_FOR\_REVIEW}). \textit{Twist:} the user corrects inverter\_id, resume\_note mid-utterance (barge-in), which the agent must catch and repair.\\ \textbf{Agent.} 12 tool-calls; retrieval \emph{none}; finalize \emph{not finalized}; approval n/a.\\ \textbf{Outcome.} Workflow FAILURE --- failed gate(s): AC. required knowledge retrieval not satisfied --- kb\_search never called; 1 infra/WS drop(s).}\end{failbox}
\begin{failbox}{apexv1\_110 --- Inspection to work-order and customer summary --- Workflow FAILURE}{\scriptsize \textbf{Task.} Inspection to work-order and customer summary --- inspect archetype, field service coordinator, manufacturing/field ops. Controls: autonomy=approval-gated commit, knowledge burden=small-search retrieval, tool burden=moderate, risk=consequential action.\\ \textbf{Required.} produce the work order work product \texttt{wo\_1} with 11 required fields (e.g.\ asset\_id=`CHILLER-3'; symptom=`intermittent shutdown'; finding=`loose sensor connector'; initial\_recommendation=`no replacement needed'); retrieve the governing policy/record (knowledge retrieval via \texttt{kb\_search}); obtain user approval, then commit/submit. \textit{Twist:} the user corrects initial\_recommendation, part\_needed, revised\_recommendation mid-utterance (barge-in), which the agent must catch and repair.\\ \textbf{Agent.} 16 tool-calls; retrieval \emph{none}; finalize committed; approval sought.\\ \textbf{Outcome.} Workflow FAILURE --- failed gate(s): PV, AC, AV. required knowledge retrieval not satisfied --- kb\_search never called; field(s) left stale after correction: initial\_recommendation; artifact incomplete: 10/11 fields correct; wrong/missing: initial\_recommendation (stale).}\end{failbox}
\begin{failbox}{apexv1\_111 --- SaaS outage triage coordination --- Workflow FAILURE}{\scriptsize \textbf{Task.} SaaS outage triage coordination --- coordinate archetype, incident commander assistant, Software/SaaS. Controls: autonomy=low-risk execute, knowledge burden=small-search retrieval, tool burden=moderate, risk=routine.\\ \textbf{Required.} produce the timeline work product \texttt{inc\_1} with 10 required fields (e.g.\ incident\_id=`OPS-19'; current\_impact=`API 5xx errors'; suspected\_service=`cart service'; blast\_radius=`US and EU regions'); retrieve the governing policy/record (knowledge retrieval via \texttt{kb\_search}); obtain user approval, then commit/submit. \textit{Twist:} the user corrects suspected\_service, blast\_radius, severity mid-utterance (barge-in), which the agent must catch and repair.\\ \textbf{Agent.} 13 tool-calls; retrieval \emph{none}; finalize committed; approval sought.\\ \textbf{Outcome.} Workflow FAILURE --- failed gate(s): AC. required knowledge retrieval not satisfied --- kb\_search never called; 1 infra/WS drop(s).}\end{failbox}
\begin{failbox}{apexv1\_112 --- Access anomaly escalation coordination --- Workflow FAILURE}{\scriptsize \textbf{Task.} Access anomaly escalation coordination --- coordinate archetype, security operations coordinator, Software/SaaS. Controls: autonomy=low-risk execute, knowledge burden=multi-document reasoning, tool burden=moderate, risk=special review.\\ \textbf{Required.} produce the timeline work product \texttt{inc\_1} with 10 required fields (e.g.\ user\_account=`acct\_5521'; reported\_claim=`user says account compromised'; log\_evidence=`anomalous login from new location'; established\_fact=`anomalous login only, compromise n'); retrieve the governing policy/record (knowledge retrieval via \texttt{kb\_search}); obtain user approval, then commit/submit. \textit{Twist:} the user corrects established\_fact, reported\_claim mid-utterance (barge-in), which the agent must catch and repair.\\ \textbf{Agent.} 12 tool-calls; retrieval \emph{none}; finalize committed; approval sought.\\ \textbf{Outcome.} Workflow FAILURE --- failed gate(s): AC. required knowledge retrieval not satisfied --- kb\_search never called; 1 infra/WS drop(s).}\end{failbox}
\begin{failbox}{apexv1\_113 --- Warehouse shipment-delay incident --- Workflow FAILURE}{\scriptsize \textbf{Task.} Warehouse shipment-delay incident --- coordinate archetype, operations incident coordinator, manufacturing/field ops. Controls: autonomy=draft-and-confirm, knowledge burden=small-search retrieval, tool burden=moderate, risk=routine.\\ \textbf{Required.} produce the timeline work product \texttt{inc\_1} with 10 required fields (e.g.\ shipment\_id=`SHP-3300'; original\_eta=`Tuesday 6am'; carrier\_eta=`Tuesday 1pm'; customer\_cutoff=`Tuesday 3pm, cannot move'); retrieve the governing policy/record (knowledge retrieval via \texttt{kb\_search}); reach the required terminal state (\texttt{READY\_FOR\_REVIEW}). \textit{Twist:} the user corrects carrier\_eta, risk, recovery\_plan mid-utterance (barge-in), which the agent must catch and repair.\\ \textbf{Agent.} 11 tool-calls; retrieval \emph{none}; finalize \emph{not finalized}; approval n/a.\\ \textbf{Outcome.} Workflow FAILURE --- failed gate(s): AC, AV. required knowledge retrieval not satisfied --- kb\_search never called; artifact incomplete: 9/10 fields correct; wrong/missing: shipment\_id (got `SHP-30300' vs `SHP-3300'); 1 infra/WS drop(s).}\end{failbox}
\begin{failbox}{apexv1\_114 --- Customer data-sync incident escalation --- Workflow FAILURE}{\scriptsize \textbf{Task.} Customer data-sync incident escalation --- coordinate archetype, support incident coordinator, Software/SaaS. Controls: autonomy=low-risk execute, knowledge burden=small-search retrieval, tool burden=moderate, risk=routine.\\ \textbf{Required.} produce the timeline work product \texttt{inc\_1} with 11 required fields (e.g.\ account=`Delta Systems'; reported\_scope=`customer says all records affected'; telemetry\_scope=`telemetry shows one region only'; established\_scope=`one region confirmed by telemetry,'); retrieve the governing policy/record (knowledge retrieval via \texttt{kb\_search}); obtain user approval, then commit/submit. \textit{Twist:} the user corrects established\_scope, reported\_scope mid-utterance (barge-in), which the agent must catch and repair.\\ \textbf{Agent.} 15 tool-calls; retrieval \emph{none}; finalize committed; approval sought.\\ \textbf{Outcome.} Workflow FAILURE --- failed gate(s): PV, AC. required knowledge retrieval not satisfied --- kb\_search never called; process-compliance/critical gate violated (no\_unauthorized\_commit, artifact\_not\_stale); tool-call failure: submit\_inc\_1; 1 infra/WS drop(s).}\end{failbox}
\begin{failbox}{apexv1\_115 --- Supply-chain component shortage response --- Workflow FAILURE}{\scriptsize \textbf{Task.} Supply-chain component shortage response --- coordinate archetype, supply-chain coordinator, manufacturing/field ops. Controls: autonomy=draft-and-confirm, knowledge burden=small-search retrieval, tool burden=moderate, risk=routine.\\ \textbf{Required.} produce the timeline work product \texttt{inc\_1} with 10 required fields (e.g.\ component=`power module PM-9'; shortage\_qty=`500'; supplier\_available=`200'; production\_priority=`Line A over Line B'); retrieve the governing policy/record (knowledge retrieval via \texttt{kb\_search}); reach the required terminal state (\texttt{READY\_FOR\_REVIEW}). \textit{Twist:} the user corrects allocation, recovery\_plan, supplier\_available mid-utterance (barge-in), which the agent must catch and repair.\\ \textbf{Agent.} 12 tool-calls; retrieval \emph{none}; finalize \emph{not finalized}; approval n/a.\\ \textbf{Outcome.} Workflow FAILURE --- failed gate(s): AC, AV. required knowledge retrieval not satisfied --- kb\_search never called; artifact incomplete: 9/10 fields correct; wrong/missing: allocation (got `300 to Line A; remainder to Line B' vs `200 to Line A first'); 1 infra/WS drop(s).}\end{failbox}
\begin{failbox}{apexv1\_116 --- Facilities water-leak escalation --- Workflow FAILURE}{\scriptsize \textbf{Task.} Facilities water-leak escalation --- coordinate archetype, facilities incident coordinator, general enterprise. Controls: autonomy=approval-gated commit, knowledge burden=supplied evidence, tool burden=moderate, risk=special review.\\ \textbf{Required.} produce the timeline work product \texttt{inc\_1} with 11 required fields (e.g.\ location=`3rd floor east'; leak\_source=`burst pipe above ceiling'; initial\_priority=`standard cleanup'; electrical\_exposure=`water near a live panel'); retrieve the governing policy/record (knowledge retrieval via \texttt{kb\_search}); obtain user approval, then commit/submit. \textit{Twist:} the user corrects electrical\_exposure, escalated\_priority, initial\_priority mid-utterance (barge-in), which the agent must catch and repair.\\ \textbf{Agent.} 12 tool-calls; retrieval done; finalize committed; approval sought.\\ \textbf{Outcome.} Workflow FAILURE --- failed gate(s): AV. artifact incomplete: 10/11 fields correct; wrong/missing: initial\_priority (got `Emergency' vs `standard cleanup'); 1 infra/WS drop(s).}\end{failbox}
\begin{failbox}{apexv1\_117 --- Payment-processing outage coordination --- Workflow FAILURE}{\scriptsize \textbf{Task.} Payment-processing outage coordination --- coordinate archetype, payments operations incident coordinator, Software/SaaS. Controls: autonomy=draft-and-confirm, knowledge burden=small-search retrieval, tool burden=moderate, risk=consequential action.\\ \textbf{Required.} produce the timeline work product \texttt{inc\_1} with 10 required fields (e.g.\ incident\_id=`PAY-88'; rail\_card=`card rail degraded'; rail\_ach=`ACH rail recovered'; overall\_status=`partial recovery, card still degra'); retrieve the governing policy/record (knowledge retrieval via \texttt{kb\_search}); reach the required terminal state (\texttt{READY\_FOR\_REVIEW}). \textit{Twist:} the user corrects overall\_status, rail\_ach, status\_wording mid-utterance (barge-in), which the agent must catch and repair.\\ \textbf{Agent.} 11 tool-calls; retrieval \emph{none}; finalize \emph{not finalized}; approval n/a.\\ \textbf{Outcome.} Workflow FAILURE --- failed gate(s): TS, AC, AV. required knowledge retrieval not satisfied --- kb\_search never called; workflow not finalized --- never submitted/committed; artifact incomplete: 9/10 fields correct (lifecycle<READY\_FOR\_REVIEW(got DRAFT)); wrong/missing: incident\_id (got `PAY-908' vs `PAY-88'), status\_wording (stale).}\end{failbox}
\begin{failbox}{apexv1\_118 --- Production quality hold coordination --- Workflow FAILURE}{\scriptsize \textbf{Task.} Production quality hold coordination --- coordinate archetype, manufacturing incident coordinator, manufacturing/field ops. Controls: autonomy=approval-gated commit, knowledge burden=small-search retrieval, tool burden=moderate, risk=consequential action.\\ \textbf{Required.} produce the timeline work product \texttt{inc\_1} with 11 required fields (e.g.\ incident\_id=`QH-14'; defect=`coating adhesion failure'; initial\_hold\_scope=`all lots this week'; test\_result=`only lots 5 to 8 affected'); retrieve the governing policy/record (knowledge retrieval via \texttt{kb\_search}); obtain user approval, then commit/submit. \textit{Twist:} the user corrects disposition, revised\_hold\_scope, test\_result mid-utterance (barge-in), which the agent must catch and repair.\\ \textbf{Agent.} 10 tool-calls; retrieval \emph{none}; finalize \emph{not finalized}; approval \emph{not sought}.\\ \textbf{Outcome.} Workflow FAILURE --- failed gate(s): TS, AC, AV. required knowledge retrieval not satisfied --- kb\_search never called; workflow not finalized --- never submitted/committed; artifact incomplete: 9/11 fields correct (lifecycle<COMMITTED(got DRAFT)); wrong/missing: initial\_hold\_scope (got `Lots 5 through 8' vs `all lots this week'), status (missing); 1 infra/WS drop(s).}\end{failbox}
\begin{passbox}{apexv1\_119 --- Live event AV failure coordination --- Workflow SUCCESS}{\scriptsize \textbf{Task.} Live event AV failure coordination --- coordinate archetype, event operations coordinator, general enterprise. Controls: autonomy=draft-and-confirm, knowledge burden=none, tool burden=light, risk=routine.\\ \textbf{Required.} produce the timeline work product \texttt{inc\_1} with 10 required fields (e.g.\ event=`keynote session'; failure=`main room projector and audio down'; backup\_room=`Room B available'; backup\_capacity=`180'); reach the required terminal state (\texttt{READY\_FOR\_REVIEW}). \textit{Twist:} the user corrects vip\_constraint, vip\_handling mid-utterance (barge-in), which the agent must catch and repair.\\ \textbf{Agent.} 13 tool-calls; retrieval \emph{none}; finalize \emph{not finalized}; approval n/a.\\ \textbf{Outcome.} Workflow SUCCESS --- all gates pass; artifact field accuracy 10/10.}\end{passbox}
\begin{failbox}{apexv1\_120 --- Near-miss incident escalation and follow-up --- Workflow FAILURE}{\scriptsize \textbf{Task.} Near-miss incident escalation and follow-up --- coordinate archetype, safety operations coordinator, manufacturing/field ops. Controls: autonomy=low-risk execute, knowledge burden=small-search retrieval, tool burden=moderate, risk=special review.\\ \textbf{Required.} produce the timeline work product \texttt{inc\_1} with 11 required fields (e.g.\ incident\_id=`NM-77'; equipment\_id=`PRESS-7'; event=`guard bypass near-miss'; injury=`no injury'); retrieve the governing policy/record (knowledge retrieval via \texttt{kb\_search}); obtain user approval, then commit/submit. \textit{Twist:} the user corrects equipment\_id, corrected\_cause, draft\_cause mid-utterance (barge-in), which the agent must catch and repair.\\ \textbf{Agent.} 10 tool-calls; retrieval done; finalize \emph{not finalized}; approval \emph{not sought}.\\ \textbf{Outcome.} Workflow FAILURE --- failed gate(s): TS, AV. workflow not finalized --- never submitted/committed; artifact incomplete: 10/11 fields correct (lifecycle<COMMITTED(got DRAFT)); wrong/missing: corrected\_cause (stale), status (missing); 1 infra/WS drop(s).}\end{failbox}
\clearpage
\subsection*{Grok-Voice-Think-2.0}
\begin{failbox}{apexv1\_001 --- Benefits enrollment with dependent correction --- Workflow FAILURE}{\scriptsize \textbf{Task.} Benefits enrollment with dependent correction --- form-fill archetype, benefits specialist, workplace/HR. Controls: autonomy=approval-gated commit, knowledge burden=small-search retrieval, tool burden=moderate, risk=consequential action.\\ \textbf{Required.} produce the structured form work product \texttt{enr\_1} with 14 required fields (e.g.\ employee\_id=`E4471'; legal\_name=`Morgan Reyes'; date\_of\_birth=`1988-07-09'; medical\_plan=`HDHP'); retrieve the governing policy/record (knowledge retrieval via \texttt{kb\_search}); obtain user approval, then commit/submit. \textit{Twist:} the user corrects dependent\_count, dependent\_names, medical\_plan, monthly\_premium mid-utterance (barge-in), which the agent must catch and repair.\\ \textbf{Agent.} 31 tool-calls; retrieval done; finalize committed; approval sought.\\ \textbf{Outcome.} Workflow FAILURE --- failed gate(s): PV, AC, AV. required knowledge retrieval not satisfied --- searched but gold document not retrieved; committed/submitted without seeking approval; artifact incomplete: 12/14 fields correct; wrong/missing: dependent\_names (got `daughter' vs `Jamie Reyes and Casey Reyes'), monthly\_premium (got `360' vs `225.0'); tool-call failure: submit\_enr\_1.}\end{failbox}
\begin{failbox}{apexv1\_002 --- Expense report from receipts and spoken narrative --- Workflow FAILURE}{\scriptsize \textbf{Task.} Expense report from receipts and spoken narrative --- form-fill archetype, finance operations specialist, general enterprise. Controls: autonomy=approval-gated commit, knowledge burden=supplied evidence, tool burden=moderate, risk=consequential action.\\ \textbf{Required.} produce the structured form work product \texttt{exp\_1} with 11 required fields (e.g.\ employee\_id=`E9910'; report\_period=`March 3 to March 6'; purpose=`client onsite in Denver'; airfare=`410.0'); retrieve the governing policy/record (knowledge retrieval via \texttt{kb\_search}); obtain user approval, then commit/submit. \textit{Twist:} the user corrects hotel, total, cost\_center mid-utterance (barge-in), which the agent must catch and repair.\\ \textbf{Agent.} 20 tool-calls; retrieval done; finalize committed; approval sought.\\ \textbf{Outcome.} Workflow FAILURE --- failed gate(s): AV. artifact incomplete: 10/11 fields correct; wrong/missing: total (got `1145' vs `1085.0').}\end{failbox}
\begin{failbox}{apexv1\_003 --- New vendor onboarding packet --- Workflow FAILURE}{\scriptsize \textbf{Task.} New vendor onboarding packet --- form-fill archetype, procurement operations specialist, general enterprise. Controls: autonomy=draft-and-confirm, knowledge burden=small-search retrieval, tool burden=moderate, risk=sensitive-data simulation.\\ \textbf{Required.} produce the structured form work product \texttt{ven\_1} with 10 required fields (e.g.\ vendor\_name=`Cedar Works LLC'; tax\_id=`88-4412290'; address=`72 Mill Road, Suite 4'; remittance\_email=`billing@cedarworks.example'); retrieve the governing policy/record (knowledge retrieval via \texttt{kb\_search}); reach the required terminal state (\texttt{READY\_FOR\_REVIEW}). \textit{Twist:} the user corrects remittance\_email, account\_number mid-utterance (barge-in), which the agent must catch and repair.\\ \textbf{Agent.} 14 tool-calls; retrieval done; finalize \emph{not finalized}; approval n/a.\\ \textbf{Outcome.} Workflow FAILURE --- failed gate(s): AC. required knowledge retrieval not satisfied --- searched but gold document not retrieved.}\end{failbox}
\begin{failbox}{apexv1\_004 --- Business travel approval request --- Workflow FAILURE}{\scriptsize \textbf{Task.} Business travel approval request --- form-fill archetype, travel coordinator, general enterprise. Controls: autonomy=draft-and-confirm, knowledge burden=small-search retrieval, tool burden=moderate, risk=routine.\\ \textbf{Required.} produce the structured form work product \texttt{trv\_1} with 10 required fields (e.g.\ traveler=`Priya Nair'; destination=`Austin'; purpose=`customer quarterly review'; meeting\_date=`2026-04-15'); retrieve the governing policy/record (knowledge retrieval via \texttt{kb\_search}); reach the required terminal state (\texttt{READY\_FOR\_REVIEW}). \textit{Twist:} the user corrects depart\_date, meeting\_date, return\_date mid-utterance (barge-in), which the agent must catch and repair.\\ \textbf{Agent.} 23 tool-calls; retrieval done; finalize \emph{not finalized}; approval n/a.\\ \textbf{Outcome.} Workflow FAILURE --- failed gate(s): AV. artifact incomplete: 8/10 fields correct; wrong/missing: depart\_date (got `April 13th, 2026' vs `2026-04-14'), return\_date (got `April 15th, 2026' vs `2026-04-16').}\end{failbox}
\begin{failbox}{apexv1\_005 --- Privileged software-access request --- Workflow FAILURE}{\scriptsize \textbf{Task.} Privileged software-access request --- form-fill archetype, IT access coordinator, Software/SaaS. Controls: autonomy=approval-gated commit, knowledge burden=multi-document reasoning, tool burden=moderate, risk=consequential action.\\ \textbf{Required.} produce the structured form work product \texttt{acc\_1} with 10 required fields (e.g.\ requester=`Sam Okafor'; project=`Q2 revenue analytics'; requested\_access=`analytics read-only'; justified\_role=`AnalyticsViewer'); retrieve the governing policy/record (knowledge retrieval via \texttt{kb\_search}); obtain user approval, then commit/submit. \textit{Twist:} the user corrects requested\_access, requested\_access, duration mid-utterance (barge-in), which the agent must catch and repair.\\ \textbf{Agent.} 30 tool-calls; retrieval done; finalize committed; approval sought.\\ \textbf{Outcome.} Workflow FAILURE --- failed gate(s): PV, AV. committed/submitted without seeking approval; artifact incomplete: 9/10 fields correct; wrong/missing: duration (got `90 days' vs `60 days'); tool-call failure: submit\_acc\_1.}\end{failbox}
\begin{failbox}{apexv1\_006 --- Warranty claim application --- Workflow FAILURE}{\scriptsize \textbf{Task.} Warranty claim application --- form-fill archetype, warranty operations specialist, manufacturing/field ops. Controls: autonomy=draft-and-confirm, knowledge burden=small-search retrieval, tool burden=moderate, risk=routine.\\ \textbf{Required.} produce the structured form work product \texttt{war\_1} with 9 required fields (e.g.\ customer\_name=`Robin Vale'; product\_model=`TurboMix 500'; serial\_number=`TMX500-88231'; purchase\_date=`2025-11-02'); retrieve the governing policy/record (knowledge retrieval via \texttt{kb\_search}); reach the required terminal state (\texttt{READY\_FOR\_REVIEW}). \textit{Twist:} the user corrects serial\_number, warranty\_policy mid-utterance (barge-in), which the agent must catch and repair.\\ \textbf{Agent.} 13 tool-calls; retrieval done; finalize \emph{not finalized}; approval n/a.\\ \textbf{Outcome.} Workflow FAILURE --- failed gate(s): AV. artifact incomplete: 8/9 fields correct; wrong/missing: contact\_phone (got `550-5173' vs `555-0173').}\end{failbox}
\begin{failbox}{apexv1\_007 --- Parental-leave administration packet --- Workflow FAILURE}{\scriptsize \textbf{Task.} Parental-leave administration packet --- form-fill archetype, HR operations specialist, workplace/HR. Controls: autonomy=draft-and-confirm, knowledge burden=multi-document reasoning, tool burden=moderate, risk=sensitive-data simulation.\\ \textbf{Required.} produce the structured form work product \texttt{lev\_1} with 9 required fields (e.g.\ employee\_id=`E3320'; leave\_type=`parental leave'; leave\_start=`2026-05-01'; leave\_end=`2026-07-31'); retrieve the governing policy/record (knowledge retrieval via \texttt{kb\_search}); reach the required terminal state (\texttt{READY\_FOR\_REVIEW}). \textit{Twist:} the user corrects fmla\_weeks, leave\_end mid-utterance (barge-in), which the agent must catch and repair.\\ \textbf{Agent.} 24 tool-calls; retrieval done; finalize \emph{not finalized}; approval n/a.\\ \textbf{Outcome.} Workflow FAILURE --- failed gate(s): AV. artifact incomplete: 8/9 fields correct; wrong/missing: fmla\_weeks (got `12' vs `13').}\end{failbox}
\begin{failbox}{apexv1\_008 --- Customer account setup and billing profile --- Workflow FAILURE}{\scriptsize \textbf{Task.} Customer account setup and billing profile --- form-fill archetype, account operations specialist, Software/SaaS. Controls: autonomy=draft-and-confirm, knowledge burden=supplied evidence, tool burden=moderate, risk=sensitive-data simulation.\\ \textbf{Required.} produce the structured form work product \texttt{acct\_1} with 9 required fields (e.g.\ company=`Northwind Retail'; billing\_contact=`Ada Lin'; technical\_contact=`Ben Cho'; billing\_country=`Germany'); retrieve the governing policy/record (knowledge retrieval via \texttt{kb\_search}); reach the required terminal state (\texttt{READY\_FOR\_REVIEW}). \textit{Twist:} the user corrects billing\_country, tax\_id, tax\_rate mid-utterance (barge-in), which the agent must catch and repair.\\ \textbf{Agent.} 16 tool-calls; retrieval done; finalize \emph{not finalized}; approval n/a.\\ \textbf{Outcome.} Workflow FAILURE --- failed gate(s): AV. artifact incomplete: 7/9 fields correct; wrong/missing: tax\_id (got `IE1234567XA' vs `DE811234567'), tax\_rate (got `23\%' vs `19.0').}\end{failbox}
\begin{failbox}{apexv1\_009 --- Conference reimbursement packet --- Workflow FAILURE}{\scriptsize \textbf{Task.} Conference reimbursement packet --- form-fill archetype, operations coordinator, professional services. Controls: autonomy=draft-and-confirm, knowledge burden=supplied evidence, tool burden=moderate, risk=routine.\\ \textbf{Required.} produce the structured form work product \texttt{rmb\_1} with 9 required fields (e.g.\ attendee=`Noa Grant'; conference=`DataCon 2026'; registration\_fee=`300.0'; workshop\_fee=`175.0'); retrieve the governing policy/record (knowledge retrieval via \texttt{kb\_search}); reach the required terminal state (\texttt{READY\_FOR\_REVIEW}). \textit{Twist:} the user corrects total, workshop\_fee mid-utterance (barge-in), which the agent must catch and repair.\\ \textbf{Agent.} 20 tool-calls; retrieval done; finalize \emph{not finalized}; approval n/a.\\ \textbf{Outcome.} Workflow FAILURE --- failed gate(s): AC, AV. required knowledge retrieval not satisfied --- searched but gold document not retrieved; artifact incomplete: 6/9 fields correct; wrong/missing: attendee (got `caller' vs `Noa Grant'), workshop\_approver (got `Noah Grant' vs `Manager Kim'), total (got `730' vs `755.0').}\end{failbox}
\begin{passbox}{apexv1\_010 --- Facility access badge request --- Workflow SUCCESS}{\scriptsize \textbf{Task.} Facility access badge request --- form-fill archetype, facilities coordinator, general enterprise. Controls: autonomy=approval-gated commit, knowledge burden=small-search retrieval, tool burden=moderate, risk=consequential action.\\ \textbf{Required.} produce the structured form work product \texttt{bdg\_1} with 10 required fields (e.g.\ contractor\_name=`Rowan Tate'; company=`BrightHVAC'; sponsor=`Facilities lead Dana'; access\_zones=`mechanical rooms'); retrieve the governing policy/record (knowledge retrieval via \texttt{kb\_search}); obtain user approval, then commit/submit. \textit{Twist:} the user corrects requested\_hours, requested\_hours mid-utterance (barge-in), which the agent must catch and repair.\\ \textbf{Agent.} 14 tool-calls; retrieval done; finalize committed; approval sought.\\ \textbf{Outcome.} Workflow SUCCESS --- all gates pass; artifact field accuracy 10/10.}\end{passbox}
\begin{passbox}{apexv1\_011 --- Software engineer recruiter screen --- Workflow SUCCESS}{\scriptsize \textbf{Task.} Software engineer recruiter screen --- interview archetype, recruiter, Software/SaaS. Controls: autonomy=prepare-only, knowledge burden=none, tool burden=light, risk=sensitive-data simulation.\\ \textbf{Required.} produce the evidence matrix work product \texttt{rec\_1} with 13 required fields (e.g.\ years\_experience=`8'; primary\_language=`Python'; system\_design\_example=`designed a multi-region ingestion '; scale\_metric=`half a million daily events'); reach the required terminal state (\texttt{READY\_FOR\_REVIEW}). \textit{Twist:} the user corrects team\_size, scale\_metric mid-utterance (barge-in), which the agent must catch and repair.\\ \textbf{Agent.} 24 tool-calls; retrieval \emph{none}; finalize \emph{not finalized}; approval n/a.\\ \textbf{Outcome.} Workflow SUCCESS --- all gates pass; artifact field accuracy 13/13.}\end{passbox}
\begin{failbox}{apexv1\_012 --- Customer-success manager recruiter screen --- Workflow FAILURE}{\scriptsize \textbf{Task.} Customer-success manager recruiter screen --- interview archetype, recruiter, Software/SaaS. Controls: autonomy=prepare-only, knowledge burden=none, tool burden=light, risk=sensitive-data simulation.\\ \textbf{Required.} produce the evidence matrix work product \texttt{rec\_1} with 12 required fields (e.g.\ years\_experience=`6'; book\_of\_business=`20 enterprise accounts'; retention\_metric=`92 percent gross retention'; customer\_save\_example=`recovered a churning key account'); reach the required terminal state (\texttt{READY\_FOR\_REVIEW}). \textit{Twist:} the user corrects retention\_metric, open\_gap mid-utterance (barge-in), which the agent must catch and repair.\\ \textbf{Agent.} 20 tool-calls; retrieval \emph{none}; finalize \emph{not finalized}; approval n/a.\\ \textbf{Outcome.} Workflow FAILURE --- failed gate(s): AV. artifact incomplete: 11/12 fields correct; wrong/missing: open\_gap (got `no direct quota-carrying experience, but' vs `quota-carrying confirmed via renewals').}\end{failbox}
\begin{failbox}{apexv1\_013 --- Warehouse supervisor screen --- Workflow FAILURE}{\scriptsize \textbf{Task.} Warehouse supervisor screen --- interview archetype, recruiter, manufacturing/field ops. Controls: autonomy=prepare-only, knowledge burden=none, tool burden=light, risk=sensitive-data simulation.\\ \textbf{Required.} produce the evidence matrix work product \texttt{rec\_1} with 12 required fields (e.g.\ years\_experience=`10'; team\_size=`25'; shift\_scheduling=`built rotating three-shift coverag'; safety\_record=`300 days incident-free'); reach the required terminal state (\texttt{READY\_FOR\_REVIEW}). \textit{Twist:} the user corrects current\_start\_year, throughput\_metric mid-utterance (barge-in), which the agent must catch and repair.\\ \textbf{Agent.} 16 tool-calls; retrieval \emph{none}; finalize \emph{not finalized}; approval n/a.\\ \textbf{Outcome.} Workflow FAILURE --- failed gate(s): AV. artifact incomplete: 11/12 fields correct; wrong/missing: conflict\_example (got `resolved team dispute over shift assignm' vs `de-escalated a union grievance').}\end{failbox}
\begin{failbox}{apexv1\_014 --- Internal transfer evidence interview --- Workflow FAILURE}{\scriptsize \textbf{Task.} Internal transfer evidence interview --- interview archetype, HR business partner, general enterprise. Controls: autonomy=prepare-only, knowledge burden=supplied evidence, tool burden=moderate, risk=sensitive-data simulation.\\ \textbf{Required.} produce the evidence matrix work product \texttt{rec\_1} with 12 required fields (e.g.\ current\_role=`senior analyst'; target\_team=`platform reliability'; project\_apollo=`led the payments platform migratio'; transferable\_skill=`incident command'); retrieve the governing policy/record (knowledge retrieval via \texttt{kb\_search}); reach the required terminal state (\texttt{READY\_FOR\_REVIEW}). \textit{Twist:} the user corrects project\_apollo, impact\_metric mid-utterance (barge-in), which the agent must catch and repair.\\ \textbf{Agent.} 20 tool-calls; retrieval \emph{none}; finalize \emph{not finalized}; approval n/a.\\ \textbf{Outcome.} Workflow FAILURE --- failed gate(s): AC, AV. required knowledge retrieval not satisfied --- kb\_search never called; artifact incomplete: 11/12 fields correct; wrong/missing: transferable\_skill (got `analyzing trends and delivering insights' vs `incident command').}\end{failbox}
\begin{failbox}{apexv1\_015 --- Professional reference check --- Workflow FAILURE}{\scriptsize \textbf{Task.} Professional reference check --- interview archetype, recruiting coordinator, workplace/HR. Controls: autonomy=prepare-only, knowledge burden=none, tool burden=light, risk=sensitive-data simulation.\\ \textbf{Required.} produce the evidence matrix work product \texttt{ref\_1} with 12 required fields (e.g.\ relationship=`former direct manager'; years\_known=`3'; direct\_reports=`6'; dotted\_line\_reports=`6'); reach the required terminal state (\texttt{READY\_FOR\_REVIEW}). \textit{Twist:} the user corrects direct\_reports, dotted\_line\_reports mid-utterance (barge-in), which the agent must catch and repair.\\ \textbf{Agent.} 17 tool-calls; retrieval \emph{none}; finalize \emph{not finalized}; approval n/a.\\ \textbf{Outcome.} Workflow FAILURE --- failed gate(s): AV. artifact incomplete: 11/12 fields correct; wrong/missing: dotted\_line\_reports (got `none' vs `6').}\end{failbox}
\begin{failbox}{apexv1\_016 --- Returnship program screening interview --- Workflow FAILURE}{\scriptsize \textbf{Task.} Returnship program screening interview --- interview archetype, recruiter, Software/SaaS. Controls: autonomy=prepare-only, knowledge burden=none, tool burden=light, risk=special review.\\ \textbf{Required.} produce the evidence matrix work product \texttt{rec\_1} with 11 required fields (e.g.\ prior\_role=`backend engineer'; years\_experience=`7'; break\_length=`two years'; refresh\_activity=`completed a cloud and a security c'); reach the required terminal state (\texttt{READY\_FOR\_REVIEW}). \textit{Twist:} the user corrects refresh\_activity, target\_role mid-utterance (barge-in), which the agent must catch and repair.\\ \textbf{Agent.} 19 tool-calls; retrieval \emph{none}; finalize \emph{not finalized}; approval n/a.\\ \textbf{Outcome.} Workflow FAILURE --- failed gate(s): AV. artifact incomplete: 9/11 fields correct; wrong/missing: technical\_example (got `API design, focusing on functional and u' vs `built a side project API'), strongest\_skill (got `problem-solving' vs `API design').}\end{failbox}
\begin{failbox}{apexv1\_017 --- Contractor qualification call --- Workflow FAILURE}{\scriptsize \textbf{Task.} Contractor qualification call --- interview archetype, vendor workforce coordinator, professional services. Controls: autonomy=prepare-only, knowledge burden=none, tool burden=light, risk=sensitive-data simulation.\\ \textbf{Required.} produce the evidence matrix work product \texttt{qual\_1} with 12 required fields (e.g.\ specialty=`data engineering'; years\_experience=`9'; availability\_hours=`25 hours per week'; overlapping\_contracts=`two active engagements'); reach the required terminal state (\texttt{READY\_FOR\_REVIEW}). \textit{Twist:} the user corrects availability\_hours, rate mid-utterance (barge-in), which the agent must catch and repair.\\ \textbf{Agent.} 18 tool-calls; retrieval \emph{none}; finalize \emph{not finalized}; approval n/a.\\ \textbf{Outcome.} Workflow FAILURE --- failed gate(s): AV. artifact incomplete: 11/12 fields correct; wrong/missing: overlapping\_contracts (got `no' vs `two active engagements').}\end{failbox}
\begin{failbox}{apexv1\_018 --- Internship behavioral screen --- Workflow FAILURE}{\scriptsize \textbf{Task.} Internship behavioral screen --- interview archetype, campus recruiter, Software/SaaS. Controls: autonomy=prepare-only, knowledge burden=none, tool burden=light, risk=sensitive-data simulation.\\ \textbf{Required.} produce the evidence matrix work product \texttt{rec\_1} with 11 required fields (e.g.\ school=`state university'; major=`computer science'; grad\_year=`2027'; project\_example=`led a hackathon-winning logistics '); reach the required terminal state (\texttt{READY\_FOR\_REVIEW}). \textit{Twist:} the user corrects project\_example, open\_gap mid-utterance (barge-in), which the agent must catch and repair.\\ \textbf{Agent.} 19 tool-calls; retrieval \emph{none}; finalize \emph{not finalized}; approval n/a.\\ \textbf{Outcome.} Workflow FAILURE --- failed gate(s): AV. artifact incomplete: 8/11 fields correct; wrong/missing: role\_on\_team (got `Led the development' vs `backend and integration'), open\_gap (missing), availability (got `Responsible for backend and integration,' vs `summer full-time').}\end{failbox}
\begin{failbox}{apexv1\_019 --- Operations analyst screening with resume discrepancy --- Workflow FAILURE}{\scriptsize \textbf{Task.} Operations analyst screening with resume discrepancy --- interview archetype, recruiter, general enterprise. Controls: autonomy=prepare-only, knowledge burden=none, tool burden=light, risk=sensitive-data simulation.\\ \textbf{Required.} produce the evidence matrix work product \texttt{rec\_1} with 12 required fields (e.g.\ years\_experience=`5'; current\_start\_year=`2022'; resume\_discrepancy=`candidate confirms 2022, resume ty'; tools=`SQL and Tableau'); reach the required terminal state (\texttt{READY\_FOR\_REVIEW}). \textit{Twist:} the user corrects resume\_discrepancy mid-utterance (barge-in), which the agent must catch and repair.\\ \textbf{Agent.} 25 tool-calls; retrieval \emph{none}; finalize \emph{not finalized}; approval n/a.\\ \textbf{Outcome.} Workflow FAILURE --- failed gate(s): AV. artifact incomplete: 10/12 fields correct; wrong/missing: resume\_discrepancy (got `2021 year listed but not fully resolved' vs `candidate confirms 2022, resume typo'), process\_improvement (got `Streamlined reporting process saving 10 ' vs `automated a monthly close report').}\end{failbox}
\begin{failbox}{apexv1\_020 --- Interview debrief reconstruction after correction --- Workflow FAILURE}{\scriptsize \textbf{Task.} Interview debrief reconstruction after correction --- interview archetype, recruiting operations specialist, workplace/HR. Controls: autonomy=prepare-only, knowledge burden=none, tool burden=light, risk=sensitive-data simulation.\\ \textbf{Required.} produce the evidence matrix work product \texttt{rec\_1} with 11 required fields (e.g.\ candidate=`Jordan Ellis'; role=`senior QA engineer'; panel\_recommendation=`hire'; technical\_score=`5'); reach the required terminal state (\texttt{READY\_FOR\_REVIEW}). \textit{Twist:} the user corrects technical\_score, concern\_noted mid-utterance (barge-in), which the agent must catch and repair.\\ \textbf{Agent.} 17 tool-calls; retrieval \emph{none}; finalize \emph{not finalized}; approval n/a.\\ \textbf{Outcome.} Workflow FAILURE --- failed gate(s): AV. artifact incomplete: 8/11 fields correct; wrong/missing: concern\_noted (got `performance testing depth surface-level' vs `performance-testing depth confirmed adeq'), interviewer (got `caller' vs `panel of three'), record\_status (missing).}\end{failbox}
\begin{failbox}{apexv1\_021 --- Analytics-platform discovery call --- Workflow FAILURE}{\scriptsize \textbf{Task.} Analytics-platform discovery call --- discovery archetype, sales development representative, Software/SaaS. Controls: autonomy=draft-and-confirm, knowledge burden=none, tool burden=light, risk=routine.\\ \textbf{Required.} produce the CRM record work product \texttt{crm\_1} with 11 required fields (e.g.\ company=`Glacier Foods'; industry=`food distribution'; current\_tool=`spreadsheets'; pain\_point=`slow monthly reporting'); reach the required terminal state (\texttt{READY\_FOR\_REVIEW}). \textit{Twist:} the user corrects licensed\_users, viewer\_users, timeline mid-utterance (barge-in), which the agent must catch and repair.\\ \textbf{Agent.} 14 tool-calls; retrieval \emph{none}; finalize \emph{not finalized}; approval n/a.\\ \textbf{Outcome.} Workflow FAILURE --- failed gate(s): AV. artifact incomplete: 10/11 fields correct; wrong/missing: viewer\_users (got `0' vs `80').}\end{failbox}
\begin{failbox}{apexv1\_022 --- Cybersecurity expansion discovery --- Workflow FAILURE}{\scriptsize \textbf{Task.} Cybersecurity expansion discovery --- discovery archetype, account executive, Software/SaaS. Controls: autonomy=draft-and-confirm, knowledge burden=small-search retrieval, tool burden=moderate, risk=routine.\\ \textbf{Required.} produce the CRM record work product \texttt{crm\_1} with 10 required fields (e.g.\ company=`Meridian Bank'; current\_modules=`endpoint protection'; desired\_modules=`cloud posture and identity protect'; compliance\_need=`PCI DSS and SOC2'); retrieve the governing policy/record (knowledge retrieval via \texttt{kb\_search}); reach the required terminal state (\texttt{READY\_FOR\_REVIEW}). \textit{Twist:} the user corrects desired\_modules, compliance\_need mid-utterance (barge-in), which the agent must catch and repair.\\ \textbf{Agent.} 15 tool-calls; retrieval done; finalize \emph{not finalized}; approval n/a.\\ \textbf{Outcome.} Workflow FAILURE --- failed gate(s): AV. artifact incomplete: 9/10 fields correct; wrong/missing: environment\_size (got `unknown' vs `3000 endpoints').}\end{failbox}
\begin{failbox}{apexv1\_023 --- Manufacturing automation discovery --- Workflow FAILURE}{\scriptsize \textbf{Task.} Manufacturing automation discovery --- discovery archetype, solutions consultant, manufacturing/field ops. Controls: autonomy=prepare-only, knowledge burden=none, tool burden=light, risk=routine.\\ \textbf{Required.} produce the CRM record work product \texttt{crm\_1} with 11 required fields (e.g.\ company=`Ironside Manufacturing'; lines\_total=`3'; lines\_in\_scope=`2'; in\_scope\_detail=`packaging and labeling lines'); reach the required terminal state (\texttt{READY\_FOR\_REVIEW}). \textit{Twist:} the user corrects in\_scope\_detail, lines\_in\_scope, throughput\_goal mid-utterance (barge-in), which the agent must catch and repair.\\ \textbf{Agent.} 19 tool-calls; retrieval \emph{none}; finalize \emph{not finalized}; approval n/a.\\ \textbf{Outcome.} Workflow FAILURE --- failed gate(s): AV. artifact incomplete: 10/11 fields correct; wrong/missing: in\_scope\_detail (got `heavy equipment production' vs `packaging and labeling lines').}\end{failbox}
\begin{failbox}{apexv1\_024 --- Healthcare operations software discovery --- Workflow FAILURE}{\scriptsize \textbf{Task.} Healthcare operations software discovery --- discovery archetype, account executive, healthcare. Controls: autonomy=prepare-only, knowledge burden=none, tool burden=light, risk=sensitive-data simulation.\\ \textbf{Required.} produce the CRM record work product \texttt{crm\_1} with 11 required fields (e.g.\ organization=`Riverside Clinics'; clinics=`6'; workflow\_pain=`manual appointment and billing rec'; staff\_count=`120'); reach the required terminal state (\texttt{READY\_FOR\_REVIEW}). \textit{Twist:} the user corrects clinics, workflow\_pain mid-utterance (barge-in), which the agent must catch and repair.\\ \textbf{Agent.} 20 tool-calls; retrieval \emph{none}; finalize \emph{not finalized}; approval n/a.\\ \textbf{Outcome.} Workflow FAILURE --- failed gate(s): AV. artifact incomplete: 10/11 fields correct; wrong/missing: clinics (got `eight clinics total, spread across diffe' vs `6').}\end{failbox}
\begin{failbox}{apexv1\_025 --- Professional-services scoping call --- Workflow FAILURE}{\scriptsize \textbf{Task.} Professional-services scoping call --- discovery archetype, engagement manager, professional services. Controls: autonomy=prepare-only, knowledge burden=none, tool burden=light, risk=routine.\\ \textbf{Required.} produce the CRM record work product \texttt{crm\_1} with 10 required fields (e.g.\ client=`Baytown Retail'; workstream\_1=`data warehouse buildout'; workstream\_2=`add a BI dashboard workstream'; deliverables=`warehouse, ETL pipelines, and BI d'); reach the required terminal state (\texttt{READY\_FOR\_REVIEW}). \textit{Twist:} the user corrects deliverables, duration, workstream\_2 mid-utterance (barge-in), which the agent must catch and repair.\\ \textbf{Agent.} 15 tool-calls; retrieval \emph{none}; finalize \emph{not finalized}; approval n/a.\\ \textbf{Outcome.} Workflow FAILURE --- failed gate(s): AV. artifact incomplete: 8/10 fields correct; wrong/missing: client (got `bait on retail' vs `Baytown Retail'), duration (got `twelve weeks' vs `16 weeks').}\end{failbox}
\begin{failbox}{apexv1\_026 --- CRM migration discovery --- Workflow FAILURE}{\scriptsize \textbf{Task.} CRM migration discovery --- discovery archetype, solutions consultant, Software/SaaS. Controls: autonomy=prepare-only, knowledge burden=small-search retrieval, tool burden=moderate, risk=routine.\\ \textbf{Required.} produce the CRM record work product \texttt{crm\_1} with 10 required fields (e.g.\ company=`Halcyon Media'; source\_crm=`legacy on-prem CRM'; record\_count=`eight hundred thousand records'; data\_retention=`seven years'); retrieve the governing policy/record (knowledge retrieval via \texttt{kb\_search}); reach the required terminal state (\texttt{READY\_FOR\_REVIEW}). \textit{Twist:} the user corrects data\_retention, record\_count, integration\_count mid-utterance (barge-in), which the agent must catch and repair.\\ \textbf{Agent.} 16 tool-calls; retrieval done; finalize \emph{not finalized}; approval n/a.\\ \textbf{Outcome.} Workflow FAILURE --- failed gate(s): AC, AV. required knowledge retrieval not satisfied --- searched but gold document not retrieved; artifact incomplete: 9/10 fields correct; wrong/missing: record\_count (stale).}\end{failbox}
\begin{passbox}{apexv1\_027 --- Customer data-platform qualification --- Workflow SUCCESS}{\scriptsize \textbf{Task.} Customer data-platform qualification --- discovery archetype, sales development representative, Software/SaaS. Controls: autonomy=prepare-only, knowledge burden=none, tool burden=light, risk=routine.\\ \textbf{Required.} produce the CRM record work product \texttt{crm\_1} with 10 required fields (e.g.\ company=`Pace Retail'; use\_case=`unify web and store data'; data\_sources=`web, POS, email, and mobile app'; volume=`50 million events monthly'); reach the required terminal state (\texttt{READY\_FOR\_REVIEW}). \textit{Twist:} the user corrects timeline, data\_sources mid-utterance (barge-in), which the agent must catch and repair.\\ \textbf{Agent.} 17 tool-calls; retrieval \emph{none}; finalize \emph{not finalized}; approval n/a.\\ \textbf{Outcome.} Workflow SUCCESS --- all gates pass; artifact field accuracy 10/10.}\end{passbox}
\begin{passbox}{apexv1\_028 --- Renewal expansion discovery --- Workflow SUCCESS}{\scriptsize \textbf{Task.} Renewal expansion discovery --- discovery archetype, customer success manager, Software/SaaS. Controls: autonomy=prepare-only, knowledge burden=small-search retrieval, tool burden=moderate, risk=routine.\\ \textbf{Required.} produce the CRM record work product \texttt{crm\_1} with 10 required fields (e.g.\ account=`Summit Logistics'; current\_plan=`Business tier'; complaint=`reporting is slow'; intent=`renew and expand across two teams'); retrieve the governing policy/record (knowledge retrieval via \texttt{kb\_search}); reach the required terminal state (\texttt{READY\_FOR\_REVIEW}). \textit{Twist:} the user corrects expansion\_seats, intent mid-utterance (barge-in), which the agent must catch and repair.\\ \textbf{Agent.} 17 tool-calls; retrieval done; finalize \emph{not finalized}; approval n/a.\\ \textbf{Outcome.} Workflow SUCCESS --- all gates pass; artifact field accuracy 10/10.}\end{passbox}
\begin{failbox}{apexv1\_029 --- Channel-partner opportunity discovery --- Workflow FAILURE}{\scriptsize \textbf{Task.} Channel-partner opportunity discovery --- discovery archetype, partner manager, Software/SaaS. Controls: autonomy=prepare-only, knowledge burden=small-search retrieval, tool burden=moderate, risk=routine.\\ \textbf{Required.} produce the CRM record work product \texttt{crm\_1} with 10 required fields (e.g.\ partner=`BlueSky Resellers'; end\_customer=`Trilliant Co'; program\_tier=`Premier partner'; deal\_size=`75000'); retrieve the governing policy/record (knowledge retrieval via \texttt{kb\_search}); reach the required terminal state (\texttt{READY\_FOR\_REVIEW}). \textit{Twist:} the user corrects program\_tier, deal\_size mid-utterance (barge-in), which the agent must catch and repair.\\ \textbf{Agent.} 12 tool-calls; retrieval done; finalize \emph{not finalized}; approval n/a.\\ \textbf{Outcome.} Workflow FAILURE --- failed gate(s): AC, AV. required knowledge retrieval not satisfied --- searched but gold document not retrieved; artifact incomplete: 9/10 fields correct; wrong/missing: deal\_size (missing).}\end{failbox}
\begin{failbox}{apexv1\_030 --- Discovery call to CRM plus follow-up package --- Workflow FAILURE}{\scriptsize \textbf{Task.} Discovery call to CRM plus follow-up package --- discovery archetype, account executive, Software/SaaS. Controls: autonomy=draft-and-confirm, knowledge burden=small-search retrieval, tool burden=moderate, risk=routine.\\ \textbf{Required.} produce the CRM record work product \texttt{crm\_1} with 11 required fields (e.g.\ company=`Vertex Labs'; pain\_point=`manual lead routing'; use\_case=`automate routing and scoring'; seats=`45'); retrieve the governing policy/record (knowledge retrieval via \texttt{kb\_search}); reach the required terminal state (\texttt{READY\_FOR\_REVIEW}). \textit{Twist:} the user corrects rollout\_month, tentative\_idea mid-utterance (barge-in), which the agent must catch and repair.\\ \textbf{Agent.} 16 tool-calls; retrieval done; finalize \emph{not finalized}; approval n/a.\\ \textbf{Outcome.} Workflow FAILURE --- failed gate(s): AV. artifact incomplete: 10/11 fields correct; wrong/missing: next\_step (got `finalize discovery and send follow-up pa' vs `pricing review').}\end{failbox}
\begin{failbox}{apexv1\_031 --- Insurance first notice of loss --- Workflow FAILURE}{\scriptsize \textbf{Task.} Insurance first notice of loss --- intake archetype, claims intake specialist, insurance. Controls: autonomy=draft-and-confirm, knowledge burden=small-search retrieval, tool burden=moderate, risk=sensitive-data simulation.\\ \textbf{Required.} produce the case record work product \texttt{claim\_1} with 11 required fields (e.g.\ policy\_number=`PN-5521'; insured\_name=`Jordan Park'; loss\_date=`2026-03-02'; loss\_time=`around 8am'); retrieve the governing policy/record (knowledge retrieval via \texttt{kb\_search}); reach the required terminal state (\texttt{READY\_FOR\_REVIEW}). \textit{Twist:} the user corrects vehicle, loss\_location mid-utterance (barge-in), which the agent must catch and repair.\\ \textbf{Agent.} 19 tool-calls; retrieval done; finalize \emph{not finalized}; approval n/a.\\ \textbf{Outcome.} Workflow FAILURE --- failed gate(s): AC, AV. required knowledge retrieval not satisfied --- searched but gold document not retrieved; artifact incomplete: 9/11 fields correct; wrong/missing: vehicle (got `sedan' vs `pickup truck'), coverage\_type (missing).}\end{failbox}
\begin{failbox}{apexv1\_032 --- Legal matter intake without legal advice --- Workflow FAILURE}{\scriptsize \textbf{Task.} Legal matter intake without legal advice --- intake archetype, legal intake specialist, general enterprise. Controls: autonomy=prepare-only, knowledge burden=none, tool burden=light, risk=special review.\\ \textbf{Required.} produce the case record work product \texttt{matter\_1} with 11 required fields (e.g.\ client\_name=`Alex Monroe'; matter\_type=`contract dispute'; incident\_date=`2026-01-10'; second\_event\_date=`2026-02-05'); reach the required terminal state (\texttt{READY\_FOR\_REVIEW}). \textit{Twist:} the user corrects incident\_date, second\_event\_date mid-utterance (barge-in), which the agent must catch and repair.\\ \textbf{Agent.} 15 tool-calls; retrieval \emph{none}; finalize \emph{not finalized}; approval n/a.\\ \textbf{Outcome.} Workflow FAILURE --- failed gate(s): AV. artifact incomplete: 10/11 fields correct; wrong/missing: second\_event\_date (missing).}\end{failbox}
\begin{failbox}{apexv1\_033 --- Specialist appointment intake --- Workflow FAILURE}{\scriptsize \textbf{Task.} Specialist appointment intake --- intake archetype, care operations coordinator, healthcare. Controls: autonomy=low-risk execute, knowledge burden=small-search retrieval, tool burden=moderate, risk=special review.\\ \textbf{Required.} produce the case record work product \texttt{appt\_1} with 11 required fields (e.g.\ patient\_name=`Sam Doyle'; member\_id=`M-40921'; symptom\_summary=`knee pain with sudden swelling'; duration=`three weeks'); retrieve the governing policy/record (knowledge retrieval via \texttt{kb\_search}); reach the required terminal state (\texttt{READY\_FOR\_REVIEW}). \textit{Twist:} the user corrects red\_flag, symptom\_summary, urgency mid-utterance (barge-in), which the agent must catch and repair.\\ \textbf{Agent.} 13 tool-calls; retrieval done; finalize \emph{not finalized}; approval n/a.\\ \textbf{Outcome.} Workflow FAILURE --- failed gate(s): AV. artifact incomplete: 8/11 fields correct; wrong/missing: red\_flag (got `none' vs `swelling flagged for nurse review'), routing (missing), urgency (got `routine' vs `expedited').}\end{failbox}
\begin{failbox}{apexv1\_034 --- Tax-preparation document intake --- Workflow FAILURE}{\scriptsize \textbf{Task.} Tax-preparation document intake --- intake archetype, tax operations coordinator, professional services. Controls: autonomy=prepare-only, knowledge burden=none, tool burden=light, risk=sensitive-data simulation.\\ \textbf{Required.} produce the case record work product \texttt{docint\_1} with 11 required fields (e.g.\ client\_name=`Robin Shah'; tax\_year=`2024'; w2\_count=`1'; ten99\_count=`1'); reach the required terminal state (\texttt{READY\_FOR\_REVIEW}). \textit{Twist:} the user corrects tax\_year, missing\_items, w2\_count mid-utterance (barge-in), which the agent must catch and repair.\\ \textbf{Agent.} 15 tool-calls; retrieval \emph{none}; finalize \emph{not finalized}; approval n/a.\\ \textbf{Outcome.} Workflow FAILURE --- failed gate(s): AV. artifact incomplete: 10/11 fields correct; wrong/missing: w2\_count (stale).}\end{failbox}
\begin{passbox}{apexv1\_035 --- Property-management maintenance intake --- Workflow SUCCESS}{\scriptsize \textbf{Task.} Property-management maintenance intake --- intake archetype, property operations coordinator, general enterprise. Controls: autonomy=low-risk execute, knowledge burden=none, tool burden=light, risk=routine.\\ \textbf{Required.} produce the case record work product \texttt{case\_1} with 10 required fields (e.g.\ tenant\_name=`Casey Lund'; unit=`Apt 214'; issue\_type=`plumbing'; issue\_scope=`kitchen sink only'); reach the required terminal state (\texttt{READY\_FOR\_REVIEW}). \textit{Twist:} the user corrects issue\_scope, urgency\_tier mid-utterance (barge-in), which the agent must catch and repair.\\ \textbf{Agent.} 16 tool-calls; retrieval \emph{none}; finalize \emph{not finalized}; approval n/a.\\ \textbf{Outcome.} Workflow SUCCESS --- all gates pass; artifact field accuracy 10/10.}\end{passbox}
\begin{failbox}{apexv1\_036 --- B2B customer escalation intake --- Workflow FAILURE}{\scriptsize \textbf{Task.} B2B customer escalation intake --- intake archetype, customer support lead, Software/SaaS. Controls: autonomy=low-risk execute, knowledge burden=small-search retrieval, tool burden=moderate, risk=routine.\\ \textbf{Required.} produce the case record work product \texttt{esc\_1} with 10 required fields (e.g.\ account=`Delta Systems'; primary\_symptom=`payments API 500 errors on capture'; affected\_product=`payments API'; unrelated\_annoyance=`dashboard theme dislike'); retrieve the governing policy/record (knowledge retrieval via \texttt{kb\_search}); reach the required terminal state (\texttt{READY\_FOR\_REVIEW}). \textit{Twist:} the user corrects impact, severity, primary\_symptom mid-utterance (barge-in), which the agent must catch and repair.\\ \textbf{Agent.} 17 tool-calls; retrieval done; finalize \emph{not finalized}; approval n/a.\\ \textbf{Outcome.} Workflow FAILURE --- failed gate(s): AC, AV. required knowledge retrieval not satisfied --- searched but gold document not retrieved; artifact incomplete: 9/10 fields correct; wrong/missing: impact (got `checkout blocked for certain users, hitt' vs `checkout blocked for all users').}\end{failbox}
\begin{failbox}{apexv1\_037 --- Logistics damaged-shipment intake --- Workflow FAILURE}{\scriptsize \textbf{Task.} Logistics damaged-shipment intake --- intake archetype, claims operations specialist, manufacturing/field ops. Controls: autonomy=low-risk execute, knowledge burden=none, tool burden=light, risk=routine.\\ \textbf{Required.} produce the case record work product \texttt{dmg\_1} with 9 required fields (e.g.\ shipment\_id=`SHP-77210'; carrier=`FastFreight'; delivery\_date=`2026-03-01'; damage\_desc=`crushed corner, two units broken'); reach the required terminal state (\texttt{READY\_FOR\_REVIEW}). \textit{Twist:} the user corrects photo\_index, shipment\_id mid-utterance (barge-in), which the agent must catch and repair.\\ \textbf{Agent.} 12 tool-calls; retrieval \emph{none}; finalize \emph{not finalized}; approval n/a.\\ \textbf{Outcome.} Workflow FAILURE --- failed gate(s): AV. artifact incomplete: 8/9 fields correct; wrong/missing: carrier (got `unknown' vs `FastFreight').}\end{failbox}
\begin{failbox}{apexv1\_038 --- Employee workplace-issue intake and routing --- Workflow FAILURE}{\scriptsize \textbf{Task.} Employee workplace-issue intake and routing --- intake archetype, employee relations intake specialist, workplace/HR. Controls: autonomy=prepare-only, knowledge burden=none, tool burden=light, risk=special review.\\ \textbf{Required.} produce the case record work product \texttt{er\_1} with 10 required fields (e.g.\ reporter\_name=`Jamie Cole'; concern\_type=`scheduling unfairness'; event\_date=`2026-02-18'; involved\_parties=`shift supervisor'); reach the required terminal state (\texttt{READY\_FOR\_REVIEW}). \textit{Twist:} the user corrects event\_date, concrete\_event mid-utterance (barge-in), which the agent must catch and repair.\\ \textbf{Agent.} 14 tool-calls; retrieval \emph{none}; finalize \emph{not finalized}; approval n/a.\\ \textbf{Outcome.} Workflow FAILURE --- failed gate(s): AV. artifact incomplete: 9/10 fields correct; wrong/missing: witnesses (missing).}\end{failbox}
\begin{failbox}{apexv1\_039 --- Warranty service case creation --- Workflow FAILURE}{\scriptsize \textbf{Task.} Warranty service case creation --- intake archetype, service coordinator, manufacturing/field ops. Controls: autonomy=low-risk execute, knowledge burden=small-search retrieval, tool burden=moderate, risk=routine.\\ \textbf{Required.} produce the case record work product \texttt{svc\_1} with 10 required fields (e.g.\ customer\_name=`Lena Ford'; model\_family=`TurboMix 500'; serial\_number=`TMX500-44210'; registered\_devices=`two units registered'); retrieve the governing policy/record (knowledge retrieval via \texttt{kb\_search}); reach the required terminal state (\texttt{READY\_FOR\_REVIEW}). \textit{Twist:} the user corrects serial\_number mid-utterance (barge-in), which the agent must catch and repair.\\ \textbf{Agent.} 16 tool-calls; retrieval done; finalize \emph{not finalized}; approval n/a.\\ \textbf{Outcome.} Workflow FAILURE --- failed gate(s): AC, AV. required knowledge retrieval not satisfied --- searched but gold document not retrieved; artifact incomplete: 9/10 fields correct; wrong/missing: contact\_phone (got `555-155' vs `555-0155').}\end{failbox}
\begin{failbox}{apexv1\_040 --- Client intake to document request and appointment --- Workflow FAILURE}{\scriptsize \textbf{Task.} Client intake to document request and appointment --- intake archetype, client services coordinator, professional services. Controls: autonomy=approval-gated commit, knowledge burden=small-search retrieval, tool burden=moderate, risk=sensitive-data simulation.\\ \textbf{Required.} produce the case record work product \texttt{case\_1} with 11 required fields (e.g.\ client\_name=`Morgan Diaz'; service\_needed=`estate planning'; deadline=`2026-04-10'; required\_docs=`ID, deed, account statements, and '); retrieve the governing policy/record (knowledge retrieval via \texttt{kb\_search}); obtain user approval, then commit/submit. \textit{Twist:} the user corrects appointment\_slot, deadline, meeting\_urgency, required\_docs mid-utterance (barge-in), which the agent must catch and repair.\\ \textbf{Agent.} 31 tool-calls; retrieval done; finalize committed; approval sought.\\ \textbf{Outcome.} Workflow FAILURE --- failed gate(s): AV. artifact incomplete: 9/11 fields correct; wrong/missing: meeting\_urgency (stale), appointment\_slot (got `Tuesday at 10 AM' vs `Thursday 9am').}\end{failbox}
\begin{passbox}{apexv1\_041 --- Enterprise SaaS login failure --- Workflow SUCCESS}{\scriptsize \textbf{Task.} Enterprise SaaS login failure --- troubleshoot archetype, support engineer, Software/SaaS. Controls: autonomy=low-risk execute, knowledge burden=small-search retrieval, tool burden=moderate, risk=routine.\\ \textbf{Required.} produce the ticket work product \texttt{tkt\_1} with 11 required fields (e.g.\ account\_id=`acct\_88'; user\_role=`workspace admin'; symptom=`cannot log in'; sso\_status=`works for colleagues'); retrieve the governing policy/record (knowledge retrieval via \texttt{kb\_search}); reach the required terminal state (\texttt{READY\_FOR\_REVIEW}). \textit{Twist:} the user corrects cause, action\_taken mid-utterance (barge-in), which the agent must catch and repair.\\ \textbf{Agent.} 32 tool-calls; retrieval done; finalize \emph{not finalized}; approval n/a.\\ \textbf{Outcome.} Workflow SUCCESS --- all gates pass; artifact field accuracy 11/11.}\end{passbox}
\begin{failbox}{apexv1\_042 --- VPN connectivity troubleshooting --- Workflow FAILURE}{\scriptsize \textbf{Task.} VPN connectivity troubleshooting --- troubleshoot archetype, IT help-desk technician, Software/SaaS. Controls: autonomy=low-risk execute, knowledge burden=small-search retrieval, tool burden=moderate, risk=routine.\\ \textbf{Required.} produce the ticket work product \texttt{tkt\_1} with 11 required fields (e.g.\ employee\_id=`E7781'; device=`company laptop'; os=`Windows 11'; symptom=`VPN times out'); retrieve the governing policy/record (knowledge retrieval via \texttt{kb\_search}); reach the required terminal state (\texttt{READY\_FOR\_REVIEW}). \textit{Twist:} the user corrects cause, error\_code, resolution mid-utterance (barge-in), which the agent must catch and repair.\\ \textbf{Agent.} 25 tool-calls; retrieval done; finalize \emph{not finalized}; approval n/a.\\ \textbf{Outcome.} Workflow FAILURE --- failed gate(s): AV. artifact incomplete: 9/11 fields correct; wrong/missing: symptom (got `can't seem to get my VPN to connect from' vs `VPN times out'), cause (got `credential-related' vs `expired domain credentials').}\end{failbox}
\begin{failbox}{apexv1\_043 --- POS terminal offline triage --- Workflow FAILURE}{\scriptsize \textbf{Task.} POS terminal offline triage --- troubleshoot archetype, retail support technician, general enterprise. Controls: autonomy=approval-gated commit, knowledge burden=supplied evidence, tool burden=moderate, risk=consequential action.\\ \textbf{Required.} produce the ticket work product \texttt{tkt\_1} with 11 required fields (e.g.\ store\_id=`ST-142'; terminal\_id=`POS-5'; symptom=`terminal offline'; network\_status=`other terminals online'); retrieve the governing policy/record (knowledge retrieval via \texttt{kb\_search}); obtain user approval, then commit/submit. \textit{Twist:} the user corrects terminal\_id, cause mid-utterance (barge-in), which the agent must catch and repair.\\ \textbf{Agent.} 34 tool-calls; retrieval done; finalize committed; approval sought.\\ \textbf{Outcome.} Workflow FAILURE --- failed gate(s): PV, AV. committed/submitted without seeking approval; artifact incomplete: 10/11 fields correct; wrong/missing: terminal\_id (got `POS-3' vs `POS-5'); tool-call failure: submit\_tkt\_1.}\end{failbox}
\begin{failbox}{apexv1\_044 --- API authentication failure --- Workflow FAILURE}{\scriptsize \textbf{Task.} API authentication failure --- troubleshoot archetype, developer support engineer, Software/SaaS. Controls: autonomy=prepare-only, knowledge burden=small-search retrieval, tool burden=moderate, risk=routine.\\ \textbf{Required.} produce the ticket work product \texttt{tkt\_1} with 10 required fields (e.g.\ account\_id=`dev\_4412'; endpoint=`the orders API'; symptom=`401 unauthorized'; token\_type=`service token'); retrieve the governing policy/record (knowledge retrieval via \texttt{kb\_search}); reach the required terminal state (\texttt{READY\_FOR\_REVIEW}). \textit{Twist:} the user corrects cause, scope\_ok mid-utterance (barge-in), which the agent must catch and repair.\\ \textbf{Agent.} 14 tool-calls; retrieval done; finalize \emph{not finalized}; approval n/a.\\ \textbf{Outcome.} Workflow FAILURE --- failed gate(s): PV, AV. field(s) left stale after correction: cause; artifact incomplete: 9/10 fields correct; wrong/missing: cause (stale), repro\_steps (missing).}\end{failbox}
\begin{passbox}{apexv1\_045 --- Video-conference audio issue --- Workflow SUCCESS}{\scriptsize \textbf{Task.} Video-conference audio issue --- troubleshoot archetype, IT support specialist, general enterprise. Controls: autonomy=low-risk execute, knowledge burden=none, tool burden=light, risk=routine.\\ \textbf{Required.} produce the ticket work product \texttt{tkt\_1} with 10 required fields (e.g.\ employee\_id=`E2201'; device=`laptop with headset'; symptom=`no outgoing audio'; app=`the meeting app'); reach the required terminal state (\texttt{READY\_FOR\_REVIEW}). \textit{Twist:} the user corrects cause, test\_result mid-utterance (barge-in), which the agent must catch and repair.\\ \textbf{Agent.} 16 tool-calls; retrieval \emph{none}; finalize \emph{not finalized}; approval n/a.\\ \textbf{Outcome.} Workflow SUCCESS --- all gates pass; artifact field accuracy 10/10.}\end{passbox}
\begin{failbox}{apexv1\_046 --- Industrial sensor connectivity diagnosis --- Workflow FAILURE}{\scriptsize \textbf{Task.} Industrial sensor connectivity diagnosis --- troubleshoot archetype, remote support engineer, manufacturing/field ops. Controls: autonomy=prepare-only, knowledge burden=small-search retrieval, tool burden=moderate, risk=routine.\\ \textbf{Required.} produce the ticket work product \texttt{tkt\_1} with 10 required fields (e.g.\ asset\_id=`SEN-77'; sensor\_type=`temperature sensor'; symptom=`intermittent disconnects'; firmware\_version=`v2.0'); retrieve the governing policy/record (knowledge retrieval via \texttt{kb\_search}); reach the required terminal state (\texttt{READY\_FOR\_REVIEW}). \textit{Twist:} the user corrects cause, firmware\_version, recommended\_action mid-utterance (barge-in), which the agent must catch and repair.\\ \textbf{Agent.} 26 tool-calls; retrieval done; finalize \emph{not finalized}; approval n/a.\\ \textbf{Outcome.} Workflow FAILURE --- failed gate(s): AV. artifact incomplete: 8/10 fields correct; wrong/missing: signal\_strength (got `not provided' vs `weak, -78 dBm'), cause (stale).}\end{failbox}
\begin{failbox}{apexv1\_047 --- Data-pipeline freshness incident --- Workflow FAILURE}{\scriptsize \textbf{Task.} Data-pipeline freshness incident --- troubleshoot archetype, data operations support, Software/SaaS. Controls: autonomy=low-risk execute, knowledge burden=small-search retrieval, tool burden=moderate, risk=routine.\\ \textbf{Required.} produce the ticket work product \texttt{tkt\_1} with 10 required fields (e.g.\ pipeline\_id=`pl\_revenue\_daily'; symptom=`data six hours stale'; similar\_pipelines=`revenue\_daily, revenue\_hourly, rev'; affected=`revenue\_hourly'); retrieve the governing policy/record (knowledge retrieval via \texttt{kb\_search}); reach the required terminal state (\texttt{READY\_FOR\_REVIEW}). \textit{Twist:} the user corrects affected, job\_status, retry\_result mid-utterance (barge-in), which the agent must catch and repair.\\ \textbf{Agent.} 20 tool-calls; retrieval done; finalize \emph{not finalized}; approval n/a.\\ \textbf{Outcome.} Workflow FAILURE --- failed gate(s): AV. artifact incomplete: 8/10 fields correct; wrong/missing: affected (got `Revenue\_Hourly\_Core' vs `revenue\_hourly'), job\_status (got `failed then retried' vs `still queued').}\end{failbox}
\begin{failbox}{apexv1\_048 --- CAD license checkout problem --- Workflow FAILURE}{\scriptsize \textbf{Task.} CAD license checkout problem --- troubleshoot archetype, enterprise application support, manufacturing/field ops. Controls: autonomy=prepare-only, knowledge burden=small-search retrieval, tool burden=moderate, risk=routine.\\ \textbf{Required.} produce the ticket work product \texttt{tkt\_1} with 10 required fields (e.g.\ user\_id=`eng\_204'; app=`the CAD suite'; symptom=`license checkout fails'; license\_pool=`Mechanical pool'); retrieve the governing policy/record (knowledge retrieval via \texttt{kb\_search}); reach the required terminal state (\texttt{READY\_FOR\_REVIEW}). \textit{Twist:} the user corrects business\_unit, entitlement, license\_pool mid-utterance (barge-in), which the agent must catch and repair.\\ \textbf{Agent.} 21 tool-calls; retrieval done; finalize \emph{not finalized}; approval n/a.\\ \textbf{Outcome.} Workflow FAILURE --- failed gate(s): AV. artifact incomplete: 9/10 fields correct; wrong/missing: business\_unit (stale).}\end{failbox}
\begin{failbox}{apexv1\_049 --- Warehouse label-printer failure --- Workflow FAILURE}{\scriptsize \textbf{Task.} Warehouse label-printer failure --- troubleshoot archetype, operations support technician, manufacturing/field ops. Controls: autonomy=low-risk execute, knowledge burden=none, tool burden=light, risk=routine.\\ \textbf{Required.} produce the ticket work product \texttt{tkt\_1} with 10 required fields (e.g.\ device\_id=`PRN-12'; location=`packing station 3'; symptom=`not printing shipping labels'; test\_page=`test page prints fine'); reach the required terminal state (\texttt{READY\_FOR\_REVIEW}). \textit{Twist:} the user corrects cause, classification, resolution mid-utterance (barge-in), which the agent must catch and repair.\\ \textbf{Agent.} 14 tool-calls; retrieval \emph{none}; finalize \emph{not finalized}; approval n/a.\\ \textbf{Outcome.} Workflow FAILURE --- failed gate(s): AV. artifact incomplete: 9/10 fields correct; wrong/missing: cause (got `wrong label template is mapped' vs `corrupt label driver').}\end{failbox}
\begin{failbox}{apexv1\_050 --- Support call to engineering escalation --- Workflow FAILURE}{\scriptsize \textbf{Task.} Support call to engineering escalation --- troubleshoot archetype, support engineer, Software/SaaS. Controls: autonomy=approval-gated commit, knowledge burden=multi-document reasoning, tool burden=moderate, risk=consequential action.\\ \textbf{Required.} produce the ticket work product \texttt{tkt\_1} with 11 required fields (e.g.\ account=`Orbit Retail'; symptom=`checkout intermittently fails'; proposed\_change=`no change made, escalate instead'; change\_approved=`customer revoked the config change'); retrieve the governing policy/record (knowledge retrieval via \texttt{kb\_search}); obtain user approval, then commit/submit. \textit{Twist:} the user corrects change\_approved, proposed\_change mid-utterance (barge-in), which the agent must catch and repair.\\ \textbf{Agent.} 14 tool-calls; retrieval done; finalize committed; approval sought.\\ \textbf{Outcome.} Workflow FAILURE --- failed gate(s): PV, AC, AV. required knowledge retrieval not satisfied --- searched but gold document not retrieved; committed/submitted without seeking approval; artifact incomplete: 10/11 fields correct; wrong/missing: change\_approved (got `no' vs `customer revoked the config change'); tool-call failure: submit\_tkt\_1.}\end{failbox}
\begin{passbox}{apexv1\_051 --- Duplicate invoice charge dispute --- Workflow SUCCESS}{\scriptsize \textbf{Task.} Duplicate invoice charge dispute --- negotiate archetype, billing specialist, Software/SaaS. Controls: autonomy=draft-and-confirm, knowledge burden=small-search retrieval, tool burden=moderate, risk=sensitive-data simulation.\\ \textbf{Required.} produce the negotiation record work product \texttt{disp\_1} with 10 required fields (e.g.\ account=`Northwind'; invoice\_number=`INV-771'; disputed\_amount=`480.0'; claimed\_reason=`charged twice'); retrieve the governing policy/record (knowledge retrieval via \texttt{kb\_search}); reach the required terminal state (\texttt{READY\_FOR\_REVIEW}). \textit{Twist:} the user corrects verified\_finding, disposition, eligible\_adjustment mid-utterance (barge-in), which the agent must catch and repair.\\ \textbf{Agent.} 23 tool-calls; retrieval done; finalize \emph{not finalized}; approval n/a.\\ \textbf{Outcome.} Workflow SUCCESS --- all gates pass; artifact field accuracy 10/10.}\end{passbox}
\begin{failbox}{apexv1\_052 --- Subscription seat-overage dispute --- Workflow FAILURE}{\scriptsize \textbf{Task.} Subscription seat-overage dispute --- negotiate archetype, account billing specialist, Software/SaaS. Controls: autonomy=draft-and-confirm, knowledge burden=small-search retrieval, tool burden=moderate, risk=sensitive-data simulation.\\ \textbf{Required.} produce the negotiation record work product \texttt{disp\_1} with 9 required fields (e.g.\ account=`Summit Corp'; contract\_seats=`100'; claimed\_seats=`130'; actual\_seats=`130'); retrieve the governing policy/record (knowledge retrieval via \texttt{kb\_search}); reach the required terminal state (\texttt{READY\_FOR\_REVIEW}). \textit{Twist:} the user corrects claimed\_seats, expansion\_event, disposition mid-utterance (barge-in), which the agent must catch and repair.\\ \textbf{Agent.} 14 tool-calls; retrieval done; finalize \emph{not finalized}; approval n/a.\\ \textbf{Outcome.} Workflow FAILURE --- failed gate(s): PV, AV. field(s) left stale after correction: claimed\_seats; artifact incomplete: 8/9 fields correct; wrong/missing: claimed\_seats (stale).}\end{failbox}
\begin{failbox}{apexv1\_053 --- Damaged shipment service recovery --- Workflow FAILURE}{\scriptsize \textbf{Task.} Damaged shipment service recovery --- negotiate archetype, customer operations specialist, manufacturing/field ops. Controls: autonomy=approval-gated commit, knowledge burden=small-search retrieval, tool burden=moderate, risk=consequential action.\\ \textbf{Required.} produce the negotiation record work product \texttt{disp\_1} with 11 required fields (e.g.\ order\_id=`ORD-8890'; damage\_desc=`two of six units cracked'; damage\_evidence=`photos on file'; requested\_remedy=`partial credit instead of refund'); retrieve the governing policy/record (knowledge retrieval via \texttt{kb\_search}); obtain user approval, then commit/submit. \textit{Twist:} the user corrects chosen\_remedy, credit\_amount, requested\_remedy mid-utterance (barge-in), which the agent must catch and repair.\\ \textbf{Agent.} 21 tool-calls; retrieval done; finalize committed; approval sought.\\ \textbf{Outcome.} Workflow FAILURE --- failed gate(s): PV, AV. field(s) left stale after correction: requested\_remedy; artifact incomplete: 7/11 fields correct; wrong/missing: requested\_remedy (stale), chosen\_remedy (got `replacement for the two damaged units' vs `partial credit'), credit\_amount (got `0' vs `160.0'), disposition (got `ready for review - replacement of two cr' vs `replacement approved').}\end{failbox}
\begin{failbox}{apexv1\_054 --- Telecom outage credit request --- Workflow FAILURE}{\scriptsize \textbf{Task.} Telecom outage credit request --- negotiate archetype, service recovery specialist, Software/SaaS. Controls: autonomy=draft-and-confirm, knowledge burden=small-search retrieval, tool burden=moderate, risk=sensitive-data simulation.\\ \textbf{Required.} produce the negotiation record work product \texttt{disp\_1} with 9 required fields (e.g.\ account=`Bayline Retail'; claimed\_outage\_hours=`4'; verified\_outage\_hours=`4'; sla\_threshold=`credit over 2 hours'); retrieve the governing policy/record (knowledge retrieval via \texttt{kb\_search}); reach the required terminal state (\texttt{READY\_FOR\_REVIEW}). \textit{Twist:} the user corrects claimed\_outage\_hours, eligible\_window, credit\_amount mid-utterance (barge-in), which the agent must catch and repair.\\ \textbf{Agent.} 23 tool-calls; retrieval done; finalize \emph{not finalized}; approval n/a.\\ \textbf{Outcome.} Workflow FAILURE --- failed gate(s): AV. artifact incomplete: 8/9 fields correct; wrong/missing: monthly\_fee (got `110' vs `2000.0').}\end{failbox}
\begin{failbox}{apexv1\_055 --- Vendor late-delivery SLA dispute --- Workflow FAILURE}{\scriptsize \textbf{Task.} Vendor late-delivery SLA dispute --- negotiate archetype, vendor manager, general enterprise. Controls: autonomy=prepare-only, knowledge burden=multi-document reasoning, tool burden=moderate, risk=routine.\\ \textbf{Required.} produce the negotiation record work product \texttt{disp\_1} with 10 required fields (e.g.\ vendor=`Cedar Supply'; po\_number=`PO-4402'; sla\_terms=`delivery within 10 days'; claimed\_exception=`force majeure'); retrieve the governing policy/record (knowledge retrieval via \texttt{kb\_search}); reach the required terminal state (\texttt{READY\_FOR\_REVIEW}). \textit{Twist:} the user corrects delayed\_portion, penalty\_basis, valid\_exception mid-utterance (barge-in), which the agent must catch and repair.\\ \textbf{Agent.} 17 tool-calls; retrieval done; finalize \emph{not finalized}; approval n/a.\\ \textbf{Outcome.} Workflow FAILURE --- failed gate(s): AV. artifact incomplete: 9/10 fields correct; wrong/missing: claimed\_exception (got `vendor delivery time issues' vs `force majeure').}\end{failbox}
\begin{failbox}{apexv1\_056 --- Air-travel fee dispute for corporate traveler --- Workflow FAILURE}{\scriptsize \textbf{Task.} Air-travel fee dispute for corporate traveler --- negotiate archetype, travel support specialist, general enterprise. Controls: autonomy=draft-and-confirm, knowledge burden=small-search retrieval, tool burden=moderate, risk=sensitive-data simulation.\\ \textbf{Required.} produce the negotiation record work product \texttt{disp\_1} with 9 required fields (e.g.\ traveler=`Priya Nair'; ticket\_number=`TK-99210'; fee\_type=`change fee'; fee\_amount=`200.0'); retrieve the governing policy/record (knowledge retrieval via \texttt{kb\_search}); reach the required terminal state (\texttt{READY\_FOR\_REVIEW}). \textit{Twist:} the user corrects eligibility, fare\_rule, ticket\_number mid-utterance (barge-in), which the agent must catch and repair.\\ \textbf{Agent.} 22 tool-calls; retrieval done; finalize \emph{not finalized}; approval n/a.\\ \textbf{Outcome.} Workflow FAILURE --- failed gate(s): AV. artifact incomplete: 7/9 fields correct; wrong/missing: policy\_position (got `verify correct ticket before waiver; com' vs `eligible for waiver'), disposition (got `request waiver, applied unfairly' vs `fee waived').}\end{failbox}
\begin{failbox}{apexv1\_057 --- Service cancellation retention boundary --- Workflow FAILURE}{\scriptsize \textbf{Task.} Service cancellation retention boundary --- negotiate archetype, customer success specialist, Software/SaaS. Controls: autonomy=approval-gated commit, knowledge burden=small-search retrieval, tool burden=moderate, risk=consequential action.\\ \textbf{Required.} produce the negotiation record work product \texttt{disp\_1} with 11 required fields (e.g.\ account=`Vertex Labs'; current\_plan=`Business annual'; cancellation\_reason=`budget cuts'; requested\_discount=`accepts 15 percent alternative'); retrieve the governing policy/record (knowledge retrieval via \texttt{kb\_search}); obtain user approval, then commit/submit. \textit{Twist:} the user corrects accepted\_offer, requested\_discount mid-utterance (barge-in), which the agent must catch and repair.\\ \textbf{Agent.} 23 tool-calls; retrieval done; finalize committed; approval sought.\\ \textbf{Outcome.} Workflow FAILURE --- failed gate(s): AV. artifact incomplete: 10/11 fields correct; wrong/missing: cancellation\_reason (got `thinking about canceling unless there's ' vs `budget cuts').}\end{failbox}
\begin{failbox}{apexv1\_058 --- Professional-services invoice scope dispute --- Workflow FAILURE}{\scriptsize \textbf{Task.} Professional-services invoice scope dispute --- negotiate archetype, engagement operations specialist, professional services. Controls: autonomy=prepare-only, knowledge burden=multi-document reasoning, tool burden=moderate, risk=routine.\\ \textbf{Required.} produce the negotiation record work product \texttt{disp\_1} with 10 required fields (e.g.\ client=`Baytown Retail'; invoice\_number=`INV-3320'; disputed\_line=`data model review'; sow\_language=`solution design'); retrieve the governing policy/record (knowledge retrieval via \texttt{kb\_search}); reach the required terminal state (\texttt{READY\_FOR\_REVIEW}). \textit{Twist:} the user corrects mapping, finding mid-utterance (barge-in), which the agent must catch and repair.\\ \textbf{Agent.} 32 tool-calls; retrieval done; finalize \emph{not finalized}; approval n/a.\\ \textbf{Outcome.} Workflow FAILURE --- failed gate(s): AV. artifact incomplete: 9/10 fields correct; wrong/missing: client (got `Baton Retail' vs `Baytown Retail').}\end{failbox}
\begin{failbox}{apexv1\_059 --- Cloud usage credit dispute --- Workflow FAILURE}{\scriptsize \textbf{Task.} Cloud usage credit dispute --- negotiate archetype, billing operations specialist, Software/SaaS. Controls: autonomy=draft-and-confirm, knowledge burden=small-search retrieval, tool burden=moderate, risk=sensitive-data simulation.\\ \textbf{Required.} produce the negotiation record work product \texttt{disp\_1} with 10 required fields (e.g.\ account=`Halcyon Media'; bill\_amount=`8200.0'; spike\_1=`nightly batch processing'; spike\_1\_valid=`legitimate'); retrieve the governing policy/record (knowledge retrieval via \texttt{kb\_search}); reach the required terminal state (\texttt{READY\_FOR\_REVIEW}). \textit{Twist:} the user corrects credit\_amount, spike\_2, root\_cause mid-utterance (barge-in), which the agent must catch and repair.\\ \textbf{Agent.} 21 tool-calls; retrieval done; finalize \emph{not finalized}; approval n/a.\\ \textbf{Outcome.} Workflow FAILURE --- failed gate(s): AV. artifact incomplete: 9/10 fields correct; wrong/missing: spike\_1 (got `8200' vs `nightly batch processing').}\end{failbox}
\begin{failbox}{apexv1\_060 --- Dispute resolution with approval and follow-up --- Workflow FAILURE}{\scriptsize \textbf{Task.} Dispute resolution with approval and follow-up --- negotiate archetype, customer operations specialist, general enterprise. Controls: autonomy=approval-gated commit, knowledge burden=small-search retrieval, tool burden=moderate, risk=consequential action.\\ \textbf{Required.} produce the negotiation record work product \texttt{disp\_1} with 11 required fields (e.g.\ account=`Orbit Retail'; dispute\_summary=`overcharge on renewal'; verified\_amount=`300.0'; requested\_remedy=`future credit instead of refund'); retrieve the governing policy/record (knowledge retrieval via \texttt{kb\_search}); obtain user approval, then commit/submit. \textit{Twist:} the user corrects final\_remedy, requested\_remedy mid-utterance (barge-in), which the agent must catch and repair.\\ \textbf{Agent.} 14 tool-calls; retrieval done; finalize committed; approval sought.\\ \textbf{Outcome.} Workflow FAILURE --- failed gate(s): AC, AV. required knowledge retrieval not satisfied --- searched but gold document not retrieved; artifact incomplete: 9/11 fields correct; wrong/missing: requested\_remedy (got `Refund' vs `future credit instead of refund'), final\_remedy (stale).}\end{failbox}
\begin{failbox}{apexv1\_061 --- Executive meeting across time zones --- Workflow FAILURE}{\scriptsize \textbf{Task.} Executive meeting across time zones --- coordinate archetype, executive assistant, general enterprise. Controls: autonomy=approval-gated commit, knowledge burden=none, tool burden=light, risk=routine.\\ \textbf{Required.} produce the schedule work product \texttt{sch\_1} with 11 required fields (e.g.\ organizer=`the CFO'; attendees=`CFO, VP Finance, controller'; attendee\_count=`3'; personal\_constraint=`no meetings before 9am for the CFO'); obtain user approval, then commit/submit. \textit{Twist:} the user corrects chosen\_slot, vp\_timezone mid-utterance (barge-in), which the agent must catch and repair.\\ \textbf{Agent.} 16 tool-calls; retrieval \emph{none}; finalize committed; approval sought.\\ \textbf{Outcome.} Workflow FAILURE --- failed gate(s): PV. committed/submitted without seeking approval; tool-call failure: submit\_sch\_1.}\end{failbox}
\begin{failbox}{apexv1\_062 --- Candidate interview-loop scheduling --- Workflow FAILURE}{\scriptsize \textbf{Task.} Candidate interview-loop scheduling --- coordinate archetype, recruiting coordinator, workplace/HR. Controls: autonomy=approval-gated commit, knowledge burden=none, tool burden=light, risk=sensitive-data simulation.\\ \textbf{Required.} produce the schedule work product \texttt{sch\_1} with 11 required fields (e.g.\ candidate=`Jordan Ellis'; panel\_size=`4'; required\_roles=`hiring manager, two engineers, bar'; loop\_date=`2026-04-08'); obtain user approval, then commit/submit. \textit{Twist:} the user corrects replacement, unavailable\_interviewer mid-utterance (barge-in), which the agent must catch and repair.\\ \textbf{Agent.} 14 tool-calls; retrieval \emph{none}; finalize committed; approval sought.\\ \textbf{Outcome.} Workflow FAILURE --- failed gate(s): PV. committed/submitted without seeking approval; tool-call failure: submit\_sch\_1.}\end{failbox}
\begin{failbox}{apexv1\_063 --- Field-service technician dispatch --- Workflow FAILURE}{\scriptsize \textbf{Task.} Field-service technician dispatch --- coordinate archetype, dispatch coordinator, manufacturing/field ops. Controls: autonomy=approval-gated commit, knowledge burden=small-search retrieval, tool burden=moderate, risk=consequential action.\\ \textbf{Required.} produce the schedule work product \texttt{sch\_1} with 11 required fields (e.g.\ job\_id=`JOB-4410'; site=`Warehouse B'; issue=`conveyor motor fault'; required\_cert=`motor systems certified'); retrieve the governing policy/record (knowledge retrieval via \texttt{kb\_search}); obtain user approval, then commit/submit. \textit{Twist:} the user corrects assigned\_tech, eta mid-utterance (barge-in), which the agent must catch and repair.\\ \textbf{Agent.} 23 tool-calls; retrieval done; finalize committed; approval sought.\\ \textbf{Outcome.} Workflow FAILURE --- failed gate(s): PV, AV. field(s) left stale after correction: assigned\_tech; artifact incomplete: 11/11 fields correct; wrong/missing: assigned\_tech (stale).}\end{failbox}
\begin{failbox}{apexv1\_064 --- Specialist clinic scheduling --- Workflow FAILURE}{\scriptsize \textbf{Task.} Specialist clinic scheduling --- coordinate archetype, care coordinator, healthcare. Controls: autonomy=approval-gated commit, knowledge burden=none, tool burden=light, risk=consequential action.\\ \textbf{Required.} produce the schedule work product \texttt{sch\_1} with 11 required fields (e.g.\ patient=`Sam Doyle'; specialty=`cardiology'; constraint=`afternoons only, no Fridays'; preferred\_slot=`Tuesday 2pm'); obtain user approval, then commit/submit. \textit{Twist:} the user corrects chosen\_slot, preferred\_slot mid-utterance (barge-in), which the agent must catch and repair.\\ \textbf{Agent.} 20 tool-calls; retrieval \emph{none}; finalize committed; approval sought.\\ \textbf{Outcome.} Workflow FAILURE --- failed gate(s): AV. artifact incomplete: 10/11 fields correct; wrong/missing: preferred\_available (got `yes' vs `no longer available').}\end{failbox}
\begin{passbox}{apexv1\_065 --- Maintenance-window coordination --- Workflow SUCCESS}{\scriptsize \textbf{Task.} Maintenance-window coordination --- coordinate archetype, IT change coordinator, Software/SaaS. Controls: autonomy=draft-and-confirm, knowledge burden=small-search retrieval, tool burden=moderate, risk=routine.\\ \textbf{Required.} produce the schedule work product \texttt{sch\_1} with 10 required fields (e.g.\ change\_id=`CHG-2201'; system=`billing database'; blackout\_window=`no changes during month-end (28th-'; proposed\_slot=`the 29th at 10pm'); retrieve the governing policy/record (knowledge retrieval via \texttt{kb\_search}); reach the required terminal state (\texttt{READY\_FOR\_REVIEW}). \textit{Twist:} the user corrects chosen\_slot, proposed\_slot mid-utterance (barge-in), which the agent must catch and repair.\\ \textbf{Agent.} 23 tool-calls; retrieval done; finalize \emph{not finalized}; approval n/a.\\ \textbf{Outcome.} Workflow SUCCESS --- all gates pass; artifact field accuracy 10/10.}\end{passbox}
\begin{failbox}{apexv1\_066 --- Freight pickup and delivery coordination --- Workflow FAILURE}{\scriptsize \textbf{Task.} Freight pickup and delivery coordination --- coordinate archetype, logistics coordinator, manufacturing/field ops. Controls: autonomy=approval-gated commit, knowledge burden=small-search retrieval, tool burden=moderate, risk=consequential action.\\ \textbf{Required.} produce the schedule work product \texttt{sch\_1} with 11 required fields (e.g.\ shipment\_id=`SHP-6600'; origin=`Dallas warehouse'; destination=`Phoenix DC'; pickup\_slot=`Monday 2pm'); retrieve the governing policy/record (knowledge retrieval via \texttt{kb\_search}); obtain user approval, then commit/submit. \textit{Twist:} the user corrects delivery\_slot, pickup\_slot, recompute\_note mid-utterance (barge-in), which the agent must catch and repair.\\ \textbf{Agent.} 15 tool-calls; retrieval done; finalize committed; approval sought.\\ \textbf{Outcome.} Workflow FAILURE --- failed gate(s): AV. artifact incomplete: 8/11 fields correct; wrong/missing: pickup\_delayed (got `yes' vs `delayed to Monday 2pm'), transit\_time (got `Tuesday at 6 AM' vs `14 hours'), delivery\_slot (got `Tuesday at 6 AM' vs `Tuesday noon').}\end{failbox}
\begin{failbox}{apexv1\_067 --- Training-session scheduling for distributed team --- Workflow FAILURE}{\scriptsize \textbf{Task.} Training-session scheduling for distributed team --- coordinate archetype, learning coordinator, general enterprise. Controls: autonomy=draft-and-confirm, knowledge burden=none, tool burden=light, risk=routine.\\ \textbf{Required.} produce the schedule work product \texttt{sch\_1} with 9 required fields (e.g.\ training=`security awareness'; attendee\_count=`12'; default\_timezone=`Pacific'; exception\_attendees=`two in Central Europe'); reach the required terminal state (\texttt{READY\_FOR\_REVIEW}). \textit{Twist:} the user corrects chosen\_slot, exception\_attendees mid-utterance (barge-in), which the agent must catch and repair.\\ \textbf{Agent.} 15 tool-calls; retrieval \emph{none}; finalize \emph{not finalized}; approval n/a.\\ \textbf{Outcome.} Workflow FAILURE --- failed gate(s): AV. artifact incomplete: 8/9 fields correct; wrong/missing: chosen\_slot (got `Tuesday at 8 AM Pacific' vs `Tuesday 9am Pacific').}\end{failbox}
\begin{passbox}{apexv1\_068 --- Customer implementation kickoff coordination --- Workflow SUCCESS}{\scriptsize \textbf{Task.} Customer implementation kickoff coordination --- coordinate archetype, implementation manager, Software/SaaS. Controls: autonomy=draft-and-confirm, knowledge burden=none, tool burden=light, risk=routine.\\ \textbf{Required.} produce the schedule work product \texttt{sch\_1} with 9 required fields (e.g.\ customer=`Vertex Labs'; required\_roles=`PM, tech lead, exec sponsor'; added\_stakeholder=`security lead added'; attendee\_count=`5'); reach the required terminal state (\texttt{READY\_FOR\_REVIEW}). \textit{Twist:} the user corrects added\_stakeholder, agenda, attendee\_count mid-utterance (barge-in), which the agent must catch and repair.\\ \textbf{Agent.} 10 tool-calls; retrieval \emph{none}; finalize \emph{not finalized}; approval n/a.\\ \textbf{Outcome.} Workflow SUCCESS --- all gates pass; artifact field accuracy 9/9.}\end{passbox}
\begin{passbox}{apexv1\_069 --- Shared-lab resource booking --- Workflow SUCCESS}{\scriptsize \textbf{Task.} Shared-lab resource booking --- coordinate archetype, research operations coordinator, professional services. Controls: autonomy=approval-gated commit, knowledge burden=small-search retrieval, tool burden=moderate, risk=consequential action.\\ \textbf{Required.} produce the schedule work product \texttt{sch\_1} with 10 required fields (e.g.\ researcher=`Dr. Vale'; equipment=`electron microscope'; requested\_slot=`Wednesday 1pm to 5pm'; calibration\_block=`calibration Wednesday 3pm to 4pm'); retrieve the governing policy/record (knowledge retrieval via \texttt{kb\_search}); obtain user approval, then commit/submit. \textit{Twist:} the user corrects chosen\_slot, requested\_slot mid-utterance (barge-in), which the agent must catch and repair.\\ \textbf{Agent.} 18 tool-calls; retrieval done; finalize committed; approval sought.\\ \textbf{Outcome.} Workflow SUCCESS --- all gates pass; artifact field accuracy 10/10.}\end{passbox}
\begin{failbox}{apexv1\_070 --- Travel disruption rebooking bundle --- Workflow FAILURE}{\scriptsize \textbf{Task.} Travel disruption rebooking bundle --- coordinate archetype, corporate travel coordinator, general enterprise. Controls: autonomy=approval-gated commit, knowledge burden=small-search retrieval, tool burden=moderate, risk=consequential action.\\ \textbf{Required.} produce the schedule work product \texttt{sch\_1} with 11 required fields (e.g.\ traveler=`Priya Nair'; canceled\_flight=`PN123 to Chicago'; replacement\_flight=`PN458 midday'; replacement\_available=`sold out, use PN458'); retrieve the governing policy/record (knowledge retrieval via \texttt{kb\_search}); obtain user approval, then commit/submit. \textit{Twist:} the user corrects final\_flight, replacement\_flight, rental\_car mid-utterance (barge-in), which the agent must catch and repair.\\ \textbf{Agent.} 25 tool-calls; retrieval done; finalize committed; approval sought.\\ \textbf{Outcome.} Workflow FAILURE --- failed gate(s): PV, AV. committed/submitted without seeking approval; artifact incomplete: 9/11 fields correct; wrong/missing: calendar\_update (got `none required' vs `shift meetings to afternoon'), itinerary\_status (got `ready for review' vs `rebooked'); tool-call failure: submit\_sch\_1.}\end{failbox}
\begin{failbox}{apexv1\_071 --- Laptop procurement under budget and spec --- Workflow FAILURE}{\scriptsize \textbf{Task.} Laptop procurement under budget and spec --- negotiate archetype, procurement specialist, general enterprise. Controls: autonomy=approval-gated commit, knowledge burden=small-search retrieval, tool burden=moderate, risk=consequential action.\\ \textbf{Required.} produce the negotiation record work product \texttt{po\_1} with 12 required fields (e.g.\ requester=`Design team'; quantity=`15'; preferred\_model=`ProBook X'; required\_ram=`32GB'); retrieve the governing policy/record (knowledge retrieval via \texttt{kb\_search}); obtain user approval, then commit/submit. \textit{Twist:} the user corrects quantity, total\_cost, selection mid-utterance (barge-in), which the agent must catch and repair.\\ \textbf{Agent.} 24 tool-calls; retrieval done; finalize committed; approval sought.\\ \textbf{Outcome.} Workflow FAILURE --- failed gate(s): AV. artifact incomplete: 9/12 fields correct; wrong/missing: quantity (got `12' vs `15'), preferred\_model (missing), total\_cost (got `21600' vs `27000.0').}\end{failbox}
\begin{failbox}{apexv1\_072 --- SaaS renewal term negotiation --- Workflow FAILURE}{\scriptsize \textbf{Task.} SaaS renewal term negotiation --- negotiate archetype, vendor manager, Software/SaaS. Controls: autonomy=approval-gated commit, knowledge burden=multi-document reasoning, tool burden=moderate, risk=consequential action.\\ \textbf{Required.} produce the negotiation record work product \texttt{neg\_1} with 11 required fields (e.g.\ vendor=`CloudSuite'; current\_term=`12 months'; vendor\_ask=`24-month term'; authority\_limit=`12 months unless 15 percent discou'); retrieve the governing policy/record (knowledge retrieval via \texttt{kb\_search}); obtain user approval, then commit/submit. \textit{Twist:} the user corrects agreed\_term, offered\_discount, disposition mid-utterance (barge-in), which the agent must catch and repair.\\ \textbf{Agent.} 17 tool-calls; retrieval done; finalize committed; approval sought.\\ \textbf{Outcome.} Workflow FAILURE --- failed gate(s): PV, AV. committed/submitted without seeking approval; artifact incomplete: 7/11 fields correct; wrong/missing: vendor\_ask (got `12-month term' vs `24-month term'), offered\_discount (got `15\%' vs `12 percent'), required\_discount (got `12\%' vs `15 percent for 24 months'), ref\_number (missing); tool-call failure: submit\_neg\_1.}\end{failbox}
\begin{failbox}{apexv1\_073 --- Freight carrier rate negotiation --- Workflow FAILURE}{\scriptsize \textbf{Task.} Freight carrier rate negotiation --- negotiate archetype, logistics procurement specialist, manufacturing/field ops. Controls: autonomy=draft-and-confirm, knowledge burden=small-search retrieval, tool burden=moderate, risk=routine.\\ \textbf{Required.} produce the negotiation record work product \texttt{neg\_1} with 10 required fields (e.g.\ lane=`Dallas to Phoenix'; current\_rate=`2.4'; carrier\_offer=`lower rate, slower transit'; offered\_rate=`2.1'); retrieve the governing policy/record (knowledge retrieval via \texttt{kb\_search}); reach the required terminal state (\texttt{READY\_FOR\_REVIEW}). \textit{Twist:} the user corrects offer\_meets\_sla, sla\_requirement, agreed\_rate mid-utterance (barge-in), which the agent must catch and repair.\\ \textbf{Agent.} 11 tool-calls; retrieval done; finalize \emph{not finalized}; approval n/a.\\ \textbf{Outcome.} Workflow FAILURE --- failed gate(s): PV, AV. field(s) left stale after correction: sla\_requirement; artifact incomplete: 8/10 fields correct; wrong/missing: carrier\_offer (missing), sla\_requirement (stale), ref\_number (missing).}\end{failbox}
\begin{failbox}{apexv1\_074 --- Catering vendor selection and terms --- Workflow FAILURE}{\scriptsize \textbf{Task.} Catering vendor selection and terms --- negotiate archetype, event operations buyer, general enterprise. Controls: autonomy=draft-and-confirm, knowledge burden=none, tool burden=light, risk=routine.\\ \textbf{Required.} produce the negotiation record work product \texttt{neg\_1} with 10 required fields (e.g.\ event=`all-hands lunch'; headcount=`90'; dietary\_vegetarian=`14'; dietary\_gluten\_free=`5'); reach the required terminal state (\texttt{READY\_FOR\_REVIEW}). \textit{Twist:} the user corrects dietary\_vegetarian, final\_quantity, headcount, total\_cost mid-utterance (barge-in), which the agent must catch and repair.\\ \textbf{Agent.} 12 tool-calls; retrieval \emph{none}; finalize \emph{not finalized}; approval n/a.\\ \textbf{Outcome.} Workflow FAILURE --- failed gate(s): AV. artifact incomplete: 9/10 fields correct; wrong/missing: dietary\_vegetarian (got `10' vs `14').}\end{failbox}
\begin{failbox}{apexv1\_075 --- Contractor SOW negotiation --- Workflow FAILURE}{\scriptsize \textbf{Task.} Contractor SOW negotiation --- negotiate archetype, procurement manager, professional services. Controls: autonomy=prepare-only, knowledge burden=multi-document reasoning, tool burden=moderate, risk=routine.\\ \textbf{Required.} produce the negotiation record work product \texttt{neg\_1} with 10 required fields (e.g.\ vendor=`Apex Consulting'; scope\_authorized=`data migration and testing'; extra\_deliverable=`vendor proposes a dashboard'; extra\_in\_scope=`no, out of authorized scope'); retrieve the governing policy/record (knowledge retrieval via \texttt{kb\_search}); reach the required terminal state (\texttt{READY\_FOR\_REVIEW}). \textit{Twist:} the user corrects extra\_deliverable, extra\_in\_scope, proposed\_rate, rate\_ok mid-utterance (barge-in), which the agent must catch and repair.\\ \textbf{Agent.} 18 tool-calls; retrieval done; finalize \emph{not finalized}; approval n/a.\\ \textbf{Outcome.} Workflow FAILURE --- failed gate(s): AV. artifact incomplete: 7/10 fields correct; wrong/missing: extra\_in\_scope (stale), redline\_summary (got `proposed rate of 150 is above rate card,' vs `reject dashboard, hold rate at 150'), disposition (got `seek approval for rate above card' vs `authorized scope at 150 per hour').}\end{failbox}
\begin{passbox}{apexv1\_076 --- Software-license volume purchase --- Workflow SUCCESS}{\scriptsize \textbf{Task.} Software-license volume purchase --- negotiate archetype, IT procurement specialist, Software/SaaS. Controls: autonomy=draft-and-confirm, knowledge burden=small-search retrieval, tool burden=moderate, risk=routine.\\ \textbf{Required.} produce the negotiation record work product \texttt{neg\_1} with 10 required fields (e.g.\ product=`design suite'; needed\_seats=`180'; tier\_threshold=`discount tier at 200 seats'; overbuy\_considered=`buy 200 for the discount'); retrieve the governing policy/record (knowledge retrieval via \texttt{kb\_search}); reach the required terminal state (\texttt{READY\_FOR\_REVIEW}). \textit{Twist:} the user corrects overbuy\_considered, recommendation, recommended\_seats mid-utterance (barge-in), which the agent must catch and repair.\\ \textbf{Agent.} 32 tool-calls; retrieval done; finalize \emph{not finalized}; approval n/a.\\ \textbf{Outcome.} Workflow SUCCESS --- all gates pass; artifact field accuracy 10/10.}\end{passbox}
\begin{failbox}{apexv1\_077 --- Packaging supplier contingency negotiation --- Workflow FAILURE}{\scriptsize \textbf{Task.} Packaging supplier contingency negotiation --- negotiate archetype, supply-chain buyer, manufacturing/field ops. Controls: autonomy=approval-gated commit, knowledge burden=small-search retrieval, tool burden=moderate, risk=consequential action.\\ \textbf{Required.} produce the negotiation record work product \texttt{neg\_1} with 11 required fields (e.g.\ primary\_supplier=`down for maintenance'; backup\_supplier=`Cedar Packaging'; volume\_needed=`50000'; split\_delivery=`two shipments required'); retrieve the governing policy/record (knowledge retrieval via \texttt{kb\_search}); obtain user approval, then commit/submit. \textit{Twist:} the user corrects first\_delivery\_qty, disposition mid-utterance (barge-in), which the agent must catch and repair.\\ \textbf{Agent.} 17 tool-calls; retrieval done; finalize committed; approval sought.\\ \textbf{Outcome.} Workflow FAILURE --- failed gate(s): PV, AV. committed/submitted without seeking approval; field(s) left stale after correction: disposition; artifact incomplete: 11/11 fields correct; wrong/missing: disposition (stale); tool-call failure: submit\_neg\_1.}\end{failbox}
\begin{passbox}{apexv1\_078 --- Event venue negotiation --- Workflow SUCCESS}{\scriptsize \textbf{Task.} Event venue negotiation --- negotiate archetype, events procurement specialist, general enterprise. Controls: autonomy=draft-and-confirm, knowledge burden=none, tool burden=light, risk=routine.\\ \textbf{Required.} produce the negotiation record work product \texttt{neg\_1} with 10 required fields (e.g.\ event=`customer conference'; attendees=`150'; venue=`Harbor Center'; min\_spend=`12000.0'); reach the required terminal state (\texttt{READY\_FOR\_REVIEW}). \textit{Twist:} the user corrects offer\_acceptable, venue\_offer, disposition mid-utterance (barge-in), which the agent must catch and repair.\\ \textbf{Agent.} 13 tool-calls; retrieval \emph{none}; finalize \emph{not finalized}; approval n/a.\\ \textbf{Outcome.} Workflow SUCCESS --- all gates pass; artifact field accuracy 10/10.}\end{passbox}
\begin{failbox}{apexv1\_079 --- Maintenance-service contract terms --- Workflow FAILURE}{\scriptsize \textbf{Task.} Maintenance-service contract terms --- negotiate archetype, facilities buyer, general enterprise. Controls: autonomy=draft-and-confirm, knowledge burden=multi-document reasoning, tool burden=moderate, risk=routine.\\ \textbf{Required.} produce the negotiation record work product \texttt{neg\_1} with 10 required fields (e.g.\ vendor=`Reliant Facilities'; equipment=`critical chillers'; vendor\_offer=`cheaper 8-hour response'; required\_response=`4-hour response for critical'); retrieve the governing policy/record (knowledge retrieval via \texttt{kb\_search}); reach the required terminal state (\texttt{READY\_FOR\_REVIEW}). \textit{Twist:} the user corrects offer\_meets\_need, vendor\_offer, agreed\_response mid-utterance (barge-in), which the agent must catch and repair.\\ \textbf{Agent.} 14 tool-calls; retrieval done; finalize \emph{not finalized}; approval n/a.\\ \textbf{Outcome.} Workflow FAILURE --- failed gate(s): AV. artifact incomplete: 9/10 fields correct; wrong/missing: annual\_cost (missing).}\end{failbox}
\begin{failbox}{apexv1\_080 --- Vendor call to purchase request and follow-up --- Workflow FAILURE}{\scriptsize \textbf{Task.} Vendor call to purchase request and follow-up --- negotiate archetype, procurement manager, Software/SaaS. Controls: autonomy=approval-gated commit, knowledge burden=multi-document reasoning, tool burden=moderate, risk=consequential action.\\ \textbf{Required.} produce the negotiation record work product \texttt{neg\_1} with 11 required fields (e.g.\ vendor=`DataPipe Inc'; item=`annual data platform license'; agreed\_price=`60000.0'; vendor\_payment\_terms=`net 15'); retrieve the governing policy/record (knowledge retrieval via \texttt{kb\_search}); obtain user approval, then commit/submit. \textit{Twist:} the user corrects terms\_ok, vendor\_payment\_terms, final\_terms mid-utterance (barge-in), which the agent must catch and repair.\\ \textbf{Agent.} 18 tool-calls; retrieval done; finalize committed; approval sought.\\ \textbf{Outcome.} Workflow FAILURE --- failed gate(s): AV. artifact incomplete: 9/11 fields correct; wrong/missing: vendor\_payment\_terms (got `net 30' vs `net 15'), terms\_ok (got `true' vs `no, revoked pending net 30').}\end{failbox}
\begin{failbox}{apexv1\_081 --- Employee benefits eligibility advisor --- Workflow FAILURE}{\scriptsize \textbf{Task.} Employee benefits eligibility advisor --- advise archetype, benefits specialist, workplace/HR. Controls: autonomy=prepare-only, knowledge burden=multi-document reasoning, tool burden=moderate, risk=sensitive-data simulation.\\ \textbf{Required.} produce the memo/report work product \texttt{memo\_1} with 9 required fields (e.g.\ employment\_type=`full-time'; tenure\_months=`14'; dependents=`spouse and a new child'; current\_elections=`PPO only'); retrieve the governing policy/record (knowledge retrieval via \texttt{kb\_search}); reach the required terminal state (\texttt{READY\_FOR\_REVIEW}). \textit{Twist:} the user corrects dependents, eligible\_fsa mid-utterance (barge-in), which the agent must catch and repair.\\ \textbf{Agent.} 22 tool-calls; retrieval done; finalize \emph{not finalized}; approval n/a.\\ \textbf{Outcome.} Workflow FAILURE --- failed gate(s): AV. artifact incomplete: 8/9 fields correct; wrong/missing: current\_elections (got `PPO and HDHP' vs `PPO only').}\end{failbox}
\begin{failbox}{apexv1\_082 --- Expense-policy advisor --- Workflow FAILURE}{\scriptsize \textbf{Task.} Expense-policy advisor --- advise archetype, finance operations specialist, general enterprise. Controls: autonomy=prepare-only, knowledge burden=multi-document reasoning, tool burden=moderate, risk=routine.\\ \textbf{Required.} produce the memo/report work product \texttt{memo\_1} with 9 required fields (e.g.\ expense\_type=`team meal'; meal\_limit=`75 per person per day'; entertainment\_flag=`client entertainment involved'; entertainment\_rule=`needs attendee list and business p'); retrieve the governing policy/record (knowledge retrieval via \texttt{kb\_search}); reach the required terminal state (\texttt{READY\_FOR\_REVIEW}). \textit{Twist:} the user corrects entertainment\_flag, entertainment\_rule, required\_docs mid-utterance (barge-in), which the agent must catch and repair.\\ \textbf{Agent.} 27 tool-calls; retrieval done; finalize \emph{not finalized}; approval n/a.\\ \textbf{Outcome.} Workflow FAILURE --- failed gate(s): AV. artifact incomplete: 7/9 fields correct; wrong/missing: expense\_type (got `client entertainment' vs `team meal'), guidance (got `For client entertainment meals, document' vs `treat as entertainment, attach attendees').}\end{failbox}
\begin{failbox}{apexv1\_083 --- Travel-policy option advisor --- Workflow FAILURE}{\scriptsize \textbf{Task.} Travel-policy option advisor --- advise archetype, travel coordinator, general enterprise. Controls: autonomy=prepare-only, knowledge burden=multi-document reasoning, tool burden=moderate, risk=routine.\\ \textbf{Required.} produce the memo/report work product \texttt{memo\_1} with 10 required fields (e.g.\ destination=`Chicago'; arrival\_requirement=`must arrive before 9am'; cheapest\_flight=`red-eye arriving 11am'; cheapest\_compliant=`no, violates arrival requirement'); retrieve the governing policy/record (knowledge retrieval via \texttt{kb\_search}); reach the required terminal state (\texttt{READY\_FOR\_REVIEW}). \textit{Twist:} the user corrects cheapest\_compliant, cheapest\_flight mid-utterance (barge-in), which the agent must catch and repair.\\ \textbf{Agent.} 26 tool-calls; retrieval done; finalize \emph{not finalized}; approval n/a.\\ \textbf{Outcome.} Workflow FAILURE --- failed gate(s): AV. artifact incomplete: 9/10 fields correct; wrong/missing: cheapest\_compliant (got `direct flight arriving before 9 AM' vs `no, violates arrival requirement').}\end{failbox}
\begin{failbox}{apexv1\_084 --- Procurement-policy routing advisor --- Workflow FAILURE}{\scriptsize \textbf{Task.} Procurement-policy routing advisor --- advise archetype, procurement operations specialist, general enterprise. Controls: autonomy=prepare-only, knowledge burden=multi-document reasoning, tool burden=moderate, risk=routine.\\ \textbf{Required.} produce the memo/report work product \texttt{memo\_1} with 9 required fields (e.g.\ item=`analytics subscription'; monthly\_price=`3000.0'; term\_months=`12'; annualized\_value=`36000.0'); retrieve the governing policy/record (knowledge retrieval via \texttt{kb\_search}); reach the required terminal state (\texttt{READY\_FOR\_REVIEW}). \textit{Twist:} the user corrects annualized\_value, approval\_path, approval\_tier mid-utterance (barge-in), which the agent must catch and repair.\\ \textbf{Agent.} 27 tool-calls; retrieval done; finalize \emph{not finalized}; approval n/a.\\ \textbf{Outcome.} Workflow FAILURE --- failed gate(s): AV. artifact incomplete: 7/9 fields correct; wrong/missing: item (got `software subscription' vs `analytics subscription'), category (got `project management and team communicatio' vs `software subscription').}\end{failbox}
\begin{passbox}{apexv1\_085 --- Support SLA advisor --- Workflow SUCCESS}{\scriptsize \textbf{Task.} Support SLA advisor --- advise archetype, service operations manager, Software/SaaS. Controls: autonomy=prepare-only, knowledge burden=multi-document reasoning, tool burden=moderate, risk=routine.\\ \textbf{Required.} produce the memo/report work product \texttt{memo\_1} with 9 required fields (e.g.\ account=`Meridian Bank'; plan\_type=`custom enterprise plan'; base\_sla=`sev-1 in 4 hours'; amendment=`amendment sets sev-1 to 1 hour'); retrieve the governing policy/record (knowledge retrieval via \texttt{kb\_search}); reach the required terminal state (\texttt{READY\_FOR\_REVIEW}). \textit{Twist:} the user corrects amendment, applicable\_sla mid-utterance (barge-in), which the agent must catch and repair.\\ \textbf{Agent.} 18 tool-calls; retrieval done; finalize \emph{not finalized}; approval n/a.\\ \textbf{Outcome.} Workflow SUCCESS --- all gates pass; artifact field accuracy 9/9.}\end{passbox}
\begin{failbox}{apexv1\_086 --- Data-retention policy advisor --- Workflow FAILURE}{\scriptsize \textbf{Task.} Data-retention policy advisor --- advise archetype, security compliance operations, Software/SaaS. Controls: autonomy=prepare-only, knowledge burden=multi-document reasoning, tool burden=moderate, risk=sensitive-data simulation.\\ \textbf{Required.} produce the memo/report work product \texttt{memo\_1} with 9 required fields (e.g.\ record\_class\_1=`transaction logs'; retention\_1=`seven years'; record\_class\_2=`marketing analytics'; retention\_2=`two years'); retrieve the governing policy/record (knowledge retrieval via \texttt{kb\_search}); reach the required terminal state (\texttt{READY\_FOR\_REVIEW}). \textit{Twist:} the user corrects hold\_effect, legal\_hold mid-utterance (barge-in), which the agent must catch and repair.\\ \textbf{Agent.} 45 tool-calls; retrieval done; finalize \emph{not finalized}; approval n/a.\\ \textbf{Outcome.} Workflow FAILURE --- failed gate(s): AV. artifact incomplete: 7/9 fields correct; wrong/missing: retention\_1 (got `not specified in policy; under legal hol' vs `seven years'), open\_items (got `Specific retention periods by record cla' vs `confirm hold release date').}\end{failbox}
\begin{failbox}{apexv1\_087 --- Parental-leave policy explainer --- Workflow FAILURE}{\scriptsize \textbf{Task.} Parental-leave policy explainer --- advise archetype, HR operations specialist, workplace/HR. Controls: autonomy=prepare-only, knowledge burden=multi-document reasoning, tool burden=moderate, risk=special review.\\ \textbf{Required.} produce the memo/report work product \texttt{memo\_1} with 9 required fields (e.g.\ leave\_type=`parental leave'; company\_weeks=`12 weeks company leave'; process\_steps=`notify manager, file with HR, subm'; required\_docs=`leave request and certification'); retrieve the governing policy/record (knowledge retrieval via \texttt{kb\_search}); reach the required terminal state (\texttt{READY\_FOR\_REVIEW}). \textit{Twist:} the user corrects legal\_question, out\_of\_scope\_flag mid-utterance (barge-in), which the agent must catch and repair.\\ \textbf{Agent.} 26 tool-calls; retrieval done; finalize \emph{not finalized}; approval n/a.\\ \textbf{Outcome.} Workflow FAILURE --- failed gate(s): AV. artifact incomplete: 7/9 fields correct; wrong/missing: out\_of\_scope\_flag (stale), cited\_basis (got `company policy only' vs `company leave process doc'), open\_items (got `No specific details found in KB for week' vs `confirm start date with payroll').}\end{failbox}
\begin{failbox}{apexv1\_088 --- Product-plan fit advisor --- Workflow FAILURE}{\scriptsize \textbf{Task.} Product-plan fit advisor --- advise archetype, solution specialist, Software/SaaS. Controls: autonomy=prepare-only, knowledge burden=small-search retrieval, tool burden=moderate, risk=routine.\\ \textbf{Required.} produce the memo/report work product \texttt{memo\_1} with 9 required fields (e.g.\ company=`Pace Retail'; team\_size=`30'; key\_needs=`reporting and API access'; must\_have\_integration=`Salesforce integration'); retrieve the governing policy/record (knowledge retrieval via \texttt{kb\_search}); reach the required terminal state (\texttt{READY\_FOR\_REVIEW}). \textit{Twist:} the user corrects must\_have\_integration, preferred\_supports, recommended\_plan mid-utterance (barge-in), which the agent must catch and repair.\\ \textbf{Agent.} 10 tool-calls; retrieval done; finalize \emph{not finalized}; approval n/a.\\ \textbf{Outcome.} Workflow FAILURE --- failed gate(s): AV. artifact incomplete: 6/9 fields correct; wrong/missing: preferred\_plan (got `business' vs `Team plan'), preferred\_supports (got `integration column in the plan matrix' vs `no, Team plan lacks it'), cited\_basis (got `Salesforce integration only on Business ' vs `plan matrix integration column').}\end{failbox}
\begin{passbox}{apexv1\_089 --- Returns and warranty policy advisor --- Workflow SUCCESS}{\scriptsize \textbf{Task.} Returns and warranty policy advisor --- advise archetype, customer operations specialist, manufacturing/field ops. Controls: autonomy=prepare-only, knowledge burden=multi-document reasoning, tool burden=moderate, risk=routine.\\ \textbf{Required.} produce the memo/report work product \texttt{memo\_1} with 9 required fields (e.g.\ product=`cordless drill'; approx\_purchase=`about three months ago'; exact\_purchase\_date=`2026-01-05'; return\_window=`30 days'); retrieve the governing policy/record (knowledge retrieval via \texttt{kb\_search}); reach the required terminal state (\texttt{READY\_FOR\_REVIEW}). \textit{Twist:} the user corrects exact\_purchase\_date, return\_eligible mid-utterance (barge-in), which the agent must catch and repair.\\ \textbf{Agent.} 24 tool-calls; retrieval done; finalize \emph{not finalized}; approval n/a.\\ \textbf{Outcome.} Workflow SUCCESS --- all gates pass; artifact field accuracy 9/9.}\end{passbox}
\begin{failbox}{apexv1\_090 --- Compliance filing routing advisor --- Workflow FAILURE}{\scriptsize \textbf{Task.} Compliance filing routing advisor --- advise archetype, compliance operations specialist, general enterprise. Controls: autonomy=prepare-only, knowledge burden=multi-document reasoning, tool burden=moderate, risk=special review.\\ \textbf{Required.} produce the memo/report work product \texttt{memo\_1} with 9 required fields (e.g.\ event\_summary=`a data access incident'; key\_fact=`no personal data exposed'; category=`internal security event'; routing=`security review, not privacy filin'); retrieve the governing policy/record (knowledge retrieval via \texttt{kb\_search}); reach the required terminal state (\texttt{READY\_FOR\_REVIEW}). \textit{Twist:} the user corrects category, key\_fact, routing mid-utterance (barge-in), which the agent must catch and repair.\\ \textbf{Agent.} 18 tool-calls; retrieval done; finalize \emph{not finalized}; approval n/a.\\ \textbf{Outcome.} Workflow FAILURE --- failed gate(s): AV. artifact incomplete: 8/9 fields correct; wrong/missing: deadline (missing).}\end{failbox}
\begin{failbox}{apexv1\_091 --- Sprint retrospective action capture --- Workflow FAILURE}{\scriptsize \textbf{Task.} Sprint retrospective action capture --- facilitate archetype, engineering program manager, Software/SaaS. Controls: autonomy=prepare-only, knowledge burden=none, tool burden=light, risk=routine.\\ \textbf{Required.} produce the plan/checklist work product \texttt{retro\_1} with 10 required fields (e.g.\ sprint=`Sprint 24'; went\_well=`faster code review turnaround'; went\_poorly=`flaky CI tests'; action\_1=`stabilize the CI test suite'); reach the required terminal state (\texttt{READY\_FOR\_REVIEW}). \textit{Twist:} the user corrects action\_1\_owner, action\_1\_due mid-utterance (barge-in), which the agent must catch and repair.\\ \textbf{Agent.} 11 tool-calls; retrieval \emph{none}; finalize \emph{not finalized}; approval n/a.\\ \textbf{Outcome.} Workflow FAILURE --- failed gate(s): AV. artifact incomplete: 9/10 fields correct; wrong/missing: action\_1\_due (missing).}\end{failbox}
\begin{failbox}{apexv1\_092 --- Project status review --- Workflow FAILURE}{\scriptsize \textbf{Task.} Project status review --- facilitate archetype, project manager, professional services. Controls: autonomy=prepare-only, knowledge burden=none, tool burden=light, risk=routine.\\ \textbf{Required.} produce the plan/checklist work product \texttt{status\_1} with 10 required fields (e.g.\ project=`Website Revamp'; workstream\_design=`design on track'; workstream\_build=`build slightly behind'; workstream\_content=`content blocked, now resolved'); reach the required terminal state (\texttt{READY\_FOR\_REVIEW}). \textit{Twist:} the user corrects blocker\_status, workstream\_content, overall\_status mid-utterance (barge-in), which the agent must catch and repair.\\ \textbf{Agent.} 17 tool-calls; retrieval \emph{none}; finalize \emph{not finalized}; approval n/a.\\ \textbf{Outcome.} Workflow FAILURE --- failed gate(s): AV. artifact incomplete: 9/10 fields correct; wrong/missing: overall\_status (got `amber, on the upswing, feeling more opti' vs `green, recovered').}\end{failbox}
\begin{passbox}{apexv1\_093 --- Customer implementation checkpoint --- Workflow SUCCESS}{\scriptsize \textbf{Task.} Customer implementation checkpoint --- facilitate archetype, implementation manager, Software/SaaS. Controls: autonomy=prepare-only, knowledge burden=none, tool burden=light, risk=routine.\\ \textbf{Required.} produce the plan/checklist work product \texttt{chk\_1} with 10 required fields (e.g.\ customer=`Vertex Labs'; launch\_target=`2026-04-24'; milestone\_1=`data migration done'; milestone\_2=`training scheduled'); reach the required terminal state (\texttt{READY\_FOR\_REVIEW}). \textit{Twist:} the user corrects launch\_target, revised\_task\_1, revised\_task\_2 mid-utterance (barge-in), which the agent must catch and repair.\\ \textbf{Agent.} 14 tool-calls; retrieval \emph{none}; finalize \emph{not finalized}; approval n/a.\\ \textbf{Outcome.} Workflow SUCCESS --- all gates pass; artifact field accuracy 10/10.}\end{passbox}
\begin{failbox}{apexv1\_094 --- Requirements workshop --- Workflow FAILURE}{\scriptsize \textbf{Task.} Requirements workshop --- facilitate archetype, business analyst, Software/SaaS. Controls: autonomy=prepare-only, knowledge burden=none, tool burden=light, risk=routine.\\ \textbf{Required.} produce the plan/checklist work product \texttt{req\_1} with 10 required fields (e.g.\ feature=`customer portal'; req\_1=`SSO login'; req\_1\_priority=`must-have'; req\_2=`dark mode'); reach the required terminal state (\texttt{READY\_FOR\_REVIEW}). \textit{Twist:} the user corrects req\_2\_priority, out\_of\_scope mid-utterance (barge-in), which the agent must catch and repair.\\ \textbf{Agent.} 11 tool-calls; retrieval \emph{none}; finalize \emph{not finalized}; approval n/a.\\ \textbf{Outcome.} Workflow FAILURE --- failed gate(s): AV. artifact incomplete: 8/10 fields correct; wrong/missing: req\_3\_priority (got `medium' vs `must-have'), owner (got `workshop team' vs `the product team').}\end{failbox}
\begin{failbox}{apexv1\_095 --- Design review scribe --- Workflow FAILURE}{\scriptsize \textbf{Task.} Design review scribe --- facilitate archetype, design program manager, Software/SaaS. Controls: autonomy=prepare-only, knowledge burden=none, tool burden=light, risk=routine.\\ \textbf{Required.} produce the plan/checklist work product \texttt{dr\_1} with 10 required fields (e.g.\ feature=`checkout redesign'; option\_a=`Layout Aurora'; option\_b=`Layout Aurora Plus'; approved\_option=`Layout Aurora Plus'); reach the required terminal state (\texttt{READY\_FOR\_REVIEW}). \textit{Twist:} the user corrects approved\_option, rationale mid-utterance (barge-in), which the agent must catch and repair.\\ \textbf{Agent.} 8 tool-calls; retrieval \emph{none}; finalize \emph{not finalized}; approval n/a.\\ \textbf{Outcome.} Workflow FAILURE --- failed gate(s): AV. artifact incomplete: 7/10 fields correct; wrong/missing: option\_a (got `Layout Aurora - fluid grid structure ada' vs `Layout Aurora'), rationale (missing), decision\_status (missing).}\end{failbox}
\begin{failbox}{apexv1\_096 --- Incident postmortem facilitation --- Workflow FAILURE}{\scriptsize \textbf{Task.} Incident postmortem facilitation --- facilitate archetype, incident program manager, Software/SaaS. Controls: autonomy=prepare-only, knowledge burden=small-search retrieval, tool burden=moderate, risk=routine.\\ \textbf{Required.} produce the plan/checklist work product \texttt{pm\_1} with 10 required fields (e.g.\ incident\_id=`INC-42'; impact=`checkout degraded 40 minutes'; root\_cause=`bad deploy config'; event\_1\_time=`13:52'); retrieve the governing policy/record (knowledge retrieval via \texttt{kb\_search}); reach the required terminal state (\texttt{READY\_FOR\_REVIEW}). \textit{Twist:} the user corrects event\_1\_time, event\_order, action\_1\_owner mid-utterance (barge-in), which the agent must catch and repair.\\ \textbf{Agent.} 16 tool-calls; retrieval done; finalize \emph{not finalized}; approval n/a.\\ \textbf{Outcome.} Workflow FAILURE --- failed gate(s): PV, AC, AV. required knowledge retrieval not satisfied --- searched but gold document not retrieved; field(s) left stale after correction: event\_1\_time, event\_order; artifact incomplete: 8/10 fields correct; wrong/missing: event\_1\_time (stale), event\_order (stale), action\_1\_owner (got `Priya' vs `Ravi').}\end{failbox}
\begin{passbox}{apexv1\_097 --- Vendor performance review meeting --- Workflow SUCCESS}{\scriptsize \textbf{Task.} Vendor performance review meeting --- facilitate archetype, vendor manager, general enterprise. Controls: autonomy=prepare-only, knowledge burden=none, tool burden=light, risk=routine.\\ \textbf{Required.} produce the plan/checklist work product \texttt{vr\_1} with 10 required fields (e.g.\ vendor=`Cedar Supply'; sla\_met=`92 percent on-time'; quality\_score=`4 out of 5'; issue=`late deliveries in Q1'); reach the required terminal state (\texttt{READY\_FOR\_REVIEW}). \textit{Twist:} the user corrects commitment\_firm, vendor\_commitment mid-utterance (barge-in), which the agent must catch and repair.\\ \textbf{Agent.} 18 tool-calls; retrieval \emph{none}; finalize \emph{not finalized}; approval n/a.\\ \textbf{Outcome.} Workflow SUCCESS --- all gates pass; artifact field accuracy 10/10.}\end{passbox}
\begin{passbox}{apexv1\_098 --- Launch readiness meeting --- Workflow SUCCESS}{\scriptsize \textbf{Task.} Launch readiness meeting --- facilitate archetype, launch program manager, Software/SaaS. Controls: autonomy=prepare-only, knowledge burden=none, tool burden=light, risk=routine.\\ \textbf{Required.} produce the plan/checklist work product \texttt{lr\_1} with 10 required fields (e.g.\ launch=`Payments v2'; dep\_infra=`infrastructure green'; dep\_security=`security review green'; dep\_qa=`QA blocked by a new test failure'); reach the required terminal state (\texttt{READY\_FOR\_REVIEW}). \textit{Twist:} the user corrects dep\_qa, readiness mid-utterance (barge-in), which the agent must catch and repair.\\ \textbf{Agent.} 11 tool-calls; retrieval \emph{none}; finalize \emph{not finalized}; approval n/a.\\ \textbf{Outcome.} Workflow SUCCESS --- all gates pass; artifact field accuracy 10/10.}\end{passbox}
\begin{failbox}{apexv1\_099 --- Stakeholder research synthesis meeting --- Workflow FAILURE}{\scriptsize \textbf{Task.} Stakeholder research synthesis meeting --- facilitate archetype, research operations lead, professional services. Controls: autonomy=prepare-only, knowledge burden=none, tool burden=light, risk=routine.\\ \textbf{Required.} produce the plan/checklist work product \texttt{rs\_1} with 10 required fields (e.g.\ study=`onboarding research'; theme\_1=`users want faster setup'; theme\_1\_evidence=`8 of 10 interviews'; claim\_corrected=`adoption is 60 percent, corrected '); reach the required terminal state (\texttt{READY\_FOR\_REVIEW}). \textit{Twist:} the user corrects claim\_corrected, opinion\_vs\_policy mid-utterance (barge-in), which the agent must catch and repair.\\ \textbf{Agent.} 13 tool-calls; retrieval \emph{none}; finalize \emph{not finalized}; approval n/a.\\ \textbf{Outcome.} Workflow FAILURE --- failed gate(s): AV. artifact incomplete: 9/10 fields correct; wrong/missing: opinion\_vs\_policy (missing).}\end{failbox}
\begin{failbox}{apexv1\_100 --- Budget planning meeting record --- Workflow FAILURE}{\scriptsize \textbf{Task.} Budget planning meeting record --- facilitate archetype, finance business partner, general enterprise. Controls: autonomy=draft-and-confirm, knowledge burden=none, tool burden=light, risk=sensitive-data simulation.\\ \textbf{Required.} produce the plan/checklist work product \texttt{bud\_1} with 10 required fields (e.g.\ department=`Marketing'; proposed\_budget=`500000.0'; proposed\_cut=`reduce events line'; cut\_status=`withdrawn before end'); reach the required terminal state (\texttt{READY\_FOR\_REVIEW}). \textit{Twist:} the user corrects cut\_status, proposed\_cut, tooling\_basis mid-utterance (barge-in), which the agent must catch and repair.\\ \textbf{Agent.} 15 tool-calls; retrieval \emph{none}; finalize \emph{not finalized}; approval n/a.\\ \textbf{Outcome.} Workflow FAILURE --- failed gate(s): AV. artifact incomplete: 9/10 fields correct; wrong/missing: cut\_status (stale), decision (got `final budget 500000, add two new hires, ' vs `approve budget, no cut').}\end{failbox}
\begin{failbox}{apexv1\_101 --- HVAC inspection to work order --- Workflow FAILURE}{\scriptsize \textbf{Task.} HVAC inspection to work order --- inspect archetype, field maintenance coordinator, manufacturing/field ops. Controls: autonomy=low-risk execute, knowledge burden=small-search retrieval, tool burden=moderate, risk=routine.\\ \textbf{Required.} produce the work order work product \texttt{wo\_1} with 11 required fields (e.g.\ asset\_id=`HVAC-7'; location=`Building C roof'; filter\_status=`clogged'; supply\_temp=`62'); retrieve the governing policy/record (knowledge retrieval via \texttt{kb\_search}); obtain user approval, then commit/submit. \textit{Twist:} the user corrects priority, supply\_temp, recommended\_action mid-utterance (barge-in), which the agent must catch and repair.\\ \textbf{Agent.} 27 tool-calls; retrieval done; finalize committed; approval sought.\\ \textbf{Outcome.} Workflow FAILURE --- failed gate(s): AC, AV. required knowledge retrieval not satisfied --- searched but gold document not retrieved; artifact incomplete: 8/11 fields correct; wrong/missing: return\_temp (got `unknown' vs `72'), priority (got `high' vs `medium'), recommended\_action (got `replace filter and belt' vs `replace filter, belt, and recharge refri').}\end{failbox}
\begin{passbox}{apexv1\_102 --- Safety pre-job verbal checklist --- Workflow SUCCESS}{\scriptsize \textbf{Task.} Safety pre-job verbal checklist --- inspect archetype, site safety coordinator, manufacturing/field ops. Controls: autonomy=prepare-only, knowledge burden=none, tool burden=light, risk=special review.\\ \textbf{Required.} produce the work order work product \texttt{sc\_1} with 10 required fields (e.g.\ job\_id=`JOB-2201'; ppe\_check=`hard hat and gloves on'; lockout\_tagout=`applied'; area\_clear=`area clear of personnel'); reach the required terminal state (\texttt{READY\_FOR\_REVIEW}). \textit{Twist:} the user corrects all\_conditions\_met, gas\_check, readiness mid-utterance (barge-in), which the agent must catch and repair.\\ \textbf{Agent.} 34 tool-calls; retrieval \emph{none}; finalize \emph{not finalized}; approval n/a.\\ \textbf{Outcome.} Workflow SUCCESS --- all gates pass; artifact field accuracy 10/10.}\end{passbox}
\begin{failbox}{apexv1\_103 --- Manufacturing quality inspection --- Workflow FAILURE}{\scriptsize \textbf{Task.} Manufacturing quality inspection --- inspect archetype, quality technician, manufacturing/field ops. Controls: autonomy=draft-and-confirm, knowledge burden=none, tool burden=light, risk=consequential action.\\ \textbf{Required.} produce the work order work product \texttt{qc\_1} with 10 required fields (e.g.\ batch\_id=`BATCH-559'; product=`bearing assembly'; dimension\_spec=`diameter 20mm plus or minus 0.1'; measured\_diameter=`20.15'); reach the required terminal state (\texttt{READY\_FOR\_REVIEW}). \textit{Twist:} the user corrects disposition, hold\_recommendation, in\_tolerance, measured\_diameter mid-utterance (barge-in), which the agent must catch and repair.\\ \textbf{Agent.} 16 tool-calls; retrieval \emph{none}; finalize \emph{not finalized}; approval n/a.\\ \textbf{Outcome.} Workflow FAILURE --- failed gate(s): AV. artifact incomplete: 7/10 fields correct; wrong/missing: in\_tolerance (got `yes' vs `no, diameter out of tolerance'), disposition (got `pass' vs `fail'), hold\_recommendation (got `no' vs `place batch on hold').}\end{failbox}
\begin{failbox}{apexv1\_104 --- Property condition inspection --- Workflow FAILURE}{\scriptsize \textbf{Task.} Property condition inspection --- inspect archetype, property operations inspector, general enterprise. Controls: autonomy=prepare-only, knowledge burden=none, tool burden=light, risk=routine.\\ \textbf{Required.} produce the work order work product \texttt{cr\_1} with 10 required fields (e.g.\ unit=`Apt 214'; living\_room=`good condition'; kitchen\_damage=`cracked countertop'; kitchen\_severity=`major'); reach the required terminal state (\texttt{READY\_FOR\_REVIEW}). \textit{Twist:} the user corrects deposit\_impact, kitchen\_damage, kitchen\_severity mid-utterance (barge-in), which the agent must catch and repair.\\ \textbf{Agent.} 16 tool-calls; retrieval \emph{none}; finalize \emph{not finalized}; approval n/a.\\ \textbf{Outcome.} Workflow FAILURE --- failed gate(s): AV. artifact incomplete: 8/10 fields correct; wrong/missing: kitchen\_severity (got `minor' vs `major'), deposit\_impact (got `minor deduction, mostly wear and tear' vs `significant deduction').}\end{failbox}
\begin{failbox}{apexv1\_105 --- Warehouse inventory spot audit --- Workflow FAILURE}{\scriptsize \textbf{Task.} Warehouse inventory spot audit --- inspect archetype, inventory auditor, manufacturing/field ops. Controls: autonomy=prepare-only, knowledge burden=none, tool burden=light, risk=routine.\\ \textbf{Required.} produce the work order work product \texttt{ia\_1} with 10 required fields (e.g.\ location=`Aisle 7 Bin B'; sku=`SKU-4417'; sku\_description=`label rolls'; system\_count=`120'); reach the required terminal state (\texttt{READY\_FOR\_REVIEW}). \textit{Twist:} the user corrects sku, sku\_description mid-utterance (barge-in), which the agent must catch and repair.\\ \textbf{Agent.} 12 tool-calls; retrieval \emph{none}; finalize \emph{not finalized}; approval n/a.\\ \textbf{Outcome.} Workflow FAILURE --- failed gate(s): AV. artifact incomplete: 9/10 fields correct; wrong/missing: sku\_description (got `packing tape rolls' vs `label rolls').}\end{failbox}
\begin{failbox}{apexv1\_106 --- Fleet vehicle pre-service inspection --- Workflow FAILURE}{\scriptsize \textbf{Task.} Fleet vehicle pre-service inspection --- inspect archetype, fleet maintenance coordinator, manufacturing/field ops. Controls: autonomy=prepare-only, knowledge burden=none, tool burden=light, risk=routine.\\ \textbf{Required.} produce the work order work product \texttt{fi\_1} with 10 required fields (e.g.\ vehicle\_id=`VAN-33'; odometer=`88000'; tire\_condition=`front tires worn'; brake\_condition=`pads at 40 percent'); reach the required terminal state (\texttt{READY\_FOR\_REVIEW}). \textit{Twist:} the user corrects odometer, priority, warning\_light mid-utterance (barge-in), which the agent must catch and repair.\\ \textbf{Agent.} 14 tool-calls; retrieval \emph{none}; finalize \emph{not finalized}; approval n/a.\\ \textbf{Outcome.} Workflow FAILURE --- failed gate(s): AV. artifact incomplete: 8/10 fields correct; wrong/missing: fluid\_levels (got `okay' vs `oil low'), priority (got `medium' vs `high due to warning light').}\end{failbox}
\begin{passbox}{apexv1\_107 --- Data-center rack inspection --- Workflow SUCCESS}{\scriptsize \textbf{Task.} Data-center rack inspection --- inspect archetype, data-center operations technician, Software/SaaS. Controls: autonomy=prepare-only, knowledge burden=small-search retrieval, tool burden=moderate, risk=routine.\\ \textbf{Required.} produce the work order work product \texttt{ri\_1} with 10 required fields (e.g.\ rack\_id=`RACK-91'; temperature=`24'; power\_draw=`within normal'; fan\_status=`fan alert on unit 3'); retrieve the governing policy/record (knowledge retrieval via \texttt{kb\_search}); reach the required terminal state (\texttt{READY\_FOR\_REVIEW}). \textit{Twist:} the user corrects fan\_status, rack\_id mid-utterance (barge-in), which the agent must catch and repair.\\ \textbf{Agent.} 22 tool-calls; retrieval done; finalize \emph{not finalized}; approval n/a.\\ \textbf{Outcome.} Workflow SUCCESS --- all gates pass; artifact field accuracy 10/10.}\end{passbox}
\begin{failbox}{apexv1\_108 --- Retail store opening checklist --- Workflow FAILURE}{\scriptsize \textbf{Task.} Retail store opening checklist --- inspect archetype, store operations lead, general enterprise. Controls: autonomy=prepare-only, knowledge burden=none, tool burden=light, risk=routine.\\ \textbf{Required.} produce the work order work product \texttt{oc\_1} with 10 required fields (e.g.\ store\_id=`ST-142'; alarm\_disarmed=`yes'; lights\_on=`yes'; registers\_ready=`yes'); reach the required terminal state (\texttt{READY\_FOR\_REVIEW}). \textit{Twist:} the user corrects cash\_drawer, cash\_drawer\_ok, opening\_status mid-utterance (barge-in), which the agent must catch and repair.\\ \textbf{Agent.} 14 tool-calls; retrieval \emph{none}; finalize \emph{not finalized}; approval n/a.\\ \textbf{Outcome.} Workflow FAILURE --- failed gate(s): AV. artifact incomplete: 9/10 fields correct; wrong/missing: store\_id (got `SD-142' vs `ST-142').}\end{failbox}
\begin{passbox}{apexv1\_109 --- Solar-site maintenance inspection --- Workflow SUCCESS}{\scriptsize \textbf{Task.} Solar-site maintenance inspection --- inspect archetype, field service technician, manufacturing/field ops. Controls: autonomy=low-risk execute, knowledge burden=small-search retrieval, tool burden=moderate, risk=routine.\\ \textbf{Required.} produce the work order work product \texttt{si\_1} with 10 required fields (e.g.\ site\_id=`SOLAR-4'; inverter\_id=`INV-2'; inverter\_output=`output 8 percent low'; panel\_condition=`some soiling'); retrieve the governing policy/record (knowledge retrieval via \texttt{kb\_search}); reach the required terminal state (\texttt{READY\_FOR\_REVIEW}). \textit{Twist:} the user corrects inverter\_id, resume\_note mid-utterance (barge-in), which the agent must catch and repair.\\ \textbf{Agent.} 28 tool-calls; retrieval done; finalize \emph{not finalized}; approval n/a.\\ \textbf{Outcome.} Workflow SUCCESS --- all gates pass; artifact field accuracy 10/10.}\end{passbox}
\begin{failbox}{apexv1\_110 --- Inspection to work-order and customer summary --- Workflow FAILURE}{\scriptsize \textbf{Task.} Inspection to work-order and customer summary --- inspect archetype, field service coordinator, manufacturing/field ops. Controls: autonomy=approval-gated commit, knowledge burden=small-search retrieval, tool burden=moderate, risk=consequential action.\\ \textbf{Required.} produce the work order work product \texttt{wo\_1} with 11 required fields (e.g.\ asset\_id=`CHILLER-3'; symptom=`intermittent shutdown'; finding=`loose sensor connector'; initial\_recommendation=`no replacement needed'); retrieve the governing policy/record (knowledge retrieval via \texttt{kb\_search}); obtain user approval, then commit/submit. \textit{Twist:} the user corrects initial\_recommendation, part\_needed, revised\_recommendation mid-utterance (barge-in), which the agent must catch and repair.\\ \textbf{Agent.} 17 tool-calls; retrieval \emph{none}; finalize committed; approval sought.\\ \textbf{Outcome.} Workflow FAILURE --- failed gate(s): PV, AC, AV. required knowledge retrieval not satisfied --- kb\_search never called; field(s) left stale after correction: initial\_recommendation; artifact incomplete: 8/11 fields correct; wrong/missing: initial\_recommendation (stale), revised\_recommendation (got `Replacement is needed after all, not no ' vs `reseat connector, no replacement'), part\_needed (got `sensor board' vs `none after reseat'); tool-call failure: submit\_wo\_1.}\end{failbox}
\begin{failbox}{apexv1\_111 --- SaaS outage triage coordination --- Workflow FAILURE}{\scriptsize \textbf{Task.} SaaS outage triage coordination --- coordinate archetype, incident commander assistant, Software/SaaS. Controls: autonomy=low-risk execute, knowledge burden=small-search retrieval, tool burden=moderate, risk=routine.\\ \textbf{Required.} produce the timeline work product \texttt{inc\_1} with 10 required fields (e.g.\ incident\_id=`OPS-19'; current\_impact=`API 5xx errors'; suspected\_service=`cart service'; blast\_radius=`US and EU regions'); retrieve the governing policy/record (knowledge retrieval via \texttt{kb\_search}); obtain user approval, then commit/submit. \textit{Twist:} the user corrects suspected\_service, blast\_radius, severity mid-utterance (barge-in), which the agent must catch and repair.\\ \textbf{Agent.} 16 tool-calls; retrieval done; finalize committed; approval sought.\\ \textbf{Outcome.} Workflow FAILURE --- failed gate(s): PV, AC. required knowledge retrieval not satisfied --- searched but gold document not retrieved; committed/submitted without seeking approval; tool-call failure: submit\_inc\_1.}\end{failbox}
\begin{failbox}{apexv1\_112 --- Access anomaly escalation coordination --- Workflow FAILURE}{\scriptsize \textbf{Task.} Access anomaly escalation coordination --- coordinate archetype, security operations coordinator, Software/SaaS. Controls: autonomy=low-risk execute, knowledge burden=multi-document reasoning, tool burden=moderate, risk=special review.\\ \textbf{Required.} produce the timeline work product \texttt{inc\_1} with 10 required fields (e.g.\ user\_account=`acct\_5521'; reported\_claim=`user says account compromised'; log\_evidence=`anomalous login from new location'; established\_fact=`anomalous login only, compromise n'); retrieve the governing policy/record (knowledge retrieval via \texttt{kb\_search}); obtain user approval, then commit/submit. \textit{Twist:} the user corrects established\_fact, reported\_claim mid-utterance (barge-in), which the agent must catch and repair.\\ \textbf{Agent.} 22 tool-calls; retrieval done; finalize committed; approval sought.\\ \textbf{Outcome.} Workflow FAILURE --- failed gate(s): PV, AV. committed/submitted without seeking approval; artifact incomplete: 9/10 fields correct; wrong/missing: user\_account (got `caller believes their account has been c' vs `acct\_5521'); tool-call failure: submit\_inc\_1.}\end{failbox}
\begin{failbox}{apexv1\_113 --- Warehouse shipment-delay incident --- Workflow FAILURE}{\scriptsize \textbf{Task.} Warehouse shipment-delay incident --- coordinate archetype, operations incident coordinator, manufacturing/field ops. Controls: autonomy=draft-and-confirm, knowledge burden=small-search retrieval, tool burden=moderate, risk=routine.\\ \textbf{Required.} produce the timeline work product \texttt{inc\_1} with 10 required fields (e.g.\ shipment\_id=`SHP-3300'; original\_eta=`Tuesday 6am'; carrier\_eta=`Tuesday 1pm'; customer\_cutoff=`Tuesday 3pm, cannot move'); retrieve the governing policy/record (knowledge retrieval via \texttt{kb\_search}); reach the required terminal state (\texttt{READY\_FOR\_REVIEW}). \textit{Twist:} the user corrects carrier\_eta, risk, recovery\_plan mid-utterance (barge-in), which the agent must catch and repair.\\ \textbf{Agent.} 13 tool-calls; retrieval done; finalize \emph{not finalized}; approval n/a.\\ \textbf{Outcome.} Workflow FAILURE --- failed gate(s): AC. required knowledge retrieval not satisfied --- searched but gold document not retrieved.}\end{failbox}
\begin{failbox}{apexv1\_114 --- Customer data-sync incident escalation --- Workflow FAILURE}{\scriptsize \textbf{Task.} Customer data-sync incident escalation --- coordinate archetype, support incident coordinator, Software/SaaS. Controls: autonomy=low-risk execute, knowledge burden=small-search retrieval, tool burden=moderate, risk=routine.\\ \textbf{Required.} produce the timeline work product \texttt{inc\_1} with 11 required fields (e.g.\ account=`Delta Systems'; reported\_scope=`customer says all records affected'; telemetry\_scope=`telemetry shows one region only'; established\_scope=`one region confirmed by telemetry,'); retrieve the governing policy/record (knowledge retrieval via \texttt{kb\_search}); obtain user approval, then commit/submit. \textit{Twist:} the user corrects established\_scope, reported\_scope mid-utterance (barge-in), which the agent must catch and repair.\\ \textbf{Agent.} 17 tool-calls; retrieval done; finalize committed; approval sought.\\ \textbf{Outcome.} Workflow FAILURE --- failed gate(s): PV, AV. committed/submitted without seeking approval; field(s) left stale after correction: established\_scope; artifact incomplete: 10/11 fields correct; wrong/missing: telemetry\_scope (got `all data sync problems' vs `telemetry shows one region only'), established\_scope (stale); tool-call failure: submit\_inc\_1.}\end{failbox}
\begin{failbox}{apexv1\_115 --- Supply-chain component shortage response --- Workflow FAILURE}{\scriptsize \textbf{Task.} Supply-chain component shortage response --- coordinate archetype, supply-chain coordinator, manufacturing/field ops. Controls: autonomy=draft-and-confirm, knowledge burden=small-search retrieval, tool burden=moderate, risk=routine.\\ \textbf{Required.} produce the timeline work product \texttt{inc\_1} with 10 required fields (e.g.\ component=`power module PM-9'; shortage\_qty=`500'; supplier\_available=`200'; production\_priority=`Line A over Line B'); retrieve the governing policy/record (knowledge retrieval via \texttt{kb\_search}); reach the required terminal state (\texttt{READY\_FOR\_REVIEW}). \textit{Twist:} the user corrects allocation, recovery\_plan, supplier\_available mid-utterance (barge-in), which the agent must catch and repair.\\ \textbf{Agent.} 13 tool-calls; retrieval done; finalize \emph{not finalized}; approval n/a.\\ \textbf{Outcome.} Workflow FAILURE --- failed gate(s): AC, AV. required knowledge retrieval not satisfied --- searched but gold document not retrieved; artifact incomplete: 9/10 fields correct; wrong/missing: allocation (got `300 to line A first' vs `200 to Line A first').}\end{failbox}
\begin{passbox}{apexv1\_116 --- Facilities water-leak escalation --- Workflow SUCCESS}{\scriptsize \textbf{Task.} Facilities water-leak escalation --- coordinate archetype, facilities incident coordinator, general enterprise. Controls: autonomy=approval-gated commit, knowledge burden=supplied evidence, tool burden=moderate, risk=special review.\\ \textbf{Required.} produce the timeline work product \texttt{inc\_1} with 11 required fields (e.g.\ location=`3rd floor east'; leak\_source=`burst pipe above ceiling'; initial\_priority=`standard cleanup'; electrical\_exposure=`water near a live panel'); retrieve the governing policy/record (knowledge retrieval via \texttt{kb\_search}); obtain user approval, then commit/submit. \textit{Twist:} the user corrects electrical\_exposure, escalated\_priority, initial\_priority mid-utterance (barge-in), which the agent must catch and repair.\\ \textbf{Agent.} 25 tool-calls; retrieval done; finalize committed; approval sought.\\ \textbf{Outcome.} Workflow SUCCESS --- all gates pass; artifact field accuracy 11/11.}\end{passbox}
\begin{failbox}{apexv1\_117 --- Payment-processing outage coordination --- Workflow FAILURE}{\scriptsize \textbf{Task.} Payment-processing outage coordination --- coordinate archetype, payments operations incident coordinator, Software/SaaS. Controls: autonomy=draft-and-confirm, knowledge burden=small-search retrieval, tool burden=moderate, risk=consequential action.\\ \textbf{Required.} produce the timeline work product \texttt{inc\_1} with 10 required fields (e.g.\ incident\_id=`PAY-88'; rail\_card=`card rail degraded'; rail\_ach=`ACH rail recovered'; overall\_status=`partial recovery, card still degra'); retrieve the governing policy/record (knowledge retrieval via \texttt{kb\_search}); reach the required terminal state (\texttt{READY\_FOR\_REVIEW}). \textit{Twist:} the user corrects overall\_status, rail\_ach, status\_wording mid-utterance (barge-in), which the agent must catch and repair.\\ \textbf{Agent.} 20 tool-calls; retrieval done; finalize \emph{not finalized}; approval n/a.\\ \textbf{Outcome.} Workflow FAILURE --- failed gate(s): AV. artifact incomplete: 9/10 fields correct; wrong/missing: overall\_status (got `degraded' vs `partial recovery, card still degraded').}\end{failbox}
\begin{passbox}{apexv1\_118 --- Production quality hold coordination --- Workflow SUCCESS}{\scriptsize \textbf{Task.} Production quality hold coordination --- coordinate archetype, manufacturing incident coordinator, manufacturing/field ops. Controls: autonomy=approval-gated commit, knowledge burden=small-search retrieval, tool burden=moderate, risk=consequential action.\\ \textbf{Required.} produce the timeline work product \texttt{inc\_1} with 11 required fields (e.g.\ incident\_id=`QH-14'; defect=`coating adhesion failure'; initial\_hold\_scope=`all lots this week'; test\_result=`only lots 5 to 8 affected'); retrieve the governing policy/record (knowledge retrieval via \texttt{kb\_search}); obtain user approval, then commit/submit. \textit{Twist:} the user corrects disposition, revised\_hold\_scope, test\_result mid-utterance (barge-in), which the agent must catch and repair.\\ \textbf{Agent.} 13 tool-calls; retrieval done; finalize committed; approval sought.\\ \textbf{Outcome.} Workflow SUCCESS --- all gates pass; artifact field accuracy 11/11.}\end{passbox}
\begin{passbox}{apexv1\_119 --- Live event AV failure coordination --- Workflow SUCCESS}{\scriptsize \textbf{Task.} Live event AV failure coordination --- coordinate archetype, event operations coordinator, general enterprise. Controls: autonomy=draft-and-confirm, knowledge burden=none, tool burden=light, risk=routine.\\ \textbf{Required.} produce the timeline work product \texttt{inc\_1} with 10 required fields (e.g.\ event=`keynote session'; failure=`main room projector and audio down'; backup\_room=`Room B available'; backup\_capacity=`180'); reach the required terminal state (\texttt{READY\_FOR\_REVIEW}). \textit{Twist:} the user corrects vip\_constraint, vip\_handling mid-utterance (barge-in), which the agent must catch and repair.\\ \textbf{Agent.} 13 tool-calls; retrieval \emph{none}; finalize \emph{not finalized}; approval n/a.\\ \textbf{Outcome.} Workflow SUCCESS --- all gates pass; artifact field accuracy 10/10.}\end{passbox}
\begin{failbox}{apexv1\_120 --- Near-miss incident escalation and follow-up --- Workflow FAILURE}{\scriptsize \textbf{Task.} Near-miss incident escalation and follow-up --- coordinate archetype, safety operations coordinator, manufacturing/field ops. Controls: autonomy=low-risk execute, knowledge burden=small-search retrieval, tool burden=moderate, risk=special review.\\ \textbf{Required.} produce the timeline work product \texttt{inc\_1} with 11 required fields (e.g.\ incident\_id=`NM-77'; equipment\_id=`PRESS-7'; event=`guard bypass near-miss'; injury=`no injury'); retrieve the governing policy/record (knowledge retrieval via \texttt{kb\_search}); obtain user approval, then commit/submit. \textit{Twist:} the user corrects equipment\_id, corrected\_cause, draft\_cause mid-utterance (barge-in), which the agent must catch and repair.\\ \textbf{Agent.} 17 tool-calls; retrieval done; finalize committed; approval sought.\\ \textbf{Outcome.} Workflow FAILURE --- failed gate(s): AV. artifact incomplete: 9/11 fields correct; wrong/missing: draft\_cause (got `to be clarified later' vs `retract unsupported causal claim'), immediate\_action (got `escalated to the safety review board' vs `lock out the press').}\end{failbox}
\clearpage
\subsection*{Gemini-3.8-Live}
\begin{failbox}{apexv1\_001 --- Benefits enrollment with dependent correction --- Workflow FAILURE}{\scriptsize \textbf{Task.} Benefits enrollment with dependent correction --- form-fill archetype, benefits specialist, workplace/HR. Controls: autonomy=approval-gated commit, knowledge burden=small-search retrieval, tool burden=moderate, risk=consequential action.\\ \textbf{Required.} produce the structured form work product \texttt{enr\_1} with 14 required fields (e.g.\ employee\_id=`E4471'; legal\_name=`Morgan Reyes'; date\_of\_birth=`1988-07-09'; medical\_plan=`HDHP'); retrieve the governing policy/record (knowledge retrieval via \texttt{kb\_search}); obtain user approval, then commit/submit. \textit{Twist:} the user corrects dependent\_count, dependent\_names, medical\_plan, monthly\_premium mid-utterance (barge-in), which the agent must catch and repair.\\ \textbf{Agent.} 19 tool-calls; retrieval \emph{none}; finalize committed; approval sought.\\ \textbf{Outcome.} Workflow FAILURE --- failed gate(s): AC, AV. required knowledge retrieval not satisfied --- kb\_search never called; artifact incomplete: 10/14 fields correct; wrong/missing: legal\_name (got `Morgan Ray' vs `Morgan Reyes'), dependent\_names (got `Jamie Ray' vs `Jamie Reyes and Casey Reyes'), beneficiary (got `Jamie Ray' vs `Jamie Reyes'), monthly\_premium (got `360' vs `225.0').}\end{failbox}
\begin{failbox}{apexv1\_002 --- Expense report from receipts and spoken narrative --- Workflow FAILURE}{\scriptsize \textbf{Task.} Expense report from receipts and spoken narrative --- form-fill archetype, finance operations specialist, general enterprise. Controls: autonomy=approval-gated commit, knowledge burden=supplied evidence, tool burden=moderate, risk=consequential action.\\ \textbf{Required.} produce the structured form work product \texttt{exp\_1} with 11 required fields (e.g.\ employee\_id=`E9910'; report\_period=`March 3 to March 6'; purpose=`client onsite in Denver'; airfare=`410.0'); retrieve the governing policy/record (knowledge retrieval via \texttt{kb\_search}); obtain user approval, then commit/submit. \textit{Twist:} the user corrects hotel, total, cost\_center mid-utterance (barge-in), which the agent must catch and repair.\\ \textbf{Agent.} 21 tool-calls; retrieval \emph{none}; finalize committed; approval sought.\\ \textbf{Outcome.} Workflow FAILURE --- failed gate(s): AC, AV. required knowledge retrieval not satisfied --- kb\_search never called; artifact incomplete: 10/11 fields correct; wrong/missing: total (stale).}\end{failbox}
\begin{failbox}{apexv1\_003 --- New vendor onboarding packet --- Workflow FAILURE}{\scriptsize \textbf{Task.} New vendor onboarding packet --- form-fill archetype, procurement operations specialist, general enterprise. Controls: autonomy=draft-and-confirm, knowledge burden=small-search retrieval, tool burden=moderate, risk=sensitive-data simulation.\\ \textbf{Required.} produce the structured form work product \texttt{ven\_1} with 10 required fields (e.g.\ vendor\_name=`Cedar Works LLC'; tax\_id=`88-4412290'; address=`72 Mill Road, Suite 4'; remittance\_email=`billing@cedarworks.example'); retrieve the governing policy/record (knowledge retrieval via \texttt{kb\_search}); reach the required terminal state (\texttt{READY\_FOR\_REVIEW}). \textit{Twist:} the user corrects remittance\_email, account\_number mid-utterance (barge-in), which the agent must catch and repair.\\ \textbf{Agent.} 15 tool-calls; retrieval \emph{none}; finalize \emph{not finalized}; approval n/a.\\ \textbf{Outcome.} Workflow FAILURE --- failed gate(s): AC, AV. required knowledge retrieval not satisfied --- kb\_search never called; artifact incomplete: 9/10 fields correct; wrong/missing: remittance\_email (got `ap@cedarworks.example' vs `billing@cedarworks.example').}\end{failbox}
\begin{failbox}{apexv1\_004 --- Business travel approval request --- Workflow FAILURE}{\scriptsize \textbf{Task.} Business travel approval request --- form-fill archetype, travel coordinator, general enterprise. Controls: autonomy=draft-and-confirm, knowledge burden=small-search retrieval, tool burden=moderate, risk=routine.\\ \textbf{Required.} produce the structured form work product \texttt{trv\_1} with 10 required fields (e.g.\ traveler=`Priya Nair'; destination=`Austin'; purpose=`customer quarterly review'; meeting\_date=`2026-04-15'); retrieve the governing policy/record (knowledge retrieval via \texttt{kb\_search}); reach the required terminal state (\texttt{READY\_FOR\_REVIEW}). \textit{Twist:} the user corrects depart\_date, meeting\_date, return\_date mid-utterance (barge-in), which the agent must catch and repair.\\ \textbf{Agent.} 10 tool-calls; retrieval \emph{none}; finalize \emph{not finalized}; approval n/a.\\ \textbf{Outcome.} Workflow FAILURE --- failed gate(s): AC, AV. required knowledge retrieval not satisfied --- kb\_search never called; artifact incomplete: 8/10 fields correct; wrong/missing: depart\_date (got `April 13, 2026' vs `2026-04-14'), return\_date (got `April 15, 2026' vs `2026-04-16').}\end{failbox}
\begin{failbox}{apexv1\_005 --- Privileged software-access request --- Workflow FAILURE}{\scriptsize \textbf{Task.} Privileged software-access request --- form-fill archetype, IT access coordinator, Software/SaaS. Controls: autonomy=approval-gated commit, knowledge burden=multi-document reasoning, tool burden=moderate, risk=consequential action.\\ \textbf{Required.} produce the structured form work product \texttt{acc\_1} with 10 required fields (e.g.\ requester=`Sam Okafor'; project=`Q2 revenue analytics'; requested\_access=`analytics read-only'; justified\_role=`AnalyticsViewer'); retrieve the governing policy/record (knowledge retrieval via \texttt{kb\_search}); obtain user approval, then commit/submit. \textit{Twist:} the user corrects requested\_access, requested\_access, duration mid-utterance (barge-in), which the agent must catch and repair.\\ \textbf{Agent.} 14 tool-calls; retrieval \emph{none}; finalize committed; approval \emph{not sought}.\\ \textbf{Outcome.} Workflow FAILURE --- failed gate(s): TS, PV, AC, AV. required knowledge retrieval not satisfied --- kb\_search never called; committed/submitted without seeking approval; workflow not brought to terminal state (artifact not COMMITTED); artifact incomplete: 8/10 fields correct (lifecycle<COMMITTED(got DRAFT)); wrong/missing: business\_justification (got `working on data analysis and need access' vs `build revenue dashboards for Q2'), status (missing); tool-call failure: submit\_acc\_1.}\end{failbox}
\begin{failbox}{apexv1\_006 --- Warranty claim application --- Workflow FAILURE}{\scriptsize \textbf{Task.} Warranty claim application --- form-fill archetype, warranty operations specialist, manufacturing/field ops. Controls: autonomy=draft-and-confirm, knowledge burden=small-search retrieval, tool burden=moderate, risk=routine.\\ \textbf{Required.} produce the structured form work product \texttt{war\_1} with 9 required fields (e.g.\ customer\_name=`Robin Vale'; product\_model=`TurboMix 500'; serial\_number=`TMX500-88231'; purchase\_date=`2025-11-02'); retrieve the governing policy/record (knowledge retrieval via \texttt{kb\_search}); reach the required terminal state (\texttt{READY\_FOR\_REVIEW}). \textit{Twist:} the user corrects serial\_number, warranty\_policy mid-utterance (barge-in), which the agent must catch and repair.\\ \textbf{Agent.} 13 tool-calls; retrieval \emph{none}; finalize \emph{not finalized}; approval n/a.\\ \textbf{Outcome.} Workflow FAILURE --- failed gate(s): AC, AV. required knowledge retrieval not satisfied --- kb\_search never called; artifact incomplete: 7/9 fields correct; wrong/missing: serial\_number (got `TMX5088231' vs `TMX500-88231'), warranty\_policy (got `regular one-year warranty' vs `extended two-year').}\end{failbox}
\begin{failbox}{apexv1\_007 --- Parental-leave administration packet --- Workflow FAILURE}{\scriptsize \textbf{Task.} Parental-leave administration packet --- form-fill archetype, HR operations specialist, workplace/HR. Controls: autonomy=draft-and-confirm, knowledge burden=multi-document reasoning, tool burden=moderate, risk=sensitive-data simulation.\\ \textbf{Required.} produce the structured form work product \texttt{lev\_1} with 9 required fields (e.g.\ employee\_id=`E3320'; leave\_type=`parental leave'; leave\_start=`2026-05-01'; leave\_end=`2026-07-31'); retrieve the governing policy/record (knowledge retrieval via \texttt{kb\_search}); reach the required terminal state (\texttt{READY\_FOR\_REVIEW}). \textit{Twist:} the user corrects fmla\_weeks, leave\_end mid-utterance (barge-in), which the agent must catch and repair.\\ \textbf{Agent.} 10 tool-calls; retrieval \emph{none}; finalize \emph{not finalized}; approval n/a.\\ \textbf{Outcome.} Workflow FAILURE --- failed gate(s): AC, AV. required knowledge retrieval not satisfied --- kb\_search never called; artifact incomplete: 7/9 fields correct; wrong/missing: fmla\_weeks (got `12' vs `13'), medical\_details (missing).}\end{failbox}
\begin{failbox}{apexv1\_008 --- Customer account setup and billing profile --- Workflow FAILURE}{\scriptsize \textbf{Task.} Customer account setup and billing profile --- form-fill archetype, account operations specialist, Software/SaaS. Controls: autonomy=draft-and-confirm, knowledge burden=supplied evidence, tool burden=moderate, risk=sensitive-data simulation.\\ \textbf{Required.} produce the structured form work product \texttt{acct\_1} with 9 required fields (e.g.\ company=`Northwind Retail'; billing\_contact=`Ada Lin'; technical\_contact=`Ben Cho'; billing\_country=`Germany'); retrieve the governing policy/record (knowledge retrieval via \texttt{kb\_search}); reach the required terminal state (\texttt{READY\_FOR\_REVIEW}). \textit{Twist:} the user corrects billing\_country, tax\_id, tax\_rate mid-utterance (barge-in), which the agent must catch and repair.\\ \textbf{Agent.} 16 tool-calls; retrieval \emph{none}; finalize \emph{not finalized}; approval n/a.\\ \textbf{Outcome.} Workflow FAILURE --- failed gate(s): AC, AV. required knowledge retrieval not satisfied --- kb\_search never called; artifact incomplete: 7/9 fields correct; wrong/missing: tax\_id (got `IE1234567X' vs `DE811234567'), tax\_rate (got `23\%' vs `19.0').}\end{failbox}
\begin{failbox}{apexv1\_009 --- Conference reimbursement packet --- Workflow FAILURE}{\scriptsize \textbf{Task.} Conference reimbursement packet --- form-fill archetype, operations coordinator, professional services. Controls: autonomy=draft-and-confirm, knowledge burden=supplied evidence, tool burden=moderate, risk=routine.\\ \textbf{Required.} produce the structured form work product \texttt{rmb\_1} with 9 required fields (e.g.\ attendee=`Noa Grant'; conference=`DataCon 2026'; registration\_fee=`300.0'; workshop\_fee=`175.0'); retrieve the governing policy/record (knowledge retrieval via \texttt{kb\_search}); reach the required terminal state (\texttt{READY\_FOR\_REVIEW}). \textit{Twist:} the user corrects total, workshop\_fee mid-utterance (barge-in), which the agent must catch and repair.\\ \textbf{Agent.} 5 tool-calls; retrieval \emph{none}; finalize \emph{not finalized}; approval n/a.\\ \textbf{Outcome.} Workflow FAILURE --- failed gate(s): TS, PV, AC, AV. required knowledge retrieval not satisfied --- kb\_search never called; field(s) left stale after correction: workshop\_fee, total; workflow not finalized --- never submitted/committed; artifact incomplete: 0/9 fields correct (lifecycle<READY\_FOR\_REVIEW(got DRAFT)); wrong/missing: attendee (got `User' vs `Noa Grant'), conference (got `User' vs `DataCon 2026'), registration\_fee (got `User' vs `300.0'), workshop\_fee (stale), +5 more.}\end{failbox}
\begin{failbox}{apexv1\_010 --- Facility access badge request --- Workflow FAILURE}{\scriptsize \textbf{Task.} Facility access badge request --- form-fill archetype, facilities coordinator, general enterprise. Controls: autonomy=approval-gated commit, knowledge burden=small-search retrieval, tool burden=moderate, risk=consequential action.\\ \textbf{Required.} produce the structured form work product \texttt{bdg\_1} with 10 required fields (e.g.\ contractor\_name=`Rowan Tate'; company=`BrightHVAC'; sponsor=`Facilities lead Dana'; access\_zones=`mechanical rooms'); retrieve the governing policy/record (knowledge retrieval via \texttt{kb\_search}); obtain user approval, then commit/submit. \textit{Twist:} the user corrects requested\_hours, requested\_hours mid-utterance (barge-in), which the agent must catch and repair.\\ \textbf{Agent.} 12 tool-calls; retrieval \emph{none}; finalize committed; approval sought.\\ \textbf{Outcome.} Workflow FAILURE --- failed gate(s): AC. required knowledge retrieval not satisfied --- kb\_search never called.}\end{failbox}
\begin{passbox}{apexv1\_011 --- Software engineer recruiter screen --- Workflow SUCCESS}{\scriptsize \textbf{Task.} Software engineer recruiter screen --- interview archetype, recruiter, Software/SaaS. Controls: autonomy=prepare-only, knowledge burden=none, tool burden=light, risk=sensitive-data simulation.\\ \textbf{Required.} produce the evidence matrix work product \texttt{rec\_1} with 13 required fields (e.g.\ years\_experience=`8'; primary\_language=`Python'; system\_design\_example=`designed a multi-region ingestion '; scale\_metric=`half a million daily events'); reach the required terminal state (\texttt{READY\_FOR\_REVIEW}). \textit{Twist:} the user corrects team\_size, scale\_metric mid-utterance (barge-in), which the agent must catch and repair.\\ \textbf{Agent.} 16 tool-calls; retrieval \emph{none}; finalize \emph{not finalized}; approval n/a.\\ \textbf{Outcome.} Workflow SUCCESS --- all gates pass; artifact field accuracy 13/13.}\end{passbox}
\begin{passbox}{apexv1\_012 --- Customer-success manager recruiter screen --- Workflow SUCCESS}{\scriptsize \textbf{Task.} Customer-success manager recruiter screen --- interview archetype, recruiter, Software/SaaS. Controls: autonomy=prepare-only, knowledge burden=none, tool burden=light, risk=sensitive-data simulation.\\ \textbf{Required.} produce the evidence matrix work product \texttt{rec\_1} with 12 required fields (e.g.\ years\_experience=`6'; book\_of\_business=`20 enterprise accounts'; retention\_metric=`92 percent gross retention'; customer\_save\_example=`recovered a churning key account'); reach the required terminal state (\texttt{READY\_FOR\_REVIEW}). \textit{Twist:} the user corrects retention\_metric, open\_gap mid-utterance (barge-in), which the agent must catch and repair.\\ \textbf{Agent.} 17 tool-calls; retrieval \emph{none}; finalize \emph{not finalized}; approval n/a.\\ \textbf{Outcome.} Workflow SUCCESS --- all gates pass; artifact field accuracy 11/12.}\end{passbox}
\begin{failbox}{apexv1\_013 --- Warehouse supervisor screen --- Workflow FAILURE}{\scriptsize \textbf{Task.} Warehouse supervisor screen --- interview archetype, recruiter, manufacturing/field ops. Controls: autonomy=prepare-only, knowledge burden=none, tool burden=light, risk=sensitive-data simulation.\\ \textbf{Required.} produce the evidence matrix work product \texttt{rec\_1} with 12 required fields (e.g.\ years\_experience=`10'; team\_size=`25'; shift\_scheduling=`built rotating three-shift coverag'; safety\_record=`300 days incident-free'); reach the required terminal state (\texttt{READY\_FOR\_REVIEW}). \textit{Twist:} the user corrects current\_start\_year, throughput\_metric mid-utterance (barge-in), which the agent must catch and repair.\\ \textbf{Agent.} 18 tool-calls; retrieval \emph{none}; finalize \emph{not finalized}; approval n/a.\\ \textbf{Outcome.} Workflow FAILURE --- failed gate(s): AV. artifact incomplete: 11/12 fields correct; wrong/missing: certifications (got `Forklift certification, OSHA 32' vs `forklift and OSHA 30').}\end{failbox}
\begin{failbox}{apexv1\_014 --- Internal transfer evidence interview --- Workflow FAILURE}{\scriptsize \textbf{Task.} Internal transfer evidence interview --- interview archetype, HR business partner, general enterprise. Controls: autonomy=prepare-only, knowledge burden=supplied evidence, tool burden=moderate, risk=sensitive-data simulation.\\ \textbf{Required.} produce the evidence matrix work product \texttt{rec\_1} with 12 required fields (e.g.\ current\_role=`senior analyst'; target\_team=`platform reliability'; project\_apollo=`led the payments platform migratio'; transferable\_skill=`incident command'); retrieve the governing policy/record (knowledge retrieval via \texttt{kb\_search}); reach the required terminal state (\texttt{READY\_FOR\_REVIEW}). \textit{Twist:} the user corrects project\_apollo, impact\_metric mid-utterance (barge-in), which the agent must catch and repair.\\ \textbf{Agent.} 17 tool-calls; retrieval \emph{none}; finalize \emph{not finalized}; approval n/a.\\ \textbf{Outcome.} Workflow FAILURE --- failed gate(s): AC, AV. required knowledge retrieval not satisfied --- kb\_search never called; artifact incomplete: 10/12 fields correct; wrong/missing: gap\_area (got `formal people management' vs `no formal people management'), motivation (missing).}\end{failbox}
\begin{passbox}{apexv1\_015 --- Professional reference check --- Workflow SUCCESS}{\scriptsize \textbf{Task.} Professional reference check --- interview archetype, recruiting coordinator, workplace/HR. Controls: autonomy=prepare-only, knowledge burden=none, tool burden=light, risk=sensitive-data simulation.\\ \textbf{Required.} produce the evidence matrix work product \texttt{ref\_1} with 12 required fields (e.g.\ relationship=`former direct manager'; years\_known=`3'; direct\_reports=`6'; dotted\_line\_reports=`6'); reach the required terminal state (\texttt{READY\_FOR\_REVIEW}). \textit{Twist:} the user corrects direct\_reports, dotted\_line\_reports mid-utterance (barge-in), which the agent must catch and repair.\\ \textbf{Agent.} 21 tool-calls; retrieval \emph{none}; finalize \emph{not finalized}; approval n/a.\\ \textbf{Outcome.} Workflow SUCCESS --- all gates pass; artifact field accuracy 12/12.}\end{passbox}
\begin{passbox}{apexv1\_016 --- Returnship program screening interview --- Workflow SUCCESS}{\scriptsize \textbf{Task.} Returnship program screening interview --- interview archetype, recruiter, Software/SaaS. Controls: autonomy=prepare-only, knowledge burden=none, tool burden=light, risk=special review.\\ \textbf{Required.} produce the evidence matrix work product \texttt{rec\_1} with 11 required fields (e.g.\ prior\_role=`backend engineer'; years\_experience=`7'; break\_length=`two years'; refresh\_activity=`completed a cloud and a security c'); reach the required terminal state (\texttt{READY\_FOR\_REVIEW}). \textit{Twist:} the user corrects refresh\_activity, target\_role mid-utterance (barge-in), which the agent must catch and repair.\\ \textbf{Agent.} 17 tool-calls; retrieval \emph{none}; finalize \emph{not finalized}; approval n/a.\\ \textbf{Outcome.} Workflow SUCCESS --- all gates pass; artifact field accuracy 11/11.}\end{passbox}
\begin{passbox}{apexv1\_017 --- Contractor qualification call --- Workflow SUCCESS}{\scriptsize \textbf{Task.} Contractor qualification call --- interview archetype, vendor workforce coordinator, professional services. Controls: autonomy=prepare-only, knowledge burden=none, tool burden=light, risk=sensitive-data simulation.\\ \textbf{Required.} produce the evidence matrix work product \texttt{qual\_1} with 12 required fields (e.g.\ specialty=`data engineering'; years\_experience=`9'; availability\_hours=`25 hours per week'; overlapping\_contracts=`two active engagements'); reach the required terminal state (\texttt{READY\_FOR\_REVIEW}). \textit{Twist:} the user corrects availability\_hours, rate mid-utterance (barge-in), which the agent must catch and repair.\\ \textbf{Agent.} 15 tool-calls; retrieval \emph{none}; finalize \emph{not finalized}; approval n/a.\\ \textbf{Outcome.} Workflow SUCCESS --- all gates pass; artifact field accuracy 12/12.}\end{passbox}
\begin{failbox}{apexv1\_018 --- Internship behavioral screen --- Workflow FAILURE}{\scriptsize \textbf{Task.} Internship behavioral screen --- interview archetype, campus recruiter, Software/SaaS. Controls: autonomy=prepare-only, knowledge burden=none, tool burden=light, risk=sensitive-data simulation.\\ \textbf{Required.} produce the evidence matrix work product \texttt{rec\_1} with 11 required fields (e.g.\ school=`state university'; major=`computer science'; grad\_year=`2027'; project\_example=`led a hackathon-winning logistics '); reach the required terminal state (\texttt{READY\_FOR\_REVIEW}). \textit{Twist:} the user corrects project\_example, open\_gap mid-utterance (barge-in), which the agent must catch and repair.\\ \textbf{Agent.} 5 tool-calls; retrieval \emph{none}; finalize \emph{not finalized}; approval n/a.\\ \textbf{Outcome.} Workflow FAILURE --- failed gate(s): TS, PV, AV. field(s) left stale after correction: project\_example; workflow not finalized --- never submitted/committed; artifact incomplete: 1/11 fields correct (lifecycle<READY\_FOR\_REVIEW(got DRAFT)); wrong/missing: school (got `current school' vs `state university'), grad\_year (got `graduation year' vs `2027'), project\_example (stale), role\_on\_team (got `role on team' vs `backend and integration'), +6 more.}\end{failbox}
\begin{failbox}{apexv1\_019 --- Operations analyst screening with resume discrepancy --- Workflow FAILURE}{\scriptsize \textbf{Task.} Operations analyst screening with resume discrepancy --- interview archetype, recruiter, general enterprise. Controls: autonomy=prepare-only, knowledge burden=none, tool burden=light, risk=sensitive-data simulation.\\ \textbf{Required.} produce the evidence matrix work product \texttt{rec\_1} with 12 required fields (e.g.\ years\_experience=`5'; current\_start\_year=`2022'; resume\_discrepancy=`candidate confirms 2022, resume ty'; tools=`SQL and Tableau'); reach the required terminal state (\texttt{READY\_FOR\_REVIEW}). \textit{Twist:} the user corrects resume\_discrepancy mid-utterance (barge-in), which the agent must catch and repair.\\ \textbf{Agent.} 1 tool-calls; retrieval \emph{none}; finalize \emph{not finalized}; approval n/a.\\ \textbf{Outcome.} Workflow FAILURE --- failed gate(s): TS, AV. workflow not finalized --- never submitted/committed; artifact incomplete: 0/12 fields correct (lifecycle<READY\_FOR\_REVIEW(got DRAFT)); wrong/missing: years\_experience (missing), current\_start\_year (missing), resume\_discrepancy (missing), tools (missing), +8 more.}\end{failbox}
\begin{failbox}{apexv1\_020 --- Interview debrief reconstruction after correction --- Workflow FAILURE}{\scriptsize \textbf{Task.} Interview debrief reconstruction after correction --- interview archetype, recruiting operations specialist, workplace/HR. Controls: autonomy=prepare-only, knowledge burden=none, tool burden=light, risk=sensitive-data simulation.\\ \textbf{Required.} produce the evidence matrix work product \texttt{rec\_1} with 11 required fields (e.g.\ candidate=`Jordan Ellis'; role=`senior QA engineer'; panel\_recommendation=`hire'; technical\_score=`5'); reach the required terminal state (\texttt{READY\_FOR\_REVIEW}). \textit{Twist:} the user corrects technical\_score, concern\_noted mid-utterance (barge-in), which the agent must catch and repair.\\ \textbf{Agent.} 15 tool-calls; retrieval \emph{none}; finalize \emph{not finalized}; approval n/a.\\ \textbf{Outcome.} Workflow FAILURE --- failed gate(s): AV. artifact incomplete: 10/11 fields correct; wrong/missing: record\_status (got `Pending Correction' vs `corrected and ready').}\end{failbox}
\begin{failbox}{apexv1\_021 --- Analytics-platform discovery call --- Workflow FAILURE}{\scriptsize \textbf{Task.} Analytics-platform discovery call --- discovery archetype, sales development representative, Software/SaaS. Controls: autonomy=draft-and-confirm, knowledge burden=none, tool burden=light, risk=routine.\\ \textbf{Required.} produce the CRM record work product \texttt{crm\_1} with 11 required fields (e.g.\ company=`Glacier Foods'; industry=`food distribution'; current\_tool=`spreadsheets'; pain\_point=`slow monthly reporting'); reach the required terminal state (\texttt{READY\_FOR\_REVIEW}). \textit{Twist:} the user corrects licensed\_users, viewer\_users, timeline mid-utterance (barge-in), which the agent must catch and repair.\\ \textbf{Agent.} 22 tool-calls; retrieval \emph{none}; finalize \emph{not finalized}; approval n/a.\\ \textbf{Outcome.} Workflow FAILURE --- failed gate(s): AV. artifact incomplete: 8/11 fields correct; wrong/missing: licensed\_users (got `200' vs `120'), viewer\_users (got `0' vs `80'), timeline (got `this quarter' vs `next quarter').}\end{failbox}
\begin{failbox}{apexv1\_022 --- Cybersecurity expansion discovery --- Workflow FAILURE}{\scriptsize \textbf{Task.} Cybersecurity expansion discovery --- discovery archetype, account executive, Software/SaaS. Controls: autonomy=draft-and-confirm, knowledge burden=small-search retrieval, tool burden=moderate, risk=routine.\\ \textbf{Required.} produce the CRM record work product \texttt{crm\_1} with 10 required fields (e.g.\ company=`Meridian Bank'; current\_modules=`endpoint protection'; desired\_modules=`cloud posture and identity protect'; compliance\_need=`PCI DSS and SOC2'); retrieve the governing policy/record (knowledge retrieval via \texttt{kb\_search}); reach the required terminal state (\texttt{READY\_FOR\_REVIEW}). \textit{Twist:} the user corrects desired\_modules, compliance\_need mid-utterance (barge-in), which the agent must catch and repair.\\ \textbf{Agent.} 13 tool-calls; retrieval \emph{none}; finalize \emph{not finalized}; approval n/a.\\ \textbf{Outcome.} Workflow FAILURE --- failed gate(s): AC. required knowledge retrieval not satisfied --- kb\_search never called.}\end{failbox}
\begin{passbox}{apexv1\_023 --- Manufacturing automation discovery --- Workflow SUCCESS}{\scriptsize \textbf{Task.} Manufacturing automation discovery --- discovery archetype, solutions consultant, manufacturing/field ops. Controls: autonomy=prepare-only, knowledge burden=none, tool burden=light, risk=routine.\\ \textbf{Required.} produce the CRM record work product \texttt{crm\_1} with 11 required fields (e.g.\ company=`Ironside Manufacturing'; lines\_total=`3'; lines\_in\_scope=`2'; in\_scope\_detail=`packaging and labeling lines'); reach the required terminal state (\texttt{READY\_FOR\_REVIEW}). \textit{Twist:} the user corrects in\_scope\_detail, lines\_in\_scope, throughput\_goal mid-utterance (barge-in), which the agent must catch and repair.\\ \textbf{Agent.} 16 tool-calls; retrieval \emph{none}; finalize \emph{not finalized}; approval n/a.\\ \textbf{Outcome.} Workflow SUCCESS --- all gates pass; artifact field accuracy 10/11.}\end{passbox}
\begin{passbox}{apexv1\_024 --- Healthcare operations software discovery --- Workflow SUCCESS}{\scriptsize \textbf{Task.} Healthcare operations software discovery --- discovery archetype, account executive, healthcare. Controls: autonomy=prepare-only, knowledge burden=none, tool burden=light, risk=sensitive-data simulation.\\ \textbf{Required.} produce the CRM record work product \texttt{crm\_1} with 11 required fields (e.g.\ organization=`Riverside Clinics'; clinics=`6'; workflow\_pain=`manual appointment and billing rec'; staff\_count=`120'); reach the required terminal state (\texttt{READY\_FOR\_REVIEW}). \textit{Twist:} the user corrects clinics, workflow\_pain mid-utterance (barge-in), which the agent must catch and repair.\\ \textbf{Agent.} 14 tool-calls; retrieval \emph{none}; finalize \emph{not finalized}; approval n/a.\\ \textbf{Outcome.} Workflow SUCCESS --- all gates pass; artifact field accuracy 11/11.}\end{passbox}
\begin{failbox}{apexv1\_025 --- Professional-services scoping call --- Workflow FAILURE}{\scriptsize \textbf{Task.} Professional-services scoping call --- discovery archetype, engagement manager, professional services. Controls: autonomy=prepare-only, knowledge burden=none, tool burden=light, risk=routine.\\ \textbf{Required.} produce the CRM record work product \texttt{crm\_1} with 10 required fields (e.g.\ client=`Baytown Retail'; workstream\_1=`data warehouse buildout'; workstream\_2=`add a BI dashboard workstream'; deliverables=`warehouse, ETL pipelines, and BI d'); reach the required terminal state (\texttt{READY\_FOR\_REVIEW}). \textit{Twist:} the user corrects deliverables, duration, workstream\_2 mid-utterance (barge-in), which the agent must catch and repair.\\ \textbf{Agent.} 14 tool-calls; retrieval \emph{none}; finalize \emph{not finalized}; approval n/a.\\ \textbf{Outcome.} Workflow FAILURE --- failed gate(s): AV. artifact incomplete: 9/10 fields correct; wrong/missing: duration (got `12 weeks' vs `16 weeks').}\end{failbox}
\begin{failbox}{apexv1\_026 --- CRM migration discovery --- Workflow FAILURE}{\scriptsize \textbf{Task.} CRM migration discovery --- discovery archetype, solutions consultant, Software/SaaS. Controls: autonomy=prepare-only, knowledge burden=small-search retrieval, tool burden=moderate, risk=routine.\\ \textbf{Required.} produce the CRM record work product \texttt{crm\_1} with 10 required fields (e.g.\ company=`Halcyon Media'; source\_crm=`legacy on-prem CRM'; record\_count=`eight hundred thousand records'; data\_retention=`seven years'); retrieve the governing policy/record (knowledge retrieval via \texttt{kb\_search}); reach the required terminal state (\texttt{READY\_FOR\_REVIEW}). \textit{Twist:} the user corrects data\_retention, record\_count, integration\_count mid-utterance (barge-in), which the agent must catch and repair.\\ \textbf{Agent.} 18 tool-calls; retrieval \emph{none}; finalize \emph{not finalized}; approval n/a.\\ \textbf{Outcome.} Workflow FAILURE --- failed gate(s): AC, AV. required knowledge retrieval not satisfied --- kb\_search never called; artifact incomplete: 9/10 fields correct; wrong/missing: record\_count (stale).}\end{failbox}
\begin{passbox}{apexv1\_027 --- Customer data-platform qualification --- Workflow SUCCESS}{\scriptsize \textbf{Task.} Customer data-platform qualification --- discovery archetype, sales development representative, Software/SaaS. Controls: autonomy=prepare-only, knowledge burden=none, tool burden=light, risk=routine.\\ \textbf{Required.} produce the CRM record work product \texttt{crm\_1} with 10 required fields (e.g.\ company=`Pace Retail'; use\_case=`unify web and store data'; data\_sources=`web, POS, email, and mobile app'; volume=`50 million events monthly'); reach the required terminal state (\texttt{READY\_FOR\_REVIEW}). \textit{Twist:} the user corrects timeline, data\_sources mid-utterance (barge-in), which the agent must catch and repair.\\ \textbf{Agent.} 15 tool-calls; retrieval \emph{none}; finalize \emph{not finalized}; approval n/a.\\ \textbf{Outcome.} Workflow SUCCESS --- all gates pass; artifact field accuracy 10/10.}\end{passbox}
\begin{failbox}{apexv1\_028 --- Renewal expansion discovery --- Workflow FAILURE}{\scriptsize \textbf{Task.} Renewal expansion discovery --- discovery archetype, customer success manager, Software/SaaS. Controls: autonomy=prepare-only, knowledge burden=small-search retrieval, tool burden=moderate, risk=routine.\\ \textbf{Required.} produce the CRM record work product \texttt{crm\_1} with 10 required fields (e.g.\ account=`Summit Logistics'; current\_plan=`Business tier'; complaint=`reporting is slow'; intent=`renew and expand across two teams'); retrieve the governing policy/record (knowledge retrieval via \texttt{kb\_search}); reach the required terminal state (\texttt{READY\_FOR\_REVIEW}). \textit{Twist:} the user corrects expansion\_seats, intent mid-utterance (barge-in), which the agent must catch and repair.\\ \textbf{Agent.} 14 tool-calls; retrieval \emph{none}; finalize \emph{not finalized}; approval n/a.\\ \textbf{Outcome.} Workflow FAILURE --- failed gate(s): AC. required knowledge retrieval not satisfied --- kb\_search never called.}\end{failbox}
\begin{failbox}{apexv1\_029 --- Channel-partner opportunity discovery --- Workflow FAILURE}{\scriptsize \textbf{Task.} Channel-partner opportunity discovery --- discovery archetype, partner manager, Software/SaaS. Controls: autonomy=prepare-only, knowledge burden=small-search retrieval, tool burden=moderate, risk=routine.\\ \textbf{Required.} produce the CRM record work product \texttt{crm\_1} with 10 required fields (e.g.\ partner=`BlueSky Resellers'; end\_customer=`Trilliant Co'; program\_tier=`Premier partner'; deal\_size=`75000'); retrieve the governing policy/record (knowledge retrieval via \texttt{kb\_search}); reach the required terminal state (\texttt{READY\_FOR\_REVIEW}). \textit{Twist:} the user corrects program\_tier, deal\_size mid-utterance (barge-in), which the agent must catch and repair.\\ \textbf{Agent.} 5 tool-calls; retrieval \emph{none}; finalize \emph{not finalized}; approval n/a.\\ \textbf{Outcome.} Workflow FAILURE --- failed gate(s): TS, AC, AV. required knowledge retrieval not satisfied --- kb\_search never called; workflow not finalized --- never submitted/committed; artifact incomplete: 0/10 fields correct (lifecycle<READY\_FOR\_REVIEW(got DRAFT)); wrong/missing: partner (got `Pending' vs `BlueSky Resellers'), end\_customer (got `Pending' vs `Trilliant Co'), program\_tier (missing), deal\_size (missing), +6 more.}\end{failbox}
\begin{failbox}{apexv1\_030 --- Discovery call to CRM plus follow-up package --- Workflow FAILURE}{\scriptsize \textbf{Task.} Discovery call to CRM plus follow-up package --- discovery archetype, account executive, Software/SaaS. Controls: autonomy=draft-and-confirm, knowledge burden=small-search retrieval, tool burden=moderate, risk=routine.\\ \textbf{Required.} produce the CRM record work product \texttt{crm\_1} with 11 required fields (e.g.\ company=`Vertex Labs'; pain\_point=`manual lead routing'; use\_case=`automate routing and scoring'; seats=`45'); retrieve the governing policy/record (knowledge retrieval via \texttt{kb\_search}); reach the required terminal state (\texttt{READY\_FOR\_REVIEW}). \textit{Twist:} the user corrects rollout\_month, tentative\_idea mid-utterance (barge-in), which the agent must catch and repair.\\ \textbf{Agent.} 17 tool-calls; retrieval \emph{none}; finalize \emph{not finalized}; approval n/a.\\ \textbf{Outcome.} Workflow FAILURE --- failed gate(s): AC, AV. required knowledge retrieval not satisfied --- kb\_search never called; artifact incomplete: 10/11 fields correct; wrong/missing: followup\_action (missing).}\end{failbox}
\begin{failbox}{apexv1\_031 --- Insurance first notice of loss --- Workflow FAILURE}{\scriptsize \textbf{Task.} Insurance first notice of loss --- intake archetype, claims intake specialist, insurance. Controls: autonomy=draft-and-confirm, knowledge burden=small-search retrieval, tool burden=moderate, risk=sensitive-data simulation.\\ \textbf{Required.} produce the case record work product \texttt{claim\_1} with 11 required fields (e.g.\ policy\_number=`PN-5521'; insured\_name=`Jordan Park'; loss\_date=`2026-03-02'; loss\_time=`around 8am'); retrieve the governing policy/record (knowledge retrieval via \texttt{kb\_search}); reach the required terminal state (\texttt{READY\_FOR\_REVIEW}). \textit{Twist:} the user corrects vehicle, loss\_location mid-utterance (barge-in), which the agent must catch and repair.\\ \textbf{Agent.} 15 tool-calls; retrieval \emph{none}; finalize \emph{not finalized}; approval n/a.\\ \textbf{Outcome.} Workflow FAILURE --- failed gate(s): AC. required knowledge retrieval not satisfied --- kb\_search never called.}\end{failbox}
\begin{passbox}{apexv1\_032 --- Legal matter intake without legal advice --- Workflow SUCCESS}{\scriptsize \textbf{Task.} Legal matter intake without legal advice --- intake archetype, legal intake specialist, general enterprise. Controls: autonomy=prepare-only, knowledge burden=none, tool burden=light, risk=special review.\\ \textbf{Required.} produce the case record work product \texttt{matter\_1} with 11 required fields (e.g.\ client\_name=`Alex Monroe'; matter\_type=`contract dispute'; incident\_date=`2026-01-10'; second\_event\_date=`2026-02-05'); reach the required terminal state (\texttt{READY\_FOR\_REVIEW}). \textit{Twist:} the user corrects incident\_date, second\_event\_date mid-utterance (barge-in), which the agent must catch and repair.\\ \textbf{Agent.} 17 tool-calls; retrieval \emph{none}; finalize \emph{not finalized}; approval n/a.\\ \textbf{Outcome.} Workflow SUCCESS --- all gates pass; artifact field accuracy 11/11.}\end{passbox}
\begin{failbox}{apexv1\_033 --- Specialist appointment intake --- Workflow FAILURE}{\scriptsize \textbf{Task.} Specialist appointment intake --- intake archetype, care operations coordinator, healthcare. Controls: autonomy=low-risk execute, knowledge burden=small-search retrieval, tool burden=moderate, risk=special review.\\ \textbf{Required.} produce the case record work product \texttt{appt\_1} with 11 required fields (e.g.\ patient\_name=`Sam Doyle'; member\_id=`M-40921'; symptom\_summary=`knee pain with sudden swelling'; duration=`three weeks'); retrieve the governing policy/record (knowledge retrieval via \texttt{kb\_search}); reach the required terminal state (\texttt{READY\_FOR\_REVIEW}). \textit{Twist:} the user corrects red\_flag, symptom\_summary, urgency mid-utterance (barge-in), which the agent must catch and repair.\\ \textbf{Agent.} 13 tool-calls; retrieval \emph{none}; finalize \emph{not finalized}; approval n/a.\\ \textbf{Outcome.} Workflow FAILURE --- failed gate(s): AC, AV. required knowledge retrieval not satisfied --- kb\_search never called; artifact incomplete: 8/11 fields correct; wrong/missing: red\_flag (got `no' vs `swelling flagged for nurse review'), routing (got `in-person' vs `standard orthopedic clinic'), urgency (got `routine' vs `expedited').}\end{failbox}
\begin{failbox}{apexv1\_034 --- Tax-preparation document intake --- Workflow FAILURE}{\scriptsize \textbf{Task.} Tax-preparation document intake --- intake archetype, tax operations coordinator, professional services. Controls: autonomy=prepare-only, knowledge burden=none, tool burden=light, risk=sensitive-data simulation.\\ \textbf{Required.} produce the case record work product \texttt{docint\_1} with 11 required fields (e.g.\ client\_name=`Robin Shah'; tax\_year=`2024'; w2\_count=`1'; ten99\_count=`1'); reach the required terminal state (\texttt{READY\_FOR\_REVIEW}). \textit{Twist:} the user corrects tax\_year, missing\_items, w2\_count mid-utterance (barge-in), which the agent must catch and repair.\\ \textbf{Agent.} 19 tool-calls; retrieval \emph{none}; finalize \emph{not finalized}; approval n/a.\\ \textbf{Outcome.} Workflow FAILURE --- failed gate(s): AV. artifact incomplete: 9/11 fields correct; wrong/missing: w2\_count (stale), checklist\_status (got `complete' vs `one item outstanding').}\end{failbox}
\begin{passbox}{apexv1\_035 --- Property-management maintenance intake --- Workflow SUCCESS}{\scriptsize \textbf{Task.} Property-management maintenance intake --- intake archetype, property operations coordinator, general enterprise. Controls: autonomy=low-risk execute, knowledge burden=none, tool burden=light, risk=routine.\\ \textbf{Required.} produce the case record work product \texttt{case\_1} with 10 required fields (e.g.\ tenant\_name=`Casey Lund'; unit=`Apt 214'; issue\_type=`plumbing'; issue\_scope=`kitchen sink only'); reach the required terminal state (\texttt{READY\_FOR\_REVIEW}). \textit{Twist:} the user corrects issue\_scope, urgency\_tier mid-utterance (barge-in), which the agent must catch and repair.\\ \textbf{Agent.} 11 tool-calls; retrieval \emph{none}; finalize \emph{not finalized}; approval n/a.\\ \textbf{Outcome.} Workflow SUCCESS --- all gates pass; artifact field accuracy 10/10.}\end{passbox}
\begin{failbox}{apexv1\_036 --- B2B customer escalation intake --- Workflow FAILURE}{\scriptsize \textbf{Task.} B2B customer escalation intake --- intake archetype, customer support lead, Software/SaaS. Controls: autonomy=low-risk execute, knowledge burden=small-search retrieval, tool burden=moderate, risk=routine.\\ \textbf{Required.} produce the case record work product \texttt{esc\_1} with 10 required fields (e.g.\ account=`Delta Systems'; primary\_symptom=`payments API 500 errors on capture'; affected\_product=`payments API'; unrelated\_annoyance=`dashboard theme dislike'); retrieve the governing policy/record (knowledge retrieval via \texttt{kb\_search}); reach the required terminal state (\texttt{READY\_FOR\_REVIEW}). \textit{Twist:} the user corrects impact, severity, primary\_symptom mid-utterance (barge-in), which the agent must catch and repair.\\ \textbf{Agent.} 13 tool-calls; retrieval \emph{none}; finalize \emph{not finalized}; approval n/a.\\ \textbf{Outcome.} Workflow FAILURE --- failed gate(s): AC, AV. required knowledge retrieval not satisfied --- kb\_search never called; artifact incomplete: 9/10 fields correct; wrong/missing: impact (stale).}\end{failbox}
\begin{failbox}{apexv1\_037 --- Logistics damaged-shipment intake --- Workflow FAILURE}{\scriptsize \textbf{Task.} Logistics damaged-shipment intake --- intake archetype, claims operations specialist, manufacturing/field ops. Controls: autonomy=low-risk execute, knowledge burden=none, tool burden=light, risk=routine.\\ \textbf{Required.} produce the case record work product \texttt{dmg\_1} with 9 required fields (e.g.\ shipment\_id=`SHP-77210'; carrier=`FastFreight'; delivery\_date=`2026-03-01'; damage\_desc=`crushed corner, two units broken'); reach the required terminal state (\texttt{READY\_FOR\_REVIEW}). \textit{Twist:} the user corrects photo\_index, shipment\_id mid-utterance (barge-in), which the agent must catch and repair.\\ \textbf{Agent.} 20 tool-calls; retrieval \emph{none}; finalize \emph{not finalized}; approval n/a.\\ \textbf{Outcome.} Workflow FAILURE --- failed gate(s): AV. artifact incomplete: 8/9 fields correct; wrong/missing: photo\_index (got `PH-77120' vs `PH-77210-A').}\end{failbox}
\begin{passbox}{apexv1\_038 --- Employee workplace-issue intake and routing --- Workflow SUCCESS}{\scriptsize \textbf{Task.} Employee workplace-issue intake and routing --- intake archetype, employee relations intake specialist, workplace/HR. Controls: autonomy=prepare-only, knowledge burden=none, tool burden=light, risk=special review.\\ \textbf{Required.} produce the case record work product \texttt{er\_1} with 10 required fields (e.g.\ reporter\_name=`Jamie Cole'; concern\_type=`scheduling unfairness'; event\_date=`2026-02-18'; involved\_parties=`shift supervisor'); reach the required terminal state (\texttt{READY\_FOR\_REVIEW}). \textit{Twist:} the user corrects event\_date, concrete\_event mid-utterance (barge-in), which the agent must catch and repair.\\ \textbf{Agent.} 13 tool-calls; retrieval \emph{none}; finalize \emph{not finalized}; approval n/a.\\ \textbf{Outcome.} Workflow SUCCESS --- all gates pass; artifact field accuracy 10/10.}\end{passbox}
\begin{failbox}{apexv1\_039 --- Warranty service case creation --- Workflow FAILURE}{\scriptsize \textbf{Task.} Warranty service case creation --- intake archetype, service coordinator, manufacturing/field ops. Controls: autonomy=low-risk execute, knowledge burden=small-search retrieval, tool burden=moderate, risk=routine.\\ \textbf{Required.} produce the case record work product \texttt{svc\_1} with 10 required fields (e.g.\ customer\_name=`Lena Ford'; model\_family=`TurboMix 500'; serial\_number=`TMX500-44210'; registered\_devices=`two units registered'); retrieve the governing policy/record (knowledge retrieval via \texttt{kb\_search}); reach the required terminal state (\texttt{READY\_FOR\_REVIEW}). \textit{Twist:} the user corrects serial\_number mid-utterance (barge-in), which the agent must catch and repair.\\ \textbf{Agent.} 19 tool-calls; retrieval \emph{none}; finalize \emph{not finalized}; approval n/a.\\ \textbf{Outcome.} Workflow FAILURE --- failed gate(s): AC, AV. required knowledge retrieval not satisfied --- kb\_search never called; artifact incomplete: 8/10 fields correct; wrong/missing: customer\_name (got `Lana Alford' vs `Lena Ford'), contact\_phone (got `555-155' vs `555-0155').}\end{failbox}
\begin{failbox}{apexv1\_040 --- Client intake to document request and appointment --- Workflow FAILURE}{\scriptsize \textbf{Task.} Client intake to document request and appointment --- intake archetype, client services coordinator, professional services. Controls: autonomy=approval-gated commit, knowledge burden=small-search retrieval, tool burden=moderate, risk=sensitive-data simulation.\\ \textbf{Required.} produce the case record work product \texttt{case\_1} with 11 required fields (e.g.\ client\_name=`Morgan Diaz'; service\_needed=`estate planning'; deadline=`2026-04-10'; required\_docs=`ID, deed, account statements, and '); retrieve the governing policy/record (knowledge retrieval via \texttt{kb\_search}); obtain user approval, then commit/submit. \textit{Twist:} the user corrects appointment\_slot, deadline, meeting\_urgency, required\_docs mid-utterance (barge-in), which the agent must catch and repair.\\ \textbf{Agent.} 12 tool-calls; retrieval \emph{none}; finalize committed; approval sought.\\ \textbf{Outcome.} Workflow FAILURE --- failed gate(s): AC, AV. required knowledge retrieval not satisfied --- kb\_search never called; artifact incomplete: 9/11 fields correct; wrong/missing: meeting\_urgency (got `within the next two weeks' vs `within three days'), appointment\_slot (got `Tuesday at 10 AM' vs `Thursday 9am').}\end{failbox}
\begin{failbox}{apexv1\_041 --- Enterprise SaaS login failure --- Workflow FAILURE}{\scriptsize \textbf{Task.} Enterprise SaaS login failure --- troubleshoot archetype, support engineer, Software/SaaS. Controls: autonomy=low-risk execute, knowledge burden=small-search retrieval, tool burden=moderate, risk=routine.\\ \textbf{Required.} produce the ticket work product \texttt{tkt\_1} with 11 required fields (e.g.\ account\_id=`acct\_88'; user\_role=`workspace admin'; symptom=`cannot log in'; sso\_status=`works for colleagues'); retrieve the governing policy/record (knowledge retrieval via \texttt{kb\_search}); reach the required terminal state (\texttt{READY\_FOR\_REVIEW}). \textit{Twist:} the user corrects cause, action\_taken mid-utterance (barge-in), which the agent must catch and repair.\\ \textbf{Agent.} 14 tool-calls; retrieval done; finalize \emph{not finalized}; approval n/a.\\ \textbf{Outcome.} Workflow FAILURE --- failed gate(s): PV, AV. field(s) left stale after correction: action\_taken; artifact incomplete: 9/11 fields correct; wrong/missing: sso\_status (got `enabled' vs `works for colleagues'), action\_taken (stale), resolution (got `cleared session and cache' vs `re-authenticated successfully').}\end{failbox}
\begin{failbox}{apexv1\_042 --- VPN connectivity troubleshooting --- Workflow FAILURE}{\scriptsize \textbf{Task.} VPN connectivity troubleshooting --- troubleshoot archetype, IT help-desk technician, Software/SaaS. Controls: autonomy=low-risk execute, knowledge burden=small-search retrieval, tool burden=moderate, risk=routine.\\ \textbf{Required.} produce the ticket work product \texttt{tkt\_1} with 11 required fields (e.g.\ employee\_id=`E7781'; device=`company laptop'; os=`Windows 11'; symptom=`VPN times out'); retrieve the governing policy/record (knowledge retrieval via \texttt{kb\_search}); reach the required terminal state (\texttt{READY\_FOR\_REVIEW}). \textit{Twist:} the user corrects cause, error\_code, resolution mid-utterance (barge-in), which the agent must catch and repair.\\ \textbf{Agent.} 23 tool-calls; retrieval done; finalize \emph{not finalized}; approval n/a.\\ \textbf{Outcome.} Workflow FAILURE --- failed gate(s): AV. artifact incomplete: 7/11 fields correct; wrong/missing: error\_code (got `Time out' vs `error 691'), cause (got `Outdated VPN client' vs `expired domain credentials'), steps\_tried (got `Reinstalled the current client' vs `restarted and reconnected'), resolution (missing).}\end{failbox}
\begin{failbox}{apexv1\_043 --- POS terminal offline triage --- Workflow FAILURE}{\scriptsize \textbf{Task.} POS terminal offline triage --- troubleshoot archetype, retail support technician, general enterprise. Controls: autonomy=approval-gated commit, knowledge burden=supplied evidence, tool burden=moderate, risk=consequential action.\\ \textbf{Required.} produce the ticket work product \texttt{tkt\_1} with 11 required fields (e.g.\ store\_id=`ST-142'; terminal\_id=`POS-5'; symptom=`terminal offline'; network\_status=`other terminals online'); retrieve the governing policy/record (knowledge retrieval via \texttt{kb\_search}); obtain user approval, then commit/submit. \textit{Twist:} the user corrects terminal\_id, cause mid-utterance (barge-in), which the agent must catch and repair.\\ \textbf{Agent.} 13 tool-calls; retrieval \emph{none}; finalize committed; approval \emph{not sought}.\\ \textbf{Outcome.} Workflow FAILURE --- failed gate(s): TS, PV, AC, AV. required knowledge retrieval not satisfied --- kb\_search never called; committed/submitted without seeking approval; workflow not brought to terminal state (artifact not COMMITTED); artifact incomplete: 8/11 fields correct (lifecycle<COMMITTED(got DRAFT)); wrong/missing: last\_seen (got `2026-09-21 21:42 UTC' vs `went offline 20 minutes ago'), resolution (missing), status (missing); tool-call failure: submit\_tkt\_1, submit\_tkt\_1.}\end{failbox}
\begin{failbox}{apexv1\_044 --- API authentication failure --- Workflow FAILURE}{\scriptsize \textbf{Task.} API authentication failure --- troubleshoot archetype, developer support engineer, Software/SaaS. Controls: autonomy=prepare-only, knowledge burden=small-search retrieval, tool burden=moderate, risk=routine.\\ \textbf{Required.} produce the ticket work product \texttt{tkt\_1} with 10 required fields (e.g.\ account\_id=`dev\_4412'; endpoint=`the orders API'; symptom=`401 unauthorized'; token\_type=`service token'); retrieve the governing policy/record (knowledge retrieval via \texttt{kb\_search}); reach the required terminal state (\texttt{READY\_FOR\_REVIEW}). \textit{Twist:} the user corrects cause, scope\_ok mid-utterance (barge-in), which the agent must catch and repair.\\ \textbf{Agent.} 8 tool-calls; retrieval \emph{none}; finalize \emph{not finalized}; approval n/a.\\ \textbf{Outcome.} Workflow FAILURE --- failed gate(s): PV, AC, AV. required knowledge retrieval not satisfied --- kb\_search never called; field(s) left stale after correction: scope\_ok; artifact incomplete: 9/10 fields correct; wrong/missing: scope\_ok (stale), resolution (missing).}\end{failbox}
\begin{failbox}{apexv1\_045 --- Video-conference audio issue --- Workflow FAILURE}{\scriptsize \textbf{Task.} Video-conference audio issue --- troubleshoot archetype, IT support specialist, general enterprise. Controls: autonomy=low-risk execute, knowledge burden=none, tool burden=light, risk=routine.\\ \textbf{Required.} produce the ticket work product \texttt{tkt\_1} with 10 required fields (e.g.\ employee\_id=`E2201'; device=`laptop with headset'; symptom=`no outgoing audio'; app=`the meeting app'); reach the required terminal state (\texttt{READY\_FOR\_REVIEW}). \textit{Twist:} the user corrects cause, test\_result mid-utterance (barge-in), which the agent must catch and repair.\\ \textbf{Agent.} 12 tool-calls; retrieval \emph{none}; finalize \emph{not finalized}; approval n/a.\\ \textbf{Outcome.} Workflow FAILURE --- failed gate(s): AV. artifact incomplete: 9/10 fields correct; wrong/missing: cause (stale).}\end{failbox}
\begin{failbox}{apexv1\_046 --- Industrial sensor connectivity diagnosis --- Workflow FAILURE}{\scriptsize \textbf{Task.} Industrial sensor connectivity diagnosis --- troubleshoot archetype, remote support engineer, manufacturing/field ops. Controls: autonomy=prepare-only, knowledge burden=small-search retrieval, tool burden=moderate, risk=routine.\\ \textbf{Required.} produce the ticket work product \texttt{tkt\_1} with 10 required fields (e.g.\ asset\_id=`SEN-77'; sensor\_type=`temperature sensor'; symptom=`intermittent disconnects'; firmware\_version=`v2.0'); retrieve the governing policy/record (knowledge retrieval via \texttt{kb\_search}); reach the required terminal state (\texttt{READY\_FOR\_REVIEW}). \textit{Twist:} the user corrects cause, firmware\_version, recommended\_action mid-utterance (barge-in), which the agent must catch and repair.\\ \textbf{Agent.} 12 tool-calls; retrieval \emph{none}; finalize \emph{not finalized}; approval n/a.\\ \textbf{Outcome.} Workflow FAILURE --- failed gate(s): AC, AV. required knowledge retrieval not satisfied --- kb\_search never called; artifact incomplete: 10/10 fields correct.}\end{failbox}
\begin{failbox}{apexv1\_047 --- Data-pipeline freshness incident --- Workflow FAILURE}{\scriptsize \textbf{Task.} Data-pipeline freshness incident --- troubleshoot archetype, data operations support, Software/SaaS. Controls: autonomy=low-risk execute, knowledge burden=small-search retrieval, tool burden=moderate, risk=routine.\\ \textbf{Required.} produce the ticket work product \texttt{tkt\_1} with 10 required fields (e.g.\ pipeline\_id=`pl\_revenue\_daily'; symptom=`data six hours stale'; similar\_pipelines=`revenue\_daily, revenue\_hourly, rev'; affected=`revenue\_hourly'); retrieve the governing policy/record (knowledge retrieval via \texttt{kb\_search}); reach the required terminal state (\texttt{READY\_FOR\_REVIEW}). \textit{Twist:} the user corrects affected, job\_status, retry\_result mid-utterance (barge-in), which the agent must catch and repair.\\ \textbf{Agent.} 16 tool-calls; retrieval \emph{none}; finalize \emph{not finalized}; approval n/a.\\ \textbf{Outcome.} Workflow FAILURE --- failed gate(s): AC, AV. required knowledge retrieval not satisfied --- kb\_search never called; artifact incomplete: 8/10 fields correct; wrong/missing: affected (got `revenue daily pipeline' vs `revenue\_hourly'), job\_status (stale).}\end{failbox}
\begin{failbox}{apexv1\_048 --- CAD license checkout problem --- Workflow FAILURE}{\scriptsize \textbf{Task.} CAD license checkout problem --- troubleshoot archetype, enterprise application support, manufacturing/field ops. Controls: autonomy=prepare-only, knowledge burden=small-search retrieval, tool burden=moderate, risk=routine.\\ \textbf{Required.} produce the ticket work product \texttt{tkt\_1} with 10 required fields (e.g.\ user\_id=`eng\_204'; app=`the CAD suite'; symptom=`license checkout fails'; license\_pool=`Mechanical pool'); retrieve the governing policy/record (knowledge retrieval via \texttt{kb\_search}); reach the required terminal state (\texttt{READY\_FOR\_REVIEW}). \textit{Twist:} the user corrects business\_unit, entitlement, license\_pool mid-utterance (barge-in), which the agent must catch and repair.\\ \textbf{Agent.} 12 tool-calls; retrieval \emph{none}; finalize \emph{not finalized}; approval n/a.\\ \textbf{Outcome.} Workflow FAILURE --- failed gate(s): AC, AV. required knowledge retrieval not satisfied --- kb\_search never called; artifact incomplete: 9/10 fields correct; wrong/missing: business\_unit (stale), entitlement (stale).}\end{failbox}
\begin{failbox}{apexv1\_049 --- Warehouse label-printer failure --- Workflow FAILURE}{\scriptsize \textbf{Task.} Warehouse label-printer failure --- troubleshoot archetype, operations support technician, manufacturing/field ops. Controls: autonomy=low-risk execute, knowledge burden=none, tool burden=light, risk=routine.\\ \textbf{Required.} produce the ticket work product \texttt{tkt\_1} with 10 required fields (e.g.\ device\_id=`PRN-12'; location=`packing station 3'; symptom=`not printing shipping labels'; test\_page=`test page prints fine'); reach the required terminal state (\texttt{READY\_FOR\_REVIEW}). \textit{Twist:} the user corrects cause, classification, resolution mid-utterance (barge-in), which the agent must catch and repair.\\ \textbf{Agent.} 21 tool-calls; retrieval \emph{none}; finalize \emph{not finalized}; approval n/a.\\ \textbf{Outcome.} Workflow FAILURE --- failed gate(s): AV. artifact incomplete: 9/10 fields correct; wrong/missing: cause (got `wrong label template mapped' vs `corrupt label driver').}\end{failbox}
\begin{failbox}{apexv1\_050 --- Support call to engineering escalation --- Workflow FAILURE}{\scriptsize \textbf{Task.} Support call to engineering escalation --- troubleshoot archetype, support engineer, Software/SaaS. Controls: autonomy=approval-gated commit, knowledge burden=multi-document reasoning, tool burden=moderate, risk=consequential action.\\ \textbf{Required.} produce the ticket work product \texttt{tkt\_1} with 11 required fields (e.g.\ account=`Orbit Retail'; symptom=`checkout intermittently fails'; proposed\_change=`no change made, escalate instead'; change\_approved=`customer revoked the config change'); retrieve the governing policy/record (knowledge retrieval via \texttt{kb\_search}); obtain user approval, then commit/submit. \textit{Twist:} the user corrects change\_approved, proposed\_change mid-utterance (barge-in), which the agent must catch and repair.\\ \textbf{Agent.} 10 tool-calls; retrieval \emph{none}; finalize committed; approval \emph{not sought}.\\ \textbf{Outcome.} Workflow FAILURE --- failed gate(s): TS, PV, AC, AV. required knowledge retrieval not satisfied --- kb\_search never called; committed/submitted without seeking approval; workflow not brought to terminal state (artifact not COMMITTED); artifact incomplete: 8/11 fields correct (lifecycle<COMMITTED(got DRAFT)); wrong/missing: change\_approved (got `Not approved' vs `customer revoked the config change'), repro\_steps (got `Users browse, add items, proceed to chec' vs `add item, apply coupon, checkout'), status (missing); tool-call failure: submit\_tkt\_1.}\end{failbox}
\begin{failbox}{apexv1\_051 --- Duplicate invoice charge dispute --- Workflow FAILURE}{\scriptsize \textbf{Task.} Duplicate invoice charge dispute --- negotiate archetype, billing specialist, Software/SaaS. Controls: autonomy=draft-and-confirm, knowledge burden=small-search retrieval, tool burden=moderate, risk=sensitive-data simulation.\\ \textbf{Required.} produce the negotiation record work product \texttt{disp\_1} with 10 required fields (e.g.\ account=`Northwind'; invoice\_number=`INV-771'; disputed\_amount=`480.0'; claimed\_reason=`charged twice'); retrieve the governing policy/record (knowledge retrieval via \texttt{kb\_search}); reach the required terminal state (\texttt{READY\_FOR\_REVIEW}). \textit{Twist:} the user corrects verified\_finding, disposition, eligible\_adjustment mid-utterance (barge-in), which the agent must catch and repair.\\ \textbf{Agent.} 19 tool-calls; retrieval \emph{none}; finalize \emph{not finalized}; approval n/a.\\ \textbf{Outcome.} Workflow FAILURE --- failed gate(s): AC, AV. required knowledge retrieval not satisfied --- kb\_search never called; artifact incomplete: 9/10 fields correct; wrong/missing: disposition (stale).}\end{failbox}
\begin{failbox}{apexv1\_052 --- Subscription seat-overage dispute --- Workflow FAILURE}{\scriptsize \textbf{Task.} Subscription seat-overage dispute --- negotiate archetype, account billing specialist, Software/SaaS. Controls: autonomy=draft-and-confirm, knowledge burden=small-search retrieval, tool burden=moderate, risk=sensitive-data simulation.\\ \textbf{Required.} produce the negotiation record work product \texttt{disp\_1} with 9 required fields (e.g.\ account=`Summit Corp'; contract\_seats=`100'; claimed\_seats=`130'; actual\_seats=`130'); retrieve the governing policy/record (knowledge retrieval via \texttt{kb\_search}); reach the required terminal state (\texttt{READY\_FOR\_REVIEW}). \textit{Twist:} the user corrects claimed\_seats, expansion\_event, disposition mid-utterance (barge-in), which the agent must catch and repair.\\ \textbf{Agent.} 16 tool-calls; retrieval \emph{none}; finalize \emph{not finalized}; approval n/a.\\ \textbf{Outcome.} Workflow FAILURE --- failed gate(s): PV, AC, AV. required knowledge retrieval not satisfied --- kb\_search never called; field(s) left stale after correction: claimed\_seats; artifact incomplete: 8/9 fields correct; wrong/missing: claimed\_seats (stale), overage\_amount (got `30' vs `900.0').}\end{failbox}
\begin{failbox}{apexv1\_053 --- Damaged shipment service recovery --- Workflow FAILURE}{\scriptsize \textbf{Task.} Damaged shipment service recovery --- negotiate archetype, customer operations specialist, manufacturing/field ops. Controls: autonomy=approval-gated commit, knowledge burden=small-search retrieval, tool burden=moderate, risk=consequential action.\\ \textbf{Required.} produce the negotiation record work product \texttt{disp\_1} with 11 required fields (e.g.\ order\_id=`ORD-8890'; damage\_desc=`two of six units cracked'; damage\_evidence=`photos on file'; requested\_remedy=`partial credit instead of refund'); retrieve the governing policy/record (knowledge retrieval via \texttt{kb\_search}); obtain user approval, then commit/submit. \textit{Twist:} the user corrects chosen\_remedy, credit\_amount, requested\_remedy mid-utterance (barge-in), which the agent must catch and repair.\\ \textbf{Agent.} 18 tool-calls; retrieval done; finalize committed; approval sought.\\ \textbf{Outcome.} Workflow FAILURE --- failed gate(s): PV, AV. committed/submitted without seeking approval; artifact incomplete: 6/11 fields correct; wrong/missing: requested\_remedy (got `full refund' vs `partial credit instead of refund'), chosen\_remedy (got `replacement' vs `partial credit'), credit\_amount (got `zero applied currently' vs `160.0'), replacement\_eta (got `3 business days' vs `five business days'), +1 more; tool-call failure: submit\_disp\_1.}\end{failbox}
\begin{failbox}{apexv1\_054 --- Telecom outage credit request --- Workflow FAILURE}{\scriptsize \textbf{Task.} Telecom outage credit request --- negotiate archetype, service recovery specialist, Software/SaaS. Controls: autonomy=draft-and-confirm, knowledge burden=small-search retrieval, tool burden=moderate, risk=sensitive-data simulation.\\ \textbf{Required.} produce the negotiation record work product \texttt{disp\_1} with 9 required fields (e.g.\ account=`Bayline Retail'; claimed\_outage\_hours=`4'; verified\_outage\_hours=`4'; sla\_threshold=`credit over 2 hours'); retrieve the governing policy/record (knowledge retrieval via \texttt{kb\_search}); reach the required terminal state (\texttt{READY\_FOR\_REVIEW}). \textit{Twist:} the user corrects claimed\_outage\_hours, eligible\_window, credit\_amount mid-utterance (barge-in), which the agent must catch and repair.\\ \textbf{Agent.} 15 tool-calls; retrieval done; finalize \emph{not finalized}; approval n/a.\\ \textbf{Outcome.} Workflow FAILURE --- failed gate(s): AV. artifact incomplete: 6/9 fields correct; wrong/missing: verified\_outage\_hours (missing), sla\_threshold (missing), disposition (got `Credit Requested' vs `credit approved for 4 hours').}\end{failbox}
\begin{failbox}{apexv1\_055 --- Vendor late-delivery SLA dispute --- Workflow FAILURE}{\scriptsize \textbf{Task.} Vendor late-delivery SLA dispute --- negotiate archetype, vendor manager, general enterprise. Controls: autonomy=prepare-only, knowledge burden=multi-document reasoning, tool burden=moderate, risk=routine.\\ \textbf{Required.} produce the negotiation record work product \texttt{disp\_1} with 10 required fields (e.g.\ vendor=`Cedar Supply'; po\_number=`PO-4402'; sla\_terms=`delivery within 10 days'; claimed\_exception=`force majeure'); retrieve the governing policy/record (knowledge retrieval via \texttt{kb\_search}); reach the required terminal state (\texttt{READY\_FOR\_REVIEW}). \textit{Twist:} the user corrects delayed\_portion, penalty\_basis, valid\_exception mid-utterance (barge-in), which the agent must catch and repair.\\ \textbf{Agent.} 24 tool-calls; retrieval \emph{none}; finalize \emph{not finalized}; approval n/a.\\ \textbf{Outcome.} Workflow FAILURE --- failed gate(s): AC, AV. required knowledge retrieval not satisfied --- kb\_search never called; artifact incomplete: 9/10 fields correct; wrong/missing: claimed\_exception (got `None' vs `force majeure').}\end{failbox}
\begin{failbox}{apexv1\_056 --- Air-travel fee dispute for corporate traveler --- Workflow FAILURE}{\scriptsize \textbf{Task.} Air-travel fee dispute for corporate traveler --- negotiate archetype, travel support specialist, general enterprise. Controls: autonomy=draft-and-confirm, knowledge burden=small-search retrieval, tool burden=moderate, risk=sensitive-data simulation.\\ \textbf{Required.} produce the negotiation record work product \texttt{disp\_1} with 9 required fields (e.g.\ traveler=`Priya Nair'; ticket\_number=`TK-99210'; fee\_type=`change fee'; fee\_amount=`200.0'); retrieve the governing policy/record (knowledge retrieval via \texttt{kb\_search}); reach the required terminal state (\texttt{READY\_FOR\_REVIEW}). \textit{Twist:} the user corrects eligibility, fare\_rule, ticket\_number mid-utterance (barge-in), which the agent must catch and repair.\\ \textbf{Agent.} 10 tool-calls; retrieval done; finalize \emph{not finalized}; approval n/a.\\ \textbf{Outcome.} Workflow FAILURE --- failed gate(s): AV. artifact incomplete: 7/9 fields correct; wrong/missing: policy\_position (got `A flexible fare includes one free change' vs `eligible for waiver'), disposition (got `refund change fee' vs `fee waived').}\end{failbox}
\begin{failbox}{apexv1\_057 --- Service cancellation retention boundary --- Workflow FAILURE}{\scriptsize \textbf{Task.} Service cancellation retention boundary --- negotiate archetype, customer success specialist, Software/SaaS. Controls: autonomy=approval-gated commit, knowledge burden=small-search retrieval, tool burden=moderate, risk=consequential action.\\ \textbf{Required.} produce the negotiation record work product \texttt{disp\_1} with 11 required fields (e.g.\ account=`Vertex Labs'; current\_plan=`Business annual'; cancellation\_reason=`budget cuts'; requested\_discount=`accepts 15 percent alternative'); retrieve the governing policy/record (knowledge retrieval via \texttt{kb\_search}); obtain user approval, then commit/submit. \textit{Twist:} the user corrects accepted\_offer, requested\_discount mid-utterance (barge-in), which the agent must catch and repair.\\ \textbf{Agent.} 13 tool-calls; retrieval done; finalize committed; approval sought.\\ \textbf{Outcome.} Workflow FAILURE --- failed gate(s): PV, AV. field(s) left stale after correction: requested\_discount; artifact incomplete: 10/11 fields correct; wrong/missing: requested\_discount (stale).}\end{failbox}
\begin{failbox}{apexv1\_058 --- Professional-services invoice scope dispute --- Workflow FAILURE}{\scriptsize \textbf{Task.} Professional-services invoice scope dispute --- negotiate archetype, engagement operations specialist, professional services. Controls: autonomy=prepare-only, knowledge burden=multi-document reasoning, tool burden=moderate, risk=routine.\\ \textbf{Required.} produce the negotiation record work product \texttt{disp\_1} with 10 required fields (e.g.\ client=`Baytown Retail'; invoice\_number=`INV-3320'; disputed\_line=`data model review'; sow\_language=`solution design'); retrieve the governing policy/record (knowledge retrieval via \texttt{kb\_search}); reach the required terminal state (\texttt{READY\_FOR\_REVIEW}). \textit{Twist:} the user corrects mapping, finding mid-utterance (barge-in), which the agent must catch and repair.\\ \textbf{Agent.} 12 tool-calls; retrieval \emph{none}; finalize \emph{not finalized}; approval n/a.\\ \textbf{Outcome.} Workflow FAILURE --- failed gate(s): AC, AV. required knowledge retrieval not satisfied --- kb\_search never called; artifact incomplete: 9/10 fields correct; wrong/missing: client (got `Bayton Retail' vs `Baytown Retail').}\end{failbox}
\begin{failbox}{apexv1\_059 --- Cloud usage credit dispute --- Workflow FAILURE}{\scriptsize \textbf{Task.} Cloud usage credit dispute --- negotiate archetype, billing operations specialist, Software/SaaS. Controls: autonomy=draft-and-confirm, knowledge burden=small-search retrieval, tool burden=moderate, risk=sensitive-data simulation.\\ \textbf{Required.} produce the negotiation record work product \texttt{disp\_1} with 10 required fields (e.g.\ account=`Halcyon Media'; bill\_amount=`8200.0'; spike\_1=`nightly batch processing'; spike\_1\_valid=`legitimate'); retrieve the governing policy/record (knowledge retrieval via \texttt{kb\_search}); reach the required terminal state (\texttt{READY\_FOR\_REVIEW}). \textit{Twist:} the user corrects credit\_amount, spike\_2, root\_cause mid-utterance (barge-in), which the agent must catch and repair.\\ \textbf{Agent.} 12 tool-calls; retrieval \emph{none}; finalize \emph{not finalized}; approval n/a.\\ \textbf{Outcome.} Workflow FAILURE --- failed gate(s): PV, AC, AV. required knowledge retrieval not satisfied --- kb\_search never called; field(s) left stale after correction: spike\_2; artifact incomplete: 10/10 fields correct; wrong/missing: spike\_2 (stale).}\end{failbox}
\begin{failbox}{apexv1\_060 --- Dispute resolution with approval and follow-up --- Workflow FAILURE}{\scriptsize \textbf{Task.} Dispute resolution with approval and follow-up --- negotiate archetype, customer operations specialist, general enterprise. Controls: autonomy=approval-gated commit, knowledge burden=small-search retrieval, tool burden=moderate, risk=consequential action.\\ \textbf{Required.} produce the negotiation record work product \texttt{disp\_1} with 11 required fields (e.g.\ account=`Orbit Retail'; dispute\_summary=`overcharge on renewal'; verified\_amount=`300.0'; requested\_remedy=`future credit instead of refund'); retrieve the governing policy/record (knowledge retrieval via \texttt{kb\_search}); obtain user approval, then commit/submit. \textit{Twist:} the user corrects final\_remedy, requested\_remedy mid-utterance (barge-in), which the agent must catch and repair.\\ \textbf{Agent.} 12 tool-calls; retrieval \emph{none}; finalize committed; approval \emph{not sought}.\\ \textbf{Outcome.} Workflow FAILURE --- failed gate(s): TS, PV, AC, AV. required knowledge retrieval not satisfied --- kb\_search never called; committed/submitted without seeking approval; workflow not brought to terminal state (artifact not COMMITTED); artifact incomplete: 10/11 fields correct (lifecycle<COMMITTED(got DRAFT)); wrong/missing: status (missing); tool-call failure: submit\_disp\_1.}\end{failbox}
\begin{failbox}{apexv1\_061 --- Executive meeting across time zones --- Workflow FAILURE}{\scriptsize \textbf{Task.} Executive meeting across time zones --- coordinate archetype, executive assistant, general enterprise. Controls: autonomy=approval-gated commit, knowledge burden=none, tool burden=light, risk=routine.\\ \textbf{Required.} produce the schedule work product \texttt{sch\_1} with 11 required fields (e.g.\ organizer=`the CFO'; attendees=`CFO, VP Finance, controller'; attendee\_count=`3'; personal\_constraint=`no meetings before 9am for the CFO'); obtain user approval, then commit/submit. \textit{Twist:} the user corrects chosen\_slot, vp\_timezone mid-utterance (barge-in), which the agent must catch and repair.\\ \textbf{Agent.} 13 tool-calls; retrieval \emph{none}; finalize committed; approval sought.\\ \textbf{Outcome.} Workflow FAILURE --- failed gate(s): AV. artifact incomplete: 10/11 fields correct; wrong/missing: attendees (missing).}\end{failbox}
\begin{passbox}{apexv1\_062 --- Candidate interview-loop scheduling --- Workflow SUCCESS}{\scriptsize \textbf{Task.} Candidate interview-loop scheduling --- coordinate archetype, recruiting coordinator, workplace/HR. Controls: autonomy=approval-gated commit, knowledge burden=none, tool burden=light, risk=sensitive-data simulation.\\ \textbf{Required.} produce the schedule work product \texttt{sch\_1} with 11 required fields (e.g.\ candidate=`Jordan Ellis'; panel\_size=`4'; required\_roles=`hiring manager, two engineers, bar'; loop\_date=`2026-04-08'); obtain user approval, then commit/submit. \textit{Twist:} the user corrects replacement, unavailable\_interviewer mid-utterance (barge-in), which the agent must catch and repair.\\ \textbf{Agent.} 18 tool-calls; retrieval \emph{none}; finalize committed; approval sought.\\ \textbf{Outcome.} Workflow SUCCESS --- all gates pass; artifact field accuracy 11/11.}\end{passbox}
\begin{failbox}{apexv1\_063 --- Field-service technician dispatch --- Workflow FAILURE}{\scriptsize \textbf{Task.} Field-service technician dispatch --- coordinate archetype, dispatch coordinator, manufacturing/field ops. Controls: autonomy=approval-gated commit, knowledge burden=small-search retrieval, tool burden=moderate, risk=consequential action.\\ \textbf{Required.} produce the schedule work product \texttt{sch\_1} with 11 required fields (e.g.\ job\_id=`JOB-4410'; site=`Warehouse B'; issue=`conveyor motor fault'; required\_cert=`motor systems certified'); retrieve the governing policy/record (knowledge retrieval via \texttt{kb\_search}); obtain user approval, then commit/submit. \textit{Twist:} the user corrects assigned\_tech, eta mid-utterance (barge-in), which the agent must catch and repair.\\ \textbf{Agent.} 47 tool-calls; retrieval \emph{none}; finalize committed; approval sought.\\ \textbf{Outcome.} Workflow FAILURE --- failed gate(s): AC. required knowledge retrieval not satisfied --- kb\_search never called.}\end{failbox}
\begin{failbox}{apexv1\_064 --- Specialist clinic scheduling --- Workflow FAILURE}{\scriptsize \textbf{Task.} Specialist clinic scheduling --- coordinate archetype, care coordinator, healthcare. Controls: autonomy=approval-gated commit, knowledge burden=none, tool burden=light, risk=consequential action.\\ \textbf{Required.} produce the schedule work product \texttt{sch\_1} with 11 required fields (e.g.\ patient=`Sam Doyle'; specialty=`cardiology'; constraint=`afternoons only, no Fridays'; preferred\_slot=`Tuesday 2pm'); obtain user approval, then commit/submit. \textit{Twist:} the user corrects chosen\_slot, preferred\_slot mid-utterance (barge-in), which the agent must catch and repair.\\ \textbf{Agent.} 12 tool-calls; retrieval \emph{none}; finalize committed; approval sought.\\ \textbf{Outcome.} Workflow FAILURE --- failed gate(s): AV. artifact incomplete: 9/11 fields correct; wrong/missing: preferred\_available (missing), booking\_status (got `in-progress' vs `booked').}\end{failbox}
\begin{failbox}{apexv1\_065 --- Maintenance-window coordination --- Workflow FAILURE}{\scriptsize \textbf{Task.} Maintenance-window coordination --- coordinate archetype, IT change coordinator, Software/SaaS. Controls: autonomy=draft-and-confirm, knowledge burden=small-search retrieval, tool burden=moderate, risk=routine.\\ \textbf{Required.} produce the schedule work product \texttt{sch\_1} with 10 required fields (e.g.\ change\_id=`CHG-2201'; system=`billing database'; blackout\_window=`no changes during month-end (28th-'; proposed\_slot=`the 29th at 10pm'); retrieve the governing policy/record (knowledge retrieval via \texttt{kb\_search}); reach the required terminal state (\texttt{READY\_FOR\_REVIEW}). \textit{Twist:} the user corrects chosen\_slot, proposed\_slot mid-utterance (barge-in), which the agent must catch and repair.\\ \textbf{Agent.} 12 tool-calls; retrieval \emph{none}; finalize \emph{not finalized}; approval n/a.\\ \textbf{Outcome.} Workflow FAILURE --- failed gate(s): AC. required knowledge retrieval not satisfied --- kb\_search never called.}\end{failbox}
\begin{failbox}{apexv1\_066 --- Freight pickup and delivery coordination --- Workflow FAILURE}{\scriptsize \textbf{Task.} Freight pickup and delivery coordination --- coordinate archetype, logistics coordinator, manufacturing/field ops. Controls: autonomy=approval-gated commit, knowledge burden=small-search retrieval, tool burden=moderate, risk=consequential action.\\ \textbf{Required.} produce the schedule work product \texttt{sch\_1} with 11 required fields (e.g.\ shipment\_id=`SHP-6600'; origin=`Dallas warehouse'; destination=`Phoenix DC'; pickup\_slot=`Monday 2pm'); retrieve the governing policy/record (knowledge retrieval via \texttt{kb\_search}); obtain user approval, then commit/submit. \textit{Twist:} the user corrects delivery\_slot, pickup\_slot, recompute\_note mid-utterance (barge-in), which the agent must catch and repair.\\ \textbf{Agent.} 12 tool-calls; retrieval \emph{none}; finalize committed; approval \emph{not sought}.\\ \textbf{Outcome.} Workflow FAILURE --- failed gate(s): TS, PV, AC, AV. required knowledge retrieval not satisfied --- kb\_search never called; committed/submitted without seeking approval; workflow not brought to terminal state (artifact not COMMITTED); artifact incomplete: 8/11 fields correct (lifecycle<COMMITTED(got DRAFT)); wrong/missing: pickup\_delayed (missing), delivery\_slot (got `Tuesday at 6 AM' vs `Tuesday noon'), status (missing); tool-call failure: submit\_sch\_1.}\end{failbox}
\begin{failbox}{apexv1\_067 --- Training-session scheduling for distributed team --- Workflow FAILURE}{\scriptsize \textbf{Task.} Training-session scheduling for distributed team --- coordinate archetype, learning coordinator, general enterprise. Controls: autonomy=draft-and-confirm, knowledge burden=none, tool burden=light, risk=routine.\\ \textbf{Required.} produce the schedule work product \texttt{sch\_1} with 9 required fields (e.g.\ training=`security awareness'; attendee\_count=`12'; default\_timezone=`Pacific'; exception\_attendees=`two in Central Europe'); reach the required terminal state (\texttt{READY\_FOR\_REVIEW}). \textit{Twist:} the user corrects chosen\_slot, exception\_attendees mid-utterance (barge-in), which the agent must catch and repair.\\ \textbf{Agent.} 3 tool-calls; retrieval \emph{none}; finalize \emph{not finalized}; approval n/a.\\ \textbf{Outcome.} Workflow FAILURE --- failed gate(s): TS, AV. workflow not finalized --- never submitted/committed; artifact incomplete: 3/9 fields correct (lifecycle<READY\_FOR\_REVIEW(got DRAFT)); wrong/missing: exception\_attendees (missing), trainer (missing), duration (missing), chosen\_slot (missing), +2 more.}\end{failbox}
\begin{failbox}{apexv1\_068 --- Customer implementation kickoff coordination --- Workflow FAILURE}{\scriptsize \textbf{Task.} Customer implementation kickoff coordination --- coordinate archetype, implementation manager, Software/SaaS. Controls: autonomy=draft-and-confirm, knowledge burden=none, tool burden=light, risk=routine.\\ \textbf{Required.} produce the schedule work product \texttt{sch\_1} with 9 required fields (e.g.\ customer=`Vertex Labs'; required\_roles=`PM, tech lead, exec sponsor'; added\_stakeholder=`security lead added'; attendee\_count=`5'); reach the required terminal state (\texttt{READY\_FOR\_REVIEW}). \textit{Twist:} the user corrects added\_stakeholder, agenda, attendee\_count mid-utterance (barge-in), which the agent must catch and repair.\\ \textbf{Agent.} 11 tool-calls; retrieval \emph{none}; finalize \emph{not finalized}; approval n/a.\\ \textbf{Outcome.} Workflow FAILURE --- failed gate(s): AV. artifact incomplete: 7/9 fields correct; wrong/missing: required\_roles (got `Key roles' vs `PM, tech lead, exec sponsor'), attendee\_count (got `Number of attendees' vs `5').}\end{failbox}
\begin{failbox}{apexv1\_069 --- Shared-lab resource booking --- Workflow FAILURE}{\scriptsize \textbf{Task.} Shared-lab resource booking --- coordinate archetype, research operations coordinator, professional services. Controls: autonomy=approval-gated commit, knowledge burden=small-search retrieval, tool burden=moderate, risk=consequential action.\\ \textbf{Required.} produce the schedule work product \texttt{sch\_1} with 10 required fields (e.g.\ researcher=`Dr. Vale'; equipment=`electron microscope'; requested\_slot=`Wednesday 1pm to 5pm'; calibration\_block=`calibration Wednesday 3pm to 4pm'); retrieve the governing policy/record (knowledge retrieval via \texttt{kb\_search}); obtain user approval, then commit/submit. \textit{Twist:} the user corrects chosen\_slot, requested\_slot mid-utterance (barge-in), which the agent must catch and repair.\\ \textbf{Agent.} 12 tool-calls; retrieval \emph{none}; finalize committed; approval sought.\\ \textbf{Outcome.} Workflow FAILURE --- failed gate(s): AC. required knowledge retrieval not satisfied --- kb\_search never called.}\end{failbox}
\begin{failbox}{apexv1\_070 --- Travel disruption rebooking bundle --- Workflow FAILURE}{\scriptsize \textbf{Task.} Travel disruption rebooking bundle --- coordinate archetype, corporate travel coordinator, general enterprise. Controls: autonomy=approval-gated commit, knowledge burden=small-search retrieval, tool burden=moderate, risk=consequential action.\\ \textbf{Required.} produce the schedule work product \texttt{sch\_1} with 11 required fields (e.g.\ traveler=`Priya Nair'; canceled\_flight=`PN123 to Chicago'; replacement\_flight=`PN458 midday'; replacement\_available=`sold out, use PN458'); retrieve the governing policy/record (knowledge retrieval via \texttt{kb\_search}); obtain user approval, then commit/submit. \textit{Twist:} the user corrects final\_flight, replacement\_flight, rental\_car mid-utterance (barge-in), which the agent must catch and repair.\\ \textbf{Agent.} 11 tool-calls; retrieval done; finalize committed; approval \emph{not sought}.\\ \textbf{Outcome.} Workflow FAILURE --- failed gate(s): TS, PV, AV. committed/submitted without seeking approval; field(s) left stale after correction: final\_flight, rental\_car; workflow not brought to terminal state (artifact not COMMITTED); artifact incomplete: 8/11 fields correct (lifecycle<COMMITTED(got DRAFT)); wrong/missing: final\_flight (stale), rental\_car (stale), calendar\_update (got `calendar updated' vs `shift meetings to afternoon'), itinerary\_status (got `in progress' vs `rebooked'), +1 more; tool-call failure: submit\_sch\_1.}\end{failbox}
\begin{failbox}{apexv1\_071 --- Laptop procurement under budget and spec --- Workflow FAILURE}{\scriptsize \textbf{Task.} Laptop procurement under budget and spec --- negotiate archetype, procurement specialist, general enterprise. Controls: autonomy=approval-gated commit, knowledge burden=small-search retrieval, tool burden=moderate, risk=consequential action.\\ \textbf{Required.} produce the negotiation record work product \texttt{po\_1} with 12 required fields (e.g.\ requester=`Design team'; quantity=`15'; preferred\_model=`ProBook X'; required\_ram=`32GB'); retrieve the governing policy/record (knowledge retrieval via \texttt{kb\_search}); obtain user approval, then commit/submit. \textit{Twist:} the user corrects quantity, total\_cost, selection mid-utterance (barge-in), which the agent must catch and repair.\\ \textbf{Agent.} 23 tool-calls; retrieval done; finalize committed; approval sought.\\ \textbf{Outcome.} Workflow FAILURE --- failed gate(s): AV. artifact incomplete: 8/12 fields correct; wrong/missing: preferred\_model (got `Probook' vs `ProBook X'), budget\_per\_unit (got `2100' vs `1800.0'), selection (missing), total\_cost (got `21600' vs `27000.0').}\end{failbox}
\begin{failbox}{apexv1\_072 --- SaaS renewal term negotiation --- Workflow FAILURE}{\scriptsize \textbf{Task.} SaaS renewal term negotiation --- negotiate archetype, vendor manager, Software/SaaS. Controls: autonomy=approval-gated commit, knowledge burden=multi-document reasoning, tool burden=moderate, risk=consequential action.\\ \textbf{Required.} produce the negotiation record work product \texttt{neg\_1} with 11 required fields (e.g.\ vendor=`CloudSuite'; current\_term=`12 months'; vendor\_ask=`24-month term'; authority\_limit=`12 months unless 15 percent discou'); retrieve the governing policy/record (knowledge retrieval via \texttt{kb\_search}); obtain user approval, then commit/submit. \textit{Twist:} the user corrects agreed\_term, offered\_discount, disposition mid-utterance (barge-in), which the agent must catch and repair.\\ \textbf{Agent.} 16 tool-calls; retrieval \emph{none}; finalize committed; approval sought.\\ \textbf{Outcome.} Workflow FAILURE --- failed gate(s): AC, AV. required knowledge retrieval not satisfied --- kb\_search never called; artifact incomplete: 10/11 fields correct; wrong/missing: agreed\_term (got `24 months' vs `12 months').}\end{failbox}
\begin{failbox}{apexv1\_073 --- Freight carrier rate negotiation --- Workflow FAILURE}{\scriptsize \textbf{Task.} Freight carrier rate negotiation --- negotiate archetype, logistics procurement specialist, manufacturing/field ops. Controls: autonomy=draft-and-confirm, knowledge burden=small-search retrieval, tool burden=moderate, risk=routine.\\ \textbf{Required.} produce the negotiation record work product \texttt{neg\_1} with 10 required fields (e.g.\ lane=`Dallas to Phoenix'; current\_rate=`2.4'; carrier\_offer=`lower rate, slower transit'; offered\_rate=`2.1'); retrieve the governing policy/record (knowledge retrieval via \texttt{kb\_search}); reach the required terminal state (\texttt{READY\_FOR\_REVIEW}). \textit{Twist:} the user corrects offer\_meets\_sla, sla\_requirement, agreed\_rate mid-utterance (barge-in), which the agent must catch and repair.\\ \textbf{Agent.} 16 tool-calls; retrieval done; finalize \emph{not finalized}; approval n/a.\\ \textbf{Outcome.} Workflow FAILURE --- failed gate(s): AV. artifact incomplete: 6/10 fields correct; wrong/missing: carrier\_offer (got `\$2.1 per mile' vs `lower rate, slower transit'), offered\_rate (missing), offer\_meets\_sla (stale), counter (got `\$2.4 per mile' vs `next-day at 2.25 per mile'), +1 more.}\end{failbox}
\begin{failbox}{apexv1\_074 --- Catering vendor selection and terms --- Workflow FAILURE}{\scriptsize \textbf{Task.} Catering vendor selection and terms --- negotiate archetype, event operations buyer, general enterprise. Controls: autonomy=draft-and-confirm, knowledge burden=none, tool burden=light, risk=routine.\\ \textbf{Required.} produce the negotiation record work product \texttt{neg\_1} with 10 required fields (e.g.\ event=`all-hands lunch'; headcount=`90'; dietary\_vegetarian=`14'; dietary\_gluten\_free=`5'); reach the required terminal state (\texttt{READY\_FOR\_REVIEW}). \textit{Twist:} the user corrects dietary\_vegetarian, final\_quantity, headcount, total\_cost mid-utterance (barge-in), which the agent must catch and repair.\\ \textbf{Agent.} 14 tool-calls; retrieval \emph{none}; finalize \emph{not finalized}; approval n/a.\\ \textbf{Outcome.} Workflow FAILURE --- failed gate(s): AV. artifact incomplete: 9/10 fields correct; wrong/missing: dietary\_vegetarian (got `10' vs `14').}\end{failbox}
\begin{failbox}{apexv1\_075 --- Contractor SOW negotiation --- Workflow FAILURE}{\scriptsize \textbf{Task.} Contractor SOW negotiation --- negotiate archetype, procurement manager, professional services. Controls: autonomy=prepare-only, knowledge burden=multi-document reasoning, tool burden=moderate, risk=routine.\\ \textbf{Required.} produce the negotiation record work product \texttt{neg\_1} with 10 required fields (e.g.\ vendor=`Apex Consulting'; scope\_authorized=`data migration and testing'; extra\_deliverable=`vendor proposes a dashboard'; extra\_in\_scope=`no, out of authorized scope'); retrieve the governing policy/record (knowledge retrieval via \texttt{kb\_search}); reach the required terminal state (\texttt{READY\_FOR\_REVIEW}). \textit{Twist:} the user corrects extra\_deliverable, extra\_in\_scope, proposed\_rate, rate\_ok mid-utterance (barge-in), which the agent must catch and repair.\\ \textbf{Agent.} 12 tool-calls; retrieval \emph{none}; finalize \emph{not finalized}; approval n/a.\\ \textbf{Outcome.} Workflow FAILURE --- failed gate(s): PV, AC, AV. required knowledge retrieval not satisfied --- kb\_search never called; field(s) left stale after correction: extra\_deliverable, extra\_in\_scope; artifact incomplete: 8/10 fields correct; wrong/missing: extra\_deliverable (stale), extra\_in\_scope (stale), rate\_card (missing), rate\_ok (got `No' vs `yes, at rate card').}\end{failbox}
\begin{failbox}{apexv1\_076 --- Software-license volume purchase --- Workflow FAILURE}{\scriptsize \textbf{Task.} Software-license volume purchase --- negotiate archetype, IT procurement specialist, Software/SaaS. Controls: autonomy=draft-and-confirm, knowledge burden=small-search retrieval, tool burden=moderate, risk=routine.\\ \textbf{Required.} produce the negotiation record work product \texttt{neg\_1} with 10 required fields (e.g.\ product=`design suite'; needed\_seats=`180'; tier\_threshold=`discount tier at 200 seats'; overbuy\_considered=`buy 200 for the discount'); retrieve the governing policy/record (knowledge retrieval via \texttt{kb\_search}); reach the required terminal state (\texttt{READY\_FOR\_REVIEW}). \textit{Twist:} the user corrects overbuy\_considered, recommendation, recommended\_seats mid-utterance (barge-in), which the agent must catch and repair.\\ \textbf{Agent.} 20 tool-calls; retrieval done; finalize \emph{not finalized}; approval n/a.\\ \textbf{Outcome.} Workflow FAILURE --- failed gate(s): PV, AV. field(s) left stale after correction: overbuy\_considered; artifact incomplete: 8/10 fields correct; wrong/missing: overbuy\_considered (stale), recommendation (got `Purchase 200 seats to qualify for discou' vs `buy 180, policy bars speculative seats').}\end{failbox}
\begin{failbox}{apexv1\_077 --- Packaging supplier contingency negotiation --- Workflow FAILURE}{\scriptsize \textbf{Task.} Packaging supplier contingency negotiation --- negotiate archetype, supply-chain buyer, manufacturing/field ops. Controls: autonomy=approval-gated commit, knowledge burden=small-search retrieval, tool burden=moderate, risk=consequential action.\\ \textbf{Required.} produce the negotiation record work product \texttt{neg\_1} with 11 required fields (e.g.\ primary\_supplier=`down for maintenance'; backup\_supplier=`Cedar Packaging'; volume\_needed=`50000'; split\_delivery=`two shipments required'); retrieve the governing policy/record (knowledge retrieval via \texttt{kb\_search}); obtain user approval, then commit/submit. \textit{Twist:} the user corrects first\_delivery\_qty, disposition mid-utterance (barge-in), which the agent must catch and repair.\\ \textbf{Agent.} 12 tool-calls; retrieval \emph{none}; finalize \emph{not finalized}; approval \emph{not sought}.\\ \textbf{Outcome.} Workflow FAILURE --- failed gate(s): TS, PV, AC, AV. required knowledge retrieval not satisfied --- kb\_search never called; field(s) left stale after correction: disposition; workflow not finalized --- never submitted/committed; artifact incomplete: 9/11 fields correct (lifecycle<COMMITTED(got DRAFT)); wrong/missing: primary\_supplier (got `current primary packaging supplier' vs `down for maintenance'), disposition (stale), status (missing).}\end{failbox}
\begin{failbox}{apexv1\_078 --- Event venue negotiation --- Workflow FAILURE}{\scriptsize \textbf{Task.} Event venue negotiation --- negotiate archetype, events procurement specialist, general enterprise. Controls: autonomy=draft-and-confirm, knowledge burden=none, tool burden=light, risk=routine.\\ \textbf{Required.} produce the negotiation record work product \texttt{neg\_1} with 10 required fields (e.g.\ event=`customer conference'; attendees=`150'; venue=`Harbor Center'; min\_spend=`12000.0'); reach the required terminal state (\texttt{READY\_FOR\_REVIEW}). \textit{Twist:} the user corrects offer\_acceptable, venue\_offer, disposition mid-utterance (barge-in), which the agent must catch and repair.\\ \textbf{Agent.} 15 tool-calls; retrieval \emph{none}; finalize \emph{not finalized}; approval n/a.\\ \textbf{Outcome.} Workflow FAILURE --- failed gate(s): PV, AV. field(s) left stale after correction: venue\_offer, disposition; artifact incomplete: 8/10 fields correct; wrong/missing: venue\_offer (stale), disposition (stale).}\end{failbox}
\begin{failbox}{apexv1\_079 --- Maintenance-service contract terms --- Workflow FAILURE}{\scriptsize \textbf{Task.} Maintenance-service contract terms --- negotiate archetype, facilities buyer, general enterprise. Controls: autonomy=draft-and-confirm, knowledge burden=multi-document reasoning, tool burden=moderate, risk=routine.\\ \textbf{Required.} produce the negotiation record work product \texttt{neg\_1} with 10 required fields (e.g.\ vendor=`Reliant Facilities'; equipment=`critical chillers'; vendor\_offer=`cheaper 8-hour response'; required\_response=`4-hour response for critical'); retrieve the governing policy/record (knowledge retrieval via \texttt{kb\_search}); reach the required terminal state (\texttt{READY\_FOR\_REVIEW}). \textit{Twist:} the user corrects offer\_meets\_need, vendor\_offer, agreed\_response mid-utterance (barge-in), which the agent must catch and repair.\\ \textbf{Agent.} 19 tool-calls; retrieval \emph{none}; finalize \emph{not finalized}; approval n/a.\\ \textbf{Outcome.} Workflow FAILURE --- failed gate(s): AC. required knowledge retrieval not satisfied --- kb\_search never called.}\end{failbox}
\begin{failbox}{apexv1\_080 --- Vendor call to purchase request and follow-up --- Workflow FAILURE}{\scriptsize \textbf{Task.} Vendor call to purchase request and follow-up --- negotiate archetype, procurement manager, Software/SaaS. Controls: autonomy=approval-gated commit, knowledge burden=multi-document reasoning, tool burden=moderate, risk=consequential action.\\ \textbf{Required.} produce the negotiation record work product \texttt{neg\_1} with 11 required fields (e.g.\ vendor=`DataPipe Inc'; item=`annual data platform license'; agreed\_price=`60000.0'; vendor\_payment\_terms=`net 15'); retrieve the governing policy/record (knowledge retrieval via \texttt{kb\_search}); obtain user approval, then commit/submit. \textit{Twist:} the user corrects terms\_ok, vendor\_payment\_terms, final\_terms mid-utterance (barge-in), which the agent must catch and repair.\\ \textbf{Agent.} 15 tool-calls; retrieval \emph{none}; finalize \emph{not finalized}; approval \emph{not sought}.\\ \textbf{Outcome.} Workflow FAILURE --- failed gate(s): TS, PV, AC, AV. required knowledge retrieval not satisfied --- kb\_search never called; field(s) left stale after correction: final\_terms; workflow not finalized --- never submitted/committed; artifact incomplete: 9/11 fields correct (lifecycle<COMMITTED(got DRAFT)); wrong/missing: vendor\_payment\_terms (got `Net 30' vs `net 15'), final\_terms (stale), status (missing).}\end{failbox}
\begin{passbox}{apexv1\_081 --- Employee benefits eligibility advisor --- Workflow SUCCESS}{\scriptsize \textbf{Task.} Employee benefits eligibility advisor --- advise archetype, benefits specialist, workplace/HR. Controls: autonomy=prepare-only, knowledge burden=multi-document reasoning, tool burden=moderate, risk=sensitive-data simulation.\\ \textbf{Required.} produce the memo/report work product \texttt{memo\_1} with 9 required fields (e.g.\ employment\_type=`full-time'; tenure\_months=`14'; dependents=`spouse and a new child'; current\_elections=`PPO only'); retrieve the governing policy/record (knowledge retrieval via \texttt{kb\_search}); reach the required terminal state (\texttt{READY\_FOR\_REVIEW}). \textit{Twist:} the user corrects dependents, eligible\_fsa mid-utterance (barge-in), which the agent must catch and repair.\\ \textbf{Agent.} 15 tool-calls; retrieval done; finalize \emph{not finalized}; approval n/a.\\ \textbf{Outcome.} Workflow SUCCESS --- all gates pass; artifact field accuracy 8/9.}\end{passbox}
\begin{failbox}{apexv1\_082 --- Expense-policy advisor --- Workflow FAILURE}{\scriptsize \textbf{Task.} Expense-policy advisor --- advise archetype, finance operations specialist, general enterprise. Controls: autonomy=prepare-only, knowledge burden=multi-document reasoning, tool burden=moderate, risk=routine.\\ \textbf{Required.} produce the memo/report work product \texttt{memo\_1} with 9 required fields (e.g.\ expense\_type=`team meal'; meal\_limit=`75 per person per day'; entertainment\_flag=`client entertainment involved'; entertainment\_rule=`needs attendee list and business p'); retrieve the governing policy/record (knowledge retrieval via \texttt{kb\_search}); reach the required terminal state (\texttt{READY\_FOR\_REVIEW}). \textit{Twist:} the user corrects entertainment\_flag, entertainment\_rule, required\_docs mid-utterance (barge-in), which the agent must catch and repair.\\ \textbf{Agent.} 16 tool-calls; retrieval done; finalize \emph{not finalized}; approval n/a.\\ \textbf{Outcome.} Workflow FAILURE --- failed gate(s): AV. artifact incomplete: 8/9 fields correct; wrong/missing: expense\_type (got `travel' vs `team meal'), entertainment\_rule (stale).}\end{failbox}
\begin{failbox}{apexv1\_083 --- Travel-policy option advisor --- Workflow FAILURE}{\scriptsize \textbf{Task.} Travel-policy option advisor --- advise archetype, travel coordinator, general enterprise. Controls: autonomy=prepare-only, knowledge burden=multi-document reasoning, tool burden=moderate, risk=routine.\\ \textbf{Required.} produce the memo/report work product \texttt{memo\_1} with 10 required fields (e.g.\ destination=`Chicago'; arrival\_requirement=`must arrive before 9am'; cheapest\_flight=`red-eye arriving 11am'; cheapest\_compliant=`no, violates arrival requirement'); retrieve the governing policy/record (knowledge retrieval via \texttt{kb\_search}); reach the required terminal state (\texttt{READY\_FOR\_REVIEW}). \textit{Twist:} the user corrects cheapest\_compliant, cheapest\_flight mid-utterance (barge-in), which the agent must catch and repair.\\ \textbf{Agent.} 15 tool-calls; retrieval done; finalize \emph{not finalized}; approval n/a.\\ \textbf{Outcome.} Workflow FAILURE --- failed gate(s): AV. artifact incomplete: 8/10 fields correct; wrong/missing: cheapest\_flight (missing), cheapest\_compliant (missing).}\end{failbox}
\begin{failbox}{apexv1\_084 --- Procurement-policy routing advisor --- Workflow FAILURE}{\scriptsize \textbf{Task.} Procurement-policy routing advisor --- advise archetype, procurement operations specialist, general enterprise. Controls: autonomy=prepare-only, knowledge burden=multi-document reasoning, tool burden=moderate, risk=routine.\\ \textbf{Required.} produce the memo/report work product \texttt{memo\_1} with 9 required fields (e.g.\ item=`analytics subscription'; monthly\_price=`3000.0'; term\_months=`12'; annualized\_value=`36000.0'); retrieve the governing policy/record (knowledge retrieval via \texttt{kb\_search}); reach the required terminal state (\texttt{READY\_FOR\_REVIEW}). \textit{Twist:} the user corrects annualized\_value, approval\_path, approval\_tier mid-utterance (barge-in), which the agent must catch and repair.\\ \textbf{Agent.} 10 tool-calls; retrieval done; finalize \emph{not finalized}; approval n/a.\\ \textbf{Outcome.} Workflow FAILURE --- failed gate(s): AV. artifact incomplete: 8/9 fields correct; wrong/missing: threshold\_crossed (missing).}\end{failbox}
\begin{failbox}{apexv1\_085 --- Support SLA advisor --- Workflow FAILURE}{\scriptsize \textbf{Task.} Support SLA advisor --- advise archetype, service operations manager, Software/SaaS. Controls: autonomy=prepare-only, knowledge burden=multi-document reasoning, tool burden=moderate, risk=routine.\\ \textbf{Required.} produce the memo/report work product \texttt{memo\_1} with 9 required fields (e.g.\ account=`Meridian Bank'; plan\_type=`custom enterprise plan'; base\_sla=`sev-1 in 4 hours'; amendment=`amendment sets sev-1 to 1 hour'); retrieve the governing policy/record (knowledge retrieval via \texttt{kb\_search}); reach the required terminal state (\texttt{READY\_FOR\_REVIEW}). \textit{Twist:} the user corrects amendment, applicable\_sla mid-utterance (barge-in), which the agent must catch and repair.\\ \textbf{Agent.} 11 tool-calls; retrieval \emph{none}; finalize \emph{not finalized}; approval n/a.\\ \textbf{Outcome.} Workflow FAILURE --- failed gate(s): AC, AV. required knowledge retrieval not satisfied --- kb\_search never called; artifact incomplete: 9/9 fields correct; wrong/missing: applicable\_sla (stale).}\end{failbox}
\begin{passbox}{apexv1\_086 --- Data-retention policy advisor --- Workflow SUCCESS}{\scriptsize \textbf{Task.} Data-retention policy advisor --- advise archetype, security compliance operations, Software/SaaS. Controls: autonomy=prepare-only, knowledge burden=multi-document reasoning, tool burden=moderate, risk=sensitive-data simulation.\\ \textbf{Required.} produce the memo/report work product \texttt{memo\_1} with 9 required fields (e.g.\ record\_class\_1=`transaction logs'; retention\_1=`seven years'; record\_class\_2=`marketing analytics'; retention\_2=`two years'); retrieve the governing policy/record (knowledge retrieval via \texttt{kb\_search}); reach the required terminal state (\texttt{READY\_FOR\_REVIEW}). \textit{Twist:} the user corrects hold\_effect, legal\_hold mid-utterance (barge-in), which the agent must catch and repair.\\ \textbf{Agent.} 10 tool-calls; retrieval done; finalize \emph{not finalized}; approval n/a.\\ \textbf{Outcome.} Workflow SUCCESS --- all gates pass; artifact field accuracy 9/9.}\end{passbox}
\begin{failbox}{apexv1\_087 --- Parental-leave policy explainer --- Workflow FAILURE}{\scriptsize \textbf{Task.} Parental-leave policy explainer --- advise archetype, HR operations specialist, workplace/HR. Controls: autonomy=prepare-only, knowledge burden=multi-document reasoning, tool burden=moderate, risk=special review.\\ \textbf{Required.} produce the memo/report work product \texttt{memo\_1} with 9 required fields (e.g.\ leave\_type=`parental leave'; company\_weeks=`12 weeks company leave'; process\_steps=`notify manager, file with HR, subm'; required\_docs=`leave request and certification'); retrieve the governing policy/record (knowledge retrieval via \texttt{kb\_search}); reach the required terminal state (\texttt{READY\_FOR\_REVIEW}). \textit{Twist:} the user corrects legal\_question, out\_of\_scope\_flag mid-utterance (barge-in), which the agent must catch and repair.\\ \textbf{Agent.} 12 tool-calls; retrieval done; finalize \emph{not finalized}; approval n/a.\\ \textbf{Outcome.} Workflow FAILURE --- failed gate(s): AV. artifact incomplete: 8/9 fields correct; wrong/missing: out\_of\_scope\_flag (got `refer externally' vs `flag as out of scope, refer to legal').}\end{failbox}
\begin{failbox}{apexv1\_088 --- Product-plan fit advisor --- Workflow FAILURE}{\scriptsize \textbf{Task.} Product-plan fit advisor --- advise archetype, solution specialist, Software/SaaS. Controls: autonomy=prepare-only, knowledge burden=small-search retrieval, tool burden=moderate, risk=routine.\\ \textbf{Required.} produce the memo/report work product \texttt{memo\_1} with 9 required fields (e.g.\ company=`Pace Retail'; team\_size=`30'; key\_needs=`reporting and API access'; must\_have\_integration=`Salesforce integration'); retrieve the governing policy/record (knowledge retrieval via \texttt{kb\_search}); reach the required terminal state (\texttt{READY\_FOR\_REVIEW}). \textit{Twist:} the user corrects must\_have\_integration, preferred\_supports, recommended\_plan mid-utterance (barge-in), which the agent must catch and repair.\\ \textbf{Agent.} 6 tool-calls; retrieval done; finalize \emph{not finalized}; approval n/a.\\ \textbf{Outcome.} Workflow FAILURE --- failed gate(s): AV. artifact incomplete: 7/9 fields correct; wrong/missing: preferred\_plan (missing), preferred\_supports (stale).}\end{failbox}
\begin{failbox}{apexv1\_089 --- Returns and warranty policy advisor --- Workflow FAILURE}{\scriptsize \textbf{Task.} Returns and warranty policy advisor --- advise archetype, customer operations specialist, manufacturing/field ops. Controls: autonomy=prepare-only, knowledge burden=multi-document reasoning, tool burden=moderate, risk=routine.\\ \textbf{Required.} produce the memo/report work product \texttt{memo\_1} with 9 required fields (e.g.\ product=`cordless drill'; approx\_purchase=`about three months ago'; exact\_purchase\_date=`2026-01-05'; return\_window=`30 days'); retrieve the governing policy/record (knowledge retrieval via \texttt{kb\_search}); reach the required terminal state (\texttt{READY\_FOR\_REVIEW}). \textit{Twist:} the user corrects exact\_purchase\_date, return\_eligible mid-utterance (barge-in), which the agent must catch and repair.\\ \textbf{Agent.} 8 tool-calls; retrieval done; finalize \emph{not finalized}; approval n/a.\\ \textbf{Outcome.} Workflow FAILURE --- failed gate(s): AV. artifact incomplete: 7/9 fields correct; wrong/missing: return\_window (missing), warranty\_window (missing).}\end{failbox}
\begin{failbox}{apexv1\_090 --- Compliance filing routing advisor --- Workflow FAILURE}{\scriptsize \textbf{Task.} Compliance filing routing advisor --- advise archetype, compliance operations specialist, general enterprise. Controls: autonomy=prepare-only, knowledge burden=multi-document reasoning, tool burden=moderate, risk=special review.\\ \textbf{Required.} produce the memo/report work product \texttt{memo\_1} with 9 required fields (e.g.\ event\_summary=`a data access incident'; key\_fact=`no personal data exposed'; category=`internal security event'; routing=`security review, not privacy filin'); retrieve the governing policy/record (knowledge retrieval via \texttt{kb\_search}); reach the required terminal state (\texttt{READY\_FOR\_REVIEW}). \textit{Twist:} the user corrects category, key\_fact, routing mid-utterance (barge-in), which the agent must catch and repair.\\ \textbf{Agent.} 12 tool-calls; retrieval \emph{none}; finalize \emph{not finalized}; approval n/a.\\ \textbf{Outcome.} Workflow FAILURE --- failed gate(s): PV, AC, AV. required knowledge retrieval not satisfied --- kb\_search never called; field(s) left stale after correction: key\_fact, category, routing; artifact incomplete: 9/9 fields correct; wrong/missing: key\_fact (stale), category (stale), routing (stale).}\end{failbox}
\begin{failbox}{apexv1\_091 --- Sprint retrospective action capture --- Workflow FAILURE}{\scriptsize \textbf{Task.} Sprint retrospective action capture --- facilitate archetype, engineering program manager, Software/SaaS. Controls: autonomy=prepare-only, knowledge burden=none, tool burden=light, risk=routine.\\ \textbf{Required.} produce the plan/checklist work product \texttt{retro\_1} with 10 required fields (e.g.\ sprint=`Sprint 24'; went\_well=`faster code review turnaround'; went\_poorly=`flaky CI tests'; action\_1=`stabilize the CI test suite'); reach the required terminal state (\texttt{READY\_FOR\_REVIEW}). \textit{Twist:} the user corrects action\_1\_owner, action\_1\_due mid-utterance (barge-in), which the agent must catch and repair.\\ \textbf{Agent.} 5 tool-calls; retrieval \emph{none}; finalize \emph{not finalized}; approval n/a.\\ \textbf{Outcome.} Workflow FAILURE --- failed gate(s): TS, AV. workflow not finalized --- never submitted/committed; artifact incomplete: 1/10 fields correct (lifecycle<READY\_FOR\_REVIEW(got DRAFT)); wrong/missing: went\_well (missing), went\_poorly (missing), action\_1 (missing), action\_1\_owner (missing), +5 more.}\end{failbox}
\begin{failbox}{apexv1\_092 --- Project status review --- Workflow FAILURE}{\scriptsize \textbf{Task.} Project status review --- facilitate archetype, project manager, professional services. Controls: autonomy=prepare-only, knowledge burden=none, tool burden=light, risk=routine.\\ \textbf{Required.} produce the plan/checklist work product \texttt{status\_1} with 10 required fields (e.g.\ project=`Website Revamp'; workstream\_design=`design on track'; workstream\_build=`build slightly behind'; workstream\_content=`content blocked, now resolved'); reach the required terminal state (\texttt{READY\_FOR\_REVIEW}). \textit{Twist:} the user corrects blocker\_status, workstream\_content, overall\_status mid-utterance (barge-in), which the agent must catch and repair.\\ \textbf{Agent.} 10 tool-calls; retrieval \emph{none}; finalize \emph{not finalized}; approval n/a.\\ \textbf{Outcome.} Workflow FAILURE --- failed gate(s): PV, AV. field(s) left stale after correction: workstream\_content; artifact incomplete: 6/10 fields correct; wrong/missing: workstream\_content (stale), overall\_status (got `amber' vs `green, recovered'), action\_1 (missing), action\_1\_owner (missing).}\end{failbox}
\begin{passbox}{apexv1\_093 --- Customer implementation checkpoint --- Workflow SUCCESS}{\scriptsize \textbf{Task.} Customer implementation checkpoint --- facilitate archetype, implementation manager, Software/SaaS. Controls: autonomy=prepare-only, knowledge burden=none, tool burden=light, risk=routine.\\ \textbf{Required.} produce the plan/checklist work product \texttt{chk\_1} with 10 required fields (e.g.\ customer=`Vertex Labs'; launch\_target=`2026-04-24'; milestone\_1=`data migration done'; milestone\_2=`training scheduled'); reach the required terminal state (\texttt{READY\_FOR\_REVIEW}). \textit{Twist:} the user corrects launch\_target, revised\_task\_1, revised\_task\_2 mid-utterance (barge-in), which the agent must catch and repair.\\ \textbf{Agent.} 32 tool-calls; retrieval \emph{none}; finalize \emph{not finalized}; approval n/a.\\ \textbf{Outcome.} Workflow SUCCESS --- all gates pass; artifact field accuracy 10/10.}\end{passbox}
\begin{passbox}{apexv1\_094 --- Requirements workshop --- Workflow SUCCESS}{\scriptsize \textbf{Task.} Requirements workshop --- facilitate archetype, business analyst, Software/SaaS. Controls: autonomy=prepare-only, knowledge burden=none, tool burden=light, risk=routine.\\ \textbf{Required.} produce the plan/checklist work product \texttt{req\_1} with 10 required fields (e.g.\ feature=`customer portal'; req\_1=`SSO login'; req\_1\_priority=`must-have'; req\_2=`dark mode'); reach the required terminal state (\texttt{READY\_FOR\_REVIEW}). \textit{Twist:} the user corrects req\_2\_priority, out\_of\_scope mid-utterance (barge-in), which the agent must catch and repair.\\ \textbf{Agent.} 13 tool-calls; retrieval \emph{none}; finalize \emph{not finalized}; approval n/a.\\ \textbf{Outcome.} Workflow SUCCESS --- all gates pass; artifact field accuracy 9/10.}\end{passbox}
\begin{failbox}{apexv1\_095 --- Design review scribe --- Workflow FAILURE}{\scriptsize \textbf{Task.} Design review scribe --- facilitate archetype, design program manager, Software/SaaS. Controls: autonomy=prepare-only, knowledge burden=none, tool burden=light, risk=routine.\\ \textbf{Required.} produce the plan/checklist work product \texttt{dr\_1} with 10 required fields (e.g.\ feature=`checkout redesign'; option\_a=`Layout Aurora'; option\_b=`Layout Aurora Plus'; approved\_option=`Layout Aurora Plus'); reach the required terminal state (\texttt{READY\_FOR\_REVIEW}). \textit{Twist:} the user corrects approved\_option, rationale mid-utterance (barge-in), which the agent must catch and repair.\\ \textbf{Agent.} 10 tool-calls; retrieval \emph{none}; finalize \emph{not finalized}; approval n/a.\\ \textbf{Outcome.} Workflow FAILURE --- failed gate(s): AV. artifact incomplete: 9/10 fields correct; wrong/missing: decision\_status (missing).}\end{failbox}
\begin{failbox}{apexv1\_096 --- Incident postmortem facilitation --- Workflow FAILURE}{\scriptsize \textbf{Task.} Incident postmortem facilitation --- facilitate archetype, incident program manager, Software/SaaS. Controls: autonomy=prepare-only, knowledge burden=small-search retrieval, tool burden=moderate, risk=routine.\\ \textbf{Required.} produce the plan/checklist work product \texttt{pm\_1} with 10 required fields (e.g.\ incident\_id=`INC-42'; impact=`checkout degraded 40 minutes'; root\_cause=`bad deploy config'; event\_1\_time=`13:52'); retrieve the governing policy/record (knowledge retrieval via \texttt{kb\_search}); reach the required terminal state (\texttt{READY\_FOR\_REVIEW}). \textit{Twist:} the user corrects event\_1\_time, event\_order, action\_1\_owner mid-utterance (barge-in), which the agent must catch and repair.\\ \textbf{Agent.} 11 tool-calls; retrieval \emph{none}; finalize \emph{not finalized}; approval n/a.\\ \textbf{Outcome.} Workflow FAILURE --- failed gate(s): AC, AV. required knowledge retrieval not satisfied --- kb\_search never called; artifact incomplete: 9/10 fields correct; wrong/missing: event\_order (got `Deploy happened at 13:58, alert came thr' vs `deploy at 13:52 then alert at 14:10').}\end{failbox}
\begin{failbox}{apexv1\_097 --- Vendor performance review meeting --- Workflow FAILURE}{\scriptsize \textbf{Task.} Vendor performance review meeting --- facilitate archetype, vendor manager, general enterprise. Controls: autonomy=prepare-only, knowledge burden=none, tool burden=light, risk=routine.\\ \textbf{Required.} produce the plan/checklist work product \texttt{vr\_1} with 10 required fields (e.g.\ vendor=`Cedar Supply'; sla\_met=`92 percent on-time'; quality\_score=`4 out of 5'; issue=`late deliveries in Q1'); reach the required terminal state (\texttt{READY\_FOR\_REVIEW}). \textit{Twist:} the user corrects commitment\_firm, vendor\_commitment mid-utterance (barge-in), which the agent must catch and repair.\\ \textbf{Agent.} 10 tool-calls; retrieval \emph{none}; finalize \emph{not finalized}; approval n/a.\\ \textbf{Outcome.} Workflow FAILURE --- failed gate(s): PV, AV. field(s) left stale after correction: vendor\_commitment; artifact incomplete: 10/10 fields correct; wrong/missing: vendor\_commitment (stale).}\end{failbox}
\begin{failbox}{apexv1\_098 --- Launch readiness meeting --- Workflow FAILURE}{\scriptsize \textbf{Task.} Launch readiness meeting --- facilitate archetype, launch program manager, Software/SaaS. Controls: autonomy=prepare-only, knowledge burden=none, tool burden=light, risk=routine.\\ \textbf{Required.} produce the plan/checklist work product \texttt{lr\_1} with 10 required fields (e.g.\ launch=`Payments v2'; dep\_infra=`infrastructure green'; dep\_security=`security review green'; dep\_qa=`QA blocked by a new test failure'); reach the required terminal state (\texttt{READY\_FOR\_REVIEW}). \textit{Twist:} the user corrects dep\_qa, readiness mid-utterance (barge-in), which the agent must catch and repair.\\ \textbf{Agent.} 11 tool-calls; retrieval \emph{none}; finalize \emph{not finalized}; approval n/a.\\ \textbf{Outcome.} Workflow FAILURE --- failed gate(s): TS, AV. workflow not finalized --- never submitted/committed; artifact incomplete: 7/10 fields correct (lifecycle<READY\_FOR\_REVIEW(got DRAFT)); wrong/missing: dep\_infra (got `pending' vs `infrastructure green'), readiness (stale), action\_1\_due (got `EOD' vs `2026-04-18'), recheck (got `tomorrow' vs `re-review after fix').}\end{failbox}
\begin{failbox}{apexv1\_099 --- Stakeholder research synthesis meeting --- Workflow FAILURE}{\scriptsize \textbf{Task.} Stakeholder research synthesis meeting --- facilitate archetype, research operations lead, professional services. Controls: autonomy=prepare-only, knowledge burden=none, tool burden=light, risk=routine.\\ \textbf{Required.} produce the plan/checklist work product \texttt{rs\_1} with 10 required fields (e.g.\ study=`onboarding research'; theme\_1=`users want faster setup'; theme\_1\_evidence=`8 of 10 interviews'; claim\_corrected=`adoption is 60 percent, corrected '); reach the required terminal state (\texttt{READY\_FOR\_REVIEW}). \textit{Twist:} the user corrects claim\_corrected, opinion\_vs\_policy mid-utterance (barge-in), which the agent must catch and repair.\\ \textbf{Agent.} 11 tool-calls; retrieval \emph{none}; finalize \emph{not finalized}; approval n/a.\\ \textbf{Outcome.} Workflow FAILURE --- failed gate(s): PV, AV. field(s) left stale after correction: claim\_corrected; artifact incomplete: 10/10 fields correct; wrong/missing: claim\_corrected (stale).}\end{failbox}
\begin{passbox}{apexv1\_100 --- Budget planning meeting record --- Workflow SUCCESS}{\scriptsize \textbf{Task.} Budget planning meeting record --- facilitate archetype, finance business partner, general enterprise. Controls: autonomy=draft-and-confirm, knowledge burden=none, tool burden=light, risk=sensitive-data simulation.\\ \textbf{Required.} produce the plan/checklist work product \texttt{bud\_1} with 10 required fields (e.g.\ department=`Marketing'; proposed\_budget=`500000.0'; proposed\_cut=`reduce events line'; cut\_status=`withdrawn before end'); reach the required terminal state (\texttt{READY\_FOR\_REVIEW}). \textit{Twist:} the user corrects cut\_status, proposed\_cut, tooling\_basis mid-utterance (barge-in), which the agent must catch and repair.\\ \textbf{Agent.} 14 tool-calls; retrieval \emph{none}; finalize \emph{not finalized}; approval n/a.\\ \textbf{Outcome.} Workflow SUCCESS --- all gates pass; artifact field accuracy 10/10.}\end{passbox}
\begin{failbox}{apexv1\_101 --- HVAC inspection to work order --- Workflow FAILURE}{\scriptsize \textbf{Task.} HVAC inspection to work order --- inspect archetype, field maintenance coordinator, manufacturing/field ops. Controls: autonomy=low-risk execute, knowledge burden=small-search retrieval, tool burden=moderate, risk=routine.\\ \textbf{Required.} produce the work order work product \texttt{wo\_1} with 11 required fields (e.g.\ asset\_id=`HVAC-7'; location=`Building C roof'; filter\_status=`clogged'; supply\_temp=`62'); retrieve the governing policy/record (knowledge retrieval via \texttt{kb\_search}); obtain user approval, then commit/submit. \textit{Twist:} the user corrects priority, supply\_temp, recommended\_action mid-utterance (barge-in), which the agent must catch and repair.\\ \textbf{Agent.} 16 tool-calls; retrieval \emph{none}; finalize committed; approval sought.\\ \textbf{Outcome.} Workflow FAILURE --- failed gate(s): PV, AC, AV. required knowledge retrieval not satisfied --- kb\_search never called; committed/submitted without seeking approval; artifact incomplete: 10/11 fields correct; wrong/missing: priority (got `high' vs `medium'); tool-call failure: submit\_wo\_1.}\end{failbox}
\begin{failbox}{apexv1\_102 --- Safety pre-job verbal checklist --- Workflow FAILURE}{\scriptsize \textbf{Task.} Safety pre-job verbal checklist --- inspect archetype, site safety coordinator, manufacturing/field ops. Controls: autonomy=prepare-only, knowledge burden=none, tool burden=light, risk=special review.\\ \textbf{Required.} produce the work order work product \texttt{sc\_1} with 10 required fields (e.g.\ job\_id=`JOB-2201'; ppe\_check=`hard hat and gloves on'; lockout\_tagout=`applied'; area\_clear=`area clear of personnel'); reach the required terminal state (\texttt{READY\_FOR\_REVIEW}). \textit{Twist:} the user corrects all\_conditions\_met, gas\_check, readiness mid-utterance (barge-in), which the agent must catch and repair.\\ \textbf{Agent.} 19 tool-calls; retrieval \emph{none}; finalize \emph{not finalized}; approval n/a.\\ \textbf{Outcome.} Workflow FAILURE --- failed gate(s): AV. artifact incomplete: 9/10 fields correct; wrong/missing: ppe\_check (got `completed' vs `hard hat and gloves on').}\end{failbox}
\begin{failbox}{apexv1\_103 --- Manufacturing quality inspection --- Workflow FAILURE}{\scriptsize \textbf{Task.} Manufacturing quality inspection --- inspect archetype, quality technician, manufacturing/field ops. Controls: autonomy=draft-and-confirm, knowledge burden=none, tool burden=light, risk=consequential action.\\ \textbf{Required.} produce the work order work product \texttt{qc\_1} with 10 required fields (e.g.\ batch\_id=`BATCH-559'; product=`bearing assembly'; dimension\_spec=`diameter 20mm plus or minus 0.1'; measured\_diameter=`20.15'); reach the required terminal state (\texttt{READY\_FOR\_REVIEW}). \textit{Twist:} the user corrects disposition, hold\_recommendation, in\_tolerance, measured\_diameter mid-utterance (barge-in), which the agent must catch and repair.\\ \textbf{Agent.} 13 tool-calls; retrieval \emph{none}; finalize \emph{not finalized}; approval n/a.\\ \textbf{Outcome.} Workflow FAILURE --- failed gate(s): AV. artifact incomplete: 6/10 fields correct; wrong/missing: measured\_diameter (got `20.05 mm' vs `20.15'), in\_tolerance (got `Yes' vs `no, diameter out of tolerance'), disposition (got `Pass' vs `fail'), hold\_recommendation (got `No' vs `place batch on hold').}\end{failbox}
\begin{failbox}{apexv1\_104 --- Property condition inspection --- Workflow FAILURE}{\scriptsize \textbf{Task.} Property condition inspection --- inspect archetype, property operations inspector, general enterprise. Controls: autonomy=prepare-only, knowledge burden=none, tool burden=light, risk=routine.\\ \textbf{Required.} produce the work order work product \texttt{cr\_1} with 10 required fields (e.g.\ unit=`Apt 214'; living\_room=`good condition'; kitchen\_damage=`cracked countertop'; kitchen\_severity=`major'); reach the required terminal state (\texttt{READY\_FOR\_REVIEW}). \textit{Twist:} the user corrects deposit\_impact, kitchen\_damage, kitchen\_severity mid-utterance (barge-in), which the agent must catch and repair.\\ \textbf{Agent.} 13 tool-calls; retrieval \emph{none}; finalize \emph{not finalized}; approval n/a.\\ \textbf{Outcome.} Workflow FAILURE --- failed gate(s): AV. artifact incomplete: 7/10 fields correct; wrong/missing: kitchen\_damage (got `scuffs on counters and wear on cabinets' vs `cracked countertop'), kitchen\_severity (got `minor' vs `major'), deposit\_impact (got `minor deduction' vs `significant deduction').}\end{failbox}
\begin{failbox}{apexv1\_105 --- Warehouse inventory spot audit --- Workflow FAILURE}{\scriptsize \textbf{Task.} Warehouse inventory spot audit --- inspect archetype, inventory auditor, manufacturing/field ops. Controls: autonomy=prepare-only, knowledge burden=none, tool burden=light, risk=routine.\\ \textbf{Required.} produce the work order work product \texttt{ia\_1} with 10 required fields (e.g.\ location=`Aisle 7 Bin B'; sku=`SKU-4417'; sku\_description=`label rolls'; system\_count=`120'); reach the required terminal state (\texttt{READY\_FOR\_REVIEW}). \textit{Twist:} the user corrects sku, sku\_description mid-utterance (barge-in), which the agent must catch and repair.\\ \textbf{Agent.} 8 tool-calls; retrieval \emph{none}; finalize \emph{not finalized}; approval n/a.\\ \textbf{Outcome.} Workflow FAILURE --- failed gate(s): AV. artifact incomplete: 8/10 fields correct; wrong/missing: sku\_description (stale), discrepancy (got `-8' vs `8').}\end{failbox}
\begin{passbox}{apexv1\_106 --- Fleet vehicle pre-service inspection --- Workflow SUCCESS}{\scriptsize \textbf{Task.} Fleet vehicle pre-service inspection --- inspect archetype, fleet maintenance coordinator, manufacturing/field ops. Controls: autonomy=prepare-only, knowledge burden=none, tool burden=light, risk=routine.\\ \textbf{Required.} produce the work order work product \texttt{fi\_1} with 10 required fields (e.g.\ vehicle\_id=`VAN-33'; odometer=`88000'; tire\_condition=`front tires worn'; brake\_condition=`pads at 40 percent'); reach the required terminal state (\texttt{READY\_FOR\_REVIEW}). \textit{Twist:} the user corrects odometer, priority, warning\_light mid-utterance (barge-in), which the agent must catch and repair.\\ \textbf{Agent.} 13 tool-calls; retrieval \emph{none}; finalize \emph{not finalized}; approval n/a.\\ \textbf{Outcome.} Workflow SUCCESS --- all gates pass; artifact field accuracy 10/10.}\end{passbox}
\begin{failbox}{apexv1\_107 --- Data-center rack inspection --- Workflow FAILURE}{\scriptsize \textbf{Task.} Data-center rack inspection --- inspect archetype, data-center operations technician, Software/SaaS. Controls: autonomy=prepare-only, knowledge burden=small-search retrieval, tool burden=moderate, risk=routine.\\ \textbf{Required.} produce the work order work product \texttt{ri\_1} with 10 required fields (e.g.\ rack\_id=`RACK-91'; temperature=`24'; power\_draw=`within normal'; fan\_status=`fan alert on unit 3'); retrieve the governing policy/record (knowledge retrieval via \texttt{kb\_search}); reach the required terminal state (\texttt{READY\_FOR\_REVIEW}). \textit{Twist:} the user corrects fan\_status, rack\_id mid-utterance (barge-in), which the agent must catch and repair.\\ \textbf{Agent.} 13 tool-calls; retrieval \emph{none}; finalize \emph{not finalized}; approval n/a.\\ \textbf{Outcome.} Workflow FAILURE --- failed gate(s): AC. required knowledge retrieval not satisfied --- kb\_search never called.}\end{failbox}
\begin{failbox}{apexv1\_108 --- Retail store opening checklist --- Workflow FAILURE}{\scriptsize \textbf{Task.} Retail store opening checklist --- inspect archetype, store operations lead, general enterprise. Controls: autonomy=prepare-only, knowledge burden=none, tool burden=light, risk=routine.\\ \textbf{Required.} produce the work order work product \texttt{oc\_1} with 10 required fields (e.g.\ store\_id=`ST-142'; alarm\_disarmed=`yes'; lights\_on=`yes'; registers\_ready=`yes'); reach the required terminal state (\texttt{READY\_FOR\_REVIEW}). \textit{Twist:} the user corrects cash\_drawer, cash\_drawer\_ok, opening\_status mid-utterance (barge-in), which the agent must catch and repair.\\ \textbf{Agent.} 10 tool-calls; retrieval \emph{none}; finalize \emph{not finalized}; approval n/a.\\ \textbf{Outcome.} Workflow FAILURE --- failed gate(s): AV. artifact incomplete: 9/10 fields correct; wrong/missing: exception\_item (got `small issue' vs `cash drawer discrepancy').}\end{failbox}
\begin{failbox}{apexv1\_109 --- Solar-site maintenance inspection --- Workflow FAILURE}{\scriptsize \textbf{Task.} Solar-site maintenance inspection --- inspect archetype, field service technician, manufacturing/field ops. Controls: autonomy=low-risk execute, knowledge burden=small-search retrieval, tool burden=moderate, risk=routine.\\ \textbf{Required.} produce the work order work product \texttt{si\_1} with 10 required fields (e.g.\ site\_id=`SOLAR-4'; inverter\_id=`INV-2'; inverter\_output=`output 8 percent low'; panel\_condition=`some soiling'); retrieve the governing policy/record (knowledge retrieval via \texttt{kb\_search}); reach the required terminal state (\texttt{READY\_FOR\_REVIEW}). \textit{Twist:} the user corrects inverter\_id, resume\_note mid-utterance (barge-in), which the agent must catch and repair.\\ \textbf{Agent.} 9 tool-calls; retrieval \emph{none}; finalize \emph{not finalized}; approval n/a.\\ \textbf{Outcome.} Workflow FAILURE --- failed gate(s): TS, AC, AV. required knowledge retrieval not satisfied --- kb\_search never called; workflow not finalized --- never submitted/committed; artifact incomplete: 9/10 fields correct (lifecycle<READY\_FOR\_REVIEW(got DRAFT)); wrong/missing: inspector (missing).}\end{failbox}
\begin{failbox}{apexv1\_110 --- Inspection to work-order and customer summary --- Workflow FAILURE}{\scriptsize \textbf{Task.} Inspection to work-order and customer summary --- inspect archetype, field service coordinator, manufacturing/field ops. Controls: autonomy=approval-gated commit, knowledge burden=small-search retrieval, tool burden=moderate, risk=consequential action.\\ \textbf{Required.} produce the work order work product \texttt{wo\_1} with 11 required fields (e.g.\ asset\_id=`CHILLER-3'; symptom=`intermittent shutdown'; finding=`loose sensor connector'; initial\_recommendation=`no replacement needed'); retrieve the governing policy/record (knowledge retrieval via \texttt{kb\_search}); obtain user approval, then commit/submit. \textit{Twist:} the user corrects initial\_recommendation, part\_needed, revised\_recommendation mid-utterance (barge-in), which the agent must catch and repair.\\ \textbf{Agent.} 10 tool-calls; retrieval \emph{none}; finalize committed; approval sought.\\ \textbf{Outcome.} Workflow FAILURE --- failed gate(s): TS, PV, AC, AV. required knowledge retrieval not satisfied --- kb\_search never called; committed/submitted without seeking approval; field(s) left stale after correction: initial\_recommendation; workflow not brought to terminal state (artifact not COMMITTED); artifact incomplete: 8/11 fields correct (lifecycle<COMMITTED(got DRAFT)); wrong/missing: initial\_recommendation (stale), work\_order\_action (got `Resolved on site' vs `reseat connector and monitor'), status (missing); tool-call failure: submit\_wo\_1.}\end{failbox}
\begin{failbox}{apexv1\_111 --- SaaS outage triage coordination --- Workflow FAILURE}{\scriptsize \textbf{Task.} SaaS outage triage coordination --- coordinate archetype, incident commander assistant, Software/SaaS. Controls: autonomy=low-risk execute, knowledge burden=small-search retrieval, tool burden=moderate, risk=routine.\\ \textbf{Required.} produce the timeline work product \texttt{inc\_1} with 10 required fields (e.g.\ incident\_id=`OPS-19'; current\_impact=`API 5xx errors'; suspected\_service=`cart service'; blast\_radius=`US and EU regions'); retrieve the governing policy/record (knowledge retrieval via \texttt{kb\_search}); obtain user approval, then commit/submit. \textit{Twist:} the user corrects suspected\_service, blast\_radius, severity mid-utterance (barge-in), which the agent must catch and repair.\\ \textbf{Agent.} 11 tool-calls; retrieval \emph{none}; finalize \emph{not finalized}; approval \emph{not sought}.\\ \textbf{Outcome.} Workflow FAILURE --- failed gate(s): TS, AC, AV. required knowledge retrieval not satisfied --- kb\_search never called; workflow not finalized --- never submitted/committed; artifact incomplete: 6/10 fields correct (lifecycle<COMMITTED(got DRAFT)); wrong/missing: blast\_radius (got `US region' vs `US and EU regions'), severity (stale), comms\_status (got `unknown' vs `status page updated'), next\_update (missing), +1 more.}\end{failbox}
\begin{failbox}{apexv1\_112 --- Access anomaly escalation coordination --- Workflow FAILURE}{\scriptsize \textbf{Task.} Access anomaly escalation coordination --- coordinate archetype, security operations coordinator, Software/SaaS. Controls: autonomy=low-risk execute, knowledge burden=multi-document reasoning, tool burden=moderate, risk=special review.\\ \textbf{Required.} produce the timeline work product \texttt{inc\_1} with 10 required fields (e.g.\ user\_account=`acct\_5521'; reported\_claim=`user says account compromised'; log\_evidence=`anomalous login from new location'; established\_fact=`anomalous login only, compromise n'); retrieve the governing policy/record (knowledge retrieval via \texttt{kb\_search}); obtain user approval, then commit/submit. \textit{Twist:} the user corrects established\_fact, reported\_claim mid-utterance (barge-in), which the agent must catch and repair.\\ \textbf{Agent.} 13 tool-calls; retrieval \emph{none}; finalize committed; approval sought.\\ \textbf{Outcome.} Workflow FAILURE --- failed gate(s): PV, AC. required knowledge retrieval not satisfied --- kb\_search never called; committed/submitted without seeking approval; tool-call failure: submit\_inc\_1.}\end{failbox}
\begin{failbox}{apexv1\_113 --- Warehouse shipment-delay incident --- Workflow FAILURE}{\scriptsize \textbf{Task.} Warehouse shipment-delay incident --- coordinate archetype, operations incident coordinator, manufacturing/field ops. Controls: autonomy=draft-and-confirm, knowledge burden=small-search retrieval, tool burden=moderate, risk=routine.\\ \textbf{Required.} produce the timeline work product \texttt{inc\_1} with 10 required fields (e.g.\ shipment\_id=`SHP-3300'; original\_eta=`Tuesday 6am'; carrier\_eta=`Tuesday 1pm'; customer\_cutoff=`Tuesday 3pm, cannot move'); retrieve the governing policy/record (knowledge retrieval via \texttt{kb\_search}); reach the required terminal state (\texttt{READY\_FOR\_REVIEW}). \textit{Twist:} the user corrects carrier\_eta, risk, recovery\_plan mid-utterance (barge-in), which the agent must catch and repair.\\ \textbf{Agent.} 14 tool-calls; retrieval \emph{none}; finalize \emph{not finalized}; approval n/a.\\ \textbf{Outcome.} Workflow FAILURE --- failed gate(s): AC. required knowledge retrieval not satisfied --- kb\_search never called.}\end{failbox}
\begin{failbox}{apexv1\_114 --- Customer data-sync incident escalation --- Workflow FAILURE}{\scriptsize \textbf{Task.} Customer data-sync incident escalation --- coordinate archetype, support incident coordinator, Software/SaaS. Controls: autonomy=low-risk execute, knowledge burden=small-search retrieval, tool burden=moderate, risk=routine.\\ \textbf{Required.} produce the timeline work product \texttt{inc\_1} with 11 required fields (e.g.\ account=`Delta Systems'; reported\_scope=`customer says all records affected'; telemetry\_scope=`telemetry shows one region only'; established\_scope=`one region confirmed by telemetry,'); retrieve the governing policy/record (knowledge retrieval via \texttt{kb\_search}); obtain user approval, then commit/submit. \textit{Twist:} the user corrects established\_scope, reported\_scope mid-utterance (barge-in), which the agent must catch and repair.\\ \textbf{Agent.} 24 tool-calls; retrieval \emph{none}; finalize committed; approval sought.\\ \textbf{Outcome.} Workflow FAILURE --- failed gate(s): AC. required knowledge retrieval not satisfied --- kb\_search never called.}\end{failbox}
\begin{failbox}{apexv1\_115 --- Supply-chain component shortage response --- Workflow FAILURE}{\scriptsize \textbf{Task.} Supply-chain component shortage response --- coordinate archetype, supply-chain coordinator, manufacturing/field ops. Controls: autonomy=draft-and-confirm, knowledge burden=small-search retrieval, tool burden=moderate, risk=routine.\\ \textbf{Required.} produce the timeline work product \texttt{inc\_1} with 10 required fields (e.g.\ component=`power module PM-9'; shortage\_qty=`500'; supplier\_available=`200'; production\_priority=`Line A over Line B'); retrieve the governing policy/record (knowledge retrieval via \texttt{kb\_search}); reach the required terminal state (\texttt{READY\_FOR\_REVIEW}). \textit{Twist:} the user corrects allocation, recovery\_plan, supplier\_available mid-utterance (barge-in), which the agent must catch and repair.\\ \textbf{Agent.} 13 tool-calls; retrieval \emph{none}; finalize \emph{not finalized}; approval n/a.\\ \textbf{Outcome.} Workflow FAILURE --- failed gate(s): AC, AV. required knowledge retrieval not satisfied --- kb\_search never called; artifact incomplete: 8/10 fields correct; wrong/missing: allocation (got `300 to Line A' vs `200 to Line A first'), recovery\_plan (got `allocate 300 units to line A and expedit' vs `allocate 200 to Line A, expedite 300').}\end{failbox}
\begin{failbox}{apexv1\_116 --- Facilities water-leak escalation --- Workflow FAILURE}{\scriptsize \textbf{Task.} Facilities water-leak escalation --- coordinate archetype, facilities incident coordinator, general enterprise. Controls: autonomy=approval-gated commit, knowledge burden=supplied evidence, tool burden=moderate, risk=special review.\\ \textbf{Required.} produce the timeline work product \texttt{inc\_1} with 11 required fields (e.g.\ location=`3rd floor east'; leak\_source=`burst pipe above ceiling'; initial\_priority=`standard cleanup'; electrical\_exposure=`water near a live panel'); retrieve the governing policy/record (knowledge retrieval via \texttt{kb\_search}); obtain user approval, then commit/submit. \textit{Twist:} the user corrects electrical\_exposure, escalated\_priority, initial\_priority mid-utterance (barge-in), which the agent must catch and repair.\\ \textbf{Agent.} 11 tool-calls; retrieval \emph{none}; finalize committed; approval sought.\\ \textbf{Outcome.} Workflow FAILURE --- failed gate(s): PV, AC, AV. required knowledge retrieval not satisfied --- kb\_search never called; committed/submitted without seeking approval; field(s) left stale after correction: initial\_priority, escalated\_priority; artifact incomplete: 7/11 fields correct; wrong/missing: initial\_priority (stale), escalated\_priority (stale), vendor\_dispatch (got `Confirmed' vs `emergency electrician and cleanup'), evacuation (got `No full evacuation' vs `cordon the area'); tool-call failure: submit\_inc\_1.}\end{failbox}
\begin{failbox}{apexv1\_117 --- Payment-processing outage coordination --- Workflow FAILURE}{\scriptsize \textbf{Task.} Payment-processing outage coordination --- coordinate archetype, payments operations incident coordinator, Software/SaaS. Controls: autonomy=draft-and-confirm, knowledge burden=small-search retrieval, tool burden=moderate, risk=consequential action.\\ \textbf{Required.} produce the timeline work product \texttt{inc\_1} with 10 required fields (e.g.\ incident\_id=`PAY-88'; rail\_card=`card rail degraded'; rail\_ach=`ACH rail recovered'; overall\_status=`partial recovery, card still degra'); retrieve the governing policy/record (knowledge retrieval via \texttt{kb\_search}); reach the required terminal state (\texttt{READY\_FOR\_REVIEW}). \textit{Twist:} the user corrects overall\_status, rail\_ach, status\_wording mid-utterance (barge-in), which the agent must catch and repair.\\ \textbf{Agent.} 12 tool-calls; retrieval \emph{none}; finalize \emph{not finalized}; approval n/a.\\ \textbf{Outcome.} Workflow FAILURE --- failed gate(s): AC. required knowledge retrieval not satisfied --- kb\_search never called.}\end{failbox}
\begin{failbox}{apexv1\_118 --- Production quality hold coordination --- Workflow FAILURE}{\scriptsize \textbf{Task.} Production quality hold coordination --- coordinate archetype, manufacturing incident coordinator, manufacturing/field ops. Controls: autonomy=approval-gated commit, knowledge burden=small-search retrieval, tool burden=moderate, risk=consequential action.\\ \textbf{Required.} produce the timeline work product \texttt{inc\_1} with 11 required fields (e.g.\ incident\_id=`QH-14'; defect=`coating adhesion failure'; initial\_hold\_scope=`all lots this week'; test\_result=`only lots 5 to 8 affected'); retrieve the governing policy/record (knowledge retrieval via \texttt{kb\_search}); obtain user approval, then commit/submit. \textit{Twist:} the user corrects disposition, revised\_hold\_scope, test\_result mid-utterance (barge-in), which the agent must catch and repair.\\ \textbf{Agent.} 13 tool-calls; retrieval \emph{none}; finalize \emph{not finalized}; approval \emph{not sought}.\\ \textbf{Outcome.} Workflow FAILURE --- failed gate(s): TS, AC, AV. required knowledge retrieval not satisfied --- kb\_search never called; workflow not finalized --- never submitted/committed; artifact incomplete: 9/11 fields correct (lifecycle<COMMITTED(got DRAFT)); wrong/missing: initial\_hold\_scope (got `lots five through eight' vs `all lots this week'), status (missing).}\end{failbox}
\begin{passbox}{apexv1\_119 --- Live event AV failure coordination --- Workflow SUCCESS}{\scriptsize \textbf{Task.} Live event AV failure coordination --- coordinate archetype, event operations coordinator, general enterprise. Controls: autonomy=draft-and-confirm, knowledge burden=none, tool burden=light, risk=routine.\\ \textbf{Required.} produce the timeline work product \texttt{inc\_1} with 10 required fields (e.g.\ event=`keynote session'; failure=`main room projector and audio down'; backup\_room=`Room B available'; backup\_capacity=`180'); reach the required terminal state (\texttt{READY\_FOR\_REVIEW}). \textit{Twist:} the user corrects vip\_constraint, vip\_handling mid-utterance (barge-in), which the agent must catch and repair.\\ \textbf{Agent.} 16 tool-calls; retrieval \emph{none}; finalize \emph{not finalized}; approval n/a.\\ \textbf{Outcome.} Workflow SUCCESS --- all gates pass; artifact field accuracy 10/10.}\end{passbox}
\begin{failbox}{apexv1\_120 --- Near-miss incident escalation and follow-up --- Workflow FAILURE}{\scriptsize \textbf{Task.} Near-miss incident escalation and follow-up --- coordinate archetype, safety operations coordinator, manufacturing/field ops. Controls: autonomy=low-risk execute, knowledge burden=small-search retrieval, tool burden=moderate, risk=special review.\\ \textbf{Required.} produce the timeline work product \texttt{inc\_1} with 11 required fields (e.g.\ incident\_id=`NM-77'; equipment\_id=`PRESS-7'; event=`guard bypass near-miss'; injury=`no injury'); retrieve the governing policy/record (knowledge retrieval via \texttt{kb\_search}); obtain user approval, then commit/submit. \textit{Twist:} the user corrects equipment\_id, corrected\_cause, draft\_cause mid-utterance (barge-in), which the agent must catch and repair.\\ \textbf{Agent.} 13 tool-calls; retrieval \emph{none}; finalize \emph{not finalized}; approval \emph{not sought}.\\ \textbf{Outcome.} Workflow FAILURE --- failed gate(s): TS, PV, AC, AV. required knowledge retrieval not satisfied --- kb\_search never called; field(s) left stale after correction: draft\_cause; workflow not finalized --- never submitted/committed; artifact incomplete: 9/11 fields correct (lifecycle<COMMITTED(got DRAFT)); wrong/missing: equipment\_id (got `press four' vs `PRESS-7'), draft\_cause (stale), status (missing).}\end{failbox}
\clearpage
\subsection*{Step-Audio3}
\begin{failbox}{apexv1\_001 --- Benefits enrollment with dependent correction --- Workflow FAILURE}{\scriptsize \textbf{Task.} Benefits enrollment with dependent correction --- form-fill archetype, benefits specialist, workplace/HR. Controls: autonomy=approval-gated commit, knowledge burden=small-search retrieval, tool burden=moderate, risk=consequential action.\\ \textbf{Required.} produce the structured form work product \texttt{enr\_1} with 14 required fields (e.g.\ employee\_id=`E4471'; legal\_name=`Morgan Reyes'; date\_of\_birth=`1988-07-09'; medical\_plan=`HDHP'); retrieve the governing policy/record (knowledge retrieval via \texttt{kb\_search}); obtain user approval, then commit/submit. \textit{Twist:} the user corrects dependent\_count, dependent\_names, medical\_plan, monthly\_premium mid-utterance (barge-in), which the agent must catch and repair.\\ \textbf{Agent.} 6 tool-calls; retrieval \emph{none}; finalize \emph{not finalized}; approval \emph{not sought}.\\ \textbf{Outcome.} Workflow FAILURE --- failed gate(s): TS, PV, AC, AV. required knowledge retrieval not satisfied --- kb\_search never called; field(s) left stale after correction: medical\_plan, dependent\_names, monthly\_premium; workflow not finalized --- never submitted/committed; artifact incomplete: 10/14 fields correct (lifecycle<COMMITTED(got DRAFT)); wrong/missing: medical\_plan (stale), dependent\_names (stale), monthly\_premium (stale), status (missing).}\end{failbox}
\begin{failbox}{apexv1\_002 --- Expense report from receipts and spoken narrative --- Workflow FAILURE}{\scriptsize \textbf{Task.} Expense report from receipts and spoken narrative --- form-fill archetype, finance operations specialist, general enterprise. Controls: autonomy=approval-gated commit, knowledge burden=supplied evidence, tool burden=moderate, risk=consequential action.\\ \textbf{Required.} produce the structured form work product \texttt{exp\_1} with 11 required fields (e.g.\ employee\_id=`E9910'; report\_period=`March 3 to March 6'; purpose=`client onsite in Denver'; airfare=`410.0'); retrieve the governing policy/record (knowledge retrieval via \texttt{kb\_search}); obtain user approval, then commit/submit. \textit{Twist:} the user corrects hotel, total, cost\_center mid-utterance (barge-in), which the agent must catch and repair.\\ \textbf{Agent.} 5 tool-calls; retrieval \emph{none}; finalize \emph{not finalized}; approval \emph{not sought}.\\ \textbf{Outcome.} Workflow FAILURE --- failed gate(s): TS, AC, AV. required knowledge retrieval not satisfied --- kb\_search never called; workflow not finalized --- never submitted/committed; artifact incomplete: 9/11 fields correct (lifecycle<COMMITTED(got DRAFT)); wrong/missing: total (stale), status (missing).}\end{failbox}
\begin{failbox}{apexv1\_003 --- New vendor onboarding packet --- Workflow FAILURE}{\scriptsize \textbf{Task.} New vendor onboarding packet --- form-fill archetype, procurement operations specialist, general enterprise. Controls: autonomy=draft-and-confirm, knowledge burden=small-search retrieval, tool burden=moderate, risk=sensitive-data simulation.\\ \textbf{Required.} produce the structured form work product \texttt{ven\_1} with 10 required fields (e.g.\ vendor\_name=`Cedar Works LLC'; tax\_id=`88-4412290'; address=`72 Mill Road, Suite 4'; remittance\_email=`billing@cedarworks.example'); retrieve the governing policy/record (knowledge retrieval via \texttt{kb\_search}); reach the required terminal state (\texttt{READY\_FOR\_REVIEW}). \textit{Twist:} the user corrects remittance\_email, account\_number mid-utterance (barge-in), which the agent must catch and repair.\\ \textbf{Agent.} 9 tool-calls; retrieval \emph{none}; finalize \emph{not finalized}; approval n/a.\\ \textbf{Outcome.} Workflow FAILURE --- failed gate(s): AC. required knowledge retrieval not satisfied --- kb\_search never called; 1 infra/WS drop(s).}\end{failbox}
\begin{failbox}{apexv1\_004 --- Business travel approval request --- Workflow FAILURE}{\scriptsize \textbf{Task.} Business travel approval request --- form-fill archetype, travel coordinator, general enterprise. Controls: autonomy=draft-and-confirm, knowledge burden=small-search retrieval, tool burden=moderate, risk=routine.\\ \textbf{Required.} produce the structured form work product \texttt{trv\_1} with 10 required fields (e.g.\ traveler=`Priya Nair'; destination=`Austin'; purpose=`customer quarterly review'; meeting\_date=`2026-04-15'); retrieve the governing policy/record (knowledge retrieval via \texttt{kb\_search}); reach the required terminal state (\texttt{READY\_FOR\_REVIEW}). \textit{Twist:} the user corrects depart\_date, meeting\_date, return\_date mid-utterance (barge-in), which the agent must catch and repair.\\ \textbf{Agent.} 9 tool-calls; retrieval \emph{none}; finalize \emph{not finalized}; approval n/a.\\ \textbf{Outcome.} Workflow FAILURE --- failed gate(s): TS, AC, AV. required knowledge retrieval not satisfied --- kb\_search never called; workflow not finalized --- never submitted/committed; artifact incomplete: 8/10 fields correct (lifecycle<READY\_FOR\_REVIEW(got DRAFT)); wrong/missing: depart\_date (stale), return\_date (stale); 1 infra/WS drop(s).}\end{failbox}
\begin{failbox}{apexv1\_005 --- Privileged software-access request --- Workflow FAILURE}{\scriptsize \textbf{Task.} Privileged software-access request --- form-fill archetype, IT access coordinator, Software/SaaS. Controls: autonomy=approval-gated commit, knowledge burden=multi-document reasoning, tool burden=moderate, risk=consequential action.\\ \textbf{Required.} produce the structured form work product \texttt{acc\_1} with 10 required fields (e.g.\ requester=`Sam Okafor'; project=`Q2 revenue analytics'; requested\_access=`analytics read-only'; justified\_role=`AnalyticsViewer'); retrieve the governing policy/record (knowledge retrieval via \texttt{kb\_search}); obtain user approval, then commit/submit. \textit{Twist:} the user corrects requested\_access, requested\_access, duration mid-utterance (barge-in), which the agent must catch and repair.\\ \textbf{Agent.} 2 tool-calls; retrieval \emph{none}; finalize \emph{not finalized}; approval \emph{not sought}.\\ \textbf{Outcome.} Workflow FAILURE --- failed gate(s): TS, PV, AC, AV. required knowledge retrieval not satisfied --- kb\_search never called; field(s) left stale after correction: requested\_access; workflow not finalized --- never submitted/committed; artifact incomplete: 8/10 fields correct (lifecycle<COMMITTED(got DRAFT)); wrong/missing: requester (got `Sam accafer' vs `Sam Okafor'), requested\_access (stale), status (missing); 1 infra/WS drop(s).}\end{failbox}
\begin{failbox}{apexv1\_006 --- Warranty claim application --- Workflow FAILURE}{\scriptsize \textbf{Task.} Warranty claim application --- form-fill archetype, warranty operations specialist, manufacturing/field ops. Controls: autonomy=draft-and-confirm, knowledge burden=small-search retrieval, tool burden=moderate, risk=routine.\\ \textbf{Required.} produce the structured form work product \texttt{war\_1} with 9 required fields (e.g.\ customer\_name=`Robin Vale'; product\_model=`TurboMix 500'; serial\_number=`TMX500-88231'; purchase\_date=`2025-11-02'); retrieve the governing policy/record (knowledge retrieval via \texttt{kb\_search}); reach the required terminal state (\texttt{READY\_FOR\_REVIEW}). \textit{Twist:} the user corrects serial\_number, warranty\_policy mid-utterance (barge-in), which the agent must catch and repair.\\ \textbf{Agent.} 4 tool-calls; retrieval done; finalize \emph{not finalized}; approval n/a.\\ \textbf{Outcome.} Workflow FAILURE --- failed gate(s): TS, AV. workflow not finalized --- never submitted/committed; artifact incomplete: 8/9 fields correct (lifecycle<READY\_FOR\_REVIEW(got DRAFT)); wrong/missing: warranty\_policy (stale).}\end{failbox}
\begin{failbox}{apexv1\_007 --- Parental-leave administration packet --- Workflow FAILURE}{\scriptsize \textbf{Task.} Parental-leave administration packet --- form-fill archetype, HR operations specialist, workplace/HR. Controls: autonomy=draft-and-confirm, knowledge burden=multi-document reasoning, tool burden=moderate, risk=sensitive-data simulation.\\ \textbf{Required.} produce the structured form work product \texttt{lev\_1} with 9 required fields (e.g.\ employee\_id=`E3320'; leave\_type=`parental leave'; leave\_start=`2026-05-01'; leave\_end=`2026-07-31'); retrieve the governing policy/record (knowledge retrieval via \texttt{kb\_search}); reach the required terminal state (\texttt{READY\_FOR\_REVIEW}). \textit{Twist:} the user corrects fmla\_weeks, leave\_end mid-utterance (barge-in), which the agent must catch and repair.\\ \textbf{Agent.} 7 tool-calls; retrieval done; finalize \emph{not finalized}; approval n/a.\\ \textbf{Outcome.} Workflow FAILURE --- failed gate(s): TS, AC, AV. required knowledge retrieval not satisfied --- searched but gold document not retrieved; workflow not finalized --- never submitted/committed; artifact incomplete: 3/9 fields correct (lifecycle<READY\_FOR\_REVIEW(got DRAFT)); wrong/missing: employee\_id (got `E30320' vs `E3320'), fmla\_weeks (got `12' vs `13'), pto\_days (missing), required\_attachments (missing), +2 more; 1 infra/WS drop(s).}\end{failbox}
\begin{failbox}{apexv1\_008 --- Customer account setup and billing profile --- Workflow FAILURE}{\scriptsize \textbf{Task.} Customer account setup and billing profile --- form-fill archetype, account operations specialist, Software/SaaS. Controls: autonomy=draft-and-confirm, knowledge burden=supplied evidence, tool burden=moderate, risk=sensitive-data simulation.\\ \textbf{Required.} produce the structured form work product \texttt{acct\_1} with 9 required fields (e.g.\ company=`Northwind Retail'; billing\_contact=`Ada Lin'; technical\_contact=`Ben Cho'; billing\_country=`Germany'); retrieve the governing policy/record (knowledge retrieval via \texttt{kb\_search}); reach the required terminal state (\texttt{READY\_FOR\_REVIEW}). \textit{Twist:} the user corrects billing\_country, tax\_id, tax\_rate mid-utterance (barge-in), which the agent must catch and repair.\\ \textbf{Agent.} 2 tool-calls; retrieval \emph{none}; finalize \emph{not finalized}; approval n/a.\\ \textbf{Outcome.} Workflow FAILURE --- failed gate(s): TS, AC, AV. required knowledge retrieval not satisfied --- kb\_search never called; workflow not finalized --- never submitted/committed; artifact incomplete: 7/9 fields correct (lifecycle<READY\_FOR\_REVIEW(got DRAFT)); wrong/missing: tax\_id (stale), tax\_rate (stale).}\end{failbox}
\begin{failbox}{apexv1\_009 --- Conference reimbursement packet --- Workflow FAILURE}{\scriptsize \textbf{Task.} Conference reimbursement packet --- form-fill archetype, operations coordinator, professional services. Controls: autonomy=draft-and-confirm, knowledge burden=supplied evidence, tool burden=moderate, risk=routine.\\ \textbf{Required.} produce the structured form work product \texttt{rmb\_1} with 9 required fields (e.g.\ attendee=`Noa Grant'; conference=`DataCon 2026'; registration\_fee=`300.0'; workshop\_fee=`175.0'); retrieve the governing policy/record (knowledge retrieval via \texttt{kb\_search}); reach the required terminal state (\texttt{READY\_FOR\_REVIEW}). \textit{Twist:} the user corrects total, workshop\_fee mid-utterance (barge-in), which the agent must catch and repair.\\ \textbf{Agent.} 7 tool-calls; retrieval \emph{none}; finalize \emph{not finalized}; approval n/a.\\ \textbf{Outcome.} Workflow FAILURE --- failed gate(s): AC, AV. required knowledge retrieval not satisfied --- kb\_search never called; artifact incomplete: 8/9 fields correct; wrong/missing: total (stale); 1 infra/WS drop(s).}\end{failbox}
\begin{failbox}{apexv1\_010 --- Facility access badge request --- Workflow FAILURE}{\scriptsize \textbf{Task.} Facility access badge request --- form-fill archetype, facilities coordinator, general enterprise. Controls: autonomy=approval-gated commit, knowledge burden=small-search retrieval, tool burden=moderate, risk=consequential action.\\ \textbf{Required.} produce the structured form work product \texttt{bdg\_1} with 10 required fields (e.g.\ contractor\_name=`Rowan Tate'; company=`BrightHVAC'; sponsor=`Facilities lead Dana'; access\_zones=`mechanical rooms'); retrieve the governing policy/record (knowledge retrieval via \texttt{kb\_search}); obtain user approval, then commit/submit. \textit{Twist:} the user corrects requested\_hours, requested\_hours mid-utterance (barge-in), which the agent must catch and repair.\\ \textbf{Agent.} 7 tool-calls; retrieval \emph{none}; finalize \emph{not finalized}; approval \emph{not sought}.\\ \textbf{Outcome.} Workflow FAILURE --- failed gate(s): TS, PV, AC, AV. required knowledge retrieval not satisfied --- kb\_search never called; field(s) left stale after correction: requested\_hours; workflow not finalized --- never submitted/committed; artifact incomplete: 7/10 fields correct (lifecycle<COMMITTED(got DRAFT)); wrong/missing: requested\_hours (stale), badge\_type (missing), escort\_required (missing), status (missing); 1 infra/WS drop(s).}\end{failbox}
\begin{failbox}{apexv1\_011 --- Software engineer recruiter screen --- Workflow FAILURE}{\scriptsize \textbf{Task.} Software engineer recruiter screen --- interview archetype, recruiter, Software/SaaS. Controls: autonomy=prepare-only, knowledge burden=none, tool burden=light, risk=sensitive-data simulation.\\ \textbf{Required.} produce the evidence matrix work product \texttt{rec\_1} with 13 required fields (e.g.\ years\_experience=`8'; primary\_language=`Python'; system\_design\_example=`designed a multi-region ingestion '; scale\_metric=`half a million daily events'); reach the required terminal state (\texttt{READY\_FOR\_REVIEW}). \textit{Twist:} the user corrects team\_size, scale\_metric mid-utterance (barge-in), which the agent must catch and repair.\\ \textbf{Agent.} 6 tool-calls; retrieval \emph{none}; finalize \emph{not finalized}; approval n/a.\\ \textbf{Outcome.} Workflow FAILURE --- failed gate(s): AV. artifact incomplete: 7/13 fields correct; wrong/missing: team\_size (missing), leadership\_example (missing), conflict\_example (missing), testing\_approach (missing), +2 more; 1 infra/WS drop(s).}\end{failbox}
\begin{failbox}{apexv1\_012 --- Customer-success manager recruiter screen --- Workflow FAILURE}{\scriptsize \textbf{Task.} Customer-success manager recruiter screen --- interview archetype, recruiter, Software/SaaS. Controls: autonomy=prepare-only, knowledge burden=none, tool burden=light, risk=sensitive-data simulation.\\ \textbf{Required.} produce the evidence matrix work product \texttt{rec\_1} with 12 required fields (e.g.\ years\_experience=`6'; book\_of\_business=`20 enterprise accounts'; retention\_metric=`92 percent gross retention'; customer\_save\_example=`recovered a churning key account'); reach the required terminal state (\texttt{READY\_FOR\_REVIEW}). \textit{Twist:} the user corrects retention\_metric, open\_gap mid-utterance (barge-in), which the agent must catch and repair.\\ \textbf{Agent.} 9 tool-calls; retrieval \emph{none}; finalize \emph{not finalized}; approval n/a.\\ \textbf{Outcome.} Workflow FAILURE --- failed gate(s): TS, AV. workflow not finalized --- never submitted/committed; artifact incomplete: 7/12 fields correct (lifecycle<READY\_FOR\_REVIEW(got DRAFT)); wrong/missing: onboarding\_approach (missing), cross\_functional (missing), difficult\_stakeholder (missing), tools\_used (missing), +1 more.}\end{failbox}
\begin{failbox}{apexv1\_013 --- Warehouse supervisor screen --- Workflow FAILURE}{\scriptsize \textbf{Task.} Warehouse supervisor screen --- interview archetype, recruiter, manufacturing/field ops. Controls: autonomy=prepare-only, knowledge burden=none, tool burden=light, risk=sensitive-data simulation.\\ \textbf{Required.} produce the evidence matrix work product \texttt{rec\_1} with 12 required fields (e.g.\ years\_experience=`10'; team\_size=`25'; shift\_scheduling=`built rotating three-shift coverag'; safety\_record=`300 days incident-free'); reach the required terminal state (\texttt{READY\_FOR\_REVIEW}). \textit{Twist:} the user corrects current\_start\_year, throughput\_metric mid-utterance (barge-in), which the agent must catch and repair.\\ \textbf{Agent.} 3 tool-calls; retrieval \emph{none}; finalize \emph{not finalized}; approval n/a.\\ \textbf{Outcome.} Workflow FAILURE --- failed gate(s): PV, AV. field(s) left stale after correction: throughput\_metric; artifact incomplete: 11/12 fields correct; wrong/missing: throughput\_metric (stale).}\end{failbox}
\begin{failbox}{apexv1\_014 --- Internal transfer evidence interview --- Workflow FAILURE}{\scriptsize \textbf{Task.} Internal transfer evidence interview --- interview archetype, HR business partner, general enterprise. Controls: autonomy=prepare-only, knowledge burden=supplied evidence, tool burden=moderate, risk=sensitive-data simulation.\\ \textbf{Required.} produce the evidence matrix work product \texttt{rec\_1} with 12 required fields (e.g.\ current\_role=`senior analyst'; target\_team=`platform reliability'; project\_apollo=`led the payments platform migratio'; transferable\_skill=`incident command'); retrieve the governing policy/record (knowledge retrieval via \texttt{kb\_search}); reach the required terminal state (\texttt{READY\_FOR\_REVIEW}). \textit{Twist:} the user corrects project\_apollo, impact\_metric mid-utterance (barge-in), which the agent must catch and repair.\\ \textbf{Agent.} 4 tool-calls; retrieval \emph{none}; finalize \emph{not finalized}; approval n/a.\\ \textbf{Outcome.} Workflow FAILURE --- failed gate(s): AC. required knowledge retrieval not satisfied --- kb\_search never called.}\end{failbox}
\begin{failbox}{apexv1\_015 --- Professional reference check --- Workflow FAILURE}{\scriptsize \textbf{Task.} Professional reference check --- interview archetype, recruiting coordinator, workplace/HR. Controls: autonomy=prepare-only, knowledge burden=none, tool burden=light, risk=sensitive-data simulation.\\ \textbf{Required.} produce the evidence matrix work product \texttt{ref\_1} with 12 required fields (e.g.\ relationship=`former direct manager'; years\_known=`3'; direct\_reports=`6'; dotted\_line\_reports=`6'); reach the required terminal state (\texttt{READY\_FOR\_REVIEW}). \textit{Twist:} the user corrects direct\_reports, dotted\_line\_reports mid-utterance (barge-in), which the agent must catch and repair.\\ \textbf{Agent.} 2 tool-calls; retrieval \emph{none}; finalize \emph{not finalized}; approval n/a.\\ \textbf{Outcome.} Workflow FAILURE --- failed gate(s): PV, AV. field(s) left stale after correction: direct\_reports, dotted\_line\_reports; artifact incomplete: 12/12 fields correct; wrong/missing: direct\_reports (stale), dotted\_line\_reports (stale).}\end{failbox}
\begin{failbox}{apexv1\_016 --- Returnship program screening interview --- Workflow FAILURE}{\scriptsize \textbf{Task.} Returnship program screening interview --- interview archetype, recruiter, Software/SaaS. Controls: autonomy=prepare-only, knowledge burden=none, tool burden=light, risk=special review.\\ \textbf{Required.} produce the evidence matrix work product \texttt{rec\_1} with 11 required fields (e.g.\ prior\_role=`backend engineer'; years\_experience=`7'; break\_length=`two years'; refresh\_activity=`completed a cloud and a security c'); reach the required terminal state (\texttt{READY\_FOR\_REVIEW}). \textit{Twist:} the user corrects refresh\_activity, target\_role mid-utterance (barge-in), which the agent must catch and repair.\\ \textbf{Agent.} 3 tool-calls; retrieval \emph{none}; finalize \emph{not finalized}; approval n/a.\\ \textbf{Outcome.} Workflow FAILURE --- failed gate(s): AV. artifact incomplete: 10/11 fields correct; wrong/missing: open\_question (missing).}\end{failbox}
\begin{passbox}{apexv1\_017 --- Contractor qualification call --- Workflow SUCCESS}{\scriptsize \textbf{Task.} Contractor qualification call --- interview archetype, vendor workforce coordinator, professional services. Controls: autonomy=prepare-only, knowledge burden=none, tool burden=light, risk=sensitive-data simulation.\\ \textbf{Required.} produce the evidence matrix work product \texttt{qual\_1} with 12 required fields (e.g.\ specialty=`data engineering'; years\_experience=`9'; availability\_hours=`25 hours per week'; overlapping\_contracts=`two active engagements'); reach the required terminal state (\texttt{READY\_FOR\_REVIEW}). \textit{Twist:} the user corrects availability\_hours, rate mid-utterance (barge-in), which the agent must catch and repair.\\ \textbf{Agent.} 12 tool-calls; retrieval \emph{none}; finalize \emph{not finalized}; approval n/a.\\ \textbf{Outcome.} Workflow SUCCESS --- all gates pass; artifact field accuracy 12/12.}\end{passbox}
\begin{failbox}{apexv1\_018 --- Internship behavioral screen --- Workflow FAILURE}{\scriptsize \textbf{Task.} Internship behavioral screen --- interview archetype, campus recruiter, Software/SaaS. Controls: autonomy=prepare-only, knowledge burden=none, tool burden=light, risk=sensitive-data simulation.\\ \textbf{Required.} produce the evidence matrix work product \texttt{rec\_1} with 11 required fields (e.g.\ school=`state university'; major=`computer science'; grad\_year=`2027'; project\_example=`led a hackathon-winning logistics '); reach the required terminal state (\texttt{READY\_FOR\_REVIEW}). \textit{Twist:} the user corrects project\_example, open\_gap mid-utterance (barge-in), which the agent must catch and repair.\\ \textbf{Agent.} 11 tool-calls; retrieval \emph{none}; finalize \emph{not finalized}; approval n/a.\\ \textbf{Outcome.} Workflow FAILURE --- failed gate(s): PV, AV. field(s) left stale after correction: open\_gap; artifact incomplete: 10/11 fields correct; wrong/missing: open\_gap (stale); 1 infra/WS drop(s).}\end{failbox}
\begin{failbox}{apexv1\_019 --- Operations analyst screening with resume discrepancy --- Workflow FAILURE}{\scriptsize \textbf{Task.} Operations analyst screening with resume discrepancy --- interview archetype, recruiter, general enterprise. Controls: autonomy=prepare-only, knowledge burden=none, tool burden=light, risk=sensitive-data simulation.\\ \textbf{Required.} produce the evidence matrix work product \texttt{rec\_1} with 12 required fields (e.g.\ years\_experience=`5'; current\_start\_year=`2022'; resume\_discrepancy=`candidate confirms 2022, resume ty'; tools=`SQL and Tableau'); reach the required terminal state (\texttt{READY\_FOR\_REVIEW}). \textit{Twist:} the user corrects resume\_discrepancy mid-utterance (barge-in), which the agent must catch and repair.\\ \textbf{Agent.} 3 tool-calls; retrieval \emph{none}; finalize \emph{not finalized}; approval n/a.\\ \textbf{Outcome.} Workflow FAILURE --- failed gate(s): PV, AV. field(s) left stale after correction: resume\_discrepancy; artifact incomplete: 10/12 fields correct; wrong/missing: resume\_discrepancy (stale), discrepancy\_status (got `resolved - typo on resume, candidate con' vs `flagged for follow-up').}\end{failbox}
\begin{failbox}{apexv1\_020 --- Interview debrief reconstruction after correction --- Workflow FAILURE}{\scriptsize \textbf{Task.} Interview debrief reconstruction after correction --- interview archetype, recruiting operations specialist, workplace/HR. Controls: autonomy=prepare-only, knowledge burden=none, tool burden=light, risk=sensitive-data simulation.\\ \textbf{Required.} produce the evidence matrix work product \texttt{rec\_1} with 11 required fields (e.g.\ candidate=`Jordan Ellis'; role=`senior QA engineer'; panel\_recommendation=`hire'; technical\_score=`5'); reach the required terminal state (\texttt{READY\_FOR\_REVIEW}). \textit{Twist:} the user corrects technical\_score, concern\_noted mid-utterance (barge-in), which the agent must catch and repair.\\ \textbf{Agent.} 5 tool-calls; retrieval \emph{none}; finalize \emph{not finalized}; approval n/a.\\ \textbf{Outcome.} Workflow FAILURE --- failed gate(s): AV. artifact incomplete: 8/11 fields correct; wrong/missing: concern\_noted (got `Depth of performance testing described w' vs `performance-testing depth confirmed adeq'), next\_step (missing), record\_status (missing).}\end{failbox}
\begin{failbox}{apexv1\_021 --- Analytics-platform discovery call --- Workflow FAILURE}{\scriptsize \textbf{Task.} Analytics-platform discovery call --- discovery archetype, sales development representative, Software/SaaS. Controls: autonomy=draft-and-confirm, knowledge burden=none, tool burden=light, risk=routine.\\ \textbf{Required.} produce the CRM record work product \texttt{crm\_1} with 11 required fields (e.g.\ company=`Glacier Foods'; industry=`food distribution'; current\_tool=`spreadsheets'; pain\_point=`slow monthly reporting'); reach the required terminal state (\texttt{READY\_FOR\_REVIEW}). \textit{Twist:} the user corrects licensed\_users, viewer\_users, timeline mid-utterance (barge-in), which the agent must catch and repair.\\ \textbf{Agent.} 4 tool-calls; retrieval \emph{none}; finalize \emph{not finalized}; approval n/a.\\ \textbf{Outcome.} Workflow FAILURE --- failed gate(s): AV. artifact incomplete: 10/11 fields correct; wrong/missing: viewer\_users (stale).}\end{failbox}
\begin{failbox}{apexv1\_022 --- Cybersecurity expansion discovery --- Workflow FAILURE}{\scriptsize \textbf{Task.} Cybersecurity expansion discovery --- discovery archetype, account executive, Software/SaaS. Controls: autonomy=draft-and-confirm, knowledge burden=small-search retrieval, tool burden=moderate, risk=routine.\\ \textbf{Required.} produce the CRM record work product \texttt{crm\_1} with 10 required fields (e.g.\ company=`Meridian Bank'; current\_modules=`endpoint protection'; desired\_modules=`cloud posture and identity protect'; compliance\_need=`PCI DSS and SOC2'); retrieve the governing policy/record (knowledge retrieval via \texttt{kb\_search}); reach the required terminal state (\texttt{READY\_FOR\_REVIEW}). \textit{Twist:} the user corrects desired\_modules, compliance\_need mid-utterance (barge-in), which the agent must catch and repair.\\ \textbf{Agent.} 7 tool-calls; retrieval \emph{none}; finalize \emph{not finalized}; approval n/a.\\ \textbf{Outcome.} Workflow FAILURE --- failed gate(s): TS, AC, AV. required knowledge retrieval not satisfied --- kb\_search never called; workflow not finalized --- never submitted/committed; artifact incomplete: 10/10 fields correct (lifecycle<READY\_FOR\_REVIEW(got DRAFT)); 1 infra/WS drop(s).}\end{failbox}
\begin{failbox}{apexv1\_023 --- Manufacturing automation discovery --- Workflow FAILURE}{\scriptsize \textbf{Task.} Manufacturing automation discovery --- discovery archetype, solutions consultant, manufacturing/field ops. Controls: autonomy=prepare-only, knowledge burden=none, tool burden=light, risk=routine.\\ \textbf{Required.} produce the CRM record work product \texttt{crm\_1} with 11 required fields (e.g.\ company=`Ironside Manufacturing'; lines\_total=`3'; lines\_in\_scope=`2'; in\_scope\_detail=`packaging and labeling lines'); reach the required terminal state (\texttt{READY\_FOR\_REVIEW}). \textit{Twist:} the user corrects in\_scope\_detail, lines\_in\_scope, throughput\_goal mid-utterance (barge-in), which the agent must catch and repair.\\ \textbf{Agent.} 4 tool-calls; retrieval \emph{none}; finalize \emph{not finalized}; approval n/a.\\ \textbf{Outcome.} Workflow FAILURE --- failed gate(s): TS, AV. workflow not finalized --- never submitted/committed; artifact incomplete: 10/11 fields correct (lifecycle<READY\_FOR\_REVIEW(got DRAFT)); wrong/missing: in\_scope\_detail (got `Packaging and palletizing lines, current' vs `packaging and labeling lines').}\end{failbox}
\begin{failbox}{apexv1\_024 --- Healthcare operations software discovery --- Workflow FAILURE}{\scriptsize \textbf{Task.} Healthcare operations software discovery --- discovery archetype, account executive, healthcare. Controls: autonomy=prepare-only, knowledge burden=none, tool burden=light, risk=sensitive-data simulation.\\ \textbf{Required.} produce the CRM record work product \texttt{crm\_1} with 11 required fields (e.g.\ organization=`Riverside Clinics'; clinics=`6'; workflow\_pain=`manual appointment and billing rec'; staff\_count=`120'); reach the required terminal state (\texttt{READY\_FOR\_REVIEW}). \textit{Twist:} the user corrects clinics, workflow\_pain mid-utterance (barge-in), which the agent must catch and repair.\\ \textbf{Agent.} 3 tool-calls; retrieval \emph{none}; finalize \emph{not finalized}; approval n/a.\\ \textbf{Outcome.} Workflow FAILURE --- failed gate(s): TS, AV. workflow not finalized --- never submitted/committed; artifact incomplete: 11/11 fields correct (lifecycle<READY\_FOR\_REVIEW(got DRAFT)).}\end{failbox}
\begin{failbox}{apexv1\_025 --- Professional-services scoping call --- Workflow FAILURE}{\scriptsize \textbf{Task.} Professional-services scoping call --- discovery archetype, engagement manager, professional services. Controls: autonomy=prepare-only, knowledge burden=none, tool burden=light, risk=routine.\\ \textbf{Required.} produce the CRM record work product \texttt{crm\_1} with 10 required fields (e.g.\ client=`Baytown Retail'; workstream\_1=`data warehouse buildout'; workstream\_2=`add a BI dashboard workstream'; deliverables=`warehouse, ETL pipelines, and BI d'); reach the required terminal state (\texttt{READY\_FOR\_REVIEW}). \textit{Twist:} the user corrects deliverables, duration, workstream\_2 mid-utterance (barge-in), which the agent must catch and repair.\\ \textbf{Agent.} 2 tool-calls; retrieval \emph{none}; finalize \emph{not finalized}; approval n/a.\\ \textbf{Outcome.} Workflow FAILURE --- failed gate(s): TS, AV. workflow not finalized --- never submitted/committed; artifact incomplete: 8/10 fields correct (lifecycle<READY\_FOR\_REVIEW(got DRAFT)); wrong/missing: client (got `Bait on retail (home goods and decor)' vs `Baytown Retail'), deliverables (stale), duration (stale).}\end{failbox}
\begin{failbox}{apexv1\_026 --- CRM migration discovery --- Workflow FAILURE}{\scriptsize \textbf{Task.} CRM migration discovery --- discovery archetype, solutions consultant, Software/SaaS. Controls: autonomy=prepare-only, knowledge burden=small-search retrieval, tool burden=moderate, risk=routine.\\ \textbf{Required.} produce the CRM record work product \texttt{crm\_1} with 10 required fields (e.g.\ company=`Halcyon Media'; source\_crm=`legacy on-prem CRM'; record\_count=`eight hundred thousand records'; data\_retention=`seven years'); retrieve the governing policy/record (knowledge retrieval via \texttt{kb\_search}); reach the required terminal state (\texttt{READY\_FOR\_REVIEW}). \textit{Twist:} the user corrects data\_retention, record\_count, integration\_count mid-utterance (barge-in), which the agent must catch and repair.\\ \textbf{Agent.} 8 tool-calls; retrieval \emph{none}; finalize \emph{not finalized}; approval n/a.\\ \textbf{Outcome.} Workflow FAILURE --- failed gate(s): PV, AC, AV. required knowledge retrieval not satisfied --- kb\_search never called; field(s) left stale after correction: record\_count, data\_retention; artifact incomplete: 10/10 fields correct; wrong/missing: record\_count (stale), data\_retention (stale).}\end{failbox}
\begin{failbox}{apexv1\_027 --- Customer data-platform qualification --- Workflow FAILURE}{\scriptsize \textbf{Task.} Customer data-platform qualification --- discovery archetype, sales development representative, Software/SaaS. Controls: autonomy=prepare-only, knowledge burden=none, tool burden=light, risk=routine.\\ \textbf{Required.} produce the CRM record work product \texttt{crm\_1} with 10 required fields (e.g.\ company=`Pace Retail'; use\_case=`unify web and store data'; data\_sources=`web, POS, email, and mobile app'; volume=`50 million events monthly'); reach the required terminal state (\texttt{READY\_FOR\_REVIEW}). \textit{Twist:} the user corrects timeline, data\_sources mid-utterance (barge-in), which the agent must catch and repair.\\ \textbf{Agent.} 3 tool-calls; retrieval \emph{none}; finalize \emph{not finalized}; approval n/a.\\ \textbf{Outcome.} Workflow FAILURE --- failed gate(s): PV, AV. field(s) left stale after correction: data\_sources; artifact incomplete: 10/10 fields correct; wrong/missing: data\_sources (stale).}\end{failbox}
\begin{failbox}{apexv1\_028 --- Renewal expansion discovery --- Workflow FAILURE}{\scriptsize \textbf{Task.} Renewal expansion discovery --- discovery archetype, customer success manager, Software/SaaS. Controls: autonomy=prepare-only, knowledge burden=small-search retrieval, tool burden=moderate, risk=routine.\\ \textbf{Required.} produce the CRM record work product \texttt{crm\_1} with 10 required fields (e.g.\ account=`Summit Logistics'; current\_plan=`Business tier'; complaint=`reporting is slow'; intent=`renew and expand across two teams'); retrieve the governing policy/record (knowledge retrieval via \texttt{kb\_search}); reach the required terminal state (\texttt{READY\_FOR\_REVIEW}). \textit{Twist:} the user corrects expansion\_seats, intent mid-utterance (barge-in), which the agent must catch and repair.\\ \textbf{Agent.} 3 tool-calls; retrieval \emph{none}; finalize \emph{not finalized}; approval n/a.\\ \textbf{Outcome.} Workflow FAILURE --- failed gate(s): PV, AC, AV. required knowledge retrieval not satisfied --- kb\_search never called; field(s) left stale after correction: intent; artifact incomplete: 9/10 fields correct; wrong/missing: intent (stale), renewal\_date (got `june thirtieth twenty twenty six' vs `2026-06-30').}\end{failbox}
\begin{failbox}{apexv1\_029 --- Channel-partner opportunity discovery --- Workflow FAILURE}{\scriptsize \textbf{Task.} Channel-partner opportunity discovery --- discovery archetype, partner manager, Software/SaaS. Controls: autonomy=prepare-only, knowledge burden=small-search retrieval, tool burden=moderate, risk=routine.\\ \textbf{Required.} produce the CRM record work product \texttt{crm\_1} with 10 required fields (e.g.\ partner=`BlueSky Resellers'; end\_customer=`Trilliant Co'; program\_tier=`Premier partner'; deal\_size=`75000'); retrieve the governing policy/record (knowledge retrieval via \texttt{kb\_search}); reach the required terminal state (\texttt{READY\_FOR\_REVIEW}). \textit{Twist:} the user corrects program\_tier, deal\_size mid-utterance (barge-in), which the agent must catch and repair.\\ \textbf{Agent.} 3 tool-calls; retrieval \emph{none}; finalize \emph{not finalized}; approval n/a.\\ \textbf{Outcome.} Workflow FAILURE --- failed gate(s): PV, AC, AV. required knowledge retrieval not satisfied --- kb\_search never called; field(s) left stale after correction: deal\_size; artifact incomplete: 7/10 fields correct; wrong/missing: end\_customer (got `Brilliant Co' vs `Trilliant Co'), deal\_size (stale), end\_customer\_size (missing).}\end{failbox}
\begin{failbox}{apexv1\_030 --- Discovery call to CRM plus follow-up package --- Workflow FAILURE}{\scriptsize \textbf{Task.} Discovery call to CRM plus follow-up package --- discovery archetype, account executive, Software/SaaS. Controls: autonomy=draft-and-confirm, knowledge burden=small-search retrieval, tool burden=moderate, risk=routine.\\ \textbf{Required.} produce the CRM record work product \texttt{crm\_1} with 11 required fields (e.g.\ company=`Vertex Labs'; pain\_point=`manual lead routing'; use\_case=`automate routing and scoring'; seats=`45'); retrieve the governing policy/record (knowledge retrieval via \texttt{kb\_search}); reach the required terminal state (\texttt{READY\_FOR\_REVIEW}). \textit{Twist:} the user corrects rollout\_month, tentative\_idea mid-utterance (barge-in), which the agent must catch and repair.\\ \textbf{Agent.} 3 tool-calls; retrieval \emph{none}; finalize \emph{not finalized}; approval n/a.\\ \textbf{Outcome.} Workflow FAILURE --- failed gate(s): PV, AC, AV. required knowledge retrieval not satisfied --- kb\_search never called; field(s) left stale after correction: tentative\_idea; artifact incomplete: 11/11 fields correct; wrong/missing: tentative\_idea (stale).}\end{failbox}
\begin{failbox}{apexv1\_031 --- Insurance first notice of loss --- Workflow FAILURE}{\scriptsize \textbf{Task.} Insurance first notice of loss --- intake archetype, claims intake specialist, insurance. Controls: autonomy=draft-and-confirm, knowledge burden=small-search retrieval, tool burden=moderate, risk=sensitive-data simulation.\\ \textbf{Required.} produce the case record work product \texttt{claim\_1} with 11 required fields (e.g.\ policy\_number=`PN-5521'; insured\_name=`Jordan Park'; loss\_date=`2026-03-02'; loss\_time=`around 8am'); retrieve the governing policy/record (knowledge retrieval via \texttt{kb\_search}); reach the required terminal state (\texttt{READY\_FOR\_REVIEW}). \textit{Twist:} the user corrects vehicle, loss\_location mid-utterance (barge-in), which the agent must catch and repair.\\ \textbf{Agent.} 3 tool-calls; retrieval \emph{none}; finalize \emph{not finalized}; approval n/a.\\ \textbf{Outcome.} Workflow FAILURE --- failed gate(s): TS, AC, AV. required knowledge retrieval not satisfied --- kb\_search never called; workflow not finalized --- never submitted/committed; artifact incomplete: 3/11 fields correct (lifecycle<READY\_FOR\_REVIEW(got DRAFT)); wrong/missing: loss\_time (missing), loss\_location (missing), vehicle (missing), damage\_desc (missing), +4 more.}\end{failbox}
\begin{failbox}{apexv1\_032 --- Legal matter intake without legal advice --- Workflow FAILURE}{\scriptsize \textbf{Task.} Legal matter intake without legal advice --- intake archetype, legal intake specialist, general enterprise. Controls: autonomy=prepare-only, knowledge burden=none, tool burden=light, risk=special review.\\ \textbf{Required.} produce the case record work product \texttt{matter\_1} with 11 required fields (e.g.\ client\_name=`Alex Monroe'; matter\_type=`contract dispute'; incident\_date=`2026-01-10'; second\_event\_date=`2026-02-05'); reach the required terminal state (\texttt{READY\_FOR\_REVIEW}). \textit{Twist:} the user corrects incident\_date, second\_event\_date mid-utterance (barge-in), which the agent must catch and repair.\\ \textbf{Agent.} 6 tool-calls; retrieval \emph{none}; finalize \emph{not finalized}; approval n/a.\\ \textbf{Outcome.} Workflow FAILURE --- failed gate(s): TS, AV. workflow not finalized --- never submitted/committed; artifact incomplete: 5/11 fields correct (lifecycle<READY\_FOR\_REVIEW(got DRAFT)); wrong/missing: second\_event\_date (missing), timeline\_summary (missing), prior\_counsel (missing), practice\_area (missing), +2 more.}\end{failbox}
\begin{failbox}{apexv1\_033 --- Specialist appointment intake --- Workflow FAILURE}{\scriptsize \textbf{Task.} Specialist appointment intake --- intake archetype, care operations coordinator, healthcare. Controls: autonomy=low-risk execute, knowledge burden=small-search retrieval, tool burden=moderate, risk=special review.\\ \textbf{Required.} produce the case record work product \texttt{appt\_1} with 11 required fields (e.g.\ patient\_name=`Sam Doyle'; member\_id=`M-40921'; symptom\_summary=`knee pain with sudden swelling'; duration=`three weeks'); retrieve the governing policy/record (knowledge retrieval via \texttt{kb\_search}); reach the required terminal state (\texttt{READY\_FOR\_REVIEW}). \textit{Twist:} the user corrects red\_flag, symptom\_summary, urgency mid-utterance (barge-in), which the agent must catch and repair.\\ \textbf{Agent.} 3 tool-calls; retrieval done; finalize \emph{not finalized}; approval n/a.\\ \textbf{Outcome.} Workflow FAILURE --- failed gate(s): TS, AC, AV. required knowledge retrieval not satisfied --- searched but gold document not retrieved; workflow not finalized --- never submitted/committed; artifact incomplete: 8/11 fields correct (lifecycle<READY\_FOR\_REVIEW(got DRAFT)); wrong/missing: routing (missing), urgency (stale), referral\_on\_file (missing).}\end{failbox}
\begin{failbox}{apexv1\_034 --- Tax-preparation document intake --- Workflow FAILURE}{\scriptsize \textbf{Task.} Tax-preparation document intake --- intake archetype, tax operations coordinator, professional services. Controls: autonomy=prepare-only, knowledge burden=none, tool burden=light, risk=sensitive-data simulation.\\ \textbf{Required.} produce the case record work product \texttt{docint\_1} with 11 required fields (e.g.\ client\_name=`Robin Shah'; tax\_year=`2024'; w2\_count=`1'; ten99\_count=`1'); reach the required terminal state (\texttt{READY\_FOR\_REVIEW}). \textit{Twist:} the user corrects tax\_year, missing\_items, w2\_count mid-utterance (barge-in), which the agent must catch and repair.\\ \textbf{Agent.} 7 tool-calls; retrieval \emph{none}; finalize \emph{not finalized}; approval n/a.\\ \textbf{Outcome.} Workflow FAILURE --- failed gate(s): AV. artifact incomplete: 8/11 fields correct; wrong/missing: w2\_count (got `2' vs `1'), advice\_given (missing), checklist\_status (missing); 1 infra/WS drop(s).}\end{failbox}
\begin{failbox}{apexv1\_035 --- Property-management maintenance intake --- Workflow FAILURE}{\scriptsize \textbf{Task.} Property-management maintenance intake --- intake archetype, property operations coordinator, general enterprise. Controls: autonomy=low-risk execute, knowledge burden=none, tool burden=light, risk=routine.\\ \textbf{Required.} produce the case record work product \texttt{case\_1} with 10 required fields (e.g.\ tenant\_name=`Casey Lund'; unit=`Apt 214'; issue\_type=`plumbing'; issue\_scope=`kitchen sink only'); reach the required terminal state (\texttt{READY\_FOR\_REVIEW}). \textit{Twist:} the user corrects issue\_scope, urgency\_tier mid-utterance (barge-in), which the agent must catch and repair.\\ \textbf{Agent.} 8 tool-calls; retrieval \emph{none}; finalize \emph{not finalized}; approval n/a.\\ \textbf{Outcome.} Workflow FAILURE --- failed gate(s): TS, PV, AV. field(s) left stale after correction: issue\_scope; workflow not finalized --- never submitted/committed; artifact incomplete: 9/10 fields correct (lifecycle<READY\_FOR\_REVIEW(got DRAFT)); wrong/missing: issue\_type (got `kitchen sink' vs `plumbing'), issue\_scope (stale).}\end{failbox}
\begin{failbox}{apexv1\_036 --- B2B customer escalation intake --- Workflow FAILURE}{\scriptsize \textbf{Task.} B2B customer escalation intake --- intake archetype, customer support lead, Software/SaaS. Controls: autonomy=low-risk execute, knowledge burden=small-search retrieval, tool burden=moderate, risk=routine.\\ \textbf{Required.} produce the case record work product \texttt{esc\_1} with 10 required fields (e.g.\ account=`Delta Systems'; primary\_symptom=`payments API 500 errors on capture'; affected\_product=`payments API'; unrelated\_annoyance=`dashboard theme dislike'); retrieve the governing policy/record (knowledge retrieval via \texttt{kb\_search}); reach the required terminal state (\texttt{READY\_FOR\_REVIEW}). \textit{Twist:} the user corrects impact, severity, primary\_symptom mid-utterance (barge-in), which the agent must catch and repair.\\ \textbf{Agent.} 4 tool-calls; retrieval \emph{none}; finalize \emph{not finalized}; approval n/a.\\ \textbf{Outcome.} Workflow FAILURE --- failed gate(s): AC, AV. required knowledge retrieval not satisfied --- kb\_search never called; artifact incomplete: 9/10 fields correct; wrong/missing: impact (stale).}\end{failbox}
\begin{failbox}{apexv1\_037 --- Logistics damaged-shipment intake --- Workflow FAILURE}{\scriptsize \textbf{Task.} Logistics damaged-shipment intake --- intake archetype, claims operations specialist, manufacturing/field ops. Controls: autonomy=low-risk execute, knowledge burden=none, tool burden=light, risk=routine.\\ \textbf{Required.} produce the case record work product \texttt{dmg\_1} with 9 required fields (e.g.\ shipment\_id=`SHP-77210'; carrier=`FastFreight'; delivery\_date=`2026-03-01'; damage\_desc=`crushed corner, two units broken'); reach the required terminal state (\texttt{READY\_FOR\_REVIEW}). \textit{Twist:} the user corrects photo\_index, shipment\_id mid-utterance (barge-in), which the agent must catch and repair.\\ \textbf{Agent.} 1 tool-calls; retrieval \emph{none}; finalize \emph{not finalized}; approval n/a.\\ \textbf{Outcome.} Workflow FAILURE --- failed gate(s): AV. artifact incomplete: 0/9 fields correct; wrong/missing: shipment\_id (missing), carrier (missing), delivery\_date (missing), damage\_desc (missing), +5 more.}\end{failbox}
\begin{failbox}{apexv1\_038 --- Employee workplace-issue intake and routing --- Workflow FAILURE}{\scriptsize \textbf{Task.} Employee workplace-issue intake and routing --- intake archetype, employee relations intake specialist, workplace/HR. Controls: autonomy=prepare-only, knowledge burden=none, tool burden=light, risk=special review.\\ \textbf{Required.} produce the case record work product \texttt{er\_1} with 10 required fields (e.g.\ reporter\_name=`Jamie Cole'; concern\_type=`scheduling unfairness'; event\_date=`2026-02-18'; involved\_parties=`shift supervisor'); reach the required terminal state (\texttt{READY\_FOR\_REVIEW}). \textit{Twist:} the user corrects event\_date, concrete\_event mid-utterance (barge-in), which the agent must catch and repair.\\ \textbf{Agent.} 9 tool-calls; retrieval \emph{none}; finalize \emph{not finalized}; approval n/a.\\ \textbf{Outcome.} Workflow FAILURE --- failed gate(s): TS, PV, AV. field(s) left stale after correction: concrete\_event; workflow not finalized --- never submitted/committed; artifact incomplete: 9/10 fields correct (lifecycle<READY\_FOR\_REVIEW(got DRAFT)); wrong/missing: concrete\_event (stale), findings\_made (missing).}\end{failbox}
\begin{failbox}{apexv1\_039 --- Warranty service case creation --- Workflow FAILURE}{\scriptsize \textbf{Task.} Warranty service case creation --- intake archetype, service coordinator, manufacturing/field ops. Controls: autonomy=low-risk execute, knowledge burden=small-search retrieval, tool burden=moderate, risk=routine.\\ \textbf{Required.} produce the case record work product \texttt{svc\_1} with 10 required fields (e.g.\ customer\_name=`Lena Ford'; model\_family=`TurboMix 500'; serial\_number=`TMX500-44210'; registered\_devices=`two units registered'); retrieve the governing policy/record (knowledge retrieval via \texttt{kb\_search}); reach the required terminal state (\texttt{READY\_FOR\_REVIEW}). \textit{Twist:} the user corrects serial\_number mid-utterance (barge-in), which the agent must catch and repair.\\ \textbf{Agent.} 5 tool-calls; retrieval \emph{none}; finalize \emph{not finalized}; approval n/a.\\ \textbf{Outcome.} Workflow FAILURE --- failed gate(s): TS, AC, AV. required knowledge retrieval not satisfied --- kb\_search never called; workflow not finalized --- never submitted/committed; artifact incomplete: 4/10 fields correct (lifecycle<READY\_FOR\_REVIEW(got DRAFT)); wrong/missing: registered\_devices (missing), purchase\_date (missing), warranty\_status (missing), service\_type (missing), +2 more; 1 infra/WS drop(s).}\end{failbox}
\begin{failbox}{apexv1\_040 --- Client intake to document request and appointment --- Workflow FAILURE}{\scriptsize \textbf{Task.} Client intake to document request and appointment --- intake archetype, client services coordinator, professional services. Controls: autonomy=approval-gated commit, knowledge burden=small-search retrieval, tool burden=moderate, risk=sensitive-data simulation.\\ \textbf{Required.} produce the case record work product \texttt{case\_1} with 11 required fields (e.g.\ client\_name=`Morgan Diaz'; service\_needed=`estate planning'; deadline=`2026-04-10'; required\_docs=`ID, deed, account statements, and '); retrieve the governing policy/record (knowledge retrieval via \texttt{kb\_search}); obtain user approval, then commit/submit. \textit{Twist:} the user corrects appointment\_slot, deadline, meeting\_urgency, required\_docs mid-utterance (barge-in), which the agent must catch and repair.\\ \textbf{Agent.} 3 tool-calls; retrieval \emph{none}; finalize \emph{not finalized}; approval \emph{not sought}.\\ \textbf{Outcome.} Workflow FAILURE --- failed gate(s): TS, AC, AV. required knowledge retrieval not satisfied --- kb\_search never called; workflow not finalized --- never submitted/committed; artifact incomplete: 6/11 fields correct (lifecycle<COMMITTED(got DRAFT)); wrong/missing: required\_docs (stale), meeting\_urgency (stale), appointment\_slot (got `Tuesday at 10 AM' vs `Thursday 9am'), case\_status (got `Ready for review' vs `opened'), +2 more.}\end{failbox}
\begin{failbox}{apexv1\_041 --- Enterprise SaaS login failure --- Workflow FAILURE}{\scriptsize \textbf{Task.} Enterprise SaaS login failure --- troubleshoot archetype, support engineer, Software/SaaS. Controls: autonomy=low-risk execute, knowledge burden=small-search retrieval, tool burden=moderate, risk=routine.\\ \textbf{Required.} produce the ticket work product \texttt{tkt\_1} with 11 required fields (e.g.\ account\_id=`acct\_88'; user\_role=`workspace admin'; symptom=`cannot log in'; sso\_status=`works for colleagues'); retrieve the governing policy/record (knowledge retrieval via \texttt{kb\_search}); reach the required terminal state (\texttt{READY\_FOR\_REVIEW}). \textit{Twist:} the user corrects cause, action\_taken mid-utterance (barge-in), which the agent must catch and repair.\\ \textbf{Agent.} 3 tool-calls; retrieval done; finalize \emph{not finalized}; approval n/a.\\ \textbf{Outcome.} Workflow FAILURE --- failed gate(s): TS, AV. workflow not finalized --- never submitted/committed; artifact incomplete: 8/11 fields correct (lifecycle<READY\_FOR\_REVIEW(got DRAFT)); wrong/missing: last\_change (missing), cause (missing), action\_taken (missing).}\end{failbox}
\begin{failbox}{apexv1\_042 --- VPN connectivity troubleshooting --- Workflow FAILURE}{\scriptsize \textbf{Task.} VPN connectivity troubleshooting --- troubleshoot archetype, IT help-desk technician, Software/SaaS. Controls: autonomy=low-risk execute, knowledge burden=small-search retrieval, tool burden=moderate, risk=routine.\\ \textbf{Required.} produce the ticket work product \texttt{tkt\_1} with 11 required fields (e.g.\ employee\_id=`E7781'; device=`company laptop'; os=`Windows 11'; symptom=`VPN times out'); retrieve the governing policy/record (knowledge retrieval via \texttt{kb\_search}); reach the required terminal state (\texttt{READY\_FOR\_REVIEW}). \textit{Twist:} the user corrects cause, error\_code, resolution mid-utterance (barge-in), which the agent must catch and repair.\\ \textbf{Agent.} 9 tool-calls; retrieval done; finalize \emph{not finalized}; approval n/a.\\ \textbf{Outcome.} Workflow FAILURE --- failed gate(s): TS, PV, AV. field(s) left stale after correction: cause, resolution; workflow not finalized --- never submitted/committed; artifact incomplete: 7/11 fields correct (lifecycle<READY\_FOR\_REVIEW(got DRAFT)); wrong/missing: last\_change (missing), cause (stale), steps\_tried (missing), resolution (stale).}\end{failbox}
\begin{failbox}{apexv1\_043 --- POS terminal offline triage --- Workflow FAILURE}{\scriptsize \textbf{Task.} POS terminal offline triage --- troubleshoot archetype, retail support technician, general enterprise. Controls: autonomy=approval-gated commit, knowledge burden=supplied evidence, tool burden=moderate, risk=consequential action.\\ \textbf{Required.} produce the ticket work product \texttt{tkt\_1} with 11 required fields (e.g.\ store\_id=`ST-142'; terminal\_id=`POS-5'; symptom=`terminal offline'; network\_status=`other terminals online'); retrieve the governing policy/record (knowledge retrieval via \texttt{kb\_search}); obtain user approval, then commit/submit. \textit{Twist:} the user corrects terminal\_id, cause mid-utterance (barge-in), which the agent must catch and repair.\\ \textbf{Agent.} 9 tool-calls; retrieval done; finalize \emph{not finalized}; approval \emph{not sought}.\\ \textbf{Outcome.} Workflow FAILURE --- failed gate(s): TS, AV. workflow not finalized --- never submitted/committed; artifact incomplete: 10/11 fields correct (lifecycle<COMMITTED(got DRAFT)); wrong/missing: status (missing).}\end{failbox}
\begin{failbox}{apexv1\_044 --- API authentication failure --- Workflow FAILURE}{\scriptsize \textbf{Task.} API authentication failure --- troubleshoot archetype, developer support engineer, Software/SaaS. Controls: autonomy=prepare-only, knowledge burden=small-search retrieval, tool burden=moderate, risk=routine.\\ \textbf{Required.} produce the ticket work product \texttt{tkt\_1} with 10 required fields (e.g.\ account\_id=`dev\_4412'; endpoint=`the orders API'; symptom=`401 unauthorized'; token\_type=`service token'); retrieve the governing policy/record (knowledge retrieval via \texttt{kb\_search}); reach the required terminal state (\texttt{READY\_FOR\_REVIEW}). \textit{Twist:} the user corrects cause, scope\_ok mid-utterance (barge-in), which the agent must catch and repair.\\ \textbf{Agent.} 6 tool-calls; retrieval done; finalize \emph{not finalized}; approval n/a.\\ \textbf{Outcome.} Workflow FAILURE --- failed gate(s): PV, AV. field(s) left stale after correction: scope\_ok; artifact incomplete: 9/10 fields correct; wrong/missing: scope\_ok (stale), repro\_steps (missing).}\end{failbox}
\begin{failbox}{apexv1\_045 --- Video-conference audio issue --- Workflow FAILURE}{\scriptsize \textbf{Task.} Video-conference audio issue --- troubleshoot archetype, IT support specialist, general enterprise. Controls: autonomy=low-risk execute, knowledge burden=none, tool burden=light, risk=routine.\\ \textbf{Required.} produce the ticket work product \texttt{tkt\_1} with 10 required fields (e.g.\ employee\_id=`E2201'; device=`laptop with headset'; symptom=`no outgoing audio'; app=`the meeting app'); reach the required terminal state (\texttt{READY\_FOR\_REVIEW}). \textit{Twist:} the user corrects cause, test\_result mid-utterance (barge-in), which the agent must catch and repair.\\ \textbf{Agent.} 7 tool-calls; retrieval \emph{none}; finalize \emph{not finalized}; approval n/a.\\ \textbf{Outcome.} Workflow FAILURE --- failed gate(s): AV. artifact incomplete: 9/10 fields correct; wrong/missing: test\_result (got `no test tone heard; audio output issue p' vs `test tone heard after switch').}\end{failbox}
\begin{failbox}{apexv1\_046 --- Industrial sensor connectivity diagnosis --- Workflow FAILURE}{\scriptsize \textbf{Task.} Industrial sensor connectivity diagnosis --- troubleshoot archetype, remote support engineer, manufacturing/field ops. Controls: autonomy=prepare-only, knowledge burden=small-search retrieval, tool burden=moderate, risk=routine.\\ \textbf{Required.} produce the ticket work product \texttt{tkt\_1} with 10 required fields (e.g.\ asset\_id=`SEN-77'; sensor\_type=`temperature sensor'; symptom=`intermittent disconnects'; firmware\_version=`v2.0'); retrieve the governing policy/record (knowledge retrieval via \texttt{kb\_search}); reach the required terminal state (\texttt{READY\_FOR\_REVIEW}). \textit{Twist:} the user corrects cause, firmware\_version, recommended\_action mid-utterance (barge-in), which the agent must catch and repair.\\ \textbf{Agent.} 6 tool-calls; retrieval done; finalize \emph{not finalized}; approval n/a.\\ \textbf{Outcome.} Workflow FAILURE --- failed gate(s): TS, AV. workflow not finalized --- never submitted/committed; artifact incomplete: 4/10 fields correct (lifecycle<READY\_FOR\_REVIEW(got DRAFT)); wrong/missing: signal\_strength (missing), last\_calibration (missing), cause (stale), recommended\_action (stale), +2 more; 1 infra/WS drop(s).}\end{failbox}
\begin{failbox}{apexv1\_047 --- Data-pipeline freshness incident --- Workflow FAILURE}{\scriptsize \textbf{Task.} Data-pipeline freshness incident --- troubleshoot archetype, data operations support, Software/SaaS. Controls: autonomy=low-risk execute, knowledge burden=small-search retrieval, tool burden=moderate, risk=routine.\\ \textbf{Required.} produce the ticket work product \texttt{tkt\_1} with 10 required fields (e.g.\ pipeline\_id=`pl\_revenue\_daily'; symptom=`data six hours stale'; similar\_pipelines=`revenue\_daily, revenue\_hourly, rev'; affected=`revenue\_hourly'); retrieve the governing policy/record (knowledge retrieval via \texttt{kb\_search}); reach the required terminal state (\texttt{READY\_FOR\_REVIEW}). \textit{Twist:} the user corrects affected, job\_status, retry\_result mid-utterance (barge-in), which the agent must catch and repair.\\ \textbf{Agent.} 7 tool-calls; retrieval \emph{none}; finalize \emph{not finalized}; approval n/a.\\ \textbf{Outcome.} Workflow FAILURE --- failed gate(s): TS, AC, AV. required knowledge retrieval not satisfied --- kb\_search never called; workflow not finalized --- never submitted/committed; artifact incomplete: 6/10 fields correct (lifecycle<READY\_FOR\_REVIEW(got DRAFT)); wrong/missing: similar\_pipelines (missing), job\_status (stale), resolution (missing), verified (missing).}\end{failbox}
\begin{failbox}{apexv1\_048 --- CAD license checkout problem --- Workflow FAILURE}{\scriptsize \textbf{Task.} CAD license checkout problem --- troubleshoot archetype, enterprise application support, manufacturing/field ops. Controls: autonomy=prepare-only, knowledge burden=small-search retrieval, tool burden=moderate, risk=routine.\\ \textbf{Required.} produce the ticket work product \texttt{tkt\_1} with 10 required fields (e.g.\ user\_id=`eng\_204'; app=`the CAD suite'; symptom=`license checkout fails'; license\_pool=`Mechanical pool'); retrieve the governing policy/record (knowledge retrieval via \texttt{kb\_search}); reach the required terminal state (\texttt{READY\_FOR\_REVIEW}). \textit{Twist:} the user corrects business\_unit, entitlement, license\_pool mid-utterance (barge-in), which the agent must catch and repair.\\ \textbf{Agent.} 4 tool-calls; retrieval \emph{none}; finalize \emph{not finalized}; approval n/a.\\ \textbf{Outcome.} Workflow FAILURE --- failed gate(s): TS, AC, AV. required knowledge retrieval not satisfied --- kb\_search never called; workflow not finalized --- never submitted/committed; artifact incomplete: 4/10 fields correct (lifecycle<READY\_FOR\_REVIEW(got DRAFT)); wrong/missing: business\_unit (stale), entitlement (stale), cause (missing), resolution (missing), +2 more.}\end{failbox}
\begin{failbox}{apexv1\_049 --- Warehouse label-printer failure --- Workflow FAILURE}{\scriptsize \textbf{Task.} Warehouse label-printer failure --- troubleshoot archetype, operations support technician, manufacturing/field ops. Controls: autonomy=low-risk execute, knowledge burden=none, tool burden=light, risk=routine.\\ \textbf{Required.} produce the ticket work product \texttt{tkt\_1} with 10 required fields (e.g.\ device\_id=`PRN-12'; location=`packing station 3'; symptom=`not printing shipping labels'; test\_page=`test page prints fine'); reach the required terminal state (\texttt{READY\_FOR\_REVIEW}). \textit{Twist:} the user corrects cause, classification, resolution mid-utterance (barge-in), which the agent must catch and repair.\\ \textbf{Agent.} 6 tool-calls; retrieval \emph{none}; finalize \emph{not finalized}; approval n/a.\\ \textbf{Outcome.} Workflow FAILURE --- failed gate(s): PV, AV. field(s) left stale after correction: classification, cause; artifact incomplete: 8/10 fields correct; wrong/missing: classification (stale), driver\_status (got `Reinstalled' vs `label driver misconfigured'), cause (stale).}\end{failbox}
\begin{failbox}{apexv1\_050 --- Support call to engineering escalation --- Workflow FAILURE}{\scriptsize \textbf{Task.} Support call to engineering escalation --- troubleshoot archetype, support engineer, Software/SaaS. Controls: autonomy=approval-gated commit, knowledge burden=multi-document reasoning, tool burden=moderate, risk=consequential action.\\ \textbf{Required.} produce the ticket work product \texttt{tkt\_1} with 11 required fields (e.g.\ account=`Orbit Retail'; symptom=`checkout intermittently fails'; proposed\_change=`no change made, escalate instead'; change\_approved=`customer revoked the config change'); retrieve the governing policy/record (knowledge retrieval via \texttt{kb\_search}); obtain user approval, then commit/submit. \textit{Twist:} the user corrects change\_approved, proposed\_change mid-utterance (barge-in), which the agent must catch and repair.\\ \textbf{Agent.} 3 tool-calls; retrieval \emph{none}; finalize committed; approval \emph{not sought}.\\ \textbf{Outcome.} Workflow FAILURE --- failed gate(s): TS, PV, AC, AV. required knowledge retrieval not satisfied --- kb\_search never called; committed/submitted without seeking approval; workflow not brought to terminal state (artifact not COMMITTED); artifact incomplete: 7/11 fields correct (lifecycle<COMMITTED(got DRAFT)); wrong/missing: proposed\_change (missing), change\_approved (missing), customer\_recap (missing), status (missing); tool-call failure: submit\_tkt\_1.}\end{failbox}
\begin{failbox}{apexv1\_051 --- Duplicate invoice charge dispute --- Workflow FAILURE}{\scriptsize \textbf{Task.} Duplicate invoice charge dispute --- negotiate archetype, billing specialist, Software/SaaS. Controls: autonomy=draft-and-confirm, knowledge burden=small-search retrieval, tool burden=moderate, risk=sensitive-data simulation.\\ \textbf{Required.} produce the negotiation record work product \texttt{disp\_1} with 10 required fields (e.g.\ account=`Northwind'; invoice\_number=`INV-771'; disputed\_amount=`480.0'; claimed\_reason=`charged twice'); retrieve the governing policy/record (knowledge retrieval via \texttt{kb\_search}); reach the required terminal state (\texttt{READY\_FOR\_REVIEW}). \textit{Twist:} the user corrects verified\_finding, disposition, eligible\_adjustment mid-utterance (barge-in), which the agent must catch and repair.\\ \textbf{Agent.} 3 tool-calls; retrieval \emph{none}; finalize \emph{not finalized}; approval n/a.\\ \textbf{Outcome.} Workflow FAILURE --- failed gate(s): PV, AC, AV. required knowledge retrieval not satisfied --- kb\_search never called; field(s) left stale after correction: verified\_finding, eligible\_adjustment; artifact incomplete: 9/10 fields correct; wrong/missing: verified\_finding (stale), eligible\_adjustment (stale).}\end{failbox}
\begin{failbox}{apexv1\_052 --- Subscription seat-overage dispute --- Workflow FAILURE}{\scriptsize \textbf{Task.} Subscription seat-overage dispute --- negotiate archetype, account billing specialist, Software/SaaS. Controls: autonomy=draft-and-confirm, knowledge burden=small-search retrieval, tool burden=moderate, risk=sensitive-data simulation.\\ \textbf{Required.} produce the negotiation record work product \texttt{disp\_1} with 9 required fields (e.g.\ account=`Summit Corp'; contract\_seats=`100'; claimed\_seats=`130'; actual\_seats=`130'); retrieve the governing policy/record (knowledge retrieval via \texttt{kb\_search}); reach the required terminal state (\texttt{READY\_FOR\_REVIEW}). \textit{Twist:} the user corrects claimed\_seats, expansion\_event, disposition mid-utterance (barge-in), which the agent must catch and repair.\\ \textbf{Agent.} 5 tool-calls; retrieval \emph{none}; finalize \emph{not finalized}; approval n/a.\\ \textbf{Outcome.} Workflow FAILURE --- failed gate(s): TS, PV, AC, AV. required knowledge retrieval not satisfied --- kb\_search never called; field(s) left stale after correction: claimed\_seats, expansion\_event; workflow not finalized --- never submitted/committed; artifact incomplete: 7/9 fields correct (lifecycle<READY\_FOR\_REVIEW(got DRAFT)); wrong/missing: claimed\_seats (stale), expansion\_event (stale), disposition (missing).}\end{failbox}
\begin{failbox}{apexv1\_053 --- Damaged shipment service recovery --- Workflow FAILURE}{\scriptsize \textbf{Task.} Damaged shipment service recovery --- negotiate archetype, customer operations specialist, manufacturing/field ops. Controls: autonomy=approval-gated commit, knowledge burden=small-search retrieval, tool burden=moderate, risk=consequential action.\\ \textbf{Required.} produce the negotiation record work product \texttt{disp\_1} with 11 required fields (e.g.\ order\_id=`ORD-8890'; damage\_desc=`two of six units cracked'; damage\_evidence=`photos on file'; requested\_remedy=`partial credit instead of refund'); retrieve the governing policy/record (knowledge retrieval via \texttt{kb\_search}); obtain user approval, then commit/submit. \textit{Twist:} the user corrects chosen\_remedy, credit\_amount, requested\_remedy mid-utterance (barge-in), which the agent must catch and repair.\\ \textbf{Agent.} 1 tool-calls; retrieval done; finalize \emph{not finalized}; approval \emph{not sought}.\\ \textbf{Outcome.} Workflow FAILURE --- failed gate(s): TS, AV. workflow not finalized --- never submitted/committed; artifact incomplete: 0/11 fields correct (lifecycle<COMMITTED(got EMPTY)); wrong/missing: order\_id (missing), damage\_desc (missing), damage\_evidence (missing), requested\_remedy (missing), +7 more.}\end{failbox}
\begin{failbox}{apexv1\_054 --- Telecom outage credit request --- Workflow FAILURE}{\scriptsize \textbf{Task.} Telecom outage credit request --- negotiate archetype, service recovery specialist, Software/SaaS. Controls: autonomy=draft-and-confirm, knowledge burden=small-search retrieval, tool burden=moderate, risk=sensitive-data simulation.\\ \textbf{Required.} produce the negotiation record work product \texttt{disp\_1} with 9 required fields (e.g.\ account=`Bayline Retail'; claimed\_outage\_hours=`4'; verified\_outage\_hours=`4'; sla\_threshold=`credit over 2 hours'); retrieve the governing policy/record (knowledge retrieval via \texttt{kb\_search}); reach the required terminal state (\texttt{READY\_FOR\_REVIEW}). \textit{Twist:} the user corrects claimed\_outage\_hours, eligible\_window, credit\_amount mid-utterance (barge-in), which the agent must catch and repair.\\ \textbf{Agent.} 2 tool-calls; retrieval \emph{none}; finalize \emph{not finalized}; approval n/a.\\ \textbf{Outcome.} Workflow FAILURE --- failed gate(s): TS, PV, AC, AV. required knowledge retrieval not satisfied --- kb\_search never called; field(s) left stale after correction: claimed\_outage\_hours, eligible\_window; workflow not finalized --- never submitted/committed; artifact incomplete: 4/9 fields correct (lifecycle<READY\_FOR\_REVIEW(got DRAFT)); wrong/missing: claimed\_outage\_hours (stale), verified\_outage\_hours (missing), eligible\_window (stale), credit\_amount (missing), +2 more.}\end{failbox}
\begin{failbox}{apexv1\_055 --- Vendor late-delivery SLA dispute --- Workflow FAILURE}{\scriptsize \textbf{Task.} Vendor late-delivery SLA dispute --- negotiate archetype, vendor manager, general enterprise. Controls: autonomy=prepare-only, knowledge burden=multi-document reasoning, tool burden=moderate, risk=routine.\\ \textbf{Required.} produce the negotiation record work product \texttt{disp\_1} with 10 required fields (e.g.\ vendor=`Cedar Supply'; po\_number=`PO-4402'; sla\_terms=`delivery within 10 days'; claimed\_exception=`force majeure'); retrieve the governing policy/record (knowledge retrieval via \texttt{kb\_search}); reach the required terminal state (\texttt{READY\_FOR\_REVIEW}). \textit{Twist:} the user corrects delayed\_portion, penalty\_basis, valid\_exception mid-utterance (barge-in), which the agent must catch and repair.\\ \textbf{Agent.} 9 tool-calls; retrieval \emph{none}; finalize \emph{not finalized}; approval n/a.\\ \textbf{Outcome.} Workflow FAILURE --- failed gate(s): PV, AC, AV. required knowledge retrieval not satisfied --- kb\_search never called; field(s) left stale after correction: delayed\_portion, valid\_exception; artifact incomplete: 10/10 fields correct; wrong/missing: delayed\_portion (stale), valid\_exception (stale).}\end{failbox}
\begin{failbox}{apexv1\_056 --- Air-travel fee dispute for corporate traveler --- Workflow FAILURE}{\scriptsize \textbf{Task.} Air-travel fee dispute for corporate traveler --- negotiate archetype, travel support specialist, general enterprise. Controls: autonomy=draft-and-confirm, knowledge burden=small-search retrieval, tool burden=moderate, risk=sensitive-data simulation.\\ \textbf{Required.} produce the negotiation record work product \texttt{disp\_1} with 9 required fields (e.g.\ traveler=`Priya Nair'; ticket\_number=`TK-99210'; fee\_type=`change fee'; fee\_amount=`200.0'); retrieve the governing policy/record (knowledge retrieval via \texttt{kb\_search}); reach the required terminal state (\texttt{READY\_FOR\_REVIEW}). \textit{Twist:} the user corrects eligibility, fare\_rule, ticket\_number mid-utterance (barge-in), which the agent must catch and repair.\\ \textbf{Agent.} 4 tool-calls; retrieval done; finalize \emph{not finalized}; approval n/a.\\ \textbf{Outcome.} Workflow FAILURE --- failed gate(s): AV. artifact incomplete: 8/9 fields correct; wrong/missing: traveler (got `preen air' vs `Priya Nair'), fare\_rule (stale), eligibility (stale).}\end{failbox}
\begin{failbox}{apexv1\_057 --- Service cancellation retention boundary --- Workflow FAILURE}{\scriptsize \textbf{Task.} Service cancellation retention boundary --- negotiate archetype, customer success specialist, Software/SaaS. Controls: autonomy=approval-gated commit, knowledge burden=small-search retrieval, tool burden=moderate, risk=consequential action.\\ \textbf{Required.} produce the negotiation record work product \texttt{disp\_1} with 11 required fields (e.g.\ account=`Vertex Labs'; current\_plan=`Business annual'; cancellation\_reason=`budget cuts'; requested\_discount=`accepts 15 percent alternative'); retrieve the governing policy/record (knowledge retrieval via \texttt{kb\_search}); obtain user approval, then commit/submit. \textit{Twist:} the user corrects accepted\_offer, requested\_discount mid-utterance (barge-in), which the agent must catch and repair.\\ \textbf{Agent.} 0 tool-calls; retrieval \emph{none}; finalize \emph{not finalized}; approval \emph{not sought}.\\ \textbf{Outcome.} Workflow FAILURE --- failed gate(s): TS, AC, AV. required knowledge retrieval not satisfied --- kb\_search never called; workflow not finalized --- never submitted/committed; artifact incomplete: 0/11 fields correct (lifecycle<COMMITTED(got EMPTY)); wrong/missing: account (missing), current\_plan (missing), cancellation\_reason (missing), requested\_discount (missing), +7 more.}\end{failbox}
\begin{failbox}{apexv1\_058 --- Professional-services invoice scope dispute --- Workflow FAILURE}{\scriptsize \textbf{Task.} Professional-services invoice scope dispute --- negotiate archetype, engagement operations specialist, professional services. Controls: autonomy=prepare-only, knowledge burden=multi-document reasoning, tool burden=moderate, risk=routine.\\ \textbf{Required.} produce the negotiation record work product \texttt{disp\_1} with 10 required fields (e.g.\ client=`Baytown Retail'; invoice\_number=`INV-3320'; disputed\_line=`data model review'; sow\_language=`solution design'); retrieve the governing policy/record (knowledge retrieval via \texttt{kb\_search}); reach the required terminal state (\texttt{READY\_FOR\_REVIEW}). \textit{Twist:} the user corrects mapping, finding mid-utterance (barge-in), which the agent must catch and repair.\\ \textbf{Agent.} 4 tool-calls; retrieval \emph{none}; finalize \emph{not finalized}; approval n/a.\\ \textbf{Outcome.} Workflow FAILURE --- failed gate(s): PV, AC, AV. required knowledge retrieval not satisfied --- kb\_search never called; field(s) left stale after correction: mapping; artifact incomplete: 8/10 fields correct; wrong/missing: client (got `Bateon Retail' vs `Baytown Retail'), mapping (stale), disposition (missing).}\end{failbox}
\begin{failbox}{apexv1\_059 --- Cloud usage credit dispute --- Workflow FAILURE}{\scriptsize \textbf{Task.} Cloud usage credit dispute --- negotiate archetype, billing operations specialist, Software/SaaS. Controls: autonomy=draft-and-confirm, knowledge burden=small-search retrieval, tool burden=moderate, risk=sensitive-data simulation.\\ \textbf{Required.} produce the negotiation record work product \texttt{disp\_1} with 10 required fields (e.g.\ account=`Halcyon Media'; bill\_amount=`8200.0'; spike\_1=`nightly batch processing'; spike\_1\_valid=`legitimate'); retrieve the governing policy/record (knowledge retrieval via \texttt{kb\_search}); reach the required terminal state (\texttt{READY\_FOR\_REVIEW}). \textit{Twist:} the user corrects credit\_amount, spike\_2, root\_cause mid-utterance (barge-in), which the agent must catch and repair.\\ \textbf{Agent.} 10 tool-calls; retrieval \emph{none}; finalize \emph{not finalized}; approval n/a.\\ \textbf{Outcome.} Workflow FAILURE --- failed gate(s): PV, AC, AV. required knowledge retrieval not satisfied --- kb\_search never called; field(s) left stale after correction: spike\_2, credit\_amount; artifact incomplete: 10/10 fields correct; wrong/missing: spike\_2 (stale), credit\_amount (stale).}\end{failbox}
\begin{failbox}{apexv1\_060 --- Dispute resolution with approval and follow-up --- Workflow FAILURE}{\scriptsize \textbf{Task.} Dispute resolution with approval and follow-up --- negotiate archetype, customer operations specialist, general enterprise. Controls: autonomy=approval-gated commit, knowledge burden=small-search retrieval, tool burden=moderate, risk=consequential action.\\ \textbf{Required.} produce the negotiation record work product \texttt{disp\_1} with 11 required fields (e.g.\ account=`Orbit Retail'; dispute\_summary=`overcharge on renewal'; verified\_amount=`300.0'; requested\_remedy=`future credit instead of refund'); retrieve the governing policy/record (knowledge retrieval via \texttt{kb\_search}); obtain user approval, then commit/submit. \textit{Twist:} the user corrects final\_remedy, requested\_remedy mid-utterance (barge-in), which the agent must catch and repair.\\ \textbf{Agent.} 1 tool-calls; retrieval \emph{none}; finalize \emph{not finalized}; approval \emph{not sought}.\\ \textbf{Outcome.} Workflow FAILURE --- failed gate(s): TS, AC, AV. required knowledge retrieval not satisfied --- kb\_search never called; workflow not finalized --- never submitted/committed; artifact incomplete: 3/11 fields correct (lifecycle<COMMITTED(got DRAFT)); wrong/missing: verified\_amount (missing), requested\_remedy (missing), final\_remedy (missing), approver (missing), +4 more.}\end{failbox}
\begin{failbox}{apexv1\_061 --- Executive meeting across time zones --- Workflow FAILURE}{\scriptsize \textbf{Task.} Executive meeting across time zones --- coordinate archetype, executive assistant, general enterprise. Controls: autonomy=approval-gated commit, knowledge burden=none, tool burden=light, risk=routine.\\ \textbf{Required.} produce the schedule work product \texttt{sch\_1} with 11 required fields (e.g.\ organizer=`the CFO'; attendees=`CFO, VP Finance, controller'; attendee\_count=`3'; personal\_constraint=`no meetings before 9am for the CFO'); obtain user approval, then commit/submit. \textit{Twist:} the user corrects chosen\_slot, vp\_timezone mid-utterance (barge-in), which the agent must catch and repair.\\ \textbf{Agent.} 9 tool-calls; retrieval \emph{none}; finalize \emph{not finalized}; approval \emph{not sought}.\\ \textbf{Outcome.} Workflow FAILURE --- failed gate(s): TS, AV. workflow not finalized --- never submitted/committed; artifact incomplete: 9/11 fields correct (lifecycle<COMMITTED(got DRAFT)); wrong/missing: chosen\_slot (got `Wednesday 12 PM Eastern' vs `Wednesday 1pm Eastern'), status (missing).}\end{failbox}
\begin{failbox}{apexv1\_062 --- Candidate interview-loop scheduling --- Workflow FAILURE}{\scriptsize \textbf{Task.} Candidate interview-loop scheduling --- coordinate archetype, recruiting coordinator, workplace/HR. Controls: autonomy=approval-gated commit, knowledge burden=none, tool burden=light, risk=sensitive-data simulation.\\ \textbf{Required.} produce the schedule work product \texttt{sch\_1} with 11 required fields (e.g.\ candidate=`Jordan Ellis'; panel\_size=`4'; required\_roles=`hiring manager, two engineers, bar'; loop\_date=`2026-04-08'); obtain user approval, then commit/submit. \textit{Twist:} the user corrects replacement, unavailable\_interviewer mid-utterance (barge-in), which the agent must catch and repair.\\ \textbf{Agent.} 9 tool-calls; retrieval \emph{none}; finalize committed; approval sought.\\ \textbf{Outcome.} Workflow FAILURE --- failed gate(s): PV, AV. field(s) left stale after correction: unavailable\_interviewer, replacement; artifact incomplete: 11/11 fields correct; wrong/missing: unavailable\_interviewer (stale), replacement (stale).}\end{failbox}
\begin{failbox}{apexv1\_063 --- Field-service technician dispatch --- Workflow FAILURE}{\scriptsize \textbf{Task.} Field-service technician dispatch --- coordinate archetype, dispatch coordinator, manufacturing/field ops. Controls: autonomy=approval-gated commit, knowledge burden=small-search retrieval, tool burden=moderate, risk=consequential action.\\ \textbf{Required.} produce the schedule work product \texttt{sch\_1} with 11 required fields (e.g.\ job\_id=`JOB-4410'; site=`Warehouse B'; issue=`conveyor motor fault'; required\_cert=`motor systems certified'); retrieve the governing policy/record (knowledge retrieval via \texttt{kb\_search}); obtain user approval, then commit/submit. \textit{Twist:} the user corrects assigned\_tech, eta mid-utterance (barge-in), which the agent must catch and repair.\\ \textbf{Agent.} 8 tool-calls; retrieval \emph{none}; finalize committed; approval sought.\\ \textbf{Outcome.} Workflow FAILURE --- failed gate(s): PV, AC, AV. required knowledge retrieval not satisfied --- kb\_search never called; field(s) left stale after correction: assigned\_tech, eta; artifact incomplete: 11/11 fields correct; wrong/missing: assigned\_tech (stale), eta (stale).}\end{failbox}
\begin{failbox}{apexv1\_064 --- Specialist clinic scheduling --- Workflow FAILURE}{\scriptsize \textbf{Task.} Specialist clinic scheduling --- coordinate archetype, care coordinator, healthcare. Controls: autonomy=approval-gated commit, knowledge burden=none, tool burden=light, risk=consequential action.\\ \textbf{Required.} produce the schedule work product \texttt{sch\_1} with 11 required fields (e.g.\ patient=`Sam Doyle'; specialty=`cardiology'; constraint=`afternoons only, no Fridays'; preferred\_slot=`Tuesday 2pm'); obtain user approval, then commit/submit. \textit{Twist:} the user corrects chosen\_slot, preferred\_slot mid-utterance (barge-in), which the agent must catch and repair.\\ \textbf{Agent.} 6 tool-calls; retrieval \emph{none}; finalize \emph{not finalized}; approval \emph{not sought}.\\ \textbf{Outcome.} Workflow FAILURE --- failed gate(s): TS, PV, AV. field(s) left stale after correction: preferred\_slot; workflow not finalized --- never submitted/committed; artifact incomplete: 6/11 fields correct (lifecycle<COMMITTED(got DRAFT)); wrong/missing: preferred\_slot (stale), preferred\_available (missing), provider (got `Dr. Newian' vs `Dr. Nguyen'), referral (missing), +2 more.}\end{failbox}
\begin{failbox}{apexv1\_065 --- Maintenance-window coordination --- Workflow FAILURE}{\scriptsize \textbf{Task.} Maintenance-window coordination --- coordinate archetype, IT change coordinator, Software/SaaS. Controls: autonomy=draft-and-confirm, knowledge burden=small-search retrieval, tool burden=moderate, risk=routine.\\ \textbf{Required.} produce the schedule work product \texttt{sch\_1} with 10 required fields (e.g.\ change\_id=`CHG-2201'; system=`billing database'; blackout\_window=`no changes during month-end (28th-'; proposed\_slot=`the 29th at 10pm'); retrieve the governing policy/record (knowledge retrieval via \texttt{kb\_search}); reach the required terminal state (\texttt{READY\_FOR\_REVIEW}). \textit{Twist:} the user corrects chosen\_slot, proposed\_slot mid-utterance (barge-in), which the agent must catch and repair.\\ \textbf{Agent.} 2 tool-calls; retrieval done; finalize \emph{not finalized}; approval n/a.\\ \textbf{Outcome.} Workflow FAILURE --- failed gate(s): TS, PV, AV. field(s) left stale after correction: proposed\_slot, chosen\_slot; workflow not finalized --- never submitted/committed; artifact incomplete: 9/10 fields correct (lifecycle<READY\_FOR\_REVIEW(got DRAFT)); wrong/missing: proposed\_slot (stale), chosen\_slot (stale), stakeholders (got `finance, esari' vs `finance and SRE').}\end{failbox}
\begin{failbox}{apexv1\_066 --- Freight pickup and delivery coordination --- Workflow FAILURE}{\scriptsize \textbf{Task.} Freight pickup and delivery coordination --- coordinate archetype, logistics coordinator, manufacturing/field ops. Controls: autonomy=approval-gated commit, knowledge burden=small-search retrieval, tool burden=moderate, risk=consequential action.\\ \textbf{Required.} produce the schedule work product \texttt{sch\_1} with 11 required fields (e.g.\ shipment\_id=`SHP-6600'; origin=`Dallas warehouse'; destination=`Phoenix DC'; pickup\_slot=`Monday 2pm'); retrieve the governing policy/record (knowledge retrieval via \texttt{kb\_search}); obtain user approval, then commit/submit. \textit{Twist:} the user corrects delivery\_slot, pickup\_slot, recompute\_note mid-utterance (barge-in), which the agent must catch and repair.\\ \textbf{Agent.} 5 tool-calls; retrieval \emph{none}; finalize \emph{not finalized}; approval \emph{not sought}.\\ \textbf{Outcome.} Workflow FAILURE --- failed gate(s): TS, AC, AV. required knowledge retrieval not satisfied --- kb\_search never called; workflow not finalized --- never submitted/committed; artifact incomplete: 6/11 fields correct (lifecycle<COMMITTED(got DRAFT)); wrong/missing: pickup\_delayed (missing), delivery\_slot (got `Tuesday 6 AM' vs `Tuesday noon'), recompute\_note (stale), booking\_status (missing), +1 more.}\end{failbox}
\begin{failbox}{apexv1\_067 --- Training-session scheduling for distributed team --- Workflow FAILURE}{\scriptsize \textbf{Task.} Training-session scheduling for distributed team --- coordinate archetype, learning coordinator, general enterprise. Controls: autonomy=draft-and-confirm, knowledge burden=none, tool burden=light, risk=routine.\\ \textbf{Required.} produce the schedule work product \texttt{sch\_1} with 9 required fields (e.g.\ training=`security awareness'; attendee\_count=`12'; default\_timezone=`Pacific'; exception\_attendees=`two in Central Europe'); reach the required terminal state (\texttt{READY\_FOR\_REVIEW}). \textit{Twist:} the user corrects chosen\_slot, exception\_attendees mid-utterance (barge-in), which the agent must catch and repair.\\ \textbf{Agent.} 2 tool-calls; retrieval \emph{none}; finalize \emph{not finalized}; approval n/a.\\ \textbf{Outcome.} Workflow FAILURE --- failed gate(s): TS, AV. workflow not finalized --- never submitted/committed; artifact incomplete: 3/9 fields correct (lifecycle<READY\_FOR\_REVIEW(got DRAFT)); wrong/missing: exception\_attendees (missing), trainer (missing), duration (missing), chosen\_slot (missing), +2 more.}\end{failbox}
\begin{passbox}{apexv1\_068 --- Customer implementation kickoff coordination --- Workflow SUCCESS}{\scriptsize \textbf{Task.} Customer implementation kickoff coordination --- coordinate archetype, implementation manager, Software/SaaS. Controls: autonomy=draft-and-confirm, knowledge burden=none, tool burden=light, risk=routine.\\ \textbf{Required.} produce the schedule work product \texttt{sch\_1} with 9 required fields (e.g.\ customer=`Vertex Labs'; required\_roles=`PM, tech lead, exec sponsor'; added\_stakeholder=`security lead added'; attendee\_count=`5'); reach the required terminal state (\texttt{READY\_FOR\_REVIEW}). \textit{Twist:} the user corrects added\_stakeholder, agenda, attendee\_count mid-utterance (barge-in), which the agent must catch and repair.\\ \textbf{Agent.} 4 tool-calls; retrieval \emph{none}; finalize \emph{not finalized}; approval n/a.\\ \textbf{Outcome.} Workflow SUCCESS --- all gates pass; artifact field accuracy 9/9.}\end{passbox}
\begin{failbox}{apexv1\_069 --- Shared-lab resource booking --- Workflow FAILURE}{\scriptsize \textbf{Task.} Shared-lab resource booking --- coordinate archetype, research operations coordinator, professional services. Controls: autonomy=approval-gated commit, knowledge burden=small-search retrieval, tool burden=moderate, risk=consequential action.\\ \textbf{Required.} produce the schedule work product \texttt{sch\_1} with 10 required fields (e.g.\ researcher=`Dr. Vale'; equipment=`electron microscope'; requested\_slot=`Wednesday 1pm to 5pm'; calibration\_block=`calibration Wednesday 3pm to 4pm'); retrieve the governing policy/record (knowledge retrieval via \texttt{kb\_search}); obtain user approval, then commit/submit. \textit{Twist:} the user corrects chosen\_slot, requested\_slot mid-utterance (barge-in), which the agent must catch and repair.\\ \textbf{Agent.} 4 tool-calls; retrieval done; finalize \emph{not finalized}; approval \emph{not sought}.\\ \textbf{Outcome.} Workflow FAILURE --- failed gate(s): TS, PV, AV. field(s) left stale after correction: requested\_slot, chosen\_slot; workflow not finalized --- never submitted/committed; artifact incomplete: 9/10 fields correct (lifecycle<COMMITTED(got DRAFT)); wrong/missing: requested\_slot (stale), chosen\_slot (stale), status (missing).}\end{failbox}
\begin{failbox}{apexv1\_070 --- Travel disruption rebooking bundle --- Workflow FAILURE}{\scriptsize \textbf{Task.} Travel disruption rebooking bundle --- coordinate archetype, corporate travel coordinator, general enterprise. Controls: autonomy=approval-gated commit, knowledge burden=small-search retrieval, tool burden=moderate, risk=consequential action.\\ \textbf{Required.} produce the schedule work product \texttt{sch\_1} with 11 required fields (e.g.\ traveler=`Priya Nair'; canceled\_flight=`PN123 to Chicago'; replacement\_flight=`PN458 midday'; replacement\_available=`sold out, use PN458'); retrieve the governing policy/record (knowledge retrieval via \texttt{kb\_search}); obtain user approval, then commit/submit. \textit{Twist:} the user corrects final\_flight, replacement\_flight, rental\_car mid-utterance (barge-in), which the agent must catch and repair.\\ \textbf{Agent.} 5 tool-calls; retrieval done; finalize \emph{not finalized}; approval \emph{not sought}.\\ \textbf{Outcome.} Workflow FAILURE --- failed gate(s): TS, AV. workflow not finalized --- never submitted/committed; artifact incomplete: 9/11 fields correct (lifecycle<COMMITTED(got DRAFT)); wrong/missing: final\_flight (stale), itinerary\_status (missing), status (missing).}\end{failbox}
\begin{failbox}{apexv1\_071 --- Laptop procurement under budget and spec --- Workflow FAILURE}{\scriptsize \textbf{Task.} Laptop procurement under budget and spec --- negotiate archetype, procurement specialist, general enterprise. Controls: autonomy=approval-gated commit, knowledge burden=small-search retrieval, tool burden=moderate, risk=consequential action.\\ \textbf{Required.} produce the negotiation record work product \texttt{po\_1} with 12 required fields (e.g.\ requester=`Design team'; quantity=`15'; preferred\_model=`ProBook X'; required\_ram=`32GB'); retrieve the governing policy/record (knowledge retrieval via \texttt{kb\_search}); obtain user approval, then commit/submit. \textit{Twist:} the user corrects quantity, total\_cost, selection mid-utterance (barge-in), which the agent must catch and repair.\\ \textbf{Agent.} 6 tool-calls; retrieval done; finalize \emph{not finalized}; approval sought.\\ \textbf{Outcome.} Workflow FAILURE --- failed gate(s): TS, AV. workflow not finalized --- never submitted/committed; artifact incomplete: 7/12 fields correct (lifecycle<COMMITTED(got DRAFT)); wrong/missing: requester (missing), vendor (missing), selection (missing), total\_cost (got `21600' vs `27000.0'), +1 more.}\end{failbox}
\begin{failbox}{apexv1\_072 --- SaaS renewal term negotiation --- Workflow FAILURE}{\scriptsize \textbf{Task.} SaaS renewal term negotiation --- negotiate archetype, vendor manager, Software/SaaS. Controls: autonomy=approval-gated commit, knowledge burden=multi-document reasoning, tool burden=moderate, risk=consequential action.\\ \textbf{Required.} produce the negotiation record work product \texttt{neg\_1} with 11 required fields (e.g.\ vendor=`CloudSuite'; current\_term=`12 months'; vendor\_ask=`24-month term'; authority\_limit=`12 months unless 15 percent discou'); retrieve the governing policy/record (knowledge retrieval via \texttt{kb\_search}); obtain user approval, then commit/submit. \textit{Twist:} the user corrects agreed\_term, offered\_discount, disposition mid-utterance (barge-in), which the agent must catch and repair.\\ \textbf{Agent.} 10 tool-calls; retrieval \emph{none}; finalize \emph{not finalized}; approval \emph{not sought}.\\ \textbf{Outcome.} Workflow FAILURE --- failed gate(s): TS, PV, AC, AV. required knowledge retrieval not satisfied --- kb\_search never called; field(s) left stale after correction: disposition; workflow not finalized --- never submitted/committed; artifact incomplete: 9/11 fields correct (lifecycle<COMMITTED(got DRAFT)); wrong/missing: disposition (stale), status (missing).}\end{failbox}
\begin{failbox}{apexv1\_073 --- Freight carrier rate negotiation --- Workflow FAILURE}{\scriptsize \textbf{Task.} Freight carrier rate negotiation --- negotiate archetype, logistics procurement specialist, manufacturing/field ops. Controls: autonomy=draft-and-confirm, knowledge burden=small-search retrieval, tool burden=moderate, risk=routine.\\ \textbf{Required.} produce the negotiation record work product \texttt{neg\_1} with 10 required fields (e.g.\ lane=`Dallas to Phoenix'; current\_rate=`2.4'; carrier\_offer=`lower rate, slower transit'; offered\_rate=`2.1'); retrieve the governing policy/record (knowledge retrieval via \texttt{kb\_search}); reach the required terminal state (\texttt{READY\_FOR\_REVIEW}). \textit{Twist:} the user corrects offer\_meets\_sla, sla\_requirement, agreed\_rate mid-utterance (barge-in), which the agent must catch and repair.\\ \textbf{Agent.} 9 tool-calls; retrieval \emph{none}; finalize \emph{not finalized}; approval n/a.\\ \textbf{Outcome.} Workflow FAILURE --- failed gate(s): PV, AC, AV. required knowledge retrieval not satisfied --- kb\_search never called; field(s) left stale after correction: sla\_requirement, agreed\_rate; artifact incomplete: 9/10 fields correct; wrong/missing: offered\_rate (missing), sla\_requirement (stale), agreed\_rate (stale).}\end{failbox}
\begin{failbox}{apexv1\_074 --- Catering vendor selection and terms --- Workflow FAILURE}{\scriptsize \textbf{Task.} Catering vendor selection and terms --- negotiate archetype, event operations buyer, general enterprise. Controls: autonomy=draft-and-confirm, knowledge burden=none, tool burden=light, risk=routine.\\ \textbf{Required.} produce the negotiation record work product \texttt{neg\_1} with 10 required fields (e.g.\ event=`all-hands lunch'; headcount=`90'; dietary\_vegetarian=`14'; dietary\_gluten\_free=`5'); reach the required terminal state (\texttt{READY\_FOR\_REVIEW}). \textit{Twist:} the user corrects dietary\_vegetarian, final\_quantity, headcount, total\_cost mid-utterance (barge-in), which the agent must catch and repair.\\ \textbf{Agent.} 7 tool-calls; retrieval \emph{none}; finalize \emph{not finalized}; approval n/a.\\ \textbf{Outcome.} Workflow FAILURE --- failed gate(s): AV. artifact incomplete: 9/10 fields correct; wrong/missing: dietary\_vegetarian (got `10' vs `14'); 1 infra/WS drop(s).}\end{failbox}
\begin{failbox}{apexv1\_075 --- Contractor SOW negotiation --- Workflow FAILURE}{\scriptsize \textbf{Task.} Contractor SOW negotiation --- negotiate archetype, procurement manager, professional services. Controls: autonomy=prepare-only, knowledge burden=multi-document reasoning, tool burden=moderate, risk=routine.\\ \textbf{Required.} produce the negotiation record work product \texttt{neg\_1} with 10 required fields (e.g.\ vendor=`Apex Consulting'; scope\_authorized=`data migration and testing'; extra\_deliverable=`vendor proposes a dashboard'; extra\_in\_scope=`no, out of authorized scope'); retrieve the governing policy/record (knowledge retrieval via \texttt{kb\_search}); reach the required terminal state (\texttt{READY\_FOR\_REVIEW}). \textit{Twist:} the user corrects extra\_deliverable, extra\_in\_scope, proposed\_rate, rate\_ok mid-utterance (barge-in), which the agent must catch and repair.\\ \textbf{Agent.} 6 tool-calls; retrieval \emph{none}; finalize \emph{not finalized}; approval n/a.\\ \textbf{Outcome.} Workflow FAILURE --- failed gate(s): PV, AC, AV. required knowledge retrieval not satisfied --- kb\_search never called; field(s) left stale after correction: extra\_deliverable, extra\_in\_scope, proposed\_rate; artifact incomplete: 8/10 fields correct; wrong/missing: extra\_deliverable (stale), extra\_in\_scope (stale), proposed\_rate (stale), rate\_ok (stale), +1 more.}\end{failbox}
\begin{failbox}{apexv1\_076 --- Software-license volume purchase --- Workflow FAILURE}{\scriptsize \textbf{Task.} Software-license volume purchase --- negotiate archetype, IT procurement specialist, Software/SaaS. Controls: autonomy=draft-and-confirm, knowledge burden=small-search retrieval, tool burden=moderate, risk=routine.\\ \textbf{Required.} produce the negotiation record work product \texttt{neg\_1} with 10 required fields (e.g.\ product=`design suite'; needed\_seats=`180'; tier\_threshold=`discount tier at 200 seats'; overbuy\_considered=`buy 200 for the discount'); retrieve the governing policy/record (knowledge retrieval via \texttt{kb\_search}); reach the required terminal state (\texttt{READY\_FOR\_REVIEW}). \textit{Twist:} the user corrects overbuy\_considered, recommendation, recommended\_seats mid-utterance (barge-in), which the agent must catch and repair.\\ \textbf{Agent.} 7 tool-calls; retrieval done; finalize \emph{not finalized}; approval n/a.\\ \textbf{Outcome.} Workflow FAILURE --- failed gate(s): PV, AV. field(s) left stale after correction: overbuy\_considered, recommended\_seats, recommendation; artifact incomplete: 9/10 fields correct; wrong/missing: overbuy\_considered (stale), recommended\_seats (stale), recommendation (stale).}\end{failbox}
\begin{failbox}{apexv1\_077 --- Packaging supplier contingency negotiation --- Workflow FAILURE}{\scriptsize \textbf{Task.} Packaging supplier contingency negotiation --- negotiate archetype, supply-chain buyer, manufacturing/field ops. Controls: autonomy=approval-gated commit, knowledge burden=small-search retrieval, tool burden=moderate, risk=consequential action.\\ \textbf{Required.} produce the negotiation record work product \texttt{neg\_1} with 11 required fields (e.g.\ primary\_supplier=`down for maintenance'; backup\_supplier=`Cedar Packaging'; volume\_needed=`50000'; split\_delivery=`two shipments required'); retrieve the governing policy/record (knowledge retrieval via \texttt{kb\_search}); obtain user approval, then commit/submit. \textit{Twist:} the user corrects first\_delivery\_qty, disposition mid-utterance (barge-in), which the agent must catch and repair.\\ \textbf{Agent.} 0 tool-calls; retrieval \emph{none}; finalize \emph{not finalized}; approval \emph{not sought}.\\ \textbf{Outcome.} Workflow FAILURE --- failed gate(s): TS, AC, AV. required knowledge retrieval not satisfied --- kb\_search never called; workflow not finalized --- never submitted/committed; artifact incomplete: 0/11 fields correct (lifecycle<COMMITTED(got EMPTY)); wrong/missing: primary\_supplier (missing), backup\_supplier (missing), volume\_needed (missing), split\_delivery (missing), +7 more.}\end{failbox}
\begin{failbox}{apexv1\_078 --- Event venue negotiation --- Workflow FAILURE}{\scriptsize \textbf{Task.} Event venue negotiation --- negotiate archetype, events procurement specialist, general enterprise. Controls: autonomy=draft-and-confirm, knowledge burden=none, tool burden=light, risk=routine.\\ \textbf{Required.} produce the negotiation record work product \texttt{neg\_1} with 10 required fields (e.g.\ event=`customer conference'; attendees=`150'; venue=`Harbor Center'; min\_spend=`12000.0'); reach the required terminal state (\texttt{READY\_FOR\_REVIEW}). \textit{Twist:} the user corrects offer\_acceptable, venue\_offer, disposition mid-utterance (barge-in), which the agent must catch and repair.\\ \textbf{Agent.} 0 tool-calls; retrieval \emph{none}; finalize \emph{not finalized}; approval n/a.\\ \textbf{Outcome.} Workflow FAILURE --- failed gate(s): TS, AV. workflow not finalized --- never submitted/committed; artifact incomplete: 0/10 fields correct (lifecycle<READY\_FOR\_REVIEW(got EMPTY)); wrong/missing: event (missing), attendees (missing), venue (missing), min\_spend (missing), +6 more.}\end{failbox}
\begin{failbox}{apexv1\_079 --- Maintenance-service contract terms --- Workflow FAILURE}{\scriptsize \textbf{Task.} Maintenance-service contract terms --- negotiate archetype, facilities buyer, general enterprise. Controls: autonomy=draft-and-confirm, knowledge burden=multi-document reasoning, tool burden=moderate, risk=routine.\\ \textbf{Required.} produce the negotiation record work product \texttt{neg\_1} with 10 required fields (e.g.\ vendor=`Reliant Facilities'; equipment=`critical chillers'; vendor\_offer=`cheaper 8-hour response'; required\_response=`4-hour response for critical'); retrieve the governing policy/record (knowledge retrieval via \texttt{kb\_search}); reach the required terminal state (\texttt{READY\_FOR\_REVIEW}). \textit{Twist:} the user corrects offer\_meets\_need, vendor\_offer, agreed\_response mid-utterance (barge-in), which the agent must catch and repair.\\ \textbf{Agent.} 2 tool-calls; retrieval \emph{none}; finalize \emph{not finalized}; approval n/a.\\ \textbf{Outcome.} Workflow FAILURE --- failed gate(s): TS, PV, AC, AV. required knowledge retrieval not satisfied --- kb\_search never called; field(s) left stale after correction: offer\_meets\_need, agreed\_response; workflow not finalized --- never submitted/committed; artifact incomplete: 10/10 fields correct (lifecycle<READY\_FOR\_REVIEW(got DRAFT)); wrong/missing: offer\_meets\_need (stale), agreed\_response (stale).}\end{failbox}
\begin{failbox}{apexv1\_080 --- Vendor call to purchase request and follow-up --- Workflow FAILURE}{\scriptsize \textbf{Task.} Vendor call to purchase request and follow-up --- negotiate archetype, procurement manager, Software/SaaS. Controls: autonomy=approval-gated commit, knowledge burden=multi-document reasoning, tool burden=moderate, risk=consequential action.\\ \textbf{Required.} produce the negotiation record work product \texttt{neg\_1} with 11 required fields (e.g.\ vendor=`DataPipe Inc'; item=`annual data platform license'; agreed\_price=`60000.0'; vendor\_payment\_terms=`net 15'); retrieve the governing policy/record (knowledge retrieval via \texttt{kb\_search}); obtain user approval, then commit/submit. \textit{Twist:} the user corrects terms\_ok, vendor\_payment\_terms, final\_terms mid-utterance (barge-in), which the agent must catch and repair.\\ \textbf{Agent.} 7 tool-calls; retrieval done; finalize \emph{not finalized}; approval \emph{not sought}.\\ \textbf{Outcome.} Workflow FAILURE --- failed gate(s): TS, AV. workflow not finalized --- never submitted/committed; artifact incomplete: 7/11 fields correct (lifecycle<COMMITTED(got DRAFT)); wrong/missing: vendor\_payment\_terms (got `Net 30' vs `net 15'), terms\_ok (stale), final\_terms (missing), status (missing).}\end{failbox}
\begin{failbox}{apexv1\_081 --- Employee benefits eligibility advisor --- Workflow FAILURE}{\scriptsize \textbf{Task.} Employee benefits eligibility advisor --- advise archetype, benefits specialist, workplace/HR. Controls: autonomy=prepare-only, knowledge burden=multi-document reasoning, tool burden=moderate, risk=sensitive-data simulation.\\ \textbf{Required.} produce the memo/report work product \texttt{memo\_1} with 9 required fields (e.g.\ employment\_type=`full-time'; tenure\_months=`14'; dependents=`spouse and a new child'; current\_elections=`PPO only'); retrieve the governing policy/record (knowledge retrieval via \texttt{kb\_search}); reach the required terminal state (\texttt{READY\_FOR\_REVIEW}). \textit{Twist:} the user corrects dependents, eligible\_fsa mid-utterance (barge-in), which the agent must catch and repair.\\ \textbf{Agent.} 3 tool-calls; retrieval done; finalize \emph{not finalized}; approval n/a.\\ \textbf{Outcome.} Workflow FAILURE --- failed gate(s): PV, AV. field(s) left stale after correction: dependents, eligible\_fsa; artifact incomplete: 8/9 fields correct; wrong/missing: dependents (stale), eligible\_fsa (stale).}\end{failbox}
\begin{failbox}{apexv1\_082 --- Expense-policy advisor --- Workflow FAILURE}{\scriptsize \textbf{Task.} Expense-policy advisor --- advise archetype, finance operations specialist, general enterprise. Controls: autonomy=prepare-only, knowledge burden=multi-document reasoning, tool burden=moderate, risk=routine.\\ \textbf{Required.} produce the memo/report work product \texttt{memo\_1} with 9 required fields (e.g.\ expense\_type=`team meal'; meal\_limit=`75 per person per day'; entertainment\_flag=`client entertainment involved'; entertainment\_rule=`needs attendee list and business p'); retrieve the governing policy/record (knowledge retrieval via \texttt{kb\_search}); reach the required terminal state (\texttt{READY\_FOR\_REVIEW}). \textit{Twist:} the user corrects entertainment\_flag, entertainment\_rule, required\_docs mid-utterance (barge-in), which the agent must catch and repair.\\ \textbf{Agent.} 2 tool-calls; retrieval \emph{none}; finalize \emph{not finalized}; approval n/a.\\ \textbf{Outcome.} Workflow FAILURE --- failed gate(s): TS, PV, AC, AV. required knowledge retrieval not satisfied --- kb\_search never called; field(s) left stale after correction: entertainment\_flag, entertainment\_rule, required\_docs; workflow not finalized --- never submitted/committed; artifact incomplete: 4/9 fields correct (lifecycle<READY\_FOR\_REVIEW(got DRAFT)); wrong/missing: expense\_type (missing), entertainment\_flag (stale), entertainment\_rule (stale), required\_docs (stale), +3 more.}\end{failbox}
\begin{failbox}{apexv1\_083 --- Travel-policy option advisor --- Workflow FAILURE}{\scriptsize \textbf{Task.} Travel-policy option advisor --- advise archetype, travel coordinator, general enterprise. Controls: autonomy=prepare-only, knowledge burden=multi-document reasoning, tool burden=moderate, risk=routine.\\ \textbf{Required.} produce the memo/report work product \texttt{memo\_1} with 10 required fields (e.g.\ destination=`Chicago'; arrival\_requirement=`must arrive before 9am'; cheapest\_flight=`red-eye arriving 11am'; cheapest\_compliant=`no, violates arrival requirement'); retrieve the governing policy/record (knowledge retrieval via \texttt{kb\_search}); reach the required terminal state (\texttt{READY\_FOR\_REVIEW}). \textit{Twist:} the user corrects cheapest\_compliant, cheapest\_flight mid-utterance (barge-in), which the agent must catch and repair.\\ \textbf{Agent.} 2 tool-calls; retrieval done; finalize \emph{not finalized}; approval n/a.\\ \textbf{Outcome.} Workflow FAILURE --- failed gate(s): TS, PV, AV. field(s) left stale after correction: cheapest\_flight, cheapest\_compliant; workflow not finalized --- never submitted/committed; artifact incomplete: 7/10 fields correct (lifecycle<READY\_FOR\_REVIEW(got DRAFT)); wrong/missing: cheapest\_flight (stale), cheapest\_compliant (stale), recommendation (missing), cited\_basis (missing).}\end{failbox}
\begin{failbox}{apexv1\_084 --- Procurement-policy routing advisor --- Workflow FAILURE}{\scriptsize \textbf{Task.} Procurement-policy routing advisor --- advise archetype, procurement operations specialist, general enterprise. Controls: autonomy=prepare-only, knowledge burden=multi-document reasoning, tool burden=moderate, risk=routine.\\ \textbf{Required.} produce the memo/report work product \texttt{memo\_1} with 9 required fields (e.g.\ item=`analytics subscription'; monthly\_price=`3000.0'; term\_months=`12'; annualized\_value=`36000.0'); retrieve the governing policy/record (knowledge retrieval via \texttt{kb\_search}); reach the required terminal state (\texttt{READY\_FOR\_REVIEW}). \textit{Twist:} the user corrects annualized\_value, approval\_path, approval\_tier mid-utterance (barge-in), which the agent must catch and repair.\\ \textbf{Agent.} 6 tool-calls; retrieval done; finalize \emph{not finalized}; approval n/a.\\ \textbf{Outcome.} Workflow FAILURE --- failed gate(s): PV, AV. field(s) left stale after correction: annualized\_value, approval\_tier, approval\_path; artifact incomplete: 8/9 fields correct; wrong/missing: annualized\_value (stale), approval\_tier (stale), approval\_path (stale).}\end{failbox}
\begin{failbox}{apexv1\_085 --- Support SLA advisor --- Workflow FAILURE}{\scriptsize \textbf{Task.} Support SLA advisor --- advise archetype, service operations manager, Software/SaaS. Controls: autonomy=prepare-only, knowledge burden=multi-document reasoning, tool burden=moderate, risk=routine.\\ \textbf{Required.} produce the memo/report work product \texttt{memo\_1} with 9 required fields (e.g.\ account=`Meridian Bank'; plan\_type=`custom enterprise plan'; base\_sla=`sev-1 in 4 hours'; amendment=`amendment sets sev-1 to 1 hour'); retrieve the governing policy/record (knowledge retrieval via \texttt{kb\_search}); reach the required terminal state (\texttt{READY\_FOR\_REVIEW}). \textit{Twist:} the user corrects amendment, applicable\_sla mid-utterance (barge-in), which the agent must catch and repair.\\ \textbf{Agent.} 8 tool-calls; retrieval done; finalize \emph{not finalized}; approval n/a.\\ \textbf{Outcome.} Workflow FAILURE --- failed gate(s): TS, PV, AV. field(s) left stale after correction: amendment; workflow not finalized --- never submitted/committed; artifact incomplete: 8/9 fields correct (lifecycle<READY\_FOR\_REVIEW(got DRAFT)); wrong/missing: amendment (stale), guidance (missing).}\end{failbox}
\begin{failbox}{apexv1\_086 --- Data-retention policy advisor --- Workflow FAILURE}{\scriptsize \textbf{Task.} Data-retention policy advisor --- advise archetype, security compliance operations, Software/SaaS. Controls: autonomy=prepare-only, knowledge burden=multi-document reasoning, tool burden=moderate, risk=sensitive-data simulation.\\ \textbf{Required.} produce the memo/report work product \texttt{memo\_1} with 9 required fields (e.g.\ record\_class\_1=`transaction logs'; retention\_1=`seven years'; record\_class\_2=`marketing analytics'; retention\_2=`two years'); retrieve the governing policy/record (knowledge retrieval via \texttt{kb\_search}); reach the required terminal state (\texttt{READY\_FOR\_REVIEW}). \textit{Twist:} the user corrects hold\_effect, legal\_hold mid-utterance (barge-in), which the agent must catch and repair.\\ \textbf{Agent.} 4 tool-calls; retrieval \emph{none}; finalize \emph{not finalized}; approval n/a.\\ \textbf{Outcome.} Workflow FAILURE --- failed gate(s): PV, AC, AV. required knowledge retrieval not satisfied --- kb\_search never called; field(s) left stale after correction: legal\_hold, hold\_effect; artifact incomplete: 8/9 fields correct; wrong/missing: legal\_hold (stale), hold\_effect (stale), open\_items (missing).}\end{failbox}
\begin{failbox}{apexv1\_087 --- Parental-leave policy explainer --- Workflow FAILURE}{\scriptsize \textbf{Task.} Parental-leave policy explainer --- advise archetype, HR operations specialist, workplace/HR. Controls: autonomy=prepare-only, knowledge burden=multi-document reasoning, tool burden=moderate, risk=special review.\\ \textbf{Required.} produce the memo/report work product \texttt{memo\_1} with 9 required fields (e.g.\ leave\_type=`parental leave'; company\_weeks=`12 weeks company leave'; process\_steps=`notify manager, file with HR, subm'; required\_docs=`leave request and certification'); retrieve the governing policy/record (knowledge retrieval via \texttt{kb\_search}); reach the required terminal state (\texttt{READY\_FOR\_REVIEW}). \textit{Twist:} the user corrects legal\_question, out\_of\_scope\_flag mid-utterance (barge-in), which the agent must catch and repair.\\ \textbf{Agent.} 7 tool-calls; retrieval done; finalize \emph{not finalized}; approval n/a.\\ \textbf{Outcome.} Workflow FAILURE --- failed gate(s): TS, AV. workflow not finalized --- never submitted/committed; artifact incomplete: 8/9 fields correct (lifecycle<READY\_FOR\_REVIEW(got DRAFT)); wrong/missing: legal\_question (missing).}\end{failbox}
\begin{failbox}{apexv1\_088 --- Product-plan fit advisor --- Workflow FAILURE}{\scriptsize \textbf{Task.} Product-plan fit advisor --- advise archetype, solution specialist, Software/SaaS. Controls: autonomy=prepare-only, knowledge burden=small-search retrieval, tool burden=moderate, risk=routine.\\ \textbf{Required.} produce the memo/report work product \texttt{memo\_1} with 9 required fields (e.g.\ company=`Pace Retail'; team\_size=`30'; key\_needs=`reporting and API access'; must\_have\_integration=`Salesforce integration'); retrieve the governing policy/record (knowledge retrieval via \texttt{kb\_search}); reach the required terminal state (\texttt{READY\_FOR\_REVIEW}). \textit{Twist:} the user corrects must\_have\_integration, preferred\_supports, recommended\_plan mid-utterance (barge-in), which the agent must catch and repair.\\ \textbf{Agent.} 5 tool-calls; retrieval done; finalize \emph{not finalized}; approval n/a.\\ \textbf{Outcome.} Workflow FAILURE --- failed gate(s): PV, AV. field(s) left stale after correction: must\_have\_integration, preferred\_supports; artifact incomplete: 7/9 fields correct; wrong/missing: must\_have\_integration (stale), preferred\_supports (stale), cited\_basis (missing).}\end{failbox}
\begin{failbox}{apexv1\_089 --- Returns and warranty policy advisor --- Workflow FAILURE}{\scriptsize \textbf{Task.} Returns and warranty policy advisor --- advise archetype, customer operations specialist, manufacturing/field ops. Controls: autonomy=prepare-only, knowledge burden=multi-document reasoning, tool burden=moderate, risk=routine.\\ \textbf{Required.} produce the memo/report work product \texttt{memo\_1} with 9 required fields (e.g.\ product=`cordless drill'; approx\_purchase=`about three months ago'; exact\_purchase\_date=`2026-01-05'; return\_window=`30 days'); retrieve the governing policy/record (knowledge retrieval via \texttt{kb\_search}); reach the required terminal state (\texttt{READY\_FOR\_REVIEW}). \textit{Twist:} the user corrects exact\_purchase\_date, return\_eligible mid-utterance (barge-in), which the agent must catch and repair.\\ \textbf{Agent.} 2 tool-calls; retrieval \emph{none}; finalize \emph{not finalized}; approval n/a.\\ \textbf{Outcome.} Workflow FAILURE --- failed gate(s): TS, PV, AC, AV. required knowledge retrieval not satisfied --- kb\_search never called; field(s) left stale after correction: exact\_purchase\_date, return\_eligible; workflow not finalized --- never submitted/committed; artifact incomplete: 3/9 fields correct (lifecycle<READY\_FOR\_REVIEW(got DRAFT)); wrong/missing: exact\_purchase\_date (stale), return\_window (missing), return\_eligible (stale), warranty\_window (missing), +3 more.}\end{failbox}
\begin{failbox}{apexv1\_090 --- Compliance filing routing advisor --- Workflow FAILURE}{\scriptsize \textbf{Task.} Compliance filing routing advisor --- advise archetype, compliance operations specialist, general enterprise. Controls: autonomy=prepare-only, knowledge burden=multi-document reasoning, tool burden=moderate, risk=special review.\\ \textbf{Required.} produce the memo/report work product \texttt{memo\_1} with 9 required fields (e.g.\ event\_summary=`a data access incident'; key\_fact=`no personal data exposed'; category=`internal security event'; routing=`security review, not privacy filin'); retrieve the governing policy/record (knowledge retrieval via \texttt{kb\_search}); reach the required terminal state (\texttt{READY\_FOR\_REVIEW}). \textit{Twist:} the user corrects category, key\_fact, routing mid-utterance (barge-in), which the agent must catch and repair.\\ \textbf{Agent.} 2 tool-calls; retrieval \emph{none}; finalize \emph{not finalized}; approval n/a.\\ \textbf{Outcome.} Workflow FAILURE --- failed gate(s): TS, AC, AV. required knowledge retrieval not satisfied --- kb\_search never called; workflow not finalized --- never submitted/committed; artifact incomplete: 9/9 fields correct (lifecycle<READY\_FOR\_REVIEW(got DRAFT)); wrong/missing: category (stale), routing (stale).}\end{failbox}
\begin{failbox}{apexv1\_091 --- Sprint retrospective action capture --- Workflow FAILURE}{\scriptsize \textbf{Task.} Sprint retrospective action capture --- facilitate archetype, engineering program manager, Software/SaaS. Controls: autonomy=prepare-only, knowledge burden=none, tool burden=light, risk=routine.\\ \textbf{Required.} produce the plan/checklist work product \texttt{retro\_1} with 10 required fields (e.g.\ sprint=`Sprint 24'; went\_well=`faster code review turnaround'; went\_poorly=`flaky CI tests'; action\_1=`stabilize the CI test suite'); reach the required terminal state (\texttt{READY\_FOR\_REVIEW}). \textit{Twist:} the user corrects action\_1\_owner, action\_1\_due mid-utterance (barge-in), which the agent must catch and repair.\\ \textbf{Agent.} 3 tool-calls; retrieval \emph{none}; finalize \emph{not finalized}; approval n/a.\\ \textbf{Outcome.} Workflow FAILURE --- failed gate(s): TS, AV. workflow not finalized --- never submitted/committed; artifact incomplete: 10/10 fields correct (lifecycle<READY\_FOR\_REVIEW(got DRAFT)).}\end{failbox}
\begin{failbox}{apexv1\_092 --- Project status review --- Workflow FAILURE}{\scriptsize \textbf{Task.} Project status review --- facilitate archetype, project manager, professional services. Controls: autonomy=prepare-only, knowledge burden=none, tool burden=light, risk=routine.\\ \textbf{Required.} produce the plan/checklist work product \texttt{status\_1} with 10 required fields (e.g.\ project=`Website Revamp'; workstream\_design=`design on track'; workstream\_build=`build slightly behind'; workstream\_content=`content blocked, now resolved'); reach the required terminal state (\texttt{READY\_FOR\_REVIEW}). \textit{Twist:} the user corrects blocker\_status, workstream\_content, overall\_status mid-utterance (barge-in), which the agent must catch and repair.\\ \textbf{Agent.} 3 tool-calls; retrieval \emph{none}; finalize \emph{not finalized}; approval n/a.\\ \textbf{Outcome.} Workflow FAILURE --- failed gate(s): TS, AV. workflow not finalized --- never submitted/committed; artifact incomplete: 10/10 fields correct (lifecycle<READY\_FOR\_REVIEW(got DRAFT)); 1 infra/WS drop(s).}\end{failbox}
\begin{passbox}{apexv1\_093 --- Customer implementation checkpoint --- Workflow SUCCESS}{\scriptsize \textbf{Task.} Customer implementation checkpoint --- facilitate archetype, implementation manager, Software/SaaS. Controls: autonomy=prepare-only, knowledge burden=none, tool burden=light, risk=routine.\\ \textbf{Required.} produce the plan/checklist work product \texttt{chk\_1} with 10 required fields (e.g.\ customer=`Vertex Labs'; launch\_target=`2026-04-24'; milestone\_1=`data migration done'; milestone\_2=`training scheduled'); reach the required terminal state (\texttt{READY\_FOR\_REVIEW}). \textit{Twist:} the user corrects launch\_target, revised\_task\_1, revised\_task\_2 mid-utterance (barge-in), which the agent must catch and repair.\\ \textbf{Agent.} 9 tool-calls; retrieval \emph{none}; finalize \emph{not finalized}; approval n/a.\\ \textbf{Outcome.} Workflow SUCCESS --- all gates pass; artifact field accuracy 10/10.}\end{passbox}
\begin{failbox}{apexv1\_094 --- Requirements workshop --- Workflow FAILURE}{\scriptsize \textbf{Task.} Requirements workshop --- facilitate archetype, business analyst, Software/SaaS. Controls: autonomy=prepare-only, knowledge burden=none, tool burden=light, risk=routine.\\ \textbf{Required.} produce the plan/checklist work product \texttt{req\_1} with 10 required fields (e.g.\ feature=`customer portal'; req\_1=`SSO login'; req\_1\_priority=`must-have'; req\_2=`dark mode'); reach the required terminal state (\texttt{READY\_FOR\_REVIEW}). \textit{Twist:} the user corrects req\_2\_priority, out\_of\_scope mid-utterance (barge-in), which the agent must catch and repair.\\ \textbf{Agent.} 5 tool-calls; retrieval \emph{none}; finalize \emph{not finalized}; approval n/a.\\ \textbf{Outcome.} Workflow FAILURE --- failed gate(s): TS, PV, AV. field(s) left stale after correction: req\_2\_priority; workflow not finalized --- never submitted/committed; artifact incomplete: 9/10 fields correct (lifecycle<READY\_FOR\_REVIEW(got DRAFT)); wrong/missing: feature (missing), req\_2\_priority (stale).}\end{failbox}
\begin{failbox}{apexv1\_095 --- Design review scribe --- Workflow FAILURE}{\scriptsize \textbf{Task.} Design review scribe --- facilitate archetype, design program manager, Software/SaaS. Controls: autonomy=prepare-only, knowledge burden=none, tool burden=light, risk=routine.\\ \textbf{Required.} produce the plan/checklist work product \texttt{dr\_1} with 10 required fields (e.g.\ feature=`checkout redesign'; option\_a=`Layout Aurora'; option\_b=`Layout Aurora Plus'; approved\_option=`Layout Aurora Plus'); reach the required terminal state (\texttt{READY\_FOR\_REVIEW}). \textit{Twist:} the user corrects approved\_option, rationale mid-utterance (barge-in), which the agent must catch and repair.\\ \textbf{Agent.} 6 tool-calls; retrieval \emph{none}; finalize \emph{not finalized}; approval n/a.\\ \textbf{Outcome.} Workflow FAILURE --- failed gate(s): AV. artifact incomplete: 9/10 fields correct; wrong/missing: decision\_status (missing).}\end{failbox}
\begin{failbox}{apexv1\_096 --- Incident postmortem facilitation --- Workflow FAILURE}{\scriptsize \textbf{Task.} Incident postmortem facilitation --- facilitate archetype, incident program manager, Software/SaaS. Controls: autonomy=prepare-only, knowledge burden=small-search retrieval, tool burden=moderate, risk=routine.\\ \textbf{Required.} produce the plan/checklist work product \texttt{pm\_1} with 10 required fields (e.g.\ incident\_id=`INC-42'; impact=`checkout degraded 40 minutes'; root\_cause=`bad deploy config'; event\_1\_time=`13:52'); retrieve the governing policy/record (knowledge retrieval via \texttt{kb\_search}); reach the required terminal state (\texttt{READY\_FOR\_REVIEW}). \textit{Twist:} the user corrects event\_1\_time, event\_order, action\_1\_owner mid-utterance (barge-in), which the agent must catch and repair.\\ \textbf{Agent.} 3 tool-calls; retrieval \emph{none}; finalize \emph{not finalized}; approval n/a.\\ \textbf{Outcome.} Workflow FAILURE --- failed gate(s): TS, AC, AV. required knowledge retrieval not satisfied --- kb\_search never called; workflow not finalized --- never submitted/committed; artifact incomplete: 9/10 fields correct (lifecycle<READY\_FOR\_REVIEW(got DRAFT)); wrong/missing: event\_order (stale).}\end{failbox}
\begin{failbox}{apexv1\_097 --- Vendor performance review meeting --- Workflow FAILURE}{\scriptsize \textbf{Task.} Vendor performance review meeting --- facilitate archetype, vendor manager, general enterprise. Controls: autonomy=prepare-only, knowledge burden=none, tool burden=light, risk=routine.\\ \textbf{Required.} produce the plan/checklist work product \texttt{vr\_1} with 10 required fields (e.g.\ vendor=`Cedar Supply'; sla\_met=`92 percent on-time'; quality\_score=`4 out of 5'; issue=`late deliveries in Q1'); reach the required terminal state (\texttt{READY\_FOR\_REVIEW}). \textit{Twist:} the user corrects commitment\_firm, vendor\_commitment mid-utterance (barge-in), which the agent must catch and repair.\\ \textbf{Agent.} 1 tool-calls; retrieval \emph{none}; finalize \emph{not finalized}; approval n/a.\\ \textbf{Outcome.} Workflow FAILURE --- failed gate(s): TS, PV, AV. field(s) left stale after correction: vendor\_commitment, commitment\_firm; workflow not finalized --- never submitted/committed; artifact incomplete: 10/10 fields correct (lifecycle<READY\_FOR\_REVIEW(got DRAFT)); wrong/missing: vendor\_commitment (stale), commitment\_firm (stale); 1 infra/WS drop(s).}\end{failbox}
\begin{failbox}{apexv1\_098 --- Launch readiness meeting --- Workflow FAILURE}{\scriptsize \textbf{Task.} Launch readiness meeting --- facilitate archetype, launch program manager, Software/SaaS. Controls: autonomy=prepare-only, knowledge burden=none, tool burden=light, risk=routine.\\ \textbf{Required.} produce the plan/checklist work product \texttt{lr\_1} with 10 required fields (e.g.\ launch=`Payments v2'; dep\_infra=`infrastructure green'; dep\_security=`security review green'; dep\_qa=`QA blocked by a new test failure'); reach the required terminal state (\texttt{READY\_FOR\_REVIEW}). \textit{Twist:} the user corrects dep\_qa, readiness mid-utterance (barge-in), which the agent must catch and repair.\\ \textbf{Agent.} 1 tool-calls; retrieval \emph{none}; finalize \emph{not finalized}; approval n/a.\\ \textbf{Outcome.} Workflow FAILURE --- failed gate(s): TS, PV, AV. field(s) left stale after correction: dep\_qa, readiness; workflow not finalized --- never submitted/committed; artifact incomplete: 9/10 fields correct (lifecycle<READY\_FOR\_REVIEW(got DRAFT)); wrong/missing: dep\_qa (stale), readiness (stale), action\_1 (missing).}\end{failbox}
\begin{failbox}{apexv1\_099 --- Stakeholder research synthesis meeting --- Workflow FAILURE}{\scriptsize \textbf{Task.} Stakeholder research synthesis meeting --- facilitate archetype, research operations lead, professional services. Controls: autonomy=prepare-only, knowledge burden=none, tool burden=light, risk=routine.\\ \textbf{Required.} produce the plan/checklist work product \texttt{rs\_1} with 10 required fields (e.g.\ study=`onboarding research'; theme\_1=`users want faster setup'; theme\_1\_evidence=`8 of 10 interviews'; claim\_corrected=`adoption is 60 percent, corrected '); reach the required terminal state (\texttt{READY\_FOR\_REVIEW}). \textit{Twist:} the user corrects claim\_corrected, opinion\_vs\_policy mid-utterance (barge-in), which the agent must catch and repair.\\ \textbf{Agent.} 1 tool-calls; retrieval \emph{none}; finalize \emph{not finalized}; approval n/a.\\ \textbf{Outcome.} Workflow FAILURE --- failed gate(s): TS, PV, AV. field(s) left stale after correction: claim\_corrected, opinion\_vs\_policy; workflow not finalized --- never submitted/committed; artifact incomplete: 10/10 fields correct (lifecycle<READY\_FOR\_REVIEW(got DRAFT)); wrong/missing: claim\_corrected (stale), opinion\_vs\_policy (stale).}\end{failbox}
\begin{failbox}{apexv1\_100 --- Budget planning meeting record --- Workflow FAILURE}{\scriptsize \textbf{Task.} Budget planning meeting record --- facilitate archetype, finance business partner, general enterprise. Controls: autonomy=draft-and-confirm, knowledge burden=none, tool burden=light, risk=sensitive-data simulation.\\ \textbf{Required.} produce the plan/checklist work product \texttt{bud\_1} with 10 required fields (e.g.\ department=`Marketing'; proposed\_budget=`500000.0'; proposed\_cut=`reduce events line'; cut\_status=`withdrawn before end'); reach the required terminal state (\texttt{READY\_FOR\_REVIEW}). \textit{Twist:} the user corrects cut\_status, proposed\_cut, tooling\_basis mid-utterance (barge-in), which the agent must catch and repair.\\ \textbf{Agent.} 3 tool-calls; retrieval \emph{none}; finalize \emph{not finalized}; approval n/a.\\ \textbf{Outcome.} Workflow FAILURE --- failed gate(s): TS, AV. workflow not finalized --- never submitted/committed; artifact incomplete: 10/10 fields correct (lifecycle<READY\_FOR\_REVIEW(got DRAFT)); wrong/missing: cut\_status (stale); 1 infra/WS drop(s).}\end{failbox}
\begin{failbox}{apexv1\_101 --- HVAC inspection to work order --- Workflow FAILURE}{\scriptsize \textbf{Task.} HVAC inspection to work order --- inspect archetype, field maintenance coordinator, manufacturing/field ops. Controls: autonomy=low-risk execute, knowledge burden=small-search retrieval, tool burden=moderate, risk=routine.\\ \textbf{Required.} produce the work order work product \texttt{wo\_1} with 11 required fields (e.g.\ asset\_id=`HVAC-7'; location=`Building C roof'; filter\_status=`clogged'; supply\_temp=`62'); retrieve the governing policy/record (knowledge retrieval via \texttt{kb\_search}); obtain user approval, then commit/submit. \textit{Twist:} the user corrects priority, supply\_temp, recommended\_action mid-utterance (barge-in), which the agent must catch and repair.\\ \textbf{Agent.} 5 tool-calls; retrieval \emph{none}; finalize \emph{not finalized}; approval \emph{not sought}.\\ \textbf{Outcome.} Workflow FAILURE --- failed gate(s): TS, AC, AV. required knowledge retrieval not satisfied --- kb\_search never called; workflow not finalized --- never submitted/committed; artifact incomplete: 9/11 fields correct (lifecycle<COMMITTED(got DRAFT)); wrong/missing: priority (stale), status (missing).}\end{failbox}
\begin{failbox}{apexv1\_102 --- Safety pre-job verbal checklist --- Workflow FAILURE}{\scriptsize \textbf{Task.} Safety pre-job verbal checklist --- inspect archetype, site safety coordinator, manufacturing/field ops. Controls: autonomy=prepare-only, knowledge burden=none, tool burden=light, risk=special review.\\ \textbf{Required.} produce the work order work product \texttt{sc\_1} with 10 required fields (e.g.\ job\_id=`JOB-2201'; ppe\_check=`hard hat and gloves on'; lockout\_tagout=`applied'; area\_clear=`area clear of personnel'); reach the required terminal state (\texttt{READY\_FOR\_REVIEW}). \textit{Twist:} the user corrects all\_conditions\_met, gas\_check, readiness mid-utterance (barge-in), which the agent must catch and repair.\\ \textbf{Agent.} 6 tool-calls; retrieval \emph{none}; finalize \emph{not finalized}; approval n/a.\\ \textbf{Outcome.} Workflow FAILURE --- failed gate(s): TS, AV. workflow not finalized --- never submitted/committed; artifact incomplete: 7/10 fields correct (lifecycle<READY\_FOR\_REVIEW(got DRAFT)); wrong/missing: readiness (stale), required\_action (got `Investigate and clear elevated gas readi' vs `ventilate and re-test'), sign\_off (missing).}\end{failbox}
\begin{failbox}{apexv1\_103 --- Manufacturing quality inspection --- Workflow FAILURE}{\scriptsize \textbf{Task.} Manufacturing quality inspection --- inspect archetype, quality technician, manufacturing/field ops. Controls: autonomy=draft-and-confirm, knowledge burden=none, tool burden=light, risk=consequential action.\\ \textbf{Required.} produce the work order work product \texttt{qc\_1} with 10 required fields (e.g.\ batch\_id=`BATCH-559'; product=`bearing assembly'; dimension\_spec=`diameter 20mm plus or minus 0.1'; measured\_diameter=`20.15'); reach the required terminal state (\texttt{READY\_FOR\_REVIEW}). \textit{Twist:} the user corrects disposition, hold\_recommendation, in\_tolerance, measured\_diameter mid-utterance (barge-in), which the agent must catch and repair.\\ \textbf{Agent.} 2 tool-calls; retrieval \emph{none}; finalize \emph{not finalized}; approval n/a.\\ \textbf{Outcome.} Workflow FAILURE --- failed gate(s): PV, AV. field(s) left stale after correction: measured\_diameter, in\_tolerance, disposition; artifact incomplete: 7/10 fields correct; wrong/missing: measured\_diameter (stale), in\_tolerance (stale), disposition (stale), hold\_recommendation (stale).}\end{failbox}
\begin{failbox}{apexv1\_104 --- Property condition inspection --- Workflow FAILURE}{\scriptsize \textbf{Task.} Property condition inspection --- inspect archetype, property operations inspector, general enterprise. Controls: autonomy=prepare-only, knowledge burden=none, tool burden=light, risk=routine.\\ \textbf{Required.} produce the work order work product \texttt{cr\_1} with 10 required fields (e.g.\ unit=`Apt 214'; living\_room=`good condition'; kitchen\_damage=`cracked countertop'; kitchen\_severity=`major'); reach the required terminal state (\texttt{READY\_FOR\_REVIEW}). \textit{Twist:} the user corrects deposit\_impact, kitchen\_damage, kitchen\_severity mid-utterance (barge-in), which the agent must catch and repair.\\ \textbf{Agent.} 8 tool-calls; retrieval \emph{none}; finalize \emph{not finalized}; approval n/a.\\ \textbf{Outcome.} Workflow FAILURE --- failed gate(s): AV. artifact incomplete: 8/10 fields correct; wrong/missing: kitchen\_severity (stale), deposit\_impact (got `Minor deduction, mostly wear and tear' vs `significant deduction').}\end{failbox}
\begin{failbox}{apexv1\_105 --- Warehouse inventory spot audit --- Workflow FAILURE}{\scriptsize \textbf{Task.} Warehouse inventory spot audit --- inspect archetype, inventory auditor, manufacturing/field ops. Controls: autonomy=prepare-only, knowledge burden=none, tool burden=light, risk=routine.\\ \textbf{Required.} produce the work order work product \texttt{ia\_1} with 10 required fields (e.g.\ location=`Aisle 7 Bin B'; sku=`SKU-4417'; sku\_description=`label rolls'; system\_count=`120'); reach the required terminal state (\texttt{READY\_FOR\_REVIEW}). \textit{Twist:} the user corrects sku, sku\_description mid-utterance (barge-in), which the agent must catch and repair.\\ \textbf{Agent.} 2 tool-calls; retrieval \emph{none}; finalize \emph{not finalized}; approval n/a.\\ \textbf{Outcome.} Workflow FAILURE --- failed gate(s): TS, AV. workflow not finalized --- never submitted/committed; artifact incomplete: 7/10 fields correct (lifecycle<READY\_FOR\_REVIEW(got DRAFT)); wrong/missing: sku\_description (stale), discrepancy (got `-8 units' vs `8'), auditor (missing).}\end{failbox}
\begin{failbox}{apexv1\_106 --- Fleet vehicle pre-service inspection --- Workflow FAILURE}{\scriptsize \textbf{Task.} Fleet vehicle pre-service inspection --- inspect archetype, fleet maintenance coordinator, manufacturing/field ops. Controls: autonomy=prepare-only, knowledge burden=none, tool burden=light, risk=routine.\\ \textbf{Required.} produce the work order work product \texttt{fi\_1} with 10 required fields (e.g.\ vehicle\_id=`VAN-33'; odometer=`88000'; tire\_condition=`front tires worn'; brake\_condition=`pads at 40 percent'); reach the required terminal state (\texttt{READY\_FOR\_REVIEW}). \textit{Twist:} the user corrects odometer, priority, warning\_light mid-utterance (barge-in), which the agent must catch and repair.\\ \textbf{Agent.} 4 tool-calls; retrieval \emph{none}; finalize \emph{not finalized}; approval n/a.\\ \textbf{Outcome.} Workflow FAILURE --- failed gate(s): TS, PV, AV. field(s) left stale after correction: warning\_light, priority; workflow not finalized --- never submitted/committed; artifact incomplete: 10/10 fields correct (lifecycle<READY\_FOR\_REVIEW(got DRAFT)); wrong/missing: warning\_light (stale), priority (stale); 1 infra/WS drop(s).}\end{failbox}
\begin{failbox}{apexv1\_107 --- Data-center rack inspection --- Workflow FAILURE}{\scriptsize \textbf{Task.} Data-center rack inspection --- inspect archetype, data-center operations technician, Software/SaaS. Controls: autonomy=prepare-only, knowledge burden=small-search retrieval, tool burden=moderate, risk=routine.\\ \textbf{Required.} produce the work order work product \texttt{ri\_1} with 10 required fields (e.g.\ rack\_id=`RACK-91'; temperature=`24'; power\_draw=`within normal'; fan\_status=`fan alert on unit 3'); retrieve the governing policy/record (knowledge retrieval via \texttt{kb\_search}); reach the required terminal state (\texttt{READY\_FOR\_REVIEW}). \textit{Twist:} the user corrects fan\_status, rack\_id mid-utterance (barge-in), which the agent must catch and repair.\\ \textbf{Agent.} 2 tool-calls; retrieval \emph{none}; finalize \emph{not finalized}; approval n/a.\\ \textbf{Outcome.} Workflow FAILURE --- failed gate(s): TS, AC, AV. required knowledge retrieval not satisfied --- kb\_search never called; workflow not finalized --- never submitted/committed; artifact incomplete: 2/10 fields correct (lifecycle<READY\_FOR\_REVIEW(got DRAFT)); wrong/missing: power\_draw (missing), fan\_status (stale), escalation\_trigger (missing), cabling (missing), +4 more; 1 infra/WS drop(s).}\end{failbox}
\begin{failbox}{apexv1\_108 --- Retail store opening checklist --- Workflow FAILURE}{\scriptsize \textbf{Task.} Retail store opening checklist --- inspect archetype, store operations lead, general enterprise. Controls: autonomy=prepare-only, knowledge burden=none, tool burden=light, risk=routine.\\ \textbf{Required.} produce the work order work product \texttt{oc\_1} with 10 required fields (e.g.\ store\_id=`ST-142'; alarm\_disarmed=`yes'; lights\_on=`yes'; registers\_ready=`yes'); reach the required terminal state (\texttt{READY\_FOR\_REVIEW}). \textit{Twist:} the user corrects cash\_drawer, cash\_drawer\_ok, opening\_status mid-utterance (barge-in), which the agent must catch and repair.\\ \textbf{Agent.} 1 tool-calls; retrieval \emph{none}; finalize \emph{not finalized}; approval n/a.\\ \textbf{Outcome.} Workflow FAILURE --- failed gate(s): TS, PV, AV. field(s) left stale after correction: cash\_drawer, cash\_drawer\_ok, opening\_status; workflow not finalized --- never submitted/committed; artifact incomplete: 8/10 fields correct (lifecycle<READY\_FOR\_REVIEW(got DRAFT)); wrong/missing: cash\_drawer (stale), cash\_drawer\_ok (stale), opening\_status (stale), action (missing).}\end{failbox}
\begin{failbox}{apexv1\_109 --- Solar-site maintenance inspection --- Workflow FAILURE}{\scriptsize \textbf{Task.} Solar-site maintenance inspection --- inspect archetype, field service technician, manufacturing/field ops. Controls: autonomy=low-risk execute, knowledge burden=small-search retrieval, tool burden=moderate, risk=routine.\\ \textbf{Required.} produce the work order work product \texttt{si\_1} with 10 required fields (e.g.\ site\_id=`SOLAR-4'; inverter\_id=`INV-2'; inverter\_output=`output 8 percent low'; panel\_condition=`some soiling'); retrieve the governing policy/record (knowledge retrieval via \texttt{kb\_search}); reach the required terminal state (\texttt{READY\_FOR\_REVIEW}). \textit{Twist:} the user corrects inverter\_id, resume\_note mid-utterance (barge-in), which the agent must catch and repair.\\ \textbf{Agent.} 2 tool-calls; retrieval \emph{none}; finalize \emph{not finalized}; approval n/a.\\ \textbf{Outcome.} Workflow FAILURE --- failed gate(s): TS, AC, AV. required knowledge retrieval not satisfied --- kb\_search never called; workflow not finalized --- never submitted/committed; artifact incomplete: 9/10 fields correct (lifecycle<READY\_FOR\_REVIEW(got DRAFT)); wrong/missing: site\_id (got `sold or four' vs `SOLAR-4'), resume\_note (stale).}\end{failbox}
\begin{failbox}{apexv1\_110 --- Inspection to work-order and customer summary --- Workflow FAILURE}{\scriptsize \textbf{Task.} Inspection to work-order and customer summary --- inspect archetype, field service coordinator, manufacturing/field ops. Controls: autonomy=approval-gated commit, knowledge burden=small-search retrieval, tool burden=moderate, risk=consequential action.\\ \textbf{Required.} produce the work order work product \texttt{wo\_1} with 11 required fields (e.g.\ asset\_id=`CHILLER-3'; symptom=`intermittent shutdown'; finding=`loose sensor connector'; initial\_recommendation=`no replacement needed'); retrieve the governing policy/record (knowledge retrieval via \texttt{kb\_search}); obtain user approval, then commit/submit. \textit{Twist:} the user corrects initial\_recommendation, part\_needed, revised\_recommendation mid-utterance (barge-in), which the agent must catch and repair.\\ \textbf{Agent.} 3 tool-calls; retrieval \emph{none}; finalize \emph{not finalized}; approval \emph{not sought}.\\ \textbf{Outcome.} Workflow FAILURE --- failed gate(s): TS, AC, AV. required knowledge retrieval not satisfied --- kb\_search never called; workflow not finalized --- never submitted/committed; artifact incomplete: 8/11 fields correct (lifecycle<COMMITTED(got DRAFT)); wrong/missing: initial\_recommendation (got `Reseat the connector' vs `no replacement needed'), revised\_recommendation (stale), status (missing).}\end{failbox}
\begin{failbox}{apexv1\_111 --- SaaS outage triage coordination --- Workflow FAILURE}{\scriptsize \textbf{Task.} SaaS outage triage coordination --- coordinate archetype, incident commander assistant, Software/SaaS. Controls: autonomy=low-risk execute, knowledge burden=small-search retrieval, tool burden=moderate, risk=routine.\\ \textbf{Required.} produce the timeline work product \texttt{inc\_1} with 10 required fields (e.g.\ incident\_id=`OPS-19'; current\_impact=`API 5xx errors'; suspected\_service=`cart service'; blast\_radius=`US and EU regions'); retrieve the governing policy/record (knowledge retrieval via \texttt{kb\_search}); obtain user approval, then commit/submit. \textit{Twist:} the user corrects suspected\_service, blast\_radius, severity mid-utterance (barge-in), which the agent must catch and repair.\\ \textbf{Agent.} 12 tool-calls; retrieval \emph{none}; finalize \emph{not finalized}; approval \emph{not sought}.\\ \textbf{Outcome.} Workflow FAILURE --- failed gate(s): TS, PV, AC, AV. required knowledge retrieval not satisfied --- kb\_search never called; field(s) left stale after correction: suspected\_service; workflow not finalized --- never submitted/committed; artifact incomplete: 8/10 fields correct (lifecycle<COMMITTED(got DRAFT)); wrong/missing: suspected\_service (stale), status (missing).}\end{failbox}
\begin{failbox}{apexv1\_112 --- Access anomaly escalation coordination --- Workflow FAILURE}{\scriptsize \textbf{Task.} Access anomaly escalation coordination --- coordinate archetype, security operations coordinator, Software/SaaS. Controls: autonomy=low-risk execute, knowledge burden=multi-document reasoning, tool burden=moderate, risk=special review.\\ \textbf{Required.} produce the timeline work product \texttt{inc\_1} with 10 required fields (e.g.\ user\_account=`acct\_5521'; reported\_claim=`user says account compromised'; log\_evidence=`anomalous login from new location'; established\_fact=`anomalous login only, compromise n'); retrieve the governing policy/record (knowledge retrieval via \texttt{kb\_search}); obtain user approval, then commit/submit. \textit{Twist:} the user corrects established\_fact, reported\_claim mid-utterance (barge-in), which the agent must catch and repair.\\ \textbf{Agent.} 2 tool-calls; retrieval \emph{none}; finalize \emph{not finalized}; approval \emph{not sought}.\\ \textbf{Outcome.} Workflow FAILURE --- failed gate(s): TS, AC, AV. required knowledge retrieval not satisfied --- kb\_search never called; workflow not finalized --- never submitted/committed; artifact incomplete: 2/10 fields correct (lifecycle<COMMITTED(got DRAFT)); wrong/missing: reported\_claim (missing), established\_fact (missing), uncertainty\_note (missing), sop\_step (missing), +4 more.}\end{failbox}
\begin{failbox}{apexv1\_113 --- Warehouse shipment-delay incident --- Workflow FAILURE}{\scriptsize \textbf{Task.} Warehouse shipment-delay incident --- coordinate archetype, operations incident coordinator, manufacturing/field ops. Controls: autonomy=draft-and-confirm, knowledge burden=small-search retrieval, tool burden=moderate, risk=routine.\\ \textbf{Required.} produce the timeline work product \texttt{inc\_1} with 10 required fields (e.g.\ shipment\_id=`SHP-3300'; original\_eta=`Tuesday 6am'; carrier\_eta=`Tuesday 1pm'; customer\_cutoff=`Tuesday 3pm, cannot move'); retrieve the governing policy/record (knowledge retrieval via \texttt{kb\_search}); reach the required terminal state (\texttt{READY\_FOR\_REVIEW}). \textit{Twist:} the user corrects carrier\_eta, risk, recovery\_plan mid-utterance (barge-in), which the agent must catch and repair.\\ \textbf{Agent.} 5 tool-calls; retrieval \emph{none}; finalize \emph{not finalized}; approval n/a.\\ \textbf{Outcome.} Workflow FAILURE --- failed gate(s): AC, AV. required knowledge retrieval not satisfied --- kb\_search never called; artifact incomplete: 10/10 fields correct; wrong/missing: risk (stale).}\end{failbox}
\begin{failbox}{apexv1\_114 --- Customer data-sync incident escalation --- Workflow FAILURE}{\scriptsize \textbf{Task.} Customer data-sync incident escalation --- coordinate archetype, support incident coordinator, Software/SaaS. Controls: autonomy=low-risk execute, knowledge burden=small-search retrieval, tool burden=moderate, risk=routine.\\ \textbf{Required.} produce the timeline work product \texttt{inc\_1} with 11 required fields (e.g.\ account=`Delta Systems'; reported\_scope=`customer says all records affected'; telemetry\_scope=`telemetry shows one region only'; established\_scope=`one region confirmed by telemetry,'); retrieve the governing policy/record (knowledge retrieval via \texttt{kb\_search}); obtain user approval, then commit/submit. \textit{Twist:} the user corrects established\_scope, reported\_scope mid-utterance (barge-in), which the agent must catch and repair.\\ \textbf{Agent.} 6 tool-calls; retrieval done; finalize \emph{not finalized}; approval \emph{not sought}.\\ \textbf{Outcome.} Workflow FAILURE --- failed gate(s): TS, PV, AV. field(s) left stale after correction: reported\_scope, established\_scope; workflow not finalized --- never submitted/committed; artifact incomplete: 10/11 fields correct (lifecycle<COMMITTED(got DRAFT)); wrong/missing: reported\_scope (stale), established\_scope (stale), status (missing).}\end{failbox}
\begin{failbox}{apexv1\_115 --- Supply-chain component shortage response --- Workflow FAILURE}{\scriptsize \textbf{Task.} Supply-chain component shortage response --- coordinate archetype, supply-chain coordinator, manufacturing/field ops. Controls: autonomy=draft-and-confirm, knowledge burden=small-search retrieval, tool burden=moderate, risk=routine.\\ \textbf{Required.} produce the timeline work product \texttt{inc\_1} with 10 required fields (e.g.\ component=`power module PM-9'; shortage\_qty=`500'; supplier\_available=`200'; production\_priority=`Line A over Line B'); retrieve the governing policy/record (knowledge retrieval via \texttt{kb\_search}); reach the required terminal state (\texttt{READY\_FOR\_REVIEW}). \textit{Twist:} the user corrects allocation, recovery\_plan, supplier\_available mid-utterance (barge-in), which the agent must catch and repair.\\ \textbf{Agent.} 5 tool-calls; retrieval \emph{none}; finalize \emph{not finalized}; approval n/a.\\ \textbf{Outcome.} Workflow FAILURE --- failed gate(s): TS, AC, AV. required knowledge retrieval not satisfied --- kb\_search never called; workflow not finalized --- never submitted/committed; artifact incomplete: 4/10 fields correct (lifecycle<READY\_FOR\_REVIEW(got DRAFT)); wrong/missing: allocation (stale), expedite\_option (missing), recovery\_plan (stale), cost\_impact (missing), +2 more; 1 infra/WS drop(s).}\end{failbox}
\begin{failbox}{apexv1\_116 --- Facilities water-leak escalation --- Workflow FAILURE}{\scriptsize \textbf{Task.} Facilities water-leak escalation --- coordinate archetype, facilities incident coordinator, general enterprise. Controls: autonomy=approval-gated commit, knowledge burden=supplied evidence, tool burden=moderate, risk=special review.\\ \textbf{Required.} produce the timeline work product \texttt{inc\_1} with 11 required fields (e.g.\ location=`3rd floor east'; leak\_source=`burst pipe above ceiling'; initial\_priority=`standard cleanup'; electrical\_exposure=`water near a live panel'); retrieve the governing policy/record (knowledge retrieval via \texttt{kb\_search}); obtain user approval, then commit/submit. \textit{Twist:} the user corrects electrical\_exposure, escalated\_priority, initial\_priority mid-utterance (barge-in), which the agent must catch and repair.\\ \textbf{Agent.} 1 tool-calls; retrieval \emph{none}; finalize \emph{not finalized}; approval \emph{not sought}.\\ \textbf{Outcome.} Workflow FAILURE --- failed gate(s): TS, PV, AC, AV. required knowledge retrieval not satisfied --- kb\_search never called; field(s) left stale after correction: initial\_priority, electrical\_exposure, escalated\_priority; workflow not finalized --- never submitted/committed; artifact incomplete: 9/11 fields correct (lifecycle<COMMITTED(got DRAFT)); wrong/missing: initial\_priority (stale), electrical\_exposure (stale), escalated\_priority (stale), vendor\_dispatch (missing), +1 more.}\end{failbox}
\begin{failbox}{apexv1\_117 --- Payment-processing outage coordination --- Workflow FAILURE}{\scriptsize \textbf{Task.} Payment-processing outage coordination --- coordinate archetype, payments operations incident coordinator, Software/SaaS. Controls: autonomy=draft-and-confirm, knowledge burden=small-search retrieval, tool burden=moderate, risk=consequential action.\\ \textbf{Required.} produce the timeline work product \texttt{inc\_1} with 10 required fields (e.g.\ incident\_id=`PAY-88'; rail\_card=`card rail degraded'; rail\_ach=`ACH rail recovered'; overall\_status=`partial recovery, card still degra'); retrieve the governing policy/record (knowledge retrieval via \texttt{kb\_search}); reach the required terminal state (\texttt{READY\_FOR\_REVIEW}). \textit{Twist:} the user corrects overall\_status, rail\_ach, status\_wording mid-utterance (barge-in), which the agent must catch and repair.\\ \textbf{Agent.} 9 tool-calls; retrieval \emph{none}; finalize \emph{not finalized}; approval n/a.\\ \textbf{Outcome.} Workflow FAILURE --- failed gate(s): PV, AC, AV. required knowledge retrieval not satisfied --- kb\_search never called; field(s) left stale after correction: rail\_ach, overall\_status; artifact incomplete: 9/10 fields correct; wrong/missing: rail\_ach (stale), overall\_status (stale), full\_recovery (missing).}\end{failbox}
\begin{failbox}{apexv1\_118 --- Production quality hold coordination --- Workflow FAILURE}{\scriptsize \textbf{Task.} Production quality hold coordination --- coordinate archetype, manufacturing incident coordinator, manufacturing/field ops. Controls: autonomy=approval-gated commit, knowledge burden=small-search retrieval, tool burden=moderate, risk=consequential action.\\ \textbf{Required.} produce the timeline work product \texttt{inc\_1} with 11 required fields (e.g.\ incident\_id=`QH-14'; defect=`coating adhesion failure'; initial\_hold\_scope=`all lots this week'; test\_result=`only lots 5 to 8 affected'); retrieve the governing policy/record (knowledge retrieval via \texttt{kb\_search}); obtain user approval, then commit/submit. \textit{Twist:} the user corrects disposition, revised\_hold\_scope, test\_result mid-utterance (barge-in), which the agent must catch and repair.\\ \textbf{Agent.} 8 tool-calls; retrieval \emph{none}; finalize committed; approval \emph{not sought}.\\ \textbf{Outcome.} Workflow FAILURE --- failed gate(s): TS, PV, AC, AV. required knowledge retrieval not satisfied --- kb\_search never called; committed/submitted without seeking approval; field(s) left stale after correction: test\_result, revised\_hold\_scope; workflow not brought to terminal state (artifact not COMMITTED); artifact incomplete: 9/11 fields correct (lifecycle<COMMITTED(got DRAFT)); wrong/missing: initial\_hold\_scope (got `Lots 5 through 8' vs `all lots this week'), test\_result (stale), revised\_hold\_scope (stale), status (missing); tool-call failure: submit\_inc\_1.}\end{failbox}
\begin{failbox}{apexv1\_119 --- Live event AV failure coordination --- Workflow FAILURE}{\scriptsize \textbf{Task.} Live event AV failure coordination --- coordinate archetype, event operations coordinator, general enterprise. Controls: autonomy=draft-and-confirm, knowledge burden=none, tool burden=light, risk=routine.\\ \textbf{Required.} produce the timeline work product \texttt{inc\_1} with 10 required fields (e.g.\ event=`keynote session'; failure=`main room projector and audio down'; backup\_room=`Room B available'; backup\_capacity=`180'); reach the required terminal state (\texttt{READY\_FOR\_REVIEW}). \textit{Twist:} the user corrects vip\_constraint, vip\_handling mid-utterance (barge-in), which the agent must catch and repair.\\ \textbf{Agent.} 3 tool-calls; retrieval \emph{none}; finalize \emph{not finalized}; approval n/a.\\ \textbf{Outcome.} Workflow FAILURE --- failed gate(s): TS, AV. workflow not finalized --- never submitted/committed; artifact incomplete: 10/10 fields correct (lifecycle<READY\_FOR\_REVIEW(got DRAFT)); 1 infra/WS drop(s).}\end{failbox}
\begin{failbox}{apexv1\_120 --- Near-miss incident escalation and follow-up --- Workflow FAILURE}{\scriptsize \textbf{Task.} Near-miss incident escalation and follow-up --- coordinate archetype, safety operations coordinator, manufacturing/field ops. Controls: autonomy=low-risk execute, knowledge burden=small-search retrieval, tool burden=moderate, risk=special review.\\ \textbf{Required.} produce the timeline work product \texttt{inc\_1} with 11 required fields (e.g.\ incident\_id=`NM-77'; equipment\_id=`PRESS-7'; event=`guard bypass near-miss'; injury=`no injury'); retrieve the governing policy/record (knowledge retrieval via \texttt{kb\_search}); obtain user approval, then commit/submit. \textit{Twist:} the user corrects equipment\_id, corrected\_cause, draft\_cause mid-utterance (barge-in), which the agent must catch and repair.\\ \textbf{Agent.} 6 tool-calls; retrieval \emph{none}; finalize \emph{not finalized}; approval \emph{not sought}.\\ \textbf{Outcome.} Workflow FAILURE --- failed gate(s): TS, AC, AV. required knowledge retrieval not satisfied --- kb\_search never called; workflow not finalized --- never submitted/committed; artifact incomplete: 8/11 fields correct (lifecycle<COMMITTED(got DRAFT)); wrong/missing: draft\_cause (missing), corrected\_cause (missing), status (missing); 1 infra/WS drop(s).}\end{failbox}
\clearpage
\subsection*{GPT-live-1}
\begin{failbox}{apexv1\_001 --- Benefits enrollment with dependent correction --- Workflow FAILURE}{\scriptsize \textbf{Task.} Benefits enrollment with dependent correction --- form-fill archetype, benefits specialist, workplace/HR. Controls: autonomy=approval-gated commit, knowledge burden=small-search retrieval, tool burden=moderate, risk=consequential action.\\ \textbf{Required.} produce the structured form work product \texttt{enr\_1} with 14 required fields (e.g.\ employee\_id=`E4471'; legal\_name=`Morgan Reyes'; date\_of\_birth=`1988-07-09'; medical\_plan=`HDHP'); retrieve the governing policy/record (knowledge retrieval via \texttt{kb\_search}); obtain user approval, then commit/submit. \textit{Twist:} the user corrects dependent\_count, dependent\_names, medical\_plan, monthly\_premium mid-utterance (barge-in), which the agent must catch and repair.\\ \textbf{Agent.} 15 tool-calls; retrieval done; finalize committed; approval sought.\\ \textbf{Outcome.} Workflow FAILURE --- failed gate(s): AC, AV. required knowledge retrieval not satisfied --- searched but gold document not retrieved; artifact incomplete: 7/14 fields correct; wrong/missing: legal\_name (got `Morgan Rays' vs `Morgan Reyes'), medical\_plan (got `PPO' vs `HDHP'), dependent\_count (got `1' vs `2'), dependent\_names (got `Jamie Rays' vs `Jamie Reyes and Casey Reyes'), +3 more.}\end{failbox}
\begin{failbox}{apexv1\_002 --- Expense report from receipts and spoken narrative --- Workflow FAILURE}{\scriptsize \textbf{Task.} Expense report from receipts and spoken narrative --- form-fill archetype, finance operations specialist, general enterprise. Controls: autonomy=approval-gated commit, knowledge burden=supplied evidence, tool burden=moderate, risk=consequential action.\\ \textbf{Required.} produce the structured form work product \texttt{exp\_1} with 11 required fields (e.g.\ employee\_id=`E9910'; report\_period=`March 3 to March 6'; purpose=`client onsite in Denver'; airfare=`410.0'); retrieve the governing policy/record (knowledge retrieval via \texttt{kb\_search}); obtain user approval, then commit/submit. \textit{Twist:} the user corrects hotel, total, cost\_center mid-utterance (barge-in), which the agent must catch and repair.\\ \textbf{Agent.} 14 tool-calls; retrieval \emph{none}; finalize committed; approval sought.\\ \textbf{Outcome.} Workflow FAILURE --- failed gate(s): TS, PV, AC, AV. required knowledge retrieval not satisfied --- kb\_search never called; process-compliance/critical gate violated (no\_unauthorized\_commit, artifact\_not\_stale, artifact\_not\_stale); workflow not brought to terminal state (artifact not COMMITTED); artifact incomplete: 7/11 fields correct (lifecycle<COMMITTED(got DRAFT)); wrong/missing: hotel (got `540' vs `480.0'), cost\_center (got `CC4402' vs `CC-4419'), total (got `1145' vs `1085.0'), status (missing); tool-call failure: submit\_exp\_1.}\end{failbox}
\begin{failbox}{apexv1\_003 --- New vendor onboarding packet --- Workflow FAILURE}{\scriptsize \textbf{Task.} New vendor onboarding packet --- form-fill archetype, procurement operations specialist, general enterprise. Controls: autonomy=draft-and-confirm, knowledge burden=small-search retrieval, tool burden=moderate, risk=sensitive-data simulation.\\ \textbf{Required.} produce the structured form work product \texttt{ven\_1} with 10 required fields (e.g.\ vendor\_name=`Cedar Works LLC'; tax\_id=`88-4412290'; address=`72 Mill Road, Suite 4'; remittance\_email=`billing@cedarworks.example'); retrieve the governing policy/record (knowledge retrieval via \texttt{kb\_search}); reach the required terminal state (\texttt{READY\_FOR\_REVIEW}). \textit{Twist:} the user corrects remittance\_email, account\_number mid-utterance (barge-in), which the agent must catch and repair.\\ \textbf{Agent.} 12 tool-calls; retrieval \emph{none}; finalize \emph{not finalized}; approval n/a.\\ \textbf{Outcome.} Workflow FAILURE --- failed gate(s): AC. required knowledge retrieval not satisfied --- kb\_search never called.}\end{failbox}
\begin{failbox}{apexv1\_004 --- Business travel approval request --- Workflow FAILURE}{\scriptsize \textbf{Task.} Business travel approval request --- form-fill archetype, travel coordinator, general enterprise. Controls: autonomy=draft-and-confirm, knowledge burden=small-search retrieval, tool burden=moderate, risk=routine.\\ \textbf{Required.} produce the structured form work product \texttt{trv\_1} with 10 required fields (e.g.\ traveler=`Priya Nair'; destination=`Austin'; purpose=`customer quarterly review'; meeting\_date=`2026-04-15'); retrieve the governing policy/record (knowledge retrieval via \texttt{kb\_search}); reach the required terminal state (\texttt{READY\_FOR\_REVIEW}). \textit{Twist:} the user corrects depart\_date, meeting\_date, return\_date mid-utterance (barge-in), which the agent must catch and repair.\\ \textbf{Agent.} 11 tool-calls; retrieval done; finalize \emph{not finalized}; approval n/a.\\ \textbf{Outcome.} Workflow FAILURE --- failed gate(s): AV. artifact incomplete: 1/10 fields correct; wrong/missing: traveler (missing), destination (missing), purpose (missing), meeting\_date (missing), +5 more.}\end{failbox}
\begin{failbox}{apexv1\_005 --- Privileged software-access request --- Workflow FAILURE}{\scriptsize \textbf{Task.} Privileged software-access request --- form-fill archetype, IT access coordinator, Software/SaaS. Controls: autonomy=approval-gated commit, knowledge burden=multi-document reasoning, tool burden=moderate, risk=consequential action.\\ \textbf{Required.} produce the structured form work product \texttt{acc\_1} with 10 required fields (e.g.\ requester=`Sam Okafor'; project=`Q2 revenue analytics'; requested\_access=`analytics read-only'; justified\_role=`AnalyticsViewer'); retrieve the governing policy/record (knowledge retrieval via \texttt{kb\_search}); obtain user approval, then commit/submit. \textit{Twist:} the user corrects requested\_access, requested\_access, duration mid-utterance (barge-in), which the agent must catch and repair.\\ \textbf{Agent.} 14 tool-calls; retrieval done; finalize \emph{not finalized}; approval sought.\\ \textbf{Outcome.} Workflow FAILURE --- failed gate(s): TS, AV. workflow not finalized --- never submitted/committed; artifact incomplete: 1/10 fields correct (lifecycle<COMMITTED(got DRAFT)); wrong/missing: project (missing), requested\_access (missing), justified\_role (missing), business\_justification (missing), +5 more.}\end{failbox}
\begin{failbox}{apexv1\_006 --- Warranty claim application --- Workflow FAILURE}{\scriptsize \textbf{Task.} Warranty claim application --- form-fill archetype, warranty operations specialist, manufacturing/field ops. Controls: autonomy=draft-and-confirm, knowledge burden=small-search retrieval, tool burden=moderate, risk=routine.\\ \textbf{Required.} produce the structured form work product \texttt{war\_1} with 9 required fields (e.g.\ customer\_name=`Robin Vale'; product\_model=`TurboMix 500'; serial\_number=`TMX500-88231'; purchase\_date=`2025-11-02'); retrieve the governing policy/record (knowledge retrieval via \texttt{kb\_search}); reach the required terminal state (\texttt{READY\_FOR\_REVIEW}). \textit{Twist:} the user corrects serial\_number, warranty\_policy mid-utterance (barge-in), which the agent must catch and repair.\\ \textbf{Agent.} 9 tool-calls; retrieval done; finalize \emph{not finalized}; approval n/a.\\ \textbf{Outcome.} Workflow FAILURE --- failed gate(s): AV. artifact incomplete: 8/9 fields correct; wrong/missing: serial\_number (got `TMX 500-88213' vs `TMX500-88231').}\end{failbox}
\begin{failbox}{apexv1\_007 --- Parental-leave administration packet --- Workflow FAILURE}{\scriptsize \textbf{Task.} Parental-leave administration packet --- form-fill archetype, HR operations specialist, workplace/HR. Controls: autonomy=draft-and-confirm, knowledge burden=multi-document reasoning, tool burden=moderate, risk=sensitive-data simulation.\\ \textbf{Required.} produce the structured form work product \texttt{lev\_1} with 9 required fields (e.g.\ employee\_id=`E3320'; leave\_type=`parental leave'; leave\_start=`2026-05-01'; leave\_end=`2026-07-31'); retrieve the governing policy/record (knowledge retrieval via \texttt{kb\_search}); reach the required terminal state (\texttt{READY\_FOR\_REVIEW}). \textit{Twist:} the user corrects fmla\_weeks, leave\_end mid-utterance (barge-in), which the agent must catch and repair.\\ \textbf{Agent.} 13 tool-calls; retrieval done; finalize \emph{not finalized}; approval n/a.\\ \textbf{Outcome.} Workflow FAILURE --- failed gate(s): AC, AV. required knowledge retrieval not satisfied --- searched but gold document not retrieved; artifact incomplete: 6/9 fields correct; wrong/missing: leave\_end (got `2026-07-24' vs `2026-07-31'), fmla\_weeks (got `12' vs `13'), medical\_details (missing).}\end{failbox}
\begin{failbox}{apexv1\_008 --- Customer account setup and billing profile --- Workflow FAILURE}{\scriptsize \textbf{Task.} Customer account setup and billing profile --- form-fill archetype, account operations specialist, Software/SaaS. Controls: autonomy=draft-and-confirm, knowledge burden=supplied evidence, tool burden=moderate, risk=sensitive-data simulation.\\ \textbf{Required.} produce the structured form work product \texttt{acct\_1} with 9 required fields (e.g.\ company=`Northwind Retail'; billing\_contact=`Ada Lin'; technical\_contact=`Ben Cho'; billing\_country=`Germany'); retrieve the governing policy/record (knowledge retrieval via \texttt{kb\_search}); reach the required terminal state (\texttt{READY\_FOR\_REVIEW}). \textit{Twist:} the user corrects billing\_country, tax\_id, tax\_rate mid-utterance (barge-in), which the agent must catch and repair.\\ \textbf{Agent.} 8 tool-calls; retrieval done; finalize \emph{not finalized}; approval n/a.\\ \textbf{Outcome.} Workflow FAILURE --- failed gate(s): TS, AC, AV. required knowledge retrieval not satisfied --- searched but gold document not retrieved; workflow not finalized --- never submitted/committed; artifact incomplete: 6/9 fields correct (lifecycle<READY\_FOR\_REVIEW(got DRAFT)); wrong/missing: billing\_country (got `Ireland' vs `Germany'), tax\_id (got `IE1234567X' vs `DE811234567'), tax\_rate (got `23\%' vs `19.0').}\end{failbox}
\begin{failbox}{apexv1\_009 --- Conference reimbursement packet --- Workflow FAILURE}{\scriptsize \textbf{Task.} Conference reimbursement packet --- form-fill archetype, operations coordinator, professional services. Controls: autonomy=draft-and-confirm, knowledge burden=supplied evidence, tool burden=moderate, risk=routine.\\ \textbf{Required.} produce the structured form work product \texttt{rmb\_1} with 9 required fields (e.g.\ attendee=`Noa Grant'; conference=`DataCon 2026'; registration\_fee=`300.0'; workshop\_fee=`175.0'); retrieve the governing policy/record (knowledge retrieval via \texttt{kb\_search}); reach the required terminal state (\texttt{READY\_FOR\_REVIEW}). \textit{Twist:} the user corrects total, workshop\_fee mid-utterance (barge-in), which the agent must catch and repair.\\ \textbf{Agent.} 10 tool-calls; retrieval done; finalize \emph{not finalized}; approval n/a.\\ \textbf{Outcome.} Workflow FAILURE --- failed gate(s): AC, AV. required knowledge retrieval not satisfied --- searched but gold document not retrieved; artifact incomplete: 5/9 fields correct; wrong/missing: workshop\_fee (got `150' vs `175.0'), hotel (got `730' vs `220.0'), total (got `1240' vs `755.0'), receipt\_status (missing).}\end{failbox}
\begin{passbox}{apexv1\_010 --- Facility access badge request --- Workflow SUCCESS}{\scriptsize \textbf{Task.} Facility access badge request --- form-fill archetype, facilities coordinator, general enterprise. Controls: autonomy=approval-gated commit, knowledge burden=small-search retrieval, tool burden=moderate, risk=consequential action.\\ \textbf{Required.} produce the structured form work product \texttt{bdg\_1} with 10 required fields (e.g.\ contractor\_name=`Rowan Tate'; company=`BrightHVAC'; sponsor=`Facilities lead Dana'; access\_zones=`mechanical rooms'); retrieve the governing policy/record (knowledge retrieval via \texttt{kb\_search}); obtain user approval, then commit/submit. \textit{Twist:} the user corrects requested\_hours, requested\_hours mid-utterance (barge-in), which the agent must catch and repair.\\ \textbf{Agent.} 13 tool-calls; retrieval done; finalize committed; approval sought.\\ \textbf{Outcome.} Workflow SUCCESS --- all gates pass; artifact field accuracy 10/10.}\end{passbox}
\begin{failbox}{apexv1\_011 --- Software engineer recruiter screen --- Workflow FAILURE}{\scriptsize \textbf{Task.} Software engineer recruiter screen --- interview archetype, recruiter, Software/SaaS. Controls: autonomy=prepare-only, knowledge burden=none, tool burden=light, risk=sensitive-data simulation.\\ \textbf{Required.} produce the evidence matrix work product \texttt{rec\_1} with 13 required fields (e.g.\ years\_experience=`8'; primary\_language=`Python'; system\_design\_example=`designed a multi-region ingestion '; scale\_metric=`half a million daily events'); reach the required terminal state (\texttt{READY\_FOR\_REVIEW}). \textit{Twist:} the user corrects team\_size, scale\_metric mid-utterance (barge-in), which the agent must catch and repair.\\ \textbf{Agent.} 19 tool-calls; retrieval \emph{none}; finalize \emph{not finalized}; approval n/a.\\ \textbf{Outcome.} Workflow FAILURE --- failed gate(s): AV. artifact incomplete: 11/13 fields correct; wrong/missing: scale\_metric (got `one million daily events' vs `half a million daily events'), team\_size (got `eight people' vs `6').}\end{failbox}
\begin{failbox}{apexv1\_012 --- Customer-success manager recruiter screen --- Workflow FAILURE}{\scriptsize \textbf{Task.} Customer-success manager recruiter screen --- interview archetype, recruiter, Software/SaaS. Controls: autonomy=prepare-only, knowledge burden=none, tool burden=light, risk=sensitive-data simulation.\\ \textbf{Required.} produce the evidence matrix work product \texttt{rec\_1} with 12 required fields (e.g.\ years\_experience=`6'; book\_of\_business=`20 enterprise accounts'; retention\_metric=`92 percent gross retention'; customer\_save\_example=`recovered a churning key account'); reach the required terminal state (\texttt{READY\_FOR\_REVIEW}). \textit{Twist:} the user corrects retention\_metric, open\_gap mid-utterance (barge-in), which the agent must catch and repair.\\ \textbf{Agent.} 17 tool-calls; retrieval \emph{none}; finalize \emph{not finalized}; approval n/a.\\ \textbf{Outcome.} Workflow FAILURE --- failed gate(s): AV. artifact incomplete: 11/12 fields correct; wrong/missing: open\_gap (got `Lack of specific quota carrying experien' vs `quota-carrying confirmed via renewals').}\end{failbox}
\begin{failbox}{apexv1\_013 --- Warehouse supervisor screen --- Workflow FAILURE}{\scriptsize \textbf{Task.} Warehouse supervisor screen --- interview archetype, recruiter, manufacturing/field ops. Controls: autonomy=prepare-only, knowledge burden=none, tool burden=light, risk=sensitive-data simulation.\\ \textbf{Required.} produce the evidence matrix work product \texttt{rec\_1} with 12 required fields (e.g.\ years\_experience=`10'; team\_size=`25'; shift\_scheduling=`built rotating three-shift coverag'; safety\_record=`300 days incident-free'); reach the required terminal state (\texttt{READY\_FOR\_REVIEW}). \textit{Twist:} the user corrects current\_start\_year, throughput\_metric mid-utterance (barge-in), which the agent must catch and repair.\\ \textbf{Agent.} 21 tool-calls; retrieval \emph{none}; finalize \emph{not finalized}; approval n/a.\\ \textbf{Outcome.} Workflow FAILURE --- failed gate(s): AV. artifact incomplete: 1/12 fields correct; wrong/missing: years\_experience (missing), team\_size (missing), shift\_scheduling (missing), safety\_record (missing), +7 more.}\end{failbox}
\begin{failbox}{apexv1\_014 --- Internal transfer evidence interview --- Workflow FAILURE}{\scriptsize \textbf{Task.} Internal transfer evidence interview --- interview archetype, HR business partner, general enterprise. Controls: autonomy=prepare-only, knowledge burden=supplied evidence, tool burden=moderate, risk=sensitive-data simulation.\\ \textbf{Required.} produce the evidence matrix work product \texttt{rec\_1} with 12 required fields (e.g.\ current\_role=`senior analyst'; target\_team=`platform reliability'; project\_apollo=`led the payments platform migratio'; transferable\_skill=`incident command'); retrieve the governing policy/record (knowledge retrieval via \texttt{kb\_search}); reach the required terminal state (\texttt{READY\_FOR\_REVIEW}). \textit{Twist:} the user corrects project\_apollo, impact\_metric mid-utterance (barge-in), which the agent must catch and repair.\\ \textbf{Agent.} 12 tool-calls; retrieval done; finalize \emph{not finalized}; approval n/a.\\ \textbf{Outcome.} Workflow FAILURE --- failed gate(s): AC, AV. required knowledge retrieval not satisfied --- searched but gold document not retrieved; artifact incomplete: 10/12 fields correct; wrong/missing: gap\_area (got `Formal people management experience' vs `no formal people management'), motivation (got `Interested in contributing to the team' vs `want deeper systems work').}\end{failbox}
\begin{failbox}{apexv1\_015 --- Professional reference check --- Workflow FAILURE}{\scriptsize \textbf{Task.} Professional reference check --- interview archetype, recruiting coordinator, workplace/HR. Controls: autonomy=prepare-only, knowledge burden=none, tool burden=light, risk=sensitive-data simulation.\\ \textbf{Required.} produce the evidence matrix work product \texttt{ref\_1} with 12 required fields (e.g.\ relationship=`former direct manager'; years\_known=`3'; direct\_reports=`6'; dotted\_line\_reports=`6'); reach the required terminal state (\texttt{READY\_FOR\_REVIEW}). \textit{Twist:} the user corrects direct\_reports, dotted\_line\_reports mid-utterance (barge-in), which the agent must catch and repair.\\ \textbf{Agent.} 9 tool-calls; retrieval \emph{none}; finalize \emph{not finalized}; approval n/a.\\ \textbf{Outcome.} Workflow FAILURE --- failed gate(s): TS, AV. workflow not finalized --- never submitted/committed; artifact incomplete: 8/12 fields correct (lifecycle<READY\_FOR\_REVIEW(got DRAFT)); wrong/missing: strengths (got `reliability' vs `calm under pressure and organized'), notable\_achievement (got `unknown' vs `turned around a failing project'), integrity\_flag (got `unknown' vs `no concerns'), overall\_rating (got `unknown' vs `strong recommend').}\end{failbox}
\begin{failbox}{apexv1\_016 --- Returnship program screening interview --- Workflow FAILURE}{\scriptsize \textbf{Task.} Returnship program screening interview --- interview archetype, recruiter, Software/SaaS. Controls: autonomy=prepare-only, knowledge burden=none, tool burden=light, risk=special review.\\ \textbf{Required.} produce the evidence matrix work product \texttt{rec\_1} with 11 required fields (e.g.\ prior\_role=`backend engineer'; years\_experience=`7'; break\_length=`two years'; refresh\_activity=`completed a cloud and a security c'); reach the required terminal state (\texttt{READY\_FOR\_REVIEW}). \textit{Twist:} the user corrects refresh\_activity, target\_role mid-utterance (barge-in), which the agent must catch and repair.\\ \textbf{Agent.} 18 tool-calls; retrieval \emph{none}; finalize \emph{not finalized}; approval n/a.\\ \textbf{Outcome.} Workflow FAILURE --- failed gate(s): AV. artifact incomplete: 1/11 fields correct; wrong/missing: prior\_role (missing), years\_experience (missing), break\_length (missing), refresh\_activity (missing), +6 more.}\end{failbox}
\begin{failbox}{apexv1\_017 --- Contractor qualification call --- Workflow FAILURE}{\scriptsize \textbf{Task.} Contractor qualification call --- interview archetype, vendor workforce coordinator, professional services. Controls: autonomy=prepare-only, knowledge burden=none, tool burden=light, risk=sensitive-data simulation.\\ \textbf{Required.} produce the evidence matrix work product \texttt{qual\_1} with 12 required fields (e.g.\ specialty=`data engineering'; years\_experience=`9'; availability\_hours=`25 hours per week'; overlapping\_contracts=`two active engagements'); reach the required terminal state (\texttt{READY\_FOR\_REVIEW}). \textit{Twist:} the user corrects availability\_hours, rate mid-utterance (barge-in), which the agent must catch and repair.\\ \textbf{Agent.} 15 tool-calls; retrieval \emph{none}; finalize \emph{not finalized}; approval n/a.\\ \textbf{Outcome.} Workflow FAILURE --- failed gate(s): AV. artifact incomplete: 10/12 fields correct; wrong/missing: availability\_hours (got `20 hours/week' vs `25 hours per week'), rate (got `145' vs `155.0').}\end{failbox}
\begin{failbox}{apexv1\_018 --- Internship behavioral screen --- Workflow FAILURE}{\scriptsize \textbf{Task.} Internship behavioral screen --- interview archetype, campus recruiter, Software/SaaS. Controls: autonomy=prepare-only, knowledge burden=none, tool burden=light, risk=sensitive-data simulation.\\ \textbf{Required.} produce the evidence matrix work product \texttt{rec\_1} with 11 required fields (e.g.\ school=`state university'; major=`computer science'; grad\_year=`2027'; project\_example=`led a hackathon-winning logistics '); reach the required terminal state (\texttt{READY\_FOR\_REVIEW}). \textit{Twist:} the user corrects project\_example, open\_gap mid-utterance (barge-in), which the agent must catch and repair.\\ \textbf{Agent.} 13 tool-calls; retrieval \emph{none}; finalize \emph{not finalized}; approval n/a.\\ \textbf{Outcome.} Workflow FAILURE --- failed gate(s): AV. artifact incomplete: 1/11 fields correct; wrong/missing: school (missing), major (missing), grad\_year (missing), project\_example (missing), +6 more.}\end{failbox}
\begin{failbox}{apexv1\_019 --- Operations analyst screening with resume discrepancy --- Workflow FAILURE}{\scriptsize \textbf{Task.} Operations analyst screening with resume discrepancy --- interview archetype, recruiter, general enterprise. Controls: autonomy=prepare-only, knowledge burden=none, tool burden=light, risk=sensitive-data simulation.\\ \textbf{Required.} produce the evidence matrix work product \texttt{rec\_1} with 12 required fields (e.g.\ years\_experience=`5'; current\_start\_year=`2022'; resume\_discrepancy=`candidate confirms 2022, resume ty'; tools=`SQL and Tableau'); reach the required terminal state (\texttt{READY\_FOR\_REVIEW}). \textit{Twist:} the user corrects resume\_discrepancy mid-utterance (barge-in), which the agent must catch and repair.\\ \textbf{Agent.} 15 tool-calls; retrieval \emph{none}; finalize \emph{not finalized}; approval n/a.\\ \textbf{Outcome.} Workflow FAILURE --- failed gate(s): AV. artifact incomplete: 1/12 fields correct; wrong/missing: years\_experience (missing), current\_start\_year (missing), resume\_discrepancy (missing), tools (missing), +7 more.}\end{failbox}
\begin{failbox}{apexv1\_020 --- Interview debrief reconstruction after correction --- Workflow FAILURE}{\scriptsize \textbf{Task.} Interview debrief reconstruction after correction --- interview archetype, recruiting operations specialist, workplace/HR. Controls: autonomy=prepare-only, knowledge burden=none, tool burden=light, risk=sensitive-data simulation.\\ \textbf{Required.} produce the evidence matrix work product \texttt{rec\_1} with 11 required fields (e.g.\ candidate=`Jordan Ellis'; role=`senior QA engineer'; panel\_recommendation=`hire'; technical\_score=`5'); reach the required terminal state (\texttt{READY\_FOR\_REVIEW}). \textit{Twist:} the user corrects technical\_score, concern\_noted mid-utterance (barge-in), which the agent must catch and repair.\\ \textbf{Agent.} 21 tool-calls; retrieval \emph{none}; finalize \emph{not finalized}; approval n/a.\\ \textbf{Outcome.} Workflow FAILURE --- failed gate(s): AV. artifact incomplete: 8/11 fields correct; wrong/missing: technical\_score (got `4' vs `5'), concern\_noted (got `Concern about depth of performance testi' vs `performance-testing depth confirmed adeq'), record\_status (missing).}\end{failbox}
\begin{failbox}{apexv1\_021 --- Analytics-platform discovery call --- Workflow FAILURE}{\scriptsize \textbf{Task.} Analytics-platform discovery call --- discovery archetype, sales development representative, Software/SaaS. Controls: autonomy=draft-and-confirm, knowledge burden=none, tool burden=light, risk=routine.\\ \textbf{Required.} produce the CRM record work product \texttt{crm\_1} with 11 required fields (e.g.\ company=`Glacier Foods'; industry=`food distribution'; current\_tool=`spreadsheets'; pain\_point=`slow monthly reporting'); reach the required terminal state (\texttt{READY\_FOR\_REVIEW}). \textit{Twist:} the user corrects licensed\_users, viewer\_users, timeline mid-utterance (barge-in), which the agent must catch and repair.\\ \textbf{Agent.} 16 tool-calls; retrieval \emph{none}; finalize \emph{not finalized}; approval n/a.\\ \textbf{Outcome.} Workflow FAILURE --- failed gate(s): AV. artifact incomplete: 8/11 fields correct; wrong/missing: licensed\_users (got `200' vs `120'), viewer\_users (got `0' vs `80'), timeline (got `this quarter' vs `next quarter').}\end{failbox}
\begin{passbox}{apexv1\_022 --- Cybersecurity expansion discovery --- Workflow SUCCESS}{\scriptsize \textbf{Task.} Cybersecurity expansion discovery --- discovery archetype, account executive, Software/SaaS. Controls: autonomy=draft-and-confirm, knowledge burden=small-search retrieval, tool burden=moderate, risk=routine.\\ \textbf{Required.} produce the CRM record work product \texttt{crm\_1} with 10 required fields (e.g.\ company=`Meridian Bank'; current\_modules=`endpoint protection'; desired\_modules=`cloud posture and identity protect'; compliance\_need=`PCI DSS and SOC2'); retrieve the governing policy/record (knowledge retrieval via \texttt{kb\_search}); reach the required terminal state (\texttt{READY\_FOR\_REVIEW}). \textit{Twist:} the user corrects desired\_modules, compliance\_need mid-utterance (barge-in), which the agent must catch and repair.\\ \textbf{Agent.} 13 tool-calls; retrieval done; finalize \emph{not finalized}; approval n/a.\\ \textbf{Outcome.} Workflow SUCCESS --- all gates pass; artifact field accuracy 10/10.}\end{passbox}
\begin{failbox}{apexv1\_023 --- Manufacturing automation discovery --- Workflow FAILURE}{\scriptsize \textbf{Task.} Manufacturing automation discovery --- discovery archetype, solutions consultant, manufacturing/field ops. Controls: autonomy=prepare-only, knowledge burden=none, tool burden=light, risk=routine.\\ \textbf{Required.} produce the CRM record work product \texttt{crm\_1} with 11 required fields (e.g.\ company=`Ironside Manufacturing'; lines\_total=`3'; lines\_in\_scope=`2'; in\_scope\_detail=`packaging and labeling lines'); reach the required terminal state (\texttt{READY\_FOR\_REVIEW}). \textit{Twist:} the user corrects in\_scope\_detail, lines\_in\_scope, throughput\_goal mid-utterance (barge-in), which the agent must catch and repair.\\ \textbf{Agent.} 14 tool-calls; retrieval \emph{none}; finalize \emph{not finalized}; approval n/a.\\ \textbf{Outcome.} Workflow FAILURE --- failed gate(s): AV. artifact incomplete: 1/11 fields correct; wrong/missing: company (missing), lines\_total (missing), lines\_in\_scope (missing), in\_scope\_detail (missing), +6 more.}\end{failbox}
\begin{failbox}{apexv1\_024 --- Healthcare operations software discovery --- Workflow FAILURE}{\scriptsize \textbf{Task.} Healthcare operations software discovery --- discovery archetype, account executive, healthcare. Controls: autonomy=prepare-only, knowledge burden=none, tool burden=light, risk=sensitive-data simulation.\\ \textbf{Required.} produce the CRM record work product \texttt{crm\_1} with 11 required fields (e.g.\ organization=`Riverside Clinics'; clinics=`6'; workflow\_pain=`manual appointment and billing rec'; staff\_count=`120'); reach the required terminal state (\texttt{READY\_FOR\_REVIEW}). \textit{Twist:} the user corrects clinics, workflow\_pain mid-utterance (barge-in), which the agent must catch and repair.\\ \textbf{Agent.} 13 tool-calls; retrieval \emph{none}; finalize \emph{not finalized}; approval n/a.\\ \textbf{Outcome.} Workflow FAILURE --- failed gate(s): AV. artifact incomplete: 10/11 fields correct; wrong/missing: clinics (got `eight' vs `6').}\end{failbox}
\begin{failbox}{apexv1\_025 --- Professional-services scoping call --- Workflow FAILURE}{\scriptsize \textbf{Task.} Professional-services scoping call --- discovery archetype, engagement manager, professional services. Controls: autonomy=prepare-only, knowledge burden=none, tool burden=light, risk=routine.\\ \textbf{Required.} produce the CRM record work product \texttt{crm\_1} with 10 required fields (e.g.\ client=`Baytown Retail'; workstream\_1=`data warehouse buildout'; workstream\_2=`add a BI dashboard workstream'; deliverables=`warehouse, ETL pipelines, and BI d'); reach the required terminal state (\texttt{READY\_FOR\_REVIEW}). \textit{Twist:} the user corrects deliverables, duration, workstream\_2 mid-utterance (barge-in), which the agent must catch and repair.\\ \textbf{Agent.} 13 tool-calls; retrieval \emph{none}; finalize \emph{not finalized}; approval n/a.\\ \textbf{Outcome.} Workflow FAILURE --- failed gate(s): AV. artifact incomplete: 1/10 fields correct; wrong/missing: client (missing), workstream\_1 (missing), workstream\_2 (missing), deliverables (missing), +5 more.}\end{failbox}
\begin{failbox}{apexv1\_026 --- CRM migration discovery --- Workflow FAILURE}{\scriptsize \textbf{Task.} CRM migration discovery --- discovery archetype, solutions consultant, Software/SaaS. Controls: autonomy=prepare-only, knowledge burden=small-search retrieval, tool burden=moderate, risk=routine.\\ \textbf{Required.} produce the CRM record work product \texttt{crm\_1} with 10 required fields (e.g.\ company=`Halcyon Media'; source\_crm=`legacy on-prem CRM'; record\_count=`eight hundred thousand records'; data\_retention=`seven years'); retrieve the governing policy/record (knowledge retrieval via \texttt{kb\_search}); reach the required terminal state (\texttt{READY\_FOR\_REVIEW}). \textit{Twist:} the user corrects data\_retention, record\_count, integration\_count mid-utterance (barge-in), which the agent must catch and repair.\\ \textbf{Agent.} 14 tool-calls; retrieval done; finalize \emph{not finalized}; approval n/a.\\ \textbf{Outcome.} Workflow FAILURE --- failed gate(s): AC, AV. required knowledge retrieval not satisfied --- searched but gold document not retrieved; artifact incomplete: 9/10 fields correct; wrong/missing: integration\_count (got `4' vs `6').}\end{failbox}
\begin{failbox}{apexv1\_027 --- Customer data-platform qualification --- Workflow FAILURE}{\scriptsize \textbf{Task.} Customer data-platform qualification --- discovery archetype, sales development representative, Software/SaaS. Controls: autonomy=prepare-only, knowledge burden=none, tool burden=light, risk=routine.\\ \textbf{Required.} produce the CRM record work product \texttt{crm\_1} with 10 required fields (e.g.\ company=`Pace Retail'; use\_case=`unify web and store data'; data\_sources=`web, POS, email, and mobile app'; volume=`50 million events monthly'); reach the required terminal state (\texttt{READY\_FOR\_REVIEW}). \textit{Twist:} the user corrects timeline, data\_sources mid-utterance (barge-in), which the agent must catch and repair.\\ \textbf{Agent.} 7 tool-calls; retrieval \emph{none}; finalize \emph{not finalized}; approval n/a.\\ \textbf{Outcome.} Workflow FAILURE --- failed gate(s): AV. artifact incomplete: 8/10 fields correct; wrong/missing: use\_case (got `modern and efficient shopping experience' vs `unify web and store data'), optional\_team\_size (missing).}\end{failbox}
\begin{failbox}{apexv1\_028 --- Renewal expansion discovery --- Workflow FAILURE}{\scriptsize \textbf{Task.} Renewal expansion discovery --- discovery archetype, customer success manager, Software/SaaS. Controls: autonomy=prepare-only, knowledge burden=small-search retrieval, tool burden=moderate, risk=routine.\\ \textbf{Required.} produce the CRM record work product \texttt{crm\_1} with 10 required fields (e.g.\ account=`Summit Logistics'; current\_plan=`Business tier'; complaint=`reporting is slow'; intent=`renew and expand across two teams'); retrieve the governing policy/record (knowledge retrieval via \texttt{kb\_search}); reach the required terminal state (\texttt{READY\_FOR\_REVIEW}). \textit{Twist:} the user corrects expansion\_seats, intent mid-utterance (barge-in), which the agent must catch and repair.\\ \textbf{Agent.} 14 tool-calls; retrieval done; finalize \emph{not finalized}; approval n/a.\\ \textbf{Outcome.} Workflow FAILURE --- failed gate(s): AV. artifact incomplete: 1/10 fields correct; wrong/missing: account (missing), current\_plan (missing), complaint (missing), intent (missing), +5 more.}\end{failbox}
\begin{failbox}{apexv1\_029 --- Channel-partner opportunity discovery --- Workflow FAILURE}{\scriptsize \textbf{Task.} Channel-partner opportunity discovery --- discovery archetype, partner manager, Software/SaaS. Controls: autonomy=prepare-only, knowledge burden=small-search retrieval, tool burden=moderate, risk=routine.\\ \textbf{Required.} produce the CRM record work product \texttt{crm\_1} with 10 required fields (e.g.\ partner=`BlueSky Resellers'; end\_customer=`Trilliant Co'; program\_tier=`Premier partner'; deal\_size=`75000'); retrieve the governing policy/record (knowledge retrieval via \texttt{kb\_search}); reach the required terminal state (\texttt{READY\_FOR\_REVIEW}). \textit{Twist:} the user corrects program\_tier, deal\_size mid-utterance (barge-in), which the agent must catch and repair.\\ \textbf{Agent.} 10 tool-calls; retrieval done; finalize \emph{not finalized}; approval n/a.\\ \textbf{Outcome.} Workflow FAILURE --- failed gate(s): AV. artifact incomplete: 7/10 fields correct; wrong/missing: program\_tier (got `Gold' vs `Premier partner'), deal\_size (got `60000' vs `75000'), end\_customer\_size (missing).}\end{failbox}
\begin{failbox}{apexv1\_030 --- Discovery call to CRM plus follow-up package --- Workflow FAILURE}{\scriptsize \textbf{Task.} Discovery call to CRM plus follow-up package --- discovery archetype, account executive, Software/SaaS. Controls: autonomy=draft-and-confirm, knowledge burden=small-search retrieval, tool burden=moderate, risk=routine.\\ \textbf{Required.} produce the CRM record work product \texttt{crm\_1} with 11 required fields (e.g.\ company=`Vertex Labs'; pain\_point=`manual lead routing'; use\_case=`automate routing and scoring'; seats=`45'); retrieve the governing policy/record (knowledge retrieval via \texttt{kb\_search}); reach the required terminal state (\texttt{READY\_FOR\_REVIEW}). \textit{Twist:} the user corrects rollout\_month, tentative\_idea mid-utterance (barge-in), which the agent must catch and repair.\\ \textbf{Agent.} 20 tool-calls; retrieval \emph{none}; finalize \emph{not finalized}; approval n/a.\\ \textbf{Outcome.} Workflow FAILURE --- failed gate(s): AC, AV. required knowledge retrieval not satisfied --- kb\_search never called; artifact incomplete: 1/11 fields correct; wrong/missing: company (missing), pain\_point (missing), use\_case (missing), seats (missing), +6 more.}\end{failbox}
\begin{failbox}{apexv1\_031 --- Insurance first notice of loss --- Workflow FAILURE}{\scriptsize \textbf{Task.} Insurance first notice of loss --- intake archetype, claims intake specialist, insurance. Controls: autonomy=draft-and-confirm, knowledge burden=small-search retrieval, tool burden=moderate, risk=sensitive-data simulation.\\ \textbf{Required.} produce the case record work product \texttt{claim\_1} with 11 required fields (e.g.\ policy\_number=`PN-5521'; insured\_name=`Jordan Park'; loss\_date=`2026-03-02'; loss\_time=`around 8am'); retrieve the governing policy/record (knowledge retrieval via \texttt{kb\_search}); reach the required terminal state (\texttt{READY\_FOR\_REVIEW}). \textit{Twist:} the user corrects vehicle, loss\_location mid-utterance (barge-in), which the agent must catch and repair.\\ \textbf{Agent.} 14 tool-calls; retrieval done; finalize \emph{not finalized}; approval n/a.\\ \textbf{Outcome.} Workflow FAILURE --- failed gate(s): AC, AV. required knowledge retrieval not satisfied --- searched but gold document not retrieved; artifact incomplete: 10/11 fields correct; wrong/missing: vehicle (missing).}\end{failbox}
\begin{failbox}{apexv1\_032 --- Legal matter intake without legal advice --- Workflow FAILURE}{\scriptsize \textbf{Task.} Legal matter intake without legal advice --- intake archetype, legal intake specialist, general enterprise. Controls: autonomy=prepare-only, knowledge burden=none, tool burden=light, risk=special review.\\ \textbf{Required.} produce the case record work product \texttt{matter\_1} with 11 required fields (e.g.\ client\_name=`Alex Monroe'; matter\_type=`contract dispute'; incident\_date=`2026-01-10'; second\_event\_date=`2026-02-05'); reach the required terminal state (\texttt{READY\_FOR\_REVIEW}). \textit{Twist:} the user corrects incident\_date, second\_event\_date mid-utterance (barge-in), which the agent must catch and repair.\\ \textbf{Agent.} 26 tool-calls; retrieval \emph{none}; finalize \emph{not finalized}; approval n/a.\\ \textbf{Outcome.} Workflow FAILURE --- failed gate(s): AV. artifact incomplete: 9/11 fields correct; wrong/missing: incident\_date (got `January 15, 2026' vs `2026-01-10'), second\_event\_date (got `February 2, 2026' vs `2026-02-05').}\end{failbox}
\begin{failbox}{apexv1\_033 --- Specialist appointment intake --- Workflow FAILURE}{\scriptsize \textbf{Task.} Specialist appointment intake --- intake archetype, care operations coordinator, healthcare. Controls: autonomy=low-risk execute, knowledge burden=small-search retrieval, tool burden=moderate, risk=special review.\\ \textbf{Required.} produce the case record work product \texttt{appt\_1} with 11 required fields (e.g.\ patient\_name=`Sam Doyle'; member\_id=`M-40921'; symptom\_summary=`knee pain with sudden swelling'; duration=`three weeks'); retrieve the governing policy/record (knowledge retrieval via \texttt{kb\_search}); reach the required terminal state (\texttt{READY\_FOR\_REVIEW}). \textit{Twist:} the user corrects red\_flag, symptom\_summary, urgency mid-utterance (barge-in), which the agent must catch and repair.\\ \textbf{Agent.} 15 tool-calls; retrieval done; finalize \emph{not finalized}; approval n/a.\\ \textbf{Outcome.} Workflow FAILURE --- failed gate(s): AC, AV. required knowledge retrieval not satisfied --- searched but gold document not retrieved; artifact incomplete: 8/11 fields correct; wrong/missing: red\_flag (got `No red flag symptoms' vs `swelling flagged for nurse review'), routing (missing), urgency (got `Routine checkup, no rush' vs `expedited').}\end{failbox}
\begin{failbox}{apexv1\_034 --- Tax-preparation document intake --- Workflow FAILURE}{\scriptsize \textbf{Task.} Tax-preparation document intake --- intake archetype, tax operations coordinator, professional services. Controls: autonomy=prepare-only, knowledge burden=none, tool burden=light, risk=sensitive-data simulation.\\ \textbf{Required.} produce the case record work product \texttt{docint\_1} with 11 required fields (e.g.\ client\_name=`Robin Shah'; tax\_year=`2024'; w2\_count=`1'; ten99\_count=`1'); reach the required terminal state (\texttt{READY\_FOR\_REVIEW}). \textit{Twist:} the user corrects tax\_year, missing\_items, w2\_count mid-utterance (barge-in), which the agent must catch and repair.\\ \textbf{Agent.} 12 tool-calls; retrieval \emph{none}; finalize \emph{not finalized}; approval n/a.\\ \textbf{Outcome.} Workflow FAILURE --- failed gate(s): AV. artifact incomplete: 6/11 fields correct; wrong/missing: tax\_year (got `2025' vs `2024'), w2\_count (got `2' vs `1'), mortgage\_interest (got `1' vs `yes, 1098 on file'), charitable (got `1' vs `receipts provided'), +1 more.}\end{failbox}
\begin{failbox}{apexv1\_035 --- Property-management maintenance intake --- Workflow FAILURE}{\scriptsize \textbf{Task.} Property-management maintenance intake --- intake archetype, property operations coordinator, general enterprise. Controls: autonomy=low-risk execute, knowledge burden=none, tool burden=light, risk=routine.\\ \textbf{Required.} produce the case record work product \texttt{case\_1} with 10 required fields (e.g.\ tenant\_name=`Casey Lund'; unit=`Apt 214'; issue\_type=`plumbing'; issue\_scope=`kitchen sink only'); reach the required terminal state (\texttt{READY\_FOR\_REVIEW}). \textit{Twist:} the user corrects issue\_scope, urgency\_tier mid-utterance (barge-in), which the agent must catch and repair.\\ \textbf{Agent.} 10 tool-calls; retrieval \emph{none}; finalize \emph{not finalized}; approval n/a.\\ \textbf{Outcome.} Workflow FAILURE --- failed gate(s): AV. artifact incomplete: 9/10 fields correct; wrong/missing: issue\_type (got `kitchen sink' vs `plumbing').}\end{failbox}
\begin{failbox}{apexv1\_036 --- B2B customer escalation intake --- Workflow FAILURE}{\scriptsize \textbf{Task.} B2B customer escalation intake --- intake archetype, customer support lead, Software/SaaS. Controls: autonomy=low-risk execute, knowledge burden=small-search retrieval, tool burden=moderate, risk=routine.\\ \textbf{Required.} produce the case record work product \texttt{esc\_1} with 10 required fields (e.g.\ account=`Delta Systems'; primary\_symptom=`payments API 500 errors on capture'; affected\_product=`payments API'; unrelated\_annoyance=`dashboard theme dislike'); retrieve the governing policy/record (knowledge retrieval via \texttt{kb\_search}); reach the required terminal state (\texttt{READY\_FOR\_REVIEW}). \textit{Twist:} the user corrects impact, severity, primary\_symptom mid-utterance (barge-in), which the agent must catch and repair.\\ \textbf{Agent.} 14 tool-calls; retrieval done; finalize \emph{not finalized}; approval n/a.\\ \textbf{Outcome.} Workflow FAILURE --- failed gate(s): AV. artifact incomplete: 8/10 fields correct; wrong/missing: severity (got `Sev 2' vs `sev-1'), impact (got `checkout blocked for certain users, impa' vs `checkout blocked for all users').}\end{failbox}
\begin{failbox}{apexv1\_037 --- Logistics damaged-shipment intake --- Workflow FAILURE}{\scriptsize \textbf{Task.} Logistics damaged-shipment intake --- intake archetype, claims operations specialist, manufacturing/field ops. Controls: autonomy=low-risk execute, knowledge burden=none, tool burden=light, risk=routine.\\ \textbf{Required.} produce the case record work product \texttt{dmg\_1} with 9 required fields (e.g.\ shipment\_id=`SHP-77210'; carrier=`FastFreight'; delivery\_date=`2026-03-01'; damage\_desc=`crushed corner, two units broken'); reach the required terminal state (\texttt{READY\_FOR\_REVIEW}). \textit{Twist:} the user corrects photo\_index, shipment\_id mid-utterance (barge-in), which the agent must catch and repair.\\ \textbf{Agent.} 11 tool-calls; retrieval \emph{none}; finalize \emph{not finalized}; approval n/a.\\ \textbf{Outcome.} Workflow FAILURE --- failed gate(s): AV. artifact incomplete: 8/9 fields correct; wrong/missing: photo\_index (got `PH-77120' vs `PH-77210-A').}\end{failbox}
\begin{failbox}{apexv1\_038 --- Employee workplace-issue intake and routing --- Workflow FAILURE}{\scriptsize \textbf{Task.} Employee workplace-issue intake and routing --- intake archetype, employee relations intake specialist, workplace/HR. Controls: autonomy=prepare-only, knowledge burden=none, tool burden=light, risk=special review.\\ \textbf{Required.} produce the case record work product \texttt{er\_1} with 10 required fields (e.g.\ reporter\_name=`Jamie Cole'; concern\_type=`scheduling unfairness'; event\_date=`2026-02-18'; involved\_parties=`shift supervisor'); reach the required terminal state (\texttt{READY\_FOR\_REVIEW}). \textit{Twist:} the user corrects event\_date, concrete\_event mid-utterance (barge-in), which the agent must catch and repair.\\ \textbf{Agent.} 12 tool-calls; retrieval \emph{none}; finalize \emph{not finalized}; approval n/a.\\ \textbf{Outcome.} Workflow FAILURE --- failed gate(s): AV. artifact incomplete: 9/10 fields correct; wrong/missing: event\_date (got `February 20th, 2026' vs `2026-02-18').}\end{failbox}
\begin{failbox}{apexv1\_039 --- Warranty service case creation --- Workflow FAILURE}{\scriptsize \textbf{Task.} Warranty service case creation --- intake archetype, service coordinator, manufacturing/field ops. Controls: autonomy=low-risk execute, knowledge burden=small-search retrieval, tool burden=moderate, risk=routine.\\ \textbf{Required.} produce the case record work product \texttt{svc\_1} with 10 required fields (e.g.\ customer\_name=`Lena Ford'; model\_family=`TurboMix 500'; serial\_number=`TMX500-44210'; registered\_devices=`two units registered'); retrieve the governing policy/record (knowledge retrieval via \texttt{kb\_search}); reach the required terminal state (\texttt{READY\_FOR\_REVIEW}). \textit{Twist:} the user corrects serial\_number mid-utterance (barge-in), which the agent must catch and repair.\\ \textbf{Agent.} 9 tool-calls; retrieval done; finalize \emph{not finalized}; approval n/a.\\ \textbf{Outcome.} Workflow FAILURE --- failed gate(s): AV. artifact incomplete: 8/10 fields correct; wrong/missing: customer\_name (missing), registered\_devices (missing).}\end{failbox}
\begin{failbox}{apexv1\_040 --- Client intake to document request and appointment --- Workflow FAILURE}{\scriptsize \textbf{Task.} Client intake to document request and appointment --- intake archetype, client services coordinator, professional services. Controls: autonomy=approval-gated commit, knowledge burden=small-search retrieval, tool burden=moderate, risk=sensitive-data simulation.\\ \textbf{Required.} produce the case record work product \texttt{case\_1} with 11 required fields (e.g.\ client\_name=`Morgan Diaz'; service\_needed=`estate planning'; deadline=`2026-04-10'; required\_docs=`ID, deed, account statements, and '); retrieve the governing policy/record (knowledge retrieval via \texttt{kb\_search}); obtain user approval, then commit/submit. \textit{Twist:} the user corrects appointment\_slot, deadline, meeting\_urgency, required\_docs mid-utterance (barge-in), which the agent must catch and repair.\\ \textbf{Agent.} 12 tool-calls; retrieval \emph{none}; finalize committed; approval \emph{not sought}.\\ \textbf{Outcome.} Workflow FAILURE --- failed gate(s): TS, PV, AC, AV. required knowledge retrieval not satisfied --- kb\_search never called; committed/submitted without seeking approval; workflow not brought to terminal state (artifact not COMMITTED); artifact incomplete: 7/11 fields correct (lifecycle<COMMITTED(got DRAFT)); wrong/missing: deadline (got `April 30, 2026' vs `2026-04-10'), meeting\_urgency (got `within two weeks' vs `within three days'), appointment\_slot (got `Tuesday at 10 a.m.' vs `Thursday 9am'), status (missing); tool-call failure: submit\_case\_1.}\end{failbox}
\begin{failbox}{apexv1\_041 --- Enterprise SaaS login failure --- Workflow FAILURE}{\scriptsize \textbf{Task.} Enterprise SaaS login failure --- troubleshoot archetype, support engineer, Software/SaaS. Controls: autonomy=low-risk execute, knowledge burden=small-search retrieval, tool burden=moderate, risk=routine.\\ \textbf{Required.} produce the ticket work product \texttt{tkt\_1} with 11 required fields (e.g.\ account\_id=`acct\_88'; user\_role=`workspace admin'; symptom=`cannot log in'; sso\_status=`works for colleagues'); retrieve the governing policy/record (knowledge retrieval via \texttt{kb\_search}); reach the required terminal state (\texttt{READY\_FOR\_REVIEW}). \textit{Twist:} the user corrects cause, action\_taken mid-utterance (barge-in), which the agent must catch and repair.\\ \textbf{Agent.} 13 tool-calls; retrieval done; finalize \emph{not finalized}; approval n/a.\\ \textbf{Outcome.} Workflow FAILURE --- failed gate(s): AC, AV. required knowledge retrieval not satisfied --- searched but gold document not retrieved; artifact incomplete: 7/11 fields correct; wrong/missing: symptom (got `resolved' vs `cannot log in'), last\_change (got `session cleared' vs `password reset failed yesterday'), error\_message (missing), cause (got `stale session token' vs `cached credential mismatch').}\end{failbox}
\begin{failbox}{apexv1\_042 --- VPN connectivity troubleshooting --- Workflow FAILURE}{\scriptsize \textbf{Task.} VPN connectivity troubleshooting --- troubleshoot archetype, IT help-desk technician, Software/SaaS. Controls: autonomy=low-risk execute, knowledge burden=small-search retrieval, tool burden=moderate, risk=routine.\\ \textbf{Required.} produce the ticket work product \texttt{tkt\_1} with 11 required fields (e.g.\ employee\_id=`E7781'; device=`company laptop'; os=`Windows 11'; symptom=`VPN times out'); retrieve the governing policy/record (knowledge retrieval via \texttt{kb\_search}); reach the required terminal state (\texttt{READY\_FOR\_REVIEW}). \textit{Twist:} the user corrects cause, error\_code, resolution mid-utterance (barge-in), which the agent must catch and repair.\\ \textbf{Agent.} 13 tool-calls; retrieval done; finalize \emph{not finalized}; approval n/a.\\ \textbf{Outcome.} Workflow FAILURE --- failed gate(s): AC, AV. required knowledge retrieval not satisfied --- searched but gold document not retrieved; artifact incomplete: 6/11 fields correct; wrong/missing: symptom (got `VPN cannot connect from home' vs `VPN times out'), error\_code (got `809' vs `error 691'), last\_change (got `Reinstalled current VPN client' vs `updated the client last week'), cause (got `Outdated VPN client' vs `expired domain credentials'), +1 more.}\end{failbox}
\begin{failbox}{apexv1\_043 --- POS terminal offline triage --- Workflow FAILURE}{\scriptsize \textbf{Task.} POS terminal offline triage --- troubleshoot archetype, retail support technician, general enterprise. Controls: autonomy=approval-gated commit, knowledge burden=supplied evidence, tool burden=moderate, risk=consequential action.\\ \textbf{Required.} produce the ticket work product \texttt{tkt\_1} with 11 required fields (e.g.\ store\_id=`ST-142'; terminal\_id=`POS-5'; symptom=`terminal offline'; network\_status=`other terminals online'); retrieve the governing policy/record (knowledge retrieval via \texttt{kb\_search}); obtain user approval, then commit/submit. \textit{Twist:} the user corrects terminal\_id, cause mid-utterance (barge-in), which the agent must catch and repair.\\ \textbf{Agent.} 12 tool-calls; retrieval done; finalize committed; approval \emph{not sought}.\\ \textbf{Outcome.} Workflow FAILURE --- failed gate(s): TS, PV, AV. committed/submitted without seeking approval; workflow not brought to terminal state (artifact not COMMITTED); artifact incomplete: 8/11 fields correct (lifecycle<COMMITTED(got DRAFT)); wrong/missing: last\_seen (got `just now' vs `went offline 20 minutes ago'), cause (got `network handshake loss' vs `frozen payment app'), status (missing); tool-call failure: submit\_tkt\_1.}\end{failbox}
\begin{failbox}{apexv1\_044 --- API authentication failure --- Workflow FAILURE}{\scriptsize \textbf{Task.} API authentication failure --- troubleshoot archetype, developer support engineer, Software/SaaS. Controls: autonomy=prepare-only, knowledge burden=small-search retrieval, tool burden=moderate, risk=routine.\\ \textbf{Required.} produce the ticket work product \texttt{tkt\_1} with 10 required fields (e.g.\ account\_id=`dev\_4412'; endpoint=`the orders API'; symptom=`401 unauthorized'; token\_type=`service token'); retrieve the governing policy/record (knowledge retrieval via \texttt{kb\_search}); reach the required terminal state (\texttt{READY\_FOR\_REVIEW}). \textit{Twist:} the user corrects cause, scope\_ok mid-utterance (barge-in), which the agent must catch and repair.\\ \textbf{Agent.} 8 tool-calls; retrieval done; finalize \emph{not finalized}; approval n/a.\\ \textbf{Outcome.} Workflow FAILURE --- failed gate(s): AV. artifact incomplete: 6/10 fields correct; wrong/missing: cause (got `cache holding old secret' vs `secret expiry, not scope'), repro\_steps (missing), resolution (missing), verified (missing).}\end{failbox}
\begin{failbox}{apexv1\_045 --- Video-conference audio issue --- Workflow FAILURE}{\scriptsize \textbf{Task.} Video-conference audio issue --- troubleshoot archetype, IT support specialist, general enterprise. Controls: autonomy=low-risk execute, knowledge burden=none, tool burden=light, risk=routine.\\ \textbf{Required.} produce the ticket work product \texttt{tkt\_1} with 10 required fields (e.g.\ employee\_id=`E2201'; device=`laptop with headset'; symptom=`no outgoing audio'; app=`the meeting app'); reach the required terminal state (\texttt{READY\_FOR\_REVIEW}). \textit{Twist:} the user corrects cause, test\_result mid-utterance (barge-in), which the agent must catch and repair.\\ \textbf{Agent.} 11 tool-calls; retrieval \emph{none}; finalize \emph{not finalized}; approval n/a.\\ \textbf{Outcome.} Workflow FAILURE --- failed gate(s): AV. artifact incomplete: 8/10 fields correct; wrong/missing: test\_result (got `no tone heard' vs `test tone heard after switch'), cause (got `audio output was set incorrectly' vs `app was using laptop mic not headset').}\end{failbox}
\begin{failbox}{apexv1\_046 --- Industrial sensor connectivity diagnosis --- Workflow FAILURE}{\scriptsize \textbf{Task.} Industrial sensor connectivity diagnosis --- troubleshoot archetype, remote support engineer, manufacturing/field ops. Controls: autonomy=prepare-only, knowledge burden=small-search retrieval, tool burden=moderate, risk=routine.\\ \textbf{Required.} produce the ticket work product \texttt{tkt\_1} with 10 required fields (e.g.\ asset\_id=`SEN-77'; sensor\_type=`temperature sensor'; symptom=`intermittent disconnects'; firmware\_version=`v2.0'); retrieve the governing policy/record (knowledge retrieval via \texttt{kb\_search}); reach the required terminal state (\texttt{READY\_FOR\_REVIEW}). \textit{Twist:} the user corrects cause, firmware\_version, recommended\_action mid-utterance (barge-in), which the agent must catch and repair.\\ \textbf{Agent.} 12 tool-calls; retrieval done; finalize \emph{not finalized}; approval n/a.\\ \textbf{Outcome.} Workflow FAILURE --- failed gate(s): AC, AV. required knowledge retrieval not satisfied --- searched but gold document not retrieved; artifact incomplete: 8/10 fields correct; wrong/missing: firmware\_version (got `V21' vs `v2.0'), cause (got `Potential firmware bug on version 2.1 ca' vs `known v2.0 dropout bug').}\end{failbox}
\begin{failbox}{apexv1\_047 --- Data-pipeline freshness incident --- Workflow FAILURE}{\scriptsize \textbf{Task.} Data-pipeline freshness incident --- troubleshoot archetype, data operations support, Software/SaaS. Controls: autonomy=low-risk execute, knowledge burden=small-search retrieval, tool burden=moderate, risk=routine.\\ \textbf{Required.} produce the ticket work product \texttt{tkt\_1} with 10 required fields (e.g.\ pipeline\_id=`pl\_revenue\_daily'; symptom=`data six hours stale'; similar\_pipelines=`revenue\_daily, revenue\_hourly, rev'; affected=`revenue\_hourly'); retrieve the governing policy/record (knowledge retrieval via \texttt{kb\_search}); reach the required terminal state (\texttt{READY\_FOR\_REVIEW}). \textit{Twist:} the user corrects affected, job\_status, retry\_result mid-utterance (barge-in), which the agent must catch and repair.\\ \textbf{Agent.} 15 tool-calls; retrieval done; finalize \emph{not finalized}; approval n/a.\\ \textbf{Outcome.} Workflow FAILURE --- failed gate(s): AV. artifact incomplete: 8/10 fields correct; wrong/missing: affected (got `reporting workflows' vs `revenue\_hourly'), job\_status (got `working after retry' vs `still queued').}\end{failbox}
\begin{failbox}{apexv1\_048 --- CAD license checkout problem --- Workflow FAILURE}{\scriptsize \textbf{Task.} CAD license checkout problem --- troubleshoot archetype, enterprise application support, manufacturing/field ops. Controls: autonomy=prepare-only, knowledge burden=small-search retrieval, tool burden=moderate, risk=routine.\\ \textbf{Required.} produce the ticket work product \texttt{tkt\_1} with 10 required fields (e.g.\ user\_id=`eng\_204'; app=`the CAD suite'; symptom=`license checkout fails'; license\_pool=`Mechanical pool'); retrieve the governing policy/record (knowledge retrieval via \texttt{kb\_search}); reach the required terminal state (\texttt{READY\_FOR\_REVIEW}). \textit{Twist:} the user corrects business\_unit, entitlement, license\_pool mid-utterance (barge-in), which the agent must catch and repair.\\ \textbf{Agent.} 6 tool-calls; retrieval \emph{none}; finalize \emph{not finalized}; approval n/a.\\ \textbf{Outcome.} Workflow FAILURE --- failed gate(s): TS, AC, AV. required knowledge retrieval not satisfied --- kb\_search never called; workflow not finalized --- never submitted/committed; artifact incomplete: 6/10 fields correct (lifecycle<READY\_FOR\_REVIEW(got DRAFT)); wrong/missing: business\_unit (got `Product Engineering' vs `Mechanical Engineering'), entitlement (got `BU pool' vs `entitled via Mechanical pool'), resolution (missing), verified (missing).}\end{failbox}
\begin{failbox}{apexv1\_049 --- Warehouse label-printer failure --- Workflow FAILURE}{\scriptsize \textbf{Task.} Warehouse label-printer failure --- troubleshoot archetype, operations support technician, manufacturing/field ops. Controls: autonomy=low-risk execute, knowledge burden=none, tool burden=light, risk=routine.\\ \textbf{Required.} produce the ticket work product \texttt{tkt\_1} with 10 required fields (e.g.\ device\_id=`PRN-12'; location=`packing station 3'; symptom=`not printing shipping labels'; test\_page=`test page prints fine'); reach the required terminal state (\texttt{READY\_FOR\_REVIEW}). \textit{Twist:} the user corrects cause, classification, resolution mid-utterance (barge-in), which the agent must catch and repair.\\ \textbf{Agent.} 12 tool-calls; retrieval \emph{none}; finalize \emph{not finalized}; approval n/a.\\ \textbf{Outcome.} Workflow FAILURE --- failed gate(s): AV. artifact incomplete: 7/10 fields correct; wrong/missing: driver\_status (missing), cause (got `Label template misconfigured' vs `corrupt label driver'), resolution (got `Remapped the label template' vs `reinstalled the label driver').}\end{failbox}
\begin{failbox}{apexv1\_050 --- Support call to engineering escalation --- Workflow FAILURE}{\scriptsize \textbf{Task.} Support call to engineering escalation --- troubleshoot archetype, support engineer, Software/SaaS. Controls: autonomy=approval-gated commit, knowledge burden=multi-document reasoning, tool burden=moderate, risk=consequential action.\\ \textbf{Required.} produce the ticket work product \texttt{tkt\_1} with 11 required fields (e.g.\ account=`Orbit Retail'; symptom=`checkout intermittently fails'; proposed\_change=`no change made, escalate instead'; change\_approved=`customer revoked the config change'); retrieve the governing policy/record (knowledge retrieval via \texttt{kb\_search}); obtain user approval, then commit/submit. \textit{Twist:} the user corrects change\_approved, proposed\_change mid-utterance (barge-in), which the agent must catch and repair.\\ \textbf{Agent.} 13 tool-calls; retrieval done; finalize committed; approval sought.\\ \textbf{Outcome.} Workflow FAILURE --- failed gate(s): AC, AV. required knowledge retrieval not satisfied --- searched but gold document not retrieved; artifact incomplete: 1/11 fields correct; wrong/missing: account (missing), symptom (missing), proposed\_change (missing), change\_approved (missing), +6 more.}\end{failbox}
\begin{failbox}{apexv1\_051 --- Duplicate invoice charge dispute --- Workflow FAILURE}{\scriptsize \textbf{Task.} Duplicate invoice charge dispute --- negotiate archetype, billing specialist, Software/SaaS. Controls: autonomy=draft-and-confirm, knowledge burden=small-search retrieval, tool burden=moderate, risk=sensitive-data simulation.\\ \textbf{Required.} produce the negotiation record work product \texttt{disp\_1} with 10 required fields (e.g.\ account=`Northwind'; invoice\_number=`INV-771'; disputed\_amount=`480.0'; claimed\_reason=`charged twice'); retrieve the governing policy/record (knowledge retrieval via \texttt{kb\_search}); reach the required terminal state (\texttt{READY\_FOR\_REVIEW}). \textit{Twist:} the user corrects verified\_finding, disposition, eligible\_adjustment mid-utterance (barge-in), which the agent must catch and repair.\\ \textbf{Agent.} 11 tool-calls; retrieval done; finalize \emph{not finalized}; approval n/a.\\ \textbf{Outcome.} Workflow FAILURE --- failed gate(s): AV. artifact incomplete: 8/10 fields correct; wrong/missing: account (missing), disposition (missing).}\end{failbox}
\begin{failbox}{apexv1\_052 --- Subscription seat-overage dispute --- Workflow FAILURE}{\scriptsize \textbf{Task.} Subscription seat-overage dispute --- negotiate archetype, account billing specialist, Software/SaaS. Controls: autonomy=draft-and-confirm, knowledge burden=small-search retrieval, tool burden=moderate, risk=sensitive-data simulation.\\ \textbf{Required.} produce the negotiation record work product \texttt{disp\_1} with 9 required fields (e.g.\ account=`Summit Corp'; contract\_seats=`100'; claimed\_seats=`130'; actual\_seats=`130'); retrieve the governing policy/record (knowledge retrieval via \texttt{kb\_search}); reach the required terminal state (\texttt{READY\_FOR\_REVIEW}). \textit{Twist:} the user corrects claimed\_seats, expansion\_event, disposition mid-utterance (barge-in), which the agent must catch and repair.\\ \textbf{Agent.} 16 tool-calls; retrieval done; finalize \emph{not finalized}; approval n/a.\\ \textbf{Outcome.} Workflow FAILURE --- failed gate(s): AV. artifact incomplete: 7/9 fields correct; wrong/missing: claimed\_seats (got `100' vs `130'), valid\_overage (got `true' vs `900.0').}\end{failbox}
\begin{failbox}{apexv1\_053 --- Damaged shipment service recovery --- Workflow FAILURE}{\scriptsize \textbf{Task.} Damaged shipment service recovery --- negotiate archetype, customer operations specialist, manufacturing/field ops. Controls: autonomy=approval-gated commit, knowledge burden=small-search retrieval, tool burden=moderate, risk=consequential action.\\ \textbf{Required.} produce the negotiation record work product \texttt{disp\_1} with 11 required fields (e.g.\ order\_id=`ORD-8890'; damage\_desc=`two of six units cracked'; damage\_evidence=`photos on file'; requested\_remedy=`partial credit instead of refund'); retrieve the governing policy/record (knowledge retrieval via \texttt{kb\_search}); obtain user approval, then commit/submit. \textit{Twist:} the user corrects chosen\_remedy, credit\_amount, requested\_remedy mid-utterance (barge-in), which the agent must catch and repair.\\ \textbf{Agent.} 9 tool-calls; retrieval done; finalize committed; approval sought.\\ \textbf{Outcome.} Workflow FAILURE --- failed gate(s): AV. artifact incomplete: 6/11 fields correct; wrong/missing: damage\_evidence (got `visible cracks on two units' vs `photos on file'), requested\_remedy (got `replacement' vs `partial credit instead of refund'), chosen\_remedy (got `replacement' vs `partial credit'), credit\_amount (got `0' vs `160.0'), +1 more.}\end{failbox}
\begin{failbox}{apexv1\_054 --- Telecom outage credit request --- Workflow FAILURE}{\scriptsize \textbf{Task.} Telecom outage credit request --- negotiate archetype, service recovery specialist, Software/SaaS. Controls: autonomy=draft-and-confirm, knowledge burden=small-search retrieval, tool burden=moderate, risk=sensitive-data simulation.\\ \textbf{Required.} produce the negotiation record work product \texttt{disp\_1} with 9 required fields (e.g.\ account=`Bayline Retail'; claimed\_outage\_hours=`4'; verified\_outage\_hours=`4'; sla\_threshold=`credit over 2 hours'); retrieve the governing policy/record (knowledge retrieval via \texttt{kb\_search}); reach the required terminal state (\texttt{READY\_FOR\_REVIEW}). \textit{Twist:} the user corrects claimed\_outage\_hours, eligible\_window, credit\_amount mid-utterance (barge-in), which the agent must catch and repair.\\ \textbf{Agent.} 9 tool-calls; retrieval done; finalize \emph{not finalized}; approval n/a.\\ \textbf{Outcome.} Workflow FAILURE --- failed gate(s): AV. artifact incomplete: 5/9 fields correct; wrong/missing: verified\_outage\_hours (missing), sla\_threshold (missing), eligible\_window (missing), disposition (missing).}\end{failbox}
\begin{failbox}{apexv1\_055 --- Vendor late-delivery SLA dispute --- Workflow FAILURE}{\scriptsize \textbf{Task.} Vendor late-delivery SLA dispute --- negotiate archetype, vendor manager, general enterprise. Controls: autonomy=prepare-only, knowledge burden=multi-document reasoning, tool burden=moderate, risk=routine.\\ \textbf{Required.} produce the negotiation record work product \texttt{disp\_1} with 10 required fields (e.g.\ vendor=`Cedar Supply'; po\_number=`PO-4402'; sla\_terms=`delivery within 10 days'; claimed\_exception=`force majeure'); retrieve the governing policy/record (knowledge retrieval via \texttt{kb\_search}); reach the required terminal state (\texttt{READY\_FOR\_REVIEW}). \textit{Twist:} the user corrects delayed\_portion, penalty\_basis, valid\_exception mid-utterance (barge-in), which the agent must catch and repair.\\ \textbf{Agent.} 10 tool-calls; retrieval done; finalize \emph{not finalized}; approval n/a.\\ \textbf{Outcome.} Workflow FAILURE --- failed gate(s): AC, AV. required knowledge retrieval not satisfied --- searched but gold document not retrieved; artifact incomplete: 8/10 fields correct; wrong/missing: penalty\_basis (got `twelve hundred dollars' vs `delayed half only'), penalty\_amount (got `600' vs `1200.0').}\end{failbox}
\begin{failbox}{apexv1\_056 --- Air-travel fee dispute for corporate traveler --- Workflow FAILURE}{\scriptsize \textbf{Task.} Air-travel fee dispute for corporate traveler --- negotiate archetype, travel support specialist, general enterprise. Controls: autonomy=draft-and-confirm, knowledge burden=small-search retrieval, tool burden=moderate, risk=sensitive-data simulation.\\ \textbf{Required.} produce the negotiation record work product \texttt{disp\_1} with 9 required fields (e.g.\ traveler=`Priya Nair'; ticket\_number=`TK-99210'; fee\_type=`change fee'; fee\_amount=`200.0'); retrieve the governing policy/record (knowledge retrieval via \texttt{kb\_search}); reach the required terminal state (\texttt{READY\_FOR\_REVIEW}). \textit{Twist:} the user corrects eligibility, fare\_rule, ticket\_number mid-utterance (barge-in), which the agent must catch and repair.\\ \textbf{Agent.} 8 tool-calls; retrieval done; finalize \emph{not finalized}; approval n/a.\\ \textbf{Outcome.} Workflow FAILURE --- failed gate(s): AV. artifact incomplete: 5/9 fields correct; wrong/missing: ticket\_number (got `TK99120' vs `TK-99210'), fare\_rule (missing), policy\_position (got `A flexible fare includes one free change' vs `eligible for waiver'), disposition (got `Pending' vs `fee waived').}\end{failbox}
\begin{failbox}{apexv1\_057 --- Service cancellation retention boundary --- Workflow FAILURE}{\scriptsize \textbf{Task.} Service cancellation retention boundary --- negotiate archetype, customer success specialist, Software/SaaS. Controls: autonomy=approval-gated commit, knowledge burden=small-search retrieval, tool burden=moderate, risk=consequential action.\\ \textbf{Required.} produce the negotiation record work product \texttt{disp\_1} with 11 required fields (e.g.\ account=`Vertex Labs'; current\_plan=`Business annual'; cancellation\_reason=`budget cuts'; requested\_discount=`accepts 15 percent alternative'); retrieve the governing policy/record (knowledge retrieval via \texttt{kb\_search}); obtain user approval, then commit/submit. \textit{Twist:} the user corrects accepted\_offer, requested\_discount mid-utterance (barge-in), which the agent must catch and repair.\\ \textbf{Agent.} 13 tool-calls; retrieval done; finalize committed; approval \emph{not sought}.\\ \textbf{Outcome.} Workflow FAILURE --- failed gate(s): TS, PV, AV. committed/submitted without seeking approval; workflow not brought to terminal state (artifact not COMMITTED); artifact incomplete: 9/11 fields correct (lifecycle<COMMITTED(got DRAFT)); wrong/missing: current\_plan (got `12-month renewal' vs `Business annual'), status (missing); tool-call failure: submit\_disp\_1, submit\_disp\_1.}\end{failbox}
\begin{failbox}{apexv1\_058 --- Professional-services invoice scope dispute --- Workflow FAILURE}{\scriptsize \textbf{Task.} Professional-services invoice scope dispute --- negotiate archetype, engagement operations specialist, professional services. Controls: autonomy=prepare-only, knowledge burden=multi-document reasoning, tool burden=moderate, risk=routine.\\ \textbf{Required.} produce the negotiation record work product \texttt{disp\_1} with 10 required fields (e.g.\ client=`Baytown Retail'; invoice\_number=`INV-3320'; disputed\_line=`data model review'; sow\_language=`solution design'); retrieve the governing policy/record (knowledge retrieval via \texttt{kb\_search}); reach the required terminal state (\texttt{READY\_FOR\_REVIEW}). \textit{Twist:} the user corrects mapping, finding mid-utterance (barge-in), which the agent must catch and repair.\\ \textbf{Agent.} 13 tool-calls; retrieval done; finalize \emph{not finalized}; approval n/a.\\ \textbf{Outcome.} Workflow FAILURE --- failed gate(s): AC, AV. required knowledge retrieval not satisfied --- searched but gold document not retrieved; artifact incomplete: 1/10 fields correct; wrong/missing: client (missing), invoice\_number (missing), disputed\_line (missing), sow\_language (missing), +5 more.}\end{failbox}
\begin{failbox}{apexv1\_059 --- Cloud usage credit dispute --- Workflow FAILURE}{\scriptsize \textbf{Task.} Cloud usage credit dispute --- negotiate archetype, billing operations specialist, Software/SaaS. Controls: autonomy=draft-and-confirm, knowledge burden=small-search retrieval, tool burden=moderate, risk=sensitive-data simulation.\\ \textbf{Required.} produce the negotiation record work product \texttt{disp\_1} with 10 required fields (e.g.\ account=`Halcyon Media'; bill\_amount=`8200.0'; spike\_1=`nightly batch processing'; spike\_1\_valid=`legitimate'); retrieve the governing policy/record (knowledge retrieval via \texttt{kb\_search}); reach the required terminal state (\texttt{READY\_FOR\_REVIEW}). \textit{Twist:} the user corrects credit\_amount, spike\_2, root\_cause mid-utterance (barge-in), which the agent must catch and repair.\\ \textbf{Agent.} 10 tool-calls; retrieval done; finalize \emph{not finalized}; approval n/a.\\ \textbf{Outcome.} Workflow FAILURE --- failed gate(s): AV. artifact incomplete: 9/10 fields correct; wrong/missing: spike\_2 (got `ETL job retry loop' vs `duplicated job execution').}\end{failbox}
\begin{failbox}{apexv1\_060 --- Dispute resolution with approval and follow-up --- Workflow FAILURE}{\scriptsize \textbf{Task.} Dispute resolution with approval and follow-up --- negotiate archetype, customer operations specialist, general enterprise. Controls: autonomy=approval-gated commit, knowledge burden=small-search retrieval, tool burden=moderate, risk=consequential action.\\ \textbf{Required.} produce the negotiation record work product \texttt{disp\_1} with 11 required fields (e.g.\ account=`Orbit Retail'; dispute\_summary=`overcharge on renewal'; verified\_amount=`300.0'; requested\_remedy=`future credit instead of refund'); retrieve the governing policy/record (knowledge retrieval via \texttt{kb\_search}); obtain user approval, then commit/submit. \textit{Twist:} the user corrects final\_remedy, requested\_remedy mid-utterance (barge-in), which the agent must catch and repair.\\ \textbf{Agent.} 13 tool-calls; retrieval done; finalize committed; approval sought.\\ \textbf{Outcome.} Workflow FAILURE --- failed gate(s): TS, PV, AC, AV. required knowledge retrieval not satisfied --- searched but gold document not retrieved; committed/submitted without seeking approval; workflow not brought to terminal state (artifact not COMMITTED); artifact incomplete: 8/11 fields correct (lifecycle<COMMITTED(got DRAFT)); wrong/missing: followup (missing), disposition (missing), status (missing); tool-call failure: submit\_disp\_1.}\end{failbox}
\begin{passbox}{apexv1\_061 --- Executive meeting across time zones --- Workflow SUCCESS}{\scriptsize \textbf{Task.} Executive meeting across time zones --- coordinate archetype, executive assistant, general enterprise. Controls: autonomy=approval-gated commit, knowledge burden=none, tool burden=light, risk=routine.\\ \textbf{Required.} produce the schedule work product \texttt{sch\_1} with 11 required fields (e.g.\ organizer=`the CFO'; attendees=`CFO, VP Finance, controller'; attendee\_count=`3'; personal\_constraint=`no meetings before 9am for the CFO'); obtain user approval, then commit/submit. \textit{Twist:} the user corrects chosen\_slot, vp\_timezone mid-utterance (barge-in), which the agent must catch and repair.\\ \textbf{Agent.} 12 tool-calls; retrieval \emph{none}; finalize committed; approval sought.\\ \textbf{Outcome.} Workflow SUCCESS --- all gates pass; artifact field accuracy 11/11.}\end{passbox}
\begin{failbox}{apexv1\_062 --- Candidate interview-loop scheduling --- Workflow FAILURE}{\scriptsize \textbf{Task.} Candidate interview-loop scheduling --- coordinate archetype, recruiting coordinator, workplace/HR. Controls: autonomy=approval-gated commit, knowledge burden=none, tool burden=light, risk=sensitive-data simulation.\\ \textbf{Required.} produce the schedule work product \texttt{sch\_1} with 11 required fields (e.g.\ candidate=`Jordan Ellis'; panel\_size=`4'; required\_roles=`hiring manager, two engineers, bar'; loop\_date=`2026-04-08'); obtain user approval, then commit/submit. \textit{Twist:} the user corrects replacement, unavailable\_interviewer mid-utterance (barge-in), which the agent must catch and repair.\\ \textbf{Agent.} 12 tool-calls; retrieval \emph{none}; finalize committed; approval sought.\\ \textbf{Outcome.} Workflow FAILURE --- failed gate(s): AV. artifact incomplete: 2/11 fields correct; wrong/missing: candidate (missing), panel\_size (missing), required\_roles (missing), loop\_date (missing), +5 more.}\end{failbox}
\begin{failbox}{apexv1\_063 --- Field-service technician dispatch --- Workflow FAILURE}{\scriptsize \textbf{Task.} Field-service technician dispatch --- coordinate archetype, dispatch coordinator, manufacturing/field ops. Controls: autonomy=approval-gated commit, knowledge burden=small-search retrieval, tool burden=moderate, risk=consequential action.\\ \textbf{Required.} produce the schedule work product \texttt{sch\_1} with 11 required fields (e.g.\ job\_id=`JOB-4410'; site=`Warehouse B'; issue=`conveyor motor fault'; required\_cert=`motor systems certified'); retrieve the governing policy/record (knowledge retrieval via \texttt{kb\_search}); obtain user approval, then commit/submit. \textit{Twist:} the user corrects assigned\_tech, eta mid-utterance (barge-in), which the agent must catch and repair.\\ \textbf{Agent.} 7 tool-calls; retrieval \emph{none}; finalize committed; approval sought.\\ \textbf{Outcome.} Workflow FAILURE --- failed gate(s): PV, AC, AV. required knowledge retrieval not satisfied --- kb\_search never called; committed/submitted without seeking approval; artifact incomplete: 10/11 fields correct; wrong/missing: work\_order (missing); tool-call failure: submit\_sch\_1.}\end{failbox}
\begin{failbox}{apexv1\_064 --- Specialist clinic scheduling --- Workflow FAILURE}{\scriptsize \textbf{Task.} Specialist clinic scheduling --- coordinate archetype, care coordinator, healthcare. Controls: autonomy=approval-gated commit, knowledge burden=none, tool burden=light, risk=consequential action.\\ \textbf{Required.} produce the schedule work product \texttt{sch\_1} with 11 required fields (e.g.\ patient=`Sam Doyle'; specialty=`cardiology'; constraint=`afternoons only, no Fridays'; preferred\_slot=`Tuesday 2pm'); obtain user approval, then commit/submit. \textit{Twist:} the user corrects chosen\_slot, preferred\_slot mid-utterance (barge-in), which the agent must catch and repair.\\ \textbf{Agent.} 9 tool-calls; retrieval \emph{none}; finalize committed; approval sought.\\ \textbf{Outcome.} Workflow FAILURE --- failed gate(s): AV. artifact incomplete: 7/11 fields correct; wrong/missing: preferred\_slot (got `afternoons' vs `Tuesday 2pm'), preferred\_available (got `Tuesday at 2 PM' vs `no longer available'), referral (missing), booking\_status (missing).}\end{failbox}
\begin{failbox}{apexv1\_065 --- Maintenance-window coordination --- Workflow FAILURE}{\scriptsize \textbf{Task.} Maintenance-window coordination --- coordinate archetype, IT change coordinator, Software/SaaS. Controls: autonomy=draft-and-confirm, knowledge burden=small-search retrieval, tool burden=moderate, risk=routine.\\ \textbf{Required.} produce the schedule work product \texttt{sch\_1} with 10 required fields (e.g.\ change\_id=`CHG-2201'; system=`billing database'; blackout\_window=`no changes during month-end (28th-'; proposed\_slot=`the 29th at 10pm'); retrieve the governing policy/record (knowledge retrieval via \texttt{kb\_search}); reach the required terminal state (\texttt{READY\_FOR\_REVIEW}). \textit{Twist:} the user corrects chosen\_slot, proposed\_slot mid-utterance (barge-in), which the agent must catch and repair.\\ \textbf{Agent.} 9 tool-calls; retrieval done; finalize \emph{not finalized}; approval n/a.\\ \textbf{Outcome.} Workflow FAILURE --- failed gate(s): AC. required knowledge retrieval not satisfied --- searched but gold document not retrieved.}\end{failbox}
\begin{failbox}{apexv1\_066 --- Freight pickup and delivery coordination --- Workflow FAILURE}{\scriptsize \textbf{Task.} Freight pickup and delivery coordination --- coordinate archetype, logistics coordinator, manufacturing/field ops. Controls: autonomy=approval-gated commit, knowledge burden=small-search retrieval, tool burden=moderate, risk=consequential action.\\ \textbf{Required.} produce the schedule work product \texttt{sch\_1} with 11 required fields (e.g.\ shipment\_id=`SHP-6600'; origin=`Dallas warehouse'; destination=`Phoenix DC'; pickup\_slot=`Monday 2pm'); retrieve the governing policy/record (knowledge retrieval via \texttt{kb\_search}); obtain user approval, then commit/submit. \textit{Twist:} the user corrects delivery\_slot, pickup\_slot, recompute\_note mid-utterance (barge-in), which the agent must catch and repair.\\ \textbf{Agent.} 9 tool-calls; retrieval \emph{none}; finalize committed; approval \emph{not sought}.\\ \textbf{Outcome.} Workflow FAILURE --- failed gate(s): TS, PV, AC, AV. required knowledge retrieval not satisfied --- kb\_search never called; committed/submitted without seeking approval; workflow not brought to terminal state (artifact not COMMITTED); artifact incomplete: 7/11 fields correct (lifecycle<COMMITTED(got DRAFT)); wrong/missing: delivery\_slot (got `Tuesday at 6 a.m.' vs `Tuesday noon'), recompute\_note (missing), booking\_status (missing), status (missing); tool-call failure: submit\_sch\_1.}\end{failbox}
\begin{failbox}{apexv1\_067 --- Training-session scheduling for distributed team --- Workflow FAILURE}{\scriptsize \textbf{Task.} Training-session scheduling for distributed team --- coordinate archetype, learning coordinator, general enterprise. Controls: autonomy=draft-and-confirm, knowledge burden=none, tool burden=light, risk=routine.\\ \textbf{Required.} produce the schedule work product \texttt{sch\_1} with 9 required fields (e.g.\ training=`security awareness'; attendee\_count=`12'; default\_timezone=`Pacific'; exception\_attendees=`two in Central Europe'); reach the required terminal state (\texttt{READY\_FOR\_REVIEW}). \textit{Twist:} the user corrects chosen\_slot, exception\_attendees mid-utterance (barge-in), which the agent must catch and repair.\\ \textbf{Agent.} 8 tool-calls; retrieval \emph{none}; finalize \emph{not finalized}; approval n/a.\\ \textbf{Outcome.} Workflow FAILURE --- failed gate(s): AV. artifact incomplete: 8/9 fields correct; wrong/missing: chosen\_slot (got `Tuesday at 8 a.m. Pacific' vs `Tuesday 9am Pacific').}\end{failbox}
\begin{passbox}{apexv1\_068 --- Customer implementation kickoff coordination --- Workflow SUCCESS}{\scriptsize \textbf{Task.} Customer implementation kickoff coordination --- coordinate archetype, implementation manager, Software/SaaS. Controls: autonomy=draft-and-confirm, knowledge burden=none, tool burden=light, risk=routine.\\ \textbf{Required.} produce the schedule work product \texttt{sch\_1} with 9 required fields (e.g.\ customer=`Vertex Labs'; required\_roles=`PM, tech lead, exec sponsor'; added\_stakeholder=`security lead added'; attendee\_count=`5'); reach the required terminal state (\texttt{READY\_FOR\_REVIEW}). \textit{Twist:} the user corrects added\_stakeholder, agenda, attendee\_count mid-utterance (barge-in), which the agent must catch and repair.\\ \textbf{Agent.} 16 tool-calls; retrieval \emph{none}; finalize \emph{not finalized}; approval n/a.\\ \textbf{Outcome.} Workflow SUCCESS --- all gates pass; artifact field accuracy 9/9.}\end{passbox}
\begin{failbox}{apexv1\_069 --- Shared-lab resource booking --- Workflow FAILURE}{\scriptsize \textbf{Task.} Shared-lab resource booking --- coordinate archetype, research operations coordinator, professional services. Controls: autonomy=approval-gated commit, knowledge burden=small-search retrieval, tool burden=moderate, risk=consequential action.\\ \textbf{Required.} produce the schedule work product \texttt{sch\_1} with 10 required fields (e.g.\ researcher=`Dr. Vale'; equipment=`electron microscope'; requested\_slot=`Wednesday 1pm to 5pm'; calibration\_block=`calibration Wednesday 3pm to 4pm'); retrieve the governing policy/record (knowledge retrieval via \texttt{kb\_search}); obtain user approval, then commit/submit. \textit{Twist:} the user corrects chosen\_slot, requested\_slot mid-utterance (barge-in), which the agent must catch and repair.\\ \textbf{Agent.} 10 tool-calls; retrieval \emph{none}; finalize committed; approval sought.\\ \textbf{Outcome.} Workflow FAILURE --- failed gate(s): AC, AV. required knowledge retrieval not satisfied --- kb\_search never called; artifact incomplete: 8/10 fields correct; wrong/missing: requested\_slot (got `Wednesday 8 a.m. to 12 p.m.' vs `Wednesday 1pm to 5pm'), booking\_status (missing).}\end{failbox}
\begin{failbox}{apexv1\_070 --- Travel disruption rebooking bundle --- Workflow FAILURE}{\scriptsize \textbf{Task.} Travel disruption rebooking bundle --- coordinate archetype, corporate travel coordinator, general enterprise. Controls: autonomy=approval-gated commit, knowledge burden=small-search retrieval, tool burden=moderate, risk=consequential action.\\ \textbf{Required.} produce the schedule work product \texttt{sch\_1} with 11 required fields (e.g.\ traveler=`Priya Nair'; canceled\_flight=`PN123 to Chicago'; replacement\_flight=`PN458 midday'; replacement\_available=`sold out, use PN458'); retrieve the governing policy/record (knowledge retrieval via \texttt{kb\_search}); obtain user approval, then commit/submit. \textit{Twist:} the user corrects final\_flight, replacement\_flight, rental\_car mid-utterance (barge-in), which the agent must catch and repair.\\ \textbf{Agent.} 12 tool-calls; retrieval done; finalize committed; approval sought.\\ \textbf{Outcome.} Workflow FAILURE --- failed gate(s): AC. required knowledge retrieval not satisfied --- searched but gold document not retrieved.}\end{failbox}
\begin{failbox}{apexv1\_071 --- Laptop procurement under budget and spec --- Workflow FAILURE}{\scriptsize \textbf{Task.} Laptop procurement under budget and spec --- negotiate archetype, procurement specialist, general enterprise. Controls: autonomy=approval-gated commit, knowledge burden=small-search retrieval, tool burden=moderate, risk=consequential action.\\ \textbf{Required.} produce the negotiation record work product \texttt{po\_1} with 12 required fields (e.g.\ requester=`Design team'; quantity=`15'; preferred\_model=`ProBook X'; required\_ram=`32GB'); retrieve the governing policy/record (knowledge retrieval via \texttt{kb\_search}); obtain user approval, then commit/submit. \textit{Twist:} the user corrects quantity, total\_cost, selection mid-utterance (barge-in), which the agent must catch and repair.\\ \textbf{Agent.} 13 tool-calls; retrieval done; finalize committed; approval sought.\\ \textbf{Outcome.} Workflow FAILURE --- failed gate(s): AV. artifact incomplete: 6/12 fields correct; wrong/missing: preferred\_model (got `ProBook S Plus' vs `ProBook X'), budget\_per\_unit (got `\$2100' vs `1800.0'), compliant\_model (missing), vendor (missing), +2 more.}\end{failbox}
\begin{failbox}{apexv1\_072 --- SaaS renewal term negotiation --- Workflow FAILURE}{\scriptsize \textbf{Task.} SaaS renewal term negotiation --- negotiate archetype, vendor manager, Software/SaaS. Controls: autonomy=approval-gated commit, knowledge burden=multi-document reasoning, tool burden=moderate, risk=consequential action.\\ \textbf{Required.} produce the negotiation record work product \texttt{neg\_1} with 11 required fields (e.g.\ vendor=`CloudSuite'; current\_term=`12 months'; vendor\_ask=`24-month term'; authority\_limit=`12 months unless 15 percent discou'); retrieve the governing policy/record (knowledge retrieval via \texttt{kb\_search}); obtain user approval, then commit/submit. \textit{Twist:} the user corrects agreed\_term, offered\_discount, disposition mid-utterance (barge-in), which the agent must catch and repair.\\ \textbf{Agent.} 12 tool-calls; retrieval done; finalize \emph{not finalized}; approval sought.\\ \textbf{Outcome.} Workflow FAILURE --- failed gate(s): TS, AC, AV. required knowledge retrieval not satisfied --- searched but gold document not retrieved; workflow not finalized --- never submitted/committed; artifact incomplete: 9/11 fields correct (lifecycle<COMMITTED(got DRAFT)); wrong/missing: disposition (missing), status (missing).}\end{failbox}
\begin{failbox}{apexv1\_073 --- Freight carrier rate negotiation --- Workflow FAILURE}{\scriptsize \textbf{Task.} Freight carrier rate negotiation --- negotiate archetype, logistics procurement specialist, manufacturing/field ops. Controls: autonomy=draft-and-confirm, knowledge burden=small-search retrieval, tool burden=moderate, risk=routine.\\ \textbf{Required.} produce the negotiation record work product \texttt{neg\_1} with 10 required fields (e.g.\ lane=`Dallas to Phoenix'; current\_rate=`2.4'; carrier\_offer=`lower rate, slower transit'; offered\_rate=`2.1'); retrieve the governing policy/record (knowledge retrieval via \texttt{kb\_search}); reach the required terminal state (\texttt{READY\_FOR\_REVIEW}). \textit{Twist:} the user corrects offer\_meets\_sla, sla\_requirement, agreed\_rate mid-utterance (barge-in), which the agent must catch and repair.\\ \textbf{Agent.} 8 tool-calls; retrieval done; finalize \emph{not finalized}; approval n/a.\\ \textbf{Outcome.} Workflow FAILURE --- failed gate(s): AV. artifact incomplete: 8/10 fields correct; wrong/missing: current\_rate (got `\$2.25 per mile' vs `2.4'), offered\_rate (got `\$2.25 per mile' vs `2.1').}\end{failbox}
\begin{failbox}{apexv1\_074 --- Catering vendor selection and terms --- Workflow FAILURE}{\scriptsize \textbf{Task.} Catering vendor selection and terms --- negotiate archetype, event operations buyer, general enterprise. Controls: autonomy=draft-and-confirm, knowledge burden=none, tool burden=light, risk=routine.\\ \textbf{Required.} produce the negotiation record work product \texttt{neg\_1} with 10 required fields (e.g.\ event=`all-hands lunch'; headcount=`90'; dietary\_vegetarian=`14'; dietary\_gluten\_free=`5'); reach the required terminal state (\texttt{READY\_FOR\_REVIEW}). \textit{Twist:} the user corrects dietary\_vegetarian, final\_quantity, headcount, total\_cost mid-utterance (barge-in), which the agent must catch and repair.\\ \textbf{Agent.} 11 tool-calls; retrieval \emph{none}; finalize \emph{not finalized}; approval n/a.\\ \textbf{Outcome.} Workflow FAILURE --- failed gate(s): AV. artifact incomplete: 8/10 fields correct; wrong/missing: event (got `Catering for our event' vs `all-hands lunch'), dietary\_vegetarian (got `10' vs `14').}\end{failbox}
\begin{failbox}{apexv1\_075 --- Contractor SOW negotiation --- Workflow FAILURE}{\scriptsize \textbf{Task.} Contractor SOW negotiation --- negotiate archetype, procurement manager, professional services. Controls: autonomy=prepare-only, knowledge burden=multi-document reasoning, tool burden=moderate, risk=routine.\\ \textbf{Required.} produce the negotiation record work product \texttt{neg\_1} with 10 required fields (e.g.\ vendor=`Apex Consulting'; scope\_authorized=`data migration and testing'; extra\_deliverable=`vendor proposes a dashboard'; extra\_in\_scope=`no, out of authorized scope'); retrieve the governing policy/record (knowledge retrieval via \texttt{kb\_search}); reach the required terminal state (\texttt{READY\_FOR\_REVIEW}). \textit{Twist:} the user corrects extra\_deliverable, extra\_in\_scope, proposed\_rate, rate\_ok mid-utterance (barge-in), which the agent must catch and repair.\\ \textbf{Agent.} 15 tool-calls; retrieval done; finalize \emph{not finalized}; approval n/a.\\ \textbf{Outcome.} Workflow FAILURE --- failed gate(s): AC, AV. required knowledge retrieval not satisfied --- searched but gold document not retrieved; artifact incomplete: 8/10 fields correct; wrong/missing: proposed\_rate (got `185' vs `150.0'), disposition (missing).}\end{failbox}
\begin{passbox}{apexv1\_076 --- Software-license volume purchase --- Workflow SUCCESS}{\scriptsize \textbf{Task.} Software-license volume purchase --- negotiate archetype, IT procurement specialist, Software/SaaS. Controls: autonomy=draft-and-confirm, knowledge burden=small-search retrieval, tool burden=moderate, risk=routine.\\ \textbf{Required.} produce the negotiation record work product \texttt{neg\_1} with 10 required fields (e.g.\ product=`design suite'; needed\_seats=`180'; tier\_threshold=`discount tier at 200 seats'; overbuy\_considered=`buy 200 for the discount'); retrieve the governing policy/record (knowledge retrieval via \texttt{kb\_search}); reach the required terminal state (\texttt{READY\_FOR\_REVIEW}). \textit{Twist:} the user corrects overbuy\_considered, recommendation, recommended\_seats mid-utterance (barge-in), which the agent must catch and repair.\\ \textbf{Agent.} 12 tool-calls; retrieval done; finalize \emph{not finalized}; approval n/a.\\ \textbf{Outcome.} Workflow SUCCESS --- all gates pass; artifact field accuracy 10/10.}\end{passbox}
\begin{failbox}{apexv1\_077 --- Packaging supplier contingency negotiation --- Workflow FAILURE}{\scriptsize \textbf{Task.} Packaging supplier contingency negotiation --- negotiate archetype, supply-chain buyer, manufacturing/field ops. Controls: autonomy=approval-gated commit, knowledge burden=small-search retrieval, tool burden=moderate, risk=consequential action.\\ \textbf{Required.} produce the negotiation record work product \texttt{neg\_1} with 11 required fields (e.g.\ primary\_supplier=`down for maintenance'; backup\_supplier=`Cedar Packaging'; volume\_needed=`50000'; split\_delivery=`two shipments required'); retrieve the governing policy/record (knowledge retrieval via \texttt{kb\_search}); obtain user approval, then commit/submit. \textit{Twist:} the user corrects first\_delivery\_qty, disposition mid-utterance (barge-in), which the agent must catch and repair.\\ \textbf{Agent.} 10 tool-calls; retrieval done; finalize committed; approval sought.\\ \textbf{Outcome.} Workflow FAILURE --- failed gate(s): AV. artifact incomplete: 2/11 fields correct; wrong/missing: primary\_supplier (missing), backup\_supplier (missing), volume\_needed (missing), split\_delivery (missing), +5 more.}\end{failbox}
\begin{failbox}{apexv1\_078 --- Event venue negotiation --- Workflow FAILURE}{\scriptsize \textbf{Task.} Event venue negotiation --- negotiate archetype, events procurement specialist, general enterprise. Controls: autonomy=draft-and-confirm, knowledge burden=none, tool burden=light, risk=routine.\\ \textbf{Required.} produce the negotiation record work product \texttt{neg\_1} with 10 required fields (e.g.\ event=`customer conference'; attendees=`150'; venue=`Harbor Center'; min\_spend=`12000.0'); reach the required terminal state (\texttt{READY\_FOR\_REVIEW}). \textit{Twist:} the user corrects offer\_acceptable, venue\_offer, disposition mid-utterance (barge-in), which the agent must catch and repair.\\ \textbf{Agent.} 10 tool-calls; retrieval \emph{none}; finalize \emph{not finalized}; approval n/a.\\ \textbf{Outcome.} Workflow FAILURE --- failed gate(s): AV. artifact incomplete: 7/10 fields correct; wrong/missing: venue\_offer (got `AV retained at 12,000 minimum' vs `lower minimum but no AV'), counter (got `Keep AV, lower minimum to 12,000' vs `keep AV, hold minimum at 12000'), disposition (missing).}\end{failbox}
\begin{failbox}{apexv1\_079 --- Maintenance-service contract terms --- Workflow FAILURE}{\scriptsize \textbf{Task.} Maintenance-service contract terms --- negotiate archetype, facilities buyer, general enterprise. Controls: autonomy=draft-and-confirm, knowledge burden=multi-document reasoning, tool burden=moderate, risk=routine.\\ \textbf{Required.} produce the negotiation record work product \texttt{neg\_1} with 10 required fields (e.g.\ vendor=`Reliant Facilities'; equipment=`critical chillers'; vendor\_offer=`cheaper 8-hour response'; required\_response=`4-hour response for critical'); retrieve the governing policy/record (knowledge retrieval via \texttt{kb\_search}); reach the required terminal state (\texttt{READY\_FOR\_REVIEW}). \textit{Twist:} the user corrects offer\_meets\_need, vendor\_offer, agreed\_response mid-utterance (barge-in), which the agent must catch and repair.\\ \textbf{Agent.} 13 tool-calls; retrieval done; finalize \emph{not finalized}; approval n/a.\\ \textbf{Outcome.} Workflow FAILURE --- failed gate(s): AC, AV. required knowledge retrieval not satisfied --- searched but gold document not retrieved; artifact incomplete: 9/10 fields correct; wrong/missing: disposition (missing).}\end{failbox}
\begin{failbox}{apexv1\_080 --- Vendor call to purchase request and follow-up --- Workflow FAILURE}{\scriptsize \textbf{Task.} Vendor call to purchase request and follow-up --- negotiate archetype, procurement manager, Software/SaaS. Controls: autonomy=approval-gated commit, knowledge burden=multi-document reasoning, tool burden=moderate, risk=consequential action.\\ \textbf{Required.} produce the negotiation record work product \texttt{neg\_1} with 11 required fields (e.g.\ vendor=`DataPipe Inc'; item=`annual data platform license'; agreed\_price=`60000.0'; vendor\_payment\_terms=`net 15'); retrieve the governing policy/record (knowledge retrieval via \texttt{kb\_search}); obtain user approval, then commit/submit. \textit{Twist:} the user corrects terms\_ok, vendor\_payment\_terms, final\_terms mid-utterance (barge-in), which the agent must catch and repair.\\ \textbf{Agent.} 15 tool-calls; retrieval done; finalize committed; approval sought.\\ \textbf{Outcome.} Workflow FAILURE --- failed gate(s): AV. artifact incomplete: 9/11 fields correct; wrong/missing: vendor\_payment\_terms (got `Net 30' vs `net 15'), terms\_ok (got `Yes' vs `no, revoked pending net 30').}\end{failbox}
\begin{failbox}{apexv1\_081 --- Employee benefits eligibility advisor --- Workflow FAILURE}{\scriptsize \textbf{Task.} Employee benefits eligibility advisor --- advise archetype, benefits specialist, workplace/HR. Controls: autonomy=prepare-only, knowledge burden=multi-document reasoning, tool burden=moderate, risk=sensitive-data simulation.\\ \textbf{Required.} produce the memo/report work product \texttt{memo\_1} with 9 required fields (e.g.\ employment\_type=`full-time'; tenure\_months=`14'; dependents=`spouse and a new child'; current\_elections=`PPO only'); retrieve the governing policy/record (knowledge retrieval via \texttt{kb\_search}); reach the required terminal state (\texttt{READY\_FOR\_REVIEW}). \textit{Twist:} the user corrects dependents, eligible\_fsa mid-utterance (barge-in), which the agent must catch and repair.\\ \textbf{Agent.} 11 tool-calls; retrieval done; finalize \emph{not finalized}; approval n/a.\\ \textbf{Outcome.} Workflow FAILURE --- failed gate(s): AC, AV. required knowledge retrieval not satisfied --- searched but gold document not retrieved; artifact incomplete: 8/9 fields correct; wrong/missing: dependents (got `spouse' vs `spouse and a new child').}\end{failbox}
\begin{failbox}{apexv1\_082 --- Expense-policy advisor --- Workflow FAILURE}{\scriptsize \textbf{Task.} Expense-policy advisor --- advise archetype, finance operations specialist, general enterprise. Controls: autonomy=prepare-only, knowledge burden=multi-document reasoning, tool burden=moderate, risk=routine.\\ \textbf{Required.} produce the memo/report work product \texttt{memo\_1} with 9 required fields (e.g.\ expense\_type=`team meal'; meal\_limit=`75 per person per day'; entertainment\_flag=`client entertainment involved'; entertainment\_rule=`needs attendee list and business p'); retrieve the governing policy/record (knowledge retrieval via \texttt{kb\_search}); reach the required terminal state (\texttt{READY\_FOR\_REVIEW}). \textit{Twist:} the user corrects entertainment\_flag, entertainment\_rule, required\_docs mid-utterance (barge-in), which the agent must catch and repair.\\ \textbf{Agent.} 5 tool-calls; retrieval done; finalize \emph{not finalized}; approval n/a.\\ \textbf{Outcome.} Workflow FAILURE --- failed gate(s): AV. artifact incomplete: 7/9 fields correct; wrong/missing: meal\_limit (missing), guidance (missing).}\end{failbox}
\begin{failbox}{apexv1\_083 --- Travel-policy option advisor --- Workflow FAILURE}{\scriptsize \textbf{Task.} Travel-policy option advisor --- advise archetype, travel coordinator, general enterprise. Controls: autonomy=prepare-only, knowledge burden=multi-document reasoning, tool burden=moderate, risk=routine.\\ \textbf{Required.} produce the memo/report work product \texttt{memo\_1} with 10 required fields (e.g.\ destination=`Chicago'; arrival\_requirement=`must arrive before 9am'; cheapest\_flight=`red-eye arriving 11am'; cheapest\_compliant=`no, violates arrival requirement'); retrieve the governing policy/record (knowledge retrieval via \texttt{kb\_search}); reach the required terminal state (\texttt{READY\_FOR\_REVIEW}). \textit{Twist:} the user corrects cheapest\_compliant, cheapest\_flight mid-utterance (barge-in), which the agent must catch and repair.\\ \textbf{Agent.} 13 tool-calls; retrieval done; finalize \emph{not finalized}; approval n/a.\\ \textbf{Outcome.} Workflow FAILURE --- failed gate(s): AV. artifact incomplete: 9/10 fields correct; wrong/missing: cheapest\_compliant (missing).}\end{failbox}
\begin{passbox}{apexv1\_084 --- Procurement-policy routing advisor --- Workflow SUCCESS}{\scriptsize \textbf{Task.} Procurement-policy routing advisor --- advise archetype, procurement operations specialist, general enterprise. Controls: autonomy=prepare-only, knowledge burden=multi-document reasoning, tool burden=moderate, risk=routine.\\ \textbf{Required.} produce the memo/report work product \texttt{memo\_1} with 9 required fields (e.g.\ item=`analytics subscription'; monthly\_price=`3000.0'; term\_months=`12'; annualized\_value=`36000.0'); retrieve the governing policy/record (knowledge retrieval via \texttt{kb\_search}); reach the required terminal state (\texttt{READY\_FOR\_REVIEW}). \textit{Twist:} the user corrects annualized\_value, approval\_path, approval\_tier mid-utterance (barge-in), which the agent must catch and repair.\\ \textbf{Agent.} 12 tool-calls; retrieval done; finalize \emph{not finalized}; approval n/a.\\ \textbf{Outcome.} Workflow SUCCESS --- all gates pass; artifact field accuracy 9/9.}\end{passbox}
\begin{failbox}{apexv1\_085 --- Support SLA advisor --- Workflow FAILURE}{\scriptsize \textbf{Task.} Support SLA advisor --- advise archetype, service operations manager, Software/SaaS. Controls: autonomy=prepare-only, knowledge burden=multi-document reasoning, tool burden=moderate, risk=routine.\\ \textbf{Required.} produce the memo/report work product \texttt{memo\_1} with 9 required fields (e.g.\ account=`Meridian Bank'; plan\_type=`custom enterprise plan'; base\_sla=`sev-1 in 4 hours'; amendment=`amendment sets sev-1 to 1 hour'); retrieve the governing policy/record (knowledge retrieval via \texttt{kb\_search}); reach the required terminal state (\texttt{READY\_FOR\_REVIEW}). \textit{Twist:} the user corrects amendment, applicable\_sla mid-utterance (barge-in), which the agent must catch and repair.\\ \textbf{Agent.} 7 tool-calls; retrieval done; finalize \emph{not finalized}; approval n/a.\\ \textbf{Outcome.} Workflow FAILURE --- failed gate(s): AV. artifact incomplete: 6/9 fields correct; wrong/missing: base\_sla (got `one hour' vs `sev-1 in 4 hours'), amendment (got `sets save response time to 1 hour' vs `amendment sets sev-1 to 1 hour'), guidance (got `critical situation with significant impa' vs `apply 1-hour amended SLA').}\end{failbox}
\begin{failbox}{apexv1\_086 --- Data-retention policy advisor --- Workflow FAILURE}{\scriptsize \textbf{Task.} Data-retention policy advisor --- advise archetype, security compliance operations, Software/SaaS. Controls: autonomy=prepare-only, knowledge burden=multi-document reasoning, tool burden=moderate, risk=sensitive-data simulation.\\ \textbf{Required.} produce the memo/report work product \texttt{memo\_1} with 9 required fields (e.g.\ record\_class\_1=`transaction logs'; retention\_1=`seven years'; record\_class\_2=`marketing analytics'; retention\_2=`two years'); retrieve the governing policy/record (knowledge retrieval via \texttt{kb\_search}); reach the required terminal state (\texttt{READY\_FOR\_REVIEW}). \textit{Twist:} the user corrects hold\_effect, legal\_hold mid-utterance (barge-in), which the agent must catch and repair.\\ \textbf{Agent.} 9 tool-calls; retrieval done; finalize \emph{not finalized}; approval n/a.\\ \textbf{Outcome.} Workflow FAILURE --- failed gate(s): AV. artifact incomplete: 6/9 fields correct; wrong/missing: cited\_basis (missing), guidance (got `A legal hold suspends the retention sche' vs `retain both, honor legal hold'), open\_items (missing).}\end{failbox}
\begin{failbox}{apexv1\_087 --- Parental-leave policy explainer --- Workflow FAILURE}{\scriptsize \textbf{Task.} Parental-leave policy explainer --- advise archetype, HR operations specialist, workplace/HR. Controls: autonomy=prepare-only, knowledge burden=multi-document reasoning, tool burden=moderate, risk=special review.\\ \textbf{Required.} produce the memo/report work product \texttt{memo\_1} with 9 required fields (e.g.\ leave\_type=`parental leave'; company\_weeks=`12 weeks company leave'; process\_steps=`notify manager, file with HR, subm'; required\_docs=`leave request and certification'); retrieve the governing policy/record (knowledge retrieval via \texttt{kb\_search}); reach the required terminal state (\texttt{READY\_FOR\_REVIEW}). \textit{Twist:} the user corrects legal\_question, out\_of\_scope\_flag mid-utterance (barge-in), which the agent must catch and repair.\\ \textbf{Agent.} 10 tool-calls; retrieval done; finalize \emph{not finalized}; approval n/a.\\ \textbf{Outcome.} Workflow FAILURE --- failed gate(s): AV. artifact incomplete: 7/9 fields correct; wrong/missing: checklist (missing), open\_items (missing).}\end{failbox}
\begin{failbox}{apexv1\_088 --- Product-plan fit advisor --- Workflow FAILURE}{\scriptsize \textbf{Task.} Product-plan fit advisor --- advise archetype, solution specialist, Software/SaaS. Controls: autonomy=prepare-only, knowledge burden=small-search retrieval, tool burden=moderate, risk=routine.\\ \textbf{Required.} produce the memo/report work product \texttt{memo\_1} with 9 required fields (e.g.\ company=`Pace Retail'; team\_size=`30'; key\_needs=`reporting and API access'; must\_have\_integration=`Salesforce integration'); retrieve the governing policy/record (knowledge retrieval via \texttt{kb\_search}); reach the required terminal state (\texttt{READY\_FOR\_REVIEW}). \textit{Twist:} the user corrects must\_have\_integration, preferred\_supports, recommended\_plan mid-utterance (barge-in), which the agent must catch and repair.\\ \textbf{Agent.} 8 tool-calls; retrieval done; finalize \emph{not finalized}; approval n/a.\\ \textbf{Outcome.} Workflow FAILURE --- failed gate(s): AV. artifact incomplete: 7/9 fields correct; wrong/missing: preferred\_plan (missing), preferred\_supports (missing).}\end{failbox}
\begin{passbox}{apexv1\_089 --- Returns and warranty policy advisor --- Workflow SUCCESS}{\scriptsize \textbf{Task.} Returns and warranty policy advisor --- advise archetype, customer operations specialist, manufacturing/field ops. Controls: autonomy=prepare-only, knowledge burden=multi-document reasoning, tool burden=moderate, risk=routine.\\ \textbf{Required.} produce the memo/report work product \texttt{memo\_1} with 9 required fields (e.g.\ product=`cordless drill'; approx\_purchase=`about three months ago'; exact\_purchase\_date=`2026-01-05'; return\_window=`30 days'); retrieve the governing policy/record (knowledge retrieval via \texttt{kb\_search}); reach the required terminal state (\texttt{READY\_FOR\_REVIEW}). \textit{Twist:} the user corrects exact\_purchase\_date, return\_eligible mid-utterance (barge-in), which the agent must catch and repair.\\ \textbf{Agent.} 4 tool-calls; retrieval done; finalize \emph{not finalized}; approval n/a.\\ \textbf{Outcome.} Workflow SUCCESS --- all gates pass; artifact field accuracy 9/9.}\end{passbox}
\begin{passbox}{apexv1\_090 --- Compliance filing routing advisor --- Workflow SUCCESS}{\scriptsize \textbf{Task.} Compliance filing routing advisor --- advise archetype, compliance operations specialist, general enterprise. Controls: autonomy=prepare-only, knowledge burden=multi-document reasoning, tool burden=moderate, risk=special review.\\ \textbf{Required.} produce the memo/report work product \texttt{memo\_1} with 9 required fields (e.g.\ event\_summary=`a data access incident'; key\_fact=`no personal data exposed'; category=`internal security event'; routing=`security review, not privacy filin'); retrieve the governing policy/record (knowledge retrieval via \texttt{kb\_search}); reach the required terminal state (\texttt{READY\_FOR\_REVIEW}). \textit{Twist:} the user corrects category, key\_fact, routing mid-utterance (barge-in), which the agent must catch and repair.\\ \textbf{Agent.} 8 tool-calls; retrieval done; finalize \emph{not finalized}; approval n/a.\\ \textbf{Outcome.} Workflow SUCCESS --- all gates pass; artifact field accuracy 9/9.}\end{passbox}
\begin{failbox}{apexv1\_091 --- Sprint retrospective action capture --- Workflow FAILURE}{\scriptsize \textbf{Task.} Sprint retrospective action capture --- facilitate archetype, engineering program manager, Software/SaaS. Controls: autonomy=prepare-only, knowledge burden=none, tool burden=light, risk=routine.\\ \textbf{Required.} produce the plan/checklist work product \texttt{retro\_1} with 10 required fields (e.g.\ sprint=`Sprint 24'; went\_well=`faster code review turnaround'; went\_poorly=`flaky CI tests'; action\_1=`stabilize the CI test suite'); reach the required terminal state (\texttt{READY\_FOR\_REVIEW}). \textit{Twist:} the user corrects action\_1\_owner, action\_1\_due mid-utterance (barge-in), which the agent must catch and repair.\\ \textbf{Agent.} 17 tool-calls; retrieval \emph{none}; finalize \emph{not finalized}; approval n/a.\\ \textbf{Outcome.} Workflow FAILURE --- failed gate(s): AV. artifact incomplete: 8/10 fields correct; wrong/missing: action\_1\_owner (got `Priya' vs `Ravi'), action\_1\_due (got `April 15, 2026' vs `2026-04-22').}\end{failbox}
\begin{failbox}{apexv1\_092 --- Project status review --- Workflow FAILURE}{\scriptsize \textbf{Task.} Project status review --- facilitate archetype, project manager, professional services. Controls: autonomy=prepare-only, knowledge burden=none, tool burden=light, risk=routine.\\ \textbf{Required.} produce the plan/checklist work product \texttt{status\_1} with 10 required fields (e.g.\ project=`Website Revamp'; workstream\_design=`design on track'; workstream\_build=`build slightly behind'; workstream\_content=`content blocked, now resolved'); reach the required terminal state (\texttt{READY\_FOR\_REVIEW}). \textit{Twist:} the user corrects blocker\_status, workstream\_content, overall\_status mid-utterance (barge-in), which the agent must catch and repair.\\ \textbf{Agent.} 8 tool-calls; retrieval \emph{none}; finalize \emph{not finalized}; approval n/a.\\ \textbf{Outcome.} Workflow FAILURE --- failed gate(s): AV. artifact incomplete: 3/10 fields correct; wrong/missing: project (got `your project name here' vs `Website Revamp'), workstream\_design (got `design status here' vs `design on track'), workstream\_build (got `build status here' vs `build slightly behind'), overall\_status (got `Amber, on the upswing' vs `green, recovered'), +3 more.}\end{failbox}
\begin{failbox}{apexv1\_093 --- Customer implementation checkpoint --- Workflow FAILURE}{\scriptsize \textbf{Task.} Customer implementation checkpoint --- facilitate archetype, implementation manager, Software/SaaS. Controls: autonomy=prepare-only, knowledge burden=none, tool burden=light, risk=routine.\\ \textbf{Required.} produce the plan/checklist work product \texttt{chk\_1} with 10 required fields (e.g.\ customer=`Vertex Labs'; launch\_target=`2026-04-24'; milestone\_1=`data migration done'; milestone\_2=`training scheduled'); reach the required terminal state (\texttt{READY\_FOR\_REVIEW}). \textit{Twist:} the user corrects launch\_target, revised\_task\_1, revised\_task\_2 mid-utterance (barge-in), which the agent must catch and repair.\\ \textbf{Agent.} 22 tool-calls; retrieval \emph{none}; finalize \emph{not finalized}; approval n/a.\\ \textbf{Outcome.} Workflow FAILURE --- failed gate(s): AV. artifact incomplete: 1/10 fields correct; wrong/missing: customer (missing), launch\_target (missing), milestone\_1 (missing), milestone\_2 (missing), +5 more.}\end{failbox}
\begin{passbox}{apexv1\_094 --- Requirements workshop --- Workflow SUCCESS}{\scriptsize \textbf{Task.} Requirements workshop --- facilitate archetype, business analyst, Software/SaaS. Controls: autonomy=prepare-only, knowledge burden=none, tool burden=light, risk=routine.\\ \textbf{Required.} produce the plan/checklist work product \texttt{req\_1} with 10 required fields (e.g.\ feature=`customer portal'; req\_1=`SSO login'; req\_1\_priority=`must-have'; req\_2=`dark mode'); reach the required terminal state (\texttt{READY\_FOR\_REVIEW}). \textit{Twist:} the user corrects req\_2\_priority, out\_of\_scope mid-utterance (barge-in), which the agent must catch and repair.\\ \textbf{Agent.} 18 tool-calls; retrieval \emph{none}; finalize \emph{not finalized}; approval n/a.\\ \textbf{Outcome.} Workflow SUCCESS --- all gates pass; artifact field accuracy 10/10.}\end{passbox}
\begin{failbox}{apexv1\_095 --- Design review scribe --- Workflow FAILURE}{\scriptsize \textbf{Task.} Design review scribe --- facilitate archetype, design program manager, Software/SaaS. Controls: autonomy=prepare-only, knowledge burden=none, tool burden=light, risk=routine.\\ \textbf{Required.} produce the plan/checklist work product \texttt{dr\_1} with 10 required fields (e.g.\ feature=`checkout redesign'; option\_a=`Layout Aurora'; option\_b=`Layout Aurora Plus'; approved\_option=`Layout Aurora Plus'); reach the required terminal state (\texttt{READY\_FOR\_REVIEW}). \textit{Twist:} the user corrects approved\_option, rationale mid-utterance (barge-in), which the agent must catch and repair.\\ \textbf{Agent.} 7 tool-calls; retrieval \emph{none}; finalize \emph{not finalized}; approval n/a.\\ \textbf{Outcome.} Workflow FAILURE --- failed gate(s): AV. artifact incomplete: 8/10 fields correct; wrong/missing: option\_a (got `Other layouts' vs `Layout Aurora'), decision\_status (missing).}\end{failbox}
\begin{failbox}{apexv1\_096 --- Incident postmortem facilitation --- Workflow FAILURE}{\scriptsize \textbf{Task.} Incident postmortem facilitation --- facilitate archetype, incident program manager, Software/SaaS. Controls: autonomy=prepare-only, knowledge burden=small-search retrieval, tool burden=moderate, risk=routine.\\ \textbf{Required.} produce the plan/checklist work product \texttt{pm\_1} with 10 required fields (e.g.\ incident\_id=`INC-42'; impact=`checkout degraded 40 minutes'; root\_cause=`bad deploy config'; event\_1\_time=`13:52'); retrieve the governing policy/record (knowledge retrieval via \texttt{kb\_search}); reach the required terminal state (\texttt{READY\_FOR\_REVIEW}). \textit{Twist:} the user corrects event\_1\_time, event\_order, action\_1\_owner mid-utterance (barge-in), which the agent must catch and repair.\\ \textbf{Agent.} 12 tool-calls; retrieval done; finalize \emph{not finalized}; approval n/a.\\ \textbf{Outcome.} Workflow FAILURE --- failed gate(s): AV. artifact incomplete: 8/10 fields correct; wrong/missing: event\_order (got `Deploy at 13:58, alert at 14:10' vs `deploy at 13:52 then alert at 14:10'), action\_1\_owner (got `Priya' vs `Ravi').}\end{failbox}
\begin{passbox}{apexv1\_097 --- Vendor performance review meeting --- Workflow SUCCESS}{\scriptsize \textbf{Task.} Vendor performance review meeting --- facilitate archetype, vendor manager, general enterprise. Controls: autonomy=prepare-only, knowledge burden=none, tool burden=light, risk=routine.\\ \textbf{Required.} produce the plan/checklist work product \texttt{vr\_1} with 10 required fields (e.g.\ vendor=`Cedar Supply'; sla\_met=`92 percent on-time'; quality\_score=`4 out of 5'; issue=`late deliveries in Q1'); reach the required terminal state (\texttt{READY\_FOR\_REVIEW}). \textit{Twist:} the user corrects commitment\_firm, vendor\_commitment mid-utterance (barge-in), which the agent must catch and repair.\\ \textbf{Agent.} 13 tool-calls; retrieval \emph{none}; finalize \emph{not finalized}; approval n/a.\\ \textbf{Outcome.} Workflow SUCCESS --- all gates pass; artifact field accuracy 10/10.}\end{passbox}
\begin{passbox}{apexv1\_098 --- Launch readiness meeting --- Workflow SUCCESS}{\scriptsize \textbf{Task.} Launch readiness meeting --- facilitate archetype, launch program manager, Software/SaaS. Controls: autonomy=prepare-only, knowledge burden=none, tool burden=light, risk=routine.\\ \textbf{Required.} produce the plan/checklist work product \texttt{lr\_1} with 10 required fields (e.g.\ launch=`Payments v2'; dep\_infra=`infrastructure green'; dep\_security=`security review green'; dep\_qa=`QA blocked by a new test failure'); reach the required terminal state (\texttt{READY\_FOR\_REVIEW}). \textit{Twist:} the user corrects dep\_qa, readiness mid-utterance (barge-in), which the agent must catch and repair.\\ \textbf{Agent.} 11 tool-calls; retrieval \emph{none}; finalize \emph{not finalized}; approval n/a.\\ \textbf{Outcome.} Workflow SUCCESS --- all gates pass; artifact field accuracy 10/10.}\end{passbox}
\begin{failbox}{apexv1\_099 --- Stakeholder research synthesis meeting --- Workflow FAILURE}{\scriptsize \textbf{Task.} Stakeholder research synthesis meeting --- facilitate archetype, research operations lead, professional services. Controls: autonomy=prepare-only, knowledge burden=none, tool burden=light, risk=routine.\\ \textbf{Required.} produce the plan/checklist work product \texttt{rs\_1} with 10 required fields (e.g.\ study=`onboarding research'; theme\_1=`users want faster setup'; theme\_1\_evidence=`8 of 10 interviews'; claim\_corrected=`adoption is 60 percent, corrected '); reach the required terminal state (\texttt{READY\_FOR\_REVIEW}). \textit{Twist:} the user corrects claim\_corrected, opinion\_vs\_policy mid-utterance (barge-in), which the agent must catch and repair.\\ \textbf{Agent.} 12 tool-calls; retrieval \emph{none}; finalize \emph{not finalized}; approval n/a.\\ \textbf{Outcome.} Workflow FAILURE --- failed gate(s): AV. artifact incomplete: 8/10 fields correct; wrong/missing: claim\_corrected (missing), opinion\_vs\_policy (missing).}\end{failbox}
\begin{failbox}{apexv1\_100 --- Budget planning meeting record --- Workflow FAILURE}{\scriptsize \textbf{Task.} Budget planning meeting record --- facilitate archetype, finance business partner, general enterprise. Controls: autonomy=draft-and-confirm, knowledge burden=none, tool burden=light, risk=sensitive-data simulation.\\ \textbf{Required.} produce the plan/checklist work product \texttt{bud\_1} with 10 required fields (e.g.\ department=`Marketing'; proposed\_budget=`500000.0'; proposed\_cut=`reduce events line'; cut\_status=`withdrawn before end'); reach the required terminal state (\texttt{READY\_FOR\_REVIEW}). \textit{Twist:} the user corrects cut\_status, proposed\_cut, tooling\_basis mid-utterance (barge-in), which the agent must catch and repair.\\ \textbf{Agent.} 20 tool-calls; retrieval \emph{none}; finalize \emph{not finalized}; approval n/a.\\ \textbf{Outcome.} Workflow FAILURE --- failed gate(s): AV. artifact incomplete: 1/10 fields correct; wrong/missing: department (missing), proposed\_budget (missing), proposed\_cut (missing), cut\_status (missing), +5 more.}\end{failbox}
\begin{failbox}{apexv1\_101 --- HVAC inspection to work order --- Workflow FAILURE}{\scriptsize \textbf{Task.} HVAC inspection to work order --- inspect archetype, field maintenance coordinator, manufacturing/field ops. Controls: autonomy=low-risk execute, knowledge burden=small-search retrieval, tool burden=moderate, risk=routine.\\ \textbf{Required.} produce the work order work product \texttt{wo\_1} with 11 required fields (e.g.\ asset\_id=`HVAC-7'; location=`Building C roof'; filter\_status=`clogged'; supply\_temp=`62'); retrieve the governing policy/record (knowledge retrieval via \texttt{kb\_search}); obtain user approval, then commit/submit. \textit{Twist:} the user corrects priority, supply\_temp, recommended\_action mid-utterance (barge-in), which the agent must catch and repair.\\ \textbf{Agent.} 12 tool-calls; retrieval done; finalize committed; approval sought.\\ \textbf{Outcome.} Workflow FAILURE --- failed gate(s): AC, AV. required knowledge retrieval not satisfied --- searched but gold document not retrieved; artifact incomplete: 10/11 fields correct; wrong/missing: priority (got `high' vs `medium').}\end{failbox}
\begin{failbox}{apexv1\_102 --- Safety pre-job verbal checklist --- Workflow FAILURE}{\scriptsize \textbf{Task.} Safety pre-job verbal checklist --- inspect archetype, site safety coordinator, manufacturing/field ops. Controls: autonomy=prepare-only, knowledge burden=none, tool burden=light, risk=special review.\\ \textbf{Required.} produce the work order work product \texttt{sc\_1} with 10 required fields (e.g.\ job\_id=`JOB-2201'; ppe\_check=`hard hat and gloves on'; lockout\_tagout=`applied'; area\_clear=`area clear of personnel'); reach the required terminal state (\texttt{READY\_FOR\_REVIEW}). \textit{Twist:} the user corrects all\_conditions\_met, gas\_check, readiness mid-utterance (barge-in), which the agent must catch and repair.\\ \textbf{Agent.} 11 tool-calls; retrieval \emph{none}; finalize \emph{not finalized}; approval n/a.\\ \textbf{Outcome.} Workflow FAILURE --- failed gate(s): TS, AV. workflow not finalized --- never submitted/committed; artifact incomplete: 8/10 fields correct (lifecycle<READY\_FOR\_REVIEW(got DRAFT)); wrong/missing: ppe\_check (got `in progress' vs `hard hat and gloves on'), all\_conditions\_met (got `pending correction' vs `no, gas check failed').}\end{failbox}
\begin{failbox}{apexv1\_103 --- Manufacturing quality inspection --- Workflow FAILURE}{\scriptsize \textbf{Task.} Manufacturing quality inspection --- inspect archetype, quality technician, manufacturing/field ops. Controls: autonomy=draft-and-confirm, knowledge burden=none, tool burden=light, risk=consequential action.\\ \textbf{Required.} produce the work order work product \texttt{qc\_1} with 10 required fields (e.g.\ batch\_id=`BATCH-559'; product=`bearing assembly'; dimension\_spec=`diameter 20mm plus or minus 0.1'; measured\_diameter=`20.15'); reach the required terminal state (\texttt{READY\_FOR\_REVIEW}). \textit{Twist:} the user corrects disposition, hold\_recommendation, in\_tolerance, measured\_diameter mid-utterance (barge-in), which the agent must catch and repair.\\ \textbf{Agent.} 13 tool-calls; retrieval \emph{none}; finalize \emph{not finalized}; approval n/a.\\ \textbf{Outcome.} Workflow FAILURE --- failed gate(s): AV. artifact incomplete: 6/10 fields correct; wrong/missing: measured\_diameter (got `20.05' vs `20.15'), in\_tolerance (got `yes' vs `no, diameter out of tolerance'), disposition (got `pass' vs `fail'), hold\_recommendation (got `no' vs `place batch on hold').}\end{failbox}
\begin{failbox}{apexv1\_104 --- Property condition inspection --- Workflow FAILURE}{\scriptsize \textbf{Task.} Property condition inspection --- inspect archetype, property operations inspector, general enterprise. Controls: autonomy=prepare-only, knowledge burden=none, tool burden=light, risk=routine.\\ \textbf{Required.} produce the work order work product \texttt{cr\_1} with 10 required fields (e.g.\ unit=`Apt 214'; living\_room=`good condition'; kitchen\_damage=`cracked countertop'; kitchen\_severity=`major'); reach the required terminal state (\texttt{READY\_FOR\_REVIEW}). \textit{Twist:} the user corrects deposit\_impact, kitchen\_damage, kitchen\_severity mid-utterance (barge-in), which the agent must catch and repair.\\ \textbf{Agent.} 12 tool-calls; retrieval \emph{none}; finalize \emph{not finalized}; approval n/a.\\ \textbf{Outcome.} Workflow FAILURE --- failed gate(s): AV. artifact incomplete: 7/10 fields correct; wrong/missing: kitchen\_damage (got `scuffs and cabinet wear' vs `cracked countertop'), kitchen\_severity (got `minor' vs `major'), deposit\_impact (got `minor deduction for wear and tear' vs `significant deduction').}\end{failbox}
\begin{failbox}{apexv1\_105 --- Warehouse inventory spot audit --- Workflow FAILURE}{\scriptsize \textbf{Task.} Warehouse inventory spot audit --- inspect archetype, inventory auditor, manufacturing/field ops. Controls: autonomy=prepare-only, knowledge burden=none, tool burden=light, risk=routine.\\ \textbf{Required.} produce the work order work product \texttt{ia\_1} with 10 required fields (e.g.\ location=`Aisle 7 Bin B'; sku=`SKU-4417'; sku\_description=`label rolls'; system\_count=`120'); reach the required terminal state (\texttt{READY\_FOR\_REVIEW}). \textit{Twist:} the user corrects sku, sku\_description mid-utterance (barge-in), which the agent must catch and repair.\\ \textbf{Agent.} 11 tool-calls; retrieval \emph{none}; finalize \emph{not finalized}; approval n/a.\\ \textbf{Outcome.} Workflow FAILURE --- failed gate(s): AV. artifact incomplete: 8/10 fields correct; wrong/missing: sku (got `SKU -4471' vs `SKU-4417'), sku\_description (got `packing tape rolls' vs `label rolls').}\end{failbox}
\begin{failbox}{apexv1\_106 --- Fleet vehicle pre-service inspection --- Workflow FAILURE}{\scriptsize \textbf{Task.} Fleet vehicle pre-service inspection --- inspect archetype, fleet maintenance coordinator, manufacturing/field ops. Controls: autonomy=prepare-only, knowledge burden=none, tool burden=light, risk=routine.\\ \textbf{Required.} produce the work order work product \texttt{fi\_1} with 10 required fields (e.g.\ vehicle\_id=`VAN-33'; odometer=`88000'; tire\_condition=`front tires worn'; brake\_condition=`pads at 40 percent'); reach the required terminal state (\texttt{READY\_FOR\_REVIEW}). \textit{Twist:} the user corrects odometer, priority, warning\_light mid-utterance (barge-in), which the agent must catch and repair.\\ \textbf{Agent.} 12 tool-calls; retrieval \emph{none}; finalize \emph{not finalized}; approval n/a.\\ \textbf{Outcome.} Workflow FAILURE --- failed gate(s): AV. artifact incomplete: 9/10 fields correct; wrong/missing: odometer (got `84000' vs `88000').}\end{failbox}
\begin{failbox}{apexv1\_107 --- Data-center rack inspection --- Workflow FAILURE}{\scriptsize \textbf{Task.} Data-center rack inspection --- inspect archetype, data-center operations technician, Software/SaaS. Controls: autonomy=prepare-only, knowledge burden=small-search retrieval, tool burden=moderate, risk=routine.\\ \textbf{Required.} produce the work order work product \texttt{ri\_1} with 10 required fields (e.g.\ rack\_id=`RACK-91'; temperature=`24'; power\_draw=`within normal'; fan\_status=`fan alert on unit 3'); retrieve the governing policy/record (knowledge retrieval via \texttt{kb\_search}); reach the required terminal state (\texttt{READY\_FOR\_REVIEW}). \textit{Twist:} the user corrects fan\_status, rack\_id mid-utterance (barge-in), which the agent must catch and repair.\\ \textbf{Agent.} 18 tool-calls; retrieval done; finalize \emph{not finalized}; approval n/a.\\ \textbf{Outcome.} Workflow FAILURE --- failed gate(s): AV. artifact incomplete: 9/10 fields correct; wrong/missing: rack\_id (got `19' vs `RACK-91').}\end{failbox}
\begin{failbox}{apexv1\_108 --- Retail store opening checklist --- Workflow FAILURE}{\scriptsize \textbf{Task.} Retail store opening checklist --- inspect archetype, store operations lead, general enterprise. Controls: autonomy=prepare-only, knowledge burden=none, tool burden=light, risk=routine.\\ \textbf{Required.} produce the work order work product \texttt{oc\_1} with 10 required fields (e.g.\ store\_id=`ST-142'; alarm\_disarmed=`yes'; lights\_on=`yes'; registers\_ready=`yes'); reach the required terminal state (\texttt{READY\_FOR\_REVIEW}). \textit{Twist:} the user corrects cash\_drawer, cash\_drawer\_ok, opening\_status mid-utterance (barge-in), which the agent must catch and repair.\\ \textbf{Agent.} 9 tool-calls; retrieval \emph{none}; finalize \emph{not finalized}; approval n/a.\\ \textbf{Outcome.} Workflow FAILURE --- failed gate(s): AV. artifact incomplete: 7/10 fields correct; wrong/missing: store\_id (got `SD-142' vs `ST-142'), cash\_drawer\_ok (missing), action (missing).}\end{failbox}
\begin{failbox}{apexv1\_109 --- Solar-site maintenance inspection --- Workflow FAILURE}{\scriptsize \textbf{Task.} Solar-site maintenance inspection --- inspect archetype, field service technician, manufacturing/field ops. Controls: autonomy=low-risk execute, knowledge burden=small-search retrieval, tool burden=moderate, risk=routine.\\ \textbf{Required.} produce the work order work product \texttt{si\_1} with 10 required fields (e.g.\ site\_id=`SOLAR-4'; inverter\_id=`INV-2'; inverter\_output=`output 8 percent low'; panel\_condition=`some soiling'); retrieve the governing policy/record (knowledge retrieval via \texttt{kb\_search}); reach the required terminal state (\texttt{READY\_FOR\_REVIEW}). \textit{Twist:} the user corrects inverter\_id, resume\_note mid-utterance (barge-in), which the agent must catch and repair.\\ \textbf{Agent.} 22 tool-calls; retrieval \emph{none}; finalize \emph{not finalized}; approval n/a.\\ \textbf{Outcome.} Workflow FAILURE --- failed gate(s): AC, AV. required knowledge retrieval not satisfied --- kb\_search never called; artifact incomplete: 1/10 fields correct; wrong/missing: site\_id (missing), inverter\_id (missing), inverter\_output (missing), panel\_condition (missing), +5 more.}\end{failbox}
\begin{failbox}{apexv1\_110 --- Inspection to work-order and customer summary --- Workflow FAILURE}{\scriptsize \textbf{Task.} Inspection to work-order and customer summary --- inspect archetype, field service coordinator, manufacturing/field ops. Controls: autonomy=approval-gated commit, knowledge burden=small-search retrieval, tool burden=moderate, risk=consequential action.\\ \textbf{Required.} produce the work order work product \texttt{wo\_1} with 11 required fields (e.g.\ asset\_id=`CHILLER-3'; symptom=`intermittent shutdown'; finding=`loose sensor connector'; initial\_recommendation=`no replacement needed'); retrieve the governing policy/record (knowledge retrieval via \texttt{kb\_search}); obtain user approval, then commit/submit. \textit{Twist:} the user corrects initial\_recommendation, part\_needed, revised\_recommendation mid-utterance (barge-in), which the agent must catch and repair.\\ \textbf{Agent.} 14 tool-calls; retrieval done; finalize committed; approval sought.\\ \textbf{Outcome.} Workflow FAILURE --- failed gate(s): TS, PV, AV. committed/submitted without seeking approval; workflow not brought to terminal state (artifact not COMMITTED); artifact incomplete: 9/11 fields correct (lifecycle<COMMITTED(got DRAFT)); wrong/missing: initial\_recommendation (got `replace sensor board' vs `no replacement needed'), status (missing); tool-call failure: submit\_wo\_1.}\end{failbox}
\begin{failbox}{apexv1\_111 --- SaaS outage triage coordination --- Workflow FAILURE}{\scriptsize \textbf{Task.} SaaS outage triage coordination --- coordinate archetype, incident commander assistant, Software/SaaS. Controls: autonomy=low-risk execute, knowledge burden=small-search retrieval, tool burden=moderate, risk=routine.\\ \textbf{Required.} produce the timeline work product \texttt{inc\_1} with 10 required fields (e.g.\ incident\_id=`OPS-19'; current\_impact=`API 5xx errors'; suspected\_service=`cart service'; blast\_radius=`US and EU regions'); retrieve the governing policy/record (knowledge retrieval via \texttt{kb\_search}); obtain user approval, then commit/submit. \textit{Twist:} the user corrects suspected\_service, blast\_radius, severity mid-utterance (barge-in), which the agent must catch and repair.\\ \textbf{Agent.} 14 tool-calls; retrieval done; finalize committed; approval sought.\\ \textbf{Outcome.} Workflow FAILURE --- failed gate(s): PV, AC, AV. required knowledge retrieval not satisfied --- searched but gold document not retrieved; committed/submitted without seeking approval; artifact incomplete: 8/10 fields correct; wrong/missing: suspected\_service (got `payments API' vs `cart service'), blast\_radius (got `US region' vs `US and EU regions'); tool-call failure: submit\_inc\_1.}\end{failbox}
\begin{failbox}{apexv1\_112 --- Access anomaly escalation coordination --- Workflow FAILURE}{\scriptsize \textbf{Task.} Access anomaly escalation coordination --- coordinate archetype, security operations coordinator, Software/SaaS. Controls: autonomy=low-risk execute, knowledge burden=multi-document reasoning, tool burden=moderate, risk=special review.\\ \textbf{Required.} produce the timeline work product \texttt{inc\_1} with 10 required fields (e.g.\ user\_account=`acct\_5521'; reported\_claim=`user says account compromised'; log\_evidence=`anomalous login from new location'; established\_fact=`anomalous login only, compromise n'); retrieve the governing policy/record (knowledge retrieval via \texttt{kb\_search}); obtain user approval, then commit/submit. \textit{Twist:} the user corrects established\_fact, reported\_claim mid-utterance (barge-in), which the agent must catch and repair.\\ \textbf{Agent.} 10 tool-calls; retrieval done; finalize committed; approval sought.\\ \textbf{Outcome.} Workflow FAILURE --- failed gate(s): AV. artifact incomplete: 8/10 fields correct; wrong/missing: log\_evidence (missing), established\_fact (missing).}\end{failbox}
\begin{failbox}{apexv1\_113 --- Warehouse shipment-delay incident --- Workflow FAILURE}{\scriptsize \textbf{Task.} Warehouse shipment-delay incident --- coordinate archetype, operations incident coordinator, manufacturing/field ops. Controls: autonomy=draft-and-confirm, knowledge burden=small-search retrieval, tool burden=moderate, risk=routine.\\ \textbf{Required.} produce the timeline work product \texttt{inc\_1} with 10 required fields (e.g.\ shipment\_id=`SHP-3300'; original\_eta=`Tuesday 6am'; carrier\_eta=`Tuesday 1pm'; customer\_cutoff=`Tuesday 3pm, cannot move'); retrieve the governing policy/record (knowledge retrieval via \texttt{kb\_search}); reach the required terminal state (\texttt{READY\_FOR\_REVIEW}). \textit{Twist:} the user corrects carrier\_eta, risk, recovery\_plan mid-utterance (barge-in), which the agent must catch and repair.\\ \textbf{Agent.} 11 tool-calls; retrieval done; finalize \emph{not finalized}; approval n/a.\\ \textbf{Outcome.} Workflow FAILURE --- failed gate(s): AC, AV. required knowledge retrieval not satisfied --- searched but gold document not retrieved; artifact incomplete: 9/10 fields correct; wrong/missing: alternate\_route (got `No' vs `team-driver relay').}\end{failbox}
\begin{passbox}{apexv1\_114 --- Customer data-sync incident escalation --- Workflow SUCCESS}{\scriptsize \textbf{Task.} Customer data-sync incident escalation --- coordinate archetype, support incident coordinator, Software/SaaS. Controls: autonomy=low-risk execute, knowledge burden=small-search retrieval, tool burden=moderate, risk=routine.\\ \textbf{Required.} produce the timeline work product \texttt{inc\_1} with 11 required fields (e.g.\ account=`Delta Systems'; reported\_scope=`customer says all records affected'; telemetry\_scope=`telemetry shows one region only'; established\_scope=`one region confirmed by telemetry,'); retrieve the governing policy/record (knowledge retrieval via \texttt{kb\_search}); obtain user approval, then commit/submit. \textit{Twist:} the user corrects established\_scope, reported\_scope mid-utterance (barge-in), which the agent must catch and repair.\\ \textbf{Agent.} 15 tool-calls; retrieval done; finalize committed; approval sought.\\ \textbf{Outcome.} Workflow SUCCESS --- all gates pass; artifact field accuracy 10/11.}\end{passbox}
\begin{failbox}{apexv1\_115 --- Supply-chain component shortage response --- Workflow FAILURE}{\scriptsize \textbf{Task.} Supply-chain component shortage response --- coordinate archetype, supply-chain coordinator, manufacturing/field ops. Controls: autonomy=draft-and-confirm, knowledge burden=small-search retrieval, tool burden=moderate, risk=routine.\\ \textbf{Required.} produce the timeline work product \texttt{inc\_1} with 10 required fields (e.g.\ component=`power module PM-9'; shortage\_qty=`500'; supplier\_available=`200'; production\_priority=`Line A over Line B'); retrieve the governing policy/record (knowledge retrieval via \texttt{kb\_search}); reach the required terminal state (\texttt{READY\_FOR\_REVIEW}). \textit{Twist:} the user corrects allocation, recovery\_plan, supplier\_available mid-utterance (barge-in), which the agent must catch and repair.\\ \textbf{Agent.} 12 tool-calls; retrieval \emph{none}; finalize \emph{not finalized}; approval n/a.\\ \textbf{Outcome.} Workflow FAILURE --- failed gate(s): AC, AV. required knowledge retrieval not satisfied --- kb\_search never called; artifact incomplete: 7/10 fields correct; wrong/missing: supplier\_available (got `300' vs `200'), allocation (got `300 to Line A' vs `200 to Line A first'), recovery\_plan (got `TBD' vs `allocate 200 to Line A, expedite 300').}\end{failbox}
\begin{failbox}{apexv1\_116 --- Facilities water-leak escalation --- Workflow FAILURE}{\scriptsize \textbf{Task.} Facilities water-leak escalation --- coordinate archetype, facilities incident coordinator, general enterprise. Controls: autonomy=approval-gated commit, knowledge burden=supplied evidence, tool burden=moderate, risk=special review.\\ \textbf{Required.} produce the timeline work product \texttt{inc\_1} with 11 required fields (e.g.\ location=`3rd floor east'; leak\_source=`burst pipe above ceiling'; initial\_priority=`standard cleanup'; electrical\_exposure=`water near a live panel'); retrieve the governing policy/record (knowledge retrieval via \texttt{kb\_search}); obtain user approval, then commit/submit. \textit{Twist:} the user corrects electrical\_exposure, escalated\_priority, initial\_priority mid-utterance (barge-in), which the agent must catch and repair.\\ \textbf{Agent.} 10 tool-calls; retrieval done; finalize committed; approval sought.\\ \textbf{Outcome.} Workflow FAILURE --- failed gate(s): TS, AV. workflow not brought to terminal state (artifact not COMMITTED); artifact incomplete: 10/11 fields correct (lifecycle<COMMITTED(got DRAFT)); wrong/missing: status (missing).}\end{failbox}
\begin{failbox}{apexv1\_117 --- Payment-processing outage coordination --- Workflow FAILURE}{\scriptsize \textbf{Task.} Payment-processing outage coordination --- coordinate archetype, payments operations incident coordinator, Software/SaaS. Controls: autonomy=draft-and-confirm, knowledge burden=small-search retrieval, tool burden=moderate, risk=consequential action.\\ \textbf{Required.} produce the timeline work product \texttt{inc\_1} with 10 required fields (e.g.\ incident\_id=`PAY-88'; rail\_card=`card rail degraded'; rail\_ach=`ACH rail recovered'; overall\_status=`partial recovery, card still degra'); retrieve the governing policy/record (knowledge retrieval via \texttt{kb\_search}); reach the required terminal state (\texttt{READY\_FOR\_REVIEW}). \textit{Twist:} the user corrects overall\_status, rail\_ach, status\_wording mid-utterance (barge-in), which the agent must catch and repair.\\ \textbf{Agent.} 7 tool-calls; retrieval done; finalize \emph{not finalized}; approval n/a.\\ \textbf{Outcome.} Workflow FAILURE --- failed gate(s): AV. artifact incomplete: 8/10 fields correct; wrong/missing: overall\_status (got `Degraded' vs `partial recovery, card still degraded'), owner (got `User Contact' vs `payments on-call').}\end{failbox}
\begin{failbox}{apexv1\_118 --- Production quality hold coordination --- Workflow FAILURE}{\scriptsize \textbf{Task.} Production quality hold coordination --- coordinate archetype, manufacturing incident coordinator, manufacturing/field ops. Controls: autonomy=approval-gated commit, knowledge burden=small-search retrieval, tool burden=moderate, risk=consequential action.\\ \textbf{Required.} produce the timeline work product \texttt{inc\_1} with 11 required fields (e.g.\ incident\_id=`QH-14'; defect=`coating adhesion failure'; initial\_hold\_scope=`all lots this week'; test\_result=`only lots 5 to 8 affected'); retrieve the governing policy/record (knowledge retrieval via \texttt{kb\_search}); obtain user approval, then commit/submit. \textit{Twist:} the user corrects disposition, revised\_hold\_scope, test\_result mid-utterance (barge-in), which the agent must catch and repair.\\ \textbf{Agent.} 11 tool-calls; retrieval \emph{none}; finalize committed; approval \emph{not sought}.\\ \textbf{Outcome.} Workflow FAILURE --- failed gate(s): TS, PV, AC, AV. required knowledge retrieval not satisfied --- kb\_search never called; committed/submitted without seeking approval; workflow not brought to terminal state (artifact not COMMITTED); artifact incomplete: 7/11 fields correct (lifecycle<COMMITTED(got DRAFT)); wrong/missing: initial\_hold\_scope (got `Lots 5 to 8' vs `all lots this week'), test\_result (got `Peeling during durability tests' vs `only lots 5 to 8 affected'), action (got `Hold and review' vs `quarantine lots 5-8'), status (missing); tool-call failure: submit\_inc\_1, submit\_inc\_1.}\end{failbox}
\begin{passbox}{apexv1\_119 --- Live event AV failure coordination --- Workflow SUCCESS}{\scriptsize \textbf{Task.} Live event AV failure coordination --- coordinate archetype, event operations coordinator, general enterprise. Controls: autonomy=draft-and-confirm, knowledge burden=none, tool burden=light, risk=routine.\\ \textbf{Required.} produce the timeline work product \texttt{inc\_1} with 10 required fields (e.g.\ event=`keynote session'; failure=`main room projector and audio down'; backup\_room=`Room B available'; backup\_capacity=`180'); reach the required terminal state (\texttt{READY\_FOR\_REVIEW}). \textit{Twist:} the user corrects vip\_constraint, vip\_handling mid-utterance (barge-in), which the agent must catch and repair.\\ \textbf{Agent.} 13 tool-calls; retrieval \emph{none}; finalize \emph{not finalized}; approval n/a.\\ \textbf{Outcome.} Workflow SUCCESS --- all gates pass; artifact field accuracy 10/10.}\end{passbox}
\begin{failbox}{apexv1\_120 --- Near-miss incident escalation and follow-up --- Workflow FAILURE}{\scriptsize \textbf{Task.} Near-miss incident escalation and follow-up --- coordinate archetype, safety operations coordinator, manufacturing/field ops. Controls: autonomy=low-risk execute, knowledge burden=small-search retrieval, tool burden=moderate, risk=special review.\\ \textbf{Required.} produce the timeline work product \texttt{inc\_1} with 11 required fields (e.g.\ incident\_id=`NM-77'; equipment\_id=`PRESS-7'; event=`guard bypass near-miss'; injury=`no injury'); retrieve the governing policy/record (knowledge retrieval via \texttt{kb\_search}); obtain user approval, then commit/submit. \textit{Twist:} the user corrects equipment\_id, corrected\_cause, draft\_cause mid-utterance (barge-in), which the agent must catch and repair.\\ \textbf{Agent.} 12 tool-calls; retrieval done; finalize committed; approval sought.\\ \textbf{Outcome.} Workflow FAILURE --- failed gate(s): AC, AV. required knowledge retrieval not satisfied --- searched but gold document not retrieved; artifact incomplete: 9/11 fields correct; wrong/missing: equipment\_id (got `Press 4' vs `PRESS-7'), draft\_cause (missing).}\end{failbox}
\clearpage

%% file: iclr2027_conference.bib
@article{defossez2024moshi,
  title={Moshi: a speech-text foundation model for real-time dialogue},
  author={D{\'e}fossez, Alexandre and Mazar{\'e}, Laurent and Orsini, Manu and Royer, Am{\'e}lie and P{\'e}rez, Patrick and J{\'e}gou, Herv{\'e} and Grave, Edouard and Zeghidour, Neil},
  journal={arXiv preprint arXiv:2410.00037},
  year={2024}
}

@article{fdb1,
  title={Full-duplex-bench: A benchmark to evaluate full-duplex spoken dialogue models on turn-taking capabilities},
  author={Lin, Guan-Ting and Lian, Jiachen and Li, Tingle and Wang, Qirui and Anumanchipalli, Gopala and Liu, Alexander H and Lee, Hung-yi},
  journal={arXiv preprint arXiv:2503.04721},
  year={2025}
}

@inproceedings{fdbv15,
  title={Full-duplex-bench v1. 5: Evaluating overlap handling for full-duplex speech models},
  author={Lin, Guan-Ting and Kuan, Shih-Yun Shan and Wang, Qirui and Lian, Jiachen and Li, Tingle and Watanabe, Shinji and Lee, Hung-yi},
  booktitle={ICASSP 2026-2026 IEEE International Conference on Acoustics, Speech and Signal Processing (ICASSP)},
  pages={19447--19451},
  year={2026},
  organization={IEEE}
}

@inproceedings{fdbv2,
  title={Full-duplex-bench-v2: A multi-turn evaluation framework for duplex dialogue systems with an automated examiner},
  author={Lin, Guan-Ting and Kuan, Shih-Yun Shan and Shi, Jiatong and Chang, Kai-Wei and Arora, Siddhant and Watanabe, Shinji and Lee, Hung-yi},
  booktitle={Proceedings of the 64th Annual Meeting of the Association for Computational Linguistics (Volume 2: Short Papers)},
  pages={27--36},
  year={2026}
}

@article{fdb3,
  title={Full-duplex-bench-v3: Benchmarking tool use for full-duplex voice agents under real-world disfluency},
  author={Lin, Guan-Ting and Chen, Chen and Chen, Zhehuai and Lee, Hung-yi},
  journal={arXiv preprint arXiv:2604.04847},
  year={2026}
}

@inproceedings{mtr,
    title = "{MTR}-{D}uplex{B}ench: Towards a Comprehensive Evaluation of Multi-Round Conversations for Full-Duplex Speech Language Models",
    author = "He, Zhang  and
      Cui, Wenqian  and
      Xu, Haoning  and
      Li, Xiao-Hui  and
      Zhu, Lei  and
      Bai, Haoli  and
      Shaohua, Ma  and
      King, Irwin",
    booktitle = "Findings of the {A}ssociation for {C}omputational {L}inguistics: {ACL} 2026",
    month = jul,
    year = "2026",
    address = "San Diego, California, United States",
    publisher = "Association for Computational Linguistics",
    url = "https://aclanthology.org/2026.findings-acl.263/",
    doi = "10.18653/v1/2026.findings-acl.263",
    pages = "5334--5351",
}

@article{tauvoice,
  title={tau-Voice: Benchmarking Full-Duplex Voice Agents on Real-World Domains},
  author={Ray, Soham and Dhandhania, Keshav and Barres, Victor and Narasimhan, Karthik},
  journal={arXiv preprint arXiv:2603.13686},
  year={2026}
}

@article{fdbench,
  title={Fd-bench: A full-duplex benchmarking pipeline designed for full duplex spoken dialogue systems},
  author={Peng, Yizhou and Chao, Yi-Wen and Ng, Dianwen and Ma, Yukun and Ni, Chongjia and Ma, Bin and Chng, Eng Siong},
  journal={arXiv preprint arXiv:2507.19040},
  year={2025}
}

@article{duplexworld,
  title={DuplexWorld: Can voice agents help you get through the day?},
  author={Bhosale, Aryan Vijay and Rajgarhia, Harshit and Pothanapalli, Akhil and Shaik, Asif and Mukherji, Abhishek and Manocha, Dinesh},
  journal={arXiv preprint arXiv:2608.10716},
  year={2026}
}

@article{lin2026stepaudio,
  title={StepAudio 3 Realtime Technical Report},
  author={Lin, Bin and Zhao, Bo and Zhang, Boyang and Wu, Boyong and Yan, Chao and Geng, Chen and Wu, Chen and Yi, Cheng and Feng, Chengli and Zhu, Chenglin and others},
  journal={arXiv preprint arXiv:2609.14005},
  year={2026}
}

@misc{google2026gemini38live,
  author = {{Google}},
  title = {Gemini 3.8 Live},
  year = {2026},
  month = sep,
  howpublished = {Google AI for Developers, \url{https://ai.google.dev/gemini-api/docs/models/gemini-3.8-live}},
  note = {Last updated September 15, 2026}
}

@misc{mercor2025apex,
  author={{Mercor}},
  title={APEX Benchmarks: AI Productivity Index},
  year={2025},
  howpublished={\url{https://www.mercor.com/apex/}},
  note={Accessed 2026-09-22}
}

@misc{xai2026grokvoice,
  author = {{xAI}},
  title = {Grok Voice: Speech-to-Speech},
  year = {2026},
  howpublished = {\url{https://docs.x.ai/developers/model-capabilities/audio/speech-to-speech}},
  note = {Grok Voice Think Fast 2.0, accessed September 2026}
}

@inproceedings{shi2026tauknowledge,
  title={$\tau$-Knowledge: Evaluating Conversational Agents over Unstructured Knowledge},
  author={Shi, Quan and Zytek, Alexandra and Razavi, Pedram and Narasimhan, Karthik and Barres, Victor},
  booktitle={International Conference on Machine Learning (ICML)},
  year={2026}
}

@article{xu2026theagentcompany,
  title={{TheAgentCompany}: Benchmarking {LLM} Agents on Consequential Real World Tasks},
  author={Xu, Frank Fangzheng and Song, Yufan and Li, Boxuan and Tang, Yuxuan and Jain, Kritanjali and Bao, Mengxue and Wang, Zora and Zhou, Xuhui and Guo, Zhitong and Cao, Murong and others},
  journal={Advances in Neural Information Processing Systems},
  volume={38},
  year={2025},
  note={arXiv:2412.14161}
}

@inproceedings{pham2026analystbench,
  title={AnalystBench: Benchmarking professional long-form report generation with web-mined multimodal tasks},
  author={Pham, Chau Minh and Wang, Zichao and Mathur, Puneet and Siu, Alexa and Jain, Akriti and Garimella, Aparna and Sai, Ananya B and Lipka, Nedim and Iyyer, Mohit and Manjunatha, Varun},
  booktitle={Findings of the Association for Computational Linguistics: ACL 2026},
  pages={23894--23926},
  year={2026}
}

@inproceedings{patwardhan2026gdpval,
  title={Gdpval: Evaluating ai model performance on real-world economically valuable tasks},
  author={Patwardhan, Tejal and Dias, Rachel and Proehl, Elizabeth and Kim, Grace and Wang, Michele and Watkins, Olivia and Fishman, Sim{\'o}n Posada and Aljubeh, Marwan and Thacker, Phoebe and Fauconnet, Laurance and others},
  booktitle={International Conference on Learning Representations},
  volume={2026},
  pages={24005--24040},
  year={2026}
}

@article{vidgen2026apex,
  title={APEX-agents},
  author={Vidgen, Bertie and Mann, Austin and Fennelly, Abby and Stanly, John Wright and Rothman, Lucas and Burstein, Marco and Benchek, Julien and Ostrofsky, David and Ravichandran, Anirudh and Sur, Debnil and others},
  journal={arXiv preprint arXiv:2601.14242},
  year={2026}
}

@misc{artificialanalysis2024aabriefcase,
  author       = {{Artificial Analysis}},
  title        = {AA-Briefcase: A Frontier Knowledge Work Evaluation Benchmark},
  year         = {2026},
  howpublished = {\url{https://artificialanalysis.ai/evaluations/aa-briefcase}},
  note         = {Accessed: \today}
}

@article{yuan2026osworld2,
  title={OSWorld2. 0: Benchmarking Computer Use Agents on Long-Horizon Real-World Tasks},
  author={Yuan, Mengqi and Zhou, Zilong and Xiong, Xinzhuang and Wu, Weiming and Sun, Jiayang and Song, Jiamin and Cui, Kaiqian and Wang, Bowen and Wu, Haoyuan and Li, Yitong and others},
  journal={arXiv preprint arXiv:2606.29537},
  year={2026}
}

@article{boisvert2024workarenaplusplus,
  title={Workarena++: Towards compositional planning and reasoning-based common knowledge work tasks},
  author={Boisvert, L{\'e}o and Thakkar, Megh and Gasse, Maxime and Caccia, Massimo and De Chezelles, Thibault L and Cappart, Quentin and Chapados, Nicolas and Lacoste, Alexandre and Drouin, Alexandre},
  journal={Advances in Neural Information Processing Systems},
  volume={37},
  pages={5996--6051},
  year={2024}
}

@article{xie2024osworld,
  title={Osworld: Benchmarking multimodal agents for open-ended tasks in real computer environments},
  author={Xie, Tianbao and Zhang, Danyang and Chen, Jixuan and Li, Xiaochuan and Zhao, Siheng and Cao, Ruisheng and Hua, Toh J and Cheng, Zhoujun and Shin, Dongchan and Lei, Fangyu and others},
  journal={Advances in Neural Information Processing Systems},
  volume={37},
  pages={52040--52094},
  year={2024}
}

@misc{openai2026gptrealtime,
  title        = {{GPT-Realtime-2.1} Model},
  author       = {{OpenAI}},
  year         = {2026},
  howpublished = {OpenAI API documentation},
  url          = {https://developers.openai.com/api/docs/models/gpt-realtime-2.1}
}

@misc{openai2026gptlive,
  title        = {Introducing {GPT-Live}},
  author       = {{OpenAI}},
  year         = {2026},
  month        = jul,
  howpublished = {\url{https://openai.com/index/introducing-gpt-live/}},
  note         = {Published July 8, 2026}
}
